%% file: main.tex
\documentclass[11pt,a4page,oneside]{book}

\input{header_packages}

\input{header_macros}
\input{header_letters}

\usepackage{tabularray}

\usepackage{scalerel,stackengine,amsmath}

\newboolean{final}
\setboolean{final}{true}

\usepackage{fullpage}

\graphicspath{{./Figures/}}

\definecolor{verylightgray}{gray}{0.95}

\begin{document}
\title{\Huge\sc Mathematical theory of deep learning\vspace{6ex}}

\author[1]{Philipp Petersen}
\author[2]{Jakob Zech}

\affil[1]{\normalsize                                                                                Universit\"at Wien, Fakult\"at für Mathematik\\
  1090 Wien, Austria\\
  \texttt{philipp.petersen@univie.ac.at }
}     

\affil[2]{\normalsize                                                                                Universit\"at Heidelberg, Interdisziplin\"ares Zentrum f\"ur Wissenschaftliches Rechnen\\
  69120 Heidelberg, Germany\\
  \texttt{jakob.zech@uni-heidelberg.de}
}     

\date{\today}

\maketitle

\setcounter{tocdepth}{1}
\tableofcontents

\numberwithin{equation}{chapter}
\numberwithin{equation}{section}

\include{preface}

\include{Notation}

\include{Introduction}

\include{FeedForwardNNs}

\include{UniversalApproximation}

\include{Splines}

\include{ReLUNNs}

\include{NumberOfPieces}

\include{DeepReLUNNs}

\include{HighDimApprox}

\include{Interpolation}

\include{Optimization}

\include{WideNetworks}

\include{LossLandscapeAnalysis}

\include{ShapeOfNNSpaces}
\include{GeneralizationOfNNs}

\include{GenInOverparameterization}

\include{RobustnessAndAdversarialEx}

\include{Architectures}

\include{Appendix}

\bibliographystyle{abbrv} \bibliography{main}
\end{document}

%% file: header_packages.tex
\usepackage[english]{babel}

\usepackage[colorinlistoftodos]{todonotes}

\usepackage[pdfpagelabels]{hyperref}
\hypersetup{colorlinks = true,
  linktoc = all, linkcolor = black, citecolor = black}

\usepackage[normalem]{ulem}
\usepackage[toc,page]{appendix}
\usepackage{dsfont} %
\usepackage{bm}
\usepackage{authblk}

\usepackage{amsmath}
\usepackage{amsthm}
\usepackage{amsfonts}
\usepackage{upgreek}
\usepackage{mathtools} %

\usepackage{algorithm}
\usepackage{algorithmic}

\usepackage{pgfplots}
\usepackage{xcolor}
\usepackage{graphicx}
\usepackage{amssymb}
\usepackage{tikz}
\usetikzlibrary{patterns}
\usetikzlibrary{arrows,calc} \usetikzlibrary{decorations,arrows}
\usetikzlibrary{decorations.pathreplacing}

\usepackage{caption}
\usepackage{subcaption}

\usepackage{lipsum}
\usepackage{ifthen}
\usepackage{xifthen} 
\usepackage{xparse}
\usepackage{pgffor}
\usepackage{xspace} %
\usepackage{footnote}

\numberwithin{equation}{section}

\usepackage{enumitem}

%% file: header_macros.tex
\theoremstyle{plain}

\definecolor{lightblue}{rgb}{0,0.2,0.4}

\usepackage{framed}
\makesavenoteenv{shaded}
\newtheorem{prototheorem}{Theorem}[chapter]

\makeatletter
\renewenvironment{proof}[1][]{
  \par\pushQED{\qed}%
  \normalfont \topsep6\p@\@plus6\p@\relax
  \trivlist
  \item[\hskip\labelsep\bfseries
    \proofname
    \if\relax\detokenize{#1}\relax\else\space #1\fi
    \@addpunct{.}]%
  \ignorespaces
}{
  \popQED\endtrivlist\@endpefalse
}
\makeatother

\newenvironment{theorem}[1][]{\colorlet{shadecolor}{blue!8}\begin{shaded}\begin{prototheorem}[#1]}{\end{prototheorem}\end{shaded}}

\newtheorem{protolemma}[prototheorem]{Lemma}
\newenvironment{lemma}[1][]{\colorlet{shadecolor}{blue!10}\begin{shaded}\begin{protolemma}[#1]}{\end{protolemma}\end{shaded}}

\newtheorem{protocorollary}[prototheorem]{Corollary}
\newenvironment{corollary}[1][]{\colorlet{shadecolor}{blue!10}\begin{shaded}\begin{protocorollary}[#1]}{\end{protocorollary}\end{shaded}}

\newtheorem{protoproposition}[prototheorem]{Proposition}
\newenvironment{proposition}[1][]{\colorlet{shadecolor}{blue!10}\begin{shaded}\begin{protoproposition}[#1]}{\end{protoproposition}\end{shaded}}

\theoremstyle{definition}
\newtheorem{protodefinition}[prototheorem]{Definition}
\newenvironment{definition}[1][]{\colorlet{shadecolor}{blue!10}\begin{shaded}\begin{protodefinition}[#1]}{\end{protodefinition}\end{shaded}}

\newtheorem{assumption}[prototheorem]{Assumption}
\newtheorem{exercise}[prototheorem]{Exercise}
\newtheorem{protoexample}[prototheorem]{Example}
\newenvironment{example}[1][]{%
  \pushQED{\qed}%
  \begin{protoexample}[#1]%
}{%
  \popQED%
  \end{protoexample}%
}

\theoremstyle{remark}
\newtheorem{remark}[prototheorem]{Remark}

\newcommand{\dup}[3][]{\left\langle #2, #3\right\rangle_{#1}}
\newcommand{\inp}[3][]{\left\langle #2, #3\right\rangle_{#1}}
\newcommand{\inpc}[3][]{\left\langle #2, #3\right\rangle_{#1}}

\newcommand{\norm}[2][]{\| #2 \|_{#1}}
\newcommand{\normc}[2][]{\left\| #2 \right\|_{#1}}

\newcommand{\set}[2]{\{#1\,|\,#2\}}
\newcommand{\setc}[2]{\left\{#1\, \middle|\,#2\right\}}

\newcommand{\dd}{\,\mathrm{d}}

\newcommand{\dfn}{\vcentcolon=}
\newcommand{\dfnn}{=\vcentcolon}

\DeclareMathOperator{\spa}{span}

\DeclareMathOperator{\supp}{supp}

\DeclareMathOperator{\argmin}{argmin}

\DeclareMathOperator{\ii}{i}

\newcommand{\cc}{{\rm cc}}
\newcommand{\co}{{\rm co}}
\newcommand{\aff}{{\rm aff}}

\newcommand{\depth}{{\rm depth}}
\newcommand{\wdth}{{\rm width}}
\newcommand{\size}{{\rm size}}
\newcommand{\Bnul}{\boldsymbol{0}}
\newcommand{\Bone}{\boldsymbol{1}}

\newcommand{\nId}[1]{{\Phi^{\rm id}_{#1}}} %
\newcommand{\nmin}[1]{{\Phi^{\min}_{#1}}} %
\newcommand{\nmax}[1]{{\Phi^{\max}_{#1}}} %
\newcommand{\ntim}[1]{{\Phi^{\times}_{#1}}} %
\newcommand{\tntim}[1]{{\tilde\Phi^{\times}_{#1}}} %

%% file: header_letters.tex
\newcommand{\objF}{F}

\newcommand{\gauss}{{\rm N}}

\newcommand{\risk}{\mathcal{R}}

\newcommand{\eps}{\varepsilon}
\newcommand{\C}{\mathbb{C}}
\newcommand{\N}{\mathbb{N}}
\newcommand{\Z}{\mathbb{Z}}
\newcommand{\R}{\mathbb{R}}
\newcommand{\Q}{\mathbb{Q}}

\renewcommand{\P}{\mathbb{P}}
\newcommand{\ind}{\mathds{1}}

\DeclareFontEncoding{FMS}{}{}
\DeclareFontSubstitution{FMS}{futm}{m}{n}
\DeclareFontEncoding{FMX}{}{}
\DeclareFontSubstitution{FMX}{futm}{m}{n}
\DeclareSymbolFont{fouriersymbols}{FMS}{futm}{m}{n}
\DeclareSymbolFont{fourierlargesymbols}{FMX}{futm}{m}{n}
\DeclareMathDelimiter{\VERT}{\mathord}{fouriersymbols}{152}{fourierlargesymbols}{147}

\providecommand{\bbE}{\mathbb{E}}

\providecommand{\bbH}{\mathbb{H}}

\providecommand{\bbK}{\mathbb{K}}

\providecommand{\bbP}{\mathbb{P}}
\providecommand{\bbQ}{\mathbb{Q}}

\providecommand{\bbV}{\mathbb{V}}

\providecommand{\CA}{\mathcal{A}}

\providecommand{\CC}{\mathcal{C}}
\providecommand{\CD}{\mathcal{D}}

\providecommand{\CF}{\mathcal{F}}

\providecommand{\CG}{\mathcal{G}}
\providecommand{\CH}{\mathcal{H}}
\providecommand{\CI}{\mathcal{I}}

\providecommand{\CL}{\mathcal{L}}
\providecommand{\CM}{\mathcal{M}}
\providecommand{\CN}{\mathcal{N}}

\providecommand{\CP}{\mathcal{P}}

\providecommand{\CR}{\mathcal{R}}
\providecommand{\CS}{\mathcal{S}}
\providecommand{\CT}{\mathcal{T}}

\providecommand{\CV}{\mathcal{V}}
\providecommand{\CW}{\mathcal{W}}

\providecommand{\BA}{{\boldsymbol{A}}}

\providecommand{\BC}{{\boldsymbol{C}}}
\providecommand{\BD}{{\boldsymbol{D}}}

\providecommand{\BG}{{\boldsymbol{G}}}
\providecommand{\BH}{{\boldsymbol{H}}}
\providecommand{\BI}{{\boldsymbol{I}}}

\providecommand{\BK}{{\boldsymbol{K}}}

\providecommand{\BM}{{\boldsymbol{M}}}

\providecommand{\BP}{{\boldsymbol{P}}}
\providecommand{\BQ}{{\boldsymbol{Q}}}

\providecommand{\BU}{{\boldsymbol{U}}}
\providecommand{\BV}{{\boldsymbol{V}}}
\providecommand{\BW}{{\boldsymbol{W}}}
\providecommand{\BX}{{\boldsymbol{X}}}
\providecommand{\BY}{{\boldsymbol{Y}}}
\providecommand{\BZ}{{\boldsymbol{Z}}}

\providecommand{\Ba}{{\boldsymbol{a}}}
\providecommand{\Bb}{{\boldsymbol{b}}}
\providecommand{\Bc}{{\boldsymbol{c}}}

\providecommand{\Be}{{\boldsymbol{e}}}

\providecommand{\Bg}{{\boldsymbol{g}}}

\providecommand{\Bk}{{\boldsymbol{k}}}

\providecommand{\Bm}{{\boldsymbol{m}}}

\providecommand{\Bp}{{\boldsymbol{p}}}
\providecommand{\Bq}{{\boldsymbol{q}}}

\providecommand{\Bs}{{\boldsymbol{s}}}
\providecommand{\Bt}{{\boldsymbol{t}}}
\providecommand{\Bu}{{\boldsymbol{u}}}
\providecommand{\Bv}{{\boldsymbol{v}}}
\providecommand{\Bw}{{\boldsymbol{w}}}
\providecommand{\Bx}{{\boldsymbol{x}}}
\providecommand{\By}{{\boldsymbol{y}}}
\providecommand{\Bz}{{\boldsymbol{z}}}

\newcommand{\VI}{{\mathbf{I}}}

\newcommand{\VV}{{\mathbf{V}}}

\newcommand{\BSigma}  {{\boldsymbol{\mathit{\Sigma}}}}

\newcommand{\Balpha}     {{\boldsymbol{\alpha}}}
\newcommand{\Bbeta}      {{\boldsymbol{\beta}}}

\newcommand{\Bzeta}      {{\boldsymbol{\zeta}}}
\newcommand{\Beta}       {{\boldsymbol{\eta}}}                %
\newcommand{\Btheta}     {{\boldsymbol{\theta}}}

\newcommand{\Bmu}        {{\boldsymbol{\mu}}}
\newcommand{\Bnu}        {{\boldsymbol{\nu}}}
\newcommand{\Bxi}        {{\boldsymbol{\xi}}}

\DeclareFontFamily{U}{mathx}{\hyphenchar\font45}
\DeclareFontShape{U}{mathx}{m}{n}{
      <5> <6> <7> <8> <9> <10>
      <10.95> <12> <14.4> <17.28> <20.74> <24.88>
      mathx10
      }{}
\DeclareSymbolFont{mathx}{U}{mathx}{m}{n}
\DeclareMathAccent{\widecheck}{0}{mathx}{"71}

%% file: preface.tex
\chapter*{Preface}
This book serves as an introduction to the key ideas in the mathematical analysis of deep learning. It is designed to help students and researchers to quickly familiarize themselves with the area and to provide a foundation for the development of university courses on the mathematics of deep learning. Our main goal in the composition of this book was to present various rigorous, but easy to grasp, results that help to build an understanding of fundamental mathematical concepts in deep learning. To achieve this, we prioritize simplicity over generality.

As a mathematical introduction to deep learning, this book does not aim to give an exhaustive survey of the entire (and rapidly growing) field, and some important research directions are missing.
In particular, we have favored mathematical results over empirical research, even though an accurate account of the theory of deep learning requires both.

The book is intended for students and researchers in mathematics and related areas. While we believe that every diligent researcher or student will be able to work through this manuscript, it should be emphasized that a familiarity with analysis, linear algebra, probability theory, and basic functional analysis is recommended for an optimal reading experience. To assist readers, a review of key concepts in probability theory and functional analysis is provided in the appendix.

The material is structured around the three main pillars of deep learning theory: Approximation theory, Optimization theory, and Statistical Learning theory. This structure, which corresponds to the three error terms typically occuring in the theoretical analysis of deep learning models, is inspired by other recent texts on the topic following the same outline \cite{doi:10.1073/pnas.1907369117,telgarskynotes,jentzen2023mathematical}.
More specifically, Chapter \ref{chap:intro} provides an overview and introduces key questions  for understand deep learning. Chapters \ref{chap:FFNNs} - \ref{chap:Interpolation} explore results in approximation theory, %
Chapters \ref{chap:training} - \ref{chap:shape} discuss optimization theory for deep learning, and Chapters \ref{chap:VC} - \ref{chap:adversarial} address the statistical aspects of deep learning.
In the final Chapter \ref{chap:ModArchitectures}, we discuss various modifications to the computational architectures of the previous chapters, which are relevant in practice.

This book is the result of a series of lectures %
given by the authors.
Parts of the material were presented by P.P.\ in a lecture titled ``Neural Network Theory'' at the University of Vienna, and by J.Z.\ in a lecture titled ``Theory of Deep Learning'' at Heidelberg University.
The lecture notes of these courses formed the basis of this book. We are grateful to the many colleagues and students who contributed to this book through insightful discussions and valuable suggestions. We would like to offer special thanks to the following individuals:

Jonathan Garcia Rebellon,
Jakob Lanser,
Andr\'es Felipe Lerma Pineda,
Marvin Ko{\ss},
Martin Mauser,
Davide Modesto,
Martina Neuman,
Bruno Perreaux,
Johannes Asmus Petersen,
Milutin Popovic,
Tuan Quach,
Tim Rakowski,
Lorenz Riess,
Jakob Fabian Rohner,
Jonas Schuhmann,
Peter \v{S}koln\'ik,
Matej Vedak,
Simon Weissmann,
Josephine Westermann,
Ashia Wilson.

%% file: Notation.tex
\section*{Notation}

\DefTblrTemplate{caption}{default}{}
\DefTblrTemplate{conthead}{default}{}
\DefTblrTemplate{capcont}{default}{}

\begin{longtblr}[
  caption = {Notation used in this book},
	label = {tab:notation},
	]{
		colspec = {|t{0.15\linewidth}|m{0.53\linewidth}|b{0.23\linewidth}|},
		rowhead = 1,
		hlines,
		row{even} = {verylightgray},
		row{1} = {lightgray},
	} 
	\textbf{Symbol} & \textbf{Description} & \textbf{Reference}\\

    $\CA$& vector of layer widths & Definition \ref{def:realizationetc}\\
    $\mathfrak{A}$& a sigma-algebra & Definition \ref{def:sigmaalgebra} \\
    $\aff(S)$& affine hull of $S$ & \eqref{eq:affHull}\\
    $\mathfrak{B}_d$& the Borel sigma-algebra on $\R^d$ & Section \ref{sec:sigtopmeas} \\
    $\mathcal{B}^n$& B-Splines of order $n$ & Definition \ref{def:multivariateBsplines}\\
    $B_r(x)$ & ball of radius $r\ge 0$ around $x$ in a metric space $X$ & \eqref{eq:ball} \\
    $B_r^d$& ball of radius $r\ge 0$ around $\Bnul$ in $\R^d$ &  \\
    $B_1^{k,d}$& ball of radius $1$ around $0$ in $C^k([0,1]^d)$ &  \eqref{eq:B1kd} \\
    $C^{0,s}(\Omega)$& $s$-H\"older continuous functions from $\Omega\to\R$ & Definition \ref{def:hoelder} \\
    $C^k(\Omega)$& $k$-times continuously differentiable functions from $\Omega\to\R$ & Definition \ref{def:ckSpace} \\
    $C^{k,s}(\Omega)$& $C^k(\Omega)$ functions $f$ for which $f^{(k)}\in C^{0,s}(\Omega)$ & Definition \ref{def:ckalphaSpace}\\
    $C^\infty_c(\Omega)$& infinitely differentiable functions from $\Omega\to\R$ with compact support in $\Omega$ & \\        
    $f_n \xrightarrow{\cc} f$ & compact convergence of $f_n$ to $f$ & Definition \ref{def:compactConvergence}  \\
    $\co(S)$ & convex hull of a set $S$ & \eqref{eq:convexhull}\\
    $f * g$ & convolution of $f$ and $g$ & \eqref{eq:convolution}\\
    $\CD$& data distribution & \eqref{eq:riskDef0}, Section \ref{sec:LearningSetup}\\
    $D^\Balpha f$ & partial derivative of $f$ w.r.t.\ multiindex $\Balpha$ & \\
    ${\rm depth}(\Phi)$ & depth of $\Phi$ & Definition \ref{def:nn}\\
    $\eps_{\mathrm{approx}}$& approximation error &\eqref{eq:biasVarianceTradeOff}\\
    $\eps_{\mathrm{gen}}$ & generalization error & \eqref{eq:biasVarianceTradeOff}\\
    $\eps_{\mathrm{int}}$& interpolation error & \eqref{eq:interpolationVarianceTradeOff}\\
    $\bbE[X]$ & expectation of random variable $X$& \eqref{eq:expectationDefinition}\\
    $\bbE[X|Y]$ & conditional expectation of random variable $X$& Subsection \ref{sec:conddist}\\
    $\mathcal{G}(S, \eps, X)$ & $\eps$-covering number of a set $S \subseteq X$ & Definition \ref{def:coveringNumber}\\
    $\gamma_{k,d,N}$ & continuous nonlinear $N$-width & \eqref{eq:nonlnwidth} \\    
    $\Gamma_C$ & Barron space with constant $C$ & Section \ref{sec:BarronClass}\\
    $\nabla_x f$ & gradient of $f$ w.r.t.\ $x$, in finite dimensions equal to $(\frac{\partial f}{\partial x})^\top$ &\\
    $\oslash$& componentwise (Hadamard) division & \\ %
    $\otimes$& componentwise (Hadamard) product & \\ %
    $h_S$& empirical risk minimizer for a sample $S$& Definition \ref{def:empiricalRiskMinimizer}\\
    $\BI_d$ & $d\times d$ identity matrix & \\    
    $\nId{L}$ & identity ReLU neural network & Lemma \ref{lemma:identity}\\
    $\ind_S$ & indicator function of the set $S$ &\\     %
    $\dup{\cdot}{\cdot}$& Euclidean inner product on $\R^d$ & \\
    $\dup[H]{\cdot}{\cdot}$& inner product on a vector space $H$ & Definition \ref{def:innerProduct}\\
    $k_\CT$ & maximal number of elements shared by a single node of a triangulation & \eqref{eq:kT}\\
    $\hat K_n(\Bx,\Bx')$ & empirical tangent kernel& \eqref{eq:etk}\\
    $\Lambda_{\mathcal{A}, \sigma, S, \mathcal{L}}$& loss landscape defining function &Definition \ref{def:lossLandscape}\\
    ${\rm Lip}(f)$& Lipschitz constant of a function $f$ & \eqref{eq:definitionOfLipschitzConstant} \\
    ${\rm Lip}_M(\Omega)$ & $M$-Lipschitz continuous functions on $\Omega$ &    \eqref{eq:lipM}\\
    $\mathcal{L}$ & general loss function & Section \ref{sec:LearningSetup}\\
    $\mathcal{L}_{0-1}$ & 0-1 loss & Section \ref{sec:LearningSetup}\\
    $ \mathcal{L}_{\rm ce}$ & binary cross entropy loss & Section \ref{sec:LearningSetup}\\
    $\mathcal{L}_2$ & square loss & Section \ref{sec:LearningSetup}\\
    $\ell^p(\N)$& space of $p$-summable sequences indexed over $\N$ & Section \ref{app:BanachSpaces}\\    
    $L^p(\Omega)$& Lebesgue space over $\Omega$ & Section \ref{app:BanachSpaces}\\
    $\CM$ & piecewise continuous and locally bounded functions & Definition \ref{eq:M}\\
    $\CN_d^m(\sigma;L,n)$ & set of multilayer perceptrons with $d$-dim input, $m$-dim output, activation function $\sigma$, depth $L$, and width $n$ & Definition \ref{def:CN}\\
    $\CN_d^m(\sigma;L)$ & union of $\CN_d^m(\sigma;L,n)$ for all $n\in\N$ & Definition \ref{def:CN}\\
    $\CN(\sigma;\CA, B)$ & set of neural networks with architecture $\CA$, activation function $\sigma$ and all weights bounded in modulus by $B$& Definition \ref{def:realizationetc}\\
    $\CN^{\rm sp}(\sigma;L,B,s)$ & set of neural networks with depth at most $L$, at most $s$ nonzero weights, and all weights bounded in modulus by $B$ & Definition \ref{def:Nsp}\\    
    $\CN^{{\rm sp},*}(\sigma; L, B,s)$ & neural networks in $\CN^{\rm sp}(\sigma;B,s)$ with range in $[-1,1]$ & \eqref{eq:CNstar} \\
    $\N$& positive natural numbers & \\
    $\N_0$& natural numbers including zero & \\
    $\gauss(\Bm,\BC)$& multivariate normal distribution with mean $\Bm\in\R^d$ and covariance $\BC\in\R^{d\times d}$ &\\
    $n_\CA$ & number of parameters of a neural network with layer widths described by $\CA$ & Definition \ref{def:realizationetc}\\
    $\norm[]{\cdot}$& Euclidean norm for vectors in $\R^d$ and spectral norm for matrices in $\R^{n\times d}$ & \\
    $\norm[F]{\cdot}$& Frobenius norm for matrices & \\
    $\norm[\infty]{\cdot}$& $\infty$-norm on $\R^d$ or supremum norm for functions & \eqref{eq:inftynorm} \\
    $\norm[p]{\cdot}$& $p$-norm on $\R^d$ & \\
    $\norm[X]{\cdot}$& norm on a vector space $X$ & \\
    $\Bnul$& zero vector or zero matrix &\\
    $O(\cdot)$& Landau notation & \\ %
    $\omega(\eta)$ & patch of the node $\eta$ & \eqref{eq:patchDefinition}\\
    $\Omega_\Lambda(c)$& sublevel set of loss landscape& Definition \ref{def:spuriousValley}\\
    ${\Bone}$& constant $1$ vector& \\
    $\partial f(\Bx)$ & set of subgradients of $f$ at $\Bx$ & Definition \ref{def:subgrad} \\
    $\frac{\partial f}{\partial\Bx}$ & partial derivative; if $f:\R^d\to\R^k$, then $\frac{\partial f(\Bx)}{\partial\Bx}\in\R^{k\times d}$ and analogous for tensor valued $f$ and $\Bx$ &  \\
    $\CP_n(\R^d)$ or $\CP_n$ & space of multivariate polynomials of degree $n$ on $\R^d$ & Example \ref{ex:poly}\\
    $\CP(\R^d)$ or $\CP$ & space of multivariate polynomials of arbitrary degree on $\R^d$ & Example \ref{ex:poly} \\        
    $\bbP_X$ & distribution of random variable $X$ & Definition \ref{def:distribution}\\
    $\bbP[A]$ & probability of event $A$ & Definition  \ref{def:probMeasure}\\
    $\bbP[A|B]$ & conditional probability of event $A$ given $B$ & Definition  \ref{eq:PAB}\\
    $\mathcal{PN}(\CA, B)$ & parameter set of neural networks with architecture $\CA$ and all weights bounded in modulus by $B$& Definition \ref{def:realizationetc}\\
    ${\rm Pieces}(f, \Omega)$ & number of pieces of $f$ on $\Omega$& Definition \ref{def:NumOfPieces}\\
    $\Phi(\Bx)$ & model (e.g.\ neural network) in terms of input $\Bx$ (parameter dependence suppressed) \\
    $\Phi(\Bx,\Bw)$ & model (e.g.\ neural network) in terms of input $\Bx$ and parameters $\Bw$\\    
    $\Phi^{\rm lin}$ & linearization around initialization& \eqref{eq:Philin}\\
    $\nmin{n}$& minimum neural network& Lemma \ref{lemma:minmaxn}\\
    $\ntim{\eps}$& multiplication neural network & Lemma \ref{lemma:mult}\\
    $\ntim{n,\eps}$ & multiplication of $n$ numbers neural network & Proposition \ref{prop:multn}\\
    $\Phi_2\circ\Phi_1$ & composition of neural networks &Lemma \ref{lemma:composition}\\ %
    $\Phi_2\bullet\Phi_1$ & sparse composition of neural networks & Lemma \ref{lemma:composition}\\ %
    $(\Phi_1, \dots, \Phi_m)$& parallelization of neural networks & \eqref{eq:map_parallel}\\ %
    $\BA^\dagger$& pseudoinverse of a matrix $\BA$ & \\
    $\Q$& rational numbers & \\
    $\R$& real numbers & \\
    $\R_-$& non-positive real numbers & \\
    $\R_+$& non-negative real numbers & \\
    $R_\sigma$ & Realization map & Definition \ref{def:realizationetc}\\
    $R^*$ & Bayes risk & \eqref{eq:BayesRiskGenSection}\\
    $\mathcal{R}(h)$ & risk of hypothesis $h$ & Definition \ref{def:genError}\\
    $\widehat{\mathcal{R}}_S(h)$ & empirical risk of $h$ for sample $S$ & \eqref{eq:empiricalRiskDef0}, Definition \ref{def:erisk}\\
    $\CS_n$& cardinal B-spline & Definition \ref{def:CardiBSpline}\\
    $\CS_{\ell, \Bt, n}^d$& multivariate cardinal B-spline & Definition \ref{def:multivariateBsplines}\\
    $|S|$ & cardinality of an arbitrary set $S$, or Lebesgue measure of $S\subseteq\R^d$&\\             %
    $\mathring{S}$& interior of a set $S$& \\
    $\overline{S}$& closure of a set $S$& \\
    $\partial{S}$& boundary of a set $S$& \\    
    $S^c$& complement of a set $S$& \\
    $S^\perp$& orthogonal complement of a set $S$ & Definition \ref{def:orthogonal}\\
    $\sigma$ & general activation function &\\    
    $\sigma_{a}$ & parametric ReLU activation function & Section \ref{sec:activationFunctions}\\
    $\sigma_{\rm ReLU}$ & ReLU activation function & Section \ref{sec:activationFunctions}\\
    ${\rm sign}$& sign function &\\
    $s_{\rm max}(\BA)$& maximal singular value of a matrix $\BA$ &\\
    $s_{\rm min}(\BA)$& minimal (positive) singular value of a matrix $\BA$ &\\    
    $\size(\Phi)$ & number of free network parameters in $\Phi$ & Definition \ref{def:SizeOfNNDefinition} \\
    ${\rm span}(S)$& linear hull or span of $S$&\\
    $\CT$ & triangulation & Definition \ref{def:mesh}\\
    $\CV$& set of nodes in a triangulation & Definition \ref{def:mesh} \\    
    $\bbV[X]$ & variance of random variable $X$ & Section \ref{sec:distAndExp}\\
    $\mathrm{VCdim}(\mathcal{H})$& VC dimension of a set of functions $\mathcal{H}$ & Definition \ref{def:VCDim} \\
    $\CW$& distribution of weight initialization & Section \ref{sec:ntksetting} \\
    $\BW^{(\ell)}, \Bb^{(\ell)}$ & weights and biases in layer $\ell$ of a neural network & Definition \ref{def:nn}\\
    ${\rm width}(\Phi)$ & width of $\Phi$ & Definition \ref{def:nn}\\
    $\Bx^{(\ell)}$ & output of $\ell$-th layer of a neural network & Definition \ref{def:nn}\\
    $\bar\Bx^{(\ell)}$ & preactivations & \eqref{eq:defxj} \\
    $X'$ & dual space to a normed space $X$ & Definition \ref{def:DualSpace}\\
    \end{longtblr}

%% file: Introduction.tex
\chapter{Introduction}\label{chap:intro}

\section{Mathematics of deep learning}
In 2012, a deep learning architecture revolutionized the field of computer vision by achieving unprecedented performance in the ImageNet Large Scale Visual Recognition Challenge (ILSVRC) \cite{krizhevsky2012imagenet}.
 The deep learning architecture, known as AlexNet, significantly outperformed all competing approaches. 
 A few years later, in March 2016, a deep learning-based architecture called AlphaGo defeated the best Go player at the time, Lee Sedol, in a five-game match \cite{silver2016mastering}.
 Go is a highly complex board game with a vast number of possible moves, making it a challenging problem for artificial intelligence. 
Because of this complexity, many researchers believed that defeating a top human Go player was a feat that would only be achieved decades later.

These breakthroughs along with many others, have sparked interest among scientists across (almost) all disciplines.
Prominent examples include DeepMind's AlphaFold \cite{jumper2021highly}, which revolutionized protein structure prediction in 2020 and earned its developers the
Nobel Prize in Chemistry in 2024, the unprecedented language capabilities of large language models like GPT-3 (and later versions) \cite{vaswani2017attention,brown2020language}, and the emergence of generative AI models like Stable Diffusion, Midjourney, DALL-E, and Gemini 2.5 Flash Image (better known as Nano Banana). Likewise, while mathematical research on neural networks has a long history, these groundbreaking developments revived interest in the theoretical underpinnings of deep learning among mathematicians. %
However, initially, there was a clear consensus in the mathematics community:
\emph{We do not understand why this technology works so well! In fact, there are
  many mathematical reasons that, at least superficially,
  should prevent the observed success.}

Over the past decade the field has matured,
and mathematicians %
have gained a more profound understanding of
deep learning, %
although many open questions remain. 
Recent years have %
brought various new explanations and insights into the inner workings of %
these models.
Before discussing them in detail in the following chapters,
 we %
 first give a high-level introduction to deep learning,
 with a focus on the supervised learning framework for function approximation,
   which is the central theme of this book.

\section{High-level overview of deep learning}
\label{sec:highlevOverview}
Deep learning refers to the application of deep neural networks trained by gradient-based methods, to identify unknown input-output relationships.
This approach has three key ingredients: \textit{deep neural networks, gradient-based training,
  and prediction}.
 We now explain each of these ingredients separately.

\begin{figure}[htb]
	\centering
	\includegraphics[width = 0.7\textwidth]{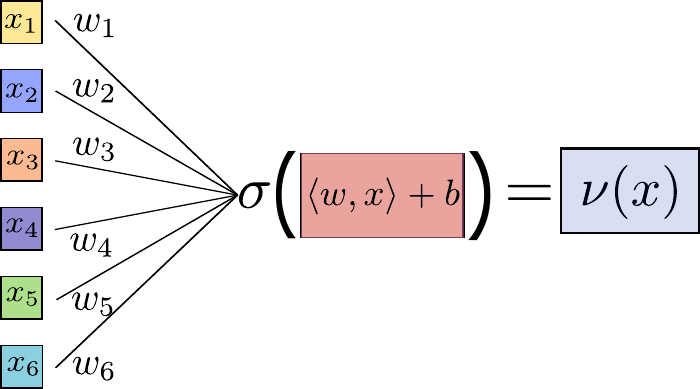}
	\caption{Illustration of a single neuron $\nu$.
          The neuron receives six inputs $(x_1, \dots, x_6) = \Bx$,
 computes their weighted sum $\sum_{j=1}^6x_j w_j$, 
 adds a bias $b$, and finally applies the activation function $\sigma$
 to produce the output $\nu(x)$.
}
	\label{fig:Neuron}
\end{figure}

\paragraph{\bf Deep Neural Networks}
Deep neural networks are %
formed by a combination of neurons.
 A \textbf{neuron} is a function of the form 
\begin{align}\label{eq:neuron}
	\R^d \ni \Bx \mapsto \nu(\Bx) = \sigma( \Bw^\top \Bx +b ),
\end{align}
where $\Bw \in \R^d$ is a {\bf weight vector}, $b\in \R$ is called {\bf bias}, and the function $\sigma$ is %
referred to as an \textbf{activation function}.
 This concept is due to McCulloch and Pitts \cite{mcculloch1943logical} and is a mathematical model for biological neurons.
 If %
 we consider
   $\sigma$ %
   to be the Heaviside function, i.e.,
   $\sigma = \ind_{\R_+}$ with $\R_+\dfn [0,\infty)$, then the neuron ``fires'' if the weighted sum of the inputs $\Bx$ surpasses the threshold $-b$.
 We depict a neuron in Figure \ref{fig:Neuron}.
 Note that if we fix $d$ and $\sigma$, then the set of neurons can be naturally parameterized %
 by the $d+1$ real values $w_1,\dots,w_d,b\in\R$.

Neural networks are functions formed by connecting neurons, where the output of one neuron becomes the input to another. One simple but very common type of neural network is the so-called feedforward neural network.
 This structure distinguishes itself by having the neurons grouped in layers, and the inputs to neurons in the $(\ell+1)$-st layer are exclusively neurons from the $\ell$-th layer. 

We start by defining a \textbf{shallow feedforward neural network} as an affine transformation applied to the output of a set of neurons that share the same input and the same activation function.
Here, an \textbf{affine transformation} is a map $T:\R^p\to\R^q$ such that $T(\Bx)=\BW\Bx+\Bb$ for some $\BW\in\R^{q\times p}$, $\Bb\in\R^q$ where $p$, $q \in \N$.

\begin{figure}[htb]
	\centering
	\includegraphics[width = 0.8\textwidth]{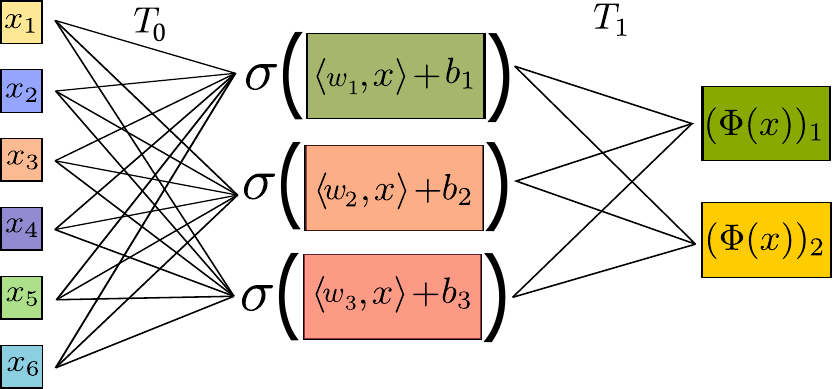}
	\caption{Illustration of a shallow neural network.
          The affine transformation $T_0$ is of the form $(x_1, \dots, x_6) = \Bx\mapsto \BW\Bx + \Bb$, where the rows of $\BW$ are the weight vectors $\Bw_1$, $\Bw_2$, $\Bw_3$ for each respective neuron.
 }
	\label{fig:ShallowNeuralNet}
      \end{figure}

Formally, a shallow feedforward neural network is, therefore, a map $\Phi$ of the form 
\[
  \R^d \ni \Bx \mapsto \Phi(\Bx) = %
  T_1\circ\sigma\circ T_0(\Bx)
\]
where $T_0$, $T_1$ are affine transformations and the application of $\sigma$ is understood to be 
in each %
  component
  of $T_1(\Bx)$. %
 A visualization of a shallow neural network is given in Figure \ref{fig:ShallowNeuralNet}.

 A  \textbf{deep feedforward neural network} is constructed by %
compositions of %
 shallow neural networks. This yields a map of the type
\[
  \R^d \ni \Bx \mapsto \Phi(\Bx) = %
  T_{L+1}\circ\sigma\circ \cdots \circ T_1 \circ\sigma\circ T_0(\Bx),
\]
where $L \in \N$ and $(T_j)_{j= 0}^{L+1}$ are affine transformations.
 The number of compositions $L$ is referred to as the \textbf{number of layers} of the deep neural network.
Similar to a single neuron, (deep) neural networks can be viewed as a parameterized function class, with the {\bf parameters} being the entries of the matrices and vectors determining the affine transformations $(T_j)_{j= 0}^{L+1}$.

\paragraph{\bf Gradient-based training}
After defining the structure or {\bf architecture} of the neural
  network, e.g., the activation function and the number of layers, the
  second step of deep learning consists of determining %
  suitable values
  for its parameters.
  In practice, this is achieved by
minimizing an {\bf objective function}. 
In \textbf{supervised learning}, %
which will be our focus, this
objective %
depends on
  a collection of input-output pairs, commonly known as {\bf training data} or simply as a \textbf{sample}. 
Concretely, let %
$S = (\Bx_i, \By_i)_{i=1}^m$ be %
a sample, where $\Bx_i \in \R^d$ represents the inputs and $\By_i \in \R^k$ the corresponding outputs with $d$, $k \in \N$.
Our goal is to find a deep neural network $\Phi$ such that %
\begin{align}\label{eq:protoPrediction}
	\Phi(\Bx_i) \approx \By_i\qquad\text{for all } i=1,\dots,m
\end{align}
in a %
meaningful sense.
For example, we could %
interpret
``$\approx$'' 
to mean closeness
with respect to the Euclidean norm,
or more generally, that $\mathcal{L}(\Phi(\Bx_i), \By_i)$ is small for a function $\mathcal{L}$
measuring the dissimilarity between its inputs.
Such a function $\mathcal{L}$ is called a \textbf{loss function}.
A standard way of %
achieving
\eqref{eq:protoPrediction} is by minimizing the so-called \textbf{empirical risk of $\Phi$} with respect to the sample $S$ defined as
\begin{align}\label{eq:empiricalRiskDef0}
\widehat{\mathcal{R}}_S(\Phi) = \frac{1}{m}\sum_{i=1}^m\mathcal{L}(\Phi(\Bx_i), \By_i).
\end{align}
This quantity serves as a measure of how well $\Phi$ predicts $\By_1, \dots, \By_m$ at the {\bf training points} $\Bx_1, \dots, \Bx_m$.

\begin{figure}[htb]
	\centering
	\includegraphics[width = 0.9\textwidth]{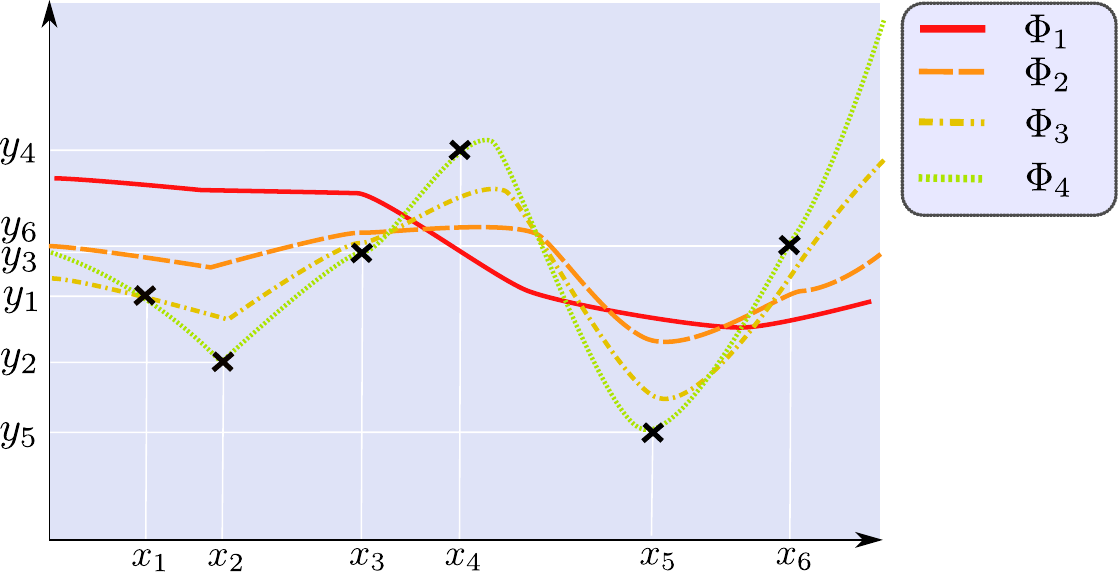}
	\caption{A sequence of one dimensional neural networks $\Phi_1, \dots, \Phi_4$ %
          that successively minimizes the empirical risk for the sample $S = (x_i,y_i)_{i=1}^6$.}
	\label{fig:LearningGradientDescent}
\end{figure}

If $\mathcal{L}$ is differentiable, and for all $\Bx_i$ %
the output $\Phi(\Bx_i)$ depends %
differentiably on the parameters of the neural network, %
then the gradient of the empirical risk $\widehat{\mathcal{R}}_S(\Phi)$
  with respect to the parameters is well-defined.
This gradient can be efficiently computed using a technique called \textbf{backpropagation}. 
This allows to minimize
\eqref{eq:empiricalRiskDef0} by optimization algorithms such as (stochastic) gradient descent. %
They produce a sequence of neural networks parameters,
  and corresponding neural network functions $\Phi_1, \Phi_2, \dots $,
for which the empirical risk is expected to decrease.
Figure \ref{fig:LearningGradientDescent} illustrates a possible behavior of this sequence.

\paragraph{\bf Prediction}
The final part of deep learning concerns the question %
of whether
we have actually learned something by the procedure above.
Suppose that our optimization routine has either converged or %
has been terminated, yielding a neural network $\Phi_*$. 
While the optimization aimed to minimize the empirical risk on the training sample $S$, our ultimate interest is not in how well $\Phi_*$ performs on $S$. %
Rather, we are interested in its performance on new %
data points $(\Bx_{\rm new}, \By_{\rm new})$ outside of $S$.

To make meaningful statements about this, %
we assume %
 existence of a \textbf{data distribution} $\mathcal{D}$ on the input-output space---in our case,
 this is $\R^d \times \R^k$---such that both the elements of $S$ and all other data points %
are drawn from this distribution.
In other words, we treat $S$ %
as an i.i.d.\ draw from $\mathcal{D}$, and $(\Bx_{\rm new}, \By_{\rm new})$
also as sampled independently from $\CD$. %
If we %
want $\Phi_*$ to perform well on average, then this amounts to controlling the following expression
 
\begin{align}\label{eq:riskDef0}
	\mathcal{R}(\Phi_*) = \mathbb{E}_{(\Bx_{\rm new}, \By_{\rm new})  \sim \mathcal{D}}[\mathcal{L}(\Phi_*(\Bx_{\rm new}), \By_{\rm new})],
\end{align}
which is called the \textbf{risk} of $\Phi_*$.
If the risk is not much larger than the empirical risk, then we say that the neural network $\Phi_*$ has a small \textbf{generalization error}.
On the other hand, if the risk is much larger than the empirical risk, then we say that $\Phi_*$  \textbf{overfits} the training data, meaning that $\Phi_*$ has memorized the training samples, but does not generalize well to data outside of the training set.

\section{Why does it work?}
It is natural to wonder why the deep learning pipeline, as outlined in the previous subsection,
ultimately succeeds in learning, i.e., achieving a small risk.
Is it true that for a given sample $(\Bx_i,\By_i)_{i=1}^m$ there exist a neural network such that $\Phi(\Bx_i) \approx \By_i$ for all $i = 1, \dots m$? 
Does the optimization routine produce
a meaningful result? 
Can we control the risk, knowing only that the empirical risk is small?

While most of these questions can be answered affirmatively under certain assumptions, these assumptions often do not apply to deep learning in practice.
We next explore some potential explanations and show that they lead to even more questions.

\paragraph{\bf Approximation}
A fundamental result in the study of
          neural networks is the so-called universal approximation theorem, which will be discussed in Chapter \ref{chap:UA}.
 This result states that every continuous function on a compact domain can be approximated arbitrarily well (in a uniform sense) by a shallow neural network.

 This result, however, does not %
 address the practically relevant question of efficiency.
      For example, if we %
      aim for computational efficiency,
      then we may be interested in identifying the smallest %
      neural network that fits the data.
      This naturally raises the question:
      \emph{What is the role of the architecture for the expressive capabilities of neural networks? }
      Furthermore, %
      viewing empirical risk minimization as an approximation problem, we are confronted with a central challenge in
      approximation theory: %
      the \emph{curse of dimensionality}. 
      Function approximation in high dimensions is %
      notoriously difficult and
      becomes exponentially harder %
      as the dimensionality increases.
      Yet, many successful deep learning architectures operate in this high-dimensional regime.
      \emph{Why do these neural networks appear to overcome this so-called curse?}

        \paragraph{\bf Optimization}
          While gradient descent can sometimes be %
          proven to converge to a global minimum,
          as we will discuss in Chapter \ref{chap:training},
          this typically requires the objective function to be at least convex. 
          However, there is no reason to believe that for example the empirical risk is a convex function of the network parameters.
          In fact, due to the repeatedly occurring compositions with the nonlinear activation function in the network, %
          the empirical risk is typically \emph{highly non-linear and not convex}. 
	Therefore, %
        there is %
        generally no guarantee that the optimization routine %
        will converge to a global minimum, and
        it may get stuck in a local (and non-global) minimum or %
        a saddle point. 
	\emph{Why is the output of the optimization nonetheless often meaningful in practice?}
        
        \paragraph{\bf Generalization}
          In traditional statistical learning theory, which we will review in Chapter \ref{chap:VC}, %
          the extent to which the risk exceeds the empirical risk, can be bounded a priori;
            such bounds are often expressed in terms of a notion of complexity of the set of admissible functions (the class of neural networks) %
          divided by the number of training samples.
			For the class of neural networks of a fixed architecture,
            the complexity roughly amounts to the number
          of neural network parameters.
          In practice, typically neural networks with \emph{more} parameters than training samples are used. %
          This is dubbed the \emph{overparameterized regime}. 
          In this regime, the classical estimates described above are void.
          
          Why is it that, nonetheless, \emph{deep overparameterized architectures
            are capable of making
            accurate predictions} on unseen data?
          Furthermore, while deep architectures often generalize well, they sometimes fail
          spectacularly on specific, carefully crafted examples. In image classification tasks, these examples may differ only slightly from correctly classified images in a way that is not perceptible to the human eye.
          Such %
          examples are known as
          \emph{adversarial examples},
          and their existence poses a great challenge for applications of deep learning.

\section{Outline and philosophy}
This book addresses the questions raised in the previous section, providing answers that are mathematically rigorous and accessible. Our focus will be on provable statements, presented in a manner that prioritizes simplicity and clarity over generality. We will sometimes illustrate key ideas only in special cases, or under strong assumptions, both to avoid an overly technical exposition, and because definitive answers are often not yet available. In the following, we summarize the content of each chapter and highlight parts pertaining to the questions stated in the previous section.

	{\bf Chapter \ref{chap:FFNNs}: Feedforward neural networks.
 }
	In this %
        chapter, we introduce the main object of study of this %
        book---the feedforward neural network.
 	
	{\bf Chapter \ref{chap:UA}: Universal approximation.
        } We present the classical view of function approximation by neural networks, and give two instances of so-called universal approximation results. Such statements describe the ability of neural networks to approximate every function of a given class to arbitrary accuracy, given that the network size is sufficiently large. The first result, which holds under very broad assumptions on the activation function, is on uniform approximation of continuous functions on compact domains. The second result shows that for a very specific activation function, the network size can be chosen independently of the desired accuracy, highlighting that universal approximation needs to be interpreted with caution.
	
	{\bf Chapter \ref{chap:Splines}: Splines.} Going beyond universal approximation, this chapter starts to explore approximation rates of neural networks. Specifically, we examine how well certain functions can be approximated relative to the number of parameters in the network.
        For so-called sigmoidal activation functions, we establish a link between neural-network- and spline-approximation.
        This reveals that smoother functions require fewer network parameters.
        However, achieving this increased efficiency necessitates the use of deeper neural networks.
        This observation offers a first glimpse into the \emph{importance of depth in deep learning}.
	
      {\bf Chapter \ref{chap:ReLUNNs}: ReLU neural networks.}
      This chapter focuses on one of the most popular activation functions in practice---the ReLU.
      We prove that the class of ReLU networks is equal to the set of continuous piecewise linear functions, thus providing a theoretical foundation for their expressive power. Furthermore, given a continuous piecewise linear function, we investigate the necessary width and depth of a ReLU network to represent it. Finally, we leverage approximation theory for piecewise linear functions to derive convergence rates for approximating H\"older continuous functions.
 
	{\bf Chapter \ref{chap:AffPieces}: Affine pieces for ReLU neural networks.} Having gained some intuition about ReLU neural networks, in this chapter, we address some potential limitations.
	We analyze ReLU neural networks by counting the number of affine regions that they generate. The key insight of this chapter is that deep neural networks can generate exponentially more regions than shallow ones. This observation provides \emph{further evidence for the potential advantages of depth} in neural network architectures.
        
	{\bf Chapter \ref{chap:DReLUNN}: Deep ReLU neural networks.}
	Having identified the ability of deep ReLU neural networks to generate a large number of affine regions, we investigate whether this translates into an actual advantage in function approximation. Indeed, for approximating smooth functions, we prove substantially better approximation rates than we obtained for shallow neural networks. This adds again to our
 \emph{understanding of depth and its connections to expressive power} of neural network architectures.
 
 {\bf Chapter \ref{chap:hdapp}: High-dimensional approximation.} The
 convergence rates established in the previous chapters deteriorate
 significantly in high-dimensional settings. This chapter examines
 three scenarios under which neural networks can provably
 \emph{overcome the curse of dimensionality}.
 
 {\bf Chapter \ref{chap:Interpolation}: Interpolation.}
 In this chapter we shift our perspective from approximation to exact interpolation of the training data. We analyze conditions under which exact interpolation is possible, and discuss the implications for empirical risk minimization. Furthermore, we present a constructive proof showing that ReLU networks can express an optimal interpolant of the data (in a specific sense).
        
 {\bf Chapter \ref{chap:training}: Training of neural networks.}
 We start to examine the training process of deep learning. First, we study the fundamentals of (stochastic) gradient descent and convex optimization.
Additionally, we examine accelerated methods and highlight the key principles behind popular training algorithms such as Adam.
Finally, we discuss how the backpropagation algorithm can be used to implement these optimization algorithms for training neural networks. 
        
{\bf Chapter \ref{chap:wideNets}: Wide neural networks and the neural tangent kernel.}
This chapter introduces the neural tangent kernel as a tool for analyzing the training behavior of neural networks. We begin by revisiting linear and kernel regression for the approximation of functions based on data. Additionally we discuss the effect of adding a regularization term to the objective function. Afterwards, we show for certain architectures of sufficient width, that the training dynamics of gradient descent resemble those of kernel regression and converge to a global minimum.
This analysis provides insights into why, under certain conditions, we can train neural networks \emph{without getting stuck in (bad) local minima}, despite the non-convexity of the objective function. Finally, we discuss a well-known link between neural networks and Gaussian processes, giving some indication why overparameterized networks \emph{do not necessarily overfit} in practice.
 
{\bf Chapter \ref{chap:LossLandscapes}: Loss landscape analysis.} In this chapter, we present an alternative view on the optimization problem, by analyzing the loss landscape---the empirical risk as a function of the neural network parameters.
We give theoretical arguments showing that increasing overparameterization leads to greater connectivity between the valleys and basins of the loss landscape.
        Consequently, overparameterized architectures make it easier to reach a region
        where all minima are global minima. 
        Additionally, we observe that most stationary points associated with non-global minima
        are saddle points.
        This sheds further light on the empirically observed fact that deep architectures can often be optimized \emph{without getting stuck in non-global minima}.

	{\bf Chapter \ref{chap:shape}: Shape of neural network spaces.}
        While Chapters \ref{chap:wideNets} and \ref{chap:LossLandscapes} highlight
        potential reasons for the success of neural network training,
        in this chapter, we show that the set of neural networks of a fixed architecture has some undesirable properties from an optimization perspective. Specifically, we show that this set is typically non-convex.
        Moreover, in general it does not possess
        the best-approximation property, meaning that there might not exist
        a neural network within the set yielding the best approximation
        for a given function.
	
	{\bf Chapter \ref{chap:VC} : Generalization properties of deep neural networks.
 }
	To understand
        why deep neural networks successfully
        generalize to unseen data points (outside of the training set), we
        study classical statistical learning theory, with a focus on
        neural network functions as the hypothesis class.
        We then show how to establish generalization bounds for deep learning,
        providing theoretical insights into the \emph{performance on unseen data}.
        
	{\bf Chapter \ref{chap:GenOverparameterized}: Generalization in the overparameterized regime.
 }
 The generalization bounds of the previous chapter are not meaningful when the number of parameters of a neural network surpasses the number of training samples.
 However, this overparameterized regime is where many successful network architectures operate.
 To gain a deeper understanding of generalization in this regime,
  we describe the phenomenon of double descent
  and present a potential explanation. This addresses the question of why deep neural networks \emph{perform well despite being highly overparameterized}.
 
	{\bf Chapter \ref{chap:adversarial}: Robustness and adversarial examples.
 }
We explore the existence of adversarial examples---inputs designed to deceive neural networks.  We provide some \emph{theoretical explanations of why adversarial examples arise}, and discuss potential strategies to prevent them.

{\bf Chapter \ref{chap:ModArchitectures}: Modern architectures.
}
In the final chapter, we present some of the modifications to the neural network architectures that have been most successful in practice. We discuss so-called residual connections, convolutional neural networks, and finally transformers.

\section{Material not covered in this book}

This %
book studies some central topics of deep learning but leaves out even more.
 Interesting questions associated with the field that were omitted, as well as some pointers to related works are listed below:
 
{\bf Advanced architectures:}
The (deep) feedforward neural network is far from the only type of neural network.
 In practice, architectures %
 must be adapted to the type of data.
 We will discuss three advanced types of architectures in Chapter \ref{chap:ModArchitectures}: Residual neural networks, convolutional neural networks, and transformers.  
 However, in practice there are many more options. Notably, we omit all discussion of graph neural networks \cite{bronstein2021geometric}, which are a natural choice for graph-based data.
 Moreover, for sequence-based inputs, we only discuss transformers, but leave out very established alternatives such as Long Short-Term Memory (LSTM) networks \cite{hochreiter1997long}.	 

{\bf Unsupervised %
  and
  Reinforcement Learning:}
While this book focuses on supervised learning, where each data point $x_i$ has a label $y_i$,
  there is a vast field of machine learning called unsupervised learning, where labels are absent.
  Classical %
  unsupervised learning %
  problems include clustering and dimensionality reduction \cite[Chapters 22/23]{understanding}.

A popular area in deep learning, where no labels are used, is physics-informed neural networks \cite{raissi2019physics}.
 Here, a neural network is %
 trained to satisfy a partial differential equation (PDE), %
 with the loss function quantifying the deviation from this PDE.
	
	Finally, reinforcement learning is a technique where an agent can interact with an environment and %
        receives feedback based on its actions.
        The actions are %
        guided by a so-called policy, which is to be learned, \cite[Chapter 17]{mohri2018foundations}.
 In deep reinforcement learning, this policy is modeled by a deep neural network.
 Reinforcement learning is the basis of the aforementioned AlphaGo.

{\bf Interpretability/Explainability and Fairness:}
The use of deep neural networks in critical decision-making processes, such as allocating scarce resources (e.g., organ transplants in medicine, financial credit approval, hiring decisions) or engineering (e.g., optimizing bridge structures, autonomous vehicle navigation, predictive maintenance), necessitates an understanding of their decision-making process. This is crucial for both practical and ethical reasons.

Practically, understanding how a model arrives at a decision can help us improve its performance and mitigate problems. It allows us to ensure that the model performs according to our intentions and does not produce undesirable outcomes. For example, in bridge design, understanding why a model suggests or rejects a particular configuration can help engineers identify potential vulnerabilities, ultimately leading to safer and more efficient designs. Ethically, transparent decision-making is crucial, especially when the outcomes have significant consequences for individuals or society; biases present in the data or model design can lead to discriminatory outcomes, making explainability essential.

However, explaining the predictions of deep neural networks is not straightforward. Despite knowledge of the network weights and biases, the repeated and complex interplay of linear transformations and non-linear activation functions often renders these models black boxes. A comprehensive overview of various techniques for interpretability, not only for deep neural networks, can be found in \cite{molnar2020interpretable}. Regarding the topic of fairness, we refer for instance to \cite{9113719,barocas-hardt-narayanan}.

	{\bf Implementation:} %
          While this %
          book
          focuses on provable theoretical results,
          the field of deep learning is strongly driven by applications,
            and a thorough understanding of deep learning cannot be achieved without practical experience.
 For this, there exist numerous resources with excellent %
 explanations. %
We recommend \cite{geron2017hands-on,chollet2021deep, prince2023understanding} as well as the countless online tutorials that are just a Google (or alternative) search away.
	
	{\bf Many more:} The field is evolving rapidly, and new ideas are constantly being generated and tested.
 This book cannot give a complete overview.
 However, we hope that it %
 provides the reader with %
 a solid foundation in the
 fundamental knowledge and principles %
 to quickly grasp and understand new developments in the field.

\section*{Bibliography and further reading}
Throughout this book, we will end each chapter with a short overview of related work and the references used in the chapter. 

In this introductory chapter, we %
highlight %
several other recent textbooks and works
on deep learning. %
For a historical survey on neural networks see \cite{SCHMIDHUBER201585} and also \cite{LecBen15}. For general textbooks on neural networks and deep learning, we refer to
\cite{haykin2009neural, GoodBengCour16, prince2023understanding} for more recent monographs. More mathematical introductions to the topic are given, for example, in \cite{MR1741038, jentzen2023mathematical, calin2020deep}. For the implementation of neural networks we refer for example to \cite{geron2017hands-on,chollet2021deep}.

%% file: FeedForwardNNs.tex
\chapter{Feedforward neural networks}\label{chap:FFNNs}
Feedforward neural networks, %
henceforth simply referred to as
neural networks (NNs), %
constitute the central object of study of this book. %
In %
this chapter, we provide a formal definition of neural networks, %
discuss the %
\emph{size} of a neural network, and give a brief overview of %
common activation functions.

\section{Formal definition}
We previously defined a single neuron $\nu$ in \eqref{eq:neuron} and
Figure \ref{fig:Neuron}. A neural network is constructed by connecting
multiple neurons. Let us now make precise this connection procedure.

\begin{definition}\label{def:nn}
	Let $L\in \N$, $d_0,\dots,d_{L+1}\in\N$, and
        let $\sigma\colon \R \to \R$.
        A function $\Phi \colon \R^{d_0}\to\R^{d_{L+1}}$ is called
        a {\bf neural network} if there exist matrices $\BW^{(\ell)}\in\R^{d_{\ell+1}\times d_\ell}$ and
	vectors $\Bb^{(\ell)}\in\R^{d_{\ell+1}}$, $\ell = 0, \dots, L$,
        such that with %
        \begin{subequations}\label{eq:xell}
	\begin{align}
          \Bx^{(0)} &\dfn \Bx\label{eq:inputlayer}\\
          \Bx^{(\ell)}&\dfn \sigma(\BW^{(\ell-1)}\Bx^{(\ell-1)}+\Bb^{(\ell-1)})&&\text{for }
          \ell=1, \dots, L\label{eq:defxell}\\
          \Bx^{(L+1)}&\dfn \BW^{(L)}\Bx^{(L)}+\Bb^{(L)}\label{eq:defxLplus1}
	\end{align}
      \end{subequations}
        holds
	\begin{align*}
		\Phi(\Bx) = \Bx^{(L+1)}\qquad \text{for all }\Bx\in\R^{d_0}.
	\end{align*}
        
        We call
        $L$ the \textbf{depth}, $d_{\rm max} =\max_{\ell= 1, \dots, L} d_\ell$ the
      \textbf{width}, $\sigma$ the \textbf{activation function}, and
      $(\sigma;d_0,\dots,d_{L+1})$ the \textbf{architecture} of the
      neural network $\Phi$. Moreover,
      $\BW^{(\ell)}\in\R^{d_{\ell+1}\times d_\ell}$ are the
      \textbf{weight matrices} and $\Bb^{(\ell)}\in\R^{d_{\ell+1}}$
      the \textbf{bias vectors} of $\Phi$ for $\ell = 0, \dots L$.

      Collecting all weights and biases in a single vector
      \begin{equation}\label{eq:parameters}
  \Bw=((\BW^{(0)},\Bb^{(0)}),\dots,(\BW^{(L)},\Bb^{(L)})),
\end{equation}
of suitable size, we also write
      \begin{equation}\label{eq:PhiBxBw}
        \Phi(\Bx,\Bw),
      \end{equation}
      if we wish to emphasize the dependence on $\Bw$.
    \end{definition}

  \begin{remark}
    Typically, there exist different choices of architectures, weights,
    and biases yielding the same function
    $\Phi:\R^{d_0}\to\R^{d_{L+1}}$. For this reason we cannot
    associate a unique meaning to these notions solely based on the
    \emph{function} realized by $\Phi$.

    In the following, when we refer to a neural network $\Phi$, we either mean the function it realizes or a specific realization as defined in Definition~\ref{def:nn}. In particular, when discussing properties such as depth or size, it is always understood that there exists a fixed construction of $\Phi$ that satisfies these properties.
  \end{remark}

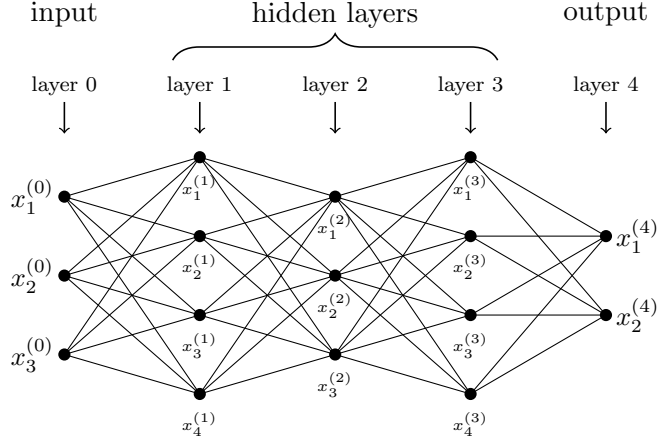
\begin{figure}
  \centering
  \input{plots/nn.tex}
  \caption{Sketch of a neural network with three hidden layers,
    and $d_0 = 3$, $d_1= 4$, $d_2=3$, $d_3 = 4$, $d_4 = 2$. The neural network has depth three and width four.}
	\label{fig:FullNN}
\end{figure}

  The architecture of a neural network is often depicted as a
    connected graph, %
    as illustrated in Figure \ref{fig:FullNN}.  The {\bf nodes} in
    such graphs represent (the output of) the neurons. They are
    arranged in {\bf layers}, with $\Bx^{(\ell)}$ in Definition
    \ref{def:nn} corresponding to the neurons in layer $\ell$. We also
    refer to $\Bx^{(0)}$ in \eqref{eq:inputlayer} as the {\bf input
      layer} and to $\Bx^{(L+1)}$ in \eqref{eq:defxLplus1} as the {\bf
      output layer}.  All layers in between are referred to as the
    {\bf hidden layers} and their output is given by
    \eqref{eq:defxell}. The number of hidden layers corresponds to the
    depth. For the correct interpretation of such graphs, we note that
    by our conventions in Definition \ref{def:nn}, the activation
    function is applied after each affine transformation, except in
    the final layer.

Neural networks of depth one are called {\bf shallow}, if the depth is larger than one
they are called {\bf deep}. The notion of deep neural networks is not
used entirely consistently in the literature, and some authors use the
word deep only in case the depth is much larger than one, where the
precise meaning of ``much larger'' depends on the application.

Throughout, we only consider neural networks in the sense of Definition \ref{def:nn}.
We
emphasize however, that this is %
just one (simple but very common) type
of neural network.
Many adjustments to this construction are possible and
also widely used. For example:
\begin{itemize}
	\item We may use {\bf different activation functions} $\sigma_\ell$ in
	each layer $\ell$ or we may even use a different activation
	function for each node.
      \item {\bf Residual} neural networks allow ``skip connections'' \cite{he2016deep}.
	This means that
	information is allowed to skip layers in the sense that the nodes
	in layer $\ell$ may have $\Bx^{(0)},\dots,\Bx^{(\ell-1)}$ as their
	input (and not just $\Bx^{(\ell-1)}$), cf.~\eqref{eq:xell}.
	\item In contrast to feedforward neural networks, {\bf recurrent} neural networks allow
	information to flow backward, in the sense that
	$\Bx^{(\ell-1)},\dots,\Bx^{(L+1)}$ may serve as input for the nodes
	in layer $\ell$ (and not just $\Bx^{(\ell-1)}$).
	This creates loops
	in the flow of information, and one has to introduce a time index
	$t\in\N$, as the output of a node in time step $t$ might be
	different from the output in time step $t+1$.
\end{itemize}

Let us clarify some further common terminology used in the context of neural networks:
    \begin{itemize}
    \item {\bf parameters}: The parameters of a neural network refer
      to the set of all entries of the weight matrices and bias
      vectors.  For notational convenience, they are often collected
      in a single vector $\Bw$ as in \eqref{eq:parameters}.
These parameters are adjustable and are learned during the training process,
determining the specific function realized by the network.

\item {\bf hyperparameters}: Hyperparameters are settings that define the network's
  architecture (and training process), but are (typically) not directly learned during
  training. Examples include the depth, the number of neurons in each layer,
  and the choice of activation function. They are %
  set before training begins.
\item {\bf weights}: The term ``weights'' is often used broadly to refer to \emph{all}
  parameters of a neural network, including both the weight matrices
  and bias vectors.
\item {\bf model}: For a fixed architecture,
  every choice of network parameters
  $\Bw$ in \eqref{eq:parameters}
  defines a specific function $\Bx\mapsto \Phi(\Bx,\Bw)$. In deep
  learning this
 function is often referred to as a model. More generally,
 ``model'' can be used to describe any function
 parameterization by a set of parameters $\Bw\in\R^n$, $n \in \N$.  
\end{itemize}

\subsection{Basic operations on neural networks}

  There are various ways how neural networks can be combined with one
  another. The next proposition addresses this for linear
  combinations, compositions, and parallelization. The formal proof, which is a good
  exercise to familiarize oneself with neural networks, is left as
  Exercise \ref{ex:VectorSpaceAndCompositions}.

\begin{proposition}\label{prop:VectorSpaceAndCompositions}
		For two neural networks $\Phi_1$, $\Phi_2$, with architectures 
		$$
		(\sigma;d_0^1, d_1^1, \dots, d_{L_1+1}^1)\quad \text{ and }\quad
		(\sigma;d_0^2, d_1^2, \dots, d_{L_2+1}^2)
		$$
		respectively, it holds that
		\begin{enumerate}
                \item for all $\alpha \in \R$ %
                  exists a neural network
                  $\Phi_\alpha$ %
                  with
                          architecture $(\sigma;d_0^1, d_1^1, \dots, d_{L_1+1}^1)$ such that
			\[
			\Phi_\alpha(\Bx) = \alpha \Phi_1(\Bx) \qquad  \text{ for all } \Bx \in \R^{d_0^1},
			\]
			\item\label{item:parallelPhi} if $d_0^1 = d_0^2 \eqqcolon d_0$ and $L_1 = L_2 \eqqcolon L$, then there exists a neural network $\Phi_{\rm parallel}$ with architecture $(\sigma;d_0, d_1^1 + d_1^2, \dots, d_{L+1}^1 + d_{L+1}^2)$ such that 
			\[ 
			\Phi_{\rm parallel}(\Bx) = (\Phi_1(\Bx), \Phi_2(\Bx)) \qquad  \text{ for all } \Bx \in \R^{d_0},
			\]
			\item if $d_0^1 = d_0^2 \eqqcolon d_0$,  $L_1 = L_2 \eqqcolon L$, and $d_{L+1}^1 = d_{L+1}^2 \eqqcolon d_{L+1}$, then there exists a %
                          neural network $\Phi_{\rm sum}$ with architecture $(\sigma;d_0, d_1^1 + d_1^2, \dots, d_{L}^1 + d_{L}^2,  d_{L+1})$ such that 
			\[ 
			\Phi_{\rm sum}(\Bx) = \Phi_1(\Bx) +  \Phi_2(\Bx) \qquad \text{ for all } \Bx \in \R^{d_0},
			\]
			\item\label{item:compositionPhi} if $d_{L_1+1}^1 = d_0^2$, then there exists a %
                          neural network $\Phi_{\rm comp}$ with architecture $(\sigma; d_0^1, d_1^1, \dots, d_{L_1}^1, d_1^2, \dots, d_{L_2+1}^2)$ such that 
			\[ 
			\Phi_{\rm comp}(\Bx) = \Phi_2 \circ \Phi_1(\Bx) \qquad  \text{ for all } \Bx \in \R^{d_0^1}.
			\]
		\end{enumerate}
	\end{proposition}

\section{Notion of size}

Neural networks provide a framework to parametrize
functions. Ultimately, our goal is to find a neural network %
that fits some underlying input-output relation. As mentioned above,
the architecture (depth, width and activation function) is
typically chosen a priori and considered fixed. During training of
the neural network, %
its parameters (weights and biases) are suitably adapted by some %
algorithm. %
Depending on the application, on top of the stated architecture
choices, further restrictions on the weights and biases can be
desirable. %
For example, the following two appear frequently:

\begin{itemize}
\item {\bf weight sharing}: This is a technique where specific entries
  of the weight matrices (or bias vectors) are constrained to be equal.
  Formally, this means imposing conditions of the form
	$W_{k,l}^{(i)}=W^{(j)}_{s,t}$, i.e.\ %
	the entry $(k,l)$ of the $i$th weight matrix is equal to the entry
	at position $(s,t)$ of weight matrix $j$.
	We denote this assumption
	by $(i,k,l)\sim (j,s,t)$, paying tribute to the trivial fact that
	``$\sim$'' is an equivalence relation.
        During training, shared weights are updated jointly,
        meaning that any change to one weight
        is simultaneously applied to all other weights of this class.
        Weight sharing can also be applied to the entries of bias vectors.
      \item {\bf sparsity}: This %
        refers to imposing a sparsity structure on the weight matrices (or bias vectors).
        Specifically, we a priori set
	$W_{k,l}^{(i)}=0$ for certain $(k,l,i)$, i.e.\ we impose
	entry $(k,l)$ of the $i$th weight matrix to be $0$.
        These zero-valued entries are considered fixed,
          and are not adjusted during training.
	The
	condition $W_{k,l}^{(i)}=0$ corresponds to node $l$ of layer $i-1$
	\emph{not} serving as an input to node $k$ in layer $i$.
	If we represent the neural network as a graph, this is indicated by not
        connecting the corresponding nodes.
        Sparsity can also be imposed on the bias vectors.
        Mathematically, this can be described by introducing a sparsity vector $\Bs$ consisting of zeros and ones, and then considering the network $\Phi(\Bx,\Bw\odot\Bs)$, where $\odot$ denotes elementwise multiplication.
\end{itemize}
Both of these restrictions decrease the number of learnable parameters
in the neural network.
The number of parameters can be seen as a measure of
the complexity of the represented function class.
For this reason, we
introduce ${\rm size}(\Phi)$ as a notion for the number of learnable
parameters.
Formally (with $|S|$ denoting the cardinality of a set $S$):

\begin{definition}\label{def:SizeOfNNDefinition}
	Let $\Phi$ be as in Definition~\ref{def:nn}.
	Then the {\bf size} of $\Phi$ is
	\begin{align}\label{eq:size}
		\size(\Phi)\dfn \left|\left(\set{(i,k,l)}{W_{k,l}^{(i)}\neq 0}\cup
		\set{(i,k)}{b_{k}^{(i)}\neq 0}
		\right)\big/\sim\right|.
	\end{align}
      \end{definition}

\section{Activation functions}\label{sec:activationFunctions}

Activation functions are a crucial part of neural networks, as they introduce nonlinearity into the model. If an affine %
activation function were used, the resulting neural network function would also be affine and hence very restricted in what it can represent.

The choice of activation function can have a significant impact on the performance, but there does not seem to be a universally optimal one. %
We next discuss %
a few important activation functions and highlight some common issues associated with them.

\begin{figure}[htb]
  \centering
  \subfloat[Sigmoid]{\includegraphics[width=0.32\textwidth]{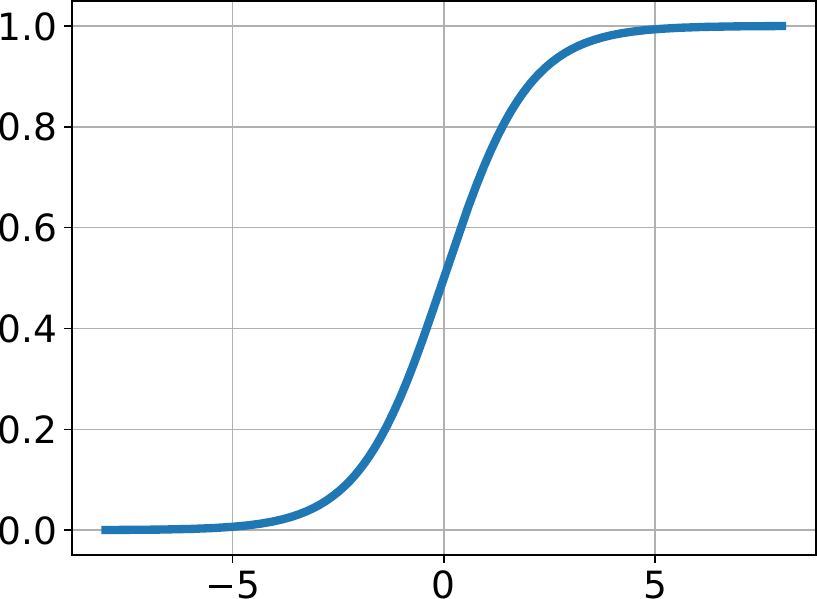}}
  \hfill
  \subfloat[ReLU and SiLU]{\includegraphics[width=0.32\textwidth]{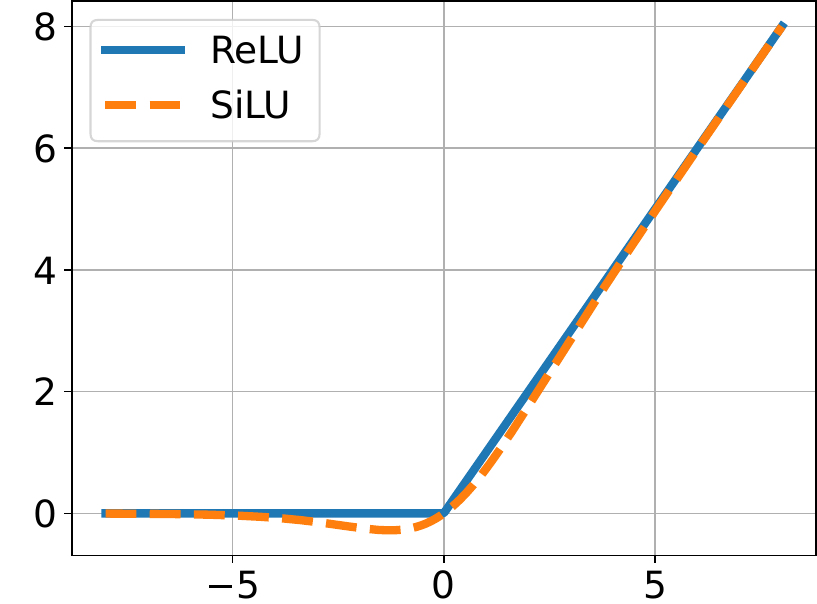}}
  \hfill
  \subfloat[Leaky ReLU]{\includegraphics[width=0.32\textwidth]{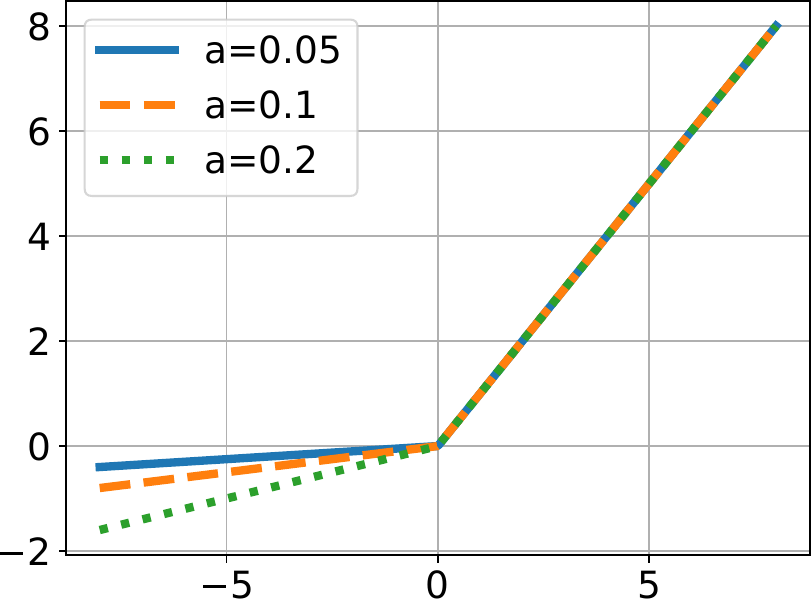}}  
  \caption{Different activation functions.
  }
	\label{fig:activationFunctions}
      \end{figure}

{\bf Sigmoid:}
The sigmoid activation function is given by
\begin{equation*}
  \sigma_{\rm sig}(x) = \frac{1}{1+e^{-x}}\qquad\text{for }x \in \R,
\end{equation*}
and depicted in Figure \ref{fig:activationFunctions} (a).
Its output %
ranges
between zero and one, making it  %
interpretable as a probability.
	The sigmoid is a smooth function, which allows the application of gradient-based training.

	It has the disadvantage that its derivative becomes very small if $|x| \to \infty$.
	This can affect learning due to the so-called vanishing gradient problem.
	Consider
        the simple neural network $\Phi_n(x)  = \sigma \circ \dots \circ \sigma(x + b)$
        defined with 
        $n \in \N$ compositions of $\sigma$, and where $b\in\R$ is a bias.
        Its derivative with respect to %
        $b$ is
	\[
	\frac{\dd}{\dd b} \Phi_n(x) = \sigma'(\Phi_{n-1}(x)) \frac{\dd}{\dd b} \Phi_{n-1}(x).
	\]
        If $\sup_{x\in\R}|\sigma'(x)|\le 1-\delta$, then by induction,
        $|\frac{\dd}{\dd b} \Phi_n(x)|\le (1-\delta)^n$. %
        The opposite effect happens for activation functions with derivatives uniformly
        larger than one.
        This argument shows that
        the derivative of $\Phi_n(x,b)$ with respect to $b$ can become
        exponentially small or exponentially large when propagated
          through the
          layers. %
	This effect, %
        known as the \emph{vanishing- or exploding gradient effect}, also occurs
        for activation functions which do not admit the uniform bounds assumed above.
	However, since the sigmoid activation function exhibits areas with extremely small gradients,
        the vanishing gradient effect can be strongly exacerbated.

        {\bf ReLU (Rectified Linear Unit):} 
        The ReLU is defined as
        \begin{equation*}
          \sigma_{\rm ReLU}(x) = \max\{x, 0\}\qquad\text{for }x \in \R,
        \end{equation*}
        and depicted in Figure \ref{fig:activationFunctions} (b).
	It is piecewise linear, and due to its simplicity
        its evaluation is computationally %
        very efficient.
        It is one of the most popular activation functions in practice.
        Since its derivative is always zero or one, it does not suffer from the vanishing gradient problem to the same extent as the sigmoid function. 
	However, %
        ReLU can suffer from the so-called \emph{dead neurons} problem.
	Consider %
        the neural network 
	\[
          \Phi(x) = \sigma_{\rm ReLU}( b - \sigma_{\rm ReLU}(x))\qquad\text{for }x\in\R
	\]
        depending on the bias $b\in\R$.
	If $b < 0$, then $\Phi(x) = 0$ for all $x \in \R$.
	The neuron %
        corresponding to the
        second application of $\sigma_{\rm ReLU}$ thus %
        produces a constant signal.
	Moreover, if $b < 0$, %
        $\frac{\dd}{\dd b}\Phi(x)=0$ for all $x\in\R$.
	As a result, every negative value of $b$ %
        yields a stationary point of
        the empirical risk.
        A gradient-based method will not %
        be able to further train the parameter $b$.
        We thus refer to %
        this neuron as a dead neuron.

        {\bf SiLU (Sigmoid Linear Unit):}
	An important difference between the ReLU and the Sigmoid is that the ReLU is
        not differentiable at $0$. The SiLU activation function (also referred to as ``swish'')
        can be interpreted as a smooth approximation to the ReLU. It is defined as
        \begin{equation*}
          \sigma_{\rm SiLU}(x)\dfn x\sigma_{\rm sig}(x) =
          \frac{x}{1+e^{-x}}\qquad\text{for }x \in \R,
        \end{equation*}
        and is depicted in Figure \ref{fig:activationFunctions} (b).
        There exist various other smooth activation functions that mimic the ReLU,
        including the Softplus $x\mapsto \log(1+\exp(x))$, the GELU (Gaussian Error Linear Unit) $x\mapsto xF(x)$ where $F(x)$ denotes
        the cumulative distribution function of the standard normal distribution,
        and the Mish $x\mapsto x\tanh(\log(1+\exp(x)))$.

        {\bf Parametric ReLU or Leaky ReLU:}          
        This variant of the ReLU addresses the dead neuron problem.
        For some $a\in (0,1)$, the parametric ReLU is defined as
          \begin{equation*}
            \sigma_{a}(x) = \max\{x, ax\}\qquad\text{for }x\in\R,
          \end{equation*}
          and is depicted in Figure \ref{fig:activationFunctions} (c) for three different
          values of $a$.
	Since the output of $\sigma$ does not have flat regions like the ReLU, the dying ReLU problem is mitigated.
	If $a$ is not chosen too small, then there is less of a vanishing gradient problem than for the Sigmoid. %
        In practice, the
        additional parameter $a$ has to be fine-tuned depending on the application.
	Like the ReLU, the parametric ReLU is not differentiable at $0$. 

\section*{Bibliography and further reading}
The concept of neural networks was first introduced by McCulloch
  and Pitts in \cite{mcculloch1943logical}. Later Rosenblatt
  \cite{rosenblatt1957perceptron} introduced the perceptron, an
  artificial neuron with adjustable weights that forms the basis of
  the multilayer perceptron (a fully connected feedforward neural
  network).  The vanishing gradient problem shortly addressed in
  Section \ref{sec:activationFunctions} was discussed by Hochreiter in
  his diploma thesis \cite{Hochreiter:91} and later in
  \cite{bengio:ieeenn94,hochreiter1997long}.

\newpage
\section*{Exercises}

\begin{exercise}\label{ex:VectorSpaceAndCompositions}
  Prove Proposition \ref{prop:VectorSpaceAndCompositions}. %
\end{exercise}

\begin{exercise}
  In this exercise, we show that ReLU and parametric ReLU create
  similar sets of neural network functions.
  Fix $a>0$.
	
	\begin{enumerate}
		\item Find a set of weight matrices and bias vectors, such that the associated neural network $\Phi_1$, with the ReLU activation function $\sigma_{\rm ReLU}$ satisfies $\Phi_1(x) = \sigma_{a}(x)$ for all $x \in \R$.
		\item Find a set of weight matrices and bias vectors, such that the associated neural network $\Phi_2$, with the parametric ReLU activation function $\sigma_a$ satisfies $\Phi_2(x) = \sigma_{\rm ReLU}(x)$ for all $x \in \R$.
                \item Conclude that every ReLU neural network can be
                    expressed as a leaky ReLU neural network and vice versa.
	\end{enumerate}
\end{exercise}

\begin{exercise}
	Let $d\in \N$, and let $\Phi_1$ be a neural network with the ReLU as activation function, input dimension $d$, and output dimension $1$.
	Moreover, let $\Phi_2$ be a neural network with the sigmoid activation function, input dimension $d$, and output dimension $1$.
	Show that, if $\Phi_1 = \Phi_2$, then $\Phi_1$ is a constant function.
\end{exercise}

\begin{exercise}
  In this exercise, we show that for the sigmoid activation functions, dead-neuron-like behavior is very rare.
        Let $\Phi$ be a neural network with the sigmoid activation function. %
	Assume that $\Phi$ is a constant function.
	Show that for every $\eps >0$ there is a non-constant neural network $\widetilde{\Phi}$ with the same architecture as $\Phi$ such that for all $\ell = 0, \dots L$,
	\begin{align*}
		\norm{\BW^{(\ell)} - \widetilde{\BW}^{(\ell)}} \leq \eps \text{ and } \norm{ \Bb^{(\ell)} - \widetilde{\Bb}^{(\ell)}} \leq \eps
	\end{align*}
	where $\BW^{(\ell)}$, $\Bb^{(\ell)}$ are the weights and biases of $\Phi$ and $\widetilde{\BW}^{(\ell)}$, $\widetilde{\Bb}^{(\ell)}$ are the biases of $\widetilde{\Phi}$.

        Show that such a statement does not hold for ReLU neural networks. What about leaky ReLU?
      \end{exercise}

%% file: plots/nn.tex
\begin{tikzpicture}[scale=1.0]
\draw [black,] (0,1.5750000000000002) -- (1.8,2.1);
\draw [black,] (0,1.5750000000000002) -- (1.8,1.05);
\draw [black,] (0,1.5750000000000002) -- (1.8,0.0);
\draw [black,] (0,1.5750000000000002) -- (1.8,-1.05);
\draw [black,] (0,0.525) -- (1.8,2.1);
\draw [black,] (0,0.525) -- (1.8,1.05);
\draw [black,] (0,0.525) -- (1.8,0.0);
\draw [black,] (0,0.525) -- (1.8,-1.05);
\draw [black,] (0,-0.525) -- (1.8,2.1);
\draw [black,] (0,-0.525) -- (1.8,1.05);
\draw [black,] (0,-0.525) -- (1.8,0.0);
\draw [black,] (0,-0.525) -- (1.8,-1.05);
\draw [black,] (1.8,2.1) -- (3.6,1.5750000000000002);
\draw [black,] (1.8,2.1) -- (3.6,0.525);
\draw [black,] (1.8,2.1) -- (3.6,-0.525);
\draw [black,] (1.8,1.05) -- (3.6,1.5750000000000002);
\draw [black,] (1.8,1.05) -- (3.6,0.525);
\draw [black,] (1.8,1.05) -- (3.6,-0.525);
\draw [black,] (1.8,0.0) -- (3.6,1.5750000000000002);
\draw [black,] (1.8,0.0) -- (3.6,0.525);
\draw [black,] (1.8,0.0) -- (3.6,-0.525);
\draw [black,] (1.8,-1.05) -- (3.6,1.5750000000000002);
\draw [black,] (1.8,-1.05) -- (3.6,0.525);
\draw [black,] (1.8,-1.05) -- (3.6,-0.525);
\draw [black,] (3.6,1.5750000000000002) -- (5.4,2.1);
\draw [black,] (3.6,1.5750000000000002) -- (5.4,1.05);
\draw [black,] (3.6,1.5750000000000002) -- (5.4,0.0);
\draw [black,] (3.6,1.5750000000000002) -- (5.4,-1.05);
\draw [black,] (3.6,0.525) -- (5.4,2.1);
\draw [black,] (3.6,0.525) -- (5.4,1.05);
\draw [black,] (3.6,0.525) -- (5.4,0.0);
\draw [black,] (3.6,0.525) -- (5.4,-1.05);
\draw [black,] (3.6,-0.525) -- (5.4,2.1);
\draw [black,] (3.6,-0.525) -- (5.4,1.05);
\draw [black,] (3.6,-0.525) -- (5.4,0.0);
\draw [black,] (3.6,-0.525) -- (5.4,-1.05);
\draw [black,] (5.4,2.1) -- (7.2,1.05);
\draw [black,] (5.4,2.1) -- (7.2,0.0);
\draw [black,] (5.4,1.05) -- (7.2,1.05);
\draw [black,] (5.4,1.05) -- (7.2,0.0);
\draw [black,] (5.4,0.0) -- (7.2,1.05);
\draw [black,] (5.4,0.0) -- (7.2,0.0);
\draw [black,] (5.4,-1.05) -- (7.2,1.05);
\draw [black,] (5.4,-1.05) -- (7.2,0.0);
\fill [black] (0,1.5750000000000002) circle (2.3pt);
\fill [black] (0,0.525) circle (2.3pt);
\fill [black] (0,-0.525) circle (2.3pt);
\fill [black] (1.8,2.1) circle (2.3pt);
\fill [black] (1.8,1.05) circle (2.3pt);
\fill [black] (1.8,0.0) circle (2.3pt);
\fill [black] (1.8,-1.05) circle (2.3pt);
\fill [black] (3.6,1.5750000000000002) circle (2.3pt);
\fill [black] (3.6,0.525) circle (2.3pt);
\fill [black] (3.6,-0.525) circle (2.3pt);
\fill [black] (5.4,2.1) circle (2.3pt);
\fill [black] (5.4,1.05) circle (2.3pt);
\fill [black] (5.4,0.0) circle (2.3pt);
\fill [black] (5.4,-1.05) circle (2.3pt);
\fill [black] (7.2,1.05) circle (2.3pt);
\fill [black] (7.2,0.0) circle (2.3pt);
\node at (0,1.5750000000000002) [left] {\small $x^{(0)}_1$};
\node at (0,0.525) [left] {\small $x^{(0)}_2$};
\node at (0,-0.525) [left] {\small $x^{(0)}_3$};
\node at (1.8,2.1) [yshift=-0.4cm] {\tiny $x^{(1)}_1$};
\node at (1.8,1.05) [yshift=-0.4cm] {\tiny $x^{(1)}_2$};
\node at (1.8,0.0) [yshift=-0.4cm] {\tiny $x^{(1)}_3$};
\node at (1.8,-1.05) [yshift=-0.4cm] {\tiny $x^{(1)}_4$};
\node at (3.6,1.5750000000000002) [yshift=-0.4cm] {\tiny $x^{(2)}_1$};
\node at (3.6,0.525) [yshift=-0.4cm] {\tiny $x^{(2)}_2$};
\node at (3.6,-0.525) [yshift=-0.4cm] {\tiny $x^{(2)}_3$};
\node at (5.4,2.1) [yshift=-0.4cm] {\tiny $x^{(3)}_1$};
\node at (5.4,1.05) [yshift=-0.4cm] {\tiny $x^{(3)}_2$};
\node at (5.4,0.0) [yshift=-0.4cm] {\tiny $x^{(3)}_3$};
\node at (5.4,-1.05) [yshift=-0.4cm] {\tiny $x^{(3)}_4$};
\node at (7.2,1.05) [right] {\small $x_{1}^{(4)}$};
\node at (7.2,0.0) [right] {\small $x_{2}^{(4)}$};
\node at (0.0,3.045) {\scriptsize layer 0};
\draw [semithick,->] (0.0,2.835) -- (0.0,2.415);
\node at (1.8,3.045) {\scriptsize layer 1};
\draw [semithick,->] (1.8,2.835) -- (1.8,2.415);
\node at (3.6,3.045) {\scriptsize layer 2};
\draw [semithick,->] (3.6,2.835) -- (3.6,2.415);
\node at (5.4,3.045) {\scriptsize layer 3};
\draw [semithick,->] (5.4,2.835) -- (5.4,2.415);
\node at (7.2,3.045) {\scriptsize layer 4};
\draw [semithick,->] (7.2,2.835) -- (7.2,2.415);
\node [yshift=0.75cm] at (0,3.2500000000000004) {input};
\node [yshift=0.75cm] at (7.2,3.2500000000000004) {output};
\draw [semithick,decorate,decoration={brace,amplitude=10pt,raise=4pt},yshift=0pt] (1.4400000000000002,3.2500000000000004) -- (5.760000000000001,3.2500000000000004) node [midway,yshift=0.75cm] {hidden layers};
\end{tikzpicture}

%% file: UniversalApproximation.tex
\chapter{Universal approximation}\label{chap:UA}
\newcommand{\tocc}{\xrightarrow{\cc}}
After introducing neural networks in Chapter \ref{chap:FFNNs}, %
it is natural to inquire about their capabilities.
Specifically, we %
might wonder if there exist inherent limitations to the type of functions a neural network can represent.
Could there be a class of functions that neural networks cannot approximate? If so, it would suggest that neural networks are specialized tools, similar to how linear regression is suited for linear relationships, but not for data with nonlinear relationships.

In this chapter, primarily following \cite{LESHNO1993861}, we %
will show that %
this is not the case, and neural networks are indeed a
\emph{universal} tool.
More precisely, given sufficiently large and complex architectures, they can %
approximate almost every sensible input-output relationship.
We will formalize and prove this claim in the subsequent sections.

\section{A universal approximation theorem}\label{sec:UniversalApproximationMainSection}
To analyze what kind of functions can be approximated with neural networks,
we start by considering %
the uniform approximation of continuous functions $f:\R^d\to\R$
on compact sets. To this end, we first introduce the notion of compact convergence.

\begin{definition}\label{def:compactConvergence}
	Let $d\in \N$. A sequence of functions $f_n:\R^d\to\R$, $n\in\N$, is said to {\bf
		converge compactly} to a function $f:\R^d\to\R$, if for every
	compact $K\subseteq \R^d$ it holds that
	$\lim_{n\to\infty}\sup_{\Bx\in K} |f_n(\Bx)-f(\Bx)|=0$.
	In this case we
	write $f_n\tocc f$.
\end{definition}

\begin{definition}\label{def:ckSpace}
  Let $d\in\N$, $k \in \N_0 \cup \{\infty\}$ and
  $\Omega \subseteq \R^d$.  We denote by $C^k(\Omega)$ the set of
  functions $f \colon \Omega \to \R$, such that all partial
  derivatives up to order $k$ exist on the interior of $\Omega$ and
  extend continuously to all of $\Omega$. Moreover for $k<\infty$
  \[
    \|f\|_{C^k(\Omega)} \dfn \sup_{\Bx \in \Omega} \sup_{\substack{\Balpha \in \N_0^d \\ |\Balpha| \le k}} |D^\Balpha f(\Bx)|,
  \]
  and for \( k = \infty \), the inner supremum is taken over all \( \Balpha \in \N_0^d \).
\end{definition}

Throughout what follows, we always consider $C^0(\R^d)$ equipped with
the topology of Definition \ref{def:compactConvergence} (also see Exercise \ref{ex:comptop}), and every subset such as $C^0(D)$ with the subspace
topology: for example, if $D\subseteq\R^d$ is bounded, then
convergence in $C^0(D)$ %
refers to uniform convergence
$\lim_{n\to\infty}\sup_{x\in D}|f_n(x)-f(x)|=0$.

\subsection{Universal approximators}

As stated before, we want to show that deep neural networks can approximate every continuous function in the sense of Definition \ref{def:compactConvergence}.
We call sets of functions that satisfy this property \emph{universal approximators}.

\begin{definition}%
	Let $d\in \N$. A set of functions $\CH$ from $\R^d$ to $\R$ is a {\bf
		universal approximator} (of $C^0(\R^d)$), if for every $\eps>0$,
	every compact $K\subseteq\R^d$, and every $f\in C^0(\R^d)$,
	there exists
	$g\in\CH$ such that $\sup_{\Bx\in K}|f(\Bx)-g(\Bx)|<\eps$.
\end{definition}

For a set of (not necessarily continuous) functions $\CH$ mapping
between $\R^d$ and $\R$, we denote by $\overline{\CH}^\cc$ its closure
with respect to compact convergence.

The relationship between a universal approximator and the closure with respect to compact convergence is established in the proposition below.

\begin{proposition}\label{prop:univapp}
	Let $d\in \N$ and  $\CH$ be a set of functions from $\R^d$ to $\R$. 
	Then, $\CH$ is a universal approximator of $C^0(\R^d)$ if and only if 
	$C^0(\R^d)\subseteq \overline{\CH}^\cc$.
\end{proposition}

\begin{proof}
	Suppose that $\CH$ is a universal approximator and fix
	$f\in C^0(\R^d)$.
	For $n \in \N$, define $K_n\dfn [-n,n]^d\subseteq\R^d$.
	Then for every
	$n\in\N$ there exists $f_n\in\CH$ such that
	$\sup_{\Bx\in K_n}|f_n(\Bx)-f(\Bx)|<{1}/{n}$.
	Since for every compact
	$K\subseteq\R^d$ there exists $n_0$ such that $K\subseteq K_n$ for
	all $n\ge n_0$, it holds $f_n\tocc f$.
	The ``only if'' part of the assertion is trivial.
\end{proof}

A key tool to show that a set is a universal approximator is the Stone-Weierstrass
theorem, see for instance \cite[Sec.~5.7]{rudin-fa}.

\begin{theorem}[Stone-Weierstrass]\label{thm:stone}
	Let $d\in \N$, let $K\subseteq\R^d$ be compact, and let $\CH\subseteq C^0(K,\R)$
	satisfy that
	\begin{enumerate}[label=(\alph*)]
		\item for all $\Bx\in K$ there exists $f\in\CH$ such that $f(\Bx)\neq 0$,
		\item for all $\Bx\neq \By\in K$ there exists $f\in\CH$ such that
		$f(\Bx)\neq f(\By)$,
		\item $\CH$ is an algebra of functions, i.e., $\CH$ is closed under
		addition, multiplication and scalar multiplication.
	\end{enumerate}
	Then $\CH$ is dense in $C^0(K)$.
\end{theorem}

\begin{example}[Polynomials are a universal approximator]
	\label{ex:poly}
        For a multiindex $\Balpha=(\alpha_1,\dots,\alpha_d)\in\N_0^d$ and a
	vector $\Bx=(x_1,\dots,x_d)\in\R^d$ denote
	$\Bx^\Balpha\dfn \prod_{j=1}^d x_j^{\alpha_j}$.
	In the following,
	with $|\Balpha|\dfn \sum_{j=1}^d\alpha_j$, we write
	\begin{align*}
          \CP_n\dfn \spa\set{\Bx^\Balpha}{\Balpha\in\N_0^d,~|\Balpha|\le n}
	\end{align*}
	i.e., $\CP_n$ is the space of polynomials of degree at most $n$
	(with real coefficients).
	It is easy to check that
	$\CP\dfn\bigcup_{n\in\N}\CP_n(\R^d)$ satisfies the assumptions of
	Theorem~\ref{thm:stone} on every compact set $K\subseteq\R^d$.
	Thus
	the space of polynomials $\CP$ is a universal approximator of
	$C^0(\R^d)$, and by Proposition \ref{prop:univapp}, $\CP$ is dense
	in $C^0(\R^d)$.
	In case we wish to emphasize the dimension of the
	underlying space, in the following we will also write $\CP_n(\R^d)$
	or $\CP(\R^d)$ to denote $\CP_n$, $\CP$ respectively.
\end{example}

\subsection{Shallow neural networks}
With the necessary formalism established, %
we can now %
show that shallow neural networks of arbitrary width form a
universal approximator under certain (mild) conditions on the
activation function.
The results in this section are based on
\cite{LESHNO1993861}, and for the proofs we follow the arguments in
that paper.

We first introduce notation for the set of all functions
realized by certain architectures.

\begin{definition}\label{def:CN} 
	Let $d$, $m$, $L$, $n \in \N$ and $\sigma \colon \R \to \R$.
	The set of all functions realized by neural networks with $d$-dimensional
	input, $m$-dimensional output, depth at most $L$, width at most
	$n$, and activation function $\sigma$  is denoted by
	\begin{align*}
		\CN_d^m(\sigma;L,n)\dfn \set{\Phi:\R^d\to\R^m}{\Phi\text{ as in Def.\ \ref{def:nn}, }\depth(\Phi)\le L,~\wdth(\Phi)\le n}.
	\end{align*}
	Furthermore,
	\begin{align*}
		\CN_d^m(\sigma;L)\dfn \bigcup_{n\in\N}\CN_d^m(\sigma;L,n).
	\end{align*}
\end{definition}

In the sequel, we require the activation function $\sigma$ to belong to the set of piecewise continuous and locally bounded functions %
\begin{equation}\label{eq:M}
  \begin{aligned}
	\CM\dfn \big\{\sigma\in L_{\rm loc}^\infty(\R) \;\big|\;&\text{there exist intervals }I_1,\dots,I_M \text{ partitioning } \R,\\
                                                                &\text{s.t. }\sigma\in C^0(I_j) \text{ for all } j = 1, \dots, M \big\}.
                                                                  \end{aligned}
\end{equation}
Here, $M\in\N$ is finite, and the intervals $I_j$ are understood to
have positive (possibly infinite) Lebesgue measure, i.e.\ $I_j$ is
not allowed to be empty or a single point.
Hence, $\sigma$ is a piecewise
continuous function, and it has discontinuities at at most finitely
many points.

\begin{example}%
	\label{ex:M}
	Activation functions belonging to $\CM$ include, in particular, all continuous non-polynomial functions, which
          in turn includes all practically
          relevant activation functions such as the
	ReLU, %
        the SiLU, %
        and the Sigmoid discussed in
        Section \ref{sec:activationFunctions}.
	In these cases, we can choose $M=1$
	and $I_1=\R$.
	Discontinuous functions include for example the
	Heaviside function $x \mapsto \ind_{x>0}$ (also called a
	``perceptron'' in this context) but also
	$x \mapsto \ind_{x>0}\sin(1/x)$: Both belong to $\CM$ with $M=2$,
	$I_1=(-\infty,0]$ and $I_2=(0,\infty)$.
	We exclude for example the
	function $x \mapsto {1}/{x}$, which is not locally bounded.
\end{example}

The rest of this subsection is dedicated to proving the following
theorem that has now already been announced repeatedly. 

\begin{theorem}\label{thm:universal}
	Let $d\in\N$ and $\sigma\in\CM$.
	Then $\CN_d^1(\sigma;1)$ is a
	universal approximator of $C^0(\R^d)$ if and only if $\sigma$ is not a
	polynomial.
\end{theorem}

\begin{remark}
  We will see in Corollary \ref{cor:universal} and Exercise \ref{ex:universalInLp} that neural networks can also arbitrarily well approximate non-continuous functions with respect to suitable norms.
\end{remark}

The universal approximation theorem by Leshno, Lin, Pinkus and Schocken
\cite{LESHNO1993861}---of which Theorem \ref{thm:universal} is a special case---is even formulated for a much larger set
$\CM$, which allows for activation functions that have discontinuities
at a (possibly non-finite) set of Lebesgue measure zero.
Instead of
proving the theorem in this generality, we resort to the simpler case
stated above.
This allows to avoid some technicalities, but the main
ideas remain the same.
The proof strategy is to verify the following
three claims:
\begin{enumerate}
	\item\label{item:ua1} {\bf reduction to univariate target functions:} if $C^0(\R^1)\subseteq \overline{\CN_1^1(\sigma;1)}^{\rm cc}$
	then $C^0(\R^d)\subseteq \overline{\CN_d^1(\sigma;1)}^{\rm cc}$,
	\item\label{item:ua2} {\bf reduction to smooth activation functions:} if $\sigma\in C^\infty(\R)$ is not a polynomial
	then $C^0(\R^1)\subseteq \overline{\CN_1^1(\sigma;1)}^{\rm cc}$,
	\item\label{item:ua3} {\bf general case:} if $\sigma\in \CM$ is not a polynomial then
	there exists
	$\tilde\sigma \in C^\infty(\R)\cap \overline{\CN_1^1(\sigma;1)}^{\rm
		cc}$ which is not a polynomial.
\end{enumerate}
Upon observing that 
$\tilde\sigma\in \overline{\CN_1^1(\sigma;1)}^{\rm cc}$ implies
$\overline{\CN_1^1(\tilde\sigma,1)}^{\rm cc}\subseteq
\overline{\CN_1^1(\sigma;1)}^{\rm cc}$, it is easy to see that these
statements together with Proposition \ref{prop:univapp}
establish %
the implication ``$\Leftarrow$'' %
asserted in Theorem \ref{thm:universal}.
The reverse direction is
straightforward to check and will be the content of Exercise \ref{ex:ra}.

We start with a more general version of \ref{item:ua1} and reduce
the problem to the one dimensional case following \cite[Theorem 2.1]{LIN1993295}.

\begin{lemma}\label{lemma:1tod}
	Assume that $\CH$ is a universal approximator of $C^0(\R)$.
	Then
	for every $d\in\N$
	\begin{align*}
		\spa\set{\Bx \mapsto g(\Bw\cdot\Bx)}{\Bw\in\R^d,~g\in\CH}
	\end{align*}
	is a universal approximator of $C^0(\R^d)$.
\end{lemma}

\begin{proof}
For $k\in\N_0$, denote by $\bbH_k$ the space of all $k$-homogeneous
polynomials, that is
\begin{align*}
	\bbH_k\dfn
	\spa\setc{\R^d \ni \Bx \mapsto \Bx^\Balpha}{\Balpha\in\N_0^d,~|\Balpha|=k}.
\end{align*}
We claim that
\begin{align}\label{eq:ReductionOfDimensionToShow}
	\bbH_k\subseteq \overline{\spa\set{\R^d \ni  \Bx \mapsto  g(\Bw\cdot\Bx)}{\Bw\in\R^d,~g\in\CH}}^\cc
	\dfnn X
\end{align}
for all $k\in\N_0$.
This implies that all multivariate polynomials
belong to $X$.
An application of the Stone-Weierstrass theorem
(cp.~Example \ref{ex:poly}) and Proposition \ref{prop:univapp} then
conclude the proof.

For every $\Balpha$, $\Bbeta\in\N_0^d$ with $|\Balpha|=|\Bbeta|=k$, it
holds $D^\Bbeta \Bx^\Balpha=\delta_{\Bbeta,\Balpha} \Balpha!$, where
$\Balpha!\dfn \prod_{j=1}^d\alpha_j!$ and
$\delta_{\Bbeta,\Balpha}=1$ if $\Bbeta=\Balpha$ and
$\delta_{\Bbeta,\Balpha}=0$ otherwise.
Hence, since
$\set{\Bx \mapsto \Bx^\Balpha}{|\Balpha|=k}$ is a basis of $\bbH_k$, the set
$\set{D^\Balpha}{|\Balpha|=k}$ is a basis of its topological dual
$\bbH_k'$.
Thus each linear functional $l \in \bbH_k'$ allows the
representation $l=p(D)$ for some $p\in \bbH_k$ (here $D$ stands for
the differential).

By the multinomial formula
\begin{align*}
	(\Bw\cdot\Bx)^k =
	\left(\sum_{j=1}^d w_jx_j\right)^k=
	\sum_{\set{\Balpha\in\N_0^d}{|\Balpha|=k}} \frac{k!}{\Balpha!}
	\Bw^\Balpha\Bx^\Balpha.
\end{align*}
Therefore, we have that $(\Bx \mapsto (\Bw\cdot\Bx)^k) \in \bbH_k$. 
Moreover, for every $l=p(D)\in \bbH_k'$ and all $\Bw\in\R^d$ we have that 
\begin{align*}
	l(\Bx \mapsto (\Bw\cdot\Bx)^k)= k! p(\Bw).
\end{align*}
Hence, if $l(\Bx \mapsto(\Bw\cdot\Bx)^k)=p(D)(\Bx \mapsto(\Bw\cdot\Bx)^k)=0$ for all
$\Bw\in\R^d$, then $p\equiv 0$ and thus $l\equiv 0$.

This implies
$\spa\set{\Bx \mapsto(\Bw\cdot\Bx)^k}{\Bw\in\R^d}=\bbH_k$. 
Indeed, if there exists $h \in \bbH_k$ which is not in  $\spa\set{\Bx \mapsto (\Bw\cdot\Bx)^k}{\Bw\in\R^d}$, 
then by the theorem of Hahn-Banach (see Theorem \ref{thm:banachSeparation}), there exists a non-zero functional in $\bbH_k'$ vanishing on $\spa\set{\Bx \mapsto(\Bw\cdot\Bx)^k}{\Bw\in\R^d}$.
This contradicts the previous observation. 

By the universality of $\CH$ it is not hard to see that 
$\Bx \mapsto (\Bw\cdot\Bx)^k\in X$ for all $\Bw\in\R^d$.
Therefore, we have
$\bbH_k\subseteq X$ for all $k\in\N_0$. 
\end{proof}

By the above lemma, in order to verify that $\CN_d^1(\sigma;1)$ is a
universal approximator, it suffices to show that $\CN_1^1(\sigma;1)$
is a universal approximator.
We first show that this is the case for sigmoidal activations.

\begin{definition}%
  \label{def:sigmoidalActivation}
An activation function $\sigma:\R\to\R$ is called {\bf sigmoidal},
if $\sigma\in C^0(\R)$, $\lim_{x\to\infty}\sigma(x)=1$ and
$\lim_{x\to-\infty}\sigma(x)=0$.
\end{definition}

For sigmoidal activation functions we can now conclude the universality in the univariate case.

\begin{lemma}\label{lemma:sigmoidal}
Let $\sigma:\R\to\R$ be monotonically increasing and sigmoidal.
Then
$C^0(\R)\subseteq \overline{\CN_1^1(\sigma;1)}^{\rm cc}$\!\!\!.
\end{lemma}

We prove Lemma \ref{lemma:sigmoidal} in Exercise \ref{ex:sigmoidalStepfun1d}.
Lemma \ref{lemma:1tod} and Lemma \ref{lemma:sigmoidal} show Theorem
\ref{thm:universal} in the special case where $\sigma$ is monotonically increasing and sigmoidal.
For the general case, let us continue with
\ref{item:ua2} and consider
$C^\infty$ activations.

\begin{lemma}\label{lemma:notpolydense}
If $\sigma\in C^\infty(\R)$ and $\sigma$ is not a polynomial, then
$\CN_1^1(\sigma;1)$ is dense in $C^0(\R)$.
\end{lemma}

\begin{proof}
	Denote $X\dfn\overline{\CN_1^1(\sigma;1)}^\cc$.
We show again that
all polynomials belong to $X$.
An application of the
Stone-Weierstrass theorem then gives the statement.

Fix $b\in\R$ and denote $f_x(w)\dfn \sigma(wx+b)$
for all $x$, $w\in\R$.
By Taylor's
theorem, for $h\neq 0$
\begin{align}\label{eq:taylorlimit}
	\frac{\sigma((w+h)x+b)-\sigma(wx+b)}{h}
	&=\frac{f_x(w+h)-f_x(w)}{h}\nonumber\\
	&=f_x'(w)+\frac{h}{2}f_x''(\xi)\nonumber\\
	&=f_x'(w)+\frac{h}{2}x^2\sigma''(\xi x+b)
\end{align}
for some $\xi=\xi(h)$ between $w$ and $w+h$.
Note that the left-hand side belongs to $\CN_1^1(\sigma;1)$
as a function of $x$.
Since
$\sigma''\in C^0(\R)$, for every compact set $K\subseteq\R$
\begin{align*}
	\sup_{x\in K}\sup_{|h|\le 1}|x^2\sigma''(\xi(h) x+b)|\le
	\sup_{x\in K}\sup_{\eta\in[w-1,w+1]}|x^2\sigma''(\eta x+b)|<\infty.
\end{align*}
Letting $h\to 0$, %
as a function of $x$ the term in \eqref{eq:taylorlimit} thus
converges uniformly towards $K\ni x\mapsto f_x'(w)$.
Since $K$ was
arbitrary, $x\mapsto f_x'(w)$ belongs to $X$.
Inductively applying
the same argument to $f_x^{(k-1)}(w)$, we find that
$x\mapsto f_x^{(k)}(w)$ belongs to $X$ for all $k\in\N$,
$w\in\R$.
Observe that $f_x^{(k)}(w) =
x^k\sigma^{(k)}(wx+b)$.
Since $\sigma$ is not a polynomial, for each
$k\in\N$ there exists $b_k\in\R$ such that
$\sigma^{(k)}(b_k)\neq 0$.
Choosing $w=0$, we obtain that
$x\mapsto \sigma^{(k)}(b_k) x^k$ belongs to $X$,
and thus also
$x\mapsto x^k$ belongs to $X$.
\end{proof}

Finally, we come to the proof of \ref{item:ua3}---the claim
that there exists at least one non-polynomial $C^\infty(\R)$
function in the closure of $\CN_1^1(\sigma;1)$.
The argument is
split into two lemmata.
Denote in the following by $C_c^\infty(\R)$ the set of compactly
supported $C^\infty(\R)$ functions, and for two functions $f$, $g:\R\to\R$
  let
  \begin{equation}\label{eq:convolution}
    f*g(x)\dfn \int_{\R}f(x-y)g(y)\dd x\qquad\text{for all }x\in\R
  \end{equation}
  be the convolution of $f$ and $g$.

\begin{lemma}\label{lemma:convolution}
	Let $\sigma\in\CM$.
	Then for each $\varphi\in C_c^\infty(\R)$ it
	holds $\sigma*\varphi\in \overline{\CN_1^1(\sigma;1)}^\cc$.
\end{lemma}

\begin{proof}
	Fix $\varphi\in C_c^\infty(\R)$ and let $a>0$ such
	that $\supp\varphi\subseteq [-a,a]$.
	Denote $y_j\dfn -a+2a {j}/{n}$ for $j=0,\dots,n$ and 
	define for $x \in \R$
	\begin{align*}
		f_n(x) \coloneqq \frac{2a}{n}
		\sum_{j=0}^{n-1}\sigma(x-y_j)\varphi(y_j).
	\end{align*}
	Clearly,  $f_n \in \CN_1^1(\sigma;1).$
	We will show $f_n\tocc \sigma*\varphi$ as $n\to\infty$.
	To do so we
	verify uniform convergence of $f_n$ towards $\sigma*\varphi$ on the
	interval $[-b,b]$
	with
	$b>0$ %
	arbitrary but fixed.
	
	For $x\in [-b,b]$
	\begin{align}\label{eq:leeps}
		|\sigma*\varphi(x)-f_n(x)|
		\le \sum_{j=0}^{n-1}\left|\int_{y_{j}}^{y_{j+1}}\sigma(x-y)\varphi(y)-\sigma(x-y_j)\varphi(y_j)\dd y\right|.
	\end{align}
	Fix $\eps\in (0,1)$.
	Since $\sigma\in\CM$, there exist 
	$z_1,\dots,z_M\in\R$ such that $\sigma$ is continuous on
	$\R\backslash\{z_1,\dots,z_M\}$ (cp.~\eqref{eq:M}).
	With
	$D_\eps\dfn \bigcup_{j=1}^M(z_j-\eps,z_j+\eps)$, observe that
	$\sigma$ is uniformly continuous on the compact set
	$K_\eps\dfn[-a-b,a+b]\cap D_\eps^c$.
	Now let
	$J_c\cup J_d =\{0,\dots,n-1\}$ be a partition (depending on $x$),
	such that %
	$j\in J_c$ if and only if   $[x-y_{j+1},x-y_j]\subseteq K_\eps$.
	Hence, $j\in J_d$ implies the existence of $i\in\{1,\dots,M\}$
	such that %
        the distance of $z_i$ to $[x-y_{j+1},x-y_j]$ is at most $\eps$.
	Due to the interval $[x-y_{j+1},x-y_{j}]$
	having length ${2a}/{n}$, we can bound
	\begin{align*}
          \sum_{j\in J_d}y_{j+1}-y_{j}
          &=\left|\bigcup_{j\in J_d}[x-y_{j+1},x-y_j]\right|\\
          &\le \left|\bigcup_{i=1}^M\Big[z_i-\eps-\frac{2a}{n},z_i+\eps+\frac{2a}{n}\Big]\right|\\
          &\le M \cdot \Big(2\eps+\frac{4a}{n}\Big),
	\end{align*}
        where $|A|$ denotes the Lebesgue measure of a measurable set $A\subseteq\R$.
	Next, because of the local boundedness of $\sigma$ and the fact that
	$\varphi\in C_c^\infty$, it holds
	$\sup_{|y|\le a+b}|\sigma(y)|+\sup_{|y|\le a}|\varphi(y)|\dfnn
	\gamma<\infty$.
	Hence
	\begin{align}\label{eq:uniformconvconv}
		&|\sigma*\varphi(x)-f_n(x)| \nonumber\\
		&\qquad \le
		\sum_{j\in J_c\cup J_d}\left|\int_{y_{j}}^{y_{j+1}}\sigma(x-y)\varphi(y)-\sigma(x-y_j)\varphi(y_j)\dd y\right|
		\nonumber\\
		&\qquad \le 2\gamma^2 M \cdot \left(2\eps+\frac{4a}{n}\right) \nonumber\\
		&\qquad \qquad  +
		2a\sup_{j\in J_c} \max_{y\in [y_{j},y_{j+1}]}| \sigma(x-y)\varphi(y)-\sigma(x-y_j)\varphi(y_j)|. 
	\end{align}
	We can bound the term in the last maximum by
	\begin{align*}
		&|\sigma(x-y)\varphi(y)-\sigma(x-y_j)\varphi(y_j)|\\
		&\qquad \le |\sigma(x-y)-\sigma(x-y_j)||\varphi(y)|
		+|\sigma(x-y_j)||\varphi(y)-\varphi(y_j)|\nonumber\\
		&\qquad \le \gamma \cdot \left( \sup_{\substack{z_1,z_2\in K_\eps\\ |z_1-z_2|\le\frac{2a}{n}}} |\sigma(z_1)-\sigma(z_2)| +
		\sup_{\substack{z_1,z_2\in %
				[-a,a]
				\\ |z_1-z_2|\le\frac{2a}{n}}} |\varphi(z_1)-\varphi(z_2)|
		\right).
	\end{align*}
	Finally, uniform continuity of $\sigma$ on $K_\eps$
	and $\varphi$ on $[-a,a]$
	imply that the last term tends to $0$ as $n\to\infty$ uniformly for
	all $x\in [-b,b]$.
	This shows that there exist $C<\infty$
	(independent of $\eps$ and $x$) and $n_\eps\in\N$ (independent of
	$x$) such that the term in \eqref{eq:uniformconvconv} is bounded by
	$C\eps$ for all $n\ge n_\eps$. 
	Since $\eps$ was arbitrary, this yields the claim.
\end{proof}

\begin{lemma}\label{lemma:Mpoly}
	If $\sigma\in \CM$ and $\sigma*\varphi$ is a polynomial for all
	$\varphi\in C_c^\infty(\R)$, then $\sigma$ is a polynomial.
\end{lemma}

\begin{proof}
	Fix $-\infty<a<b<\infty$ and consider $C_c^\infty(a,b)\dfn
	\set{\varphi\in C^\infty(\R)}{\supp\varphi\subseteq [a,b]}$.
	Define
	a metric $\rho$ on $C_c^\infty(a,b)$ via
	\begin{align*}
		\rho(\varphi,\psi)\dfn \sum_{j\in\N_0} 2^{-j}
		\frac{|\varphi-\psi|_{C^j(a,b)}}{1+|\varphi-\psi|_{C^j(a,b)}},
	\end{align*}
	where
	\begin{align*}
		|\varphi|_{C^j(a,b)}
		\dfn \sup_{x\in [a,b]}|\varphi^{(j)}(x)|.
	\end{align*}
	Since the space of $j$ times differentiable functions on $[a,b]$
	is complete with respect to the norm
	$\sum_{i=0}^j|\cdot|_{C^i(a,b)}$, see for instance \cite[Satz
	104.3]{HH1}, the space $C_c^\infty(a,b)$ is complete with the metric $\rho$. 
	For $k\in\N$ set
		\begin{align*}
			V_k\dfn \set{\varphi\in C_c^\infty(a,b)}{\sigma*\varphi\in\CP_k},
		\end{align*}
		where $\CP_k\dfn \spa\set{\R \ni x \mapsto x^j}{0\le j\le k}$ denotes the space of
		polynomials of degree at most $k$.
		Then $V_k$ is closed with respect to the
		metric $\rho$.
		To see this, we need to show
                that for a converging sequence $\varphi_j \to \varphi^*$ with respect to $\rho$ and $\varphi_j  \in V_k$, it follows that $D^{k+1}(\sigma * \varphi^*)= 0$ and hence $\sigma * \varphi^*$ is a polynomial: Using $D^{k+1}(\sigma *\varphi_j) = 0$ if $\varphi_j\in V_k$, the linearity of the convolution, and the fact that $D^{k+1}(\sigma*g) = \sigma*D^{k+1}(g)$ for differentiable $g$ and if both sides are well-defined, we get 
		\begin{align*}
			&\sup_{x \in [a,b]}| D^{k+1}(\sigma * \varphi^*)(x)|\\
			 & \qquad = 	\sup_{x \in [a,b]}| \sigma * D^{k+1}( \varphi^* - \varphi_j)(x)| \\
			& \qquad \leq |b-a| \sup_{z \in [a-b, b-a]} |\sigma(z)|	\cdot \sup_{x \in [a,b]}| D^{k+1}( \varphi_j -\varphi^*)(x)|.
		\end{align*} 	
		Since $\sigma$ is locally bounded, the right hand-side converges to $0$ as $j\to\infty$.

		By assumption we have
		\begin{align*}
			\bigcup_{k\in\N}V_k = C_c^\infty(a,b).
		\end{align*}
		Baire's category theorem (Theorem \ref{thm:BaireCat}) implies the existence of $k_0\in\N$
		(depending on $a$, $b$) such that $V_{k_0}$ contains an open subset
		of $C_c^\infty(a,b)$.
		Since $V_{k_0}$ is a vector space, it must
		hold $V_{k_0}=C_c^\infty(a,b)$.

		We now show that $\varphi*\sigma\in \CP_{k_0}$ for every
		$\varphi\in C_c^\infty(\R)$; in other words, $k_0=k_0(a,b)$ can be
		chosen independent of $a$ and $b$.
		First consider a shift $s\in\R$
		and let $\tilde a\dfn a+s$ and $\tilde b\dfn b+s$.
		Then with
		$S(x)\dfn x+s$, for any $\varphi\in C_c^\infty(\tilde a,\tilde b)$
		holds $\varphi\circ S\in C_c^\infty(a,b)$, and thus
		$(\varphi\circ S)*\sigma\in\CP_{k_0}$.
		Since
		$(\varphi\circ S)*\sigma(x) = \varphi*\sigma(x+s)$, we conclude that
		$\varphi*\sigma\in\CP_{k_0}$.
		Next let
		$-\infty<\tilde a<\tilde b<\infty$ be \emph{arbitrary}.
		Then, for any
		integer $n>(\tilde b-\tilde a)/(b-a)$ we can cover
		$(\tilde a,\tilde b)$ with $n\in\N$ overlapping open intervals
		$(a_{1},b_{1}),\dots,(a_n,b_n)$, each of length $b-a$.
		Any
		$\varphi\in C_c^\infty(\tilde a,\tilde b)$ can be written as
		$\varphi=\sum_{j=1}^n\varphi_j$ where
		$\varphi_j\in C_c^\infty(a_j,b_j)$.
		Then
		$\varphi*\sigma = \sum_{j=1}^n\varphi_j*\sigma\in\CP_{k_0}$, and
		thus $\varphi*\sigma\in\CP_{k_0}$ for every
		$\varphi\in C_c^\infty(\R)$.
		
		Finally, Exercise
		\ref{ex:completeProofOfMpoly} implies $\sigma\in\CP_{k_0}$.
		\end{proof}

		Now we can put everything together to show Theorem~\ref{thm:universal}.

\begin{proof}[of Theorem~\ref{thm:universal}]
			By Exercise \ref{ex:ra} we have the implication ``$\Rightarrow$''.
			
			For the other direction we assume that $\sigma\in\CM$ is not a
			polynomial.
			Then by Lemma \ref{lemma:Mpoly} there exists
			$\varphi\in C_c^\infty(\R)$ such that $\sigma*\varphi$ is not a
			polynomial.
			According to Lemma \ref{lemma:convolution} we have
			$\sigma*\varphi\in \overline{\CN_1^1(\sigma;1)}^\cc$.
			We conclude
			with Lemma \ref{lemma:notpolydense} that $\CN_1^1(\sigma;1)$ is a
			universal approximator of $C^0(\R)$.
			
			Finally, by Lemma \ref{lemma:1tod}, $\CN_d^1(\sigma;1)$ is a universal approximator of
			$C^0(\R^d)$.
\end{proof}

\subsection{Deep neural networks}
Theorem \ref{thm:universal} shows the universal approximation capability of single-hidden-layer neural networks with activation functions $\sigma\in\CM\backslash\CP$: they can approximate every continuous function on every compact set to arbitrary precision, given sufficient width. This result directly extends to neural networks of any fixed depth $L\ge 1$. The idea is to use the fact that the identity function can be approximated with a shallow neural network. Composing a shallow neural network approximation of the target function $f$ with (multiple) shallow neural networks approximating the identity function, gives a deep neural network approximation of $f$.

Instead of directly applying Theorem \ref{thm:universal}, we first establish the following proposition regarding the approximation of the identity function. Rather than $\sigma\in\CM\backslash\CP$, it requires a different (mild) assumption on the activation function. This allows for a constructive proof, yielding explicit bounds on the neural network size, which will prove useful later in the book.

\begin{proposition}\label{prop:Identity1}
  Let $d$, $L\in \N$, let $K \subseteq\R^d$ be compact, and let
  $\sigma: \R \to \R$ be such that there exists an open set on which
  $\sigma$ is differentiable and not constant.  Then, for every
  $\eps >0$, there exists a neural network
  $\Phi \in \CN_d^d(\sigma; L, d)$ such that
  \[
    \norm[\infty]{\Phi(\Bx) - \Bx} < \eps\qquad\text{for all
    }\Bx\in K.
  \]
\end{proposition}

\begin{proof}
  The proof uses the same idea as in Lemma
    \ref{lemma:notpolydense}, where we approximate the derivative of
  the activation function by a simple neural network.
  Let us first assume $d \in \N$ and $L = 1$.
			
  Let $x^* \in \R$ be such that $\sigma$ is differentiable on a
  neighborhood of $x^*$ and $\sigma'(x^*) = \theta \neq 0$.  Moreover,
  let $\Bx^* = (x^*, \dots, x^*) \in \R^d$.  Then, for $\lambda >0$ we
  define
  \[
    \Phi_\lambda(\Bx) \coloneqq \frac{\lambda}{\theta} \sigma\left(
      \frac{\Bx}{\lambda} + \Bx^*\right) - \frac{\lambda}{\theta}
    \sigma(\Bx^*),
  \]
                      
  Then, we have, for all $\Bx \in K$,
  \begin{align}\label{eq:estimateRealisation}
    \Phi_\lambda(\Bx) - \Bx = \lambda \frac{\sigma(\Bx/\lambda + \Bx^*) - \sigma(\Bx^*)}{\theta} - \Bx. 
  \end{align}
  If %
  $x_i=0$ for $i \in \{1, \dots, d\}$, then
  \eqref{eq:estimateRealisation} shows that
  $(\Phi_\lambda(\Bx) - \Bx)_i = 0$.  Otherwise
  \[
    |(\Phi_\lambda(\Bx) - \Bx)_i| = \frac{|x_i|}{|\theta|}\left|
      \frac{\sigma(x_i/\lambda + x^*) - \sigma(x^*)}{x_i/\lambda}
      - \theta\right|.
  \]
  By the definition of the derivative, we have that
  $|(\Phi_\lambda(\Bx) - \Bx)_i| \to 0$ for $\lambda \to \infty$
  uniformly for all $\Bx \in K$ and $i \in \{1, \dots, d\}$.
  Therefore, $|\Phi_\lambda(\Bx) - \Bx| \to 0$ for
  $\lambda \to \infty$ uniformly for all $\Bx \in K$.
			
  The extension to $L > 1$ is straightforward and is the content of
  Exercise \ref{ex:ExtendToDeep}.
\end{proof}

  Using the aforementioned generalization of Proposition \ref{prop:Identity1}
  to arbitrary non-polynomial activation functions $\sigma\in\CM$, we
  obtain the following extension of Theorem \ref{thm:universal}.
  
  \begin{corollary}\label{cor:universaldeep}
    Let $d\in\N$, $L\in\N$ and $\sigma\in\CM$.  Then
    $\CN_d^1(\sigma;L)$ is a universal approximator of $C^0(\R^d)$ if
    and only if $\sigma$ is not a polynomial.
  \end{corollary}
  \begin{proof}
    We only show the implication ``$\Leftarrow$''. The other direction
    is again left as an exercise, see Exercise \ref{ex:ra}.
        
    Assume $\sigma\in\CM$ is not a polynomial, let $K\subseteq\R^d$ be
    compact, and let $f\in C^0(\R^d)$. Fix $\eps\in (0,1)$. We need to
    show that there exists a neural network $\Phi\in\CN_d^1(\sigma;L)$
    such that $\sup_{\Bx\in K}|f(\Bx)-\Phi(\Bx)|<\eps$. The case $L=1$
    holds by Theorem \ref{thm:universal}, so let $L>1$.

    By Theorem \ref{thm:universal}, there exist
    $\Phi_{\rm shallow}\in \CN_d^1(\sigma;1)$ such that
    \begin{equation}\label{eq:Phishallowerr}
      \sup_{\Bx\in K}|f(\Bx)-\Phi_{\rm shallow}(\Bx)|<\frac{\eps}{2}.
    \end{equation}
    Compactness of $\set{f(\Bx)}{\Bx\in K}$ implies that we can find
    $n>0$ such that
    \begin{equation}\label{eq:imPhi1}
      \set{\Phi_{\rm shallow}(\Bx)}{\Bx\in K}\subseteq [-n,n].
    \end{equation}

    Let $\Phi_{\rm id}\in \CN_1^1(\sigma;L-1)$ be an approximation to
    the identity such that
    \begin{equation}\label{eq:Phiiderr}
      \sup_{x\in [-n,n]}|x-\Phi_{\rm id}(x)|<\frac{\eps}{2},
    \end{equation}
    which is possible by the extension of Proposition \ref{prop:Identity1}
    to general non-polynomial activation functions $\sigma\in\CM$.
        
    Denote $\Phi\dfn \Phi_{\rm id}\circ \Phi_{\rm shallow}$.
    According to Proposition \ref{prop:VectorSpaceAndCompositions}
    \ref{item:compositionPhi} holds $\Phi\in\CN_d^1(\sigma;L)$ as
    desired.  Moreover \eqref{eq:Phishallowerr}, \eqref{eq:imPhi1},
    \eqref{eq:Phiiderr} imply
    \begin{align*}
      \sup_{\Bx\in K}|f(\Bx)-\Phi(\Bx)|
      &=\sup_{\Bx\in K}|f(\Bx)-\Phi_{\rm id}(\Phi_{\rm shallow}(\Bx))|\\
      &\le \sup_{\Bx\in K}\big(|f(\Bx)-\Phi_{\rm shallow}(\Bx)|+
        |\Phi_{\rm shallow}(\Bx)-\Phi_{\rm id}(\Phi_{\rm shallow}(\Bx))|\big)\\
      &\le \frac{\eps}{2} + \frac{\eps}{2} = \eps.
    \end{align*}
    This concludes the proof.
  \end{proof}

  \subsection{Other norms}
  In addition to the case of continuous functions, universal approximation theorems can be shown for various other function classes and topologies, which may also allow for the approximation of functions exhibiting discontinuities or singularities.
		To give but one example, we next state such a result for Lebesgue spaces on compact sets.
		The proof is left to the reader, see Exercise \ref{ex:universalInLp}.

\begin{corollary}\label{cor:universal}
  Let $d\in\N$, $L\in\N$, $p \in [1, \infty)$, and let $\sigma\in\CM$ not be a polynomial.
  Then for every $\eps>0$, every compact $K \subseteq\R^d$, and every $f \in L^p(K)$ there exists $\Phi^{f,\eps} \in \CN_d^1(\sigma;L)$ such that 
  \[
    \left(\int_{K} | f(\Bx) - \Phi(\Bx)|^p \dd\Bx\right)^{1/p} \leq \eps.
  \]
\end{corollary}

		\section{Superexpressive activations and Kolmogorov's superposition theorem %
		}\label{sec:kolmogorov}
		In the previous
		section, we saw that a %
		large class of activation functions allow for universal approximation. 
		However, %
                these results did not provide any insights into the necessary neural network size for achieving a specific accuracy.
		
		Before exploring this topic further in the following chapters, we
                next present a remarkable result that shows how the
		required neural network size is significantly influenced by the choice of
		activation function.
		The result asserts that, with the appropriate activation function, every %
                $f\in C^0(K)$ on a compact set $K\subseteq\R^d$ can be
		approximated to \emph{every desired accuracy} $\eps>0$ using a neural network
		of size $O(d^2)$; in particular the neural network size is independent of
		$\eps>0$, $K$, and $f$.
		We will first discuss the one-dimensional
		case.

\begin{proposition}\label{prop:magic}
			There exists a continuous activation function $\sigma:\R\to\R$ such
			that for every compact $K\subseteq\R$, every $\eps>0$ and every
			$f\in C^0(K)$ there exists %
			$\Phi(x)=\sigma(wx+b)\in\CN_1^1(\sigma;1,1)$ such that
			\begin{align*}
				\sup_{x\in K}|f(x)-\Phi(x)|<\eps.
			\end{align*}
		\end{proposition}

\begin{proof}
			Denote by $\tilde \CP_n$ all polynomials $p(x)=\sum_{j=0}^n q_jx^j$
			with rational coefficients, i.e.\ such that $q_j\in\bbQ$ for all
			$j=0,\dots,n$.
			Then $\tilde\CP_n$ can be identified with the
			$n$-fold Cartesian product $\bbQ\times\cdots\times\bbQ$, and thus
			$\tilde\CP_n$ is a countable set.
			Consequently also the set
			$\tilde\CP\dfn\bigcup_{n\in\N}\tilde\CP_n$ of all polynomials with
			rational coefficients is countable.
			Let $(p_i)_{i\in\Z}$ be an
			enumeration of these polynomials, and set
			\begin{align*}
				\sigma(x)\dfn
				\begin{cases}
					p_i(x-2i) &\text{if }x\in [2i,2i+1]\\
					p_i(1)(2i+2-x)+p_{i+1}(0)(x-2i-1) &\text{if }x\in (2i+1,2i+2).
				\end{cases}
			\end{align*}
			In words, $\sigma$ equals $p_i$ on even intervals $[2i, 2i+1]$
			and is linear on odd intervals $[2i+1, 2i+2]$, resulting in a
			continuous function overall.

			We first assume $K=[0,1]$.
			By Example \ref{ex:poly}, for every
			$\eps>0$ exists $p(x)=\sum_{j=1}^nr_jx^j$ such that
			$\sup_{x\in [0,1]}|p(x)-f(x)|<{\eps}/{2}$.
			Now choose
			$q_j\in\bbQ$ so close to $r_j$ such that
			$\tilde p(x)\dfn \sum_{j=1}^n q_jx^j$ satisfies
			$\sup_{x\in [0,1]}|\tilde p(x)-p(x)|<{\eps}/{2}$.
			Let $i\in\Z$
			such that $\tilde p(x)=p_i(x)$, i.e., $p_i(x)=\sigma(2i+x)$ for all
			$x\in [0,1]$.
			Then $\sup_{x\in [0,1]}|f(x)-\sigma(x+2i)|<\eps$.
			
			For general compact $K$ assume that $K\subseteq [a,b]$.
			By
			Tietze's extension theorem, $f$ allows a continuous extension to
			$[a,b]$, so without loss of generality
			$K=[a,b]$.
			By the first case we can find $i\in\Z$ such that with
			$y=(x-a)/(b-a)$ (i.e.\ $y\in [0,1]$ if
			$x\in [a,b]$)
			\begin{align*}
				\sup_{x\in [a,b]}\left|f(x)-\sigma\left(\frac{x-a}{b-a}+2i\right)\right|
				=
				\sup_{y\in [0,1]}\left|f(y\cdot (b-a)+a)-\sigma(y+2i)\right|
				<\eps,
			\end{align*}
			which gives the statement with $w={1}/(b-a)$ and
			$b=-{a}\cdot(b-a)+2i$.
		\end{proof}

		To extend this result to arbitrary dimension, we will %
		use Kolmogorov's superposition theorem.
		It states that every continuous function %
                of $d$ variables can be expressed as a composition of functions that
                each depend only on one variable.

\begin{theorem}[Kolmogorov]\label{thm:kolmogorov}
			For every $d\in\N$ there exist $2d^2+d$ monotonically increasing functions
			$\varphi_{i,j}\in C^0(\R)$, $i=1,\dots,d$, $j=1,\dots,2d+1$, such
			that for every $f\in C^0([0,1]^d)$ there exist functions
			$f_j\in C^0(\R)$, $j=1,\dots,2d+1$ satisfying
			\begin{align*}
				f(\Bx)=\sum_{j=1}^{2d+1} f_j\left(\sum_{i=1}^d \varphi_{i,j}(x_i) \right)\qquad\text{for all }\Bx\in [0,1]^d.
			\end{align*}
		\end{theorem}

                Kolmogorov's theorem was proven in \cite{MR0111809}. To avoid the most technical parts of the argument, we show a simpler statement where the inner functions are allowed to be discontinuous. While this sidesteps a key difficulty, the proof still illustrates why it is possible to reduce the problem to univariate functions.
  
  \begin{proposition}\label{prop:kolmogorov}
    For every $d\in\N$ and every $\eps>0$ there exist $d$
    monotonically increasing functions $h_j:[0,1]\to\R$, $j=1,\dots,d$, such
    that for every $L$-Lipschitz continuous $f:[0,1]^d\to\R$
    there exists $g\in C^0(\R)$ with
    \begin{equation}\label{eq:kolmogorovprop}
      \sup_{\Bx\in [0,1]^d}\Big|f(\Bx)-g\Big(\sum_{j=1}^dh_j(x_j)\Big)\Big|\le L \cdot \eps.
    \end{equation}
  \end{proposition}

  \begin{proof}
    Let $n\in\N$ and $a_i=i/n$ for $i=0,\dots,n$. For
    $\Bnu\in\Lambda_n\dfn \{0,\dots,n-1\}^d$ set
    $Q_\Bnu\dfn \times_{j=1}^d I_{\nu_j}$, where
    $I_{\nu_j}=[a_{\nu_j},a_{\nu_j+1})$ if $\nu_j+1<n$ and
    $I_{\nu_j}=[a_{\nu_j},a_{\nu_j+1}]$ if $\nu_j+1=n$.
    Thus the $Q_\Bnu$ form a disjoint partition of $[0,1]^d$.
    Denote further by
    $f_\Bnu$ the value of $f$ at the midpoint of the cube
    $Q_\Bnu$. 
    Then with
  \begin{equation*}
    F_n(\Bx)\dfn \sum_{\Bnu\in\Lambda_n}f_\Bnu \cdot \ind_{Q_\Bnu}(\Bx),
  \end{equation*}
  we have
  \begin{equation*}
    \sup_{\Bx\in [0,1]^d}|f(\Bx)-F_n(\Bx)|\le L \frac{\sqrt{d}}{n}.
  \end{equation*}
  Since $n$ was arbitrary, to finish the proof it suffices to
  show that $F_n(\Bx)$ can be written as $g(\sum_{j=1}^d h_j(x_j))$.

  To each $\Bnu\in\{0,\dots,n-1\}^d$ we assign the unique number
  \begin{equation*}
    b_\Bnu \dfn \sum_{j=1}^d \nu_j\cdot n^{j-1}.
  \end{equation*}
  We remark that $\set{b_\Bnu}{\Bnu\in\Lambda_n}=\{0,\dots,n^{d}-1\}$. Define (the $f$-dependent function) $g:\R\to\R$ via
  \begin{equation}\label{eq:koldefg}
    g(x)\dfn
    f_\Bnu\qquad\text{if }x\in [b_\Bnu-1/4,b_\Bnu+1/4],
  \end{equation}
  for all $\Bnu\in\Lambda_n$,
  and continuously extended to all of $\R$.
  For $j=1,\dots,d$ define the monotonically increasing
  (and $f$-independent functions)
  \begin{equation*}
    h_j(x_j)\dfn
    \begin{cases}
      0 &\text{if }x_j<0\\
      n^{j-1}\cdot i &\text{if }x_j\in I_i,~0\le i<n\\
      n^{j} &\text{if }x_j>1.
    \end{cases}
  \end{equation*}
  Then for any $\Bx\in Q_\Bmu$
  \begin{equation*}
    h(\Bx)\dfn \sum_{j=1}^d h_j(x_j) = \sum_{j=1}^d \mu_j\cdot n^{j-1}=b_\Bmu,
  \end{equation*}
  so that by \eqref{eq:koldefg}
  \begin{equation*}
    g\circ h(\Bx) = g(b_\Bmu)=
    f_\Bmu = F_n(\Bx).
  \end{equation*}
  This concludes the proof.
  \end{proof}

\begin{corollary}
			Let $d\in\N$.
			With the activation function $\sigma:\R\to\R$ from
			Proposition~\ref{prop:magic}, for every compact $K\subseteq\R^d$, every
			$\eps>0$ and every $f\in C^0(K)$ there exists
			$\Phi\in\CN_d^1(\sigma;2,2d^2+d)$ (i.e.\ $\wdth(\Phi)=2d^2+d$ and
			$\depth(\Phi)=2$)
			such that
			\begin{align*}
				\sup_{\Bx\in K}|f(\Bx)-\Phi(\Bx)|<\eps.
			\end{align*}
		\end{corollary}

\begin{proof}
			Without loss of generality we can assume $K=[0,1]^d$: the extension to the general case
			then follows by Tietze's extension theorem and a scaling argument as
			in the proof of Proposition~\ref{prop:magic}.
			
			Let $f_j$, $\varphi_{i,j}$, $i=1,\dots,d$, $j=1,\dots,2d+1$ be as in
			Theorem~\ref{thm:kolmogorov}.
			Fix $\eps>0$.
			Let $a>0$ be so
			large that
			\begin{align*}
				\sup_{i,j}\sup_{x\in [0,1]}|\varphi_{i,j}(x)|\le a.
			\end{align*}
			Since each $f_j$ is uniformly continuous on the compact set
			$[-da,da]$, we can find $\delta>0$ such that
			\begin{align}\label{eq:fj_delta}
				\sup_j\sup_{\substack{|y-\tilde y|<\delta\\ |y|,|\tilde y|\le da}}|f_j(y)-f_j(\tilde y)|<
				\frac{\eps}{2(2d+1)}.
			\end{align}
			
			By Proposition~\ref{prop:magic} there exist $w_{i,j}$, $b_{i,j}\in\R$ such
			that
			\begin{align}\label{eq:varphiij_err}
				\sup_{i,j}\sup_{x\in [0,1]}|\varphi_{i,j}(x)-\underbrace{\sigma(w_{i,j}x+b_{i,j})}_{\dfnn \tilde\varphi_{i,j}(x)}|<\frac{\delta}{d}
			\end{align}
			and $w_{j}$, $b_{j}\in\R$ such that
			\begin{align}\label{eq:fj_err}
				\sup_{j}\sup_{|y|\le a+\delta}|f_j(y)-\underbrace{\sigma(w_{j}y+b_{j})}_{\dfnn \tilde f_j(y)}|<\frac{\eps}{2(2d+1)}.
			\end{align}
			
			Then for all $\Bx\in [0,1]^d$ by \eqref{eq:varphiij_err}
			\begin{align*}
				\left|\sum_{i=1}^d\varphi_{i,j}(x_i)
				-\sum_{i=1}^d\tilde \varphi_{i,j}(x_i)\right|
				< d\frac{\delta}{d}=\delta.
			\end{align*}
			Thus with
			\begin{align*}
				y_j\dfn \sum_{j=1}^d\varphi_{i,j}(x_i),\qquad
				\tilde y_j\dfn \sum_{j=1}^d\tilde\varphi_{i,j}(x_i)
			\end{align*}
			it holds $|y_j-\tilde y_j|<\delta$.
			Using \eqref{eq:fj_delta} and
			\eqref{eq:fj_err} we conclude
			\begin{align*}
				&\left|f(\Bx)-\sum_{j=1}^{2d+1}\sigma \left(w_{j} \cdot \left(\sum_{i=1}^d\sigma(w_{i,j}x_i+b_{i,j})\right)+b_{j} \right)\right|= \left|\sum_{j=1}^{2d+1}(f_j(y_j)-
				\tilde f_j(\tilde y_j))\right|\nonumber\\
				&\qquad\qquad\le
				\sum_{j=1}^{2d+1}\left(|f_j(y_j)-
				f_j(\tilde y_j)|+|f_j(\tilde y_j)-\tilde f_j(\tilde y_j)| \right)\nonumber\\
				&\qquad\qquad\le
				\sum_{j=1}^{2d+1}\left(\frac{\eps}{2(2d+1)}+\frac{\eps}{2(2d+1)}\right)\le \eps.
			\end{align*}
			This concludes the proof.
		\end{proof}

		Kolmogorov's superposition theorem is intriguing %
		as it shows that approximating $d$-dimensional functions
		can be reduced to the (generally much simpler) one-dimensional case through
		compositions. 
		Neural networks, by %
		nature, are well suited to approximate functions with compositional structures.
		However, as the proof of Proposition \ref{prop:kolmogorov} illustrates,
                the functions $f_j$
		in Theorem \ref{thm:kolmogorov}, even though only one-dimensional,
		could %
                become very complex and hard to approximate themselves if $d$ is large.

                Closely related to this construction,
		the ``magic'' activation function in
		Proposition~\ref{prop:magic} encodes the information of all rational
		polynomials on the unit interval, which is why a neural network of size
		$O(1)$ suffices to approximate every function to arbitrary accuracy.
		Naturally, no practical algorithm can efficiently determine appropriate neural network weights and biases for this
		architecture. 
		As such, the results presented in Section \ref{sec:kolmogorov}
		should be taken with a pinch of salt as their practical relevance is
		highly limited. 
		Nevertheless, they highlight that while universal
		approximation is a fundamental and important property of neural
		networks, it leaves many aspects unexplored. 
		To get further insight into practically relevant architectures,
                in the following chapters, we
                investigate neural networks with activation functions such as the ReLU.
		
		\section*{Bibliography and further reading}
                 The foundation of universal approximation
                  theorems goes back to the late 1980s with seminal
                  works by Cybenko \cite{Cybenko1989},
                  Hornik et al.\ \cite{HORNIK1989359,HORNIK1991251}, Funahashi
                  \cite{FUNAHASHI1989183}
                  and Carroll and Dickinson \cite{Carroll1989ConstructionON}. These results were
                  subsequently extended to a wider range of activation
                  functions and architectures. The present analysis in
                  Section \ref{sec:UniversalApproximationMainSection}
                  closely follows the arguments in
                  \cite{LESHNO1993861}, where it was essentially shown
                  that universal approximation can be achieved if the
                  activation function is not polynomial. The proof of
                  Lemma \ref{lemma:1tod} is from \cite[Theorem 2.1]{LIN1993295}, with
                  earlier results of this type being due to \cite{MR131106}.

                  Kolmogorov's superposition theorem stated in Theorem
                  \ref{thm:kolmogorov} was originally proven in 1957 
                  \cite{MR0111809}. For a more recent and constructive
                  proof see for instance
                  \cite{Braun2009}. Kolmogorov's theorem and its
                  obvious connections to neural networks have inspired
                  various research in this field, e.g.\
                  \cite{hechtnielsen:kolmogorov,KURKOVA1992501,MONTANELLI20201,SCHMIDTHIEBER2021119,ISMAILOV2023127096},
                  with its practical relevance being debated
                  \cite{irrelevant,relevant}. The idea for the
                  ``magic'' activation function in Section
                  \ref{sec:kolmogorov} comes from
                  \cite{MAIOROV199981} where it is shown that such an activation function can even be chosen %
                  monotonically increasing.

		\newpage
		\section*{Exercises}

		\begin{exercise}\label{ex:comptop}
			Write down a generator of a (minimal) topology on $C^0(\R^d)$ such
			that $f_n\to f\in C^0(\R^d)$ if and only if $f_n\tocc f$, and show
			this equivalence.
			This topology is referred to as the
			topology of compact convergence.
		\end{exercise}
		
		\begin{exercise}\label{ex:ra}
			Show the implication ``$\Rightarrow$'' of Theorem
			\ref{thm:universal} and Corollary \ref{cor:universaldeep}.
		\end{exercise}
		
		\begin{exercise}\label{ex:sigmoidalStepfun1d}
			Prove Lemma \ref{lemma:sigmoidal}.
			\emph{Hint}: Consider $\sigma(nx)$ for
			large $n\in\N$.
		\end{exercise}
		
		\begin{exercise}\label{ex:completeProofOfMpoly}
			Let $k\in\N$, $\sigma\in\CM$ and assume that
			$\sigma *\varphi\in \CP_k$ for all $\varphi\in
			C_c^\infty(\R)$.
			Show that $\sigma\in\CP_k$.
			
			\emph{Hint}: Consider
			$\psi\in C_c^\infty(\R)$ such that $\psi\ge 0$ and
			$\int_\R\psi(x)\dd x =1$ and set
			$\psi_\eps(x):=\psi(x/\eps)/\eps$.
			Use that away from the
			discontinuities of $\sigma$ it holds
			$\psi_\eps *\sigma(x)\to\sigma(x)$ as $\eps\to 0$.
			Conclude
			that $\sigma$ is piecewise in $\CP_k$, and finally show
			that $\sigma\in C^{k}(\R)$.
			
		\end{exercise}

		\begin{exercise}\label{ex:universalInLp}
                  Prove Corollary \ref{cor:universal} with the use of Corollary \ref{cor:universaldeep}.
		\end{exercise}

		\begin{exercise}
			\label{ex:ExtendToDeep}
			Complete the proof of Proposition \ref{prop:Identity1} for $L>1$.	
		\end{exercise}

%% file: Splines.tex
\chapter{Splines}\label{chap:Splines}

In Chapter \ref{chap:UA}, we saw that sufficiently large neural networks can approximate every continuous function to arbitrary accuracy. However, these results %
did not further specify the meaning of ``sufficiently large'' or what constitutes a %
suitable architecture. Ideally, given a function $f$, and a desired accuracy $\eps>0$, we would like to have a (possibly sharp) bound on the required size, depth, and width guaranteeing the existence of a neural network approximating $f$ up to error $\eps$.

The field of approximation theory establishes such trade-offs between properties of the function $f$ (e.g., its smoothness), the approximation accuracy, and the number of parameters needed to achieve this accuracy. For example, given $k$, $d\in\N$, how many parameters are required to approximate a function $f:[0,1]^d\to\R$ with $\norm[C^k({[0,1]^d})]{f}\le 1$ up to uniform error $\eps$? Splines are known to achieve this approximation accuracy with a superposition of $O(\eps^{-d/k})$ simple (piecewise polynomial) basis functions. In this chapter, following \cite{MR1176581}, we show that certain sigmoidal neural networks can match this performance in terms of the neural network size. In fact, from an approximation theoretical viewpoint we show that the considered neural networks are at least as expressive as superpositions of splines.

\section{B-splines and smooth functions}

We introduce a simple type of spline and its approximation properties below. 

\begin{definition}\label{def:CardiBSpline}
  For $n \in \N$, the \textbf{univariate cardinal B-spline} %
  of order $n\in \N$ is given by
 \begin{align}\label{eq:UnivCardBSpline}
	\CS_n(x) \coloneqq \frac{1}{(n-1)!}  \sum_{\ell = 0}^{n} (-1)^\ell \binom{n}{\ell} \sigma_{\rm ReLU}(x-\ell)^{n-1} \qquad \text{ for } x \in \R,    
\end{align}
where %
$0^0 \dfn 0$ and $\sigma_{\rm ReLU}$ %
denotes the %
ReLU activation function. 
\end{definition}

By shifting and dilating the cardinal B-spline, we obtain a system of univariate splines.
Taking tensor products of these univariate splines yields a set of higher-dimensional functions known as the multivariate B-splines.

\begin{definition}\label{def:multivariateBsplines} For $t \in \R$ and $n$, $\ell \in \N$ we define $\CS_{\ell, t, n} \coloneqq \CS_n (2^{\ell}(\cdot - t))$. 
Additionally, %
for $d \in \N$, $ \Bt \in \R^d$, and $n$, $\ell \in \N$, 
we define the \textbf{the multivariate B-spline} $\CS_{\ell, \Bt, n}^d$ as
\[
	\CS_{\ell, \Bt, n}^d(\Bx) \coloneqq \prod_{i = 1}^d \CS_{\ell, t_i, n}(x_i)\qquad \text{ for } \Bx = (x_1, \dots x_d) \in \R^d,
\]	
and
\[
	\mathcal{B}^n \coloneqq \setc{\CS_{\ell, \Bt, n}^d}{ \ell \in \N, \Bt \in \R^d}
      \]
      is the \textbf{dictionary of B-splines of order $n$}.
    \end{definition}

Having introduced the system $\mathcal{B}^n$, we would like to understand how well we can represent each smooth function by superpositions of elements of $\mathcal{B}^n$.
  The following theorem is adapted from the more general result \cite[Theorem 7]{oswald1990degree}; also see \cite[Theorem D.3]{Marzouk2023Distribution} for a presentation closer to the present formulation.

\begin{theorem}%
  \label{thm:OswaldTheorem}
	Let $d$, $n$, $k \in \N$ such that $0< k \leq n$. 
	Then there exists $C$ such that for
        every $f \in C^k([0,1]^d)$ and every
        $N \in \N$,
        there exist $c_i \in \R$ with $|c_i| \leq C \norm[{L^\infty([0,1]^d)}]{f}$ and
        $B_i \in \mathcal{B}^n$ for $i = 1, \dots, N$, such that
        \[
          \left\|f - \sum_{i=1}^N c_i B_i \right\|_{L^\infty([0,1]^d)} \leq C N^{-\frac{k}{d}} \|f \|_{C^k[0,1]^d}.
        \]
      \end{theorem}

\begin{remark}\label{rmk:depthorder}
  There are a couple of critical concepts in Theorem \ref{thm:OswaldTheorem} that will reappear throughout this book.
  The number of parameters $N$
  determines the approximation accuracy %
  $N^{-k/d}$.
This implies that %
achieving accuracy $\eps>0$ %
requires $O(\eps^{-d/k})$ parameters (according to this upper bound),
which grows exponentially in $d$.
This %
exponential dependence on $d$ is referred to as the ``curse of dimension'' and will be discussed %
again in the subsequent chapters.
The smoothness parameter $k$ %
has the opposite effect of $d$, and improves the convergence rate.
Thus, smoother functions can be approximated with %
fewer B-splines than rougher functions.
This more efficient approximation requires the use of B-splines of %
order $n$ with $n\ge k$.
We will see in the following, that the order of the B-spline is closely linked to the concept of depth in neural networks.
\end{remark}

\section{Reapproximation of B-splines with sigmoidal activations}\label{sec:reapproxDNNSplines}

We now show that the approximation rates of B-splines can be transferred to certain neural networks. The %
following argument is based on \cite{mhaskar1993approximation}. 

\begin{definition}\label{def:higherOrderSigmoidal}
	A function $\sigma: \R \to \R$ is called \textbf{sigmoidal of order $q\in \N$}, if 
	$\sigma \in C^{q-1}(\R)$ and there exists $C >0$ such that 
	\begin{align*}
          \frac{\sigma(x)}{x^q} &\to 0 &&\text{as } x\to - \infty,\\
          \frac{\sigma(x)}{x^q} &\to 1 &&\text{as } x\to \infty,\\
          |\sigma(x)| &\leq C \cdot (1+|x|)^q &&\text{for all } x \in \R.
	\end{align*}
\end{definition}

\begin{example}
  The rectified power unit $x\mapsto\sigma_{\rm ReLU}(x)^q$ is sigmoidal of order $q$.
\end{example}

Our goal in the following is to show that neural networks %
can approximate %
a linear combination of $N$ B-splines with a number of parameters that is proportional to $N$. As an immediate consequence of Theorem \ref{thm:OswaldTheorem}, we then obtain a convergence rate for neural networks. Let us start by approximating a single univariate B-spline with a neural network of fixed size.

\begin{proposition}\label{prop:reapproxBspline}
	Let $n\in \N$, $n \geq 2$, $K>0$, and let $\sigma: \R \to \R$ be sigmoidal of order $q\geq 2$. 
	There exists a constant $C>0$ such that for every $\eps >0$ there is a neural network $\Phi^{	\CS_n}$ with activation function $\sigma$, $\lceil\log_{q}(n-1)\rceil$ layers, and size $C$, such that 
	\[
		\left\|\CS_n - \Phi^{	\CS_n}\right\|_{L^\infty([-K,K])} \leq \eps.
	\]
\end{proposition}

\begin{proof}
  By definition \eqref{eq:UnivCardBSpline}, $\CS_n$
  is a %
  linear combination of %
  $n+1$
  shifts of $\sigma_{\rm ReLU}^{n-1}$. 
  We start by approximating $\sigma_{\rm ReLU}^{n-1}$.
It is not hard to see (Exercise \ref{ex:convergenceOfTowerOfSigmoids}) that, for every $K'>0$ and every $t \in \N$
\begin{align}\label{eq:towerOfSigmoids}
	\left|a^{-q^t} \underbrace{\sigma \circ \sigma \circ \dots \circ \sigma(a x)}_{t-\text{ times}}  - \sigma_{\rm ReLU}(x)^{q^t} \right|\to 0 \qquad\text{as } a \to \infty
\end{align}
uniformly for all $x \in [-K',K']$.

Set $t \coloneqq \lceil\log_{q}(n-1)\rceil$. Then $t \geq 1$ since $n \geq 2$, and $q^t \geq n-1$.
Thus, for every $K'>0$ and $\eps >0$ there exists a neural network $\Phi^{q^t}_\eps$ %
with $\lceil\log_{q}(n-1)\rceil$ layers %
satisfying
\begin{align}\label{eq:theDerivativeTrick}
  \left|\Phi^{q^t}_\eps(x) - \sigma_{\rm ReLU}(x)^{q^t}\right| \leq \eps
  \qquad\text{for all }x \in [-K', K'].
\end{align}
This shows that we can approximate the ReLU to the power of $q^t\ge n-1$.
However, %
  our goal is to obtain an approximation of the ReLU raised %
  to the power $n-1$, which could be smaller than $q^t$.
To reduce the order, we emulate approximate derivatives of $\Phi^{q^t}_\eps$.
Concretely, we show the following claim: For all $1\leq p \leq q^t$ for every $K'>0$ and $\eps >0$ there exists a neural network $\Phi^{p}_\eps$ having $\lceil\log_{q}(n-1)\rceil$ layers and satisfying
\begin{align}\label{eq:allp}
  \left|\Phi^{p}_\eps(x) - \sigma_{\rm ReLU}(x)^{p}\right| \leq \eps
  \qquad\text{for all }x\in [-K',K'].
\end{align}
The claim holds for $p = q^t$. We now proceed by induction over $p=q^t,q^t-1,\dots$
Assume \eqref{eq:allp} holds for some $p\in\{2,\dots,q^t\}$.
Fix $\delta \geq 0$. %
Then
\begin{align*}
	&\left|\frac{\Phi^{p}_{\delta^2}(x + \delta) - \Phi^{p}_{\delta^2}(x )}{p \delta} - \sigma_{\rm ReLU}(x)^{p-1}\right|\\
	&\qquad  \leq 2\frac{\delta}{p} + \left|\frac{\sigma_{\rm ReLU}(x + \delta)^{p} -  \sigma_{\rm ReLU}(x)^{p}}{p \delta} - \sigma_{\rm ReLU}(x)^{p-1}\right|.
\end{align*} %
Hence, by the binomial theorem it follows that there exists $\delta_{*}>0$ such that
\begin{align*}
	\left|\frac{\Phi^{p}_{\delta_{*}^2}(x + \delta_{*}) - \Phi^{p}_{\delta_{*}^2}(x)}{p \delta_{*}} -  \sigma_{\rm ReLU}(x)^{p-1}\right| \leq \eps,
\end{align*}
for all $x \in [-K', K']$. 
By Proposition \ref{prop:VectorSpaceAndCompositions}, $(\Phi^{p}_{\delta_{*}^2}(x + \delta_{*}) - \Phi^{p}_{\delta_{*}^2})/(p \delta_{*})$ is a neural network with $\lceil\log_{q}(n-1)\rceil$ layers and size independent from $\eps$. 
Calling this neural network $\Phi^{p-1}_{\eps}$ shows that %
\eqref{eq:allp} holds for $p-1$, which concludes the induction argument and proves the claim.

For every neural network $\Phi$, %
every spatial translation $\Phi(\cdot-t)$ %
is a neural network %
of the same architecture.
Hence, %
every term in the sum \eqref{eq:UnivCardBSpline}
can be approximated to arbitrary accuracy by a %
neural network of a fixed size.
Since by Proposition \ref{prop:VectorSpaceAndCompositions}, sums of neural networks %
of the same depth are again neural networks %
of the same depth, the result follows.
\end{proof}

Next, we extend Proposition \ref{prop:reapproxBspline} to the multivariate splines $\CS_{\ell, \Bt, n}^d$ for arbitrary $\ell$, $d \in \N$, $\Bt \in \R^d$.

\begin{proposition}\label{prop:reapproxBsplinehighdim}
	Let $n$, $d\in \N$, $n \geq 2$, $K>0$, and let $\sigma: \R \to \R$ be sigmoidal of order $q\geq 2$. 
	Further let $\ell \in \N$ and $\Bt \in \R^d$.
	
	Then, there exists a constant $C>0$ such that for every $\eps >0$ there is a neural network $\Phi^{	\CS_{\ell, \Bt, n}^d}$ with activation function $\sigma$, $ \lceil \log_2(d)\rceil + \lceil\log_{q}(n-1)\rceil$ layers, and size $C$, such that 
	\[
	\left\|\CS_{\ell, \Bt, n}^d  - \Phi^{\CS_{\ell, \Bt, n}^d}\right\|_{L^\infty([-K,K]^d)} \leq \eps.
	\]
\end{proposition}

\begin{proof}
  By definition $\CS_{\ell, \Bt, n}^d(\Bx)=\prod_{i=1}^d\CS_{\ell,t_i,n}(x_i)$ where
    $$\CS_{\ell, t_i, n}(x_i)=  \CS_{n}(2^\ell(x_i-t_i)).$$
  By Proposition \ref{prop:reapproxBspline}
there exist a constant $C'>0$ such that for each $i = 1, \dots, d$ and all $\eps >0$, there is a neural network $\Phi^{\CS_{\ell, t_i, n}}$ with size $C'$ and $\lceil \log_q(n-1)\rceil$ layers such that 
\[
	\left\|\CS_{\ell, t_i, n}- \Phi^{\CS_{\ell, t_i, n}}\right\|_{L^\infty([-K,K]^d)} \leq \eps.
\]
If $d=1$, this shows the statement. For general $d$, %
it remains to show that the product
of the $\Phi^{\CS_{\ell, t_i, n}}$ for $i = 1, \dots, d$ can be approximated.

We first prove the following claim by induction: For every $d\in \N$, $d \geq 2$, there exists a constant $C''>0$, such that for all ${K'} \geq 1$ and all $\eps>0$ there exists a neural network
$\Phi_{{\rm mult},\eps,d}$
with size $C''$, $\lceil \log_2(d) \rceil$ layers, and activation function $\sigma$ such that for all $x_1, \dots, x_d$ with $|x_i| \leq {K'}$ for all $i = 1, \dots, d$, 
\begin{align}\label{eq:inductionClaimMultiplicationSigmoidal}
	\left|\Phi_{{\rm mult}, \eps, d}(x_1, \dots, x_d) - \prod_{i =1}^d x_i\right| < \eps.
\end{align}
For the base case, let $d = 2$.
Similar to the proof of Proposition \ref{prop:reapproxBspline}, one can show that there exists $C'''>0$ such that for every $\eps>0$ and $K'>0$ there exists a neural network $\Phi_{{\rm square}, \eps}$ with one hidden layer and size $C'''$ such that
\begin{align*}
	|\Phi_{{\rm square}, \eps} - \sigma_{\rm ReLU}(x)^2| \leq \eps \qquad\text{for all } |x| \leq {K'}.
\end{align*}
For every $x = (x_1,x_2) \in \R^2$
\begin{align}
	x_1 x_2 &= \frac{1}{2} \left((x_1+x_2)^2 - x_1^2 - x_2^2\right) \nonumber\\
	&=  \frac{1}{2} \left(\sigma_{\rm ReLU}(x_1+x_2)^2 + \sigma_{\rm ReLU}(-x_1-x_2)^2 - \sigma_{\rm ReLU}(x_1)^2 \right.\nonumber\\
	&\qquad - \left.\sigma_{\rm ReLU}(-x_1)^2 - \sigma_{\rm ReLU}(x_2)^2 - \sigma_{\rm ReLU}(-x_2)^2\right).\label{eq:multiplicationFromSquares}
\end{align}
Each term on the right-hand side can be approximated up to uniform error $\eps/6$ with a network of size $C'''$ and one hidden layer.
By Proposition \ref{prop:VectorSpaceAndCompositions}, we conclude that there exists a neural network $\Phi_{{\rm mult}, \eps, 2}$ satisfying %
\eqref{eq:inductionClaimMultiplicationSigmoidal} for $d=2$.

Assume the induction hypothesis \eqref{eq:inductionClaimMultiplicationSigmoidal} holds for $d-1\ge 1$, and let
$\eps >0$ and ${K'}\geq 1$. %
We have
\begin{align}\label{eq:adfiugasuih}
		\prod_{i =1}^d x_i = \prod_{i = 1}^{\lfloor d/2 \rfloor} x_i  \cdot \prod_{i = \lfloor d/2 \rfloor + 1}^{d} x_i.
\end{align}
We will now approximate each of the terms in the product on the right-hand side of \eqref{eq:adfiugasuih} by a neural network using the induction assumption. 

For simplicity assume in the following that $\lceil\log_2(\lfloor d/2 \rfloor)\rceil = \lceil\log_2(d - \lfloor d/2 \rfloor)\rceil$. The general case can be addressed via Proposition \ref{prop:Identity1}. By the induction assumption there then exist neural networks $\Phi_{{\rm mult}, 1}$ and $\Phi_{{\rm mult}, 2}$ both with $\lceil\log_2(\lfloor d/2 \rfloor)\rceil$ layers, such that for all $x_i$ with $|x_i| \leq {K'}$ for $i = 1, \dots, d$
\begin{align*}
	\left|\Phi_{{\rm mult}, 1}(x_1, \dots, x_{\lfloor d/2 \rfloor}) - \prod_{i = 1}^{\lfloor d/2 \rfloor} x_i\right|  &< \frac{\eps}{4 (({K'})^{\lfloor d/2 \rfloor} + \eps)}, \\
	\left|\Phi_{{\rm mult}, 2}(x_{\lfloor d/2 \rfloor+1}, \dots, x_{d}) - \prod_{i = \lfloor d/2 \rfloor + 1}^{d} x_i\right|  &< \frac{\eps}{4 (({K'})^{\lfloor d/2 \rfloor} + \eps)}.
\end{align*}
By Proposition \ref{prop:VectorSpaceAndCompositions}, %
$\Phi_{{\rm mult}, \eps, d} \coloneqq \Phi_{{\rm mult}, \eps/2, 2} \circ  (\Phi_{{\rm mult}, 1}, \Phi_{{\rm mult}, 2})$ is a neural network with $1 + \lceil\log_2(\lfloor d/2 \rfloor)\rceil = \lceil\log_2(d)\rceil$ layers.
By construction, the size of $\Phi_{{\rm mult},\eps,d}$ does not depend on $K'$ or $\eps$.
Thus, to complete the induction, %
it only remains to show \eqref{eq:inductionClaimMultiplicationSigmoidal}. 

For all $a$, $b$, $c$, $d \in \R$ holds
\[
	|ab - cd| \leq |a| |b-d| + |d| |a-c|.
\]
Hence, for $x_1, \dots, x_d$ with $|x_i| \leq {K'}$ for all $i = 1, \dots, d$, we have that
\begin{align*}
	&\left|\prod_{i =1}^d x_i - \Phi_{{\rm mult}, \eps, d}(x_1, \dots, x_d)\right| \\
	&\leq \frac{\eps}{2} + \left|\prod_{i = 1}^{\lfloor d/2 \rfloor} x_i  \cdot \prod_{i = \lfloor d/2 \rfloor + 1}^{d} x_i - \Phi_{{\rm mult}, 1} (x_1, \dots, x_{\lfloor d/2 \rfloor}) \Phi_{{\rm mult}, 2} (x_{\lfloor d/2 \rfloor+1}, \dots, x_d ) \right|\\
	&\leq \frac{\eps}{2} + |{K'}|^{\lfloor d/2 \rfloor} \frac{\eps}{4 (({K'})^{\lfloor d/2 \rfloor} + \eps)}  + (|{K'}|^{\lceil d/2 \rceil} + \eps) \frac{\eps}{4 (({K'})^{\lfloor d/2 \rfloor} + \eps)} < \eps.
\end{align*}
This completes the proof of \eqref{eq:inductionClaimMultiplicationSigmoidal}. 

The overall result follows by using Proposition \ref{prop:VectorSpaceAndCompositions} to show that the multiplication network can be composed with a neural network comprised of the $\Phi^{\CS_{\ell, t_i, n}}$ for $i = 1, \dots, d$. 
Since in no step above the size of the individual networks was dependent on the approximation accuracy, this is also true for the final network. 
\end{proof}

Proposition \ref{prop:reapproxBsplinehighdim} shows that we can approximate a single multivariate B-spline with a neural network with a size that is independent %
of the accuracy.
Combining this observation with Theorem \ref{thm:OswaldTheorem} leads to the following result.

\begin{theorem}\label{thm:SmoothFunctionapproximationBysigmoidalNNs}
  Let $d$, $n$, $k \in \N$ such that $0< k \leq n$ and $n\ge 2$.
  Let $q \geq 2$, and let $\sigma$ be sigmoidal of order $q$.
  
	Then there exists $C$ such that for
        every $f \in C^k([0,1]^d)$ and every
        $N \in \N$ there exists a neural network $\Phi^N$ with
        activation function $\sigma$, $\lceil \log_2(d)\rceil + \lceil\log_{q}(k-1)\rceil$ layers, and size bounded by $CN$, such that
		\[
			\left\|f - \Phi^N \right\|_{L^\infty([0,1]^d)} \leq C N^{-\frac{k}{d}}\norm[{C^k([0,1]^d)}]{f}.
		\]
      \end{theorem}

\begin{proof}
Fix $N\in\N$. %
By Theorem \ref{thm:OswaldTheorem}, %
there exist coefficients $|c_i|\leq C\norm[{L^\infty([0,1]^d)}]{f}$ and $B_i \in \mathcal{B}^n$ for $i = 1, \dots, N$, such that 
\begin{align*}
	\left\|f - \sum_{i=1}^N c_i B_i \right\|_{L^\infty([0,1]^d)} \leq C N^{-\frac{k}{d}} \|f \|_{C^k([0,1]^d)}.
\end{align*}
Moreover, by Proposition \ref{prop:reapproxBsplinehighdim}, %
for each $i = 1, \dots, N$ exists a neural network $\Phi^{B_i}$ with $\lceil \log_2(d)\rceil + \lceil\log_{q}(k-1)\rceil$ layers, and a fixed size, which approximates $B_i$ on $[-1,1]^d \supseteq [0,1]^d$ up to error of $\eps\dfn N^{-k/d}/N$. 
The size of $\Phi^{B_i}$ is independent of $i$ and $N$.

By Proposition \ref{prop:VectorSpaceAndCompositions}, %
there exists a neural network
$\Phi^N$ that uniformly approximates $\sum_{i=1}^N c_i B_i$ up to error $\eps$ on $[0,1]^d$, and has $\lceil \log_2(d)\rceil + \lceil\log_{q}(k-1)\rceil$ layers. 
The size of this network is linear in $N$ (see Exercise \ref{ex:numberOfWeightsOfASum}).
This concludes the proof.
\end{proof}

Theorem \ref{thm:SmoothFunctionapproximationBysigmoidalNNs} shows that neural networks with higher-order sigmoidal functions can approximate smooth functions with the same accuracy as spline approximations while having a comparable number of parameters.
The network depth is required to behave like $O(\log(k))$ in terms of the smoothness parameter $k$, cp.~Remark \ref{rmk:depthorder}.

\section*{Bibliography and further reading}
The argument of linking sigmoidal activation functions with spline based approximation was first introduced in \cite{MR1176581,mhaskar1993approximation}. For further details on spline approximation, see \cite{oswald1990degree} or the book \cite{Schumaker_2007}.

The general strategy of approximating basis functions by neural networks, and then lifting approximation results for those bases %
has been employed widely in the literature, and will also reappear again in this book.
While the following chapters primarily focus on ReLU activation, we highlight a few notable approaches with non-ReLU activations based on the outlined strategy: To approximate analytic functions, \cite{mhaskar1996neural} emulates a monomial basis. To approximate periodic functions, a basis of trigonometric polynomials is recreated in \cite{mhaskar1995degree}. Wavelet bases have been emulated in \cite{pati1993analysis}.
Moreover, neural networks have been studied through the representation system of ridgelets \cite{candes1998ridgelets} %
and ridge functions \cite{ismailov2021ridge}. A general framework describing the emulation of representation systems to %
transfer approximation results was %
presented in \cite{bolcskei2019optimal}.

\newpage
\section*{Exercises}

\begin{exercise}\label{ex:convergenceOfTowerOfSigmoids}
	Show that \eqref{eq:towerOfSigmoids} holds. 
\end{exercise}

\begin{exercise}\label{ex:numberOfWeightsOfASum}
	Let $L \in \N$, $\sigma \colon \R \to \R$, and let $\Phi_1$, $\Phi_2$ be two neural networks with architecture $(\sigma; d_0, d_1^{(1)}, \dots, d_{L}^{(1)}, d_{L+1})$ and $(\sigma; d_0, d_1^{(2)}, \dots, d_{L}^{(2)}, d_{L+1})$. 
	Show that $\Phi_1 + \Phi_2$ is a neural network with $\size(\Phi_1 + \Phi_2) \leq \size(\Phi_1) + \size(\Phi_2)$.
\end{exercise}

\begin{exercise}
	Show that, for $\sigma = \sigma_{\rm ReLU}^2$ and $k \leq 2$, for all $f \in C^{k}([0,1]^d)$ all weights of the approximating neural network of Theorem \ref{thm:SmoothFunctionapproximationBysigmoidalNNs} can be bounded in absolute value by $O(\max\{2, \|f \|_{C^k([0,1]^d)}\})$.
\end{exercise}

%% file: ReLUNNs.tex
\chapter{ReLU neural networks}\label{chap:ReLUNNs}
In this chapter, we discuss feedforward neural networks using the ReLU
activation function $\sigma_{\rm ReLU}$ %
introduced in Section \ref{sec:activationFunctions}. We refer to these
functions as ReLU neural networks.  Due to its simplicity and the fact
that it reduces the vanishing and exploding gradients phenomena, %
the ReLU is one of the most widely used activation functions in
practice.

A key component of the proofs in the previous chapters was the
approximation of derivatives of the activation function to %
emulate polynomials.  Since the ReLU is piecewise linear, this trick
is not applicable.  This makes the analysis fundamentally different
from the case of smoother activation functions.  Nonetheless, we will
see that even this extremely simple activation function yields a very
rich class of functions possessing remarkable approximation
capabilities.

To formalize these results, we begin this chapter by adopting a
framework from \cite{petersen2018optimal}, which enables the tracking
of the number of network parameters for basic manipulations such as
adding up or composing two neural networks. This will allow to bound
the network complexity, when constructing more elaborate networks from
simpler ones.  With these preliminaries at hand, the rest of the
chapter is dedicated to the exploration of links between ReLU neural
networks and the class of %
``continuous piecewise linear functions''.  In Section
\ref{sec:CWPLfunctionsandRepresentations}, we will see that every such
function can be exactly represented by a ReLU neural network.
Afterwards, in Section \ref{sec:simplicialPieces} we will give a more
detailed analysis of the required network complexity.  Finally, we
will use %
these results to prove a first approximation theorem for ReLU neural
networks in Section \ref{sec:HoelderRates}.  The argument is similar
in spirit to Chapter \ref{chap:Splines}, in that we \emph{transfer}
established approximation theory for piecewise linear functions to the
class of ReLU neural networks of a certain architecture.

\section{Basic ReLU calculus}\label{sec:basic_relu}
The goal of this section is to %
formalize how to combine and manipulate ReLU neural networks.  We have
seen an instance of such a %
result already in Proposition \ref{prop:VectorSpaceAndCompositions}.
Now we want to make this result more precise under the assumption that
the activation function is the ReLU.  We sharpen Proposition
\ref{prop:VectorSpaceAndCompositions} by adding bounds on the number
of weights that the resulting neural networks have.  The following
four operations form the basis of all constructions in the sequel.
\begin{itemize}
\item \textit{Reproducing an identity:} We have seen in Proposition
  \ref{prop:Identity1} that for most activation functions, an
  approximation to the identity can be built by neural networks.  For
  ReLUs, we can have an even stronger result and reproduce the
  identity exactly.  This identity will play a crucial role in order
  to extend certain neural networks to deeper neural networks, and to
  facilitate an efficient composition operation.
\item \textit{Composition:} We saw in Proposition
  \ref{prop:VectorSpaceAndCompositions} that we can produce a
  composition of two neural networks and the resulting function is a
  neural network as well.  There we did not study the size of the
  resulting neural networks.  For ReLU activation functions, this
  composition can be done in a very efficient way leading to a neural
  network that has up to a constant not more than the number of
  weights of the two initial neural networks.
\item \textit{Parallelization:} Also the parallelization of two neural
  networks was discussed in Proposition
  \ref{prop:VectorSpaceAndCompositions}.  We will refine this notion
  and make precise the size of the resulting neural networks.
\item \textit{Linear combinations:} Similarly, for the sum of two
  neural networks, we will give precise bounds on the size of the
  resulting neural network.
\end{itemize}

\subsection{Identity}
We start with expressing the identity on $\R^d$ as a neural network of
depth $L\in\N$.

\begin{lemma}[Identity]\label{lemma:identity}
  Let $L\in\N$.  Then, there exists a ReLU neural network $\nId{L}$
  such that $\nId{L}(\Bx)=\Bx$ for all $\Bx\in\R^d$.  Moreover,
  $\depth(\nId{L})=L$, $\wdth(\nId{L})=2d$, and
  $\size(\nId{L})=2d\cdot (L+1)$.
\end{lemma}

\begin{proof}
  Writing $\BI_d\in\R^{d\times d}$ for the identity matrix, we choose
  the weights
  \begin{align*}
    &(\BW^{(0)},\Bb^{(0)}),\dots,(\BW^{(L)},\Bb^{(L)}) \\
    &\qquad 
      \dfn \left(\begin{pmatrix}\BI_d\\ -\BI_d\end{pmatrix},\Bnul\right),
      \underbrace{(\BI_{2d},\Bnul),\dots,(\BI_{2d},\Bnul)}_{L-1\text{ times}},
      ((\BI_d,-\BI_d),\Bnul)
      .
  \end{align*}

  Using that $x=\sigma_{\rm ReLU}(x)-\sigma_{\rm ReLU}(-x)$ for all
  $x\in\R$ and $\sigma_{\rm ReLU}(x)=x$ for all $x\ge 0$ it is obvious
  that the neural network $\nId{L}$ associated to the weights above
  satisfies the assertion of the lemma.
\end{proof}

We will see in Exercise \ref{ex:polynomialsIdentityExact} that the
property to exactly represent the identity is not shared by sigmoidal
activation functions.  It does hold for polynomial activation
functions though; also see Proposition \ref{prop:Identity1}.

\subsection{Composition}
Assume we have two neural networks $\Phi_1$, $\Phi_2$ with corresponding architectures $(\sigma_{\rm ReLU};d_0^1, \dots, d_{L_1+1}^1)$ and
$(\sigma_{\rm ReLU};d_0^2, \dots, d_{L_1+1}^2)$ respectively.
Moreover, we assume that they have weights and biases given by
\begin{align*}
  (\BW^{(0)}_1,\Bb^{(0)}_1),\dots,(\BW^{(L_1)}_1,\Bb^{(L_1)}_1), \text{ and } (\BW^{(0)}_2,\Bb^{(0)}_2),\dots,(\BW^{(L_2)}_2,\Bb^{(L_2)}_2),
\end{align*}
respectively.  If the output dimension $d^1_{L_1+1}$ of $\Phi_1$
equals the input dimension $d_0^2$ of $\Phi_2$, we can define two
types of concatenations: First $\Phi_2\circ\Phi_1$ is the neural
network with weights and biases given by
\begin{align*}
  &\left(\BW^{(0)}_1,\Bb^{(0)}_1\right),\dots,\left(\BW^{(L_1-1)}_1,\Bb^{(L_1-1)}_1\right),\left(\BW^{(0)}_2\BW^{(L_1)}_1,\BW^{(0)}_2 \Bb^{(L_1)}_1+\Bb_2^{(0)}\right),\\
  & \qquad
    \left(\BW^{(1)}_2,\Bb^{(1)}_2\right),\dots,\left(\BW^{(L_2)}_2,\Bb^{(L_2)}_2\right).
\end{align*}

Second, $\Phi_2\bullet\Phi_1$ is the neural network defined as
$\Phi_2\circ \nId{1}\circ \Phi_1$.  In terms of weights and biases,
$\Phi_2\bullet\Phi_1$ is given as
\begin{align*}
  &
    \left(\BW^{(0)}_{1},\Bb^{(0)}_{1}\right),\dots,\left(\BW^{(L_1-1)}_1,\Bb^{(L_1-1)}_1\right),
    \left(\begin{pmatrix}
            \BW^{(L_1)}_1\\
            -\BW^{(L_1)}_1
          \end{pmatrix},
    \begin{pmatrix}
      \Bb^{(L_1)}_1\\
      -\Bb^{(L_1)}_1
    \end{pmatrix}\right),\nonumber\\
  &\quad\left(\left(\BW^{(0)}_2,-\BW^{(0)}_2\right),\Bb^{(0)}_2\right),
    \left(\BW^{(1)}_2,\Bb^{(1)}_2\right),
    \dots,\left(\BW^{(L_2)}_2,\Bb^{(L_2)}_2\right).
\end{align*}
The following lemma collects the properties of the constructions
above.

\begin{lemma}[Composition]\label{lemma:composition}
  Let $\Phi_1$, $\Phi_2$ be neural networks with architectures
  $(\sigma_{\rm ReLU};d_0^1, \dots, d_{L_1+1}^1)$ and
  $(\sigma_{\rm ReLU};d_0^2, \dots, d_{L_2+1}^2)$.  Assume
  $d_{L_1+1}^1=d_0^2$.  Then
  $\Phi_2\circ\Phi_1(\Bx)= \Phi_2\bullet\Phi_1(\Bx)=
  \Phi_2(\Phi_1(\Bx))$ for all $\Bx\in\R^{d^0_1}$.  Moreover,
  \begin{align*}
    \wdth(\Phi_2\circ\Phi_1) &\le \max\{\wdth(\Phi_1),\wdth(\Phi_2)\},\\
    \depth(\Phi_2\circ\Phi_1) &= \depth(\Phi_1)+\depth(\Phi_2),\\
    \size(\Phi_2\circ\Phi_1) &\le \size(\Phi_1)+\size(\Phi_2)+(d_{L_1}^1+1)d^1_2,
  \end{align*}
  and
  \begin{align*}
    \wdth(\Phi_2\bullet\Phi_1) &\le 2\max\{\wdth(\Phi_1),\wdth(\Phi_2)\},\\
    \depth(\Phi_2\bullet\Phi_1) &= \depth(\Phi_1)+\depth(\Phi_2)+1,\\
    \size(\Phi_2\bullet\Phi_1) &\le
                                 2(\size(\Phi_1)+\size(\Phi_2)).
  \end{align*}
\end{lemma}

\begin{proof}
  The fact that
  $\Phi_2\circ\Phi_1(\Bx)= \Phi_2\bullet\Phi_1(\Bx)=
  \Phi_2(\Phi_1(\Bx))$ for all $\Bx\in\R^{d_0^1}$ follows immediately
  from the construction.  The same can be said for the width and depth
  bounds.  To confirm the size bound, we note that
  $\BW_2^{(0)}\BW_1^{(L_1)}\in\R^{d^2_1\times d^1_{L_1}}$ and hence
  $\BW_2^{(0)}\BW_1^{(L_1)}$ has not more than $d_1^2\times d_{L_1}^1$
  (nonzero) entries.  Moreover,
  $\BW^{(0)}_2 \Bb^{(L_1)}_1+\Bb_2^{(0)}\in\R^{d_1^2}$.  Thus, the
  $L_1$-th layer of $\Phi_2\circ\Phi_1(\Bx)$ has at most
  $d_1^2\times (1+d_{L_1}^1)$ entries.  The rest is obvious from the
  construction.
\end{proof}

Interpreting linear transformations as neural networks of depth $0$,
the previous lemma is also valid in case $\Phi_1$ or $\Phi_2$ is a
linear mapping.

\subsection{Parallelization}\label{sec:parallel}
Let $(\Phi_i)_{i=1}^m$ be neural networks with architectures
$(\sigma_{\rm ReLU};d_0^i, \dots, d_{L_i+1}^i)$, respectively.  We
proceed to build a neural network $(\Phi_1, \dots, \Phi_m)$ %
realizing the function
\begin{align}\label{eq:map_parallel}
  (\Phi_1, \dots, \Phi_m) \colon \R^{\sum_{j=1}^m d^j_{0}}&\to \R^{\sum_{j=1}^m d^j_{L_j+1}}\\
  (\Bx_1,\dots,\Bx_m)&\mapsto (\Phi_1(\Bx_1),\dots,\Phi_m(\Bx_m)).
                       \nonumber
\end{align}

To do so we first assume $L_1=\dots=L_m=L$, and define
$(\Phi_1, \dots, \Phi_m)$ via the following sequence of weight-bias
tuples:
\begin{align}\label{eq:parallel}
  \left(\begin{pmatrix}
          \BW^{(0)}_1 & &\\
                      &\ddots &\\        
                      & &\BW^{(0)}_m\\
	\end{pmatrix},
  \begin{pmatrix}
    \Bb^{(0)}_1\\
    \vdots\\
    \Bb^{(0)}_m
  \end{pmatrix}
  \right),\dots,
  \left(\begin{pmatrix}
          \BW^{(L)}_1 & &\\
                      &\ddots &\\
                      && \BW^{(L)}_m
	\end{pmatrix},
  \begin{pmatrix}
    \Bb^{(L)}_1\\
    \vdots\\
    \Bb^{(L)}_m
  \end{pmatrix}
  \right)
\end{align}
where these matrices are understood as block-diagonal filled up with
zeros.  For the general case where the $\Phi_j$ might have different
depths, let $L_{\max}\dfn\max_{1\le i\le m}L_i$ and
$I\dfn\set{1\le i\le m}{L_i<L_{\max}}$.  For $j\in I^c$ set
$\widetilde{\Phi}_j\dfn\Phi_j$, and for each $j\in I$
\begin{align}\label{eq:saiduhaoshd}
  \widetilde\Phi_j&\dfn \nId{L_{\max}-L_j}\circ \Phi_j. 
\end{align}
Finally,
\begin{align}\label{eq:saiduhaoshd2}
  (\Phi_1,\dots,\Phi_m)\dfn(\widetilde\Phi_1,\dots,\widetilde\Phi_m).
\end{align}

We collect the properties of the parallelization in the lemma below.

\begin{lemma}[Parallelization]\label{lemma:parallelization}
  Let $m \in \N$ and $(\Phi_i)_{i=1}^m$ be neural networks with
  architectures $(\sigma_{\rm ReLU};d_0^i, \dots, d_{L_i+1}^i)$,
  respectively.  Then the neural network $(\Phi_1, \dots, \Phi_m)$
  satisfies
  \[
    (\Phi_1, \dots, \Phi_m)(\Bx) = (\Phi_1(\Bx_1),\dots,\Phi_m(\Bx_m))
    \text{ for all } \Bx \in \R^{\sum_{j=1}^m d^j_{0}}.
  \]
  Moreover, with $L_{\max}\dfn \max_{j\le m}L_j$ it holds that
  \begin{subequations}\label{eq:lemma:parallel}
    \begin{align}
      \wdth((\Phi_1, \dots, \Phi_m)) &\le 2\sum_{j=1}^m\wdth(\Phi_j),\\
      \depth((\Phi_1, \dots, \Phi_m)) &= \max_{j\le m}\depth(\Phi_j),\\
      \size((\Phi_1, \dots, \Phi_m)) &\le {2}\sum_{j=1}^m\size(\Phi_j)+2\sum_{j=1}^m(L_{\max}-L_j) {d_{L_j+1}^j}.
    \end{align}
  \end{subequations}
\end{lemma}

\begin{proof}
  All statements except for the bound on the size follow immediately
  from the construction.  To obtain the bound on the size, we note
  that by construction the sizes of the $(\widetilde\Phi_i)_{i=1}^m$
  in %
  \eqref{eq:saiduhaoshd} will simply be added.  The size of %
  each $\widetilde\Phi_i$ can be bounded with Lemma
  \ref{lemma:composition}.
\end{proof}

If all input dimensions $d_0^1=\dots=d_0^m \eqqcolon d_0$ are the
same, we will also use \textbf{parallelization with shared inputs} to
realize the function $\Bx\mapsto (\Phi_1(\Bx),\dots,\Phi_m(\Bx))$ from
$\R^{d_0}\to \R^{d^1_{L_1+1}+\dots+d^m_{L_m+1}}$.  In terms of the
construction \eqref{eq:parallel}, the only required change is that the
block-diagonal matrix ${\rm diag}(\BW_1^{(0)},\dots,\BW_m^{(0)})$
becomes the matrix in $\R^{\sum_{j=1}^m d_1^j\times d_0^1}$ which
stacks $\BW_1^{(0)},\dots,\BW_m^{(0)}$ on top of each other.
Similarly, we will allow $\Phi_j$ to only take some of the entries of
$\Bx$ as input.  For parallelization with shared inputs we will use
the same notation $(\Phi_j)_{j=1}^m$ as before, where the precise
meaning will always be clear from context.  Note that Lemma
\ref{lemma:parallelization} remains valid in this case.

\subsection{Linear combinations}\label{sec:lincomb}
Let $m \in \N$ and let $(\Phi_i)_{i=1}^m$ be ReLU neural networks that
have architectures $(\sigma_{\rm ReLU};d_0^i, \dots, d_{L_i+1}^i)$,
respectively.  Assume that $d^1_{L_1+1}=\dots=d^m_{L_m+1}$, i.e., all
$\Phi_1,\dots,\Phi_m$ have the same output dimension.  For scalars
$\alpha_j\in\R$, we wish to construct a ReLU neural network
$\sum_{j=1}^m\alpha_j\Phi_j$ realizing the function
\begin{align*}
  \begin{cases}
    \R^{\sum_{j=1}^m d^j_{0}}\to \R^{d^1_{L_1+1}}\\
    (\Bx_1,\dots,\Bx_m)\mapsto \sum_{j=1}^m\alpha_j\Phi_j(\Bx_j).
  \end{cases}
\end{align*}

This corresponds to the parallelization $(\Phi_1, \dots, \Phi_m)$
composed with the linear transformation
$(\Bz_1,\dots,\Bz_m)\mapsto \sum_{j=1}^m\alpha_j\Bz_j$.  The following
result holds.

\begin{lemma}[Linear combinations]\label{lemma:addition}
  Let $m \in \N$ and $(\Phi_i)_{i=1}^m$ be neural networks with
  architectures $(\sigma_{\rm ReLU};d_0^i, \dots, d_{L_i+1}^i)$,
  respectively.  Assume that $d^1_{L_1+1}=\dots=d^m_{L_m+1}$, let
  $\alpha\in\R^m$ and set $L_{\max}\dfn \max_{j\le m}L_j$.  Then,
  there exists a neural network $\sum_{j=1}^m\alpha_j\Phi_j$ such that
  $(\sum_{j=1}^m\alpha_j\Phi_j)(\Bx) =
  \sum_{j=1}^m\alpha_j\Phi_j(\Bx_j)$ for all
  $\Bx = (\Bx_j)_{j=1}^m \in \R^{\sum_{j=1}^m d^j_{0}}$.  Moreover,
  \begin{subequations}%
    \begin{align}
      \wdth\left(\sum_{j=1}^m\alpha_j\Phi_j\right) &\le 2\sum_{j=1}^m\wdth(\Phi_j),\\
      \depth\left(\sum_{j=1}^m\alpha_j\Phi_j\right) &= \max_{j\le m}\depth(\Phi_j),\\
      \size\left(\sum_{j=1}^m\alpha_j\Phi_j\right) &\le {2}\sum_{j=1}^m\size(\Phi_j)+2\sum_{j=1}^m(L_{\max}-L_j) {d_{L_j+1}^j}.
    \end{align}
  \end{subequations}
\end{lemma}

\begin{proof}
  The construction of $\sum_{j=1}^m\alpha_j\Phi_j$ is analogous to
  that of $(\Phi_1, \dots, \Phi_m)$, i.e., we first define the linear
  combination of neural networks with the same depth.  Then the
  weights are chosen as in \eqref{eq:parallel}, but with the last
  linear transformation replaced by
  \begin{align*}
    \left(( \alpha_1\BW^{(L)}_1 \cdots \alpha_m\BW^{(L)}_m),
    \sum_{j=1}^m \alpha_j\Bb^{(L)}_j
    \right).
  \end{align*}

  For general depths, we define the sum of the neural networks to be
  the sum of the extended neural networks $\widetilde{\Phi}_i$ as of
  \eqref{eq:saiduhaoshd}.  All statements of the lemma follow
  immediately from this construction.
\end{proof}

In case $d_0^1=\dots=d_0^m \eqqcolon d_0$ (all neural networks have
the same input dimension), we will also consider \textbf{linear
  combinations with shared inputs}, i.e., a neural network realizing
\[
  \Bx\mapsto \sum_{j=1}^m\alpha_j\Phi_j(\Bx)\qquad\text{for } \Bx \in
  \R^{d_0}.
\]
This requires the same minor adjustment as discussed at the end of
Section \ref{sec:parallel}.  Lemma \ref{lemma:addition} remains valid
in this case and again we do not distinguish in notation for linear
combinations with or without shared inputs.

\section{Continuous piecewise linear functions}\label{sec:CWPLfunctionsandRepresentations}
In this section, we will relate ReLU neural networks to a large class
of functions.  We first formally introduce the set of continuous
piecewise linear functions from a set $\Omega\subseteq\R^d$ to $\R$.
Note that we admit in particular $\Omega=\R^d$ in the following
definition.

\begin{definition}\label{def:cpwl}
  Let $\Omega\subseteq\R^d$, $d\in\N$.  We call a function
  $f:\Omega\to\R$ {\bf continuous, piecewise linear (cpwl)} if
  $f\in C^0(\Omega)$ and there exist $n\in\N$ affine functions
  $g_j \colon \R^d\to\R,$ $g_j(\Bx) = \Bw_j^\top\Bx+b_j$ such that for
  each $\Bx\in\Omega$ it holds that $f(\Bx)= g_j(\Bx)$ for at least
  one $j\in\{1,\dots,n\}$.  For $m>1$ we call $f:\Omega\to\R^m$ cpwl
  if and only if each component of $f$ is cpwl.
\end{definition}

\begin{remark}
  A ``continuous piecewise linear function'' as in
  Definition~\ref{def:cpwl} is actually %
  piecewise \emph{affine}.  To maintain consistency with the
  literature, we use the terminology cpwl.
\end{remark}

In the following, we will refer to the connected domains on which $f$
is equal to one of the functions $g_j$, also as {\bf regions} or {\bf
  pieces}.  If $f$ is cpwl with $q\in\N$ regions, then with $n \in \N$
denoting the number of affine functions it holds $n\le q$.

Note that, the mapping $\Bx \mapsto \sigma_{\rm ReLU}(\Bw^\top\Bx+b)$,
which is a ReLU neural network with a single neuron, is cpwl (with two
regions).  Consequently, every ReLU neural network is a repeated
composition of linear combinations of cpwl functions.  It is not hard
to see that the set of cpwl functions is closed under compositions and
linear combinations.  Hence, \emph{every ReLU neural network is a cpwl
  function.}  Interestingly, the reverse direction of this statement
is also true, meaning that \emph{every cpwl function can be
  represented by a ReLU neural network} as we shall demonstrate below.
Therefore, we can identify the class of functions realized by
arbitrary ReLU neural networks as the class of cpwl functions.

\begin{theorem}\label{thm:cpwlrelu}
  Let $d\in \N$, let $\Omega\subseteq\R^d$ be convex, and let
  $f:\Omega\to\R$ be cpwl with $n \in \N$ as in
  Definition~\ref{def:cpwl}.  Then, there exists a ReLU neural network
  $\Phi^f$ such that $\Phi^f(\Bx)=f(\Bx)$ for all $\Bx\in\Omega$ and
  \begin{equation*}
    \size(\Phi^f)=O(dn2^n),\quad
    \wdth(\Phi^f)=O(dn2^n),\quad
    \depth(\Phi^f)=O(n).
  \end{equation*}
\end{theorem}

A statement similar to Theorem \ref{thm:cpwlrelu} can be found in
\cite{arora2018understanding,MR4087799}. There, the authors give a
construction with a depth that behaves logarithmic in $d$ and is
independent of $n$, but with significantly larger bounds on the
size. As we shall see, the proof of Theorem \ref{thm:cpwlrelu} is %
a simple consequence of the following well-known result %
from \cite{TARELA199917}; also see \cite{MR1913786}, and for sharper
bounds \cite{1542439}. It states that every cpwl function can be
expressed as a finite maximum of a finite minimum of certain affine
functions.

\begin{proposition}\label{prop:cpwlrelu}
  Let $d\in \N$, $\Omega\subseteq\R^d$ be convex, and let
  $f:\Omega\to\R$ be cpwl with $n\in\N$ affine functions as in
  Definition \ref{def:cpwl}.  Then there exists $m\in\N$ and sets
  $s_j\subseteq\{1,\dots,n\}$ for $j\in \{1,\dots,m\}$, such that
  \begin{align}\label{eq:maxmin}
    f(\Bx) = \max_{1\le j\le m}\min_{i\in s_j}(g_i(\Bx))\qquad\text{ for all } \Bx\in\Omega.
  \end{align}
\end{proposition}

\begin{proof} {\bf Step 1.} We start with $d=1$, i.e.,
  $\Omega\subseteq\R$ is a (possibly unbounded) interval and for each
  $x\in \Omega$ there exists $j\in\{1,\dots,n\}$ such that with
  $g_j(x) \coloneqq w_jx+b_j$ it holds that $f(x)=g_j(x)$.  Without
  loss of generality, we can assume that $g_i \neq g_j$ for all
  $i \neq j$.  Since the graphs of the $g_j$ are lines, they intersect
  at (at most) finitely many points in $\Omega$.

  Since $f$ is continuous, we conclude that there exist finitely many
  intervals covering $\Omega$, such that $f$ coincides with one of the
  $g_j$ on each interval.  For each $x\in \Omega$ let
  \begin{align*}
    s_x\dfn \set{1\le j\le n}{g_j(x)\ge f(x)}
  \end{align*}
  and
  \begin{align*}
    f_x(y)\dfn \min_{j\in s_x}g_j(y)\qquad\text{ for all } y\in \Omega.
  \end{align*}
  Clearly, $f_x(x)=f(x)$.  We claim that, additionally,
  \begin{align}\label{eq:claimfx}
    f_x(y)\le f(y)\qquad\text{ for all } y\in \Omega.
  \end{align}
  This then shows that
  \begin{align*}
    f(y)=\max_{x\in \Omega}f_x(y) = \max_{x\in \Omega}\min_{j\in s_x}g_j(y)\qquad\text{ for all } y\in\R.
  \end{align*}
  Since there exist only finitely many possibilities to choose a
  subset of $\{1,\dots,n\}$, we conclude that \eqref{eq:maxmin} holds
  for $d=1$.

  It remains to verify the claim \eqref{eq:claimfx}.  Fix
  $y\neq x\in\Omega$.  Without loss of generality, let $x<y$ and let
  $x=x_0<\dots<x_k=y$ be such that $f|_{[x_{i-1},x_{i}]}$ equals some
  $g_j$ for each $i\in\{1,\dots,k\}$.  In order to show
  \eqref{eq:claimfx}, it suffices to prove that there exists at least
  one $j$ such that $g_j(x_0)\ge f(x_0)$ and $g_j(x_k)\le f(x_k)$.
  The claim is trivial for $k=1$.  We proceed by induction.  Suppose
  the claim holds for $k-1$, and consider the partition
  $x_0<\dots<x_k$.  Let $r\in\{1,\dots,n\}$ be such that
  $f|_{[x_0,x_1]}=g_r|_{[x_0,x_1]}$.  Applying the induction
  hypothesis to the interval $[x_1,x_k]$, we can find
  $j\in\{1,\dots,n\}$ such that $g_j(x_1)\ge f(x_1)$ and
  $g_j(x_k)\le f(x_k)$.  If $g_j(x_0)\ge f(x_0)$, then $g_j$ is the
  desired function.  Otherwise, $g_j(x_0)<f(x_0)$.  Then
  $g_r(x_0)=f(x_0)>g_j(x_0)$ and $g_r(x_1)=f(x_1)\le g_j(x_1)$.
  Therefore $g_r(x)\le g_j(x)$ for all $x\ge x_1$, and in particular
  $g_r(x_k)\le g_j(x_k)$.  Thus $g_r$ is the desired function.

  {\bf Step 2.}  For general $d\in\N$, let
  $g_j(\Bx)\dfn \Bw_j^\top\Bx+b_j$ for $j=1,\dots,n$.  For each
  $\Bx\in\Omega$, let
  \begin{align*}
    s_\Bx\dfn \set{1\le j\le n}{g_j(\Bx)\ge f(\Bx)}
  \end{align*}
  and for all $\By\in\Omega$, let
  \begin{align*}
    f_\Bx(\By)\dfn \min_{j\in s_\Bx}g_j(\By).
  \end{align*}

  For an arbitrary $1$-dimensional affine subspace $S\subseteq \R^d$
  passing through $\Bx$ consider the line (segment)
  $I\dfn S\cap\Omega$, which is connected since $\Omega$ is convex.
  By Step 1, it holds
  \begin{align*}
    f(\By) = \max_{\Bx\in\Omega}f_\Bx(\By)=\max_{\Bx\in\Omega}\min_{j\in s_\Bx}g_j(\By)
  \end{align*}
  on all of $I$.  Since $I$ was arbitrary the formula is valid for all
  $\By\in\Omega$.  This again implies \eqref{eq:maxmin} as in Step 1.
\end{proof}

\begin{remark}
  For any $a_1,\dots,a_k\in\R$ holds
  $\min\{-a_1,\dots,-a_k\}=-\max\{a_1,\dots,a_k\}$. Thus, in the
  setting of Proposition \ref{prop:cpwlrelu}, there exists
  $\tilde m\in\N$ and sets $\tilde s_j\subseteq\{1,\dots,n\}$ for
  $j=1,\dots,\tilde m$, such that for all $\Bx\in\Omega$
  \begin{align*}
    f(\Bx)=-(-f(\Bx)) &= -\max_{1\le j\le \tilde m}\min_{i\in \tilde s_j}(-g_i(\Bx))\\
                      &=-\max_{1\le j\le \tilde m}(-\max_{i\in \tilde s_j}(g_i(\Bx)))\\
                      &=\min_{1\le j\le \tilde m}(\max_{i\in \tilde s_j}(g_i(\Bx))).
  \end{align*}
\end{remark}

To prove Theorem~\ref{thm:cpwlrelu}, it therefore suffices to show
that the minimum and the maximum are expressible by ReLU neural
networks.

\begin{lemma}\label{lemma:minmax}
  For every $x$, $y\in\R$ it holds that
  \begin{align*}
    \min\{x,y\} = \sigma_{\rm ReLU}(y)-\sigma_{\rm ReLU}(-y) - \sigma_{\rm ReLU}(y-x) \in \CN_2^1( \sigma_{\rm ReLU};1,3)
  \end{align*}
  and
  \begin{align*}
    \max\{x,y\} = \sigma_{\rm ReLU}(y)-\sigma_{\rm ReLU}(-y)+\sigma_{\rm ReLU}(x-y)\in \CN_2^1( \sigma_{\rm ReLU};1,3).
  \end{align*}
\end{lemma}

\begin{proof}
  We have
  \begin{align*}
    \max\{x,y\} &= y + \begin{cases}
                         0 &\text{if }y>x\\
                         x-y &\text{if }x\ge y
                       \end{cases}\\
                &= y + \sigma_{\rm ReLU}(x-y).
  \end{align*}
  Using $y=\sigma_{\rm ReLU}(y)-\sigma_{\rm ReLU}(-y)$, the claim for
  the maximum follows.  For the minimum observe that
  $\min\{x,y\}=-\max\{-x,-y\}$.
\end{proof}

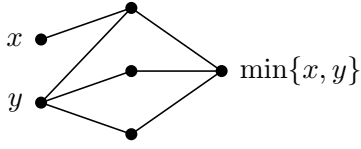
\begin{figure}
  \begin{center}
    \input{./plots/minnet.tex}
  \end{center}\caption{Sketch of the neural network in Lemma
    \ref{lemma:minmax}.  Only edges with non-zero weights are drawn.
  }\label{fig:minnet}
\end{figure}

The minimum of $n\ge 2$ inputs can be computed by repeatedly applying
the construction of Lemma \ref{lemma:minmax}.  The resulting neural
network is described in the next lemma.

\begin{lemma}\label{lemma:minmaxn}
  For every $n\ge 2$ there exists a neural network
  $\nmin{n}:\R^n\to\R$ with
  \begin{equation*}
    \size(\nmin{n})\le 16 n,\qquad \wdth(\nmin{n})\le 3n,\qquad
    \depth(\nmin{n})\le \lceil\log_2(n)\rceil
  \end{equation*}
  such that $\nmin{n}(x_1,\dots,x_n)=\min_{1\le j\le n} x_j$.
  Similarly, there exists a neural network $\nmax{n}:\R^n\to\R$
  realizing the maximum and satisfying the same %
  complexity bounds.
\end{lemma}

\begin{figure}
                                        \begin{center}
                                          \input{./plots/minnetgeneral.tex}
                                        \end{center}\caption{Construction
                                          of $\nmin{n}$ for general
                                          $n$ in Step 2 of the proof
                                          of Lemma
                                          \ref{lemma:minmaxn}.}\label{fig:generalmin}
                                      \end{figure}
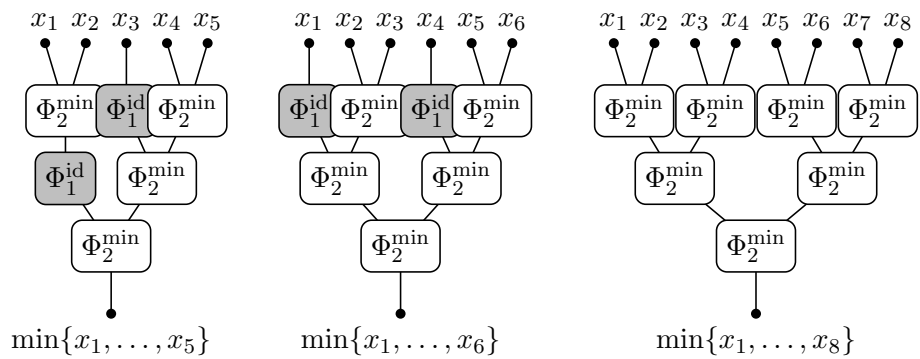

\begin{proof}
  Throughout denote by $\nmin{2}:\R^2\to\R$ the neural network from
  Lemma \ref{lemma:minmax}. It is of depth $1$ and size $7$ (since all
  biases are zero, it suffices to count the number of connections in
  Figure~\ref{fig:minnet}).

  {\bf Step 1.} Consider first the case where $n=2^k$ for some
  $k\in\N$.  We proceed by induction of $k$. For $k=1$ the claim is
  proven. For $k\ge 2$ set
  \begin{equation}\label{eq:minhat2k}
    \nmin{2^k}\dfn
    \nmin{2}\circ (\nmin{2^{k-1}},\nmin{2^{k-1}}).
  \end{equation}
  By Lemma \ref{lemma:composition} and Lemma
  \ref{lemma:parallelization} we have
  \begin{align*}
    \depth(\nmin{2^k})\le \depth(\nmin{2})+\depth(\nmin{2^{k-1}})\le\cdots\le k.
  \end{align*}
  Next, we bound the size of the neural network.  Note that all biases
  in this neural network are set to $0$, since the $\nmin{2}$ neural
  network in Lemma \ref{lemma:minmax} has no biases.  Thus, the size
  of the neural network $\nmin{2^k}$ corresponds to the number of
  connections in the graph (the number of nonzero weights).  Careful
  inspection of the neural network architecture, see
  Figure~\ref{fig:minnetn}, reveals that
  \begin{align*}
    \size(\nmin{2^k})&=4\cdot 2^{k-1}
                       +\sum_{j=0}^{k-2} 12\cdot 2^{j}
                       +3\\
                     &= 2n+12\cdot (2^{k-1}-1)+3
                       = 2n+6n-9\le 8n,
  \end{align*}
  and that $\wdth(\nmin{2^k})\le ({3}/{2}) 2^{k}$. This concludes the
  proof for the case $n=2^k$.

  {\bf Step 2.} For the general case, we first let
  \begin{equation*}
    \nmin{1}(x)\dfn x\qquad\text{for all }x\in\R
  \end{equation*}
  be the identity on $\R$, i.e.\ a linear transformation and thus
  formally a depth $0$ neural network. Then, for all $n\ge 2$
  \begin{equation}\label{eq:minhatn}
    \nmin{n}\dfn
    \nmin{2}\circ
    \begin{cases}
      (\nId{1}\circ\nmin{\lfloor\frac{n}{2}\rfloor},\nmin{\lceil\frac{n}{2}\rceil}) &\text{if }n\in\set{2^k+1}{k\in\N}\\
      (\nmin{\lfloor\frac{n}{2}\rfloor},\nmin{\lceil\frac{n}{2}\rceil}) &\text{otherwise.}
    \end{cases}
  \end{equation}
  This definition extends \eqref{eq:minhat2k} to arbitrary $n\ge 2$,
  since the first case in \eqref{eq:minhatn} never occurs if $n\ge 2$
  is a power of two.

  To analyze \eqref{eq:minhatn}, we start with the depth and claim
  that
  \begin{equation*}
    \depth(\nmin{n})=k\qquad\text{for all }2^{k-1}<n\le 2^{k}
  \end{equation*}
  and all $k\in\N$. We proceed by induction over $k$. The case $k=1$
  is clear. For the induction step, assume the statement holds for
  some fixed $k\in\N$ and fix an integer $n$ with
  $2^{k}<n\le 2^{k+1}$. Then
  \begin{equation*}
    \Big\lceil \frac{n}{2}\Big\rceil\in (2^{k-1},2^k]\cap\N
  \end{equation*}
  and
  \begin{equation*}
    \Big\lfloor \frac{n}{2}\Big\rfloor \in\begin{cases}
                                            \{2^{k-1}\} &\text{if }n=2^{k}+1\\
                                            (2^{k-1},2^k]\cap\N &\text{otherwise.}
                                          \end{cases}
                                        \end{equation*}
                                        Using the induction
                                        assumption, \eqref{eq:minhatn}
                                        and Lemmas
                                        \ref{lemma:identity} and
                                        \ref{lemma:composition}, this
                                        shows
                                        \begin{equation*}
                                          \depth(\nmin{n}) = \depth(\nmin{2}) + k = 1+k,
                                        \end{equation*}
                                        and proves the claim.

                                        For the size and width bounds,
                                        we only sketch the argument:
                                        Fix $n\in\N$ such that
                                        $2^{k-1}<n\le 2^k$. Then
                                        $\nmin{n}$ is constructed from
                                        at most as many subnetworks as
                                        $\nmin{2^k}$, but with some
                                        $\nmin{2}:\R^2\to\R$ blocks
                                        replaced by $\nId{1}:\R\to\R$,
                                        see Figure
                                        \ref{fig:generalmin}. Since
                                        $\nId{1}$ has the same depth
                                        as $\nmin{2}$, but is smaller
                                        in width and number of
                                        connections, the width and
                                        size of $\nmin{n}$ is bounded
                                        by the width and size of
                                        $\nmin{2^k}$. Due to
                                        $2^k\le 2n$, the bounds from
                                        Step 1 give the bounds stated
                                        in the lemma.

                                        {\bf Step 3.} For the maximum,
                                        define
                                        \[
                                          \nmax{n} (x_1,\dots,x_n)\dfn
                                          -\nmin{n}(-x_1,\dots,-x_n).
                                        \]
                                      \end{proof}

                                      \begin{figure}
                                        \begin{center}
                                          \input{./plots/minnetn.tex}
                                        \end{center}\caption{Architecture
                                          of the $\nmin{2^k}$ neural
                                          network in Step 1 of the
                                          proof of Lemma
                                          \ref{lemma:minmaxn} and the
                                          number of connections in
                                          each layer for $k=3$. Each
                                          grey box corresponds to $12$
                                          connections in the graph.}
                                        \label{fig:minnetn}
                                      \end{figure}
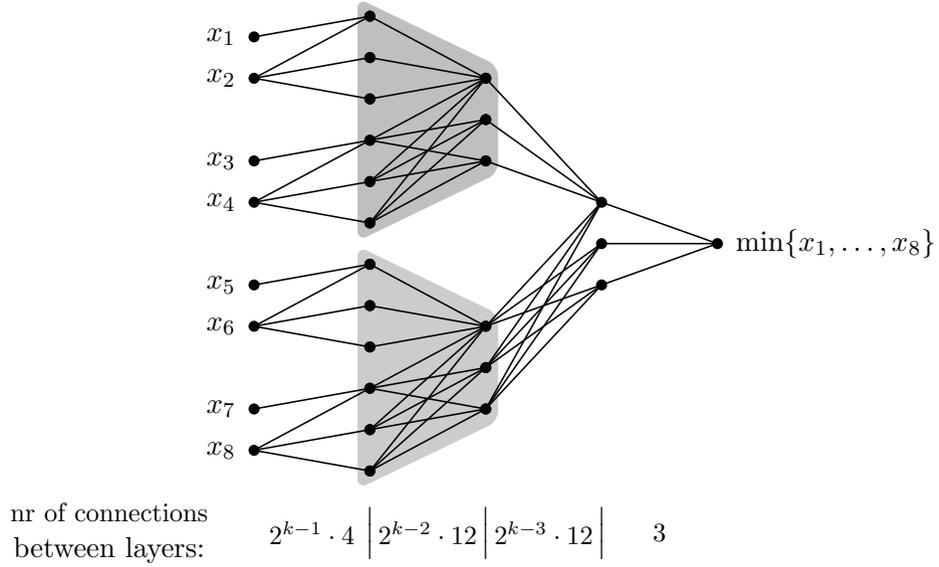

\begin{proof}[of Theorem~\ref{thm:cpwlrelu}]
  By Proposition~\ref{prop:cpwlrelu} the neural network
  \begin{align*}
    \Phi\dfn \nmax{m}\bullet (\nmin{|s_j|})_{j=1}^m\bullet
    ((\Bw_i^\top \Bx+b_i)_{i\in s_j})_{j=1}^m
  \end{align*}
  realizes the function $f$.

  Since the number of possibilities to choose subsets of
  $\{1,\dots,n\}$ equals $2^n$ we have $m\le 2^n$.  Since each $s_j$
  is a subset of $\{1,\dots,n\}$, the cardinality $|s_j|$ of $s_j$ is
  bounded by $n$.  By Lemma \ref{lemma:composition}, Lemma
  \ref{lemma:parallelization}, and Lemma \ref{lemma:minmaxn}
  \begin{align*}
    \depth(\Phi) &\le 2+\depth(\nmax{m})+\max_{1\le j\le n}\depth(\nmin{|s_j|})\\
                 &\le 1+\lceil\log_2(2^n)\rceil+\lceil\log_2(n)\rceil = O(n)
  \end{align*}
  and
  \begin{align*}
    \wdth(\Phi)&\le 2\max\Big\{\wdth(\nmax{m}),\sum_{j=1}^m\wdth(\nmin{|s_j|}),\sum_{j=1}^m\wdth((\Bw_i^\top \Bx+b_i)_{i\in s_j}))\Big\}\nonumber\\
               &\le 2\max\{3 m, 3 m n,%
                 mdn\} = O(dn2^n)
  \end{align*}
  and
  \begin{align*}
    \size(\Phi)&\le 4\Big(\size(\nmax{m})+\size((\nmin{|s_j|})_{j=1}^m)+\size{((\Bw_i^\top \Bx+b_i)_{i\in s_j})_{j=1}^m)}\Big)\nonumber\\
               &\le 4\left(16m +2\sum_{j=1}^m (16|s_j|+2\lceil\log_2(n)\rceil)+{nm(d+1)}\right) = O(dn2^n).
  \end{align*}
  This concludes the proof.
\end{proof}

\section{Simplicial pieces}\label{sec:simplicialPieces}
This section studies the case, where we do not have arbitrary cpwl
functions, but where the regions on which $f$ is affine are simplices.
Under this condition, we can construct neural networks that scale
merely \emph{linearly} in the number of such regions, which is a
serious improvement from the \emph{exponential} dependence of the size
on the number of regions that was found in Theorem \ref{thm:cpwlrelu}.

\subsection{Triangulations of $\Omega$}
For the ensuing discussion, we will consider $\Omega\subseteq\R^d$ to
be partitioned into simplices.  This partitioning will be termed a
\textbf{triangulation} of $\Omega$.  Other notions prevalent in the
literature include a \textbf{tessellation} of $\Omega$, or a
\textbf{simplicial mesh} on $\Omega$.  To give a precise definition,
let us first recall some terminology.  For a set $S\subseteq\R^d$ we
denote the {\bf convex hull} of $S$ by
\begin{align}\label{eq:convexhull}
  \co(S)\dfn \setc{\sum_{j=1}^n\alpha_j \Bx_j}{n\in\N,~\Bx_j\in S,\alpha_j\ge 0,~\sum_{j=1}^n\alpha_j=1}.
\end{align}

An \textbf{$n$-simplex} is the convex hull of $n\in \N$ points that
are independent in a specific sense.  This is made precise in the
following definition.

\begin{definition}
  Let $n\in\N_0$, $d\in\N$ and $n\le d$.  We call
  $\Bx_0,\dots,\Bx_n\in\R^d$ {\bf affinely independent} if and only if
  either $n=0$ or $n\ge 1$ and the vectors
  $\Bx_1-\Bx_0,\dots,\Bx_n-\Bx_0$ are linearly independent.  In this
  case, we call $\co(\Bx_0,\dots,\Bx_n)\dfn\co(\{\Bx_0,\dots,\Bx_n\})$
  an {\bf $n$-simplex}.
\end{definition}

As mentioned before, a triangulation refers to a partition of a space
into simplices.  We give a formal definition below.

\begin{definition}\label{def:mesh}
  Let $d\in\N$, and $\Omega\subseteq\R^d$ be compact.  Let $\CT$ be a
  finite set of $d$-simplices, %
  and for each $\tau\in\CT$ let $V(\tau)\subseteq \Omega$ have
  cardinality $d+1$ such that $\tau=\co(V(\tau))$. We call $\CT$ a
  {\bf regular triangulation} of $\Omega$, if and only if
  \begin{enumerate}
  \item %
    $\bigcup_{\tau\in\CT}\tau = \Omega$,
  \item\label{def:meshint} for all $\tau$, $\tau'\in\CT$ it holds that
    $\tau\cap\tau'=\co(V(\tau)\cap V(\tau'))$.
  \end{enumerate}
  We call $\Beta\in \CV\dfn \bigcup_{\tau\in\CT}V(\tau)$ a {\bf node}
  (or vertex) and $\tau\in\CT$ an {\bf element} of the triangulation.
\end{definition}

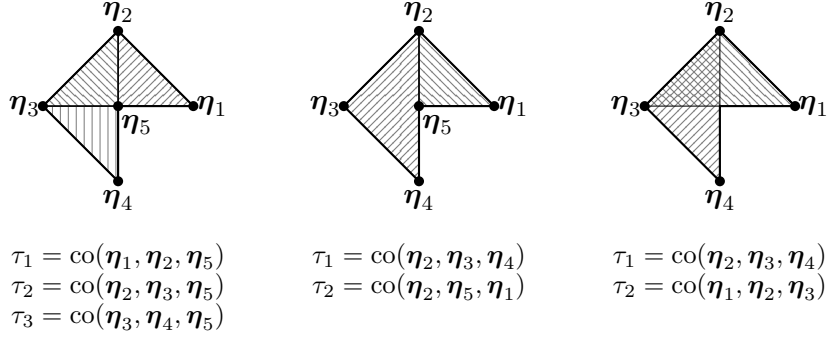
\begin{figure}
  \begin{center}
    \input{./plots/triangulation.tex}
  \end{center}\caption{The first is a regular triangulation, while the
    second and the third are not.}\label{fig:triangulation}
\end{figure}

For a regular triangulation $\CT$ with nodes $\CV$ we also introduce
the constant
\begin{align}\label{eq:kT}
  k_\CT\dfn \max_{\Beta\in\CV}|\set{\tau\in\CT}{\Beta\in\tau}|
\end{align}
corresponding to the maximal number of elements shared by a single
node.

\subsection{Size bounds for regular
  triangulations}\label{sec:bound_tri}

Throughout this subsection, let $\CT$ be a regular triangulation of
$\Omega$, and we adhere to the notation of Definition \ref{def:mesh}.
We will say that $f:\Omega\to\R$ is cpwl with respect to $\CT$ if $f$
is cpwl and $f|_\tau$ is affine for each $\tau\in\CT$.  The rest of
this subsection is dedicated to proving the following result.  It was
first shown in \cite{Longo2023DeRham} with a more technical argument,
and extends an earlier statement from \cite{MR4087799} to general
triangulations (also see Section \ref{sec:convextriangulations}).

\begin{theorem}\label{thm:simplicial}
  Let $d \in \N$, $\Omega\subseteq\R^d$ be a bounded domain, and let
  $\CT$ be a regular triangulation of $\Omega$.  Let $f:\Omega\to\R$
  be cpwl with respect to $\CT$ and $f|_{\partial\Omega}=0$.  Then
  there exists a ReLU neural network $\Phi:\Omega\to\R$ realizing $f$,
  and it holds
  \begin{align}
    \size(\Phi) = O(|\CT|),\qquad
    \wdth(\Phi)=O(|\CT|),\qquad
    \depth(\Phi) = O(1),
  \end{align}
  where the constants in the Landau notation depend on $d$ and $k_\CT$
  in \eqref{eq:kT}.
\end{theorem}

We will split the proof into several lemmata.  The strategy is to
introduce a basis of the space of cpwl functions on $\CT$ the elements
of which vanish on the boundary of $\Omega$.  We will then show that
there exist $O(|\CT|)$ basis functions, each of which can be
represented with a neural network the size of which depends only on
$k_\CT$ and $d$.  To construct this basis, we first point out that an
affine function on a simplex is uniquely defined by its values at the
nodes.

\begin{lemma}\label{lemma:linex} 
  Let $d \in \N$.  Let $\tau\dfn \co(\Beta_0,\dots,\Beta_d)$ be a
  $d$-simplex.  For every $y_0,\dots,y_{d}\in\R$, there exists a
  unique $g\in\P_1(\R^d)$ such that $g(\Beta_i)=y_i$, $i=0,\dots,d$.
\end{lemma}

\begin{proof}
  Since $\Beta_1-\Beta_0,\dots,\Beta_d-\Beta_0$ is a basis of $\R^d$,
  there is a unique $\Bw\in\R^d$ such that
  $\Bw^\top (\Beta_i-\Beta_0)=y_i-y_0$ for $i=1,\dots,d$.  Then
  $g(\Bx)\dfn \Bw^\top\Bx+(y_0-\Bw^\top\Beta_0)$ is as desired.
  Moreover, for every $g\in\P_1$ it holds that
  $g(\sum_{i=0}^d\alpha_i\Beta_i)=\sum_{i=0}^d\alpha_ig(\Beta_i)$
  whenever $\sum_{i=0}^d\alpha_i=1$ (this is in general not true if
  the coefficients do not sum to $1$).  Hence, $g$ is uniquely
  determined by its values at the nodes.
\end{proof}

Since $\Omega$ is the union of the simplices $\tau\in\CT$, every cpwl
function with respect to $\CT$ is thus uniquely defined through its
values at the nodes.  Hence, the desired basis consists of cpwl
functions $\varphi_\Beta:\Omega\to\R$ with respect to $\CT$ such that
\begin{align}\label{eq:basis}
  \varphi_\Beta(\Bmu)=\delta_{\Beta\Bmu}\qquad\text{ for all }\Bmu\in\CV,
\end{align}
where $\delta_{\Beta\Bmu}$ denotes the Kronecker delta.  Assuming
$\varphi_\Beta$ to be well-defined for the moment, we can then
represent every cpwl function $f:\Omega\to\R$ that vanishes on the
boundary $\partial\Omega$ as
\begin{align*}
  f(\Bx) = \sum_{\Beta\in\CV\cap\mathring{\Omega}}f(\Beta)\varphi_\Beta(\Bx)\qquad\text{for all }\Bx\in\Omega.
\end{align*}

Note that it suffices to sum over the set of \textbf{interior nodes}
$\CV\cap\mathring{\Omega}$, since $f(\Beta)=0$ whenever
$\Beta\in\partial\Omega$.  To formally verify existence and
well-definedness of $\varphi_\Beta$, we first need a lemma
characterizing the boundary of so-called ``patches'' of the
triangulation: For each $\Beta\in\CV$, we introduce the {\bf patch}
$\omega(\Beta)$ of the node $\Beta$ as the union of all elements
containing $\Beta$, i.e.,
\begin{align}\label{eq:patchDefinition}
  \omega(\Beta)\dfn \bigcup_{\set{\tau\in\CT}{\Beta\in\tau}}\tau.
\end{align}

\begin{lemma}\label{lemma:wboundary}
  Let $\Beta\in\CV\cap\mathring{\Omega}$ be an interior node.  Then,
  \begin{align*}
    \partial\omega(\Beta) =\bigcup_{\set{\tau\in\CT}{\Beta\in\tau}}\co(V(\tau)\backslash\{\Beta\}).
  \end{align*}
\end{lemma}

\begin{figure}
  \begin{center}
    \begin{tikzpicture}[scale=1]
      \coordinate[label=below:$\Beta$] (center) at (0,0);
      \coordinate[label=above:$\Beta_6$] (A) at (-1,1.3);
      \coordinate[label=above:$\Beta_1$] (B) at (1,1.3);
      \coordinate[label=right:$\Beta_2$] (C) at (2,0);
      \coordinate[label=below:$\Beta_3$] (D) at (1,-1.3);
      \coordinate[label=below:$\Beta_4$] (E) at (-1,-1.3);
      \coordinate[label=left:$\Beta_5$] (F) at (-2,0); \draw (center)
      -- (A) -- (B) -- cycle; \draw (center) -- (B) -- (C) -- cycle;
      \draw (center) -- (C) -- (D) -- cycle; \draw (center) -- (D) --
      (E) -- cycle; \draw (center) -- (E) -- (F) -- cycle; \draw
      (center) -- (F) -- (A) -- cycle; \draw[very thick] (A) -- (B) --
      (C) -- (D) -- (E) -- (F) -- cycle; \node at (0.7, 0.27)
      {$\tau_1$}; \node at (0.7, -0.27) {$\tau_2$}; \node at (0, -0.7)
      {$\tau_3$}; \node at (-0.7, -0.27) {$\tau_4$}; \node at (-0.7,
      0.27) {$\tau_5$}; \node at (0, 0.7) {$\tau_6$}; \draw[red,very
      thick] (B) -- (C); \node at (3.5, 0.9) [red] {\footnotesize
        $\co(V(\tau_1)\backslash\{\Beta\})=\co(\{\Beta_1,\Beta_2\})$};
      \node at (-2.3,1) {$\omega(\Beta)$};
    \end{tikzpicture}
  \end{center}
  \caption{Visualization of Lemma \ref{lemma:wboundary} in two
    dimensions.  The patch $\omega(\Beta)$ consists of the union of
    all $2$-simplices $\tau_i$ containing $\Beta$.  Its boundary
    consists of the union of all $1$-simplices made up by the nodes of
    each $\tau_i$ without the center node, i.e., the convex hulls of
    $V(\tau_i)\backslash\{\Beta\}$.}\label{fig:patch_boundary}
\end{figure}
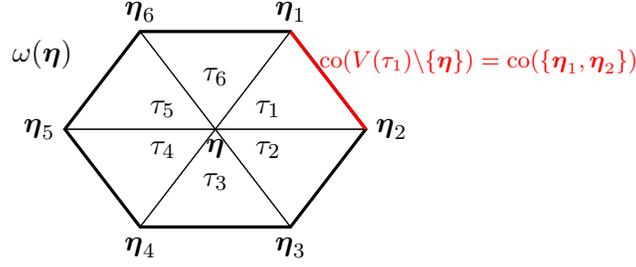

We refer to Figure \ref{fig:patch_boundary} for a visualization of
Lemma \ref{lemma:wboundary}.  The proof of Lemma \ref{lemma:wboundary}
is quite technical but nonetheless elementary.  We therefore only
outline the general argument but leave the details to the reader in
Exercise \ref{ex:boundary_tau}: The boundary of $\omega(\Beta)$ must
be contained in the union of the boundaries of all $\tau$ in the patch
$\omega(\Beta)$.  Since $\Beta$ is an interior point of $\Omega$, it
must also be an interior point of $\omega(\Beta)$.  This can be used
to show that for every
$S\dfn \{\Beta_{i_0},\dots,\Beta_{i_{k}}\}\subseteq V(\tau)$ of
cardinality $k+1\le d$, the interior of (the $k$-dimensional manifold)
$\co(S)$ belongs to the interior of $\omega(\Beta)$ whenever
$\Beta\in S$.  Using Exercise \ref{ex:boundary_tau}, it then only
remains to check that $\co(S)\subseteq\partial\omega(\Beta)$ whenever
$\Beta\notin S$, which yields the claimed formula.  We are now in
position to show well-definedness of the basis functions in
\eqref{eq:basis}.

\begin{lemma}\label{lemma:basis}
  For each interior node $\Beta\in\CV\cap \mathring{\Omega}$ there
  exists a unique cpwl function $\varphi_{\Beta}:\Omega\to\R$
  satisfying \eqref{eq:basis}.  Moreover, $\varphi_{\Beta}$ can be
  expressed by a ReLU neural network with size, width, and depth
  bounds that only depend on $d$ and $k_\CT$.
\end{lemma}

\begin{proof}
  By Lemma \ref{lemma:linex}, on each $\tau\in\CT$, the affine
  function $\varphi_\Beta|_{\tau}$ is uniquely defined through the
  values at the nodes of $\tau$.  This defines a continuous function
  $\varphi_{\Beta}:\Omega\to\R$. Indeed, whenever
  $\tau\cap\tau'\neq\emptyset$, then $\tau\cap\tau'$ is a subsimplex
  of both $\tau$ and $\tau'$ in the sense of Definition \ref{def:mesh}
  \ref{def:meshint}.  Thus, applying Lemma \ref{lemma:linex} again,
  the affine functions on $\tau$ and $\tau'$ coincide on
  $\tau\cap\tau'$.

  Using Lemma \ref{lemma:linex}, Lemma \ref{lemma:wboundary} and the
  fact that $\varphi_{\Beta}(\Bmu)=0$ whenever $\Bmu\neq\Beta$, we
  find that $\varphi_\Beta$ vanishes on the boundary of the patch
  $\omega(\Beta)\subseteq\Omega$.  Thus, $\varphi_\Beta$ vanishes on
  the boundary of $\Omega$.  Extending by zero, it becomes a cpwl
  function $\varphi_\Beta:\R^d\to\R$.  This function is nonzero only
  on elements $\tau$ for which $\Beta\in\tau$.  Hence, it is a cpwl
  function with at most $n\dfn k_\CT+1$ affine functions.  By Theorem
  \ref{thm:cpwlrelu}, $\varphi_\Beta$ can be expressed as a ReLU
  neural network with the claimed size, width and depth bounds; to
  apply Theorem \ref{thm:cpwlrelu} we used that (the extension of)
  $\varphi_\Beta$ is defined on the \emph{convex} domain $\R^d$.
\end{proof}

Finally, Theorem~\ref{thm:simplicial} is now an easy consequence of
the above lemmata.

\begin{proof}[of Theorem~\ref{thm:simplicial}]
  With
  \begin{align}\label{eq:PhiSimplicial}
    \Phi(\Bx)\dfn \sum_{\Beta\in\CV\cap\mathring{\Omega}} f(\Beta)\varphi_\Beta(\Bx) \qquad \text{ for } \Bx \in \Omega, 
  \end{align}
  it holds that $\Phi:\Omega\to\R$ satisfies $\Phi(\Beta)=f(\Beta)$
  for all $\Beta\in\CV$.  By Lemma \ref{lemma:linex} this implies that
  $f$ equals $\Phi$ on each $\tau$, and thus $f$ equals $\Phi$ on all
  of $\Omega$.  Since each element $\tau$ is the convex hull of $d+1$
  nodes $\Beta\in\CV$, the cardinality of $\CV$ is bounded by the
  cardinality of $\CT$ times $d+1$.  Thus, the summation in
  \eqref{eq:PhiSimplicial} is over $O(|\CT|)$ terms.  Using Lemma
  \ref{lemma:addition} and Lemma \ref{lemma:basis} we obtain the
  claimed bounds on size, width, and depth of the neural network.
\end{proof}

\subsection{Size bounds for locally convex triangulations}\label{sec:convextriangulations}
Assuming local convexity of the triangulation, in this section we make
the dependence of the constants in Theorem \ref{thm:simplicial}
explicit in the dimension $d$ and in the maximal number of simplices
$k_{\CT}$ touching a node, see \eqref{eq:kT}. As such the improvement
over Theorem \ref{thm:simplicial} is modest, and the reader may choose
to skip this section on a first pass. Nonetheless, the proof,
originally from \cite{MR4087799}, is entirely constructive and gives
some further insight on how ReLU networks express functions. Let us
start by stating the required convexity constraint.

\begin{definition}
  A regular triangulation $\CT$ is called {\bf locally convex} if and
  only if $\omega(\Beta)$ is convex for all interior nodes
  $\Beta\in\CV\cap\mathring{\Omega}$.
\end{definition}

The following theorem is a variant of \cite[Theorem 3.1]{MR4087799}.

\begin{theorem}\label{thm:simplicial2}
  Let $d \in \N$, and let $\Omega\subseteq\R^d$ be a bounded domain.
  Let $\CT$ be a locally convex regular triangulation of $\Omega$.
  Let $f:\Omega\to\R$ be cpwl with respect to $\CT$ and
  $f|_{\partial\Omega}=0$.  Then, there exists a constant $C>0$
  (independent of $d$, $f$ and $\CT$) and there exists a neural
  network $\Phi^f:\Omega\to\R$ such that $\Phi^f = f$,
  \begin{align*}
    \size(\Phi^f) &\le C \cdot (1+d^2 k_{\CT} |\CT|),\\
    \wdth(\Phi^f)&\le C \cdot (1+d \log(k_{\CT})|\CT|),\\
    \depth(\Phi^f)&\le C \cdot (1+\log_2(k_{\CT})).
  \end{align*}
\end{theorem}

Assume in the following that $\CT$ is a locally convex triangulation.
We will split the proof of the theorem again into a few lemmata.
First, we will show that a convex patch can be written as an
intersection of finitely many half-spaces.  Specifically, with the
\textbf{affine hull} of a set $S$ defined as
\begin{align}\label{eq:affHull}
  \aff(S)\dfn \setc{\sum_{j=1}^n\alpha_j \Bx_j}{n\in\N,~\Bx_j\in S,~\alpha_j\in\R,~\sum_{j=1}^n\alpha_j=1}
\end{align}
let in the following for $\tau\in\CT$ and $\Beta\in V(\tau)$
\begin{align*}
  H_0(\tau,\Beta)\dfn
  {\rm aff}(V(\tau)\backslash\{\Beta\})
\end{align*}
be the affine hyperplane passing through all nodes in
$V(\tau)\backslash\{\Beta\}$, and let further
\begin{align*}
  H_+(\tau,\Beta)\dfn \set{\Bx\in\R^d}{\Bx\text{ is on the same side of }H_0(\tau,\Beta)\text{ as }\Beta}\cup H_0(\tau,\Beta).
\end{align*}

\begin{lemma}\label{lemma:convexpatch}
  Let $\Beta$ be an interior node.  Then a patch $\omega(\Beta)$ is
  convex if and only if
  \begin{align}\label{eq:omega_convex}
    \omega(\Beta)=\bigcap_{\set{\tau\in\CT}{\Beta\in\tau}}H_+(\tau,\Beta).
  \end{align}
\end{lemma}

\begin{proof}
  The right-hand side is a finite intersection of (convex)
  half-spaces, and thus itself convex.  It remains to show that if
  $\omega(\Beta)$ is convex, then \eqref{eq:omega_convex} holds.  We
  start with ``$\supseteq$''.  Suppose $\Bx\notin \omega(\Beta)$.  Then
  the straight line $\co(\{\Bx,\Beta\})$ must pass through
  $\partial\omega(\Beta)$, and by Lemma \ref{lemma:wboundary} this
  implies that there exists $\tau\in\CT$ with $\Beta\in\tau$ such that
  $\co(\{\Bx,\Beta\})$ passes through
  ${\rm aff}(V(\tau)\backslash\{\Beta\})=H_0(\tau,\Beta)$.  Hence
  $\Beta$ and $\Bx$ lie on different sides of this affine hyperplane,
  which shows ``$\supseteq$''.  Now we show ``$\subseteq$''.  Let
  $\tau\in\CT$ be such that $\Beta\in\tau$ and fix $\Bx$ in the
  complement of $H_+(\tau,\Beta)$.  Suppose that
  $\Bx\in\omega(\Beta)$.  By convexity, we then have
  $\co(\{\Bx\}\cup\tau)\subseteq\omega(\Beta)$.  This implies that
  there exists a point in $\co(V(\tau)\backslash\{\Beta\})$ belonging
  to the interior of $\omega(\Beta)$.  This contradicts Lemma
  \ref{lemma:wboundary}.  Thus, $\Bx\notin\omega(\Beta)$.
\end{proof}

The above lemma allows us to explicitly construct the basis functions
$\varphi_\Beta$ in \eqref{eq:basis}.  To see this, denote in the
following for $\tau\in\CT$ and $\Beta\in V(\tau)$ by
$g_{\tau,\Beta}\in\P_1(\R^d)$ the affine function such that
\begin{align*}
  g_{\tau,\Beta}(\Bmu)=\begin{cases}
                         1 & \text{if }\Beta=\Bmu\\
                         0 &\text{if }\Beta\neq \Bmu
                       \end{cases}\qquad \text{ for all } \Bmu\in V(\tau).
\end{align*}

This function exists and is unique by Lemma \ref{lemma:linex}.
Observe that $\varphi_\Beta(\Bx)=g_{\tau,\Beta}(\Bx)$ for all
$\Bx\in\tau$.

\begin{lemma}\label{lemma:basisconvex}
  Let $\Beta\in\CV\cap\mathring{\Omega}$ be an interior node and let
  $\omega(\Beta)$ be a convex patch.  Then
  \begin{align}\label{eq:varphiimin}
    \varphi_{\Beta}(\Bx) =
    \max\left\{0,\min_{\set{\tau\in\CT}{\Beta\in\tau}}g_{\tau,\Beta}(\Bx)\right\}\qquad \text{ for all } \Bx\in\R^d.
  \end{align}
\end{lemma}

\begin{proof}
  First let $\Bx\notin\omega(\Beta)$.  By Lemma
  \ref{lemma:convexpatch} there exists $\tau\in \CT$ with
  $\Beta\in\tau$ such that $\Bx$ is in the complement of
  $H_+(\tau,\Beta)$.  Observe that
  \begin{align}\label{eq:gijsign}
    g_{\tau,\Beta}|_{H_+(\tau,\Beta)}\ge 0 %
    \qquad\text{and}
    \qquad g_{\tau,\Beta}|_{H_+(\tau,\Beta)^c}<0.
  \end{align}
  Thus
  \begin{align*}
    \min_{\set{\tau\in\CT}{\Beta\in\tau}}g_{\tau,\Beta}(\Bx) <0\qquad\text{ for all } \Bx\in\omega(\Beta)^c,
  \end{align*}
  i.e., \eqref{eq:varphiimin} holds for all
  $\Bx\in \R^d\backslash \omega(\Beta)$.  Next, let $\tau$,
  $\tau'\in\CT$ such that $\Beta\in\tau$ and $\Beta\in\tau'$.  We wish
  to show that $g_{\tau,\Beta}(\Bx)\le g_{\tau',\Beta}(\Bx)$ for all
  $\Bx\in\tau$.  Since $g_{\tau,\Beta}(\Bx)=\varphi_{\Beta}(\Bx)$ for
  all $\Bx\in\tau$, this then concludes the proof of
  \eqref{eq:varphiimin}.  By Lemma \ref{lemma:convexpatch} it holds
  \begin{align*}
    \Bmu\in H_+(\tau',\Beta)\qquad\text{for all}\qquad\Bmu\in V(\tau).
  \end{align*}
  Hence, by \eqref{eq:gijsign}
  \begin{align*}
    g_{\tau',\Beta}(\Bmu)\ge 0 = g_{\tau,\Beta}(\Bmu)\qquad\text{for all}\qquad \Bmu \in V(\tau)\backslash\{\Beta\}.
  \end{align*}

  Moreover, $g_{\tau,\Beta}(\Beta)=g_{\tau',\Beta}(\Beta)=1$.  Thus,
  $g_{\tau,\Beta}(\Bmu)\ge g_{\tau',\Beta}(\Bmu)$ for all
  $\Bmu\in V(\tau')$ and therefore
  \begin{align*}
    g_{\tau',\Beta}(\Bx)\ge g_{\tau,\Beta}(\Bx)\qquad\text{ for all } \Bx\in\co(V(\tau'))=\tau'.
  \end{align*}
\end{proof}

\begin{proof}[of Theorem \ref{thm:simplicial2}]
  For every interior node $\Beta\in\CV\cap\mathring{\Omega}$, the cpwl
  basis function $\varphi_\Beta$ in \eqref{eq:basis} can be expressed
  as in \eqref{eq:varphiimin}, i.e.,
  \begin{align*}
    \varphi_\Beta(\Bx) = \sigma\bullet
    \nmin{|\set{\tau\in\CT}{\Beta\in\tau}|}
    \bullet
    (g_{\tau,\Beta}(\Bx))_{\set{\tau\in\CT}{\Beta\in\tau}},
  \end{align*}
  where $(g_{\tau,\Beta}(\Bx))_{\set{\tau\in\CT}{\Beta\in\tau}}$
  denotes the parallelization with shared inputs of the functions
  $g_{\tau,\Beta}(\Bx)$ for all $\tau\in\CT$ such that $\Beta\in\tau$.

  For this neural network, with
  $|\set{\tau\in\CT}{\Beta\in\tau}|\le k_\CT$, we have by Lemma
  \ref{lemma:composition}
  \begin{align}
    \size(\varphi_\Beta)&\le 4\big(\size(\sigma)+\size(\nmin{|\set{\tau\in\CT}{\Beta\in\tau}|})+\size((g_{\tau,\Beta})_{\set{\tau\in\CT}{\Beta\in\tau}})\big) \nonumber\\
                        &\le 4(2+16k_\CT+k_\CT d) \label{eq:weNeedThisLaterInTheManifoldSection}
  \end{align}
  and similarly
  \begin{align}
    \depth(\varphi_\Beta)\le 4+\lceil\log_2(k_\CT)\rceil,
    \qquad\wdth(\varphi_\Beta)\le \max\{1,3k_\CT,d\}.\label{eq:weNeedThisLaterInTheManifoldSection2}
  \end{align}

  Since for every interior node, the number of simplices touching the
  node must be larger or equal to $d$, we can assume
  $\max\{k_\CT,d\}=k_{\CT}$ in the following (otherwise there exist no
  interior nodes, and the function $f$ is constant $0$).  As in the
  proof of Theorem \ref{thm:simplicial}, the neural network

\begin{align*}
  \Phi(\Bx)\dfn \sum_{\Beta\in\CV\cap\mathring{\Omega}}f(\Beta)\varphi_\Beta(\Bx)
\end{align*}
realizes the function $f$ on all of $\Omega$.  Since the number of
nodes $|\CV|$ is bounded by $(d+1)|\CT|$, an application of Lemma
\ref{lemma:addition} yields the desired bounds.
\end{proof}

\section{Convergence rates for H\"older continuous functions}\label{sec:HoelderRates}
Theorem~\ref{thm:simplicial} immediately implies convergence rates for
certain classes of (low regularity) functions.  Recall for example the
space $C^{0,s}$ of \textbf{H\"older continuous} functions.

  \begin{definition}\label{def:hoelder}
    Let $s\in (0,1]$ and $\Omega\subseteq\R^d$. Then for
    $f:\Omega\to\R$
    \begin{align}\label{eq:Hoelder}
      \norm[C^{0,s}(\Omega)]{f}\dfn
      \sup_{\Bx\in \Omega}|f(\Bx)|+
      \sup_{\Bx\neq\By\in \Omega}\frac{|f(\Bx)-f(\By)|}{\norm[2]{\Bx-\By}^s},
    \end{align}
    and we denote by $C^{0,s}(\Omega)$ the set of functions
    $f\in C^0(\Omega)$ for which $\norm[C^{0,s}(\Omega)]{f}<\infty$.
  \end{definition}

  H\"older continuous functions can be approximated well by %
  cpwl functions. This leads to the following result.

\begin{theorem}\label{thm:hoelder}
  Let $d\in\N$. There exists a constant $C=C(d)>0$ and for every
  $N\in\N$ there exists a ReLU neural network $\Phi_N(\Bx,\Bw)$ with
  $\Bw\in\R^{N}$, such that
  \begin{align}\label{eq:sizeexactN}
    \size(\Phi_N)=N,\qquad
    \wdth(\Phi_N)\le N,\qquad      
    \depth(\Phi_N)\le C
  \end{align}
  and for every $s\in (0,1]$, $f\in C^{0,s}([0,1]^d)$
  \begin{align*}
    \inf_{\Bw\in\R^{\tilde N}}\norm[{C^0([0,1]^d)}]{f(\Bx)-\Phi_N(\Bx,\Bw)}\le
    C \norm[C^{0,s}({[0,1]^d})]{f}
    N^{-\frac{s}{d}}.
  \end{align*}
\end{theorem}

\begin{proof}
  Fix $M\ge 2$ and consider the set of nodes
  $\set{{\Bnu}/{M}}{\Bnu\in\{-1,\dots,M+1\}^d}$ where
  ${\Bnu}/{M}=({\nu_1}/{M},\dots,{\nu_d}/{M})$.  These nodes suggest a
  partition of $[-1/M,1+1/M]^d$ into $(2+M)^d$ sub-hypercubes.  Each
  such sub-hypercube can be partitioned into $d!$ simplices, such that
  we obtain a regular triangulation $\CT$ with $d!(2+M)^d$ elements on
  $[0,1]^d$.  According to Theorem \ref{thm:simplicial} there exists a
  neural network $\Phi_N$ that is cpwl with respect to $\CT$ and
  $\Phi_N^f({\Bnu}/{M})=f({\Bnu}/{M})$ whenever
  $\Bnu\in\{0,\dots,M\}^d$ and $\Phi_N^f({\Bnu}/{M})=0$ for all other
  (boundary) nodes.

  Note that the underlying architecture is independent of $f$, and we
  can write $\Phi_N^f(\Bx)=\Phi_N(\Bx,\Bw_f)$ for some $f$-dependent
  parameters $\Bw_f$. For ease of notation we simply write
  $\Phi_N(\Bx)$ in the following. It holds
  \begin{align}\label{eq:sizePhis}
    \begin{split}
      \size(\Phi_N)&\dfnn N\le C |\CT|=C d!(2+M)^{d},\\
      \wdth(\Phi_N)&\le N\le C |\CT|=C d!(2+M)^{d},\\
      \depth(\Phi_N)&\le C
    \end{split}
  \end{align}
  for a constant $C$ that only depends on $d$ (since for our regular
  triangulation $\CT$, $k_\CT$ in \eqref{eq:kT} is a fixed
  $d$-dependent constant). Here we used that the width is necessarily
  bounded by the size.

  Let us bound the error.  Fix a point $\Bx\in [0,1]^d$.  Then $\Bx$
  belongs to one of the interior simplices $\tau$ of the
  triangulation.  Two nodes of the simplex have distance at most
  \begin{align*}
    \left(\sum_{j=1}^d \left(\frac{1}{M}\right)^2 \right)^{1/2}=\frac{\sqrt{d}}{M}\dfnn\eps.
  \end{align*}
  Since $\Phi_N|_{\tau}$ is the linear interpolant of $f$ at the nodes
  $V(\tau)$ of the simplex $\tau$, $\Phi_N(\Bx)$ is a convex
  combination of the $(f(\Beta))_{\Beta\in V(\tau)}$.  Fix an
  arbitrary node $\Beta_0\in V(\tau)$.  Then
  $\norm[2]{\Bx-\Beta_0}\le\eps$ and
  \begin{align*}
    |\Phi_N(\Bx)-\Phi_N(\Beta_0)|
    &\le \max_{\Beta,\Bmu\in V(\tau)}|f(\Beta)-f(\Bmu)|\\
    &\le
      \sup_{\substack{\Bx,\By\in [0,1]^d\\\norm[2]{\Bx-\By}\le \eps}}|f(\Bx)-f(\By)|\\
    &\le \norm[C^{0,s}({[0,1]^d})]{f}
      \eps^s.
  \end{align*}
  Hence, using $f(\Beta_0)=\Phi_N(\Beta_0)$,
  \begin{align} |f(\Bx)-\Phi_N(\Bx)| &\le |f(\Bx)-f(\Beta_0)|+ |\Phi_N(\Bx)-\Phi_N(\Beta_0)|\nonumber\\
                                     &\le 2 \norm[C^{0,s}({[0,1]^d})]{f} \eps^s\nonumber\\
                                     &  = 2 \norm[C^{0,s}({[0,1]^d})]{f} d^{\frac{s}{2}} M^{-s}\nonumber\\
                                     & = 2 d^{\frac{s}{2}} \norm[C^{0,s}({[0,1]^d})]{f} (M^d)^{-\frac{s}{d}}\nonumber\\
                                     &\le \tilde C
                                       N^{-\frac{s}{d}},\label{eq:errorPhis}
  \end{align}
  with $\tilde C$ solely depending on $d$, where we used that by
  \eqref{eq:sizePhis}
  \begin{equation*}
    N\le Cd!(2+M)^d\le Cd!3^d M^d.
  \end{equation*}
  The statement now follows by \eqref{eq:sizePhis} and
  \eqref{eq:errorPhis}.
\end{proof}

The principle behind Theorem \ref{thm:hoelder} can be applied in even
more generality.  Since we can represent every cpwl function on a
regular triangulation with a neural network of size $O(N)$, where $N$
denotes the number of elements, most classical (e.g.\ finite element)
approximation theory for cpwl functions can be lifted to generate
statements about ReLU approximation.  For instance, it is well-known,
that functions in the Sobolev space $H^{2}([0,1]^d)$ can be
approximated by cpwl functions on a regular triangulation in terms of
$L^2([0,1]^d)$ with the rate ${2}/{d}$, e.g., \cite[Chapter
22]{ern2021finite}.  Similar as in the proof of
Theorem~\ref{thm:hoelder}, for %
every $N\in\N$ there then exists a ReLU neural network $\Phi_N$ such
that ${\rm{size}}(\Phi_N)=N$ and for every $f\in H^2([0,1]^d)$
\begin{align*}
  \inf_{\Bw\in\R^N}\norm[{L^2([0,1]^d)}]{f-\Phi_N(\cdot,\Bw)}\le C
  \norm[{H^2([0,1]^d)}]{f}
  N^{-\frac{2}{d}}.
\end{align*}

Finally, we %
may consider how to approximate smoother functions such as
$f\in C^k([0,1]^d)$, $k>1$, with ReLU neural networks.  As discussed
in Chapter \ref{chap:Splines} for sigmoidal activation functions,
larger $k$ can lead to faster convergence.  However, we will see in
the following chapter, that the emulation of piecewise affine
functions on regular triangulations will not yield improved
approximation rates as $k$ increases.  To %
leverage such smoothness with ReLU networks, %
in Chapter \ref{chap:DReLUNN} we will first build %
networks that emulate polynomials. Surprisingly, %
it turns out that polynomials can be approximated very efficiently by
\emph{deep} ReLU neural networks.

\section*{Bibliography and further reading}
The ReLU calculus introduced in Section \ref{sec:basic_relu} was
similarly given in \cite{petersen2018optimal}. The fact that every
cpwl function can be expressed as a maximum over a minimum of linear
functions goes back to the papers \cite{MR1091933,TARELA199917}; see
also \cite{MR1913786} for an accessible presentation of this
result. Additionally, \cite{1542439} provides sharper bounds on the
number of required nestings in such representations.

The main result of Section \ref{sec:CWPLfunctionsandRepresentations},
which shows that every cpwl function can be expressed by a ReLU
network, is then a straightforward consequence. This was first
observed in \cite{arora2018understanding}, which also provided bounds
on the network size. These bounds were significantly improved in
\cite{MR4087799} for cpwl functions on triangular meshes that satisfy
a local convexity condition. Under this assumption, it was shown that
the network size essentially only grows linearly with the number of
pieces. The paper \cite{Longo2023DeRham} showed that the convexity
assumption is not necessary for this statement to hold.  We give a
similar result in Section \ref{sec:bound_tri}, using a simpler
argument than \cite{Longo2023DeRham}. The locally convex case from
\cite{MR4087799} is separately discussed in Section
\ref{sec:convextriangulations}, as it allows for further improvements
in some constants.

The implications for the approximation of H\"older continuous
functions discussed in Section \ref{sec:HoelderRates}, follows by
standard approximation theory for cpwl functions; see for example
\cite{devore1993constructive} or the finite element literature such as
\cite{ciarlet1978finite,brenner2007mathematical,ern2021finite}, which
focus on approximation in Sobolev spaces.  Additionally,
\cite{NEURIPS2020_979a3f14} provide a stronger result, where it is
shown that ReLU networks can essentially achieve twice the rate proven
in Theorem \ref{thm:hoelder}, and this is sharp.  For a general
reference on splines and piecewise polynomial approximation see for
instance \cite{Schumaker_2007}. Finally we mention that similar
convergence results can also be shown for other activation functions,
see, e.g., %
\cite{mhaskar1993approximation}.

\newpage
\section*{Exercises}

\begin{exercise} \label{ex:polynomialsIdentityExact} Let $p:\R\to\R$
  be a polynomial of degree $n\ge 1$ (with leading coefficient
  nonzero) and let $s:\R\to\R$ be a continuous sigmoidal activation
  function.  Show that the identity map $x\mapsto x:\R\to\R$ belongs
  to $\CN_1^1(p;1,n+1)$ but not to $\CN_1^1(s;L)$ for any $L\in\N$.
\end{exercise}

\begin{exercise}
  Consider cpwl functions $f:\R\to\R$ with $n\in\N_0$ breakpoints
  (points where the function is not $C^1$).  Determine the minimal
  size required to exactly express every such $f$ with a depth-$1$
  ReLU neural network.
\end{exercise}

\begin{exercise}
  Show that, the notion of affine independence %
  is invariant under permutations of the points.
\end{exercise}
\begin{exercise}
  Let $\tau={\rm co}(\Bx_0,\dots,\Bx_d)$ be a $d$-simplex.  Show that
  the coefficients $\alpha_i\ge 0$ such that $\sum_{i=0}^d\alpha_i=1$
  and $\Bx=\sum_{i=0}^d\alpha_i \Bx_i$ are unique for every
  $\Bx\in \tau$.
\end{exercise}
\begin{exercise}\label{ex:boundary_tau}
  Let $\tau={\rm co}(\Beta_0,\dots,\Beta_d)$ be a $d$-simplex.  Show
  that the boundary of $\tau$ is given by
  $\bigcup_{i=0}^d{\rm
    co}(\{\Beta_0,\dots,\Beta_d\}\backslash\{\Beta_i\})$.
\end{exercise}

%% file: plots/minnet.tex
\begin{tikzpicture}[scale=1.2]
\draw [black,semithick] (0,-0.7) -- (1,0.35);
\draw [black,semithick] (0,-0.7) -- (1,-0.35);
\draw [black,semithick] (0,-0.7) -- (1,-1.0499999999999998);
\draw [black,semithick] (0,0) -- (1,0.35);
\draw [black,semithick] (1,0.35) -- (2,-0.35);
\draw [black,semithick] (1,-0.35) -- (2,-0.35);
\draw [black,semithick] (1,-1.0499999999999998) -- (2,-0.35);
\fill [black] (0,0) circle (1.9pt);
\fill [black] (0,-0.7) circle (1.9pt);
\fill [black] (1,0.35) circle (1.9pt);
\fill [black] (1,-0.35) circle (1.9pt);
\fill [black] (1,-1.0499999999999998) circle (1.9pt);
\fill [black] (2,-0.35) circle (1.9pt);
\node at (0,0) [left,xshift=-3] {\normalsize $x$};
\node at (0,-0.7) [left,xshift=-3] {\normalsize $y$};
\node at (2,-0.35) [right,xshift=3] {\normalsize $\min\{x,y\}$};
\end{tikzpicture}

%% file: plots/minnetn.tex
\begin{tikzpicture}[scale=1.1]
\fill[black!25,rounded corners] (1.25,-1) -- (1.25,0.47) -- (2.9499999999999997,-0.35) -- (2.9499999999999997,-1.65) -- (1.25,-2.47) -- (1.25,-1);
\fill[black!20,rounded corners] (1.25,-4.0) -- (1.25,-2.5300000000000002) -- (2.9499999999999997,-3.35) -- (2.9499999999999997,-4.65) -- (1.25,-5.470000000000001) -- (1.25,-4.0);
\node [align = center] at (-1.75,-6.0) {\small nr of connections\\ between layers:};
\node at (0.7,-6.0) {\small $2^{k-1}\cdot 4$};
\node at (2.0999999999999996,-6.0) {\small $2^{k-2}\cdot 12$};
\node at (3.5,-6.0) {\small $2^{k-3}\cdot 12$};
\node at (4.8999999999999995,-6.0) {\small $3$};
\node at (1.4,-6.0) {$\Big |$};
\node at (2.8,-6.0) {$\Big |$};
\node at (4.199999999999999,-6.0) {$\Big |$};
\draw [black,semithick] (0,-0.5) -- (1.4,0.25);
\draw [black,semithick] (0,-0.5) -- (1.4,-0.25);
\draw [black,semithick] (0,-0.5) -- (1.4,-0.75);
\draw [black,semithick] (0,0) -- (1.4,0.25);
\fill [black] (0,0) circle (1.9pt);
\fill [black] (0,-0.5) circle (1.9pt);
\fill [black] (1.4,0.25) circle (1.9pt);
\fill [black] (1.4,-0.25) circle (1.9pt);
\fill [black] (1.4,-0.75) circle (1.9pt);
\draw [black,semithick] (0,-2.0) -- (1.4,-1.25);
\draw [black,semithick] (0,-2.0) -- (1.4,-1.75);
\draw [black,semithick] (0,-2.0) -- (1.4,-2.25);
\draw [black,semithick] (0,-1.5) -- (1.4,-1.25);
\fill [black] (0,-1.5) circle (1.9pt);
\fill [black] (0,-2.0) circle (1.9pt);
\fill [black] (1.4,-1.25) circle (1.9pt);
\fill [black] (1.4,-1.75) circle (1.9pt);
\fill [black] (1.4,-2.25) circle (1.9pt);
\draw [black,semithick] (0,-3.5) -- (1.4,-2.75);
\draw [black,semithick] (0,-3.5) -- (1.4,-3.25);
\draw [black,semithick] (0,-3.5) -- (1.4,-3.75);
\draw [black,semithick] (0,-3.0) -- (1.4,-2.75);
\fill [black] (0,-3.0) circle (1.9pt);
\fill [black] (0,-3.5) circle (1.9pt);
\fill [black] (1.4,-2.75) circle (1.9pt);
\fill [black] (1.4,-3.25) circle (1.9pt);
\fill [black] (1.4,-3.75) circle (1.9pt);
\draw [black,semithick] (0,-5.0) -- (1.4,-4.25);
\draw [black,semithick] (0,-5.0) -- (1.4,-4.75);
\draw [black,semithick] (0,-5.0) -- (1.4,-5.25);
\draw [black,semithick] (0,-4.5) -- (1.4,-4.25);
\fill [black] (0,-4.5) circle (1.9pt);
\fill [black] (0,-5.0) circle (1.9pt);
\fill [black] (1.4,-4.25) circle (1.9pt);
\fill [black] (1.4,-4.75) circle (1.9pt);
\fill [black] (1.4,-5.25) circle (1.9pt);
\draw [black,semithick] (1.4,-1.25) -- (2.8,-0.5);
\draw [black,semithick] (1.4,-1.25) -- (2.8,-1.0);
\draw [black,semithick] (1.4,-1.25) -- (2.8,-1.5);
\draw [black,semithick] (1.4,-1.75) -- (2.8,-0.5);
\draw [black,semithick] (1.4,-1.75) -- (2.8,-1.0);
\draw [black,semithick] (1.4,-1.75) -- (2.8,-1.5);
\draw [black,semithick] (1.4,-2.25) -- (2.8,-0.5);
\draw [black,semithick] (1.4,-2.25) -- (2.8,-1.0);
\draw [black,semithick] (1.4,-2.25) -- (2.8,-1.5);
\draw [black,semithick] (1.4,0.25) -- (2.8,-0.5);
\draw [black,semithick] (1.4,-0.25) -- (2.8,-0.5);
\draw [black,semithick] (1.4,-0.75) -- (2.8,-0.5);
\fill [black] (1.4,0.25) circle (1.9pt);
\fill [black] (1.4,-0.25) circle (1.9pt);
\fill [black] (1.4,-0.75) circle (1.9pt);
\fill [black] (1.4,-1.25) circle (1.9pt);
\fill [black] (1.4,-1.75) circle (1.9pt);
\fill [black] (1.4,-2.25) circle (1.9pt);
\fill [black] (2.8,-0.5) circle (1.9pt);
\fill [black] (2.8,-1.0) circle (1.9pt);
\fill [black] (2.8,-1.5) circle (1.9pt);
\draw [black,semithick] (1.4,-4.25) -- (2.8,-3.5);
\draw [black,semithick] (1.4,-4.25) -- (2.8,-4.0);
\draw [black,semithick] (1.4,-4.25) -- (2.8,-4.5);
\draw [black,semithick] (1.4,-4.75) -- (2.8,-3.5);
\draw [black,semithick] (1.4,-4.75) -- (2.8,-4.0);
\draw [black,semithick] (1.4,-4.75) -- (2.8,-4.5);
\draw [black,semithick] (1.4,-5.25) -- (2.8,-3.5);
\draw [black,semithick] (1.4,-5.25) -- (2.8,-4.0);
\draw [black,semithick] (1.4,-5.25) -- (2.8,-4.5);
\draw [black,semithick] (1.4,-2.75) -- (2.8,-3.5);
\draw [black,semithick] (1.4,-3.25) -- (2.8,-3.5);
\draw [black,semithick] (1.4,-3.75) -- (2.8,-3.5);
\fill [black] (1.4,-2.75) circle (1.9pt);
\fill [black] (1.4,-3.25) circle (1.9pt);
\fill [black] (1.4,-3.75) circle (1.9pt);
\fill [black] (1.4,-4.25) circle (1.9pt);
\fill [black] (1.4,-4.75) circle (1.9pt);
\fill [black] (1.4,-5.25) circle (1.9pt);
\fill [black] (2.8,-3.5) circle (1.9pt);
\fill [black] (2.8,-4.0) circle (1.9pt);
\fill [black] (2.8,-4.5) circle (1.9pt);
\draw [black,semithick] (2.8,-3.5) -- (4.199999999999999,-2.0);
\draw [black,semithick] (2.8,-3.5) -- (4.199999999999999,-2.5);
\draw [black,semithick] (2.8,-3.5) -- (4.199999999999999,-3.0);
\draw [black,semithick] (2.8,-4.0) -- (4.199999999999999,-2.0);
\draw [black,semithick] (2.8,-4.0) -- (4.199999999999999,-2.5);
\draw [black,semithick] (2.8,-4.0) -- (4.199999999999999,-3.0);
\draw [black,semithick] (2.8,-4.5) -- (4.199999999999999,-2.0);
\draw [black,semithick] (2.8,-4.5) -- (4.199999999999999,-2.5);
\draw [black,semithick] (2.8,-4.5) -- (4.199999999999999,-3.0);
\draw [black,semithick] (2.8,-0.5) -- (4.199999999999999,-2.0);
\draw [black,semithick] (2.8,-1.0) -- (4.199999999999999,-2.0);
\draw [black,semithick] (2.8,-1.5) -- (4.199999999999999,-2.0);
\fill [black] (2.8,-0.5) circle (1.9pt);
\fill [black] (2.8,-1.0) circle (1.9pt);
\fill [black] (2.8,-1.5) circle (1.9pt);
\fill [black] (2.8,-3.5) circle (1.9pt);
\fill [black] (2.8,-4.0) circle (1.9pt);
\fill [black] (2.8,-4.5) circle (1.9pt);
\fill [black] (4.199999999999999,-2.0) circle (1.9pt);
\fill [black] (4.199999999999999,-2.5) circle (1.9pt);
\fill [black] (4.199999999999999,-3.0) circle (1.9pt);
\draw [black,semithick] (4.199999999999999,-2.0) -- (5.6,-2.5);
\draw [black,semithick] (4.199999999999999,-2.5) -- (5.6,-2.5);
\draw [black,semithick] (4.199999999999999,-3.0) -- (5.6,-2.5);
\fill [black] (5.6,-2.5) circle (1.9pt);
\node at (-0.1,0.0) [left] {\normalsize $x_1$};
\node at (-0.1,-0.5) [left] {\normalsize $x_2$};
\node at (-0.1,-1.5) [left] {\normalsize $x_3$};
\node at (-0.1,-2.0) [left] {\normalsize $x_4$};
\node at (-0.1,-3.0) [left] {\normalsize $x_5$};
\node at (-0.1,-3.5) [left] {\normalsize $x_6$};
\node at (-0.1,-4.5) [left] {\normalsize $x_7$};
\node at (-0.1,-5.0) [left] {\normalsize $x_8$};
\node at (5.699999999999999,-2.5) [right] {\normalsize $\min\{x_1,\dots,x_8\}$};
\end{tikzpicture}

%% file: plots/minnetgeneral.tex
\begin{tikzpicture}[scale=0.9]
\fill (0.0,0) circle (2.1pt) node [above,yshift=1] {$x_1$};
\fill (0.6,0) circle (2.1pt) node [above,yshift=1] {$x_2$};
\fill (1.2,0) circle (2.1pt) node [above,yshift=1] {$x_3$};
\fill (1.7999999999999998,0) circle (2.1pt) node [above,yshift=1] {$x_4$};
\fill (2.4,0) circle (2.1pt) node [above,yshift=1] {$x_5$};
\draw [semithick] (0.0,0) -- (0.3,-1);
\draw [semithick] (0.6,0) -- (0.3,-1);
\draw [semithick] (1.2,0) -- (1.2,-1);
\draw [semithick] (1.7999999999999998,0) -- (2.0999999999999996,-1);
\draw [semithick] (2.4,0) -- (2.0999999999999996,-1);
\draw [semithick] (0.3,-1) -- (0.3,-2);
\draw [semithick] (1.2,-1) -- (1.65,-2);
\draw [semithick] (2.0999999999999996,-1) -- (1.65,-2);
\draw [semithick] (0.3,-2) -- (0.975,-3);
\draw [semithick] (1.65,-2) -- (0.975,-3);
\draw [semithick] (0.975,-3) -- (0.975,-4) node [below,yshift=-1] {$\min\{x_1,\dots,x_{5}\}$};
\fill (0.975,-4) circle (2.1pt);
\node[fill=white,draw, rounded corners, minimum width=0.5cm, minimum height=0.5cm,semithick] at (0.3,-1) {$\small\nmin{2}$};
\node[draw, rounded corners, minimum width=0.5cm, minimum height=0.5cm,fill=black!25,semithick] at (1.2,-1) {$\small\nId{1}$};
\node[fill=white,draw, rounded corners, minimum width=0.5cm, minimum height=0.5cm,semithick] at (2.0999999999999996,-1) {$\small\nmin{2}$};
\node[draw, rounded corners, minimum width=0.5cm, minimum height=0.5cm,fill=black!25,semithick] at (0.3,-2) {$\small\nId{1}$};
\node[fill=white,draw, rounded corners, minimum width=0.5cm, minimum height=0.5cm,semithick] at (1.65,-2) {$\small\nmin{2}$};
\node[fill=white,draw, rounded corners, minimum width=0.5cm, minimum height=0.5cm,semithick] at (0.975,-3) {$\small\nmin{2}$};
\fill (3.9,0) circle (2.1pt) node [above,yshift=1] {$x_1$};
\fill (4.5,0) circle (2.1pt) node [above,yshift=1] {$x_2$};
\fill (5.1,0) circle (2.1pt) node [above,yshift=1] {$x_3$};
\fill (5.699999999999999,0) circle (2.1pt) node [above,yshift=1] {$x_4$};
\fill (6.3,0) circle (2.1pt) node [above,yshift=1] {$x_5$};
\fill (6.9,0) circle (2.1pt) node [above,yshift=1] {$x_6$};
\draw [semithick] (3.9,0) -- (3.9,-1);
\draw [semithick] (4.5,0) -- (4.8,-1);
\draw [semithick] (5.1,0) -- (4.8,-1);
\draw [semithick] (5.699999999999999,0) -- (5.699999999999999,-1);
\draw [semithick] (6.3,0) -- (6.6,-1);
\draw [semithick] (6.9,0) -- (6.6,-1);
\draw [semithick] (3.9,-1) -- (4.35,-2);
\draw [semithick] (4.8,-1) -- (4.35,-2);
\draw [semithick] (5.699999999999999,-1) -- (6.1499999999999995,-2);
\draw [semithick] (6.6,-1) -- (6.1499999999999995,-2);
\draw [semithick] (4.35,-2) -- (5.25,-3);
\draw [semithick] (6.1499999999999995,-2) -- (5.25,-3);
\draw [semithick] (5.25,-3) -- (5.25,-4) node [below,yshift=-1] {$\min\{x_1,\dots,x_{6}\}$};
\fill (5.25,-4) circle (2.1pt);
\node[draw, rounded corners, minimum width=0.5cm, minimum height=0.5cm,fill=black!25,semithick] at (3.9,-1) {$\small\nId{1}$};
\node[fill=white,draw, rounded corners, minimum width=0.5cm, minimum height=0.5cm,semithick] at (4.8,-1) {$\small\nmin{2}$};
\node[draw, rounded corners, minimum width=0.5cm, minimum height=0.5cm,fill=black!25,semithick] at (5.699999999999999,-1) {$\small\nId{1}$};
\node[fill=white,draw, rounded corners, minimum width=0.5cm, minimum height=0.5cm,semithick] at (6.6,-1) {$\small\nmin{2}$};
\node[fill=white,draw, rounded corners, minimum width=0.5cm, minimum height=0.5cm,semithick] at (4.35,-2) {$\small\nmin{2}$};
\node[fill=white,draw, rounded corners, minimum width=0.5cm, minimum height=0.5cm,semithick] at (6.1499999999999995,-2) {$\small\nmin{2}$};
\node[fill=white,draw, rounded corners, minimum width=0.5cm, minimum height=0.5cm,semithick] at (5.25,-3) {$\small\nmin{2}$};
\fill (8.399999999999999,0) circle (2.1pt) node [above,yshift=1] {$x_1$};
\fill (8.999999999999998,0) circle (2.1pt) node [above,yshift=1] {$x_2$};
\fill (9.599999999999998,0) circle (2.1pt) node [above,yshift=1] {$x_3$};
\fill (10.2,0) circle (2.1pt) node [above,yshift=1] {$x_4$};
\fill (10.799999999999999,0) circle (2.1pt) node [above,yshift=1] {$x_5$};
\fill (11.399999999999999,0) circle (2.1pt) node [above,yshift=1] {$x_6$};
\fill (11.999999999999998,0) circle (2.1pt) node [above,yshift=1] {$x_7$};
\fill (12.599999999999998,0) circle (2.1pt) node [above,yshift=1] {$x_8$};
\draw [semithick] (8.399999999999999,0) -- (8.7,-1);
\draw [semithick] (8.999999999999998,0) -- (8.7,-1);
\draw [semithick] (9.599999999999998,0) -- (9.899999999999999,-1);
\draw [semithick] (10.2,0) -- (9.899999999999999,-1);
\draw [semithick] (10.799999999999999,0) -- (11.099999999999998,-1);
\draw [semithick] (11.399999999999999,0) -- (11.099999999999998,-1);
\draw [semithick] (11.999999999999998,0) -- (12.299999999999997,-1);
\draw [semithick] (12.599999999999998,0) -- (12.299999999999997,-1);
\draw [semithick] (8.7,-1) -- (9.299999999999999,-2);
\draw [semithick] (9.899999999999999,-1) -- (9.299999999999999,-2);
\draw [semithick] (11.099999999999998,-1) -- (11.699999999999998,-2);
\draw [semithick] (12.299999999999997,-1) -- (11.699999999999998,-2);
\draw [semithick] (9.299999999999999,-2) -- (10.499999999999998,-3);
\draw [semithick] (11.699999999999998,-2) -- (10.499999999999998,-3);
\draw [semithick] (10.499999999999998,-3) -- (10.499999999999998,-4) node [below,yshift=-1] {$\min\{x_1,\dots,x_{8}\}$};
\fill (10.499999999999998,-4) circle (2.1pt);
\node[fill=white,draw, rounded corners, minimum width=0.5cm, minimum height=0.5cm,semithick] at (8.7,-1) {$\small\nmin{2}$};
\node[fill=white,draw, rounded corners, minimum width=0.5cm, minimum height=0.5cm,semithick] at (9.899999999999999,-1) {$\small\nmin{2}$};
\node[fill=white,draw, rounded corners, minimum width=0.5cm, minimum height=0.5cm,semithick] at (11.099999999999998,-1) {$\small\nmin{2}$};
\node[fill=white,draw, rounded corners, minimum width=0.5cm, minimum height=0.5cm,semithick] at (12.299999999999997,-1) {$\small\nmin{2}$};
\node[fill=white,draw, rounded corners, minimum width=0.5cm, minimum height=0.5cm,semithick] at (9.299999999999999,-2) {$\small\nmin{2}$};
\node[fill=white,draw, rounded corners, minimum width=0.5cm, minimum height=0.5cm,semithick] at (11.699999999999998,-2) {$\small\nmin{2}$};
\node[fill=white,draw, rounded corners, minimum width=0.5cm, minimum height=0.5cm,semithick] at (10.499999999999998,-3) {$\small\nmin{2}$};
\end{tikzpicture}

%% file: plots/triangulation.tex
\begin{tikzpicture}\filldraw [pattern=north east lines,pattern color = gray] (1,0) -- (0,1) -- (0,0);
\filldraw [pattern=north west lines,pattern color = gray] (0,1) -- (-1,0) -- (0,0);
\filldraw [pattern=vertical lines,pattern color = gray] (-1,0) -- (0,-1) -- (0,0);
\draw [] (1,0) -- (0,1) -- (0,0) -- cycle;
\node at (0,-2.0) {\small $\tau_{1}=\co({\Beta_1,\Beta_2,\Beta_5})$};
\draw [] (0,1) -- (-1,0) -- (0,0) -- cycle;
\node at (0,-2.4) {\small $\tau_{2}=\co({\Beta_2,\Beta_3,\Beta_5})$};
\draw [] (-1,0) -- (0,-1) -- (0,0) -- cycle;
\node at (0,-2.8) {\small $\tau_{3}=\co({\Beta_3,\Beta_4,\Beta_5})$};
\draw [thick] (1,0) -- (0,1) -- (-1,0) -- (0,-1) -- (0,0) -- (1,0);
\fill (1,0) circle (1.9pt);
\fill (0,1) circle (1.9pt);
\fill (-1,0) circle (1.9pt);
\fill (0,-1) circle (1.9pt);
\fill (0,0) circle (1.9pt);
\node [xshift=7] at (1,0) {$\Beta_{1}$};
\node [yshift=7] at (0,1) {$\Beta_{2}$};
\node [xshift=-7] at (-1,0) {$\Beta_{3}$};
\node [yshift=-7] at (0,-1) {$\Beta_{4}$};
\node [xshift=7,yshift=-7] at (0,0) {$\Beta_{5}$};
\filldraw [pattern=north east lines,pattern color = gray] (4,1) -- (3,0) -- (4,-1);
\filldraw [pattern=north west lines,pattern color = gray] (4,1) -- (4,0) -- (5,0);
\draw [] (4,1) -- (3,0) -- (4,-1) -- cycle;
\node at (4,-2.0) {\small $\tau_{1}=\co({\Beta_2,\Beta_3,\Beta_4})$};
\draw [] (4,1) -- (4,0) -- (5,0) -- cycle;
\node at (4,-2.4) {\small $\tau_{2}=\co({\Beta_2,\Beta_5,\Beta_1})$};
\draw [thick] (5,0) -- (4,1) -- (3,0) -- (4,-1) -- (4,0) -- (5,0);
\fill (5,0) circle (1.9pt);
\fill (4,1) circle (1.9pt);
\fill (3,0) circle (1.9pt);
\fill (4,-1) circle (1.9pt);
\fill (4,0) circle (1.9pt);
\node [xshift=7] at (5,0) {$\Beta_{1}$};
\node [yshift=7] at (4,1) {$\Beta_{2}$};
\node [xshift=-7] at (3,0) {$\Beta_{3}$};
\node [yshift=-7] at (4,-1) {$\Beta_{4}$};
\node [xshift=7,yshift=-7] at (4,0) {$\Beta_{5}$};
\filldraw [pattern=north east lines,pattern color = gray] (8,1) -- (7,0) -- (8,-1);
\filldraw [pattern=north west lines,pattern color = gray] (9,0) -- (8,1) -- (7,0);
\draw [] (8,1) -- (7,0) -- (8,-1) -- cycle;
\node at (8,-2.0) {\small $\tau_{1}=\co({\Beta_2,\Beta_3,\Beta_4})$};
\draw [] (9,0) -- (8,1) -- (7,0) -- cycle;
\node at (8,-2.4) {\small $\tau_{2}=\co({\Beta_1,\Beta_2,\Beta_3})$};
\draw [thick] (9,0) -- (8,1) -- (7,0) -- (8,-1) -- (8,0) -- (9,0);
\fill (9,0) circle (1.9pt);
\fill (8,1) circle (1.9pt);
\fill (7,0) circle (1.9pt);
\fill (8,-1) circle (1.9pt);
\node [xshift=7] at (9,0) {$\Beta_{1}$};
\node [yshift=7] at (8,1) {$\Beta_{2}$};
\node [xshift=-7] at (7,0) {$\Beta_{3}$};
\node [yshift=-7] at (8,-1) {$\Beta_{4}$};
\end{tikzpicture}

%% file: NumberOfPieces.tex
\chapter{Affine pieces for ReLU neural networks}\label{chap:AffPieces}
In the previous chapters, we observed some remarkable approximation results of shallow ReLU neural networks. 
In practice, however, %
deeper architectures are more common.
To understand why, in this chapter we discuss some potential shortcomings of shallow ReLU networks compared to deep ReLU networks. 

Traditionally, an insightful approach to study limitations of ReLU neural networks has been to analyze the
number of linear regions %
these functions can generate. 

\begin{definition}\label{def:NumOfPieces}
  Let $d\in \N$, $\Omega \subseteq \R^d$, and let
  $f\colon \Omega \to \R$ be cpwl (see Definition \ref{def:cpwl}).  We
  say that $f$ has $p \in \N$ {\bf pieces (or linear regions)}, if
  $p$ is the smallest number of connected open sets
  $(\Omega_i)_{i=1}^p$ such that
  $\bigcup_{i=1}^p \overline{\Omega_i} = \Omega$, and $f|_{\Omega_i}$
  is an affine function for all $i = 1, \dots, p$.  We denote
  ${\rm Pieces}(f, \Omega) \coloneqq p$.

For $d=1$ we call every point where $f$ is not differentiable a
\textbf{break point} of $f$. 
\end{definition}

To get an accurate cpwl approximation of a function, the approximating function needs to have many pieces. The next theorem, corresponding to \cite[Theorem 2]{frenzen2010number}, quantifies this statement.

\begin{theorem}
  \label{thm:bestApproximationBypiecewiseaffineThings}
Let $-\infty<a<b<\infty$ and $f \in C^3([a,b])$ so that $f$ is not
affine.
Then there exists a constant
$C>0$ depending only on $\int_{a}^b\sqrt{|f''(x)|}\dd x$ so that
\[
	\norm[{L^\infty([a,b])}]{g-f} > C p^{-2}
\]
for all cpwl $g$ %
with at most $p\in\N$ pieces.
\end{theorem}

The proof of the theorem is left to the reader, see Exercise \ref{ex:peps}.

Theorem \ref{thm:bestApproximationBypiecewiseaffineThings} implies that %
for ReLU neural networks we need architectures allowing for
many pieces, if we want to approximate non-linear functions to high accuracy. 
How many pieces can we create %
for a fixed depth and width? We %
establish
a simple theoretical upper bound in %
Section \ref{sec:UpperBoundOnPieces}.
Subsequently, we %
investigate
under which conditions these upper bounds are attainable in Section \ref{sec:tightness}. 
Lastly, in Section  \ref{sec:NumOfPieceInPractice}, we will discuss the practical relevance of this analysis by examining how many pieces ``typical'' neural networks possess.
Surprisingly, %
it turns out that randomly initialized deep neural networks on average do not have a number of pieces that is anywhere close to the %
theoretically achievable maximum.

\section{%
Upper bounds}\label{sec:UpperBoundOnPieces}

\begin{figure}[htb]
\centering
\includegraphics[width = 0.9\textwidth]{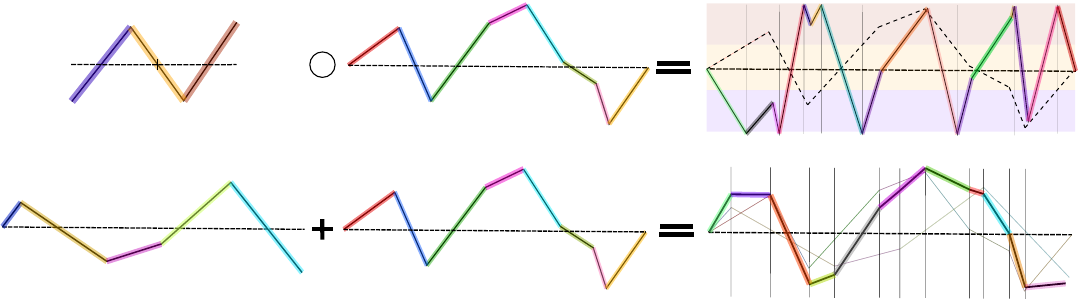}
\caption{%
\textbf{Top:} Composition of two cpwl functions $f_1 \circ f_2$ can create a piece whenever the value of $f_2$ crosses a level that is associated to a break point of $f_1$.
\textbf{Bottom:} Addition of two cpwl functions $f_1 + f_2$ produces a cpwl function that can have break points at positions where either $f_1$ or $f_2$ has a break point.}
\label{fig:numberOfPiecesGeneration}
\end{figure}

Neural networks are based on the composition and addition of neurons.
These two operations %
increase the possible number of pieces in a very specific way.
Figure \ref{fig:numberOfPiecesGeneration} depicts the two operations and their effect.
They can be described as follows:
\begin{itemize}
\item \textit{Summation:} Let $\Omega\subseteq\R$. The sum of two %
  cpwl functions $f_1$, $f_2:\Omega\to\R$ satisfies  
\begin{align}\label{eq:upperBoundPiecesSum}
	{\rm Pieces}(f_1 + f_2, \Omega) \leq {\rm Pieces}(f_1, \Omega) + {\rm Pieces}(f_2, \Omega)-1.
\end{align}
This %
holds because the sum is affine in every point where both $f_1$ and $f_2$ are affine.
Therefore, the sum has at most as many break points as $f_1$ and $f_2$ combined.
Moreover, the number of pieces of a univariate function equals the number of its break points plus one.
\item \textit{Composition:}
  Let again $\Omega \subseteq \R$.
  The composition of two functions $f_1 \colon \R^d \to \R$ and $f_2\colon \Omega \to \R^d$ satisfies  
\begin{align}\label{eq:upperBoundPiecesComposition}
	{\rm Pieces}(f_1 \circ f_2, \Omega) \leq {\rm Pieces}(f_1, \R^d) \cdot {\rm Pieces}(f_2, \Omega).
\end{align}
This is because for each of the affine pieces of $f_2$---let us call one of those pieces $A \subseteq \R$---we have that $f_2$ is either constant or injective on $A$.
If it is constant, then $f_1 \circ f_2$ is constant.
If it is injective, then ${\rm Pieces}(f_1 \circ f_2, A) = {\rm Pieces}(f_1, f_2(A)) \leq  {\rm Pieces}(f_1, \R^d)$.
Since %
this holds for all pieces of $f_2$ we %
get \eqref{eq:upperBoundPiecesComposition}.
\end{itemize} 

These considerations give the following result, which follows the argument of \cite[Lemma 2.1]{telgarsky_depth}.
We state it for general cpwl activation functions.
The ReLU activation function corresponds to $p=2$.
Recall that the notation $(\sigma;d_0,\dots,d_{L+1})$ denotes
the architecture of a feedforward neural network, see Definition \ref{def:nn}.

\begin{theorem}\label{thm:NumberOfPiecesTelgarskyStyle}
Let $L \in \N$.
Let $\sigma$ be cpwl with $p$ pieces.
Then, %
every neural network %
with architecture
$(\sigma;1, d_1, \dots, d_L, 1)$ %
has at most $(p \cdot \wdth(\Phi))^{L}$ pieces.
\end{theorem}

\begin{proof}
The proof is via induction over the depth $L$.
Let $L=1$, and let $\Phi:\R\to\R$ be a neural network of architecture $(\sigma;1,d_1,1)$. Then
\[
	\Phi(x) = \sum_{k = 1}^{d_1} w_k^{(1)} \sigma(w_k^{(0)} x + b_{k}^{(0)}) + b^{(1)}\qquad\text{for }x\in\R,
      \]
      for certain $\Bw^{(0)}$, $\Bw^{(1)}$, $\Bb^{(0)}\in\R^{d_1}$ and $b^{(1)}\in\R$.
      By \eqref{eq:upperBoundPiecesSum}, %
      ${\rm Pieces}(\Phi)\le p \cdot \wdth(\Phi)$.

For the induction step, assume the statement holds for $L\in \N$,
and let $\Phi:\R \to\R$ be a neural network of architecture
$(\sigma;1,d_1,\dots,d_{L+1},1)$.
Then, we can write
\[
	\Phi(x) = \sum_{j=1}^{d_{L+1}}w_j\sigma(h_j(x)) + b\qquad\text{for }x\in\R,
      \]
for some $\Bw\in\R^{d_{L+1}}$, $b\in\R$,  and where
each $h_j$ is a neural network of architecture $(\sigma;1,d_1,\dots,d_{L},1)$.
Using the induction hypothesis, %
each $\sigma  \,  \circ h_\ell$ has at most $p \cdot  (p  \cdot \wdth(\Phi))^{L}$ affine pieces.
Hence $\Phi$ has at most $ \wdth(\Phi) \cdot p \cdot (p\cdot \wdth(\Phi))^{L} = (p \cdot  \wdth(\Phi))^{L+1}$ %
affine pieces.
This completes the proof.
\end{proof}

Theorem \ref{thm:NumberOfPiecesTelgarskyStyle} shows that there are limits to how many pieces can be created with a certain architecture. 
It is noteworthy that the effects of the depth and the width of a neural network are vastly different. 
While increasing the width %
can polynomially increase the number of pieces, increasing the depth can result in exponential increase.
This is a first indication of the prowess of depth of neural networks. 

To understand the effect of this on the approximation problem, we apply the bound of Theorem \ref{thm:NumberOfPiecesTelgarskyStyle} to Theorem \ref{thm:bestApproximationBypiecewiseaffineThings}. 

\begin{theorem}\label{thm:FrenzenApplied}
  Let $d_0 \in \N$ and $f \in C^3([0,1]^{d_0})$. %
  Assume there exists a line segment 
  $\mathfrak{s} \subseteq [0,1]^{d_0}$
  of positive length such that
  $0<c\dfn \int_\mathfrak{s} \sqrt{|f''(x)|}\dd x$.
  Then, there exists  $C>0$ solely depending on $c$,
  such that for all ReLU neural networks $\Phi:\R^{d_0}\to\R$ with $L$ hidden layers
	\[
          \norm[{L^\infty([0,1]^{d_0})}]{f-\Phi} \geq C \cdot (2 \wdth(\Phi))^{-2L}.
	\]
\end{theorem}

Theorem \ref{thm:FrenzenApplied} %
gives a lower bound on achievable approximation rates in dependence of the depth $L$.
As target functions become smoother, we expect that we can achieve faster convergence rates (cp.~Chapter \ref{chap:Splines}).
However, without increasing the depth, it seems to be impossible to leverage such additional smoothness. 

This observation strongly indicates that deeper architectures %
can be superior. %
Before making this more concrete,
we first explore whether the upper bounds of Theorem
\ref{thm:NumberOfPiecesTelgarskyStyle} are %
also achievable.

\section{Tightness of upper bounds}\label{sec:tightness}
\begin{figure}
\begin{center}
\begin{tikzpicture}[scale=1.7]
  \draw [semithick,->] (-0.1,0) -- (1.2,0);
  \draw [semithick,->] (0,-0.1) -- (0,1.2);
  \draw [thick] (0,0) -- (0.5,1) -- (1,0);
  \node at (0,-0.06) [below] {\footnotesize $0$};
  \draw (1,0.06) -- (1,-0.06) node [below] {\footnotesize $1$};
  \draw (0.06,1) -- (-0.06,1) node [left] {\footnotesize $1$};
  \node at (0.5,1.2) {$h_1$};

  \draw [semithick,->] (-0.1+2.2,0) -- (1.2+2.2,0);
  \draw [semithick,->] (0+2.2,-0.1) -- (0+2.2,1.2);
  \draw [thick] (0+2.2,0) -- (0.25+2.2,1) -- (0.5+2.2,0) -- (0.75+2.2,1) -- (1+2.2,0);
  \node at (0+2.2,-0.06) [below] {\footnotesize $0$};
  \draw (1+2.2,0.06) -- (1+2.2,-0.06) node [below] {\footnotesize $1$};
  \draw (0.06+2.2,1) -- (-0.06+2.2,1) node [left] {\footnotesize $1$};
  \node at (0.5+2.2,1.2) {$h_2$};

  \draw [semithick,->] (-0.1+4.4,0) -- (1.2+4.4,0);
  \draw [semithick,->] (0+4.4,-0.1) -- (0+4.4,1.2);
  \draw [thick] (0,0) -- (0.5,1) -- (1,0);
  \draw [thick] (0+4.4,0) -- (0.125+4.4,1) -- (0.25+4.4,0) -- (0.375+4.4,1) -- (0.5+4.4,0) -- (0.625+4.4,1) -- (0.75+4.4,0) -- (0.875+4.4,1) -- (1+4.4,0);
  \node at (0+4.4,-0.06) [below] {\footnotesize $0$};
  \draw (1+4.4,0.06) -- (1+4.4,-0.06) node [below] {\footnotesize $1$};
  \draw (0.06+4.4,1) -- (-0.06+4.4,1) node [left] {\footnotesize $1$};
  \node at (0.5+4.4,1.2) {$h_3$};
\end{tikzpicture}
\end{center}
\caption{The functions $h_n$ in Lemma \ref{lemma:hn}.}\label{fig:hn}
\end{figure}
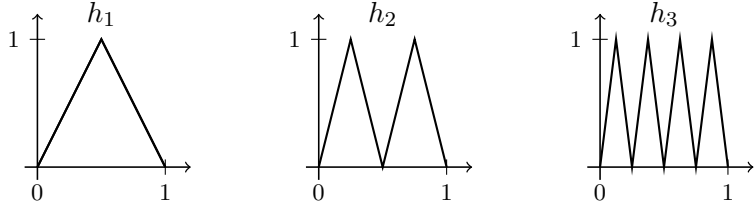

We follow \cite{telgarsky_depth} to construct a ReLU neural network, that realizes the upper bound of Theorem \ref{thm:NumberOfPiecesTelgarskyStyle}. %
First let $h_1:[0,1]\to\R$ be the hat function
\begin{align*}
	h_1(x)\dfn
	\begin{cases}
	  2x &\text{if }x\in [0,\frac{1}{2}]\\
	  2-2x &\text{if }x\in [\frac{1}{2},1].
	\end{cases}
\end{align*}
This function can be expressed by a ReLU neural network of depth one and with two nodes
\begin{subequations}\label{eq:hn}
\begin{align}\label{eq:h1}
	h_1(x)= \sigma_{\rm ReLU}(2x)-\sigma_{\rm ReLU}(4x-2)\qquad\text{ for all } x\in [0,1].
\end{align}
We recursively set
\begin{align}
  h_n\dfn h_{n-1}\circ h_1\qquad\text{for all }n\ge 2,
\end{align}
\end{subequations}
i.e., $h_n=h_1\circ\cdots\circ h_1$ is the $n$-fold composition of
$h_1$.
Since $h_1:[0,1]\to [0,1]$, we have $h_n:[0,1]\to [0,1]$
and
\begin{align*}
	h_n\in\CN_1^1(\sigma_{\rm ReLU};n,2).
\end{align*}
It turns out that this function has a rather interesting behavior.
It is a ``sawtooth'' function with $2^{n-1}$ spikes, see Figure~\ref{fig:hn}.
\begin{lemma}\label{lemma:hn}
Let $n \in \N$.
It holds for all $x\in[0,1]$
\begin{align*}
	  h_n(x) = \begin{cases}
	    2^n(x-i2^{-n}) &\text{if $i\ge 0$ is even and }x\in [i2^{-n},(i+1)2^{-n}]\\
	    2^n((i+1)2^{-n}-x) &\text{if $i\ge 1$ is odd and }x\in [i2^{-n},(i+1)2^{-n}].
	    \end{cases}
\end{align*}
\end{lemma}
\begin{proof}
The case $n=1$ holds by definition.
We proceed by induction, and assume
the statement holds for $n$.
Let $x\in[0,{1}/{2}]$
and $i\ge 0$ even such that $x\in [i2^{-(n+1)},(i+1)2^{-(n+1)}]$.
Then
$2x\in [i2^{-n},(i+1)2^{-n}]$.
Thus
\begin{align*}
	  h_{n}(h_1(x))=h_n(2x) = 2^{n}(2x-i2^{-n})
	  = 2^{n+1}(x-i2^{-n+1}).
\end{align*}
Similarly, if
$x\in[0,{1}/{2}]$ and $i\ge 1$ odd such that
$x\in [i2^{-(n+1)},(i+1)2^{-(n+1)}]$,
then $h_1(x)=2x\in [i2^{-n},(i+1)2^{-n}]$
and
\begin{align*}
	  h_{n}(h_1(x))=h_n(2x) = 2^{n}(2x-(i+1)2^{-n})
	  = 2^{n+1}(x-(i+1)2^{-n+1}).
\end{align*}
The case $x\in[{1}/{2},1]$ follows by observing that 
$h_{n+1}$ is symmetric around ${1}/{2}$.
\end{proof}

The neural network $h_n$ has size $O(n)$ and is piecewise linear on at least $2^n$ pieces.
This shows that the number of pieces can indeed increase exponentially in the neural network size, also see the upper bound in Theorem \ref{thm:NumberOfPiecesTelgarskyStyle}.

\section{Number of pieces in practice}\label{sec:NumOfPieceInPractice}

\begin{figure}
\includegraphics[width = 0.45\textwidth]{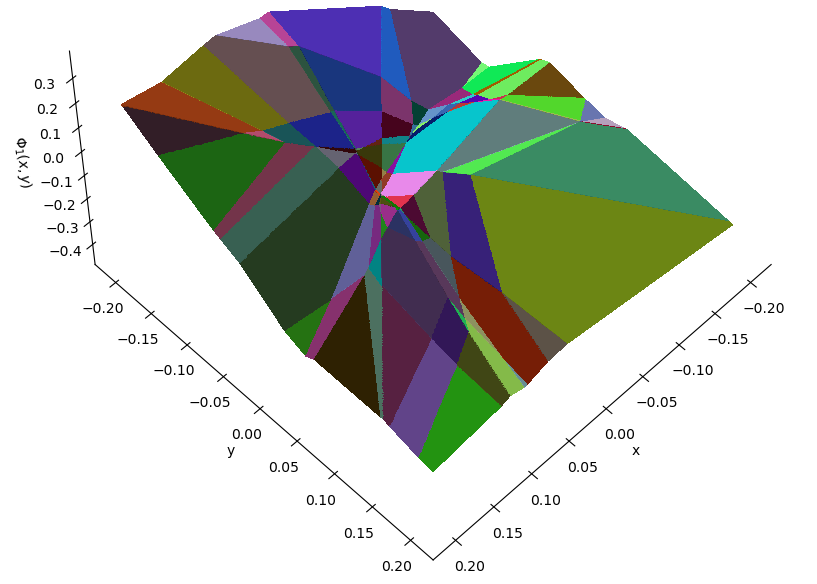} \includegraphics[width = 0.45\textwidth]{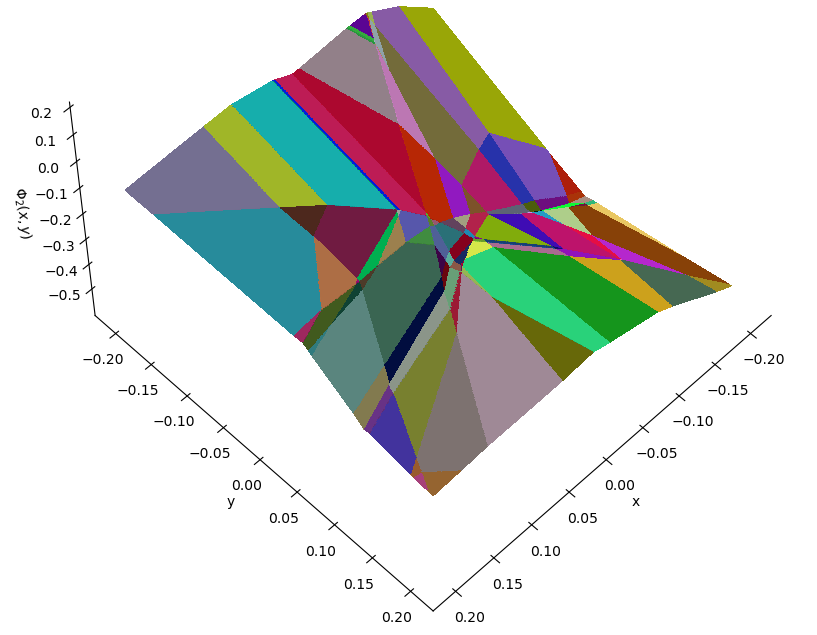}
\caption{%
Two randomly initialized neural networks $\Phi_1$ and $\Phi_2$ with architectures $(\sigma_{\rm ReLU};2, 10, 10, 1)$ and $(\sigma_{\rm ReLU};2,5,5,5,5,5,1)$.
The initialization scheme was He initialization \cite{he2015delving}. The number of linear regions equals $114$ and $110$, respectively.
}
\label{fig:NumPiecesDeepVsShallow}
\end{figure}

We have seen in Theorem \ref{thm:NumberOfPiecesTelgarskyStyle} that deep neural networks \emph{can} have many more pieces than their shallow counterparts.
This begs the question if deep neural networks tend to generate more pieces in practice.
More formally: If we %
randomly initialize the
weights of a neural network, what is the expected number of linear regions?
Will this number scale exponentially with the depth? 
This question was analyzed in \cite{hanin2019complexity}, %
and surprisingly, it was found that
the number of pieces of randomly initialized neural networks typically does \emph{not} depend exponentially on the depth. 
In Figure \ref{fig:NumPiecesDeepVsShallow}, we depict two neural networks, one shallow and one deep, that were randomly initialized according to He initialization \cite{he2015delving}. 
Both neural networks have essentially the same number of pieces ($114$ and $110$) and there is no clear indication that one has a deeper architecture than the other. 

In the following, we will give a simplified version of the main result of \cite{hanin2019complexity} to show why random deep neural networks often behave like shallow neural networks. 

We recall from Figure \ref{fig:numberOfPiecesGeneration} that pieces are generated through composition of two functions $f_1$ and $f_2$, if the values of $f_2$ cross a level that is associated to a break point of $f_1$. 
In the case of a simple neuron of the form 
\begin{align*}
	\Bx \mapsto \sigma_{\rm ReLU}(\langle \Ba, h(\Bx) \rangle + b)
\end{align*}
where $h$ is a cpwl function, $\Ba$ is a vector, %
and $b$  is a scalar, %
many pieces can be generated if $\langle\Ba, h(\Bx) \rangle $ crosses the $-b$ level
often.

If $\Ba$, $b$ are random variables, and we know that $h$ does not oscillate too much, then we can quantify the probability of $\langle \Ba, h(\Bx) \rangle $ crossing the $-b$ level often. 
The following lemma from \cite[Lemma 3.1]{karner2022limitations} provides the details.

\begin{lemma}%
  \label{lem:probabilityOfHittingALevel}
	Let $c >0$ and let $h\colon [0,c] \to \R$ be a cpwl function on $[0,c]$.
        Let $t \in \N$, let $A\subseteq\R$ be a Lebesgue measurable set, and assume that for every $y \in A$
	\[
          |\set{x \in [0,c]}{h(x) = y}| \geq t.
	\]
        
	Then,
        $c \norm[L^\infty]{h'} \geq \norm[L^1]{h'} \geq %
          |A| \cdot t$, where $|A|$ is the Lebesgue measure of $A$.
	In particular, %
        if $h$ has at most $P\in \N$ pieces and %
        $\norm[L^1]{h'}<\infty$, then %
        for all $\delta>0$, $t\le P$
	\begin{align*}
		\mathbb{P}\left[|\set{x \in [0,c]}{h(x) = U }| \geq t\right] &\leq \frac{\norm[L^1]{h'}}{\delta t},\\
		\mathbb{P}\left[|\set{x \in [0,c]}{h(x) = U}| > P \right] &= 0,
	\end{align*}
where $U$ is a uniformly distributed variable on $[-\delta/2, \delta/2]$.
\end{lemma}

\begin{proof}
	We will assume $c = 1$.
        The general case then follows by considering $\tilde{h}(x) = h(x/c)$. 
	
	Let for $(c_i)_{i=1}^{P+1} \subseteq [0,1]$ with $c_1 = 0$, $c_{P+1} = 1$ and $c_i \leq c_{i+1}$ for all $i = 1,\dots, P+1$ the pieces of $h$ be given by $( (c_i, c_{i+1}) )_{i=1}^P$. 
	We denote 
	\begin{align*}
		V_1 \coloneqq [0, c_2], \quad V_i \coloneqq (c_i, c_{i+1}] \text{ for } i = 1, \dots, P
	\end{align*}
	and for $i =1, \dots, P+1$
	\begin{align*}
		\widetilde{V}_i \coloneqq \bigcup_{j=1}^{i-1} V_j.
	\end{align*}
	We define, for $n \in \N \cup \{\infty\}$
	\begin{align*}
		T_{i, n} &\coloneqq h(V_i)\cap \setc{y \in A}{|\set{ x \in \widetilde{V}_i}{h(x) = y}| = n-1}.	
	\end{align*}
	In words, $T_{i, n}$ contains the values of $A$ that are hit on $V_i$ for the $n$th time. 
	Since $h$ is cpwl, we observe that for all $i = 1, \dots, P$
	\begin{enumerate}
	\item $T_{i, n_1} \cap T_{i, n_2} = \emptyset$ for all $n_1,n_2 \in \N \cup \{\infty\}$, $n_1\neq n_2$,
	\item $ T_{i, \infty} \cup \bigcup_{n=1}^\infty T_{i, n} = h(V_i) \cap A$,
	\item $T_{i, n} = \emptyset$ for all $P < n < \infty$,
	\item $|T_{i, \infty}| = 0$.          
	\end{enumerate}
	
	Note that, since $h$ is affine on $V_i$ it holds that $h' = |h(V_i)| / |V_i|$ on $V_i$.         
	Hence, for $t \leq P$
	\begin{align*}
          \norm[L^1]{h'} &\geq \sum_{i=1}^P |h(V_i)| \geq  \sum_{i=1}^P  |h(V_i) \cap A|\\
                         &%
                           =
                   \ \sum_{i=1}^P \left(\sum_{n=1}^\infty |T_{i, n}|\right) + |T_{i, \infty}|\\
                         &%
                           =
           \sum_{i=1}^P \sum_{n=1}^\infty |T_{i, n}|\\
                         &%
                           \geq
           \sum_{n=1}^t \sum_{i=1}^P |T_{i, n}|,
	\end{align*}
        where the first equality follows by (i), (ii), the second by (iv), and the last inequality by (iii). %
	Note that, by assumption for all $n \leq t$ every $y \in A$ is an element of $T_{i, n}$ or $T_{i, \infty}$ for some $i \leq P$. 
	Therefore, by (iv)
	\[
          \sum_{i=1}^P |T_{i, n}| \geq |A|,     
	\]
	which completes the proof.
\end{proof}

Lemma \ref{lem:probabilityOfHittingALevel} applied to neural networks essentially states that, in a single neuron, if the bias term is chosen uniformly randomly on an interval of length $\delta$, then the probability of generating at least $t$ pieces by composition scales %
reciprocal to $t$.

Next, we will analyze how Lemma \ref{lem:probabilityOfHittingALevel} implies an upper bound on the number of pieces generated in a randomly initialized neural network.
For simplicity, we only consider random biases in the following,
but mention that similar results hold if both the biases and weights are random variables \cite{hanin2019complexity}.

\begin{definition}
Let $L\in \N$, $(d_0, d_1, \dots, d_{L}, 1)\in \N^{L+2}$ %
and $\BW^{(\ell)} \in \R^{d_{\ell+1} \times d_{\ell}}$ for $\ell = 0, \dots, L$.
Furthermore,
let $\delta>0$ and let the bias vectors $\Bb^{(\ell)} \in \R^{d_{\ell+1}}$, for $\ell=0,\dots,L$, be random variables such that each entry of each $\Bb^{(\ell)}$ is independently and uniformly distributed on the interval $[-\delta/2,\delta/2]$.
We call the associated ReLU neural network %
a \textbf{random-bias neural network}. 
\end{definition}

To apply Lemma \ref{lem:probabilityOfHittingALevel} to a single neuron with random biases, we also need some bound on the derivative of the input to the neuron.

\begin{definition}
Let $L\in \N$, $(d_0, d_1, \dots, d_{L}, 1)\in \N^{L+2}$, and
$\BW^{(\ell)} \in \R^{d_{\ell+1} \times d_{\ell}}$ and $\Bb^{(\ell)} \in \R^{d_{\ell+1}}$ for $\ell = 0, \dots, L$. Moreover let $\delta>0$.

For $\ell = 1, \dots, L+1$, $i=1,\dots, d_{\ell}$ %
introduce the functions
\begin{align*}
	\eta_{\ell, i}(\Bx; (\BW^{(j)}, \Bb^{(j)})_{j = 0}^{\ell-1}) = (\BW^{(\ell-1)}\Bx^{(\ell-1)})_i\qquad \text{for } \Bx \in \R^{d_0},
\end{align*}
where $\Bx^{(\ell-1)}$ is as in \eqref{eq:xell}.
We call
\begin{align*}
	\nu\left((\BW^{(\ell)})_{\ell = 1}^L, \delta\right) \coloneqq &\max\Bigg\{ \normc[2]{\eta_{\ell, i}'(\, \cdot \, ; (\BW^{(j)}, \Bb^{(j)})_{j = 0}^{\ell-1})} \Bigg| \\
	 & (\Bb^{(j)})_{j = 0}^L \in \prod_{j = 0}^L [-\delta/2,\delta/2]^{d_{j+1}},\ell = 1, \dots, L, i=1,\dots, d_{\ell}\Bigg\}
\end{align*}
the \textbf{maximal internal derivative} of $\Phi$.
\end{definition}

We can now formulate the main result of this section.

\begin{theorem}\label{thm:numberOfPiecesInPractice}
	Let $L\in \N$ and let $(d_0, d_1, \dots, d_{L}, 1) \in \N^{L+2}$.
Let $\delta\in (0,1]$.
Let $\BW^{(\ell)} \in \R^{d_{\ell+1} \times d_{\ell}}$, for $\ell = 0, \dots, L$, be such that 
	$\nu\left((\BW^{(\ell)})_{\ell = 0}^L, \delta\right) \leq C_\nu$ for a $C_\nu >0$.
	
	For an associated random-bias neural network $\Phi$, we have that for a line segment $\mathfrak{s} \subseteq \R^{d_0}$ of length 1 
	\begin{align}\label{eq:expectedNumOfPieces}
		\mathbb{E}[{\rm Pieces}(\Phi, \mathfrak{s})]\leq 1 + d_1 + \frac{C_\nu}{\delta} (1 + (L-1) \ln(2 \wdth(\Phi))) \sum_{j = 2}^{L} d_j.
	\end{align}
\end{theorem}

\begin{proof}
Let $\BW^{(\ell)} \in \R^{d_{\ell+1} \times d_{\ell}}$ for $\ell = 0, \dots, L$.
Moreover, let $\Bb^{(\ell)} \in [-\delta/2, \delta/2]^{d_{\ell+1}}$  for $\ell = 0, \dots, L$ be uniformly distributed random variables. %
We denote
\begin{align*}
  \theta_\ell
  \colon \mathfrak{s} &\to \R^{d_\ell}\\
	\Bx &\mapsto (\eta_{\ell, i}(\Bx; (\BW^{(j)}, \Bb^{(j)})_{j = 0}^{\ell-1}))_{i=1}^{d_\ell}.
\end{align*}

	Let $\kappa\colon \mathfrak{s} \to [0,1]$ be an isomorphism. 
        Since each coordinate of $\theta_{\ell}$ is cpwl, %
        there are points $\Bx_0, \Bx_1, \ldots, \Bx_{q_\ell} \in  \mathfrak{s}$ with $\kappa(\Bx_j) < \kappa(\Bx_{j+1})$ for $j = 0, \dots, q_{\ell} - 1$, such that $\theta_{\ell}$ is affine (as a function into $\R^{d_\ell}$) on $[\kappa(\Bx_j), \kappa(\Bx_{j+1})]$ for all $j = 0, \dots, q_{\ell}-1$ as well as on $[0, \kappa(\Bx_0)]$ and $[\kappa(\Bx_{q_\ell}),1]$.
	
	We will now inductively find an upper bound on the $q_\ell$. 
	
	Let $\ell = 2$, then 
	\[
		\theta_2(\Bx) = \BW^{(1)} \sigma_{\rm ReLU}( \BW^{(0)} \Bx +  \Bb^{(0)}).
	\] 
	Since $\BW^{(1)} \cdot + b^{(1)}$ is an affine function, it follows that $\theta_2$ can only be non-affine in points where $\sigma_{\rm ReLU}( \BW^{(0)} \cdot +  \Bb^{(0)})$ is not affine. 
	Therefore, $\theta_2$ is only non-affine if one coordinate of $\BW^{(0)} \cdot +  \Bb^{(0)}$ intersects $0$ nontrivially.
This can happen at most $d_1$ times. 
	We conclude that we can choose $q_2 = d_1$. 
	
	Next, let us find an upper bound on $q_{\ell+1}$ from $q_\ell$.
Note that
	\[
		\theta_{\ell+1}(\Bx) =\BW^{(\ell)} \sigma_{\rm ReLU}( \theta_{\ell}(\Bx) + \Bb^{(\ell-1)}).
	\]
	Now $\theta_{\ell+1}$ is affine in every point $\Bx\in \mathfrak{s}$ where $\theta_{\ell}$ is affine and $(\theta_{\ell}(\Bx) + \Bb^{(\ell-1)})_i\neq 0$ for all coordinates $i = 1, \dots, d_\ell$.
As a result, we have that we can choose $q_{\ell+1}$ such that
	\[
		q_{\ell+1} \leq q_{\ell} + \big|\set{\Bx \in \mathfrak{s}}{(\theta_{\ell}(\Bx) + \Bb^{(\ell-1)})_i = 0 \text{ for at least one } i =1, \dots, d_{\ell}}\big|.
	\]
	Therefore, for $\ell \geq 2$
	\begin{align*}
		q_{\ell+1} &\leq d_{1} + \sum_{j = 3}^{\ell} \big|\set{\Bx \in \mathfrak{s}}{(\theta_{j}(\Bx) + \Bb^{(j)})_i = 0 \text{ for at least one } i =1, \dots, d_{j}}\big|\\
		&\leq d_1 + \sum_{j = 2}^{\ell} \sum_{i=1}^{d_j} \big|\set{\Bx \in \mathfrak{s}}{\eta_{j, i}(\Bx) = - \Bb^{(j)}_i }\big|.
	\end{align*}
	By Theorem \ref{thm:NumberOfPiecesTelgarskyStyle}, we have that 
	\[
          {\rm Pieces}\left(\eta_{\ell, i}( \, \cdot \, ; (\BW^{(j)}, \Bb^{(j)})_{j = 0}^{\ell-1}), \mathfrak{s}\right)
          \leq (2 \wdth(\Phi))^{\ell-1}.
	\]
	We define for $k\in \N \cup \{\infty\}$
	\begin{align*}
		p_{k,\ell, i} \coloneqq \mathbb{P} \left[\big|\set{\Bx \in \mathfrak{s}}{\eta_{\ell, i}(\Bx) = - \Bb^{(\ell)}_i }\big| \geq k\right]
	\end{align*}
	Then by Lemma \ref{lem:probabilityOfHittingALevel}
	\begin{align*}
		p_{k,\ell, i}  \leq \frac{C_\nu}{\delta k}
	\end{align*}
	and for $k >  (2 \wdth(\Phi))^{\ell-1}$
	\begin{align*}
		p_{k,\ell, i} = 0.
	\end{align*}
        
        It holds
	\begin{align*}
		&\mathbb{E}\left[\sum_{j = 2}^{L} \sum_{i=1}^{d_j} \Big|\setc{\Bx \in \mathfrak{s}}{\eta_{j, i}(\Bx) = - \Bb^{(j)}_i }\Big|\right] \\
		\leq & \sum_{j = 2}^{L}  \sum_{i=1}^{d_j} \sum_{k =1}^{\infty} k \cdot \mathbb{P}\left[ \Big|\setc{\Bx \in \mathfrak{s}}{\eta_{j, i}(\Bx) = - \Bb^{(j)}_i}\Big| = k\right]\\
		\leq & \sum_{j = 2}^{L}  \sum_{i=1}^{d_j} \sum_{k =1}^{\infty} k \cdot (p_{k,j, i} - p_{k+1,j, i}).
\end{align*}
The inner sum can be bounded by
	\begin{align*}
		\sum_{k = 1}^\infty k \cdot (p_{k,j, i} - p_{k+1,j, i}) &= \sum_{k = 1}^\infty k \cdot  p_{k,j, i} - \sum_{k = 1}^\infty k \cdot  p_{k+1,j, i}\\
		&= \sum_{k = 1}^\infty k \cdot  p_{k,j, i} - \sum_{k = 2}^\infty (k-1) \cdot  p_{k,j, i}\\
		&=  p_{1,j, i} + \sum_{k = 2}^\infty p_{k,j, i}\\
		&= \sum_{k = 1}^\infty p_{k,j, i}\\
		&\leq C_\nu \delta^{-1} \sum_{k =1}^{(2 \wdth(\Phi))^{L-1}} \frac{1}{k}\\
		&\leq C_\nu \delta^{-1} \left(1+ \int_{1}^{(2 \wdth(\Phi))^{L-1}} \frac{1}{x} \dd x\right)\\
		&\leq  C_\nu \delta^{-1} (1 + (L-1) \ln((2 \wdth(\Phi)))).
	\end{align*}
	We conclude that, in expectation, we can bound $q_{L+1}$ by 
	\begin{align*}
		d_1 + C_\nu \delta^{-1} (1 + (L-1) \ln(2 \wdth(\Phi))) \sum_{j = 2}^{L} d_j. 
	\end{align*}
	Finally, since $\theta_{L} = \Phi_{L+1}|_{\mathfrak{s}}$, it follows that 
	\begin{align*}
		{\rm Pieces}(\Phi, \mathfrak{s}) \leq q_{L + 1} + 1
	\end{align*}
	which yields the result. 
\end{proof}

\begin{remark}
We make the following observations about Theorem \ref{thm:numberOfPiecesInPractice}:
\begin{itemize}
\item \textit{Non-exponential dependence on depth:} If we consider \eqref{eq:expectedNumOfPieces}, we see that the number of pieces scales in expectation essentially like
  $O(LN)$,
  where $N$ is the total number of neurons of the architecture.
This shows that in expectation, the number of pieces is linear in the number of layers, as opposed to the exponential upper bound of Theorem \ref{thm:NumberOfPiecesTelgarskyStyle}.
	\item \textit{Maximal internal derivative:} Theorem \ref{thm:numberOfPiecesInPractice} requires the weights to be chosen such that the maximal internal derivative is bounded by a certain number.
However, if they are randomly initialized %
in such a way that with %
high probability the maximal internal derivative is bounded by a small number, then similar results can be shown.
In practice, weights in the $\ell$th layer are often initialized according to a centered normal distribution with standard deviation $\sqrt{2/d_{\ell}}$, \cite{he2015delving}. 
Due to the anti-proportionality of the variance to the width of the layers it is achieved that the internal derivatives remain bounded with high probability, independent of the width of the neural networks.
This explains the observation from Figure \ref{fig:NumPiecesDeepVsShallow}.
\end{itemize}
\end{remark}	

\section*{Bibliography and further reading}
Establishing bounds on the number of linear regions of a ReLU
  network has been a popular tool to investigate the complexity of
  ReLU neural networks, see
  \cite{montafur14linearregions,pmlr-v70-raghu17a,arora2018understanding,serra2018bounding,hanin2019complexity}.
  The bound presented in Section \ref{sec:UpperBoundOnPieces}, is
  based on \cite{telgarsky_depth}. %
  For the construction of the sawtooth function in Section
  \ref{sec:tightness}, %
  we follow the arguments in
  \cite{telgarsky_depth,telgarsky_depth2}. %
  Together with the lower bound on the number of required linear
  regions given in \cite{frenzen2010number}, this analysis shows how
  depth can be a limiting factor in terms of achievable convergence
  rates, as stated in Theorem \ref{thm:FrenzenApplied}.
Finally, the analysis of the number of pieces deep neural networks
attained with random initialization (Section
\ref{sec:NumOfPieceInPractice}) is based on \cite{hanin2019complexity}
and \cite{karner2022limitations}.

\newpage
\section*{Exercises}
\begin{exercise}\label{ex:peps}
Let $-\infty<a<b<\infty$ and let $f\in C^3([a,b])\backslash\P_1$.
Denote by $p(\eps)\in\N$ the minimal number of intervals partitioning
$[a,b]$, such that a (not necessarily continuous) piecewise linear
function on $p(\eps)$ intervals can approximate $f$ on $[a,b]$
uniformly up to error $\eps>0$.
In this exercise, we wish to show
\begin{align}\label{eq:toshowpeps}
	\liminf_{\eps\searrow 0} p(\eps)\sqrt{\eps} >0.
\end{align}
Therefore, we can find a constant $C>0$ such that $\eps\ge C p(\eps)^{-2}$
for all $\eps>0$.
This shows a variant of
Theorem \ref{thm:bestApproximationBypiecewiseaffineThings}.
Proceed as follows to prove \eqref{eq:toshowpeps}:

\begin{enumerate}
\item Fix $\eps>0$ and let
$a=x_0<x_1\dots<x_{p(\eps)}=b$ be a partitioning into
$p(\eps)$ pieces.
For $i=0,\dots,p(\eps)-1$ and $x\in [x_i,x_{i+1}]$
let
\begin{align*}
	e_i(x)\dfn f(x) - \left(f(x_i)+\frac{f(x_{i+1})-f(x_i)}{x_{i+1}-x_i}(x-x_i)\right).
\end{align*}
Show that $|e_i(x)|\le 2\eps$ for all $x\in [x_i,x_{i+1}]$.

\item With $h_i\dfn x_{i+1}-x_i$ and $m_i\dfn (x_i+x_{i+1})/{2}$ show
that
\begin{align*}
	\max_{x\in [x_i,x_{i+1}]}|e_i(x)| = \frac{h_i^2}{8} |f''(m_i)|
	+O(h_i^3).
\end{align*}

\item Assuming that $c\dfn \inf_{x\in [a,b]}|f''(x)|>0$ show that
\begin{align*}
	\liminf_{\eps\searrow 0}p(\eps)\sqrt{\eps}
	\ge \frac{1}{4}\int_a^b \sqrt{|f''(x)|}\dd x.
\end{align*}

\item Conclude that \eqref{eq:toshowpeps} holds for
general non-linear $f\in C^3([a,b])$.
\end{enumerate}
\end{exercise}

\begin{exercise}
Show that, for $L = 1$, Theorem \ref{thm:NumberOfPiecesTelgarskyStyle} holds for piecewise smooth functions, when replacing the number of affine pieces by the number of smooth pieces.  These are defined by replacing ``affine'' by ``smooth'' (meaning $C^\infty$)  in Definition \ref{def:NumOfPieces}.
\end{exercise}

\begin{exercise}
  Show that, for $L > 1$, Theorem \ref{thm:NumberOfPiecesTelgarskyStyle} does \emph{not} hold for piecewise smooth functions, when replacing the number of affine pieces by the number of smooth pieces.
\end{exercise}

\begin{exercise}
For $p \in \N$, $p > 2$ and $n\in \N$, construct a function $h^{(p)}_n$ similar to $h_n$ of \eqref{lemma:hn}, such that $h^{(p)}_n \in \CN_1^1(\sigma_{\rm ReLU};n,p)$ %
and such that $h^{(p)}_n$ has $p^n$ pieces and size $O(p^2 n)$.
\end{exercise}

%% file: DeepReLUNNs.tex
\chapter{Deep ReLU neural networks}\label{chap:DReLUNN}
In the previous chapter, we observed that many layers are a necessary prerequisite for ReLU neural networks to approximate smooth functions with high rates.
We now analyze which depth is sufficient to achieve good approximation rates for smooth functions.

To approximate smooth functions efficiently, one of the main tools in Chapter \ref{chap:Splines} was to rebuild polynomial-based functions, such as higher-order B-splines.
For smooth activation functions, we were able to reproduce polynomials by using the nonlinearity of the activation functions.
This argument certainly cannot be repeated for the \emph{piecewise linear} ReLU.
On the other hand, up until now, we have seen that deep ReLU neural networks are extremely efficient at producing the strongly oscillating sawtooth functions discussed in Lemma \ref{lemma:hn}.  
The main observation %
in this chapter is that the %
sawtooth functions %
are intimately linked to the %
squaring function, which %
again leads %
to %
polynomials. %
This observation was first made by Dmitry Yarotsky \cite{yarotsky} in 2016, and the present chapter is primarily based on this paper.

In Sections \ref{sec:squareFunction} and \ref{sec:multiplication}, we give Yarotsky's
approximation of the squaring and multiplication functions.
As a direct consequence, we show in Section \ref{sec:depthseparation} that deep ReLU neural networks can be \emph{significantly} more efficient than shallow ones in approximating analytic functions.

Using these %
tools, we %
conclude in Section \ref{sec:CksFunctions} that deep ReLU neural networks can efficiently approximate $k$-times continuously differentiable functions with H\"older continuous derivatives.

\section{The square function}\label{sec:squareFunction}
We start with the approximation of the map $x\mapsto x^2$. The construction,
  first given in \cite{yarotsky}, is based on the sawtooth functions $h_n$ defined in \eqref{eq:hn} and originally introduced in \cite{telgarsky_depth}, see Figure \ref{fig:hn}. The proof idea is visualized in Figure \ref{fig:sn}.

\begin{figure}
  \begin{center}
    \input{plots/square.tex}
\end{center}
\caption{Construction of $s_n$ in Proposition \ref{prop:sn}.}
\label{fig:sn}
\end{figure}

\begin{proposition}\label{prop:sn}
Let $n\in\N$.
Then
\begin{align*}
	  s_n(x)\dfn x-\sum_{j=1}^n \frac{h_j(x)}{2^{2j}}
\end{align*}
is a piecewise linear function on $[0,1]$ with break points
$x_{n,j}=j2^{-n}$, $j=0,\dots,2^n$.
Moreover,
$s_n(x_{n,k})=x_{n,k}^2$ for all $k=0,\dots,2^n$, i.e.\ $s_n$ is
the piecewise linear interpolant of $x^2$ on $[0,1]$.
\end{proposition}

\begin{proof}
The statement holds for $n=1$.
We proceed by induction.
Assume the statement holds for $s_n$ and let
$k\in\{0,\dots,2^{n+1}\}$.
By Lemma~\ref{lemma:hn},
$h_{n+1}(x_{n+1,k})=0$ whenever $k$ is even.
Hence for even
$k\in\{0,\dots,2^{n+1}\}$
\begin{align*}
	  s_{n+1}(x_{n+1,k}) &= x_{n+1,k}-\sum_{j=1}^{n+1} \frac{h_j(x_{n+1,k})}{2^{2j}}\\
	  & =s_n(x_{n+1,k})-\frac{h_{n+1}(x_{n+1,k})}{2^{2(n+1)}}
	  =s_n(x_{n+1,k})=x_{n+1,k}^2,
\end{align*}
where we used the induction assumption $s_n(x_{n+1,k})=x_{n+1,k}^2$
for $x_{n+1,k}=k2^{-(n+1)}=\frac{k}{2}2^{-n}=x_{n,k/2}$.

Now let $k\in\{1,\dots,2^{n+1}-1\}$ be odd.
Then by
Lemma~\ref{lemma:hn}, $h_{n+1}(x_{n+1,k})=1$.
Moreover, since $s_n$ is
linear on $[x_{n,(k-1)/{2}},x_{n,(k+1)/{2}}]=[x_{n+1,k-1},x_{n+1,k+1}]$
and $x_{n+1,k}$ is the midpoint of this interval,
\begin{align*}
	  s_{n+1}(x_{n+1,k}) &= s_n(x_{n+1,k}) - \frac{h_{n+1}(x_{n+1,k})}{2^{2(n+1)}}\nonumber\\
	                     &= \frac{1}{2}(x_{n+1,k-1}^2+x_{n+1,k+1}^2)-\frac{1}{2^{2(n+1)}}\nonumber\\
	                     &=\frac{(k-1)^2}{2^{2(n+1)+1}}+\frac{(k+1)^2}{2^{2(n+1)+1}}
	                       -\frac{2}{2^{2(n+1)+1}}\nonumber\\
	                     &= \frac{1}{2}\frac{2k^2}{2^{2(n+1)}} = \frac{k^2}{2^{2(n+1)}}=x_{n+1,k}^2.
\end{align*}
This completes the proof.
\end{proof}

As a consequence there holds the following, \cite[Proposition 2]{yarotsky}.

\begin{lemma}\label{lemma:sn}
For $n \in \N$, it holds
\begin{align*}
	  \sup_{x\in[0,1]}|x^2-s_n(x)|\le 2^{-2n-1}.
\end{align*}
Moreover $s_n\in\CN_{1}^1(\sigma_{\rm ReLU};n,3)$, and
$\size(s_n)\le 7n$ and $\depth(s_n)=n$.
\end{lemma}

\begin{proof}
Set $e_n(x)\dfn x^2-s_n(x)$.
Let
$x$ be in the interval
$[x_{n,k},x_{n,k+1}]=[k2^{-n},(k+1)2^{-n}]$
of length $2^{-n}$.
Since $s_n$ is the linear interpolant of $x^2$ on this interval,
we have
\begin{align*}
	  |e_n'(x)| = \left|2x-\frac{x_{n,k+1}^2-x_{n,k}^2}{2^{-n}}\right|
	  =\left|2x - \frac{2k+1}{2^n}\right|\le \frac{1}{2^n}.
\end{align*}
Thus $e_n:[0,1]\to\R$ has Lipschitz constant $2^{-n}$.
Since $e_n(x_{n,k})=0$ for all $k=0,\dots,2^n$,
and the length of the interval $[x_{n,k},x_{n,k+1}]$
equals $2^{-n}$ we get
\begin{align*}
	  \sup_{x\in [0,1]}|e_n(x)|\le \frac{1}{2} 2^{-n} 2^{-n}=2^{-2n-1}.
\end{align*}

Finally, to see that $s_n$ can be represented by a neural network
of the claimed architecture, note that for $n\ge 2$
\begin{align*}
	  s_n(x) = x-\sum_{j=1}^n\frac{h_j(x)}{2^{2j}}
	  = s_{n-1}(x)-\frac{h_n(x)}{2^{2n}}
	  = \sigma_{\rm ReLU}\circ s_{n-1}(x)-\frac{h_1\circ h_{n-1}(x)}{2^{2n}}.
\end{align*}
Here we used that $s_{n-1}$ is the piecewise linear interpolant of
$x^2$, so that $s_{n-1}(x)\ge 0$ and thus
$s_{n-1}(x)=\sigma_{\rm ReLU}(s_{n-1}(x))$ for all $x\in [0,1]$.
Hence $s_n$ is of depth $n$ and width $3$, see Figure \ref{fig:h1sn}.
\end{proof}

\begin{figure}
\begin{center}
\input{./plots/h1sn.tex}
\end{center}
\caption{The neural networks $h_1(x)=\sigma_{\rm ReLU}(2x)-\sigma_{\rm ReLU}(4x-2)$
and $s_n(x) = \sigma_{\rm ReLU}(s_{n-1}(x))-{h_n(x)}/{2^{2n}}$
where $h_n = h_1\circ h_{n-1}$. Figure based on \cite[Fig.~2c]{yarotsky} and \cite[Fig.~1a]{Schwab2019Deep}.}
\label{fig:h1sn}
\end{figure}

In conclusion, we have shown that $s_n:[0,1]\to [0,1]$ approximates the
square function uniformly on $[0,1]$ with exponentially decreasing
error in the neural network size.
Note that due to Theorem \ref{thm:FrenzenApplied}, this would not be possible with
a shallow neural network, which can at best interpolate $x^2$ on a
partition of $[0,1]$ with polynomially many (w.r.t.\ the neural network
size) pieces.

\section{Multiplication}\label{sec:multiplication}
According to Lemma \ref{lemma:sn}, depth can help in the approximation
of $x\mapsto x^2$, which, on first sight, seems like a rather
specific example.
However, as we shall discuss in the following,
this opens up a path towards fast approximation of functions with
high regularity, e.g., $C^k([0,1]^d)$ for some $k>1$.
The crucial
observation is that, via the polarization identity we can write the
product of two numbers as a sum of squares
\begin{align}\label{eq:parallelogram}
	x\cdot y = \frac{(x+y)^2-(x-y)^2}{4}
\end{align}
for all $x$, $y\in\R$.
Efficient approximation of the operation of
multiplication allows efficient approximation of polynomials.
Those in turn are well-known to be good approximators for functions
exhibiting $k\in\N$ derivatives.
Before exploring this idea further
in the next section, we first make precise the observation that neural networks can efficiently approximate the multiplication
of real numbers.

We start with the multiplication of two numbers, in which case neural networks of
logarithmic size in the desired accuracy are sufficient,
\cite[Proposition 3]{yarotsky}.

\begin{lemma}\label{lemma:mult}
For every $\eps>0$ there exists a ReLU neural network
$\ntim{\eps}:[-1,1]^2\to [-1,1]$ such that
\begin{align*}
	  \sup_{x,y\in [-1,1]}|x\cdot y-\ntim{\eps}(x,y)|\le \eps,
\end{align*}
and it holds $\size(\ntim{\eps})\le C \cdot (1+|\log(\eps)|)$ and
$\depth(\ntim{\eps})\le C\cdot(1+|\log(\eps)|)$ for a constant $C>0$
independent of $\eps$.
Moreover, $\ntim{\eps}(x,y)=0$ if $x=0$
or $y=0$.
\end{lemma}

\begin{proof}
With $n=\lceil|\log_4(\eps)|\rceil$, define the neural network
\begin{align}
	  \ntim{\eps}(x,y)\dfn &  s_n\left(\frac{\sigma_{\rm ReLU}(x+y)+\sigma_{\rm ReLU}(-x-y)}{2}\right) \nonumber\\
	& - s_n \left(\frac{\sigma_{\rm ReLU}(x-y)+\sigma_{\rm ReLU}(y-x)}{2}\right).\label{eq:deftimesepsn}
\end{align}
Since $|a|=\sigma_{\rm ReLU}(a)+\sigma_{\rm ReLU}(-a)$, by \eqref{eq:parallelogram} we
have for all $x$, $y\in [-1,1]$
\begin{align*}
	  \left| x\cdot y-\ntim{\eps}(x,y)\right|
	  &=
	    \left|\frac{(x+y)^2-(x-y)^2}{4}-\left(s_n\left(\frac{|x+y|}{2}\right)-s_n\left(\frac{|x-y|}{2}\right)\right)\right|\nonumber\\
	  &=
	    \left|\frac{4(\frac{x+y}{2})^2-4(\frac{x-y}{2})^2}{4}-\frac{4s_n(\frac{|x+y|}{2})-4s_n(\frac{|x-y|}{2})}{4}\right|\nonumber\\
	  &\le \frac{4(2^{-2n-1}+2^{-2n-1})}{4}=4^{-n}\le \eps,
\end{align*}
where we used $|x+y|/2$, $|x-y|/2\in [0,1]$.
We have
$\depth(\ntim{\eps})=1+\depth(s_n)=1+n\le
1+\lceil|\log_4(\eps)|\rceil$ and
$\size(\ntim{\eps}) \le C+2\size(s_n)\le C n\le C\cdot
(1-\log(\eps))$ for some constant $C>0$.

The fact that $\ntim{\eps}$ maps from $[-1,1]^2\to [-1,1]$
follows by \eqref{eq:deftimesepsn} and because
$s_n:[0,1]\to [0,1]$.
Finally, if $x=0$, then
$\ntim{\eps}(x,y)=s_n(|x+y|)-s_n(|x-y|)=s_n(|y|)-s_n(|y|)=0$. 
If $y=0$ the same argument can be made.
\end{proof}

In a similar way as in Proposition \ref{prop:reapproxBsplinehighdim} and Lemma \ref{lemma:minmaxn}, we can apply operations with two inputs in the form of a binary tree to extend them to an operation on arbitrary many inputs; see again \cite{yarotsky}, and \cite[Proposition 3.3]{Schwab2019Deep} for the specific argument considered here.

\begin{proposition}\label{prop:multn}
For every $n\ge 2$ and $\eps>0$ there exists a ReLU neural network
$\ntim{n,\eps}:[-1,1]^n\to [-1,1]$ such that
\begin{align*}
	  \sup_{x_j\in [-1,1]}\left|\prod_{j=1}^n x_j-\ntim{n,\eps}(x_1,\dots,x_n)\right|\le \eps,
\end{align*}
and it holds $\size(\ntim{n,\eps})\le Cn\cdot(1+|\log(\eps/n)|)$
and $\depth(\ntim{n,\eps})\le C\log(n)(1+|\log(\eps/n)|)$ for
a constant $C>0$ independent of $\eps$ and $n$.
\end{proposition}

\begin{proof}
We begin with the case
$n=2^k$.
For $k=1$ let
$\tntim{2,\delta}\dfn \ntim{\delta}$.
If $k\ge 2$ let
\begin{align*}
	  \tntim{2^k,\delta}\dfn \ntim{\delta}\circ
	  \left(\tntim{2^{k-1},\delta},\tntim{2^{k-1},\delta}\right).
\end{align*}
Using Lemma \ref{lemma:mult}, we find that this neural network has depth
bounded by
\begin{align*}
	  \depth\left(\tntim{2^k,\delta}\right)
	  \le
	  k\depth(\ntim{\delta})\le Ck \cdot (1+|\log(\delta)|)
	  \le C\log(n)(1+|\log(\delta)|).
\end{align*}
Observing that the number of occurrences of $\ntim{\delta}$
equals $\sum_{j=0}^{k-1}2^j\le n$, the size of
$\tntim{2^k,\delta}$ can bounded by
$Cn \size(\ntim{\delta})\le Cn \cdot (1+|\log(\delta)|)$.

To estimate the approximation error, denote with $\Bx=(x_j)_{j=1}^{2^k}$
\begin{align*}
	  e_k\dfn \sup_{x_j\in [-1,1]} \left|\prod_{j\le 2^{k}}x_j-\tntim{2^{k},\delta}(\Bx)\right|.
\end{align*}
Then, using short notation of the
type $\Bx_{\le 2^{k-1}}\dfn (x_1,\dots,x_{2^{k-1}})$,
\begin{align*}
	      e_k
	  &=\sup_{x_j\in [-1,1]}\left|\prod_{j=1}^{2^k}x_j-\ntim{\delta}\left(\tntim{2^{k-1},\delta}(\Bx_{\le 2^{k-1}}),\tntim{2^{k-1},\delta}(\Bx_{>2^{k-1}})\right) \right|\nonumber\\
	    &\le \delta +
	      \sup_{x_j\in [-1,1]}\left(
	      \left|\prod_{j\le 2^{k-1}}x_j\right|
	      e_{k-1}
	      +\left|\tntim{2^{k-1},\delta}(\Bx_{>2^{k-1}})\right|
	      e_{k-1}
	      \right)\nonumber\\
	    &\le \delta +2
	      e_{k-1}\le \delta+2(\delta+2e_{k-2})\le\dots\le \delta\sum_{j=0}^{k-2}2^j+2^{k-1}e_1\nonumber\\
          &\le 2^k \delta =n\delta.
\end{align*}
Here we used $e_1\le\delta$, and that $\tntim{2^{k-1},\delta}$
maps $[-1,1]^{2^{k-1}}$ to $[-1,1]$, which is a consequence of
Lemma \ref{lemma:mult}.

The case for general $n\ge 2$ (not necessarily $n=2^k$) is
treated similar as in Lemma \ref{lemma:minmaxn}, by replacing some
$\ntim{\delta}$ neural networks with identity neural networks.

Finally, setting $\delta\dfn {\eps}/{n}$ and
$\ntim{n,\eps}\dfn \tntim{n,\delta}$ concludes the
proof.
\end{proof}

\section{Polynomials, analytic functions and depth separation}\label{sec:depthseparation}
We now discuss a few first consequences of the above observations,
and begin with the approximation of the univariate polynomial
\begin{equation}\label{eq:polynomial}
  p(x)=\sum_{j=0}^n c_jx^j.
\end{equation}
One possibility to approximate $p$ is via the Horner scheme and
the approximate multiplication $\Phi^{\times}_\eps$ from Lemma \ref{lemma:mult},
yielding
\begin{align*}
  p(x) &= c_0+x\cdot(c_1+x\cdot(\dots +x\cdot c_n)\dots)\\
  &\simeq c_0+\Phi_\eps^\times(x,c_1+\Phi_\eps^\times(x,c_2\dots +\Phi_\eps^\times(x,c_n))\dots).
\end{align*}
This scheme requires depth $O(n)$ due to the nested multiplications. An
alternative is to approximate all monomials $1,x,\dots,x^n$ with a
binary tree using approximate multiplications $\Phi_\eps^\times$, and
combing them in the output layer, see Figure \ref{fig:monomials}. This
idea leads to a network of size $O(n\log(n))$ and depth $O(\log(n))$. The
following lemma formalizes this, see \cite[Lemma A.5]{petersen2018optimal},
\cite[Proposition III.5]{9363169}, and in particular \cite[Lemma 4.3]{opschoor2020fem}.
The proof is left as Exercise \ref{ex:polynomial}.

\begin{lemma}\label{lemma:polynomial}
  There exists a constant $C>0$, such that for
  any $\eps\in (0,1)$ and any
  polynomial $p$ of degree $n\ge 2$ as in \eqref{eq:polynomial},
  there exists a neural network $\Phi_\eps^p$ such that
  \begin{equation*}
    \sup_{x\in [-1,1]}|p(x)-\Phi_\eps^p(x)|\le C \eps \sum_{j=0}^n|c_j|
  \end{equation*}
  and $\size(\Phi_\eps^p)\le C n\log(n/\eps)$ and $\depth(\Phi_\eps^p)\le C\log(n/\eps)$.
\end{lemma}

\begin{figure}
\begin{center}
\begin{tikzpicture}
  \node at (-0.4,0) {$1^{\phantom{0}}$};
  \node at (0.4,0) {$x^{\phantom{0}}$};
  \draw [thick,->] (-0.5,-0.3) -- (-0.8,-1);
  \draw [thick,->] (-0.4,-0.3) -- (-0.125,-1);
  
  \draw [thick,->] (0.3,-0.3) -- (0.025,-1);
  \draw [thick,->] (0.4,-0.3) -- (0.7,-1);    

  \node at (-0.8,-1.2) {$1^{\phantom{0}}$};
  \node at (0,-1.2) {$x^{\phantom{0}}$};
  \node at (0.8,-1.2) {$x^2$};

  \node at (-1.6,-2.4) {$1^{\phantom{0}}$};
  \node at (-0.8,-2.4) {$x^{\phantom{0}}$};
  \node at (0.0,-2.4) {$x^2$};
  \node at (0.8,-2.4) {$x^3$};
  \node at (1.6,-2.4) {$x^4$};

  \draw [thick,->] (-1,-1.5) -- (-1.6,-2.2);
  \draw [thick,->] (-0.85,-1.5) -- (-0.85,-2.2);

  \draw [thick,->] (-0.15,-1.5) -- (-0.7,-2.2);  
  \draw [thick,->] (-0.05,-1.5) -- (-0.05,-2.2);
  \draw [thick,->] (0.05,-1.5) -- (0.6,-2.2);
  
  \draw [thick,->] (0.75,-1.5) -- (0.75,-2.2);
  \draw [thick,->] (0.85,-1.5) -- (1.5,-2.2);  
  
\end{tikzpicture}
\end{center}
\caption{Monomials $1,\dots,x^n$ with $n=2^{k}$ can be generated in
  a binary tree of depth $k$. Each node represents the product of
  its inputs, with single-input nodes interpreted as
  squares.}\label{fig:monomials}
\end{figure}

Lemma \ref{lemma:polynomial} shows that deep ReLU networks can
approximate polynomials efficiently. This leads to an interesting
implication regarding %
{\bf analytic} functions: we say that $f:[-1,1]\to\R$ is
analytic if its Taylor series around any
point $x\in [-1,1]$ converges to $f$ in a neighborhood of $x$. For
instance all polynomials, $\sin$, $\cos$, $\exp$ etc.\ are
analytic. There holds the following result \cite{MR3856963,OSZ21}.

\begin{proposition}\label{prop:analytic}
  Let $f:[-1,1]\to\R$ be analytic but not linear. Then there exist constants
  $C$, $\beta>0$ such that for every $N\in\N$, there exists a ReLU neural network $\Phi_{N}$ such that
  \begin{equation*}
    {\rm size}(\Phi_N)\le N\qquad\text{and}\qquad
    {\rm depth}(\Phi_N)\le C\sqrt{N}
  \end{equation*}
  and
  \begin{equation*}
    \sup_{x\in [-1,1]}|f(x)-\Phi_{N}(x)|\le C\exp\Big(-\beta \sqrt{N}\Big).
  \end{equation*}
\end{proposition}

\begin{proof}
  Let us show the upper bound on the deep neural network. Assume first that the convergence radius of the Taylor series of $f$ around $0$ is $r>1$. Then for all $x\in [-1,1]$
  \begin{equation*}
    f(x)=\sum_{j\in\N_0}c_j x^j\qquad\text{where}\qquad
    c_j= \frac{f^{(j)}(0)}{j!}\qquad\text{and}\qquad
    |c_j|\le C_r r^{-j},
  \end{equation*}
  for all $j\in\N_0$ and some $C_r>0$. Hence $p_n(x)\dfn \sum_{j=0}^n c_j x^j$ satisfies
  \begin{equation*}
    \sup_{x\in [-1,1]}|f(x)-p_n(x)|\le \sum_{j>n}|c_j|
    \le C_r \sum_{j>n} r^{-j}\le \frac{C_r r^{-n}}{1-r^{-1}}.
  \end{equation*}
  
  Fix $\eps\dfn r^{-n}$ and let $\Phi^{p_n}_\eps$ be the network in Lemma \ref{lemma:polynomial}. Then
  \begin{align*}
    \sup_{x\in [-1,1]}|f(x)-\Phi^{p_n}_\eps(x)|&\le \sup_{x\in [-1,1]}\big(|f(x)-p_n(x)| + |p_n(x)-\Phi^{p_n}_\eps(x)|\big)\\
                                                 &\le \Big(\frac{C_r}{1-r^{-1}}+C\sum_{j\in\N_0}|c_j| \Big)r^{-n}
    =\tilde C r^{-n}
  \end{align*}
  for some $\tilde C$ depending on $r$, $C_r$ and the constant in Lemma \ref{lemma:polynomial}, but independent of $n$.
  By Lemma \ref{lemma:polynomial}
  \begin{align*}
    \size(\Phi_\eps^{p_n})&\le \hat C n (\log(n)+n\log(r))=O(n^2)\\
    \depth(\Phi_\eps^{p_n})&\le \hat C\cdot (\log(n)+n\log(r))=O(n).
  \end{align*}
  With $N=\lceil Cn (\log(n)+n\log(r))\rceil$ we get
  $\size(\Phi_\eps^{p_n})\le N$, $\depth(\Phi_\eps^{p_n})=O(\sqrt{N})$
  and
  \begin{equation*}
    \sup_{x\in [-1,1]}|f(x)-\Phi^{p_n}_\eps(x)|\le \tilde C \exp(-\log(r) n)
    \le\tilde C \exp(-\beta n)
  \end{equation*}
  for some $\beta$ depending on $\log(r)$ and the above constants.
  Since $n\in\N$ was arbitrary, the statement follows.

  The general case, where the Taylor expansions of $f$ converges
  only locally is left as Exercise \ref{ex:analytic}.
\end{proof}

\begin{remark}
  The above discussion on the approximation of polynomials and analytic functions can be extended to the multivariate case. For polynomials we will discuss this in the proofs of the next subsection; for a more explicit statement of the approximation of multivariate polynomials with ReLU networks see \cite[Section 2.3]{OSZ21}.
  For the
  approximation of analytic functions $f:[-1,1]^d\to\R$,
  this then
  leads to an error bound of type
  $\exp(-\beta N^{1/(1+d)})$, see \cite{MR3856963,OSZ21}.
\end{remark}

The above proposition shows a type of exponential convergence when approximating analytic functions. On the other hand, we have already seen in Theorem \ref{thm:FrenzenApplied} that fixed-depth networks can in general only achieve algebraic convergence rates. This leads to a remarkable statement about the superiority of deep ReLU architectures when approximating nonlinear analytic functions: for fixed-depth networks, the number of parameters must grow faster than any polynomial compared to the required size of deep architectures. We formalize this observation in the following corollary.

\begin{corollary}\label{cor:analytic}
  Let $f:[-1,1]\to\R$ be analytic but not linear. Then there exist constants
  $C$, $\beta>0$ such that for every $\eps>0$, there exists a ReLU neural network
  $\Phi_{\rm deep}$ satisfying
  \begin{equation}\label{eq:Phideep}
    \sup_{x\in [-1,1]}|f(x)-\Phi_{\rm deep}(x)|\le C\exp\Big(-\beta \sqrt{\size(\Phi_{\rm deep})}\Big)\le \eps,
  \end{equation}
  but for any ReLU neural network $\Phi_{\rm shallow}$ of depth at most $L$ holds
  \begin{equation}\label{eq:Philower}
    \sup_{x\in [-1,1]}|f(x)-\Phi_{\rm shallow}(x)|\ge C^{-1} {\rm size}(\Phi_{\rm shallow})^{-2L}.
  \end{equation}  
\end{corollary}

\begin{proof}
  The upper bound in \eqref{eq:Phideep} is a direct consequence of Proposition
  \ref{prop:analytic}. The lower bound on \eqref{eq:Philower}
  holds by Theorem \ref{thm:FrenzenApplied}.
\end{proof}

The proposition shows that the approximation of
certain (highly relevant) functions requires
significantly more parameters when using shallow instead of deep
architectures. Such statements are known as \emph{depth separation}
results. We refer for instance to
\cite{telgarsky_depth,telgarsky_depth2,telgarskynotes}, where such a
result was shown by Telgarsky based on the sawtooth function
constructed in Section \ref{sec:tightness}. %
Lower bounds on the approximation in the spirit of
Corollary \ref{cor:analytic}
were also given in \cite{WhyDeep2017} and \cite{yarotsky}.

\section{$C^{k,s}$ functions}\label{sec:CksFunctions}
We will now discuss the implications of our observations in the
previous sections for the approximation of functions in the class
$C^{k,s}$.

\begin{definition}\label{def:ckalphaSpace}
  Let $k\in\N_0$, $s\in [0,1]$ and $\Omega\subseteq\R^d$.
  Then for $f:\Omega\to\R$
\begin{align}\label{eq:Hoelder2}
	\begin{split}
	\norm[C^{k,s}(\Omega)]{f}\dfn&
	\sup_{\Bx\in \Omega}\max_{\set{\Balpha\in \N_0^d}{|\Balpha|\le k}}
	|D^\Balpha f(\Bx)| \\
	&+
	\sup_{\Bx\neq\By\in \Omega}
	\max_{\set{\Balpha\in \N_0^d}{|\Balpha|=k}}
	\frac{|D^\Balpha f(\Bx)-D^\Balpha f(\By)|}{\norm[]{\Bx-\By}^s},
	\end{split}
\end{align}
and we denote by $C^{k,s}(\Omega)$ the set of functions $f\in C^k(\Omega)$ for
which $\norm[C^{k,s}(\Omega)]{f}<\infty$.
\end{definition}

Note that these spaces are ordered according to
\begin{align*}
	C^k(\Omega)\supseteq C^{k,s}(\Omega)\supseteq C^{k,t}(\Omega)\supseteq C^{k+1}(\Omega)
\end{align*}
for all $0<s\le t\le 1$.

In order to state our main result, we first recall a version of
Taylor's remainder formula for $C^{k,s}(\Omega)$ functions.

\begin{lemma}\label{lemma:taylor}
Let $d\in\N$, $k\in\N$, $s\in [0,1]$, $\Omega=[0,1]^d$ and
$f\in C^{k,s}(\Omega)$.
Then for all $\Ba$, $\Bx\in\Omega$
\begin{align}
	  f(\Bx)=\sum_{\set{\Balpha\in\N_0^d}{0\le |\Balpha|\le k}}
	  \frac{D^\Balpha f(\Ba)}{\Balpha!}(\Bx-\Ba)^\Balpha+R_k(\Bx)
\end{align}
where with $h\dfn \max_{i\le d}|a_i-x_i|$ we have
$|R_k(\Bx)|\le h^{k+s}\frac{d^{k+1/2}}{k!}\norm[C^{k,s}(\Omega)]{f}$.
\end{lemma}

\begin{proof}
First, for a function $g\in C^{k}(\R)$ and $a$, $t\in\R$
\begin{align*}
	  g(t)&=\sum_{j=0}^{k-1}\frac{g^{(j)}(a)}{j!} (t-a)^j + \frac{g^{(k)}(\xi)}{k!}(t-a)^{k}\\
	&=\sum_{j=0}^{k}\frac{g^{(j)}(a)}{j!} (t-a)^j + \frac{g^{(k)}(\xi)-g^{(k)}(a)}{k!}(t-a)^{k},
\end{align*}
for some $\xi$ between $a$ and $t$.
Now let $f\in C^{k,s}(\R^d)$
and $\Ba$, $\Bx\in\R^d$.
Thus with $g(t)\dfn f(\Ba+t \cdot (\Bx-\Ba))$
holds for $f(\Bx)=g(1)$
\begin{align*}
	  f(\Bx)= \sum_{j=0}^{k-1}\frac{g^{(j)}(0)}{j!} + \frac{g^{(k)}(\xi)}{k!}.
\end{align*}
By the chain rule
\begin{align*}
	  g^{(j)}(t) = \sum_{\set{\Balpha\in\N_0^d}{|\Balpha|=j}}\binom{j}{\Balpha}D^\Balpha f(\Ba+t \cdot (\Bx-\Ba))(\Bx-\Ba)^\Balpha,
\end{align*}
where we use the multivariate notations
$\binom{j}{\Balpha}=\frac{j!}{\Balpha!}=\frac{j!}{\prod_{j=1}^d\alpha_j!}$
and $(\Bx-\Ba)^\Balpha=\prod_{j=1}^d(x_j-a_j)^{\alpha_j}$.
Hence
\begin{align*}
	  f(\Bx) = &\underbrace{\sum_{\set{\Balpha\in\N_0^d}{0\le |\Balpha|\le k}}
	    \frac{D^\Balpha f(\Ba)}{\Balpha!}(\Bx-\Ba)^\Balpha}_{\in \P_k} \\
	 & \qquad +\underbrace{\sum_{|\Balpha|=k} \frac{D^\Balpha f(\Ba+\xi\cdot (\Bx-\Ba))-D^\Balpha f(\Ba)}{\Balpha!}(\Bx-\Ba)^{\Balpha}}_{\dfnn R_k},
\end{align*}
for some $\xi\in [0,1]$.
Using the definition of $h$, the remainder term can be bounded by
\begin{align*}
	  |R_k|&\le h^{k}  \max_{|\Balpha|=k}\sup_{\substack{\Bx\in\Omega\\t\in [0,1]}}|D^\Balpha f(\Ba+t\cdot (\Bx-\Ba))-D^\Balpha f(\Ba)|
	  \frac{1}{k!}\sum_{\set{\Balpha\in\N_0^d}{|\Balpha|=k}}\binom{k}{\Balpha}\nonumber\\
	       &\le h^{k+s}\frac{d^{k+\frac s 2}}{k!}\norm[C^{k,s}({\Omega})]{f},
\end{align*}
where we used \eqref{eq:Hoelder2}, $\norm[]{\Bx-\Ba}\le \sqrt{d}h$, 
and $\sum_{\set{\Balpha\in\N_0^d}{|\Balpha|=k}}\binom{k}{\Balpha}=(1+\cdots+1)^k=d^k$
by the multinomial formula.
\end{proof}

We now come to the main statement of this section.
Up to logarithmic terms, it shows the convergence rate $(k+s)/{d}$ for
approximating functions in $C^{k,s}([0,1]^d)$.

\begin{theorem}\label{thm:Cks}
  Let $d\in\N$, $k\in\N_0$, and $s\in [0,1]$.

There exists a constant $C>0$
  and for every $N\in\N$ there exists a ReLU
  neural network $\Phi_N(\Bx,\Bw)$ %
  with $\Bw\in\R^{\tilde N}$,
such that
\begin{equation*}
  \size(\Phi_N)=\tilde N\le CN\log(N)\qquad\text{and}\qquad\depth(\Phi_N)\le C \log(N),
\end{equation*}
and for every $f\in C^{k,s}(\Omega)$
\begin{align}\label{eq:Ckserr}
  \inf_{\Bw\in\R^{\tilde N}}\norm[{C^0([0,1]^d)}]{f(\Bx)-\Phi_N(\Bx,\Bw)}\le
	  C
  \norm[C^{k,s}(\Omega)]{f}
N^{-\frac{k+s}{d}}.
\end{align}
\end{theorem}

\begin{proof}
The idea of the proof is to use the so-called ``partition of unity
method'': First we will construct a partition of unity
$(\varphi_\Bnu)_{\Bnu}$, such that for an appropriately chosen $M \in \N$ each $\varphi_\Bnu$ has support
on a $O({1}/{M})$ neighborhood of a point $\Beta\in\Omega$.
On
each of these neighborhoods we will use the local Taylor
polynomial $p_\Bnu$ of $f$ around $\Beta$ to approximate the
function.
Then $\sum_{\Bnu}\varphi_\Bnu p_\Bnu$ gives an
approximation to $f$ on $\Omega$.
This approximation can be emulated by
a neural network of the type
$\sum_{\Bnu}\ntim{\eps}(\varphi_\Bnu,\hat p_\Bnu)$, where
$\hat p_\Bnu$ is an neural network approximation to the polynomial $p_\Bnu$.

It suffices to show the theorem in the case where
\[
	\max\left\{\frac{d^{k+1/2}}{k!},\exp(d)\right\}\norm[C^{k,s}(\Omega)]{f}\le
	1.
\]
The general case can then be immediately deduced by a scaling argument.

{\bf Step 1.} We construct the neural network.
Define
\begin{align}\label{eq:defMeps}
	  M\dfn \lceil N^{1/d}\rceil\qquad\text{and}\qquad\eps\dfn N^{-\frac{k+s}{d}}.
\end{align}

Consider a uniform simplicial mesh with nodes
$\set{{{\Bnu}/{M}}}{\Bnu\le M}$ where
${{\Bnu}/{M}}\dfn ({\nu_1}/{M},\dots,{\nu_d}/{M})$,
and where ``$\Bnu\le M$'' is short for
$\set{\Bnu\in\N_0^d}{\nu_i\le M \text{ for all } i\le d}$.
We denote by
$\varphi_\Bnu$ the cpwl basis function on this mesh such that
$\varphi_\Bnu({{\Bnu}/{M}})=1$ and
$\varphi_\Bnu({{\Bmu}/{M}}) = 0$ whenever
$\Bmu\neq\Bnu$.
As shown in Chapter \ref{chap:ReLUNNs},
$\varphi_\Bnu$ is a neural network of size $O(1)$.
Then
\begin{align}\label{eq:pum}
	  \sum_{\Bnu\le M}\varphi_\Bnu \equiv 1\qquad\text{on }\Omega,
\end{align}
is a partition of unity.
Moreover, observe that
\begin{align}\label{eq:suppphinu}
	  \supp(\varphi_\Bnu)\subseteq \setc{\Bx\in\Omega}{\normc[\infty]{\Bx-{\frac{\Bnu}{M}}}\le \frac{1}{M}},
\end{align}
where $\norm[\infty]{\Bx}=\max_{i\le d}|x_i|$.

For each $\Bnu\le M$ define the multivariate polynomial
\begin{align*}
	  p_\Bnu(\Bx)\dfn
	  \sum_{|\Balpha|\le k}
	  \frac{D^\Balpha f\left({\frac{\Bnu}{M}}\right)}{\Balpha!}\left(\Bx-{\frac{\Bnu}{M}}\right)^\Balpha\in \P_k,
\end{align*}
and the approximation
\begin{align*}
	  \hat p_\Bnu(\Bx)\dfn
	  \sum_{|\Balpha|\le k}
	  \frac{D^\Balpha f\left({\frac{\Bnu}{M}}\right)}{\Balpha!}
	  \ntim{|\Balpha|,\eps}\left(x_{i_{\Balpha,1}}-\frac{\nu_{i_{\Balpha,1}}}{M},\dots,x_{i_{\Balpha,k}}-\frac{\nu_{i_{\Balpha,k}}}{M}\right),
\end{align*}
where $(i_{\Balpha,1},\dots,i_{\Balpha,k})\in
\{0,\dots,d\}^k$ is arbitrary but fixed such that
$|\set{j}{i_{\Balpha,j}=r}|=\alpha_r$ for all
$r=1,\dots,d$.
Finally, define
\begin{align}\label{eq:CksPhiN}
	  \Phi_N^f\dfn \sum_{\Bnu\le M}\ntim{\eps}(\varphi_\Bnu,\hat p_\Bnu),
\end{align}
and note that the underlying architecture is independent of $f$, i.e.\
  $\Phi_N^f(\Bx)=\Phi_N(\Bx,\Bw_f)$ for some network architecture $\Phi_N$ and certain $f$-dependent parameters $\Bw_f$.

{\bf Step 2.} We bound the approximation error.
First,
for each $\Bx\in\Omega$, using \eqref{eq:pum} and \eqref{eq:suppphinu}
\begin{align*}
	  \left|f(\Bx)-\sum_{\Bnu\le M}\varphi_\Bnu(\Bx)p_\Bnu(\Bx)\right|
	  &\le \sum_{\Bnu\le M}|\varphi_\Bnu(\Bx)|
	    |p_\Bnu(\Bx)-f(\Bx)|\nonumber\\
	  &\le \max_{\Bnu\le M} \sup_{\set{\By\in\Omega}{\norm[\infty]{{\frac{\Bnu}{M}}-\By}\le \frac{1}{M}}}
	    |f(\By)-p_\Bnu(\By)|.
\end{align*}
By Lemma \ref{lemma:taylor} we obtain
\begin{align}\label{eq:PhiNerror1}
	  \sup_{\Bx\in\Omega}\left|f(\Bx)-\sum_{\Bnu\le M}\varphi_\Bnu(\Bx)p_\Bnu(\Bx)\right|
	  \le M^{-(k+s)}\frac{d^{k+\frac{1}{2}}}{k!}\norm[C^{k,s}(\Omega)]{f}\le M^{-(k+s)}.
\end{align}

Next, fix $\Bnu\le M$ and $\By\in\Omega$ such that
$\norm[\infty]{{{\Bnu}/{M}}-\By}\le {1}/{M}\le 1$.
Then by
Proposition~\ref{prop:multn}
\begin{align}\label{eq:hatpbnu}
	  |p_\Bnu(\By)-\hat p_\Bnu(\By)|
	  &\le \sum_{|\Balpha|\le k}
	  \frac{D^\Balpha f\left({\frac{\Bnu}{M}}\right)}{\Balpha!}
	  \left|\prod_{j=1}^k \left(y_{i_{\Balpha,j}}-\frac{\nu_{i_{\Balpha,j}}}{M}\right)\right.
	\nonumber
	  \\
	&\qquad \left.
	- \ \ntim{|\Balpha|,\eps}\left(y_{i_{\Balpha,1}}-\frac{\nu_{i_{\Balpha,1}}}{M},\dots,y_{i_{\Balpha,k}}-\frac{\nu_{i_{\Balpha,k}}}{M}\right)\right|\nonumber\\
	  &\le \eps \sum_{|\Balpha|\le k}
	    \frac{D^\Balpha f({\frac{\Bnu}{M}})}{\Balpha!}
	    \le \eps \exp(d)\norm[C^{k,s}(\Omega)]{f}\le \eps,
\end{align}
where we used $|D^\Balpha f({{\Bnu}/{M}})|\le \norm[C^{k,s}(\Omega)]{f}$
and
\begin{align*}
	  \sum_{\set{\Balpha\in\N_0^d}{|\Balpha|\le k}}\frac{1}{\Balpha!}=
	  \sum_{j=0}^k\frac{1}{j!}\sum_{\set{\Balpha\in\N_0^d}{|\Balpha|=j}}\frac{j!}{\Balpha!}=
	  \sum_{j=0}^k\frac{d^j}{j!}
	  \le \sum_{j=0}^\infty\frac{d^j}{j!}
	  = \exp(d).
\end{align*}
Similarly, one shows that
\begin{align*}
	  |\hat p_\Bnu(\Bx)|\le \exp(d)\norm[C^{k,s}(\Omega)]{f}\le 1\qquad\text{ for all } \Bx\in\Omega.
\end{align*}

Fix $\Bx\in\Omega$.
Then $\Bx$ belongs to a simplex of the mesh, and
thus $\Bx$ can be in the support of at most $d+1$ (the number of
nodes of a simplex) functions $\varphi_\Bnu$.
Moreover, Lemma
\ref{lemma:mult} implies that
$\supp\ntim{\eps}(\varphi_\Bnu(\cdot),\hat
p_\Bnu(\cdot))\subseteq \supp\varphi_\Bnu$.
Hence, using Lemma
\ref{lemma:mult} and \eqref{eq:hatpbnu}
\begin{align*}
	  &\left|\sum_{\Bnu\le M}\varphi_\Bnu(\Bx)p_\Bnu(\Bx)-
	    \sum_{\Bnu\le M}\ntim{\eps}(\varphi_\Bnu(\Bx),\hat p_\Bnu(\Bx))
	  \right| \nonumber\\
	  &\qquad\le \sum_{\set{\Bnu\le M}{\Bx\in\supp\varphi_\Bnu}}
	    \left(|\varphi_\Bnu(\Bx)p_\Bnu(\Bx)-
	\varphi_\Bnu(\Bx)\hat p_\Bnu(\Bx)| \right.\\ 
	&\qquad \qquad \left.
	+ \  |\varphi_\Bnu(\Bx)\hat p_\Bnu(\Bx)-\ntim{\eps}(\varphi_\Bnu(\Bx),\hat p_\Bnu(\Bx))|\right)\nonumber\\
	  &\qquad\le \eps + (d+1)\eps=(d+2)\eps.
\end{align*}
In total, together with \eqref{eq:PhiNerror1}
\begin{align*}
	  \sup_{\Bx\in\Omega}|f(\Bx)-\Phi_N^f(\Bx)|\le M^{-(k+s)}+\eps \cdot (d+2).
\end{align*}
With our choices in \eqref{eq:defMeps} this yields the error
bound \eqref{eq:Ckserr}.

{\bf Step 3.} It remains to bound the size and depth of the neural 
network in \eqref{eq:CksPhiN}.

By Lemma \ref{lemma:basis}, for each
$0\le\Bnu\le M$ we have
\begin{align}
	  \size(\varphi_\Bnu)\le C\cdot (1+k_\CT),\qquad
	  \depth(\varphi_\Bnu)\le C\cdot (1+\log(k_\CT)),
\end{align}
where $k_\CT$ is the maximal number of simplices attached to a
node in the mesh.
Note that $k_\CT$ is independent of $M$, so that
the size and depth of $\varphi_\Bnu$ are bounded by a constant
$C_\varphi$ independent of $M$.

Lemma~\ref{lemma:mult} and Proposition~\ref{prop:multn} thus imply
with our choice of $\eps=N^{-(k+s)/d}$
\begin{align*}
	  \depth(\Phi_N^f)&=\depth(\ntim{\eps})+\max_{\Bnu\le M}\depth(\varphi_\Beta)+\max_{\Bnu\le M}\depth(\hat p_\Bnu)\nonumber\\
	                &\le C\cdot(1+|\log(\eps)|+C_\varphi)+\depth(\ntim{k,\eps})\nonumber\\
	                &\le C\cdot(1+|\log(\eps)|+C_\varphi)\nonumber\\
	                &\le C\cdot(1+\log(N))
\end{align*}
for some constant $C>0$ depending on $k$ and $d$ (we use ``$C$'' to
denote a generic constant that can change its value in each line).

To bound the size, we first observe with Lemma \ref{lemma:addition} that
\begin{align*}
	  \size(\hat p_\Bnu)\le C\cdot \left(1+\sum_{|\Balpha|\le k}\size\left(\ntim{|\Balpha|,\eps}\right)\right)\le C\cdot  (1+|\log(\eps)|)
	  \end{align*}
for some $C$ depending on $k$.
Thus, for the size of $\Phi_N^f$
we obtain with $M=\lceil N^{1/d}\rceil$
\begin{align*}
	  \size(\Phi_N^f)&\le C\cdot \left(1+\sum_{\Bnu\le M}\left(\size(\ntim{\eps})+\size(\varphi_\Bnu)+\size(\hat p_\Bnu)\right)\right)\nonumber\\
	               &\le C\cdot (1+M)^d(1+|\log(\eps)|+C_\varphi)\nonumber\\
	               &\le C\cdot (1+N^{1/d})^d(1+C_\varphi+\log(N))\nonumber\\
	               &\le CN\log(N),
\end{align*}
which concludes the proof.
\end{proof}

Theorem~\ref{thm:Cks} is similar in spirit to \cite[Section 3.2]{yarotsky};
  the main differences are that \cite{yarotsky} considers the class $C^k([0,1]^d)$
  instead of $C^{k,s}([0,1]^d)$, and uses an approximate partition of unity, while
  we use the exact partition of unity constructed in Chapter \ref{chap:ReLUNNs}.
Up to logarithmic terms,
the theorem shows the convergence rate $(k+s)/{d}$.
As long as $k$ is large, in principle we can achieve arbitrarily
large (and $d$-independent if $k\ge d$) convergence rates.
In contrast to Theorem~\ref{thm:hoelder}, achieving error
$N^{-\frac{k+s}{d}}$ requires %
depth $O(\log(N))$, i.e.\ %
the neural network depth is required to increase.
This can be avoided however,
  and networks of depth $O(k/d)$ suffice
  to attain these convergence rates \cite{petersen2018optimal}.

\begin{remark}\label{rmk:Cks}
Let $L:\Bx\mapsto\BA\Bx+\Bb:\R^d\to\R^d$ be a bijective affine
transformation and set $\Omega\dfn L([0,1]^d)\subseteq\R^d$.
Then
for a function $f\in C^{k,s}(\Omega)$, by Theorem~\ref{thm:Cks} there
exists a neural network $\Phi_N^f$ such that
\begin{align*}
	  \sup_{\Bx\in\Omega}|f(\Bx)-\Phi_N^f(L^{-1}(\Bx))|
	  &=\sup_{\Bx\in [0,1]^d}|f(L(\Bx))-\Phi_N^f(\Bx)|\\
	  &\le C \norm[C^{k,s}({[0,1]^d})]{f\circ L} N^{-\frac{k+s}{d}}.
\end{align*}
Since for $\Bx\in [0,1]^d$ holds
$|f(L(\Bx))|\le \sup_{\By\in\Omega}|f(\By)|$ and, if
$\boldsymbol{0}\neq \Balpha\in\N_0^d$ is a multiindex, then 
$|D^\Balpha (f(L(\Bx))|\le \norm[]{A}^{|\Balpha|}\sup_{\By\in
\Omega}|D^\Balpha f(\By)|$, we have
$\norm[C^{k,s}({[0,1]^d})]{f\circ L}\le
(1+\norm[]{A}^{k+s})\norm[C^{k,s}(\Omega)]{f}$.
Thus the convergence
rate $N^{-\frac{k+s}{d}}$ is achieved on every set of the type
$L([0,1]^d)$ for an affine map $L$, and in particular on every
hypercube $\times_{j=1}^d[a_j,b_j]$.
\end{remark}

\section*{Bibliography and further reading}
This chapter is based on the seminal 2017 paper by Yarotsky
\cite{yarotsky}, where the construction of approximating the square
function, the multiplication, and polynomials (discussed in Sections
\ref{sec:squareFunction}, \ref{sec:multiplication},
\ref{sec:depthseparation}) was first introduced and analyzed. The
construction relies on the sawtooth function discussed in Section
\ref{sec:tightness} and originally constructed by Telgarsky in
\cite{telgarsky_depth}.  Similar results were obtained around the same
  time by Liang and Srikant via a bit extraction technique using both
  the ReLU and the Heaviside function as activation functions
  \cite{WhyDeep2017}.  These works have since sparked a large body of
  research, as they allow to lift polynomial approximation theory to
  neural network classes. Convergence results based on this type of
argument include for example
\cite{petersen2018optimal,ECKLE2019232,doi:10.1137/18M1189336,MR3856963,OSZ21}.
We also refer to \cite{pmlr-v70-telgarsky17a} for related results
  on rational approximation.

  The depth separation result in Section \ref{sec:depthseparation} is
  based on the exponential convergence rates obtained for analytic
  functions in \cite{MR3856963,OSZ21}, also see \cite[Lemma III.7]{9363169}.
  For the approximation of
    polynomials with ReLU neural networks stated in Lemma
    \ref{lemma:polynomial}, see, e.g.,
    \cite{petersen2018optimal,9363169,opschoor2020fem}, and also
    \cite{opschoordiss,OPSCHOOR2024142} for constructions based on Chebyshev
    polynomials, which can be more efficient. For further depth
  separation results, we refer to
  \cite{telgarsky_depth,telgarsky_depth2,pmlr-v49-eldan16,Safran2016DepthSI,arora2018understanding}. Moreover,
  closely related to such statements is the 1987 thesis by Håstad
  \cite{hastad1986}, which considers the limitations of logic circuits
  in terms of depth. 

  The approximation result derived in Section \ref{sec:CksFunctions}
  for $C^{k,s}$ functions follows by standard approximation
  theory for piecewise polynomial functions, and is similar as in
  \cite{yarotsky}. We point out that %
such statements
  can also be shown for other activation functions
  than ReLU; see in particular the works of Mhaskar
  \cite{mhaskar1993approximation,mhaskar1996neural} and Section 6 in
  Pinkus' Acta Numerica article \cite{MR1819645} for sigmoidal and
  smooth activations. Additionally, the more recent paper
  \cite{DERYCK2021732} specifically addresses the hyperbolic tangent
  activation. Finally, \cite{guhring2021approximation} studies general
  activation functions that allow for the construction of approximate
  partitions of unity.

  \newpage

  \section*{Exercises}

  \begin{exercise}\label{ex:depthseparation}
    We show another type of depth separation result:
    Let $d\ge 2$. Prove that there exist ReLU NNs $\Phi:\R^d\to\R$ of depth two,
    which cannot be represented exactly by ReLU NNs $\Phi:\R^d\to\R$ of depth one.

    \emph{Hint}: Show that nonzero ReLU NNs of depth one necessarily have unbounded support.
  \end{exercise}
  
  \begin{exercise}\label{ex:polynomial}
    Prove Lemma \ref{lemma:polynomial}.

    \emph{Hint}: Proceed by induction over the iteration depth in
    Figure \ref{fig:monomials}.
  \end{exercise}

  \begin{exercise}\label{ex:analytic}
    Show Proposition \ref{prop:analytic} in the general case where the Taylor series of $f$
    only converges locally (see proof of Proposition \ref{prop:analytic}).

    \emph{Hint}: Use the partition of unity method from the proof of Theorem \ref{thm:Cks}.
  \end{exercise}

%% file: plots/square.tex
\begin{tikzpicture}[scale=0.9]
\draw [<->,thick] (0,2.3) -- (0,0) -- (3.3,0);
\draw [-] (3,0.11499999999999999) -- (3,-0.11499999999999999) node [below] {\small $1$};
\draw [-] (0.165,2) -- (-0.165,2) node [left] {\small $\frac{1}{4}$};
\draw [thick,blue,dashed] (0.000,0.000) -- (0.006,0.008) -- (0.012,0.016) -- (0.018,0.024) -- (0.024,0.032) -- (0.030,0.040) -- (0.036,0.048) -- (0.042,0.056) -- (0.048,0.064) -- (0.054,0.072) -- (0.060,0.080) -- (0.066,0.088) -- (0.072,0.096) -- (0.078,0.104) -- (0.084,0.112) -- (0.090,0.120) -- (0.096,0.128) -- (0.102,0.136) -- (0.108,0.144) -- (0.114,0.152) -- (0.120,0.160) -- (0.126,0.168) -- (0.132,0.176) -- (0.138,0.184) -- (0.144,0.192) -- (0.150,0.200) -- (0.156,0.208) -- (0.162,0.216) -- (0.168,0.224) -- (0.174,0.232) -- (0.180,0.240) -- (0.186,0.248) -- (0.192,0.257) -- (0.198,0.265) -- (0.204,0.273) -- (0.210,0.281) -- (0.216,0.289) -- (0.222,0.297) -- (0.228,0.305) -- (0.234,0.313) -- (0.240,0.321) -- (0.246,0.329) -- (0.253,0.337) -- (0.259,0.345) -- (0.265,0.353) -- (0.271,0.361) -- (0.277,0.369) -- (0.283,0.377) -- (0.289,0.385) -- (0.295,0.393) -- (0.301,0.401) -- (0.307,0.409) -- (0.313,0.417) -- (0.319,0.425) -- (0.325,0.433) -- (0.331,0.441) -- (0.337,0.449) -- (0.343,0.457) -- (0.349,0.465) -- (0.355,0.473) -- (0.361,0.481) -- (0.367,0.489) -- (0.373,0.497) -- (0.379,0.505) -- (0.385,0.513) -- (0.391,0.521) -- (0.397,0.529) -- (0.403,0.537) -- (0.409,0.545) -- (0.415,0.553) -- (0.421,0.561) -- (0.427,0.569) -- (0.433,0.577) -- (0.439,0.585) -- (0.445,0.593) -- (0.451,0.601) -- (0.457,0.609) -- (0.463,0.617) -- (0.469,0.625) -- (0.475,0.633) -- (0.481,0.641) -- (0.487,0.649) -- (0.493,0.657) -- (0.499,0.665) -- (0.505,0.673) -- (0.511,0.681) -- (0.517,0.689) -- (0.523,0.697) -- (0.529,0.705) -- (0.535,0.713) -- (0.541,0.721) -- (0.547,0.729) -- (0.553,0.737) -- (0.559,0.745) -- (0.565,0.754) -- (0.571,0.762) -- (0.577,0.770) -- (0.583,0.778) -- (0.589,0.786) -- (0.595,0.794) -- (0.601,0.802) -- (0.607,0.810) -- (0.613,0.818) -- (0.619,0.826) -- (0.625,0.834) -- (0.631,0.842) -- (0.637,0.850) -- (0.643,0.858) -- (0.649,0.866) -- (0.655,0.874) -- (0.661,0.882) -- (0.667,0.890) -- (0.673,0.898) -- (0.679,0.906) -- (0.685,0.914) -- (0.691,0.922) -- (0.697,0.930) -- (0.703,0.938) -- (0.709,0.946) -- (0.715,0.954) -- (0.721,0.962) -- (0.727,0.970) -- (0.733,0.978) -- (0.739,0.986) -- (0.745,0.994) -- (0.752,1.002) -- (0.758,1.010) -- (0.764,1.018) -- (0.770,1.026) -- (0.776,1.034) -- (0.782,1.042) -- (0.788,1.050) -- (0.794,1.058) -- (0.800,1.066) -- (0.806,1.074) -- (0.812,1.082) -- (0.818,1.090) -- (0.824,1.098) -- (0.830,1.106) -- (0.836,1.114) -- (0.842,1.122) -- (0.848,1.130) -- (0.854,1.138) -- (0.860,1.146) -- (0.866,1.154) -- (0.872,1.162) -- (0.878,1.170) -- (0.884,1.178) -- (0.890,1.186) -- (0.896,1.194) -- (0.902,1.202) -- (0.908,1.210) -- (0.914,1.218) -- (0.920,1.226) -- (0.926,1.234) -- (0.932,1.242) -- (0.938,1.251) -- (0.944,1.259) -- (0.950,1.267) -- (0.956,1.275) -- (0.962,1.283) -- (0.968,1.291) -- (0.974,1.299) -- (0.980,1.307) -- (0.986,1.315) -- (0.992,1.323) -- (0.998,1.331) -- (1.004,1.339) -- (1.010,1.347) -- (1.016,1.355) -- (1.022,1.363) -- (1.028,1.371) -- (1.034,1.379) -- (1.040,1.387) -- (1.046,1.395) -- (1.052,1.403) -- (1.058,1.411) -- (1.064,1.419) -- (1.070,1.427) -- (1.076,1.435) -- (1.082,1.443) -- (1.088,1.451) -- (1.094,1.459) -- (1.100,1.467) -- (1.106,1.475) -- (1.112,1.483) -- (1.118,1.491) -- (1.124,1.499) -- (1.130,1.507) -- (1.136,1.515) -- (1.142,1.523) -- (1.148,1.531) -- (1.154,1.539) -- (1.160,1.547) -- (1.166,1.555) -- (1.172,1.563) -- (1.178,1.571) -- (1.184,1.579) -- (1.190,1.587) -- (1.196,1.595) -- (1.202,1.603) -- (1.208,1.611) -- (1.214,1.619) -- (1.220,1.627) -- (1.226,1.635) -- (1.232,1.643) -- (1.238,1.651) -- (1.244,1.659) -- (1.251,1.667) -- (1.257,1.675) -- (1.263,1.683) -- (1.269,1.691) -- (1.275,1.699) -- (1.281,1.707) -- (1.287,1.715) -- (1.293,1.723) -- (1.299,1.731) -- (1.305,1.739) -- (1.311,1.747) -- (1.317,1.756) -- (1.323,1.764) -- (1.329,1.772) -- (1.335,1.780) -- (1.341,1.788) -- (1.347,1.796) -- (1.353,1.804) -- (1.359,1.812) -- (1.365,1.820) -- (1.371,1.828) -- (1.377,1.836) -- (1.383,1.844) -- (1.389,1.852) -- (1.395,1.860) -- (1.401,1.868) -- (1.407,1.876) -- (1.413,1.884) -- (1.419,1.892) -- (1.425,1.900) -- (1.431,1.908) -- (1.437,1.916) -- (1.443,1.924) -- (1.449,1.932) -- (1.455,1.940) -- (1.461,1.948) -- (1.467,1.956) -- (1.473,1.964) -- (1.479,1.972) -- (1.485,1.980) -- (1.491,1.988) -- (1.497,1.996) -- (1.503,1.996) -- (1.509,1.988) -- (1.515,1.980) -- (1.521,1.972) -- (1.527,1.964) -- (1.533,1.956) -- (1.539,1.948) -- (1.545,1.940) -- (1.551,1.932) -- (1.557,1.924) -- (1.563,1.916) -- (1.569,1.908) -- (1.575,1.900) -- (1.581,1.892) -- (1.587,1.884) -- (1.593,1.876) -- (1.599,1.868) -- (1.605,1.860) -- (1.611,1.852) -- (1.617,1.844) -- (1.623,1.836) -- (1.629,1.828) -- (1.635,1.820) -- (1.641,1.812) -- (1.647,1.804) -- (1.653,1.796) -- (1.659,1.788) -- (1.665,1.780) -- (1.671,1.772) -- (1.677,1.764) -- (1.683,1.756) -- (1.689,1.747) -- (1.695,1.739) -- (1.701,1.731) -- (1.707,1.723) -- (1.713,1.715) -- (1.719,1.707) -- (1.725,1.699) -- (1.731,1.691) -- (1.737,1.683) -- (1.743,1.675) -- (1.749,1.667) -- (1.756,1.659) -- (1.762,1.651) -- (1.768,1.643) -- (1.774,1.635) -- (1.780,1.627) -- (1.786,1.619) -- (1.792,1.611) -- (1.798,1.603) -- (1.804,1.595) -- (1.810,1.587) -- (1.816,1.579) -- (1.822,1.571) -- (1.828,1.563) -- (1.834,1.555) -- (1.840,1.547) -- (1.846,1.539) -- (1.852,1.531) -- (1.858,1.523) -- (1.864,1.515) -- (1.870,1.507) -- (1.876,1.499) -- (1.882,1.491) -- (1.888,1.483) -- (1.894,1.475) -- (1.900,1.467) -- (1.906,1.459) -- (1.912,1.451) -- (1.918,1.443) -- (1.924,1.435) -- (1.930,1.427) -- (1.936,1.419) -- (1.942,1.411) -- (1.948,1.403) -- (1.954,1.395) -- (1.960,1.387) -- (1.966,1.379) -- (1.972,1.371) -- (1.978,1.363) -- (1.984,1.355) -- (1.990,1.347) -- (1.996,1.339) -- (2.002,1.331) -- (2.008,1.323) -- (2.014,1.315) -- (2.020,1.307) -- (2.026,1.299) -- (2.032,1.291) -- (2.038,1.283) -- (2.044,1.275) -- (2.050,1.267) -- (2.056,1.259) -- (2.062,1.251) -- (2.068,1.242) -- (2.074,1.234) -- (2.080,1.226) -- (2.086,1.218) -- (2.092,1.210) -- (2.098,1.202) -- (2.104,1.194) -- (2.110,1.186) -- (2.116,1.178) -- (2.122,1.170) -- (2.128,1.162) -- (2.134,1.154) -- (2.140,1.146) -- (2.146,1.138) -- (2.152,1.130) -- (2.158,1.122) -- (2.164,1.114) -- (2.170,1.106) -- (2.176,1.098) -- (2.182,1.090) -- (2.188,1.082) -- (2.194,1.074) -- (2.200,1.066) -- (2.206,1.058) -- (2.212,1.050) -- (2.218,1.042) -- (2.224,1.034) -- (2.230,1.026) -- (2.236,1.018) -- (2.242,1.010) -- (2.248,1.002) -- (2.255,0.994) -- (2.261,0.986) -- (2.267,0.978) -- (2.273,0.970) -- (2.279,0.962) -- (2.285,0.954) -- (2.291,0.946) -- (2.297,0.938) -- (2.303,0.930) -- (2.309,0.922) -- (2.315,0.914) -- (2.321,0.906) -- (2.327,0.898) -- (2.333,0.890) -- (2.339,0.882) -- (2.345,0.874) -- (2.351,0.866) -- (2.357,0.858) -- (2.363,0.850) -- (2.369,0.842) -- (2.375,0.834) -- (2.381,0.826) -- (2.387,0.818) -- (2.393,0.810) -- (2.399,0.802) -- (2.405,0.794) -- (2.411,0.786) -- (2.417,0.778) -- (2.423,0.770) -- (2.429,0.762) -- (2.435,0.754) -- (2.441,0.745) -- (2.447,0.737) -- (2.453,0.729) -- (2.459,0.721) -- (2.465,0.713) -- (2.471,0.705) -- (2.477,0.697) -- (2.483,0.689) -- (2.489,0.681) -- (2.495,0.673) -- (2.501,0.665) -- (2.507,0.657) -- (2.513,0.649) -- (2.519,0.641) -- (2.525,0.633) -- (2.531,0.625) -- (2.537,0.617) -- (2.543,0.609) -- (2.549,0.601) -- (2.555,0.593) -- (2.561,0.585) -- (2.567,0.577) -- (2.573,0.569) -- (2.579,0.561) -- (2.585,0.553) -- (2.591,0.545) -- (2.597,0.537) -- (2.603,0.529) -- (2.609,0.521) -- (2.615,0.513) -- (2.621,0.505) -- (2.627,0.497) -- (2.633,0.489) -- (2.639,0.481) -- (2.645,0.473) -- (2.651,0.465) -- (2.657,0.457) -- (2.663,0.449) -- (2.669,0.441) -- (2.675,0.433) -- (2.681,0.425) -- (2.687,0.417) -- (2.693,0.409) -- (2.699,0.401) -- (2.705,0.393) -- (2.711,0.385) -- (2.717,0.377) -- (2.723,0.369) -- (2.729,0.361) -- (2.735,0.353) -- (2.741,0.345) -- (2.747,0.337) -- (2.754,0.329) -- (2.760,0.321) -- (2.766,0.313) -- (2.772,0.305) -- (2.778,0.297) -- (2.784,0.289) -- (2.790,0.281) -- (2.796,0.273) -- (2.802,0.265) -- (2.808,0.257) -- (2.814,0.248) -- (2.820,0.240) -- (2.826,0.232) -- (2.832,0.224) -- (2.838,0.216) -- (2.844,0.208) -- (2.850,0.200) -- (2.856,0.192) -- (2.862,0.184) -- (2.868,0.176) -- (2.874,0.168) -- (2.880,0.160) -- (2.886,0.152) -- (2.892,0.144) -- (2.898,0.136) -- (2.904,0.128) -- (2.910,0.120) -- (2.916,0.112) -- (2.922,0.104) -- (2.928,0.096) -- (2.934,0.088) -- (2.940,0.080) -- (2.946,0.072) -- (2.952,0.064) -- (2.958,0.056) -- (2.964,0.048) -- (2.970,0.040) -- (2.976,0.032) -- (2.982,0.024) -- (2.988,0.016) -- (2.994,0.008) -- (3.000,0.000);
\draw [thick] (0,-1.1) -- (0.6,-1.1) node [right] {\small $x-x^2$};
\draw [thick,blue,dashed] (0,-1.8) -- (0.6,-1.8) node [right] {\small $\frac{h_1(x)}{4}$};
\draw [thick] (0.000,0.000) -- (0.030,0.080) -- (0.061,0.158) -- (0.091,0.235) -- (0.121,0.310) -- (0.152,0.384) -- (0.182,0.455) -- (0.212,0.526) -- (0.242,0.594) -- (0.273,0.661) -- (0.303,0.726) -- (0.333,0.790) -- (0.364,0.852) -- (0.394,0.913) -- (0.424,0.971) -- (0.455,1.028) -- (0.485,1.084) -- (0.515,1.138) -- (0.545,1.190) -- (0.576,1.241) -- (0.606,1.290) -- (0.636,1.337) -- (0.667,1.383) -- (0.697,1.427) -- (0.727,1.469) -- (0.758,1.510) -- (0.788,1.549) -- (0.818,1.587) -- (0.848,1.623) -- (0.879,1.657) -- (0.909,1.690) -- (0.939,1.721) -- (0.970,1.750) -- (1.000,1.778) -- (1.030,1.804) -- (1.061,1.828) -- (1.091,1.851) -- (1.121,1.872) -- (1.152,1.892) -- (1.182,1.910) -- (1.212,1.926) -- (1.242,1.941) -- (1.273,1.954) -- (1.303,1.966) -- (1.333,1.975) -- (1.364,1.983) -- (1.394,1.990) -- (1.424,1.995) -- (1.455,1.998) -- (1.485,2.000) -- (1.515,2.000) -- (1.545,1.998) -- (1.576,1.995) -- (1.606,1.990) -- (1.636,1.983) -- (1.667,1.975) -- (1.697,1.966) -- (1.727,1.954) -- (1.758,1.941) -- (1.788,1.926) -- (1.818,1.910) -- (1.848,1.892) -- (1.879,1.872) -- (1.909,1.851) -- (1.939,1.828) -- (1.970,1.804) -- (2.000,1.778) -- (2.030,1.750) -- (2.061,1.721) -- (2.091,1.690) -- (2.121,1.657) -- (2.152,1.623) -- (2.182,1.587) -- (2.212,1.549) -- (2.242,1.510) -- (2.273,1.469) -- (2.303,1.427) -- (2.333,1.383) -- (2.364,1.337) -- (2.394,1.290) -- (2.424,1.241) -- (2.455,1.190) -- (2.485,1.138) -- (2.515,1.084) -- (2.545,1.028) -- (2.576,0.971) -- (2.606,0.913) -- (2.636,0.852) -- (2.667,0.790) -- (2.697,0.726) -- (2.727,0.661) -- (2.758,0.594) -- (2.788,0.526) -- (2.818,0.455) -- (2.848,0.384) -- (2.879,0.310) -- (2.909,0.235) -- (2.939,0.158) -- (2.970,0.080) -- (3.000,0.000);
\draw [<->,thick] (4.2,2.3) -- (4.2,0) -- (7.5,0);
\draw [-] (7.2,0.11499999999999999) -- (7.2,-0.11499999999999999) node [below] {\small $1$};
\draw [-] (4.365,2) -- (4.035,2) node [left] {\small $\frac{1}{16}$};
\draw [thick,blue,dashed] (4.200,0.000) -- (4.206,0.016) -- (4.212,0.032) -- (4.218,0.048) -- (4.224,0.064) -- (4.230,0.080) -- (4.236,0.096) -- (4.242,0.112) -- (4.248,0.128) -- (4.254,0.144) -- (4.260,0.160) -- (4.266,0.176) -- (4.272,0.192) -- (4.278,0.208) -- (4.284,0.224) -- (4.290,0.240) -- (4.296,0.257) -- (4.302,0.273) -- (4.308,0.289) -- (4.314,0.305) -- (4.320,0.321) -- (4.326,0.337) -- (4.332,0.353) -- (4.338,0.369) -- (4.344,0.385) -- (4.350,0.401) -- (4.356,0.417) -- (4.362,0.433) -- (4.368,0.449) -- (4.374,0.465) -- (4.380,0.481) -- (4.386,0.497) -- (4.392,0.513) -- (4.398,0.529) -- (4.404,0.545) -- (4.410,0.561) -- (4.416,0.577) -- (4.422,0.593) -- (4.428,0.609) -- (4.434,0.625) -- (4.440,0.641) -- (4.446,0.657) -- (4.453,0.673) -- (4.459,0.689) -- (4.465,0.705) -- (4.471,0.721) -- (4.477,0.737) -- (4.483,0.754) -- (4.489,0.770) -- (4.495,0.786) -- (4.501,0.802) -- (4.507,0.818) -- (4.513,0.834) -- (4.519,0.850) -- (4.525,0.866) -- (4.531,0.882) -- (4.537,0.898) -- (4.543,0.914) -- (4.549,0.930) -- (4.555,0.946) -- (4.561,0.962) -- (4.567,0.978) -- (4.573,0.994) -- (4.579,1.010) -- (4.585,1.026) -- (4.591,1.042) -- (4.597,1.058) -- (4.603,1.074) -- (4.609,1.090) -- (4.615,1.106) -- (4.621,1.122) -- (4.627,1.138) -- (4.633,1.154) -- (4.639,1.170) -- (4.645,1.186) -- (4.651,1.202) -- (4.657,1.218) -- (4.663,1.234) -- (4.669,1.251) -- (4.675,1.267) -- (4.681,1.283) -- (4.687,1.299) -- (4.693,1.315) -- (4.699,1.331) -- (4.705,1.347) -- (4.711,1.363) -- (4.717,1.379) -- (4.723,1.395) -- (4.729,1.411) -- (4.735,1.427) -- (4.741,1.443) -- (4.747,1.459) -- (4.753,1.475) -- (4.759,1.491) -- (4.765,1.507) -- (4.771,1.523) -- (4.777,1.539) -- (4.783,1.555) -- (4.789,1.571) -- (4.795,1.587) -- (4.801,1.603) -- (4.807,1.619) -- (4.813,1.635) -- (4.819,1.651) -- (4.825,1.667) -- (4.831,1.683) -- (4.837,1.699) -- (4.843,1.715) -- (4.849,1.731) -- (4.855,1.747) -- (4.861,1.764) -- (4.867,1.780) -- (4.873,1.796) -- (4.879,1.812) -- (4.885,1.828) -- (4.891,1.844) -- (4.897,1.860) -- (4.903,1.876) -- (4.909,1.892) -- (4.915,1.908) -- (4.921,1.924) -- (4.927,1.940) -- (4.933,1.956) -- (4.939,1.972) -- (4.945,1.988) -- (4.952,1.996) -- (4.958,1.980) -- (4.964,1.964) -- (4.970,1.948) -- (4.976,1.932) -- (4.982,1.916) -- (4.988,1.900) -- (4.994,1.884) -- (5.000,1.868) -- (5.006,1.852) -- (5.012,1.836) -- (5.018,1.820) -- (5.024,1.804) -- (5.030,1.788) -- (5.036,1.772) -- (5.042,1.756) -- (5.048,1.739) -- (5.054,1.723) -- (5.060,1.707) -- (5.066,1.691) -- (5.072,1.675) -- (5.078,1.659) -- (5.084,1.643) -- (5.090,1.627) -- (5.096,1.611) -- (5.102,1.595) -- (5.108,1.579) -- (5.114,1.563) -- (5.120,1.547) -- (5.126,1.531) -- (5.132,1.515) -- (5.138,1.499) -- (5.144,1.483) -- (5.150,1.467) -- (5.156,1.451) -- (5.162,1.435) -- (5.168,1.419) -- (5.174,1.403) -- (5.180,1.387) -- (5.186,1.371) -- (5.192,1.355) -- (5.198,1.339) -- (5.204,1.323) -- (5.210,1.307) -- (5.216,1.291) -- (5.222,1.275) -- (5.228,1.259) -- (5.234,1.242) -- (5.240,1.226) -- (5.246,1.210) -- (5.252,1.194) -- (5.258,1.178) -- (5.264,1.162) -- (5.270,1.146) -- (5.276,1.130) -- (5.282,1.114) -- (5.288,1.098) -- (5.294,1.082) -- (5.300,1.066) -- (5.306,1.050) -- (5.312,1.034) -- (5.318,1.018) -- (5.324,1.002) -- (5.330,0.986) -- (5.336,0.970) -- (5.342,0.954) -- (5.348,0.938) -- (5.354,0.922) -- (5.360,0.906) -- (5.366,0.890) -- (5.372,0.874) -- (5.378,0.858) -- (5.384,0.842) -- (5.390,0.826) -- (5.396,0.810) -- (5.402,0.794) -- (5.408,0.778) -- (5.414,0.762) -- (5.420,0.745) -- (5.426,0.729) -- (5.432,0.713) -- (5.438,0.697) -- (5.444,0.681) -- (5.451,0.665) -- (5.457,0.649) -- (5.463,0.633) -- (5.469,0.617) -- (5.475,0.601) -- (5.481,0.585) -- (5.487,0.569) -- (5.493,0.553) -- (5.499,0.537) -- (5.505,0.521) -- (5.511,0.505) -- (5.517,0.489) -- (5.523,0.473) -- (5.529,0.457) -- (5.535,0.441) -- (5.541,0.425) -- (5.547,0.409) -- (5.553,0.393) -- (5.559,0.377) -- (5.565,0.361) -- (5.571,0.345) -- (5.577,0.329) -- (5.583,0.313) -- (5.589,0.297) -- (5.595,0.281) -- (5.601,0.265) -- (5.607,0.248) -- (5.613,0.232) -- (5.619,0.216) -- (5.625,0.200) -- (5.631,0.184) -- (5.637,0.168) -- (5.643,0.152) -- (5.649,0.136) -- (5.655,0.120) -- (5.661,0.104) -- (5.667,0.088) -- (5.673,0.072) -- (5.679,0.056) -- (5.685,0.040) -- (5.691,0.024) -- (5.697,0.008) -- (5.703,0.008) -- (5.709,0.024) -- (5.715,0.040) -- (5.721,0.056) -- (5.727,0.072) -- (5.733,0.088) -- (5.739,0.104) -- (5.745,0.120) -- (5.751,0.136) -- (5.757,0.152) -- (5.763,0.168) -- (5.769,0.184) -- (5.775,0.200) -- (5.781,0.216) -- (5.787,0.232) -- (5.793,0.248) -- (5.799,0.265) -- (5.805,0.281) -- (5.811,0.297) -- (5.817,0.313) -- (5.823,0.329) -- (5.829,0.345) -- (5.835,0.361) -- (5.841,0.377) -- (5.847,0.393) -- (5.853,0.409) -- (5.859,0.425) -- (5.865,0.441) -- (5.871,0.457) -- (5.877,0.473) -- (5.883,0.489) -- (5.889,0.505) -- (5.895,0.521) -- (5.901,0.537) -- (5.907,0.553) -- (5.913,0.569) -- (5.919,0.585) -- (5.925,0.601) -- (5.931,0.617) -- (5.937,0.633) -- (5.943,0.649) -- (5.949,0.665) -- (5.956,0.681) -- (5.962,0.697) -- (5.968,0.713) -- (5.974,0.729) -- (5.980,0.745) -- (5.986,0.762) -- (5.992,0.778) -- (5.998,0.794) -- (6.004,0.810) -- (6.010,0.826) -- (6.016,0.842) -- (6.022,0.858) -- (6.028,0.874) -- (6.034,0.890) -- (6.040,0.906) -- (6.046,0.922) -- (6.052,0.938) -- (6.058,0.954) -- (6.064,0.970) -- (6.070,0.986) -- (6.076,1.002) -- (6.082,1.018) -- (6.088,1.034) -- (6.094,1.050) -- (6.100,1.066) -- (6.106,1.082) -- (6.112,1.098) -- (6.118,1.114) -- (6.124,1.130) -- (6.130,1.146) -- (6.136,1.162) -- (6.142,1.178) -- (6.148,1.194) -- (6.154,1.210) -- (6.160,1.226) -- (6.166,1.242) -- (6.172,1.259) -- (6.178,1.275) -- (6.184,1.291) -- (6.190,1.307) -- (6.196,1.323) -- (6.202,1.339) -- (6.208,1.355) -- (6.214,1.371) -- (6.220,1.387) -- (6.226,1.403) -- (6.232,1.419) -- (6.238,1.435) -- (6.244,1.451) -- (6.250,1.467) -- (6.256,1.483) -- (6.262,1.499) -- (6.268,1.515) -- (6.274,1.531) -- (6.280,1.547) -- (6.286,1.563) -- (6.292,1.579) -- (6.298,1.595) -- (6.304,1.611) -- (6.310,1.627) -- (6.316,1.643) -- (6.322,1.659) -- (6.328,1.675) -- (6.334,1.691) -- (6.340,1.707) -- (6.346,1.723) -- (6.352,1.739) -- (6.358,1.756) -- (6.364,1.772) -- (6.370,1.788) -- (6.376,1.804) -- (6.382,1.820) -- (6.388,1.836) -- (6.394,1.852) -- (6.400,1.868) -- (6.406,1.884) -- (6.412,1.900) -- (6.418,1.916) -- (6.424,1.932) -- (6.430,1.948) -- (6.436,1.964) -- (6.442,1.980) -- (6.448,1.996) -- (6.455,1.988) -- (6.461,1.972) -- (6.467,1.956) -- (6.473,1.940) -- (6.479,1.924) -- (6.485,1.908) -- (6.491,1.892) -- (6.497,1.876) -- (6.503,1.860) -- (6.509,1.844) -- (6.515,1.828) -- (6.521,1.812) -- (6.527,1.796) -- (6.533,1.780) -- (6.539,1.764) -- (6.545,1.747) -- (6.551,1.731) -- (6.557,1.715) -- (6.563,1.699) -- (6.569,1.683) -- (6.575,1.667) -- (6.581,1.651) -- (6.587,1.635) -- (6.593,1.619) -- (6.599,1.603) -- (6.605,1.587) -- (6.611,1.571) -- (6.617,1.555) -- (6.623,1.539) -- (6.629,1.523) -- (6.635,1.507) -- (6.641,1.491) -- (6.647,1.475) -- (6.653,1.459) -- (6.659,1.443) -- (6.665,1.427) -- (6.671,1.411) -- (6.677,1.395) -- (6.683,1.379) -- (6.689,1.363) -- (6.695,1.347) -- (6.701,1.331) -- (6.707,1.315) -- (6.713,1.299) -- (6.719,1.283) -- (6.725,1.267) -- (6.731,1.251) -- (6.737,1.234) -- (6.743,1.218) -- (6.749,1.202) -- (6.755,1.186) -- (6.761,1.170) -- (6.767,1.154) -- (6.773,1.138) -- (6.779,1.122) -- (6.785,1.106) -- (6.791,1.090) -- (6.797,1.074) -- (6.803,1.058) -- (6.809,1.042) -- (6.815,1.026) -- (6.821,1.010) -- (6.827,0.994) -- (6.833,0.978) -- (6.839,0.962) -- (6.845,0.946) -- (6.851,0.930) -- (6.857,0.914) -- (6.863,0.898) -- (6.869,0.882) -- (6.875,0.866) -- (6.881,0.850) -- (6.887,0.834) -- (6.893,0.818) -- (6.899,0.802) -- (6.905,0.786) -- (6.911,0.770) -- (6.917,0.754) -- (6.923,0.737) -- (6.929,0.721) -- (6.935,0.705) -- (6.941,0.689) -- (6.947,0.673) -- (6.954,0.657) -- (6.960,0.641) -- (6.966,0.625) -- (6.972,0.609) -- (6.978,0.593) -- (6.984,0.577) -- (6.990,0.561) -- (6.996,0.545) -- (7.002,0.529) -- (7.008,0.513) -- (7.014,0.497) -- (7.020,0.481) -- (7.026,0.465) -- (7.032,0.449) -- (7.038,0.433) -- (7.044,0.417) -- (7.050,0.401) -- (7.056,0.385) -- (7.062,0.369) -- (7.068,0.353) -- (7.074,0.337) -- (7.080,0.321) -- (7.086,0.305) -- (7.092,0.289) -- (7.098,0.273) -- (7.104,0.257) -- (7.110,0.240) -- (7.116,0.224) -- (7.122,0.208) -- (7.128,0.192) -- (7.134,0.176) -- (7.140,0.160) -- (7.146,0.144) -- (7.152,0.128) -- (7.158,0.112) -- (7.164,0.096) -- (7.170,0.080) -- (7.176,0.064) -- (7.182,0.048) -- (7.188,0.032) -- (7.194,0.016) -- (7.200,0.000);
\draw [thick] (4.2,-1.1) -- (4.8,-1.1) node [right] {\small $x-x^2-\frac{h_1(x)}{4}$};
\draw [thick,blue,dashed] (4.2,-1.8) -- (4.8,-1.8) node [right] {\small $\frac{h_2(x)}{16}$};
\draw [thick] (4.200,0.000) -- (4.206,0.032) -- (4.212,0.064) -- (4.218,0.095) -- (4.224,0.126) -- (4.230,0.157) -- (4.236,0.188) -- (4.242,0.218) -- (4.248,0.248) -- (4.254,0.278) -- (4.260,0.308) -- (4.266,0.337) -- (4.272,0.366) -- (4.278,0.395) -- (4.284,0.424) -- (4.290,0.452) -- (4.296,0.480) -- (4.302,0.508) -- (4.308,0.536) -- (4.314,0.563) -- (4.320,0.590) -- (4.326,0.617) -- (4.332,0.643) -- (4.338,0.669) -- (4.344,0.696) -- (4.350,0.721) -- (4.356,0.747) -- (4.362,0.772) -- (4.368,0.797) -- (4.374,0.822) -- (4.380,0.846) -- (4.386,0.870) -- (4.392,0.894) -- (4.398,0.918) -- (4.404,0.942) -- (4.410,0.965) -- (4.416,0.988) -- (4.422,1.010) -- (4.428,1.033) -- (4.434,1.055) -- (4.440,1.077) -- (4.446,1.099) -- (4.453,1.120) -- (4.459,1.141) -- (4.465,1.162) -- (4.471,1.183) -- (4.477,1.203) -- (4.483,1.223) -- (4.489,1.243) -- (4.495,1.263) -- (4.501,1.282) -- (4.507,1.301) -- (4.513,1.320) -- (4.519,1.338) -- (4.525,1.357) -- (4.531,1.375) -- (4.537,1.393) -- (4.543,1.410) -- (4.549,1.427) -- (4.555,1.444) -- (4.561,1.461) -- (4.567,1.478) -- (4.573,1.494) -- (4.579,1.510) -- (4.585,1.526) -- (4.591,1.541) -- (4.597,1.556) -- (4.603,1.571) -- (4.609,1.586) -- (4.615,1.601) -- (4.621,1.615) -- (4.627,1.629) -- (4.633,1.642) -- (4.639,1.656) -- (4.645,1.669) -- (4.651,1.682) -- (4.657,1.695) -- (4.663,1.707) -- (4.669,1.719) -- (4.675,1.731) -- (4.681,1.743) -- (4.687,1.754) -- (4.693,1.765) -- (4.699,1.776) -- (4.705,1.787) -- (4.711,1.797) -- (4.717,1.807) -- (4.723,1.817) -- (4.729,1.826) -- (4.735,1.836) -- (4.741,1.845) -- (4.747,1.854) -- (4.753,1.862) -- (4.759,1.870) -- (4.765,1.878) -- (4.771,1.886) -- (4.777,1.894) -- (4.783,1.901) -- (4.789,1.908) -- (4.795,1.915) -- (4.801,1.921) -- (4.807,1.928) -- (4.813,1.933) -- (4.819,1.939) -- (4.825,1.945) -- (4.831,1.950) -- (4.837,1.955) -- (4.843,1.960) -- (4.849,1.964) -- (4.855,1.968) -- (4.861,1.972) -- (4.867,1.976) -- (4.873,1.979) -- (4.879,1.982) -- (4.885,1.985) -- (4.891,1.988) -- (4.897,1.990) -- (4.903,1.992) -- (4.909,1.994) -- (4.915,1.996) -- (4.921,1.997) -- (4.927,1.998) -- (4.933,1.999) -- (4.939,2.000) -- (4.945,2.000) -- (4.952,2.000) -- (4.958,2.000) -- (4.964,1.999) -- (4.970,1.999) -- (4.976,1.998) -- (4.982,1.996) -- (4.988,1.995) -- (4.994,1.993) -- (5.000,1.991) -- (5.006,1.989) -- (5.012,1.986) -- (5.018,1.984) -- (5.024,1.981) -- (5.030,1.977) -- (5.036,1.974) -- (5.042,1.970) -- (5.048,1.966) -- (5.054,1.962) -- (5.060,1.957) -- (5.066,1.952) -- (5.072,1.947) -- (5.078,1.942) -- (5.084,1.936) -- (5.090,1.931) -- (5.096,1.924) -- (5.102,1.918) -- (5.108,1.911) -- (5.114,1.905) -- (5.120,1.897) -- (5.126,1.890) -- (5.132,1.882) -- (5.138,1.874) -- (5.144,1.866) -- (5.150,1.858) -- (5.156,1.849) -- (5.162,1.840) -- (5.168,1.831) -- (5.174,1.822) -- (5.180,1.812) -- (5.186,1.802) -- (5.192,1.792) -- (5.198,1.781) -- (5.204,1.771) -- (5.210,1.760) -- (5.216,1.748) -- (5.222,1.737) -- (5.228,1.725) -- (5.234,1.713) -- (5.240,1.701) -- (5.246,1.688) -- (5.252,1.675) -- (5.258,1.662) -- (5.264,1.649) -- (5.270,1.636) -- (5.276,1.622) -- (5.282,1.608) -- (5.288,1.593) -- (5.294,1.579) -- (5.300,1.564) -- (5.306,1.549) -- (5.312,1.533) -- (5.318,1.518) -- (5.324,1.502) -- (5.330,1.486) -- (5.336,1.469) -- (5.342,1.453) -- (5.348,1.436) -- (5.354,1.419) -- (5.360,1.401) -- (5.366,1.384) -- (5.372,1.366) -- (5.378,1.348) -- (5.384,1.329) -- (5.390,1.310) -- (5.396,1.291) -- (5.402,1.272) -- (5.408,1.253) -- (5.414,1.233) -- (5.420,1.213) -- (5.426,1.193) -- (5.432,1.172) -- (5.438,1.152) -- (5.444,1.131) -- (5.451,1.109) -- (5.457,1.088) -- (5.463,1.066) -- (5.469,1.044) -- (5.475,1.022) -- (5.481,0.999) -- (5.487,0.976) -- (5.493,0.953) -- (5.499,0.930) -- (5.505,0.906) -- (5.511,0.883) -- (5.517,0.858) -- (5.523,0.834) -- (5.529,0.809) -- (5.535,0.785) -- (5.541,0.759) -- (5.547,0.734) -- (5.553,0.708) -- (5.559,0.683) -- (5.565,0.656) -- (5.571,0.630) -- (5.577,0.603) -- (5.583,0.576) -- (5.589,0.549) -- (5.595,0.522) -- (5.601,0.494) -- (5.607,0.466) -- (5.613,0.438) -- (5.619,0.409) -- (5.625,0.381) -- (5.631,0.352) -- (5.637,0.323) -- (5.643,0.293) -- (5.649,0.263) -- (5.655,0.233) -- (5.661,0.203) -- (5.667,0.172) -- (5.673,0.142) -- (5.679,0.111) -- (5.685,0.079) -- (5.691,0.048) -- (5.697,0.016) -- (5.703,0.016) -- (5.709,0.048) -- (5.715,0.079) -- (5.721,0.111) -- (5.727,0.142) -- (5.733,0.172) -- (5.739,0.203) -- (5.745,0.233) -- (5.751,0.263) -- (5.757,0.293) -- (5.763,0.323) -- (5.769,0.352) -- (5.775,0.381) -- (5.781,0.409) -- (5.787,0.438) -- (5.793,0.466) -- (5.799,0.494) -- (5.805,0.522) -- (5.811,0.549) -- (5.817,0.576) -- (5.823,0.603) -- (5.829,0.630) -- (5.835,0.656) -- (5.841,0.683) -- (5.847,0.708) -- (5.853,0.734) -- (5.859,0.759) -- (5.865,0.785) -- (5.871,0.809) -- (5.877,0.834) -- (5.883,0.858) -- (5.889,0.883) -- (5.895,0.906) -- (5.901,0.930) -- (5.907,0.953) -- (5.913,0.976) -- (5.919,0.999) -- (5.925,1.022) -- (5.931,1.044) -- (5.937,1.066) -- (5.943,1.088) -- (5.949,1.109) -- (5.956,1.131) -- (5.962,1.152) -- (5.968,1.172) -- (5.974,1.193) -- (5.980,1.213) -- (5.986,1.233) -- (5.992,1.253) -- (5.998,1.272) -- (6.004,1.291) -- (6.010,1.310) -- (6.016,1.329) -- (6.022,1.348) -- (6.028,1.366) -- (6.034,1.384) -- (6.040,1.401) -- (6.046,1.419) -- (6.052,1.436) -- (6.058,1.453) -- (6.064,1.469) -- (6.070,1.486) -- (6.076,1.502) -- (6.082,1.518) -- (6.088,1.533) -- (6.094,1.549) -- (6.100,1.564) -- (6.106,1.579) -- (6.112,1.593) -- (6.118,1.608) -- (6.124,1.622) -- (6.130,1.636) -- (6.136,1.649) -- (6.142,1.662) -- (6.148,1.675) -- (6.154,1.688) -- (6.160,1.701) -- (6.166,1.713) -- (6.172,1.725) -- (6.178,1.737) -- (6.184,1.748) -- (6.190,1.760) -- (6.196,1.771) -- (6.202,1.781) -- (6.208,1.792) -- (6.214,1.802) -- (6.220,1.812) -- (6.226,1.822) -- (6.232,1.831) -- (6.238,1.840) -- (6.244,1.849) -- (6.250,1.858) -- (6.256,1.866) -- (6.262,1.874) -- (6.268,1.882) -- (6.274,1.890) -- (6.280,1.897) -- (6.286,1.905) -- (6.292,1.911) -- (6.298,1.918) -- (6.304,1.924) -- (6.310,1.931) -- (6.316,1.936) -- (6.322,1.942) -- (6.328,1.947) -- (6.334,1.952) -- (6.340,1.957) -- (6.346,1.962) -- (6.352,1.966) -- (6.358,1.970) -- (6.364,1.974) -- (6.370,1.977) -- (6.376,1.981) -- (6.382,1.984) -- (6.388,1.986) -- (6.394,1.989) -- (6.400,1.991) -- (6.406,1.993) -- (6.412,1.995) -- (6.418,1.996) -- (6.424,1.998) -- (6.430,1.999) -- (6.436,1.999) -- (6.442,2.000) -- (6.448,2.000) -- (6.455,2.000) -- (6.461,2.000) -- (6.467,1.999) -- (6.473,1.998) -- (6.479,1.997) -- (6.485,1.996) -- (6.491,1.994) -- (6.497,1.992) -- (6.503,1.990) -- (6.509,1.988) -- (6.515,1.985) -- (6.521,1.982) -- (6.527,1.979) -- (6.533,1.976) -- (6.539,1.972) -- (6.545,1.968) -- (6.551,1.964) -- (6.557,1.960) -- (6.563,1.955) -- (6.569,1.950) -- (6.575,1.945) -- (6.581,1.939) -- (6.587,1.933) -- (6.593,1.928) -- (6.599,1.921) -- (6.605,1.915) -- (6.611,1.908) -- (6.617,1.901) -- (6.623,1.894) -- (6.629,1.886) -- (6.635,1.878) -- (6.641,1.870) -- (6.647,1.862) -- (6.653,1.854) -- (6.659,1.845) -- (6.665,1.836) -- (6.671,1.826) -- (6.677,1.817) -- (6.683,1.807) -- (6.689,1.797) -- (6.695,1.787) -- (6.701,1.776) -- (6.707,1.765) -- (6.713,1.754) -- (6.719,1.743) -- (6.725,1.731) -- (6.731,1.719) -- (6.737,1.707) -- (6.743,1.695) -- (6.749,1.682) -- (6.755,1.669) -- (6.761,1.656) -- (6.767,1.642) -- (6.773,1.629) -- (6.779,1.615) -- (6.785,1.601) -- (6.791,1.586) -- (6.797,1.571) -- (6.803,1.556) -- (6.809,1.541) -- (6.815,1.526) -- (6.821,1.510) -- (6.827,1.494) -- (6.833,1.478) -- (6.839,1.461) -- (6.845,1.444) -- (6.851,1.427) -- (6.857,1.410) -- (6.863,1.393) -- (6.869,1.375) -- (6.875,1.357) -- (6.881,1.338) -- (6.887,1.320) -- (6.893,1.301) -- (6.899,1.282) -- (6.905,1.263) -- (6.911,1.243) -- (6.917,1.223) -- (6.923,1.203) -- (6.929,1.183) -- (6.935,1.162) -- (6.941,1.141) -- (6.947,1.120) -- (6.954,1.099) -- (6.960,1.077) -- (6.966,1.055) -- (6.972,1.033) -- (6.978,1.010) -- (6.984,0.988) -- (6.990,0.965) -- (6.996,0.942) -- (7.002,0.918) -- (7.008,0.894) -- (7.014,0.870) -- (7.020,0.846) -- (7.026,0.822) -- (7.032,0.797) -- (7.038,0.772) -- (7.044,0.747) -- (7.050,0.721) -- (7.056,0.696) -- (7.062,0.669) -- (7.068,0.643) -- (7.074,0.617) -- (7.080,0.590) -- (7.086,0.563) -- (7.092,0.536) -- (7.098,0.508) -- (7.104,0.480) -- (7.110,0.452) -- (7.116,0.424) -- (7.122,0.395) -- (7.128,0.366) -- (7.134,0.337) -- (7.140,0.308) -- (7.146,0.278) -- (7.152,0.248) -- (7.158,0.218) -- (7.164,0.188) -- (7.170,0.157) -- (7.176,0.126) -- (7.182,0.095) -- (7.188,0.064) -- (7.194,0.032) -- (7.200,0.000);
\draw [<->,thick] (8.4,2.3) -- (8.4,0) -- (11.7,0);
\draw [-] (11.4,0.11499999999999999) -- (11.4,-0.11499999999999999) node [below] {\small $1$};
\draw [-] (8.565,2) -- (8.235000000000001,2) node [left] {\small $\frac{1}{64}$};
\draw [dashed,blue,thick] (8.400,0.000) -- (8.406,0.032) -- (8.412,0.064) -- (8.418,0.096) -- (8.424,0.128) -- (8.430,0.160) -- (8.436,0.192) -- (8.442,0.224) -- (8.448,0.257) -- (8.454,0.289) -- (8.460,0.321) -- (8.466,0.353) -- (8.472,0.385) -- (8.478,0.417) -- (8.484,0.449) -- (8.490,0.481) -- (8.496,0.513) -- (8.502,0.545) -- (8.508,0.577) -- (8.514,0.609) -- (8.520,0.641) -- (8.526,0.673) -- (8.532,0.705) -- (8.538,0.737) -- (8.544,0.770) -- (8.550,0.802) -- (8.556,0.834) -- (8.562,0.866) -- (8.568,0.898) -- (8.574,0.930) -- (8.580,0.962) -- (8.586,0.994) -- (8.592,1.026) -- (8.598,1.058) -- (8.604,1.090) -- (8.610,1.122) -- (8.616,1.154) -- (8.622,1.186) -- (8.628,1.218) -- (8.634,1.251) -- (8.640,1.283) -- (8.646,1.315) -- (8.653,1.347) -- (8.659,1.379) -- (8.665,1.411) -- (8.671,1.443) -- (8.677,1.475) -- (8.683,1.507) -- (8.689,1.539) -- (8.695,1.571) -- (8.701,1.603) -- (8.707,1.635) -- (8.713,1.667) -- (8.719,1.699) -- (8.725,1.731) -- (8.731,1.764) -- (8.737,1.796) -- (8.743,1.828) -- (8.749,1.860) -- (8.755,1.892) -- (8.761,1.924) -- (8.767,1.956) -- (8.773,1.988) -- (8.779,1.980) -- (8.785,1.948) -- (8.791,1.916) -- (8.797,1.884) -- (8.803,1.852) -- (8.809,1.820) -- (8.815,1.788) -- (8.821,1.756) -- (8.827,1.723) -- (8.833,1.691) -- (8.839,1.659) -- (8.845,1.627) -- (8.851,1.595) -- (8.857,1.563) -- (8.863,1.531) -- (8.869,1.499) -- (8.875,1.467) -- (8.881,1.435) -- (8.887,1.403) -- (8.893,1.371) -- (8.899,1.339) -- (8.905,1.307) -- (8.911,1.275) -- (8.917,1.242) -- (8.923,1.210) -- (8.929,1.178) -- (8.935,1.146) -- (8.941,1.114) -- (8.947,1.082) -- (8.953,1.050) -- (8.959,1.018) -- (8.965,0.986) -- (8.971,0.954) -- (8.977,0.922) -- (8.983,0.890) -- (8.989,0.858) -- (8.995,0.826) -- (9.001,0.794) -- (9.007,0.762) -- (9.013,0.729) -- (9.019,0.697) -- (9.025,0.665) -- (9.031,0.633) -- (9.037,0.601) -- (9.043,0.569) -- (9.049,0.537) -- (9.055,0.505) -- (9.061,0.473) -- (9.067,0.441) -- (9.073,0.409) -- (9.079,0.377) -- (9.085,0.345) -- (9.091,0.313) -- (9.097,0.281) -- (9.103,0.248) -- (9.109,0.216) -- (9.115,0.184) -- (9.121,0.152) -- (9.127,0.120) -- (9.133,0.088) -- (9.139,0.056) -- (9.145,0.024) -- (9.152,0.008) -- (9.158,0.040) -- (9.164,0.072) -- (9.170,0.104) -- (9.176,0.136) -- (9.182,0.168) -- (9.188,0.200) -- (9.194,0.232) -- (9.200,0.265) -- (9.206,0.297) -- (9.212,0.329) -- (9.218,0.361) -- (9.224,0.393) -- (9.230,0.425) -- (9.236,0.457) -- (9.242,0.489) -- (9.248,0.521) -- (9.254,0.553) -- (9.260,0.585) -- (9.266,0.617) -- (9.272,0.649) -- (9.278,0.681) -- (9.284,0.713) -- (9.290,0.745) -- (9.296,0.778) -- (9.302,0.810) -- (9.308,0.842) -- (9.314,0.874) -- (9.320,0.906) -- (9.326,0.938) -- (9.332,0.970) -- (9.338,1.002) -- (9.344,1.034) -- (9.350,1.066) -- (9.356,1.098) -- (9.362,1.130) -- (9.368,1.162) -- (9.374,1.194) -- (9.380,1.226) -- (9.386,1.259) -- (9.392,1.291) -- (9.398,1.323) -- (9.404,1.355) -- (9.410,1.387) -- (9.416,1.419) -- (9.422,1.451) -- (9.428,1.483) -- (9.434,1.515) -- (9.440,1.547) -- (9.446,1.579) -- (9.452,1.611) -- (9.458,1.643) -- (9.464,1.675) -- (9.470,1.707) -- (9.476,1.739) -- (9.482,1.772) -- (9.488,1.804) -- (9.494,1.836) -- (9.500,1.868) -- (9.506,1.900) -- (9.512,1.932) -- (9.518,1.964) -- (9.524,1.996) -- (9.530,1.972) -- (9.536,1.940) -- (9.542,1.908) -- (9.548,1.876) -- (9.554,1.844) -- (9.560,1.812) -- (9.566,1.780) -- (9.572,1.747) -- (9.578,1.715) -- (9.584,1.683) -- (9.590,1.651) -- (9.596,1.619) -- (9.602,1.587) -- (9.608,1.555) -- (9.614,1.523) -- (9.620,1.491) -- (9.626,1.459) -- (9.632,1.427) -- (9.638,1.395) -- (9.644,1.363) -- (9.651,1.331) -- (9.657,1.299) -- (9.663,1.267) -- (9.669,1.234) -- (9.675,1.202) -- (9.681,1.170) -- (9.687,1.138) -- (9.693,1.106) -- (9.699,1.074) -- (9.705,1.042) -- (9.711,1.010) -- (9.717,0.978) -- (9.723,0.946) -- (9.729,0.914) -- (9.735,0.882) -- (9.741,0.850) -- (9.747,0.818) -- (9.753,0.786) -- (9.759,0.754) -- (9.765,0.721) -- (9.771,0.689) -- (9.777,0.657) -- (9.783,0.625) -- (9.789,0.593) -- (9.795,0.561) -- (9.801,0.529) -- (9.807,0.497) -- (9.813,0.465) -- (9.819,0.433) -- (9.825,0.401) -- (9.831,0.369) -- (9.837,0.337) -- (9.843,0.305) -- (9.849,0.273) -- (9.855,0.240) -- (9.861,0.208) -- (9.867,0.176) -- (9.873,0.144) -- (9.879,0.112) -- (9.885,0.080) -- (9.891,0.048) -- (9.897,0.016) -- (9.903,0.016) -- (9.909,0.048) -- (9.915,0.080) -- (9.921,0.112) -- (9.927,0.144) -- (9.933,0.176) -- (9.939,0.208) -- (9.945,0.240) -- (9.951,0.273) -- (9.957,0.305) -- (9.963,0.337) -- (9.969,0.369) -- (9.975,0.401) -- (9.981,0.433) -- (9.987,0.465) -- (9.993,0.497) -- (9.999,0.529) -- (10.005,0.561) -- (10.011,0.593) -- (10.017,0.625) -- (10.023,0.657) -- (10.029,0.689) -- (10.035,0.721) -- (10.041,0.754) -- (10.047,0.786) -- (10.053,0.818) -- (10.059,0.850) -- (10.065,0.882) -- (10.071,0.914) -- (10.077,0.946) -- (10.083,0.978) -- (10.089,1.010) -- (10.095,1.042) -- (10.101,1.074) -- (10.107,1.106) -- (10.113,1.138) -- (10.119,1.170) -- (10.125,1.202) -- (10.131,1.234) -- (10.137,1.267) -- (10.143,1.299) -- (10.149,1.331) -- (10.156,1.363) -- (10.162,1.395) -- (10.168,1.427) -- (10.174,1.459) -- (10.180,1.491) -- (10.186,1.523) -- (10.192,1.555) -- (10.198,1.587) -- (10.204,1.619) -- (10.210,1.651) -- (10.216,1.683) -- (10.222,1.715) -- (10.228,1.747) -- (10.234,1.780) -- (10.240,1.812) -- (10.246,1.844) -- (10.252,1.876) -- (10.258,1.908) -- (10.264,1.940) -- (10.270,1.972) -- (10.276,1.996) -- (10.282,1.964) -- (10.288,1.932) -- (10.294,1.900) -- (10.300,1.868) -- (10.306,1.836) -- (10.312,1.804) -- (10.318,1.772) -- (10.324,1.739) -- (10.330,1.707) -- (10.336,1.675) -- (10.342,1.643) -- (10.348,1.611) -- (10.354,1.579) -- (10.360,1.547) -- (10.366,1.515) -- (10.372,1.483) -- (10.378,1.451) -- (10.384,1.419) -- (10.390,1.387) -- (10.396,1.355) -- (10.402,1.323) -- (10.408,1.291) -- (10.414,1.259) -- (10.420,1.226) -- (10.426,1.194) -- (10.432,1.162) -- (10.438,1.130) -- (10.444,1.098) -- (10.450,1.066) -- (10.456,1.034) -- (10.462,1.002) -- (10.468,0.970) -- (10.474,0.938) -- (10.480,0.906) -- (10.486,0.874) -- (10.492,0.842) -- (10.498,0.810) -- (10.504,0.778) -- (10.510,0.745) -- (10.516,0.713) -- (10.522,0.681) -- (10.528,0.649) -- (10.534,0.617) -- (10.540,0.585) -- (10.546,0.553) -- (10.552,0.521) -- (10.558,0.489) -- (10.564,0.457) -- (10.570,0.425) -- (10.576,0.393) -- (10.582,0.361) -- (10.588,0.329) -- (10.594,0.297) -- (10.600,0.265) -- (10.606,0.232) -- (10.612,0.200) -- (10.618,0.168) -- (10.624,0.136) -- (10.630,0.104) -- (10.636,0.072) -- (10.642,0.040) -- (10.648,0.008) -- (10.655,0.024) -- (10.661,0.056) -- (10.667,0.088) -- (10.673,0.120) -- (10.679,0.152) -- (10.685,0.184) -- (10.691,0.216) -- (10.697,0.248) -- (10.703,0.281) -- (10.709,0.313) -- (10.715,0.345) -- (10.721,0.377) -- (10.727,0.409) -- (10.733,0.441) -- (10.739,0.473) -- (10.745,0.505) -- (10.751,0.537) -- (10.757,0.569) -- (10.763,0.601) -- (10.769,0.633) -- (10.775,0.665) -- (10.781,0.697) -- (10.787,0.729) -- (10.793,0.762) -- (10.799,0.794) -- (10.805,0.826) -- (10.811,0.858) -- (10.817,0.890) -- (10.823,0.922) -- (10.829,0.954) -- (10.835,0.986) -- (10.841,1.018) -- (10.847,1.050) -- (10.853,1.082) -- (10.859,1.114) -- (10.865,1.146) -- (10.871,1.178) -- (10.877,1.210) -- (10.883,1.242) -- (10.889,1.275) -- (10.895,1.307) -- (10.901,1.339) -- (10.907,1.371) -- (10.913,1.403) -- (10.919,1.435) -- (10.925,1.467) -- (10.931,1.499) -- (10.937,1.531) -- (10.943,1.563) -- (10.949,1.595) -- (10.955,1.627) -- (10.961,1.659) -- (10.967,1.691) -- (10.973,1.723) -- (10.979,1.756) -- (10.985,1.788) -- (10.991,1.820) -- (10.997,1.852) -- (11.003,1.884) -- (11.009,1.916) -- (11.015,1.948) -- (11.021,1.980) -- (11.027,1.988) -- (11.033,1.956) -- (11.039,1.924) -- (11.045,1.892) -- (11.051,1.860) -- (11.057,1.828) -- (11.063,1.796) -- (11.069,1.764) -- (11.075,1.731) -- (11.081,1.699) -- (11.087,1.667) -- (11.093,1.635) -- (11.099,1.603) -- (11.105,1.571) -- (11.111,1.539) -- (11.117,1.507) -- (11.123,1.475) -- (11.129,1.443) -- (11.135,1.411) -- (11.141,1.379) -- (11.147,1.347) -- (11.154,1.315) -- (11.160,1.283) -- (11.166,1.251) -- (11.172,1.218) -- (11.178,1.186) -- (11.184,1.154) -- (11.190,1.122) -- (11.196,1.090) -- (11.202,1.058) -- (11.208,1.026) -- (11.214,0.994) -- (11.220,0.962) -- (11.226,0.930) -- (11.232,0.898) -- (11.238,0.866) -- (11.244,0.834) -- (11.250,0.802) -- (11.256,0.770) -- (11.262,0.737) -- (11.268,0.705) -- (11.274,0.673) -- (11.280,0.641) -- (11.286,0.609) -- (11.292,0.577) -- (11.298,0.545) -- (11.304,0.513) -- (11.310,0.481) -- (11.316,0.449) -- (11.322,0.417) -- (11.328,0.385) -- (11.334,0.353) -- (11.340,0.321) -- (11.346,0.289) -- (11.352,0.257) -- (11.358,0.224) -- (11.364,0.192) -- (11.370,0.160) -- (11.376,0.128) -- (11.382,0.096) -- (11.388,0.064) -- (11.394,0.032) -- (11.400,0.000);
\draw [thick] (8.4,-1.1) -- (9.0,-1.1) node [right] {\small $x-x^2-\frac{h_1(x)}{4}-\frac{h_2(x)}{16}$};
\draw [thick,blue,dashed] (8.4,-1.8) -- (9.0,-1.8) node [right] {\small $\frac{h_3(x)}{64}$};
\draw [thick] (8.400,0.000) -- (8.406,0.064) -- (8.412,0.126) -- (8.418,0.188) -- (8.424,0.248) -- (8.430,0.308) -- (8.436,0.366) -- (8.442,0.424) -- (8.448,0.480) -- (8.454,0.536) -- (8.460,0.590) -- (8.466,0.643) -- (8.472,0.696) -- (8.478,0.747) -- (8.484,0.797) -- (8.490,0.846) -- (8.496,0.894) -- (8.502,0.942) -- (8.508,0.988) -- (8.514,1.033) -- (8.520,1.077) -- (8.526,1.120) -- (8.532,1.162) -- (8.538,1.203) -- (8.544,1.243) -- (8.550,1.282) -- (8.556,1.320) -- (8.562,1.357) -- (8.568,1.393) -- (8.574,1.427) -- (8.580,1.461) -- (8.586,1.494) -- (8.592,1.526) -- (8.598,1.556) -- (8.604,1.586) -- (8.610,1.615) -- (8.616,1.642) -- (8.622,1.669) -- (8.628,1.695) -- (8.634,1.719) -- (8.640,1.743) -- (8.646,1.765) -- (8.653,1.787) -- (8.659,1.807) -- (8.665,1.826) -- (8.671,1.845) -- (8.677,1.862) -- (8.683,1.878) -- (8.689,1.894) -- (8.695,1.908) -- (8.701,1.921) -- (8.707,1.933) -- (8.713,1.945) -- (8.719,1.955) -- (8.725,1.964) -- (8.731,1.972) -- (8.737,1.979) -- (8.743,1.985) -- (8.749,1.990) -- (8.755,1.994) -- (8.761,1.997) -- (8.767,1.999) -- (8.773,2.000) -- (8.779,2.000) -- (8.785,1.999) -- (8.791,1.996) -- (8.797,1.993) -- (8.803,1.989) -- (8.809,1.984) -- (8.815,1.977) -- (8.821,1.970) -- (8.827,1.962) -- (8.833,1.952) -- (8.839,1.942) -- (8.845,1.931) -- (8.851,1.918) -- (8.857,1.905) -- (8.863,1.890) -- (8.869,1.874) -- (8.875,1.858) -- (8.881,1.840) -- (8.887,1.822) -- (8.893,1.802) -- (8.899,1.781) -- (8.905,1.760) -- (8.911,1.737) -- (8.917,1.713) -- (8.923,1.688) -- (8.929,1.662) -- (8.935,1.636) -- (8.941,1.608) -- (8.947,1.579) -- (8.953,1.549) -- (8.959,1.518) -- (8.965,1.486) -- (8.971,1.453) -- (8.977,1.419) -- (8.983,1.384) -- (8.989,1.348) -- (8.995,1.310) -- (9.001,1.272) -- (9.007,1.233) -- (9.013,1.193) -- (9.019,1.152) -- (9.025,1.109) -- (9.031,1.066) -- (9.037,1.022) -- (9.043,0.976) -- (9.049,0.930) -- (9.055,0.883) -- (9.061,0.834) -- (9.067,0.785) -- (9.073,0.734) -- (9.079,0.683) -- (9.085,0.630) -- (9.091,0.576) -- (9.097,0.522) -- (9.103,0.466) -- (9.109,0.409) -- (9.115,0.352) -- (9.121,0.293) -- (9.127,0.233) -- (9.133,0.172) -- (9.139,0.111) -- (9.145,0.048) -- (9.152,0.016) -- (9.158,0.079) -- (9.164,0.142) -- (9.170,0.203) -- (9.176,0.263) -- (9.182,0.323) -- (9.188,0.381) -- (9.194,0.438) -- (9.200,0.494) -- (9.206,0.549) -- (9.212,0.603) -- (9.218,0.656) -- (9.224,0.708) -- (9.230,0.759) -- (9.236,0.809) -- (9.242,0.858) -- (9.248,0.906) -- (9.254,0.953) -- (9.260,0.999) -- (9.266,1.044) -- (9.272,1.088) -- (9.278,1.131) -- (9.284,1.172) -- (9.290,1.213) -- (9.296,1.253) -- (9.302,1.291) -- (9.308,1.329) -- (9.314,1.366) -- (9.320,1.401) -- (9.326,1.436) -- (9.332,1.469) -- (9.338,1.502) -- (9.344,1.533) -- (9.350,1.564) -- (9.356,1.593) -- (9.362,1.622) -- (9.368,1.649) -- (9.374,1.675) -- (9.380,1.701) -- (9.386,1.725) -- (9.392,1.748) -- (9.398,1.771) -- (9.404,1.792) -- (9.410,1.812) -- (9.416,1.831) -- (9.422,1.849) -- (9.428,1.866) -- (9.434,1.882) -- (9.440,1.897) -- (9.446,1.911) -- (9.452,1.924) -- (9.458,1.936) -- (9.464,1.947) -- (9.470,1.957) -- (9.476,1.966) -- (9.482,1.974) -- (9.488,1.981) -- (9.494,1.986) -- (9.500,1.991) -- (9.506,1.995) -- (9.512,1.998) -- (9.518,1.999) -- (9.524,2.000) -- (9.530,2.000) -- (9.536,1.998) -- (9.542,1.996) -- (9.548,1.992) -- (9.554,1.988) -- (9.560,1.982) -- (9.566,1.976) -- (9.572,1.968) -- (9.578,1.960) -- (9.584,1.950) -- (9.590,1.939) -- (9.596,1.928) -- (9.602,1.915) -- (9.608,1.901) -- (9.614,1.886) -- (9.620,1.870) -- (9.626,1.854) -- (9.632,1.836) -- (9.638,1.817) -- (9.644,1.797) -- (9.651,1.776) -- (9.657,1.754) -- (9.663,1.731) -- (9.669,1.707) -- (9.675,1.682) -- (9.681,1.656) -- (9.687,1.629) -- (9.693,1.601) -- (9.699,1.571) -- (9.705,1.541) -- (9.711,1.510) -- (9.717,1.478) -- (9.723,1.444) -- (9.729,1.410) -- (9.735,1.375) -- (9.741,1.338) -- (9.747,1.301) -- (9.753,1.263) -- (9.759,1.223) -- (9.765,1.183) -- (9.771,1.141) -- (9.777,1.099) -- (9.783,1.055) -- (9.789,1.010) -- (9.795,0.965) -- (9.801,0.918) -- (9.807,0.870) -- (9.813,0.822) -- (9.819,0.772) -- (9.825,0.721) -- (9.831,0.669) -- (9.837,0.617) -- (9.843,0.563) -- (9.849,0.508) -- (9.855,0.452) -- (9.861,0.395) -- (9.867,0.337) -- (9.873,0.278) -- (9.879,0.218) -- (9.885,0.157) -- (9.891,0.095) -- (9.897,0.032) -- (9.903,0.032) -- (9.909,0.095) -- (9.915,0.157) -- (9.921,0.218) -- (9.927,0.278) -- (9.933,0.337) -- (9.939,0.395) -- (9.945,0.452) -- (9.951,0.508) -- (9.957,0.563) -- (9.963,0.617) -- (9.969,0.669) -- (9.975,0.721) -- (9.981,0.772) -- (9.987,0.822) -- (9.993,0.870) -- (9.999,0.918) -- (10.005,0.965) -- (10.011,1.010) -- (10.017,1.055) -- (10.023,1.099) -- (10.029,1.141) -- (10.035,1.183) -- (10.041,1.223) -- (10.047,1.263) -- (10.053,1.301) -- (10.059,1.338) -- (10.065,1.375) -- (10.071,1.410) -- (10.077,1.444) -- (10.083,1.478) -- (10.089,1.510) -- (10.095,1.541) -- (10.101,1.571) -- (10.107,1.601) -- (10.113,1.629) -- (10.119,1.656) -- (10.125,1.682) -- (10.131,1.707) -- (10.137,1.731) -- (10.143,1.754) -- (10.149,1.776) -- (10.156,1.797) -- (10.162,1.817) -- (10.168,1.836) -- (10.174,1.854) -- (10.180,1.870) -- (10.186,1.886) -- (10.192,1.901) -- (10.198,1.915) -- (10.204,1.928) -- (10.210,1.939) -- (10.216,1.950) -- (10.222,1.960) -- (10.228,1.968) -- (10.234,1.976) -- (10.240,1.982) -- (10.246,1.988) -- (10.252,1.992) -- (10.258,1.996) -- (10.264,1.998) -- (10.270,2.000) -- (10.276,2.000) -- (10.282,1.999) -- (10.288,1.998) -- (10.294,1.995) -- (10.300,1.991) -- (10.306,1.986) -- (10.312,1.981) -- (10.318,1.974) -- (10.324,1.966) -- (10.330,1.957) -- (10.336,1.947) -- (10.342,1.936) -- (10.348,1.924) -- (10.354,1.911) -- (10.360,1.897) -- (10.366,1.882) -- (10.372,1.866) -- (10.378,1.849) -- (10.384,1.831) -- (10.390,1.812) -- (10.396,1.792) -- (10.402,1.771) -- (10.408,1.748) -- (10.414,1.725) -- (10.420,1.701) -- (10.426,1.675) -- (10.432,1.649) -- (10.438,1.622) -- (10.444,1.593) -- (10.450,1.564) -- (10.456,1.533) -- (10.462,1.502) -- (10.468,1.469) -- (10.474,1.436) -- (10.480,1.401) -- (10.486,1.366) -- (10.492,1.329) -- (10.498,1.291) -- (10.504,1.253) -- (10.510,1.213) -- (10.516,1.172) -- (10.522,1.131) -- (10.528,1.088) -- (10.534,1.044) -- (10.540,0.999) -- (10.546,0.953) -- (10.552,0.906) -- (10.558,0.858) -- (10.564,0.809) -- (10.570,0.759) -- (10.576,0.708) -- (10.582,0.656) -- (10.588,0.603) -- (10.594,0.549) -- (10.600,0.494) -- (10.606,0.438) -- (10.612,0.381) -- (10.618,0.323) -- (10.624,0.263) -- (10.630,0.203) -- (10.636,0.142) -- (10.642,0.079) -- (10.648,0.016) -- (10.655,0.048) -- (10.661,0.111) -- (10.667,0.172) -- (10.673,0.233) -- (10.679,0.293) -- (10.685,0.352) -- (10.691,0.409) -- (10.697,0.466) -- (10.703,0.522) -- (10.709,0.576) -- (10.715,0.630) -- (10.721,0.683) -- (10.727,0.734) -- (10.733,0.785) -- (10.739,0.834) -- (10.745,0.883) -- (10.751,0.930) -- (10.757,0.976) -- (10.763,1.022) -- (10.769,1.066) -- (10.775,1.109) -- (10.781,1.152) -- (10.787,1.193) -- (10.793,1.233) -- (10.799,1.272) -- (10.805,1.310) -- (10.811,1.348) -- (10.817,1.384) -- (10.823,1.419) -- (10.829,1.453) -- (10.835,1.486) -- (10.841,1.518) -- (10.847,1.549) -- (10.853,1.579) -- (10.859,1.608) -- (10.865,1.636) -- (10.871,1.662) -- (10.877,1.688) -- (10.883,1.713) -- (10.889,1.737) -- (10.895,1.760) -- (10.901,1.781) -- (10.907,1.802) -- (10.913,1.822) -- (10.919,1.840) -- (10.925,1.858) -- (10.931,1.874) -- (10.937,1.890) -- (10.943,1.905) -- (10.949,1.918) -- (10.955,1.931) -- (10.961,1.942) -- (10.967,1.952) -- (10.973,1.962) -- (10.979,1.970) -- (10.985,1.977) -- (10.991,1.984) -- (10.997,1.989) -- (11.003,1.993) -- (11.009,1.996) -- (11.015,1.999) -- (11.021,2.000) -- (11.027,2.000) -- (11.033,1.999) -- (11.039,1.997) -- (11.045,1.994) -- (11.051,1.990) -- (11.057,1.985) -- (11.063,1.979) -- (11.069,1.972) -- (11.075,1.964) -- (11.081,1.955) -- (11.087,1.945) -- (11.093,1.933) -- (11.099,1.921) -- (11.105,1.908) -- (11.111,1.894) -- (11.117,1.878) -- (11.123,1.862) -- (11.129,1.845) -- (11.135,1.826) -- (11.141,1.807) -- (11.147,1.787) -- (11.154,1.765) -- (11.160,1.743) -- (11.166,1.719) -- (11.172,1.695) -- (11.178,1.669) -- (11.184,1.642) -- (11.190,1.615) -- (11.196,1.586) -- (11.202,1.556) -- (11.208,1.526) -- (11.214,1.494) -- (11.220,1.461) -- (11.226,1.427) -- (11.232,1.393) -- (11.238,1.357) -- (11.244,1.320) -- (11.250,1.282) -- (11.256,1.243) -- (11.262,1.203) -- (11.268,1.162) -- (11.274,1.120) -- (11.280,1.077) -- (11.286,1.033) -- (11.292,0.988) -- (11.298,0.942) -- (11.304,0.894) -- (11.310,0.846) -- (11.316,0.797) -- (11.322,0.747) -- (11.328,0.696) -- (11.334,0.643) -- (11.340,0.590) -- (11.346,0.536) -- (11.352,0.480) -- (11.358,0.424) -- (11.364,0.366) -- (11.370,0.308) -- (11.376,0.248) -- (11.382,0.188) -- (11.388,0.126) -- (11.394,0.064) -- (11.400,0.000);
\end{tikzpicture}

%% file: plots/h1sn.tex
\begin{tikzpicture}[scale=0.95]

  \draw (-4.1,0) -- (-3.3,0.4);
  \draw (-4.1,0) -- (-3.3,-0.4);
  \draw (-3.3,0.4) -- (-2.5,0);
  \draw (-3.3,-0.4) -- (-2.5,0);
  
  \fill [black] (-4.1,0) circle (1.7pt) node [left] {\footnotesize $x$};
  
  \fill [black] (-3.3,0.4) circle (1.7pt);
  \fill [black] (-3.3,-0.4) circle (1.7pt);

  \fill [black] (-2.5,0) circle (1.7pt) node [right] {\footnotesize $h_1(x)$};

  \node at (3+1,0) {$\dots$};
  
  \fill [black!20,rounded corners] (0.85,-0.85) rectangle (1.15,0.15);
  \fill [black!20,rounded corners] (1.85,-0.85) rectangle (2.15,0.15);
  \fill [black!20,rounded corners] (2.85,-0.85) rectangle (3.15,0.15);
  \fill [black!20,rounded corners] (3.85+1,-0.85) rectangle (4.15+1,0.15);

  \fill [black!20,rounded corners] (0.85,0.55) rectangle (1.15,0.85);
  \fill [black!20,rounded corners] (1.85,0.55) rectangle (2.15,0.85);
  \fill [black!20,rounded corners] (2.85,0.55) rectangle (3.15,0.85);
  \fill [black!20,rounded corners] (3.85+1,0.55) rectangle (4.15+1,0.85);

  \foreach \x in {-0.7,0,0.7}
  {\draw (0,0) -- (1,\x);}

  \fill [black] (0,0) circle (1.7pt) node [left] {\footnotesize $x$};

  \foreach \x in {-0.7,0}
  {\foreach \y in {-0.7,0,0.7}
    {\draw (1,\x) -- (2,\y);}}
  \draw (1,0.7) -- (2,0.7);

  \fill [black] (1,0.7) circle (1.7pt) node [yshift=0.5cm] {\footnotesize $x$};
  \fill [black] (1,0) circle (1.7pt);
  \fill [black] (1,-0.7) circle (1.7pt) node [yshift=-0.5cm] {\footnotesize $h_1(x)$};

  \foreach \x in {-0.7,0}
  {\foreach \y in {-0.7,0,0.7}
    {\draw (2,\x) -- (3,\y);}}
  \draw (2,0.7) -- (3,0.7);

  \fill [black] (2,0.7) circle (1.7pt) node [yshift=0.5cm] {\footnotesize $s_1(x)$};
  \fill [black] (2,0) circle (1.7pt);
  \fill [black] (2,-0.7) circle (1.7pt) node [yshift=-0.5cm] {\footnotesize $h_2(x)$};

  \fill [black] (3,0.7) circle (1.7pt) node [yshift=0.5cm] {\footnotesize $s_2(x)$};
  \fill [black] (3,0) circle (1.7pt);
  \fill [black] (3,-0.7) circle (1.7pt) node [yshift=-0.5cm] {\footnotesize $h_3(x)$};  

  \foreach \x in {-0.7,0,0.7}
  {\draw (4+1,\x) -- (5+1,0);}

  \fill [black] (4+1,0.7) circle (1.7pt) node [yshift=0.5cm] {\footnotesize $s_{n-1}(x)$};
  \fill [black] (4+1,0) circle (1.7pt);
  \fill [black] (4+1,-0.7) circle (1.7pt) node [yshift=-0.5cm] {\footnotesize $h_n(x)$};

  \fill [black] (5+1,0) circle (1.7pt) node [right] {\footnotesize $s_n(x)$};
  
\end{tikzpicture}

%% file: HighDimApprox.tex
\chapter{High-dimensional approximation}\label{chap:hdapp}
In the previous chapters,
we established convergence rates for the uniform approximation of a
function $f:[0,1]^d\to\R$ by a neural network.
For example, %
Theorem \ref{thm:Cks}
provides the error bound
$\mathcal{O}(N^{-k/d})$
in terms of the network size $N$ (up to logarithmic terms),
if $f\in C^k([0,1]^d)$.
Achieving an accuracy of $\eps>0$, therefore, necessitates 
a network size $N=O(\eps^{-d/k})$ (according to this bound).
Hence, for a fixed $\eps>0$, the size of the network needs to increase exponentially
in $d$.
This exponential dependence on the dimension $d$ is
referred to as the \textbf{Curse of Dimensionality} (CoD) \cite{bellman1952theory}. 
In this chapter, we give more details on the CoD, explain why it occurs, and discuss a few scenarios under which it can be mitigated.

Section~\ref{sec:cod} recalls classical results on nonlinear approximation from \cite{devore1998nonlinear, novak2009approximation}, showing that for smoothness spaces such as $C^k([0,1]^d)$, any neural network architecture necessarily suffers from the CoD under mild assumptions. This suggests that such function spaces are often too general in high-dimensional settings.
Motivated by this, the subsequent sections introduce narrower function classes, that are better suited for efficient approximation in high dimensions. While it is easy to construct function classes that do not suffer from the CoD, identifying meaningful and practically relevant ones is more difficult.

In Section \ref{sec:BarronClass} we examine an assumption
limiting the behavior of functions in their Fourier domain.
This assumption allows for slow but dimension independent approximation rates. Next, in Section \ref{sec:compositionalityApprox} we
consider functions with a specific compositional structure.
Concretely, these functions are constructed by compositions and linear combinations of simple low-dimensional subfunctions.
In this case, the curse of dimension is present but only through the input dimension of the subfunctions.
Finally, in Section \ref{sec:ManifoldAssumption} we study the situation where the functions are still defined on high-dimensional spaces, but %
the approximation accuracy is only measured on a lower dimensional submanifold of the high-dimensional input space.
Here, the approximation rate is governed by the smoothness and the dimension of the manifold.

\section{The curse of dimensionality}\label{sec:cod}
Consider a function $f: [0,1]^{d} \to \mathbb{R}^{m}$, where both the input dimension $d$ and output dimension $m$ may be large. Approximating $f$ amounts to approximating each of its $m$ component functions $f_i: [0,1]^d \to \mathbb{R}$ for $i = 1, \dots, m$. Consequently, algorithms typically scale linearly in $m$ with respect to computational cost, model complexity, and data requirements. For this reason, the output dimension $m$ being large is often not considered a major bottleneck. The input dimension $d$ plays a much more critical role.

In what follows, we thus focus on the set of functions $f:[0,1]^d\to\R$, belonging to the unit ball in $C^k$, i.e.\
\begin{equation}\label{eq:B1kd}
  B_1^{k,d}\dfn \set{f\in C^k([0,1]^d)}{\norm[{C^k([0,1]^d)}]{f}\le 1}
\end{equation}
for some $k\in\N\cup\{\infty\}$, cf.~Definition \ref{def:ckSpace}. More precisely, we will focus on uniform approximation, i.e.\ with respect to the norm
\begin{equation}\label{eq:inftynorm}
  \norm[{\infty}]{f}\dfn \sup_{\Bx\in [0,1]^d}|f(\Bx)|,
\end{equation}
where $d$ will always be clear from context.\footnote{For continuous $f:[0,1]^d\to\R$ the norms $\norm[\infty]{f}$ and $\norm[{C^0([0,1]^d)}]{f}$ are the same. We explicitly use $\norm[\infty]{\cdot}$ here, since approximations to $f$ will not be required to be continuous.}
Our goal is to give lower bounds on the complexity of how a neural network capable of approximating all functions in $B_1^{k,d}$ must scale in $d$.

\subsection{Data requirements}\label{sec:datareq}
Consider an algorithm $A$, which takes the $N$ function values $f(\Bx_1),\dots,f(\Bx_N)$, and tries to recreate $f\in B_1^{k,d}\subset C^k([0,1]^d)$ from it. Mathematically, the term \emph{algorithm} refers in the following simply to a mapping $A:\R^N\to %
\set{g}{g:[0,1]^d\to\R}$, meaning $A$ takes a vector in $\R^N$ and returns a function on $[0,1]^d$. In the context of %
deep learning, $A$ could for example %
map the data points $f(\Bx_1),\dots,f(\Bx_N)$
to the function realized by the neural network
trained on these data points for function regression.
In this subsection, our goal is to show the following theorem.

\begin{theorem}\label{thm:coddata}
  Let $k\in\N$. There exists $C_k>0$, such that for all $d$, $N\in\N$, all $\Bx_1,\dots,\Bx_N\in [0,1]^d$, and all maps $A:\R^N\to \set{g}{g:[0,1]^d\to\R}$
  \begin{equation*}
    \sup_{f\in B_1^{k,d}}\norm[{\infty}]{f-A(f(\Bx_1),\dots,f(\Bx_N))}\ge C_k N^{-k/d}.
  \end{equation*}
\end{theorem}

The theorem states in particular that, \emph{assuming our only prior knowledge on $f$ to be that $f\in B_1^{k,d}$}, the number of evaluation points $\Bx_1,\dots,\Bx_N$ required for the existence of an algorithm that reliably reconstructs $f$ up to a certain accuracy necessarily grows exponentially with $d$.

The underlying issue is that, as the dimension $d$ increases, the domain $[0,1]^d$ can accommodate exponentially many disjoint small regions; this allows functions to localize in increasingly many ways. As a result, the richness of the function class $B_1^{k,d}$ effectively grows exponentially with $d$. To make this precise, we will next construct an explicit sequence of such localized functions. This then %
directly leads to the lower bound in Theorem~\ref{thm:coddata}.

For $\Bx\in\R^d$, introduce the {\bf bump function}
\begin{equation}\label{eq:bumpfunction}
  \psi(\Bx)\dfn \begin{cases}
                  \exp\Big(1-\frac{1}{1-\norm{\Bx}^2}\Big) &\norm{\Bx}\le 1\\
                  0 &\text{otherwise}.
                  \end{cases}
\end{equation}

\begin{lemma}\label{lemma:bumpfunction}
  Let $d\in\N$ and set $\psi_{n}\dfn \psi(n\Bx)$ for all $n\in\N$, $\Bx\in\R^d$. Then $\psi_{n}\in C^\infty(\R^d)$, $\norm[C^0(\R^d)]{\psi_n}=1$ and
  \begin{equation*}
    \supp(\psi_{n})\subseteq \Big[-\frac{1}{n},\frac{1}{n}\Big]^d.
  \end{equation*}
  Moreover, for every $k\in\N$
  there exist $0<C_{k,1}\le C_{k,2}$ independent of $d$ such that
  \begin{equation*}
    n^{k} C_{k,1} \le \norm[C^k(\R^d)]{\psi_{n}}\le n^{k} C_{k,2}\qquad\text{for all }n\in\N.
  \end{equation*}
\end{lemma}
\begin{proof}
  It is easy to check that $\psi\in C^\infty(\R^d)$ with support on $\set{\Bx\in \R^d}{\norm{\Bx}\le 1}\subseteq [-1,1]^d$. Thus $\psi_{n}\in C^\infty(\R^d)$ with support on $[-1/n,1/n]^d$.
  Moreover, by the chain rule
  \begin{equation*}
    \norm[{C^k(\R^d)}]{\psi_{n}} = \sup_{\Bx\in\R^d}\max_{\substack{\Balpha\in\N_0^d\\|\Balpha|\le k}} |D^\Balpha \psi_{n}(\Bx)|
    = \max_{\substack{\Balpha\in\N_0^d\\|\Balpha|\le k}} n^{|\Balpha|} \underbrace{\sup_{\Bx\in\R^d} |D^\Balpha \psi(\Bx)|}_{\dfnn C_{d,\Balpha}}.
  \end{equation*}
  Note that $C_{d,\Balpha}>0$ for each $\Balpha\in\N_0^d$ since $\psi$ is not a polynomial.

  Fix $r\ge k$, a multiindex $\Balpha\in\N_0^r$ with $|\Balpha|=k$ and $j\le k$ nonzero entries. Moreover let $\Bbeta\in\N_0^r$ be a permutation of the entries of $\Balpha$ with its nonzero entries in the first $j$ positions.  Due to the symmetry of $\psi$, we have
  \begin{equation*}
    C_{r,\Balpha} =
    \sup_{\Bx\in\R^r} |D^\Balpha \psi(\Bx)|
    =
    \sup_{\Bx\in\R^r} |D^\Bbeta \psi(\Bx)|
    =
    \sup_{\Bx\in\R^r} \Big|D^\Bbeta \exp\Big(1-\frac{1}{1-\norm[]{\Bx}^2}\Big)\Big|.
  \end{equation*}
  From the last expression we see that the supremum will be achieved when $x_{j+1}=\dots=x_r=0$. Hence with $\tilde\Bbeta=(\beta_1,\dots,\beta_k)\in\N_0^k$
  \begin{equation*}
    C_{r,\Balpha}=\sup_{\Bx\in\R^r} \Big|D^\Bbeta \exp\Big(1-\frac{1}{1-\norm[]{\Bx}^2}\Big)\Big|
    =
    \sup_{\Bx\in\R^k} \Big|D^{\tilde\Bbeta} \exp\Big(1-\frac{1}{1-\norm[]{\Bx}^2}\Big)\Big|=C_{k,\tilde\Bbeta}.
  \end{equation*}
  Thus
  \begin{equation*}
    \sup_{r\ge k} \max_{\substack{\Balpha\in\N_0^r\\|\Balpha|=k}}C_{r,\Balpha}
    =\max_{\substack{\tilde\Bbeta\in\N_0^k\\|\tilde\Bbeta|=k}}C_{k,\tilde\Bbeta}.
  \end{equation*}
  Finally, for all $d$, $n\in\N$
  \begin{equation*}
    0<C_{k,1}\dfn
    \min_{r=1,\dots,k}
    \max_{\substack{\Balpha\in\N_0^r\\|\Balpha|=k}}C_{r,\Balpha}\le n^{-k}\norm[{C^k(\R^d)}]{\psi_{n}}
    \le
    \max_{r=1,\dots,k}
    \max_{|\Balpha|\le k} C_{r,\Balpha}\dfnn C_{k,2}<\infty,
  \end{equation*}  
  which gives the claim.
\end{proof}

The next lemma is an immediate consequence of the above, and provides the announced sequence of functions.

\begin{lemma}\label{lemma:psiBnun}
  For $\Bnu\in \{1,\dots,n\}^d$ let
  \begin{equation*}
    \psi_{\Bnu,n}(\Bx)\dfn \frac{1}{C_{2,k}(3n)^k}\psi_{3n}\Big(\Bx-\frac{\Bnu}{n+1}\Big).
  \end{equation*}
  Then $\supp(\psi_{\Bnu,n})\cap\supp(\psi_{\Bmu,n})=\emptyset$
  whenever $\Bmu\neq\Bnu$, and for all $\Bnu\in \{1,\dots,n\}^d$
  \begin{equation*}
    \norm[{C^0([0,1]^d)}]{\psi_{\Bnu,n}}=\frac{1}{C_{2,k}(3n)^{k}},\qquad
    \norm[{C^k([0,1]^d)}]{\psi_{\Bnu,n}}\le 1.
  \end{equation*}
\end{lemma}

\begin{proof}[of Theorem \ref{thm:coddata}]
  Fix $d$, $k\in\N$ and let $n\in\N$ be such that $n^d\le N < (n+1)^d$. The $(n+1)^d$ functions $\psi_{\Bnu,n+1}$, $\Bnu\in\{1,\dots,n+1\}^d$ have disjoint support. Thus, there exists at least one $\Bnu$ with
  \begin{equation*}
    \Bx_j\notin\supp\psi_{\Bnu,n+1}\qquad\text{for all }j=1,\dots,N,
  \end{equation*}
  and we set $f\dfn \psi_{\Bnu,n+1}\in B_1^{k,d}$. Note that
  also $-f\in B_1^{k,d}$, and
  \begin{equation*}
    g\dfn A(f(\Bx_1),\dots,f(\Bx_N))=
    A(-f(\Bx_1),\dots,-f(\Bx_N)).
  \end{equation*}
  It holds
  \begin{equation*}
    2 C_{2,k}(3(n+1))^{-k} = 2\norm[\infty]{f}\le \norm[\infty]{f-g}+\norm[\infty]{f+g}
  \end{equation*}
  and thus either
  \begin{equation*}
    \norm[\infty]{f-g}\ge (C_{2,k}(3(n+1))^{k})^{-1}\quad\text{or}\quad
    \norm[\infty]{-f-g}\ge (C_{2,k}(3(n+1))^{k})^{-1}.
  \end{equation*}
  Using $n-1 \le N^{1/d}$, we have
  \begin{equation*}
    (C_{2,k}(3(n+1))^{k})^{-1}\ge C_{2,k}^{-1} 2^{-k}(3n)^{-k}
    \ge C_{2,k}^{-1} 6^{-k}N^{-k/d}    
  \end{equation*}
  which proves the claim.
\end{proof}

\subsection{Number of parameters}\label{sec:codparam}
We now follow \cite{devore1989},
and consider approximations of the form
\begin{equation*}
  R\circ W (f)\qquad\text{for }f\in B_1^{k,d},
\end{equation*}
where $W:C^0([0,1]^d)\to\R^N$ is continuous\footnote{Throughout Sections \ref{sec:codparam} and \ref{sec:codparaminfty}, it would suffice
  to assume $W:B_1^{k,d}\to\R^N$ to be continuous w.r.t.\ the topology of $C^0([0,1]^d)$ on $B_1^{k,d}$.}, and $R:\R^N\to \set{g}{g:[0,1]^d\to\R}$
is an arbitrary map. We interpret these functions as follows:
\begin{itemize}
\item the {\bf weight map} $W$ assigns $N$ parameters (e.g., network weights) to a given target function $f$,
\item the {\bf realization map} $R$ takes an $N$-dimensional parameter $W\in\R^N$ and returns a function $g:[0,1]^d\to\R$ (e.g., the function realized by a neural network with weights $W$). If $R$ is a smooth mapping into some function space, its image can be interpreted as an $N$-dimensional submanifold of this function space.
\end{itemize}

Define the {\bf continuous nonlinear $N$-width} of $B_1^{k,d}$
\begin{equation}\label{eq:nonlnwidth}
  \gamma_{k,d,N}\dfn \inf_{\substack{W:C^0([0,1]^d)\to \R^N \text{ continuous}\\R:\R^N\to \set{g}{g:[0,1]^d\to\R}}}~
  \sup_{f\in B_1^{k,d}}~\norm[{\infty}]{f-R(W(f))}.
\end{equation}
In this section,  we prove the following lower bound on $\gamma_{k,d,N}$, \cite[Theorem 4.2]{devore1989}.

\begin{theorem}\label{thm:codparam}
  Let $k\in\N$. There exists $C_k>0$, such that
  for all $d$, $N\in\N$
  \begin{equation*}%
    \gamma_{k,d,N}
    \ge C_k N^{-k/d}.
  \end{equation*}
\end{theorem}

\begin{remark}
  Theorem \ref{thm:coddata} is a special case of Theorem \ref{thm:codparam}, with $A=R$ and
  $W(f)=(f(\Bx_1),\dots,f(\Bx_N))$.
\end{remark}

Theorem \ref{thm:codparam} answers the following question: When approximating
all functions in $B_1^{k,d}$ uniformly by a function class continuously parametrized by $N\in\N$ real numbers, 
how small can the worst-case
error be? An immediate consequence is the next corollary which gives a
lower bound of the required network size for \emph{any neural
  network architecture} (regardless of activation function, width, depth,
sparsity pattern etc.).

\begin{corollary}\label{cor:codnn}
  For every $N\in\N$ let $\Phi_N(\cdot,\Bw)$ be a feedforward neural
  network with input dimension $d$ and parameters $\Bw\in\R^N$, i.e.\
  \begin{equation*}
    \size(\Phi_N)=N,
  \end{equation*}
  (cf.~Definitions \ref{def:nn}, \ref{def:SizeOfNNDefinition}).
  Then there exists $C_k$ such that for all $d$, $N\in\N$,
  and any continuous weight selection $W:C^0([0,1]^d)\to\R^{N}$
  \begin{equation*}
    \sup_{f\in B_1^{k,d}}~\norm[{\infty}]{f-\Phi(\cdot,W(f))}\ge C_k N^{-k/d}.
  \end{equation*}
\end{corollary}

The proof of Corollary \ref{cor:codnn} simply consists of the application of Theorem \ref{thm:codparam} to the map $R:\Bw\mapsto \Phi_N(\cdot,\Bw)$. The statement shows again a curse of dimensionality: for
any continuous weight selection process, and any neural network
architecture of size at most $N\in\N$, the best achievable
uniform approximation is lower bounded by $C_k N^{-k/d}$. In
particular, to achieve a fixed accuracy $\eps$, it must hold
\begin{equation}\label{eq:Ckexponential}
  N\ge \Big(\frac{C_k}{\eps}\Big)^{\frac{d}{k}},
\end{equation}
implying exponential increase of the number of parameters in $d$.

\begin{remark}\label{rmk:ReLUessopt1}
  As we have seen in Theorem \ref{thm:Cks} (up to log factors) the convergence rate $N^{-k/d}$ is indeed achievable by feedforward ReLU neural networks. Moreover, the specific weights constructed in the proof of Theorem \ref{thm:Cks} correspond to a continuous weight selection. In this (specific) sense, ReLU neural networks are therefore essentially optimal.
\end{remark}

\begin{figure}
  \begin{center}
    \includegraphics[width=\textwidth]{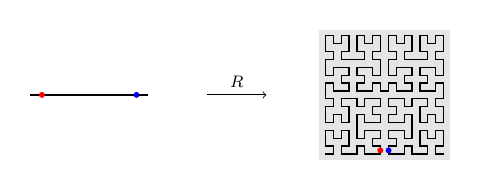}
  \end{center}
  \caption{(Near) space filling curves $R:w\mapsto R(w)$ can be used to approximate elements of a high dimensional space arbitrarily well by a map $R$ depending on a low dimensional parameter $w$. However, $R(w_1)$ and $R(w_2)$ being close, in general does not imply that the parameters $w_1$ and $w_2$ are close.}\label{fig:spacefilling}
\end{figure}

The continuity of $W$ is a crucial restriction to
make the above statements meaningful. In the case of neural networks,
it implies that if $\norm[C^0{([0,1]^d)}]{f-\tilde f}$ is small, then
also the corresponding network weights must be close. This is a
reasonable assumption: if $W$ is not continuous, then the weight
selection process is highly unstable, and therefore likely not
practically computable.  If we would allow for arbitrary maps
$W:B_1^{k,d}\to\R^N$, even under the additional assumption of
continuity of $R$, the infimum in \eqref{eq:nonlnwidth} becomes
zero. This is due to the existence of space-filling curves and similar
pathologies, see Figure \ref{fig:spacefilling}. In fact we have
already encountered an example of this in Chapter \ref{chap:UA},
specifically Proposition \ref{prop:magic}: There we saw that for some 
particular activation function, all functions in $C^0([0,1])$ can be arbitrarily
well approximated by a neural network using only a single
parameter. We encourage the reader to revisit the proof and
verify that the weight selection map is in this case (clearly)
discontinuous.

We now come to the proof of Theorem \ref{thm:codparam}. A key component of the argument is the Borsuk-Ulam theorem \cite{Borsuk1933,MR801938}, which we recall next.

\begin{theorem}[Borsuk-Ulam]
  Let $N\in\N$, $N\ge 2$, and denote the $(N-1)$-sphere by $S^{N-1}\subseteq\R^{N}$. Let $f:S^{N-1}\to\R^N$ be continuous and odd, i.e.\ $f(-\Bx)=-f(\Bx)$ for all $\Bx\in S^{N-1}$. Then there exists at least one $\Bx\in S^{N-1}$ such that $f(\Bx)=\Bnul\in\R^N$.
\end{theorem}

The next lemma corresponds to \cite[Theorem 3.1]{devore1989}. Although the $\gamma_{k,d,N}$ in \eqref{eq:nonlnwidth} seem rather intricate, the lemma gives a surprisingly simple argument to derive lower bounds on these quantities.

\begin{lemma}\label{lemma:borsukulam}
  Let $k\in\N\cup\{\infty\}$, let $X_N$ be an $N$-dimensional subspace of $C^k([0,1]^d)$ for some $N\ge 2$, and assume that for some $\rho>0$ %
  \begin{equation*}
    \set{f\in X_N}{\norm[{\infty}]{f}=\rho}\subseteq B_1^{k,d}.
  \end{equation*}
  Then $\gamma_{k,d,N}\ge \rho$.
\end{lemma}
\begin{proof}
  Fix $W:C^0([0,1]^d)\to\R^N$ continuous, and
  $R:\R^N\to \set{g}{g:[0,1]^d\to\R}$ arbitrary.

  Letting
  \begin{equation*}
    \tilde W(f)\dfn W(f)-W(-f),
  \end{equation*}
  the map $\tilde W:C^0([0,1]^d)\to\R^N$ is continuous and odd.
  Now fix a basis $\psi_1,\dots,\psi_N$ of $X_N$ and set for $(x_1,\dots,x_{N})\in S^{N-1}\subseteq\R^{N}$
  \begin{equation*}
    \phi(x_1,\dots,x_N)\dfn \rho \frac{\sum_{j=1}^N x_j \psi_j}{\norm[C^0({[0,1]^d})]{\sum_{j=1}^N x_j \psi_j}}.
  \end{equation*}
  It is easily verified that $\phi:S^{N-1}\to
  \set{f\in X_N}{\norm[{C^0([0,1]^d)}]{f}=\rho}
  \subseteq
  C^0([0,1]^d)$
  is continuous and odd. %
  Therefore
  \begin{equation*}
    \tilde W\circ \phi:S^{N-1}\to\R^N
  \end{equation*}
  is continuous and odd. By the Borsuk-Ulam Theorem, this map has at least one zero. Thus there exists
  $f_0\in \set{f\in X_N}{\norm[{\infty}]{f}=\rho}\subseteq B_1^{k,d}$ with $\tilde W(f_0)=\Bnul\in\R^N$, i.e.\
  $W(f_0)=W(-f_0)$.

  We get
  \begin{equation*}
    2\rho = 2\norm[{\infty}]{f_0}\le \norm[\infty]{f_0-R(W(f_0))}+\norm[\infty]{-f_0-R(W(-f_0))}.
  \end{equation*}
  Therefore with $f_0$, $-f_0\in B_1^{k,d}$, either
  \begin{equation*}
    \norm[\infty]{f_0-R(W(f_0))}\ge \rho\qquad\text{or}\qquad
    \norm[\infty]{-f_0-R(W(-f_0))}\ge \rho.
  \end{equation*}
  Since $W$ and $R$ were arbitrary, the claim follows.
\end{proof}

\begin{proof}[of Theorem \ref{thm:codparam}]
  Fix $n\in\N$ such that $n^d\le N\le (n+1)^d\dfn \tilde N$ and set
  with the functions from Lemma \ref{lemma:psiBnun} 
  \begin{equation*}
    X_{\tilde N}\dfn {\rm span}\set{\psi_{\Bnu,n+1}}{\Bnu\in\{1,\dots,n+1\}^d}.
  \end{equation*}
  Since the $\psi_{\Bnu,n+1}$ have disjoint support, by Lemma \ref{lemma:psiBnun} it holds with $\rho\dfn (C_{2,k}3(n+1)^k)^{-1}$
  \begin{equation*}
    \set{f\in X_{\tilde N}}{\norm[\infty]{f}=\rho}\subseteq B_1^{k,d}.
  \end{equation*}
  Hence Lemma \ref{lemma:borsukulam} gives
  \begin{equation*}
    \gamma_{k,d,N}\ge \gamma_{k,d,\tilde N}\ge \rho = (C_{2,k}3(n+1)^k)^{-1}
    \ge C_{2,k}^{-1}6^{-k}N^{-k/d},
  \end{equation*}
  where the last inequality follows as in the proof of Theorem \ref{thm:codparam}.
\end{proof}

\subsection{Infinitely differentiable functions}\label{sec:codparaminfty}
So far we have seen a curse of dimensionality when approximating functions in $C^k([0,1]^d)$ for fixed and finite $k\in\N$. Our analysis in Theorem \ref{thm:Cks}, also see \eqref{eq:Ckexponential}, suggests that large dimension $d$ can be compensated by large $k$, at least in terms of the \emph{asymptotic convergence rate}. This naturally raises the following question: if we allow $k$ to grow with $d$, can we break the curse of dimensionality? The answer is no. Even for $C^\infty$ functions it persists in general. The next theorem, which %
corresponds to \cite[Theorem 1]{novak2009approximation}, makes this precise. It shows that to achieve a fixed approximation accuracy for all functions in $B_1^{\infty,d}$ (the unit ball in $C^\infty([0,1]^d)$) the number of parameters $N$ must still grow exponentially with $d$.

\begin{theorem}\label{thm:codparaminfty}
  For all $d\in\N$, $d\ge 2$, %
  \begin{equation*}
    \gamma_{\infty,d,2^{\lfloor d/2\rfloor}} \ge 1.
  \end{equation*}
\end{theorem}

For neural networks we can conclude the following.

\begin{corollary}
  For every $d\in\N$, $d\ge 2$, let $\Phi_d( \cdot,\Bw)$ be a neural network
  with input dimension $d$ and parameters $\Bw\in\R^{2^{\lfloor d/2\rfloor}}$, i.e.\
  \begin{equation*}
    {\rm size}(\Phi_d) = 2^{\lfloor d/2\rfloor}. %
  \end{equation*}
  Then, for every $d\ge 2$
  and every continuous weight selection $W_d:C^0([0,1]^d)\to\R^{2^{\lfloor d/2\rfloor}}$
  \begin{equation*}
    \sup_{f\in B_1^{\infty,d}}\norm[\infty]{f-\Phi_d(\cdot,W_d(f))}\ge 1.
  \end{equation*}
\end{corollary}

\begin{proof}[of Theorem \ref{thm:codparaminfty}]
  Fix $d\ge 2$
  and set $s\dfn \lfloor d/2\rfloor \ge 1$.
  We will construct a space $X_d\subseteq C^\infty([0,1]^d)$
  of dimension $2^s$ such that
  \begin{equation}\label{eq:codinftytoshow}
    \set{f\in X_d}{\norm[\infty]{f}=1}\subseteq 
    \set{f\in C^\infty([0,1]^d)}{\norm[{C^\infty([0,1]^d)}]{f}\le 1}=B_1^{\infty,d}.
  \end{equation}
  Together with Lemma \ref{lemma:borsukulam}, this then concludes the proof.

  {\bf Step 1.} We claim that for $a$, $b\in\R$ and $g(z)\dfn a+bz$,
  it holds for any $k\in\N_0$
  \begin{equation*}
    \sup_{z\in [0,2]}|g(z)|\ge \sup_{z\in [0,2]}|g^{(k)}(z)|.
  \end{equation*}
  Since $g^{(k)}\equiv 0$ for $k\ge 2$, it suffices to consider $k=1$, which corresponds to 
  \begin{equation*}
    \max\{|a|,|a+2b|\} = \sup_{z\in [0,2]}|g(z)|\ge \sup_{x\in [0,2]}|g'(z)|=|b|.
  \end{equation*}
  Assuming the inequality to be wrong implies $|a|+|a+2b|<|2b|$ and leads to the contradiction
  $|2b|\le |-a|+|a+2b|<|2b|$. 

  {\bf Step 2.} For $\Ba=(a_\Bnu)_{\Bnu\in\{0,1\}^s}\in\R^{2^s}$ define
  \begin{equation*}
    f_\Ba(\Bx)\dfn \sum_{\Bnu\in\{0,1\}^s}a_\Bnu(x_{1}+x_{2})^{\nu_1}(x_3+x_4)^{\nu_2}\dots (x_{2s-1}+x_{2s})^{\nu_s}\qquad\text{for }\Bx\in [0,1]^d.
  \end{equation*}
  We will show that \eqref{eq:codinftytoshow} is satisfied for the $2^s$ dimensional function space
  \begin{equation*}
    X_d\dfn \set{f_\Ba(\Bx)}{\Ba\in\R^{2^s}}\subseteq C^\infty([0,1]^d).
  \end{equation*}

  Fix $\Ba\in\R^{2^s}$ and set
  \begin{equation*}
        g_\Ba(\Bz)\dfn \sum_{\Bnu\in\{0,1\}^s}a_\Bnu z_1^{\nu_1} z_2^{\nu_2}\dots z_s^{\nu_s}\qquad\text{for }\Bz\in [0,2]^s.
  \end{equation*}
  Fix a multiindex $\Bbeta\in\N_0^s$. Since
  $z_j\mapsto g_\Ba(z_1,\dots,z_s)$ (and any partial derivative of $g_\Ba$) is affine linear for each $j$, according to Step 1
  \begin{align*}
    \sup_{\Bz\in [0,2]^s}|D^\Bbeta g_\Ba(\Bz)|
    &=\sup_{z_2,\dots,z_s\in [0,2]}\sup_{z_1\in [0,2]}\Big|\frac{\partial^{\beta_1}}{\partial z_1^{\beta_1}}\dots \frac{\partial^{\beta_s}}{\partial z_s^{\beta_s}} g_\Ba(z_1,\dots,z_s)\Big|\\
    &\le \sup_{z_2,\dots,z_s\in [0,2]}\sup_{z_1\in [0,2]}\Big|\frac{\partial^{\beta_2}}{\partial z_2^{\beta_2}}\dots \frac{\partial^{\beta_s}}{\partial z_s^{\beta_s}} g_\Ba(z_1,\dots,z_s)\Big|\\
    &\le\dots\le \sup_{\Bz\in [0,2]^s}|g_\Ba(\Bz)|.
  \end{align*}
  Now let $\Balpha\in\N_0^d$ be arbitrary.
  Then with $\beta_j=\alpha_{2j-1}+\alpha_{2j}$ and $z_j=x_{2j-1}+x_{2j}$ for $j=1,\dots,s$,
  \begin{equation*}
    D^\Bbeta g_\Ba(z_1,\dots,z_{s}) = D^\Balpha f_\Ba(x_1,\dots,x_d).
  \end{equation*}
  Therefore
  \begin{equation*}
    \sup_{\Bx\in [0,1]^d}\sup_{\Balpha\in\N_0^d}|D^\Balpha f_\Ba(\Bx)|
    =\sup_{\Bz\in [0,2]^s}\sup_{\Bbeta\in\N_0^s}|D^\Bbeta g_\Ba(\Bz)|
    =\sup_{\Bz\in [0,2]^s}|g_\Ba(\Bz)|
    =\sup_{\Bx\in [0,1]^d}|f_\Ba(\Bx)|.
  \end{equation*}
  This shows $\norm[\infty]{f}=\norm[C^\infty({[0,1]^d})]{f}$
  for any $f\in X_d$. Thus \eqref{eq:codinftytoshow} holds.
\end{proof}

\section{The Barron class}\label{sec:BarronClass}
In \cite{barron}, Barron introduced a set of functions
that can be approximated by neural networks without a curse of dimensionality.
This set, known as the {\bf Barron class}, is characterized by a specific
  type of bounded variation. %
  To define it,
for $g\in L^1(\R^d)$ we denote by
	\begin{align*}
	 \check g(\Bw)\dfn \int_{\R^d} g(\Bx) e^{\ii \Bw^\top \Bx} \dd \Bx
	\end{align*}
	its inverse Fourier transform.
        Then, for $C>0$ the Barron class is defined as
 	\[
		\Gamma_{C} \coloneqq \setc{f\in C(\R^d)}{\exists g \in L^1(\R^d) ,~\int_{\R^d} |\Bxi| |g(\Bxi)| \dd\Bxi \leq C \text{ and } f = \check{g}}.
	\]
	We say that a function $f \in \Gamma_C$ has a finite Fourier moment, even though technically the Fourier transform of $f$ may not be well-defined, since $f$ does not need to be integrable. 
	By the Riemann-Lebesgue Lemma, \cite[Lemma 1.1.1]{grochenig2013foundations}, the condition $f\in C(\R^d)$ in the definition of $\Gamma_{C}$ is automatically satisfied if $g \in L^1(\R^d)$ as in the definition exists.
	
	The following proof approximation result for functions in $\Gamma_C$ is due to \cite{barron}. The presentation of the proof is similar to \cite[Section 5]{Petersen}.
          
	\begin{theorem}%
          \label{thm:BarronLight}
          Let $\sigma:\R\to\R$ be sigmoidal (see Definition
          \ref{def:sigmoidalActivation}) and let $f\in \Gamma_C$ for
          some $C>0$. Denote by
          $B_1^d\dfn \set{\Bx\in\R^d}{\norm{\Bx}\le 1}$ the unit
          ball. Then, for every $c > 4C^2$ and every $N\in\N$ there
          exists a neural network $\Phi^f$ with architecture
          $(\sigma; d, N, 1)$ such that
		\begin{align}\label{eq:BarronApproximationEstimate}
			\frac{1}{|B_1^d|}\int_{B_1^d} \left|f(\Bx)- \Phi^f(\Bx)\right|^2 \dd\Bx \leq \frac{c}{N},
		\end{align}
                where %
                $|B_1^d|$ is the Lebesgue measure of $B_1^d$.
	\end{theorem}
\begin{remark}
The approximation rate in \eqref{eq:BarronApproximationEstimate} can be slightly improved under some assumptions on the activation function such as powers of the ReLU, \cite{siegel2022high}.
\end{remark}

Importantly, the dimension $d$ does not enter on the right-hand side of \eqref{eq:BarronApproximationEstimate},
in particular the convergence rate is not directly affected by the
dimension, which is in stark
contrast to %
the results of the previous chapters.
However,
it should be noted, that the constant $C$ may still have some
inherent $d$-dependence, see Exercise \ref{ex:GaussianBarronNormHasConstantDependingOnD}.

The proof of Theorem \ref{thm:BarronLight} is based on a peculiar
property of high-dimensional convex sets, which is described by the
(approximate) Carath\'eodory theorem, the original version of which
  was given in \cite{Carathodory1911berDV}. The more general version stated in the
  following lemma follows \cite[Theorem 0.0.2]{vershynin2018high} and
  \cite{barron, SAF_1980-1981____A5_0}. For its statement
  recall that $\overline{\co}(G)$ denotes the the closure of the convex hull of $G$.

\begin{lemma}
  \label{lem:ConvexCombinationResult}
  Let $H$ be a Hilbert space,
  and let $G\subseteq H$ be such that for some $B>0$ it holds that
  $\norm[H]{g}\le B$ for all $g\in G$.
Let $f \in \overline{\mathrm{co}}(G)$.
Then, for every $N \in \N$ and every $c> B^2$ there exist $(g_i)_{i=1}^N \subseteq G$ %
such that
	\begin{align} \label{eq:repOfConvHull}
		\left\|f - \frac{1}{N}\sum_{i=1}^N %
          g_i\right\|_{H}^2 \leq \frac{c}{N}.
	\end{align}
\end{lemma}
\begin{proof}
  Fix $\eps>0$ and $N\in\N$.  Since $f\in\overline{\co}(G)$, there
  exist coefficients $\alpha_1,\dots,\alpha_m\in [0,1]$
  summing to $1$, and linearly independent elements
  $h_1,\dots,h_m\in G$ such that
  \begin{equation*}
    f^*\dfn \sum_{j=1}^m\alpha_jh_j
  \end{equation*}
  satisfies $\norm[H]{f-f^*}<\eps$. We claim that there exists
  $g_1,\dots,g_N$, each in $\{h_1,\dots,h_m\}$, such that
  \begin{equation}\label{eq:maureyclaim}
    \normc[H]{f^*-\frac{1}{N}\sum_{j=1}^Ng_j}^2\le\frac{B^2}{N}.
  \end{equation}
    Since $\eps>0$ was
  arbitrary, this then concludes the proof. Since there exists an
  isometric isomorphism from ${\rm span}\{h_1,\dots,h_m\}$ to $\R^m$,
  there is no loss of generality in assuming $H=\R^m$ in the following.

  Let $X_i$, $i=1,\dots,N$, be i.i.d.\ $\R^m$-valued random variables with
  \begin{equation*}
    \bbP[X_i=h_j] = \alpha_j\qquad\text{for all }i=1,\dots,m.
  \end{equation*}
  In particular $\bbE[X_i]=\sum_{j=1}^m\alpha_jh_j=f^*$ for each $i$. Moreover, 
  \begin{align}\label{eq:expectationdiffbarron}
    \bbE\left[\normc[H]{f^*-\frac{1}{N}\sum_{j=1}^N X_j}^2\right] &=
                                                                    \bbE\left[\normc[H]{\frac{1}{N}\sum_{j=1}^N (f^*-X_j)}^2\right] \nonumber\\
                                                                  &= \frac{1}{N^2}\Bigg[\sum_{j=1}^N\norm[H]{f^*-X_j}^2+\sum_{i\neq j}\dup[H]{f^*-X_i}{f^*-X_j} \Bigg]\nonumber\\
                                                                  &= \frac{1}{N} \bbE[\norm[H]{f^*-X_1}^2]\nonumber\\
                                                                  &=\frac{1}{N}\bbE[\norm[H]{f^*}-2\dup[H]{f^*}{X_1}+\norm[H]{X_1}^2]\nonumber\\
                                                                  &=\frac{1}{N}\bbE[\norm[H]{X_1}^2-\norm[H]{f^*}^2]
                                                                    \le \frac{B^2}{N}.
  \end{align}
  Here we used that the $(X_i)_{i=1}^N$ are i.i.d., 
  the fact that $\bbE[X_i]=f^*$,
  as well as $\bbE{\inp{f^*-X_i}{f^*-X_j}}=0$ if $i\neq j$.  Since the
  expectation in \eqref{eq:expectationdiffbarron} is bounded by
  $B^2/N$, there must exist at least one realization of the random
  variables $X_i\in\{h_1,\dots,h_m\}$,
denoted as $g_i$, for which \eqref{eq:maureyclaim} holds. 
\end{proof}

Lemma \ref{lem:ConvexCombinationResult} %
provides a powerful tool:
If we want to approximate a function $f$ with a superposition of $N$ elements in a set $G$, then it is sufficient to show that $f$ can be represented as an arbitrary (infinite) convex combination of elements of $G$.

Lemma \ref{lem:ConvexCombinationResult} suggests that we can prove Theorem \ref{thm:BarronLight} by showing that each function in $\Gamma_C$ %
belongs to the closure of the convex hull of all neural networks with a single neuron, i.e.\ the set of all affine transforms of the sigmoidal activation function $\sigma$.
We make a small detour before proving this result.
We first show that each function $f \in \Gamma_C$ is in the closure of the convex hull of \emph{the set of affine transforms of Heaviside functions}, i.e.\ the set
\[
	G_C \coloneqq \setc{B_1^d  \ni \Bx \mapsto \gamma \cdot \ind_{\R_+}( \langle \Ba, \Bx \rangle  + b)}{\Ba \in \R^d, b \in \R,  |\gamma| \leq 2C}.
\]
The following lemma, corresponding to %
\cite[Theorem 2]{barron}
and \cite[Lemma 5.12]{Petersen},
provides a link between $\Gamma_C$ and $G_C$.

\begin{lemma}%
  \label{lem:ApproximationOfBarronClassByHeavisides}
	Let $d\in \N$,  $C >0$ and $f \in \Gamma_C$.
Then $f|_{B_1^d}-f(\Bnul) \in \overline{\mathrm{co}}(G_C)$,
where the closure is taken with respect to the norm
\begin{align}\label{eq:L2meannorm}
  \|g\|_{{L}^{2, \diamond}(B_1^d)} \coloneqq
\left(\frac{1}{|B_1^d|} \int_{B_1^d} |g(\Bx)|^2 \dd\Bx\right)^{1/2}.
  \end{align}
\end{lemma}

\begin{proof} {\bf Step 1.} We express $f(\Bx)$ via an integral.
  
  Since $f \in \Gamma_C$, we have that there exist $g \in L^1(\R^d)$ such that for all $\Bx \in \R^d$  
	\begin{align}\label{eq:BarronDivide}
		f(\Bx) - f(\Bnul)&= \int_{\R^d}  g(\Bxi) \left(e^{\ii \langle \Bx, \Bxi\rangle} - 1\right)\dd\Bxi\nonumber\\
		&= \int_{\R^d} \left|g(\Bxi)\right| \left(e^{\ii (\langle \Bx, \Bxi\rangle + \kappa(\Bxi))} - e^{\ii\kappa(\Bxi)}\right)\dd\Bxi\nonumber\\
		&= \int_{\R^d} \left|g(\Bxi)\right| \big(\cos(\langle \Bx, \Bxi\rangle + \kappa(\Bxi)) - \cos(\kappa(\Bxi))\big) \dd\Bxi,
	\end{align}
	where $\kappa(\Bxi)$ is the phase of $g(\Bxi)$,
        i.e.\ $g(\Bxi)=|g(\Bxi)|e^{\ii \kappa(\Bxi)}$,
        and the last equality follows since $f$ is real-valued.
        Define a measure $\mu$ on $\R^d$ via its Lebesgue density
	\[
          \dd\mu(\Bxi) \coloneqq \frac{1}{C'}|\Bxi| |g(\Bxi)|  \dd\Bxi,
	\]
	where $C'\coloneqq\int|\Bxi||g(\Bxi)|\dd\Bxi\leq C$; this is possible since $f \in \Gamma_C$. %
        Then \eqref{eq:BarronDivide} leads to 
	\begin{align} \label{eq:integralProbabilityMeasureBarron}
	f(\Bx) - f(\Bnul) = C' \int_{\R^d} \frac{\cos(\langle \Bx, \Bxi \rangle + \kappa(\Bxi)) - \cos(\kappa(\Bxi))}{|\Bxi|} \dd\mu(\Bxi).
	\end{align}

        {\bf Step 2.} We show that $\Bx\mapsto f(\Bx)-f(\Bnul)$ is
          in the $L^{2, \diamond}(B_1^d)$ closure of convex
          combinations of the functions
          $\Bx\mapsto q_\Bx(\Btheta)$, where $\Btheta \in \R^d$, and
          \begin{equation}\label{eq:qx}
            \begin{aligned}
          q_\Bx:&B_1^d \to\R\\
          &\Bxi \mapsto %
            C' \frac{\cos(\langle \Bx, \Bxi \rangle+ \kappa(\Bxi)) - \cos(\kappa(\Bxi))}{|\Bxi|}.
            \end{aligned}
        \end{equation}

        The cosine function is 1-Lipschitz.
        Hence for any $\Bxi\in\R^d$ the map \eqref{eq:qx} 
          is bounded by one. %
        In addition, it is easy to see that $q_\Bx$ is well-defined and continuous even in the origin.
	Therefore, for $\Bx \in B_1^d$, the integral \eqref{eq:integralProbabilityMeasureBarron} can be approximated by a Riemann sum, i.e.,
	\begin{align}
		&\left|C' \!\int_{\R^d} q_\Bx(\Bxi) \dd\mu(\Bxi) -  C' \sum_{\Btheta \in \frac{1}{n} \Z^d } q_\Bx(\Btheta) \cdot \mu(I_\Btheta)\right| \to 0\qquad\text{as }n\to\infty \label{eq:hereWeUseVeryLargeWeights}
	\end{align}
	where $I_\Btheta \coloneqq [0,1/n)^d + \Btheta$.
        Since $\Bx \mapsto f(\Bx) - f(\Bnul)$ is continuous and thus bounded on $B_{1}^d$, we have by the dominated convergence theorem that
\begin{align}
	&\frac{1}{|B_1^d|}\int_{B_1^d} \left|f(\Bx) - f(\Bnul)- C' \sum_{\Btheta \in \frac{1}{n} \Z^d} q_\Bx(\Btheta)\cdot \mu(I_\Btheta)\right|^2 \dd\Bx \to 0.\label{eq:ThisIsintheClosureOfTheConvexHullOfGC}
\end{align}
Since $\sum_{\Btheta \in \frac{1}{n} \Z^d} \mu(I_\Btheta) =  \mu(\R^d) = 1$, the claim holds.

{\bf Step 3.} %
We prove that $\Bx \mapsto q_\Bx(\Btheta)$ is in the $L^{2, \diamond}(B_1^d)$ closure of convex combinations of $G_C$ for every $\Btheta\in\R^d$.
  Together with Step 2, this then concludes the proof.

Setting $z = \langle \Bx , \Btheta/|\Btheta| \rangle$, %
the result follows if the maps
          \begin{equation}\label{eq:tqx}
            \begin{aligned}
              h_\Btheta:&[-1,1] \to\R\\
                           &z \mapsto %
                             C' \frac{\cos(|\Btheta| z + \kappa(\Btheta)) - \cos(\kappa(\Btheta))}{|\Btheta|}
            \end{aligned}
          \end{equation}
can be approximated arbitrarily well by convex combinations of functions of the form
\begin{align}\label{eq:TheseThingsApproximatequiteWell}
	[-1,1] \ni z \mapsto \gamma \ind_{\R_+}\left( a' z + b'\right),
\end{align}
where $a'$, $b' \in \R$ and $|\gamma| \leq 2C$.
To show this define for $T \in \N$
\begin{align*}
	g_{T,+} &\coloneqq \sum_{i=1}^T \frac{\left|h_\Btheta\left(\frac{i}{T}\right) -h_\Btheta\left(\frac{i-1}{T}\right)\right|}{2C} \left( 2C \mathrm{sign}\left( h_\Btheta\left(\frac{i}{T}\right) - h_\Btheta\left(\frac{i-1}{T}\right)\right) \ind_{\R_+}\left(x - \frac{i}{T}\right)\right),\\
	g_{T,-} &\coloneqq \sum_{i=1}^T \frac{\left|h_\Btheta\left( - \frac{i}{T} \right) - h_\Btheta\left(\frac{1-i}{T}\right)\right|}{2C}  \left( 2C \mathrm{sign}\left(h_\Btheta\left(-\frac{i}{T}\right) - h_\Btheta\left(\frac{1-i}{T}\right)\right)  \ind_{\R_+}\left(-x + \frac{i}{T}\right)\right).
\end{align*}
By construction, $g_{T,-} + g_{T,+}$ is a piecewise constant approximation to $h_\Btheta$ that interpolates $h_\Btheta$ at $i/T$ for $i = 1, \dots, T$. Since $h_\Btheta$ is continuous, we have that $g_{T,-} + g_{T,+}\to h_\Btheta$ uniformly as $T\to \infty$.
Moreover, $\norm[L^\infty(\R)]{h_\Btheta'} \leq C$ and hence %
\begin{align*}
	&\sum_{i=1}^T \frac{|h_\Btheta(i/T) - h_\Btheta((i-1)/T)|}{2C} + \sum_{i=1}^T \frac{|h_\Btheta( - i/T ) - h_\Btheta((1-i)/T)|}{2C}\\
	& \leq \frac{2}{2C T} \sum_{i=1}^T \norm[L^\infty(\R)]{h_\Btheta'}  \leq 1,
\end{align*}
where we used $C'\le C$ for the last inequality. We conclude that $g_{T,-} + g_{T,+}$ is a convex combination of functions of the form  \eqref{eq:TheseThingsApproximatequiteWell}.
Hence, %
$h_\Btheta$ can be arbitrarily well approximated by convex combinations of the form \eqref{eq:TheseThingsApproximatequiteWell}. This concludes the proof.
\end{proof}

We now have all tools to complete the proof of Theorem \ref{thm:BarronLight}.
\begin{proof}[of Theorem \ref{thm:BarronLight}]
	Let $f \in \Gamma_C$. %
        By Lemma \ref{lem:ApproximationOfBarronClassByHeavisides}
	\[
          f|_{B_1^d} - f(\Bnul) \in \overline{\mathrm{co}}(G_C),
	\]
        where the closure is understood with respect to the norm \eqref{eq:L2meannorm}.
	It is not hard to see that for every %
        $g \in G_C$ %
        it holds that $\|g\|_{L^{2,\diamond}(B_1^d)}\leq 2C$.
	Applying Lemma \ref{lem:ConvexCombinationResult} with the Hilbert space $L^{2, \diamond}(B_1^d)$, we get that
        for every $N \in \N$ there exist $|\gamma_i|\leq 2C$, $\Ba_i\in \R^d$, $b_i \in \R$, for $i = 1, \dots, N$, so that
	\[
		\frac{1}{|B_1^d|} \int_{B_1^d} \left| f(\Bx) - f(\Bnul) - \frac{1}{N} \sum_{i=1}^N \gamma_i\ind_{\R_+}(\langle \Ba_i, x\rangle +b_i)\right|^2 \dd\Bx \leq \frac{4C^2}{N}.
	\]
	By Exercise \ref{ex:sigmoidalStepfun1d}, it holds that $\sigma(\lambda \cdot) \to \ind_{\R_+}$ for $\lambda \to \infty$ almost everywhere.
Thus, %
for every $\delta>0$ there exist $\tilde{\Ba}_i$, $\tilde{b}_i$, $i = 1, \dots,N$, so that

	\[
	\frac{1}{|B_1^d|} \int_{B_1^d} \left| f(\Bx) - f(\Bnul) - \frac{1}{N} \sum_{i=1}^N \gamma_i\sigma\left(\langle \tilde{\Ba}_i, \Bx\rangle + \tilde{b}_i\right)\right|^2 \dd\Bx \leq \frac{4C^2}{N} + \delta.
	\]
	The result follows by observing that
	\[
          \frac{1}{N}\sum_{i=1}^N \gamma_i\sigma\left(\langle \tilde{\Ba}_i, \Bx\rangle + \tilde{b}_i\right) + f(\Bnul)
	\]
	is a neural network with architecture $(\sigma; d,N,1)$.
\end{proof}

The dimension-independent approximation rate of Theorem \ref{thm:BarronLight} may seem surprising, especially %
when comparing to the results in Chapters \ref{chap:Splines} and \ref{chap:ReLUNNs}.
However, %
this can be explained by recognizing that the assumption of a finite Fourier moment is effectively a \emph{dimension-dependent regularity assumption}.
Indeed, the condition becomes more restrictive in higher dimensions and hence the complexity of $\Gamma_C$ does not grow with the dimension.

To further explain this, let us relate the Barron class to classical function spaces.
In \cite[Section II]{barron} it was observed that a sufficient condition is that all derivatives of order up to $\lfloor d/2 \rfloor +2$ are square-integrable.
In other words, if $f$ belongs to the Sobolev space $H^{\lfloor d/2 \rfloor +2}(\R^d)$,
then $f$ is a Barron function.
Importantly,
the functions %
must become smoother, %
as the dimension increases.
This assumption would also imply an approximation rate
of $N^{-1/2}$ in the $L^2$ norm by sums of at most $N$ B-splines, see
\cite{oswald1990degree, devore1998nonlinear}.  However, in such
  estimates some constants may still depend exponentially on $d$,
  whereas all constants in Theorem \ref{thm:BarronLight} are
  controlled independently of $d$.

Another %
notable aspect of the approximation of Barron functions is that the absolute values of the weights other than the output weights are not bounded by a constant. %
To see this, we refer to \eqref{eq:hereWeUseVeryLargeWeights}, where arbitrarily large $\theta$ need to be used.
While $\Gamma_C$ is a compact set, the set of neural networks of the specified architecture for a fixed $N\in \N$ is not parameterized with a compact parameter set.
In a certain sense, this is reminiscent of Proposition \ref{prop:magic} and Theorem \ref{thm:kolmogorov}, where arbitrarily strong approximation rates where achieved by using a very complex activation function and a non-compact parameter space.

\section{Functions with compositionality structure}\label{sec:compositionalityApprox}

As a next instance of types of functions for which the curse of dimensionality can be overcome, we study functions with compositional structure.
In words, this means that we study high-dimensional functions that are constructed by composing many low-dimensional functions. 
This point of view was proposed in \cite{poggio2017and}. 
Note that this can be a realistic assumption in many cases, such as for sensor networks, where local information is first aggregated in smaller clusters of sensors before some information is sent to a processing unit for further evaluation.

We introduce a model for compositional functions next.
Consider a directed acyclic graph $\CG$ with $M$ vertices
$\eta_1,\dots,\eta_M$ such that
\begin{itemize}
\item exactly $d$ vertices, $\eta_1,\dots,\eta_d$, have no ingoing edge,
\item each vertex has at most $m\in\N$ ingoing edges,
\item exactly one vertex, $\eta_M$, has no outgoing edge.
\end{itemize}

\begin{figure}[htb]
	\centering
	\includegraphics[width =0.6\textwidth]{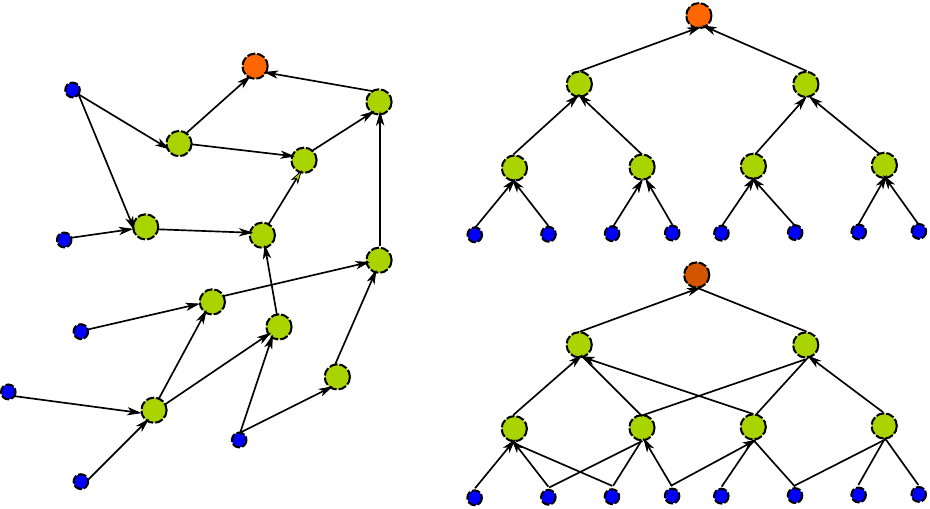}
	\caption{Three types of graphs that could be the basis of compositional functions.
The associated functions are composed of two or three-dimensional functions only.}
	\label{fig:HierarchicalFunctions}
\end{figure}

With each vertex $\eta_j$ for $j>d$ we associate a function
$f_j:\R^{d_j}\to\R$.
Here $d_j$ denotes the cardinality of the set
$S_j$, which is defined as the set of indices $i$ corresponding to
vertices $\eta_i$ for which we have an edge from $\eta_i$ to $\eta_j$.
Without loss of generality, we assume that $m\ge d_j=|S_j|\ge 1$ for
all $j>d$.
Finally, we let
\begin{subequations}\label{eq:compfunc}
\begin{align}
	F_j\dfn x_j\quad \text{ for all } \quad j\le d
\end{align}
and\footnote{The ordering of the inputs $(F_i)_{i\in S_j}$ in
\eqref{eq:compfuncb} is arbitrary but considered fixed throughout.}
\begin{align}\label{eq:compfuncb}
	F_j\dfn f_j((F_i)_{i\in S_j})\quad \text{ for all } \quad j>d.
\end{align}
\end{subequations}
Then $F_M(x_1,\dots,x_d)$ is a function from $\R^d\to\R$.
Assuming
\begin{align}\label{eq:hatfjfj}
	\norm[C^{k,s}(\R^{d_j})]{f_j}\le 1\quad \text{ for all } \quad j=d+1,\dots,M,
\end{align}
we denote the set of all functions of the type $F_M$ by $\CF^{k,s}(m,d,M)$.
Figure \ref{fig:HierarchicalFunctions} shows possible graphs of such functions.

Clearly, for $s=0$, $\CF^{k,0}(m,d,M)\subseteq C^{k}(\R^d)$ since the
composition of functions in $C^k$ belongs again to $C^k$.
A direct
application of Theorem~\ref{thm:Cks} allows to approximate
$F_M\in \CF^{k}(m,d,M)$ with a neural network of size $O(N\log(N))$ and
error $O(N^{-\frac{k}{d}})$.
Since each $f_j$ depends only on $m$
variables, intuitively we expect an error convergence of type
$O(N^{-\frac{k}{m}})$ with the constant somehow depending on the
number $M$ of vertices.
To show that this is actually possible, in the following we associate
with each node $\eta_j$ a depth $l_j\ge 0$, such that $l_j$ is the
maximum number of edges connecting $\eta_j$ to one of the nodes
$\{\eta_1,\dots,\eta_d\}$.

\begin{proposition}\label{prop:Gfunctions}
Let $k$, $m$, $d$, $M \in \N$ and $s>0$.
Let
$F_M\in\CF^{k,s}(m,d,M)$.
Then there exists a constant
$C=C(m,k+s,M)$ such that for every $N \in \N$ there exists a ReLU neural network
$\Phi^{F_M}$ such that
\begin{align*}
	\size(\Phi^{F_M})\le C N\log(N),\qquad\depth(\Phi^{F_M})\le C\log(N)
\end{align*}
and
\begin{align*}
	\sup_{\Bx\in [0,1]^d}|F_M(\Bx)-\Phi^{F_M}(\Bx)|\le N^{-\frac{k+s}{m}}.
\end{align*}
\end{proposition}
\begin{proof}
Throughout this proof we assume without loss of generality that the
indices follow a topological ordering, i.e., they are ordered such
that $S_j\subseteq\{1,\dots,j-1\}$ for all $j$ (i.e.\ the inputs of
vertex $\eta_j$ can only be vertices $\eta_i$ with $i<j$).

{\bf Step 1.}
First assume that $\hat f_j$ are functions such that with $0 < \eps \leq 1$
\begin{align}\label{eq:fj-hfj}
	|f_j(\Bx)-\hat f_j(\Bx)|\le \delta_j\dfn \eps \cdot (2m)^{-(M+1-j)}
	\quad \text{ for all } \quad \Bx\in [-2,2]^{d_j}.
\end{align}
Let $\hat F_j$ be defined as in \eqref{eq:compfunc}, but with all
$f_j$ in \eqref{eq:compfuncb} replaced by $\hat f_j$.
We now check
the error of the approximation $\hat F_M$ to $F_M$.
To do so we
proceed by induction over $j$ and show that for all
$\Bx\in [-1,1]^d$
\begin{align}\label{eq:Fi-hFi}
	|F_j(\Bx)-\hat F_j(\Bx)|\le (2m)^{-(M-j)}\eps.
\end{align}
Note that due to $\norm[C^k]{f_j}\le 1$ we have $|F_j(\Bx)|\le 1$
and thus \eqref{eq:Fi-hFi} implies in particular that
$\hat F_j(\Bx)\in [-2,2]$.

For $j=1$ it holds $F_1(x_1)=\hat F_1(x_1)=x_1$, and thus
\eqref{eq:Fi-hFi} is valid for all $x_1\in [-1,1]$.
For the
induction step, for all $\Bx\in [-1,1]^d$ by \eqref{eq:fj-hfj} and
the induction hypothesis
\begin{align*}
	|F_j(\Bx)-\hat F_j(\Bx)|
	& = |f_j((F_i)_{i\in S_j})-\hat f_j((\hat F_i)_{i\in S_j})|\nonumber\\
	& = |f_j((F_i)_{i\in S_j})-f_j((\hat F_i)_{i\in S_j})|+|f_j((\hat F_i)_{i\in S_j})-\hat f_j((\hat F_i)_{i\in S_j})|\nonumber\\
	& \le \sum_{i\in S_j}|F_i-\hat F_i|+\delta_j\nonumber\\
	&\le m \cdot (2m)^{-(M-(j-1))}\eps+(2m)^{-(M+1-j)}\eps\nonumber\\
	&\le (2m)^{-(M-j)}\eps.
\end{align*}
Here we used that $|\frac{d}{dx_r}f_j((x_i)_{i\in S_j})|\le 1$ for
all $r\in S_j$ so that by the triangle inequality and the mean value theorem
\begin{align*}
	|f_j((x_i)_{i\in S_j})-f_j((y_i)_{i\in S_j})|
	&\le \sum_{r\in S_j}|f((x_i)_{\substack{i\in S_j\\i\le r}},(y_i)_{\substack{i\in S_j\\i>r}})
	-f((x_i)_{\substack{i\in S_j\\i< r}},(y_i)_{\substack{i\in S_j\\i\ge r}})|\nonumber\\
	&\le \sum_{r\in S_j}|x_r-y_r|.
\end{align*}
This shows that \eqref{eq:Fi-hFi} holds, and thus for all
$\Bx\in [-1,1]^d$
\begin{align*}
	|F_M(\Bx)-\hat F_M (\Bx)|\le \eps.
\end{align*}

{\bf Step 2.} We sketch a construction, of how to write $\hat F_M$
from Step 1 as a neural network $\Phi^{F_M}$ of the asserted size and depth
bounds.
Fix $N\in\N$ and let
\begin{align*}
	N_j\dfn \lceil N(2m)^{\frac{m}{k+s}(M+1-j)}\rceil.
\end{align*}
By Theorem~\ref{thm:Cks}, since $d_j\le m$, we can find a neural network $\Phi^{f_j}$
satisfying %
\begin{align}\label{eq:fj-hfj2}
	\sup_{\Bx\in [-2,2]^{d_j}}|f_j(\Bx)-\Phi^{f_j}(\Bx)|\le
	N_j^{-\frac{k+s}{m}}\le 
	N^{-\frac{k+s}{m}}(2m)^{-(M+1-j)}
\end{align}
and
\begin{align*}
	\size(\Phi^{f_j}) \le C N_j\log(N_j) \le C N (2m)^{\frac{m(M+1-j)}{k+s}}\left(\log(N)+\log(2m)\frac{m(M+1-j)}{k+s}\right)
\end{align*}
as well as
\begin{align*}
	\depth(\Phi^{f_j})\le C \cdot \left(\log(N)+\log(2m)\frac{m(M+1-j)}{k+s}\right).
\end{align*}
Then %
\begin{align*}
	\sum_{j=1}^{M}\size(\Phi^{f_j})&\le 2 C N\log(N)\sum_{j=1}^M (2m)^{\frac{m(M+1-j)}{k+s}}\nonumber\\
	&\le 2C N\log(N) \sum_{j=1}^M \left((2m)^{\frac{m}{k+s}}\right)^j\nonumber\\
	&\le 2C N\log(N) (2m)^{\frac{m(M+1)}{k+s}}.
\end{align*}
Here we used
$\sum_{j=1}^M a^j\le\int_1^{M+1}\exp(\log(a)x)\dd x\le
\frac{1}{\log(a)}a^{M+1}$.

The function $\hat F_M$ from Step 1 then will yield error
$N^{-\frac{k+s}{m}}$ by \eqref{eq:fj-hfj} and \eqref{eq:fj-hfj2}.
We
observe that $\hat F_M$ can be constructed inductively as a neural network $\Phi^{F_M}$ by
propagating all values $\Phi^{F_1},\dots,\hat \Phi^{F_j}$ to all consecutive
layers using identity neural networks and then using the outputs of 
$(\Phi^{F_i})_{i\in S_{j+1}}$ as input to $\Phi^{f_{j+1}}$.
The depth of
this neural network is bounded by
\begin{align*}
	\sum_{j=1}^M\depth(\Phi^{f_j}) = O(M\log(N)).
\end{align*}
We have at most $\sum_{j=1}^M |S_j|\le mM$ values which need to be
propagated through these $O(M\log(N))$ layers, amounting to an
overhead $O(mM^2\log(N))=O(\log(N))$ for the identity neural networks.
In
all the neural network size is thus $O(N\log(N))$.
\end{proof}

\begin{remark}
From the proof we observe that the constant $C$ in
Proposition~\ref{prop:Gfunctions} behaves like
$O((2m)^{\frac{m(M+1)}{k+s}})$.
\end{remark}

\section{Functions on manifolds}\label{sec:ManifoldAssumption}

Another instance in which %
the curse of dimension can be mitigated, is if the input to the
network belongs to $\R^d$, but stems from an $m$-dimensional manifold
$\CM\subseteq\R^d$.
If we only measure the approximation error on $\CM$, then we can again show that it is $m$
rather than $d$ that determines the rate of convergence.

\begin{figure}
	\centering\includegraphics[width = 0.6\textwidth]{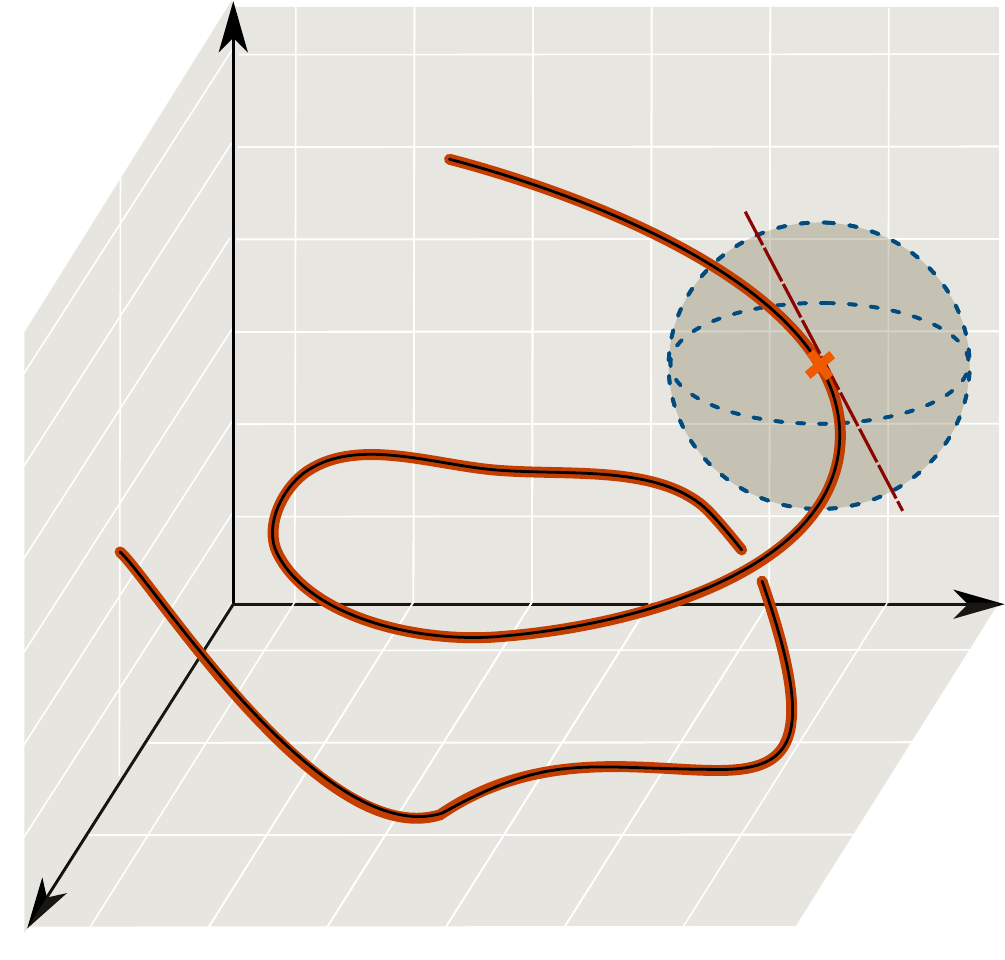}
	\put(-0.33\textwidth, 0.5\textwidth){\large\color{red} $\mathcal{M}$}
	\caption{One-dimensional sub-manifold of three-dimensional space.
At the orange point, we depict a ball and the tangent space of the manifold.
}
	\label{fig:manifoldAssumption}
      \end{figure}

To explain the idea, we assume in the following that $\CM$ is a
smooth, compact $m$-dimensional manifold in $\R^d$.
Moreover, we
suppose that there exists $\delta>0$ and finitely many points
$\Bx_1,\dots,\Bx_M\in\CM$ such that the $\delta$-balls
$B_{\delta/2}(\Bx_i)\dfn
\set{\By\in\R^d}{\norm[2]{\By-\Bx}<{\delta}/{2}}$ for
$j=1,\dots,M$ cover $\CM$ (for every $\delta>0$ such $\Bx_i$ exist since
$\CM$ is compact).
Moreover, denoting by $T_{\Bx}\CM\simeq \R^m$ the
tangential space of $\CM$ at $\Bx$, we assume $\delta>0$ to be so
small that the orthogonal projection
\begin{align}\label{eq:pij}
	\pi_j:B_{\delta}(\Bx_j)\cap\CM\to T_{\Bx_j}\CM
\end{align}
is injective, the set
$\pi_j(B_{\delta}(\Bx_j)\cap\CM)\subseteq T_{\Bx_j}\CM$ has $C^\infty$
boundary, and the inverse of $\pi_j$, i.e.
\begin{align}\label{eq:pijinv}
	\pi_j^{-1}:\pi_j(B_{\delta}(\Bx_j)\cap\CM)\to\CM
\end{align}
is $C^\infty$ (this is possible because $\CM$ is a smooth manifold). 
A visualization of this assumption is shown in Figure \ref{fig:manifoldAssumption}.

Note that $\pi_j$ in \eqref{eq:pij} is a linear map, whereas
$\pi_j^{-1}$ in \eqref{eq:pijinv} is in general non-linear.

For a function $f:\CM\to\R$ we can then write
\begin{align}\label{eq:fjpij}
  f(\Bx) = f(\pi_j^{-1}(\pi_j(\Bx)))=f_j(\pi_j(\Bx))
  \qquad\text{for all }\Bx\in B_\delta(\Bx_j)\cap\CM
\end{align}
where
\begin{align*}
	f_j\dfn f\circ \pi_j^{-1}:\pi_j(B_{\delta}(\Bx_j)\cap\CM)\to\R.
\end{align*}
In the following, for $f:\CM\to\R$, $k\in\N_0$, and $s\in [0,1)$ we let
\begin{align*}
	\norm[C^{k,s}(\CM)]{f}\dfn \sup_{j=1,\dots,M}\norm[C^{k,s}(\pi_j(B_{\delta}(\Bx_j)\cap\CM))]{f_j}.
\end{align*}
We now state the main result of this section.

\begin{proposition}
Let $d$, $k \in \N$, $s \geq 0$, and let $\CM$ be a
smooth, compact $m$-dimensional manifold in $\R^d$. 
Then there exists a constant $C>0$ such that for all $f\in C^{k,s}(\CM)$ and every $N\in\N$ there exists a ReLU neural network
$\Phi_N^f$ such that $\size(\Phi_N^f)\le CN\log(N)$,
$\depth(\Phi_N^f)\le C\log(N)$ and
\begin{align*}
	\sup_{\Bx\in\CM}|f(\Bx)-\Phi_N^f(\Bx)|\le C \norm[C^{k,s}(\CM)]{f}N^{-\frac{k+s}{m}}.
\end{align*}
\end{proposition}

\begin{proof}
Since $\CM$ is compact there exists $A>0$ such that
$\CM\subseteq [-A,A]^d$.
Similar as in the proof of
Theorem~\ref{thm:Cks}, we consider a uniform mesh with nodes
$\set{-A+2A\frac{\Bnu}{n}}{\Bnu\le n}$, and the corresponding
piecewise linear basis functions forming the partition of unity
$\sum_{\Bnu\le n}\varphi_\Bnu\equiv 1$ on $[-A,A]^d$ where
$\supp\varphi_\Bnu\subseteq
\set{\By\in\R^d}{\norm[\infty]{\frac{\Bnu}{n}-\By}\le \frac{A}{n}}$.
Let $\delta>0$ be as in the beginning of this section. 
Since $\CM$ is covered by the balls $(B_{\delta/2}(\Bx_j))_{j=1}^M$,
fixing $n \in \N$ large enough, for each $\Bnu$ such that
$\supp\varphi_\Bnu \cap\CM\neq\emptyset$ there exists
$j(\Bnu)\in\{1,\dots,M\}$ such that
$\supp\varphi_\Bnu\subseteq B_\delta(\Bx_{j(\Bnu)})$ and we set
$I_j\dfn \set{\Bnu\le n
}{j=j(\Bnu)}$.
Using \eqref{eq:fjpij} we then have for all $\Bx\in\CM$
\begin{align}\label{eq:fpum}
  f(\Bx)=%
 \sum_{\Bnu\le n}\varphi_\Bnu(\Bx) f(\Bx)
	=\sum_{j=1}^M\sum_{\Bnu\in I_j}\varphi_\Bnu(\Bx) f_j(\pi_j(\Bx)).
\end{align}

Next, we approximate the functions
$f_j$.
Let $C_j$ be the smallest ($m$-dimensional) cube in
$T_{\Bx_j}\CM\simeq\R^m$ such that
$\pi_j(B_\delta(\Bx_j)\cap\CM)\subseteq C_j$.
The function $f_j$ can be extended to a
function on $C_j$ (we will use the same notation for this extension)
such that%
\begin{align*}
	\norm[C^{k,s}(C_j)]{f}\le C \norm[C^{k,s}(\pi_j(B_\delta(\Bx_j)\cap\CM))]{f},
\end{align*}
for some constant depending on $\pi_j(B_\delta(\Bx_j)\cap\CM)$ but
independent of $f$. 
Such an extension result can, for example, be found in \cite[Chapter VI]{MR0290095}.
By Theorem~\ref{thm:Cks} (also see
Remark~\ref{rmk:Cks}), there exists a neural network $\hat f_j:C_j\to\R$ such that
\begin{align}
	\sup_{\Bx\in C_j}|f_j(\Bx)-\hat f_j(\Bx)|\le C N^{-\frac{k+s}{m}}
\end{align}
and
\begin{align*}
	\size(\hat f_j)\le C N\log(N),\qquad
	\depth(\hat f_j)\le C\log(N).
\end{align*}

To approximate $f$ in \eqref{eq:fpum} we now let with
$\eps\dfn N^{-\frac{k+s}{m}}$ %
\begin{align*}
	\Phi_N\dfn \sum_{j=1}^M\sum_{\Bnu\in I_j}\ntim{\eps}(\varphi_\Bnu,\hat f_i\circ\pi_j),
\end{align*}
where we note that $\pi_j$ is linear and thus $\hat f_j\circ\pi_j$
can be expressed by a neural network.
First let us estimate the error of this
approximation.
For $\Bx\in\CM$
\begin{align*}
	|f(\Bx)-\Phi_N(\Bx)|
	&\le\sum_{j=1}^M \sum_{\Bnu\in I_j}|\varphi_\Bnu(\Bx)f_j(\pi_j(\Bx))-\ntim{\eps}(\varphi_\Bnu(\Bx),\hat f_j(\pi_j(\Bx)))|\nonumber\\
	&\le\sum_{j=1}^M \sum_{\Bnu\in I_j}\left(|\varphi_\Bnu(\Bx)f_j(\pi_j(\Bx))
	  -\varphi_\Bnu(\Bx)\hat f_j(\pi_j(\Bx))| \right.\\
	  & \qquad \left.
	+ |\varphi_\Bnu(\Bx)\hat f_j(\pi_j(\Bx))
	  -\ntim{\eps}(\varphi_\Bnu(\Bx),\hat f_j(\pi_j(\Bx)))|\right)\nonumber\\
	&\le\sup_{i\le M}\norm[L^\infty(C_i)]{f_i-\hat f_i}
	  \sum_{j=1}^M \sum_{\Bnu\in I_j}|\varphi_\Bnu(\Bx)|
	  +
	  \sum_{j=1}^M \sum_{\set{\Bnu\in I_j}{\Bx\in\supp\varphi_\Bnu}}\eps\nonumber\\
	&\le CN^{-\frac{k+s}{m}} + d\eps\le CN^{-\frac{k+s}{m}},
\end{align*}
where we used that $\Bx$ can be in the support of at most $d$ of the
$\varphi_\Bnu$, and where $C$ is a constant depending on $d$ and
$\CM$.

Finally, let us bound the size and depth of this approximation.
Using $\size(\varphi_\Bnu)\le C$, $\depth(\varphi_\Bnu)\le C$ (see \eqref{eq:weNeedThisLaterInTheManifoldSection2})
and
$\size(\ntim{\eps})\le C\log(\eps)\le C\log(N)$ and
$\depth(\ntim{\eps})\le C\depth(\eps)\le C\log(N)$ (see Lemma
\ref{lemma:mult}) we find
\begin{align*}
	\sum_{j=1}^M\sum_{\Bnu\in I_j}\left(\size(\ntim{\eps})+\size(\varphi_\Bnu)+\size(\hat f_i\circ\pi_j)\right)
	&\le \sum_{j=1}^M\sum_{\Bnu\in I_j}C\log(N)+C+CN\log(N)\nonumber\\
	&=O(N\log(N)),
\end{align*}
which implies the bound on $\size(\Phi_N)$.
Moreover, 
\begin{align*}
	\depth(\Phi_N)&\le \depth(\ntim{\eps})+\max\left\{\depth(\varphi_\Bnu,\hat f_j)\right\}\\
	&\le C \log(N)+\log(N)=O(\log(N)).
\end{align*}
This completes the proof.
\end{proof}

\section*{Bibliography and further reading}
Section \ref{sec:cod} on the curse of dimensionality is based on the papers \cite{devore1998nonlinear, novak2009approximation}. In particular Theorems \ref{thm:coddata} and \ref{thm:codparam} are simplifications of the results in \cite{devore1998nonlinear}, and Theorem \ref{thm:codparaminfty} is a modification of the main result in \cite{novak2009approximation} which has a different assumption on $W$ (its components are assumed bounded linear); the main part of the argument is exactly the same however, and also the proofs for the other statements closely follow these references. Other %
relevant literature in this direction includes for example \cite{kolmogorov,pinkus2012n,devore1993constructive,MR1819645,Novak:2008}.

The ideas of Section \ref{sec:BarronClass} were originally developed in \cite{barron}, with an extension to $L^\infty$ approximation provided in \cite{barron1992neural}. These arguments can be extended to yield dimension-independent approximation rates for high-dimensional discontinuous functions, provided the discontinuity follows a Barron function, as shown in \cite{Petersen}. 
The Barron class has been generalized in various ways, as discussed in \cite{ma2018priori, ma2020towards, weinan2019barron, weinan2022representation, barron2018approximation}.

The compositionality assumption of Section \ref{sec:compositionalityApprox} was discussed in the form presented in \cite{poggio2017and}. An alternative approach, known as the %
hierarchical composition/interaction model, was studied in \cite{kohler2021rate}.

The manifold assumption discussed in Section \ref{sec:ManifoldAssumption} is frequently found in the literature, with notable examples including \cite{shaham2018provable, chui2018deep, chen2019efficient, schmidt2019deep, nakada2020adaptive, kohler2022estimation}.

Another prominent direction, %
  omitted in this chapter, pertains to scientific machine learning. High-dimensional functions often arise from (parametric) PDEs, which have a rich literature describing their properties and structure. Various results have shown that neural networks can leverage the inherent low-dimensionality known to exist in such problems. Efficient approximation of certain classes of high-dimensional (or even infinite-dimensional) analytic functions, ubiquitous in parametric PDEs, has been verified in \cite{Schwab2019Deep, Schwab2023Deep} based on \cite{yarotsky}. Further general analyses for high-dimensional parametric problems can be found in \cite{Opschoor2022Bayesian, kutyniok2022theoretical}, and results exploiting specific structural conditions of the underlying PDEs, e.g., in \cite{laakmann2021efficient,DeRyck2023}. Additionally, \cite{MR3856963,doi:10.1137/18M1189336,OSZ21} provide results regarding fast convergence for certain smooth functions in potentially high but finite dimensions.

For high-dimensional PDEs, elliptic problems have been addressed in \cite{grohs2022deep}, linear and semilinear parabolic evolution equations have been explored in \cite{grohs2023proof, gonon2021deep, hutzenthaler2020proof}, and stochastic differential equations in \cite{MR4283528}.

\newpage
\section*{Exercises}
\begin{exercise}\label{ex:affineTransformationsBarron}
	Let $C>0$ and $d\in \N$.
Show that, if $g \in \Gamma_C$, then 
	\[
		a^{-d} g\left(a (\cdot -\Bb)\right) \in \Gamma_C,
	\]
	for every $a\in \R_+$, $\Bb\in \R^d$.	
\end{exercise}

\begin{exercise}\label{ex:SumSBarron}
	Let $C>0$ and $d\in \N$.
Show that, for $g_i \in \Gamma_C$, $i = 1, \dots, m$ and $c= (c_i)_{i=1}^m $ it holds that 
	\[
	\sum_{i=1}^m c_i g_i \in \Gamma_{\|c\|_1 C}.
	\]
\end{exercise}

\begin{exercise}\label{ex:GaussianBarronNormHasConstantDependingOnD}	
  Show that for every $d\in\N$ the function
	$f(\Bx)\dfn \exp(-{\norm[2]{\Bx}^2}/{2})$, $\Bx\in \R^d$, belongs to
	$\Gamma_d$, and it holds $C_f=O(\sqrt{d})$, for $d \to \infty$.
\end{exercise}

\begin{exercise}
	Let $d\in \N$, and let $f(\Bx) = \sum_{i=1}^\infty c_i \sigma_{\rm ReLU}(\langle \Ba_i, \Bx \rangle + b_i)$ for $\Bx \in \R^d$ with $\|\Ba_i\| = 1, |b_i| \leq 1 $ for all $i \in \N$.
Show that for every $N \in \N$, there exists a ReLU neural network with $N$ neurons and one layer such that 
	\[
		\|f - f_N\|_{L^2(B_1^d)} \leq \frac{3 \|c\|_1}{\sqrt{N}}.
	\]
	Hence, every infinite ReLU neural network can be approximated at a rate $O(N^{1/2})$ by finite ReLU neural networks of width $N$.
\end{exercise}

\begin{exercise}	
	Let $C>0$ prove that every $f \in \Gamma_C$ is continuously differentiable.
      \end{exercise}

%% file: Interpolation.tex
\newcommand{\lip}{{\rm Lip}}

\chapter{Interpolation}\label{chap:Interpolation}
The learning problem associated to minimizing the empirical risk of
\eqref{eq:empiricalRiskDef0} is based on minimizing an error that
results from evaluating a neural network on a \emph{finite} set of
(training) points.
In contrast, all previous approximation
results %
focused on achieving uniformly small errors across the entire
domain. 
Finding neural networks that achieve a small training
error
appears to be much simpler,
since, instead of
$\|f - \Phi_n \|_\infty \to 0$ for a sequence of
neural networks $\Phi_n$, it suffices to have
$\Phi_n(\Bx_i) \to f(\Bx_i)$ for all $\Bx_i$ in the training set.

In this chapter, we study the extreme case of the aforementioned approximation problem. 
We analyze under which conditions it is possible to find a neural network that coincides with the target function $f$ at all training points.
This is referred to as
\emph{interpolation}.
To make this notion more precise, we state the following definition.

\begin{definition}[Interpolation]
	Let $d$,  $m \in \N$, and let $\Omega \subseteq \R^d$.
	We say that a set of functions $\CH \subseteq \{h \colon  \Omega \to \R\}$ \textbf{interpolates $m$ points in $\Omega$}, if
	for every  $S = (\Bx_i, y_i)_{i=1}^m \subseteq \Omega \times \R$, such that $\Bx_i \neq \Bx_j$ for $i \neq j$, there exists a function $h \in \CH$ such that $h(\Bx_i) = y_i$ for all $i = 1, \dots, m$.
	
\end{definition}

Knowing the interpolation properties of an architecture represents extremely valuable information for two reasons:
\begin{itemize} 
	\item Consider
	an architecture that interpolates $m$ points and
	let the number of training samples be bounded by $m$.
	Then \eqref{eq:empiricalRiskDef0} always has a
	solution.
	\item Consider again
	an architecture that interpolates $m$ points and
	assume that the number of training samples is \emph{less} than $m$.
	Then for every point  $\tilde{\Bx}$ not in the training set and every $y \in \R$
	there exists a minimizer $h$ of \eqref{eq:empiricalRiskDef0} that satisfies $ h(\tilde{\Bx}) = y$.
	As a consequence, without further restrictions (many of which we will discuss below), such an architecture can (in general) not generalize to unseen data.
\end{itemize}
The existence of solutions to the interpolation problem does not
follow
trivially 
from the approximation results provided in the previous chapters (even
though we will later see that there is a close connection).
We also
remark that the question of how many points neural networks with a
given architecture can interpolate is %
closely related to the so-called VC dimension, which we will study in
Chapter \ref{chap:VC}.

We start our analysis of the interpolation properties of neural networks by presenting a result similar to the universal approximation theorem but for interpolation in the following section.
In the subsequent section, we then look at interpolation with desirable properties.

\section{Universal interpolation}

Under what conditions on the activation function and architecture can
a set of neural networks interpolate $m \in \N$ points? According
to Chapter \ref{chap:UA}, particularly Theorem \ref{thm:universal},
we know that shallow neural networks can approximate every continuous
function with arbitrary accuracy, provided the neural network width is
large enough.
As the neural network's width and/or depth increases, the
architectures become increasingly powerful, leading us to expect
that at some point, they should be able to interpolate $m$
points.
However, this intuition may not be correct:
\begin{example}
	Let $\CH:=\set{f\in C^0([0,1])}{f(0)\in\Q}$.
	Then
	$\CH$ is dense in $C^0([0,1])$, but $\CH$ does not even
	interpolate one point in $[0,1]$.
	
\end{example}
Moreover, Theorem \ref{thm:universal} is an asymptotic result that
only states that a given function can be approximated for sufficiently
large neural network architectures, but it does not state how large
the architecture needs to be.

Surprisingly, Theorem \ref{thm:universal} can nonetheless be used to give a guarantee that a fixed-size architecture yields sets of neural networks that allow the interpolation of $m$ points.
  This result is due to \cite{MR1819645}; for a more detailed discussion of previous results see the bibliography section.
Due to its similarity to the universal approximation theorem and the
fact that it uses the same assumptions, we call the following theorem
the ``Universal Interpolation Theorem''.
For its statement recall
the definition of the set of allowed activation functions $\CM$ in
\eqref{eq:M} and the class $\CN_d^1(\sigma,1,n)$ of shallow neural networks
of width $n$ introduced in Definition \ref{def:CN}.

\begin{theorem}[Universal Interpolation Theorem]\label{thm:universalInterpolationThm}
	Let $d$, $n \in \N$ and let $\sigma \in \mathcal{M}$ not be a
	polynomial.
	Then $\CN_d^1(\sigma,1,n)$
	interpolates $n+1$ points in $\R^d$.
\end{theorem}

\begin{proof}
	Fix $(\Bx_i)_{i=1}^{n+1}\subseteq\R^d$ arbitrary.
	We will show that for
	any $(y_i)_{i=1}^{n+1}\subseteq\R$ there exist weights
	and biases $(\Bw_j)_{j=1}^n\subseteq \R^d$,
	$(b_j)_{j=1}^n$, $(v_j)_{j=1}^n\subseteq \R$, $c\in \R$ such that
	\begin{align}\label{eq:UITtoshow}
		\Phi(\Bx_i):=\sum_{j=1}^{n} v_j\sigma(\Bw_j^\top\Bx_i+b_j)+c=y_i\quad\text{for all }\quad i=1,\dots,n+1.
	\end{align}
	Since $\Phi\in \CN_d^1(\sigma,1,n)$ this then concludes the proof.
	
	Denote
	\begin{align}\label{eq:AMatrixThatShouldBeRegular}
		\BA\dfn\begin{pmatrix}
			1&      \sigma(\Bw_1^\top\Bx_1+b_1)&\cdots &\sigma(\Bw_{n}^\top\Bx_1+b_{n}) \\
			\vdots &\vdots&\ddots&\vdots \\
			1&      \sigma(\Bw_1^\top\Bx_{n+1}+b_1)&\cdots &\sigma(\Bw_{n}^\top\Bx_{n+1}+b_{n}) \\
		\end{pmatrix}\in\R^{(n+1)\times (n+1)}.
	\end{align}
	Then $\BA$ being regular implies that for each
	$(y_i)_{i=1}^{n+1}$ exist $c$ and $(v_j)_{j=1}^{n}$ such that
	\eqref{eq:UITtoshow} holds.
	Hence, it suffices to find
	$(\Bw_j)_{j=1}^n$ and $(b_j)_{j=1}^n$ such
	that $\BA$ is regular.
	
	To do so, we proceed by induction over $k=0,\dots,n$, to show that
	there exist $(\Bw_j)_{j=1}^k$ and $(b_j)_{j=1}^k$ such that the
	first $k+1$ columns of $\BA$ are linearly independent.
	The case
	$k=0$ is trivial.
	Next let $0<k<n$ and assume that the first $k$
	columns of $\BA$ are linearly independent.
	We wish to find
	$\Bw_{k}$, $b_{k}$ such that the first $k+1$ columns are linearly
	independent.
	Suppose such $\Bw_{k}$, $b_{k}$ do not exist and
	denote by $Y_k\subseteq\R^{n+1}$ the space spanned by the first $k$
	columns of $\BA$.
	Then for all $\Bw\in\R^n$, $b\in\R$ the vector
	$(\sigma(\Bw^\top\Bx_i+b))_{i=1}^{n+1}\in\R^{n+1}$ must belong to
	$Y_k$.
	Fix
	$\By=(y_i)_{i=1}^{n+1}\in\R^{n+1}\backslash Y_k$.
	Then
	\begin{align*}
		\inf_{\tilde\Phi\in\CN_d^1(\sigma,1)}\norm[2]{(\tilde\Phi(\Bx_i))_{i=1}^{n+1}-\By}^2&=\inf_{N,\Bw_j,b_j,v_j,c}
		\sum_{i=1}^{n+1}\Big(\sum_{j=1}^N v_j\sigma(\Bw_j^\top\Bx_i+b_j)+c-y_i\Big)^2\\
		&\ge \inf_{\tilde\By\in Y_k}\norm[2]{\tilde\By-\By}^2>0.
	\end{align*}
	Since we can find a continuous function $f:\R^d\to\R$ such that
	$f(\Bx_i)=y_i$ for all $i=1,\dots,n+1$, this contradicts Theorem
	\ref{thm:universal}.
\end{proof}

\section{Optimal interpolation and reconstruction}\label{sec:OptIntLip}
Consider a bounded domain
$\Omega\subseteq\R^d$, a function $f:\Omega\to\R$, distinct points
$\Bx_1,\dots,\Bx_m\in \Omega$, and corresponding function values
$y_i\dfn f(\Bx_i)$.
Our objective is to approximate $f$ based solely
on the data pairs $(\Bx_i,y_i)$, $i=1,\dots,m$.
In this section, we
will show that, under certain assumptions on $f$, ReLU neural networks can
express an ``optimal'' reconstruction which also turns out to be an
interpolant of the data.

\subsection{Motivation}
\begin{figure}[htb]
	\centering
	\includegraphics[width = \textwidth]{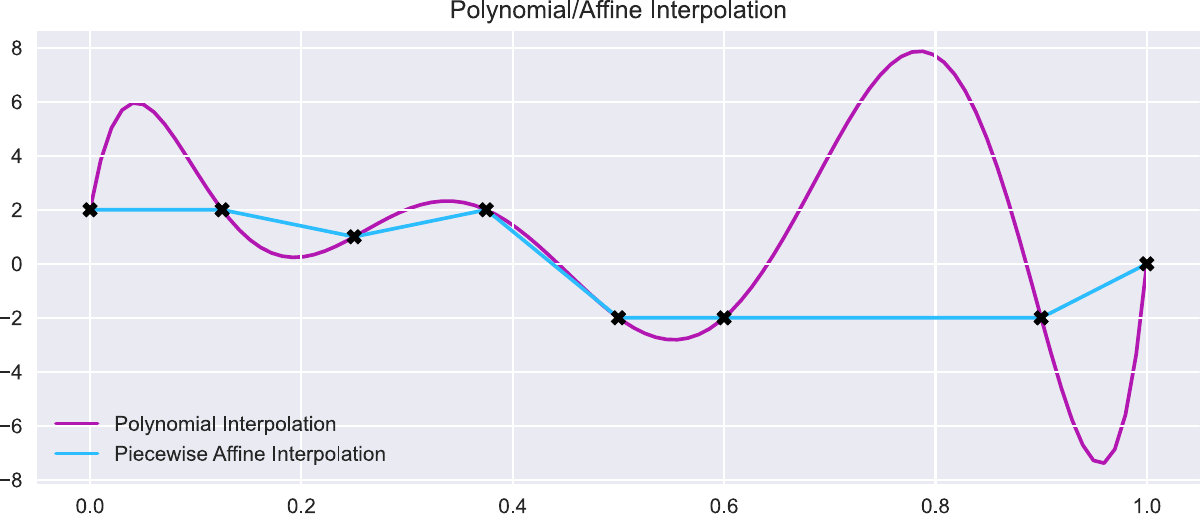}
	\caption{Interpolation of eight points by a polynomial of
		degree seven and by a piecewise affine spline.
		The
		polynomial interpolation has a significantly larger
		derivative or Lipschitz constant than the piecewise affine
		interpolator.}
	\label{fig:InterpolationGoodAndBad}	
\end{figure}

In the previous section, we observed that neural networks with
$m-1\in \N$ hidden neurons can interpolate $m$ points for every
reasonable activation function.
However, not all interpolants are
equally suitable for a given application.
For instance, consider
Figure \ref{fig:InterpolationGoodAndBad} for a comparison between
polynomial and piecewise affine interpolation on the
unit interval.

The two interpolants exhibit rather different behaviors.
In general,
there is no way of determining which constitutes a better
approximation to $f$.
In particular, given our limited information
about $f$, we cannot accurately reconstruct any additional features
that may exist between interpolation points $\Bx_1,\dots,\Bx_m$.
In
accordance with Occam's razor, it thus seems reasonable to assume that
$f$ does not exhibit extreme oscillations or behave erratically
between interpolation points.
As such, the piecewise interpolant
appears preferable in this scenario.
One way to formalize the
assumption that $f$ does not ``exhibit extreme oscillations'' is to
\emph{assume} that the Lipschitz constant
\begin{align}\label{eq:definitionOfLipschitzConstant}
	\lip(f):=\sup_{\Bx\neq \By}\frac{|f(\Bx)-f(\By)|}{\norm{\Bx-\By}}
\end{align}
of $f$ is bounded by a fixed value $M\in\R$.
Here $\norm{\cdot}$
denotes an arbitrary fixed norm on $\R^d$.

How should we choose $M$?
For every function $f:\Omega\to\R$ satisfying
\begin{align}\label{eq:interpolating}
	f(\Bx_i)=y_i\quad\text{ for all }\quad i=1,\dots,m,
\end{align}
we have
\begin{align} \label{eq:gradientCondition}
	\lip(f) = %
	\sup_{\Bx\neq \By\in\Omega} \frac{|f(\Bx) - f(\By)|}{\norm{\Bx-\By}} \geq \sup_{i \neq j} \frac{|y_i - y_j|}{\norm{\Bx_i - \Bx_j}} \eqqcolon \tilde M.
\end{align}
Because of this, we fix $M$ as a real number greater than or equal to $\tilde M$ for the remainder of our analysis.

\subsection{Optimal reconstruction for Lipschitz continuous functions}
The above considerations raise the following question: \emph{Given
	only the information that the function has Lipschitz constant at
	most $M$, what is the best reconstruction of $f$ based on the data?}
We consider here the ``best reconstruction'' to be a function that
minimizes the $L^\infty$-error in the worst case.
Specifically, with
\begin{align}\label{eq:lipM}
	\lip_M(\Omega)
	\dfn\set{f:\Omega\to\R}{\lip(f)\le M},
\end{align}
denoting the set of all functions with Lipschitz constant at most $M$,
we want to solve the following problem: Find an element in
	\begin{align}\label{eq:argminConditionBestInterpolator}
		\Phi\in
		\argmin_{h:\Omega\to\R}~\sup_{\substack{f \in \lip_M(\Omega)\\ \text{$f$ satisfies \eqref{eq:interpolating}}}} %
		~\sup_{\Bx\in\Omega}|f(\Bx)-h(\Bx)|.
	\end{align}

The next theorem shows that a function $\Phi$ as in
\eqref{eq:argminConditionBestInterpolator} indeed exists.
This $\Phi$
not only allows for an explicit formula, it also belongs to
$\lip_M(\Omega)$ and additionally interpolates the data.
Hence, it is
not just an optimal reconstruction, it is also an optimal
interpolant.
This theorem goes back to \cite{BELIAKOV200620},
which, in turn, is based on \cite{SUKHAREV197821}. 

\begin{theorem}\label{thm:optInt}
	Let  $m$, $d \in \N$, $\Omega\subseteq\R^d$, $f:\Omega\to\R$, and let $\Bx_1,\dots,\Bx_m \in \Omega$,
	$y_1,\dots,y_m\in \R$ satisfy \eqref{eq:interpolating} and \eqref{eq:gradientCondition} with $\tilde{M} > 0$.
	Further, let $M\geq \tilde{M}$. 
	
	Then %
          there exists at least one $\Phi$ in \eqref{eq:argminConditionBestInterpolator},
        which is given by
	\begin{align}\label{eq:formulaForg}
		\Phi(\Bx):=\frac{1}{2}(f_{\rm upper}(\Bx)+f_{\rm lower}(\Bx)) \qquad \text{ for } \Bx \in \Omega,
	\end{align}
	where
	\begin{align*}
		f_{\rm upper}(\Bx)&\dfn \min_{k=1,\dots,m}(y_k+M\norm{\Bx-\Bx_k})\\
		f_{\rm lower}(\Bx)&\dfn \max_{k=1,\dots,m}(y_k-M\norm{\Bx-\Bx_k}).
	\end{align*}
\end{theorem}

\begin{proof}
	First we claim that for all $h_1$, $h_2\in\lip_M(\Omega)$
	holds $\max\{h_1,h_2\}\in \lip_M(\Omega)$ as well as
	$\min\{h_1,h_2\}\in \lip_M(\Omega)$.
	Since $\min\{h_1,h_2\}=-\max\{-h_1,-h_2\}$,
	it suffices to show the claim for
	the maximum.
	We need to check that
	\begin{align}\label{eq:maxLip}
		\frac{|\max\{h_1(\Bx),h_2(\Bx)\}-\max\{h_1(\By),h_2(\By)\}|}{\norm{\Bx-\By}}\le M
	\end{align}
	for all $\Bx\neq\By\in\Omega$.
	Fix $\Bx\neq\By$.
	Without loss of
	generality we assume that
	\begin{align*}
		\max\{h_1(\Bx),h_2(\Bx)\}\ge \max\{h_1(\By),h_2(\By)\}\quad\text{and}\quad
		\max\{h_1(\Bx),h_2(\Bx)\}=h_1(\Bx).
	\end{align*}
	If $\max\{h_1(\By),h_2(\By)\}=h_1(\By)$ then the numerator in
	\eqref{eq:maxLip} equals $h_1(\Bx)-h_1(\By)$ which is bounded by
	$M\norm{\Bx-\By}$.
	If $\max\{h_1(\By),h_2(\By)\}=h_2(\By)$, then
	the numerator equals $h_1(\Bx)-h_2(\By)$ which is bounded by
	$h_1(\Bx)-h_1(\By)\le M\norm{\Bx-\By}$.
	In either case
	\eqref{eq:maxLip} holds.
	
	Clearly, $\Bx\mapsto y_k-M\norm{\Bx-\Bx_k}\in\lip_M(\Omega)$
	for each $k=1,\dots,m$ and thus
	$f_{\rm upper}$, $f_{\rm lower}\in\lip_M(\Omega)$ as well as
	$\Phi\in\lip_M(\Omega)$.
	
	Next we claim that for all $f\in\lip_M(\Omega)$ satisfying
	\eqref{eq:interpolating} holds
	\begin{align}\label{eq:fuplowbound}
		f_{\rm lower}(\Bx)\le f(\Bx)\le f_{\rm upper}(\Bx)\quad\text{ for all } \quad \Bx\in\Omega.
	\end{align}
	This is true since for every $k\in\{1,\dots,m\}$ and $\Bx\in\Omega$
	\begin{align*}
		|y_k-f(\Bx)| = |f(\Bx_k)-f(\Bx)|\le M \norm{\Bx-\Bx_k}
	\end{align*}
	so that for all $\Bx\in\Omega$
	\begin{align*}
		f(\Bx)\le \min_{k=1,\dots,m}(y_k+M\norm{\Bx-\Bx_k}),\qquad
		f(\Bx)\ge \max_{k=1,\dots,m}(y_k-M\norm{\Bx-\Bx_k}).
	\end{align*}
	
	Since
	$f_{\rm upper}$, $f_{\rm lower}\in\lip_M(\Omega)$ satisfy
	\eqref{eq:interpolating}, we conclude that for every $h:\Omega\to\R$
	holds
	\begin{align}\label{eq:optimal_lower}
		\sup_{\substack{f\in\lip_M(\Omega)\\\text{$f$ satisfies \eqref{eq:interpolating}}}}\sup_{\Bx\in\Omega}|f(\Bx)-h(\Bx)|
		&\ge \sup_{\Bx\in\Omega} \max\{|f_{\rm lower}(\Bx)-h(\Bx)|,|f_{\rm upper}(\Bx)-h(\Bx)|\}\nonumber\\
		& \ge \sup_{\Bx\in\Omega} \frac{|f_{\rm lower}(\Bx) - f_{\rm upper}(\Bx)|}{2}.
	\end{align}
	Moreover, using \eqref{eq:fuplowbound},
	\begin{align}\label{eq:optimal_upper}
		\sup_{\substack{f\in\lip_M(\Omega)\\\text{$f$ satisfies \eqref{eq:interpolating}}}}\sup_{\Bx\in\Omega}|f(\Bx)-\Phi(\Bx)|&\le
		\sup_{\Bx\in\Omega} \max\{|f_{\rm lower}(\Bx)-\Phi(\Bx)|,|f_{\rm upper}(\Bx)-\Phi(\Bx)|\} \nonumber\\
		&= \sup_{\Bx\in\Omega} \frac{|f_{\rm lower}(\Bx) - f_{\rm upper}(\Bx)|}{2}.
	\end{align}
	Finally, \eqref{eq:optimal_lower} and \eqref{eq:optimal_upper} imply
	that $\Phi$ is a solution of \eqref{eq:argminConditionBestInterpolator}.
\end{proof}

\begin{figure}[htb]
	\centering
	\includegraphics[width = \textwidth]{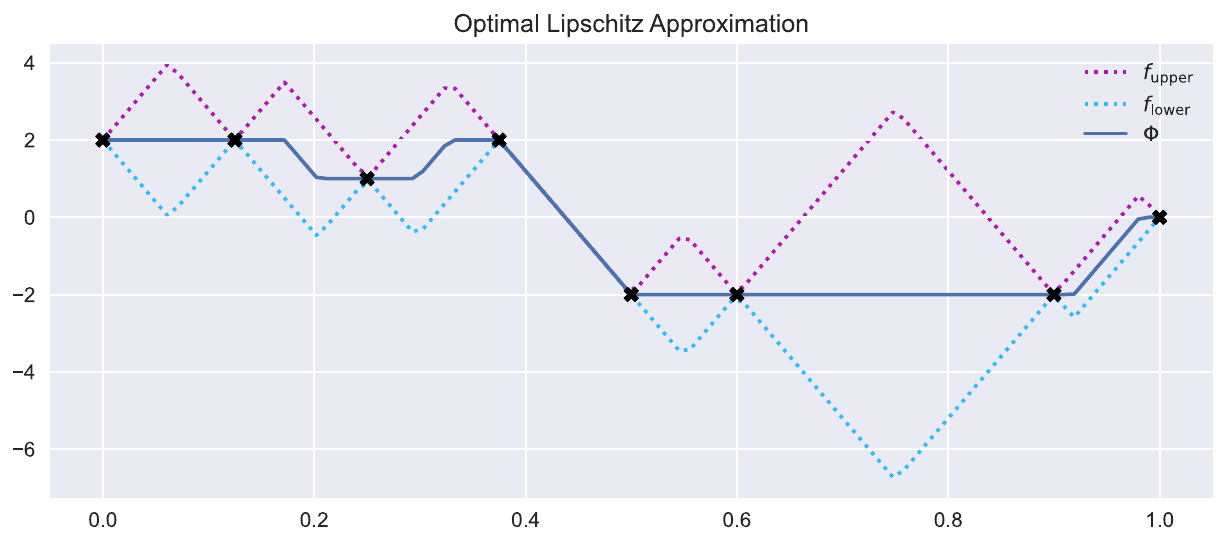}
	\caption{Interpolation of the points from Figure \ref{fig:InterpolationGoodAndBad} with the optimal Lipschitz interpolant.}
	\label{fig:InterpolationBest}
\end{figure}

\begin{figure}[htb]
	\centering
	\includegraphics[width = 0.45\textwidth]{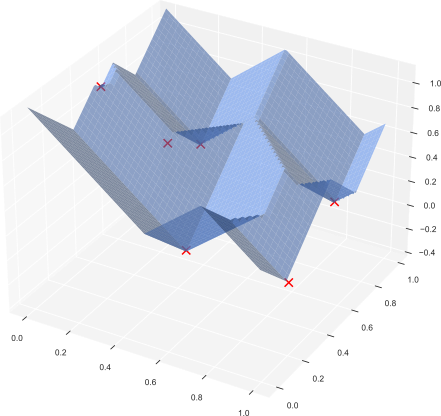} \quad \includegraphics[width = 0.45\textwidth]{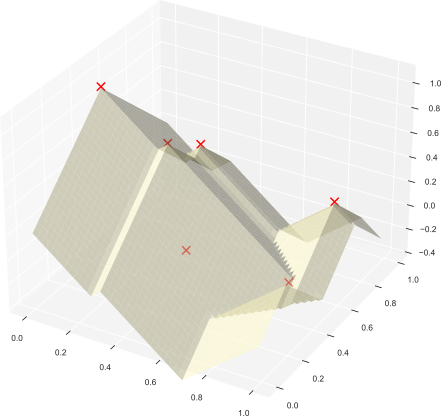}\\ 
	\includegraphics[width = 0.45\textwidth]{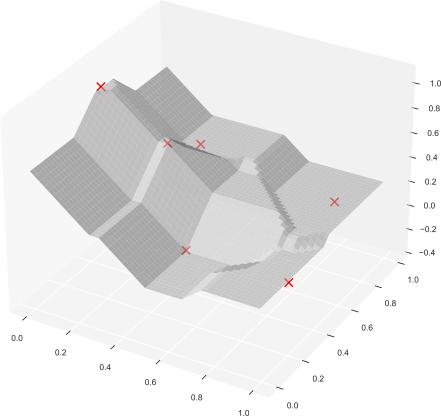}
	\caption{Two-dimensional example of the interpolation method of \eqref{eq:formulaForg}.
		From top left to bottom we see $f_{\mathrm{upper}}$, $f_{\mathrm{lower}}$, and $\Phi$.
		The interpolation points $(\Bx_i, y_i)_{i=1}^6$ are marked with red crosses.}
	\label{fig:Interpolation2D}
\end{figure}

Figure \ref{fig:InterpolationBest} depicts $f_{\rm upper}$,
$f_{\rm lower}$, and $\Phi$ for the interpolation problem shown in
Figure \ref{fig:InterpolationGoodAndBad}, while Figure
\ref{fig:Interpolation2D} provides a two-dimensional example.

\subsection{Optimal ReLU reconstructions}
So far everything was valid with an arbitrary norm on $\R^d$.
For the
next theorem, we will restrict ourselves to the $1$-norm
$\norm[1]{\Bx}=\sum_{j=1}^d|x_j|$.
Using the explicit formula of
Theorem \ref{thm:optInt}, we will now show the remarkable result that
ReLU neural networks can exactly express an optimal reconstruction (in the sense of
\eqref{eq:argminConditionBestInterpolator})
with a neural network whose size scales linearly
in the product of the dimension $d$ and the number of data points
$m$.
Additionally,
the proof is constructive, thus allowing in principle
for an explicit construction of the neural network without the
need for training.

\begin{theorem}[Optimal Lipschitz Reconstruction]\label{thm:optimalLipschitz}
	Let  $m$, $d \in \N$, $\Omega\subseteq\R^d$, $f:\Omega\to\R$, and let $\Bx_1,\dots,\Bx_m \in \Omega$,
	$y_1,\dots,y_m\in \R$ satisfy \eqref{eq:interpolating} and \eqref{eq:gradientCondition} with $\tilde{M} > 0$.
	Further, let $M\geq \tilde{M}$ and let
	$\norm{\cdot}=\norm[1]{\cdot}$ in \eqref{eq:gradientCondition} and
	\eqref{eq:lipM}.
	
	Then, there exists a ReLU neural network $\Phi \in \lip_M(\Omega)$
	that interpolates the data (i.e.\ satisfies \eqref{eq:interpolating})
	and satisfies
	\begin{align*}
          \Phi \in \argmin_{h:\Omega\to\R} \sup_{\substack{f \in \lip_M(\Omega)\\\text{$f$ satisfies \eqref{eq:interpolating}}}} \sup_{\Bx\in\Omega}|f(\Bx)-h(\Bx)|.
	\end{align*}
        Moreover, $\depth(\Phi)=O(\log(m))$, $\wdth(\Phi)=O(dm)$ and all weights of $\Phi$ are bounded in absolute value by $\max\{M,\norm[\infty]{\By}\}$.
\end{theorem}

\begin{proof}
	To prove the result, we simply need to show that the function in
	\eqref{eq:formulaForg} can be expressed as a ReLU neural network with the
	size bounds described in the theorem.
	First we notice, that there
	is a simple ReLU neural network that implements the 1-norm.
	It holds
	for all $\Bx \in \R^d$ that
	\begin{align*}
		\norm[1]{\Bx} = \sum_{i=1}^d \left(\sigma(x_i) + \sigma(-x_i)\right).
	\end{align*}
	Thus, there exists a ReLU neural network $\Phi^{\norm[1]{\cdot}}$ such that
	for all $\Bx\in\R^d$
	\begin{align*}
		\mathrm{width}(\Phi^{\norm[1]{\cdot}}) = 2d, \qquad \mathrm{depth}(\Phi^{\norm[1]{\cdot}}) = 1, \qquad \Phi^{\norm[1]{\cdot}}(\Bx) = \norm[1]{\Bx}
	\end{align*}
	As a result, there exist ReLU neural networks $\Phi_k:\R^d\to\R$,
	$k = 1, \dots, m$, such that 
	\begin{align*}
		\mathrm{width}(\Phi_k) = 2d, \qquad \mathrm{depth}(\Phi_k) = 1, \qquad 	 	\Phi_k(\Bx) = y_k + M \norm[1]{\Bx - \Bx_k}
	\end{align*}	 
	for all $\Bx \in \R^d$.
	Using the parallelization of neural 
	networks introduced in Section \ref{sec:parallel},
	there exists a ReLU neural network  $\Phi_{\rm all}\dfn
	(\Phi_1, \dots, \Phi_m)\colon \R^d \to \R^m$ such that
	\begin{align*}
		\mathrm{width}(\Phi_{\mathrm{all}}
		) &= 4md, \qquad \mathrm{depth}(\Phi_{\mathrm{all}}
                    ) = 1%
        \end{align*}
        and
        \begin{align*}
		\Phi_{\mathrm{all}}
		(\Bx) &= (y_k + M \norm[1]{\Bx - \Bx_k})_{k=1}^m \qquad \text{ for all } \Bx \in \R^d.
	\end{align*}
	Using Lemma \ref{lemma:minmaxn}, we can now find a ReLU neural network $\Phi_{\mathrm{upper}}$ such that $\Phi_{\mathrm{upper}} = f_{\mathrm{upper}}(\Bx)$ for all $\Bx \in \Omega$, $\mathrm{width}(\Phi_{\mathrm{upper}}) \leq \max\{16 m, 4md\}$, and $ \mathrm{depth}(\Phi_{\mathrm{upper}}) \leq 1 + \log(m)$.
	
	Essentially the same construction yields a ReLU neural network
	$\Phi_{\mathrm{lower}}$ with the respective properties.
	Lemma
	\ref{lemma:addition} then completes the proof.
\end{proof}

\section*{Bibliography and further reading}
The universal interpolation theorem stated in this chapter is due to \cite[Theorem 5.1]{MR1819645}. 
There exist several earlier interpolation results, which were shown under stronger assumptions:
In \cite{sartori1991simple}, the interpolation property is already linked with a rank condition on the matrix \eqref{eq:AMatrixThatShouldBeRegular}. However, no general conditions on the activation functions %
were formulated.
In \cite{itoy1996superpositionoflinearlyindependent}, the interpolation theorem is established under the assumption that the activation function $\sigma$ is continuous and nondecreasing,  $\lim_{x \to -\infty} \sigma(x) = 0$, and $\lim_{x \to \infty} \sigma(x) = 1$.  
This result was improved in \cite{huang1998upper}, which dropped the nondecreasing assumption on $\sigma$.

The main idea of the optimal Lipschitz interpolation theorem in Section \ref{sec:OptIntLip} is due to \cite{BELIAKOV200620}.
A neural network construction of Lipschitz interpolants, which however is not the optimal interpolant in the sense of \eqref{eq:argminConditionBestInterpolator},
  is given in \cite[Theorem 2.27]{jentzen2021existence}.

\newpage

\section*{Exercises}

\begin{exercise}\label{ex:1NNInTheLimit}
	Under the assumptions of Theorem \ref{thm:optInt}, we define for 
	$x \in \Omega$ the set of nearest neighbors by 
	$$
		I_x \dfn \argmin_{i = 1, \dots, m} \| x_i - x \|.
	$$
	The one-nearest-neighbor classifier $f_{\rm 1NN}$ is defined by 
	\begin{align*}
		f_{\rm 1NN}(x) = \frac{1}{2} (\min_{i\in I_x} y_i + \max_{i\in I_x} y_i).
	\end{align*}
	Let $\Phi_M$ be the function in \eqref{eq:formulaForg}. Show that for all $x \in \Omega$
	$$
	\Phi_M(x) \to f_{\rm 1NN}(x)\qquad \text{as } M \to \infty.
	$$
\end{exercise}

\begin{exercise}\label{ex:maxNormOptimalLipschitz}
  Extend Theorem \ref{thm:optimalLipschitz} to the $\norm[\infty]{\cdot}$-norm.

  \emph{Hint:} The resulting neural network will need to be deeper than the one of Theorem \ref{thm:optimalLipschitz}.
\end{exercise}

%% file: Optimization.tex
\chapter{Training of neural networks}\label{chap:training}
Up to this point, we have discussed the representation and
approximation of certain function classes using neural networks.
The
second pillar of deep learning concerns the question of how to fit a
neural network to given data, i.e., having fixed an architecture, how to find
suitable weights and biases.
This task amounts to minimizing a
so-called {\bf objective function} such as the empirical risk %
in
\eqref{eq:empiricalRiskDef0}.
Throughout this chapter we denote the
objective function by 
\begin{align*}
	\objF:\R^n\to\R,
\end{align*}
and interpret it as a function of all neural network weights and biases
collected in a vector in $\R^n$.
The goal\footnote{In reality, the goal is more nuanced: rather than merely minimizing the objective function $\objF$, we want to find a parameter $\Bw$ that yields a well-generalizing model, i.e., a small population risk, see Chapter \ref{chap:VC}. However, in this chapter we adopt the perspective of minimizing $\objF$.} is to (approximately) determine a {\bf
minimizer}, i.e., some $\Bw_*\in\R^n$ satisfying
\begin{align*}
	\objF(\Bw_*)\le \objF(\Bw)\qquad\text{for all }\Bw\in\R^n.
\end{align*}

Standard approaches primarily include variants of (stochastic) gradient descent. These are the focus of the present chapter, in which we discuss basic ideas and results in convex optimization using gradient-based algorithms.  In Sections \ref{sec:GD}, \ref{sec:SGD}, and \ref{sec:acc}, we explore gradient descent, stochastic gradient descent, and accelerated gradient descent, and provide convergence proofs for smooth and strongly convex objectives. Section \ref{sec:othertraining} discusses adaptive step sizes and explains the core principles behind popular algorithms such as Adam. Finally, Section \ref{sec:backprop} introduces the backpropagation algorithm, which enables the efficient application of gradient-based methods to neural network training.

\section{Gradient descent}\label{sec:GD}
The general idea of gradient descent is to start %
with some $\Bw_0\in\R^n$, and then apply sequential updates by moving
in the direction of \emph{steepest descent} of the objective function.
Assume for the moment that $\objF\in C^2(\R^n)$, and denote the
$k$th iterate by $\Bw_k$. Then
\begin{align}\label{eq:ThetaTaylor}
	\objF(\Bw_k+\Bv)=\objF(\Bw_k)+\Bv^\top\nabla\objF(\Bw_k)+O(\norm{\Bv}^2)\qquad \text{for }\norm{\Bv} \to 0.
\end{align}
This shows that the change in $\objF$ around $\Bw_k$ is
locally described by the gradient $\nabla\objF(\Bw_k)$. For small $\Bv$ the
  contribution of the second order term is negligible, and the
  direction $\Bv$ along which the decrease of the objective function is maximized
  equals the negative gradient $-\nabla\objF(\Bw_k)$.
Thus, $-\nabla\objF(\Bw_k)$ is also called the direction of steepest
descent. This leads to an update of the form
\begin{align}\label{eq:GD}
	\Bw_{k+1}\dfn \Bw_k-h_k \nabla\objF(\Bw_k),
\end{align}
where $h_k>0$ is referred to as the {\bf step size} or {\bf learning
rate}.
We refer to this iterative algorithm as {\bf gradient descent}.

\begin{figure}[htb]
	\centering
	\includegraphics[width = 0.49\textwidth]{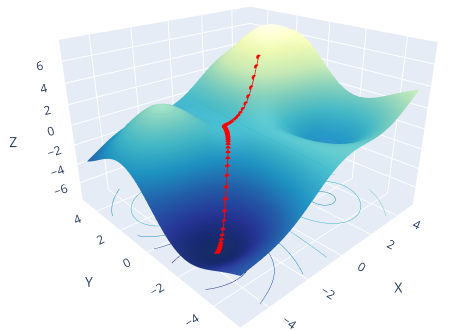} \includegraphics[width = 0.49\textwidth]{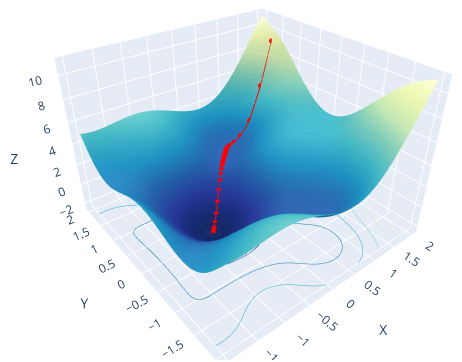}
	\caption{Two examples of gradient descent as defined in \eqref{eq:GD}.
The red points represent the $\Bw_k$. }
	\label{fig:gradientDescent}
\end{figure}

By \eqref{eq:ThetaTaylor} and \eqref{eq:GD} it holds (also see \cite[Section 1.2]{bertsekas16})
\begin{equation}\label{eq:Taylor2}
  \objF(\Bw_{k+1})=\objF(\Bw_k)-h_k\norm{\nabla\objF(\Bw_k)}^2+O(h_k^2),
\end{equation}
so that if $\nabla\objF(\Bw_k)\neq\Bnul$, a small enough step size $h_k$
ensures that the algorithm decreases the value of
the objective function. In practice,
tuning the learning rate $h_k$ can be a subtle issue as it
should strike a balance between the following dissenting requirements:
\begin{enumerate}
\item $h_k$ needs to be sufficiently small so that %
  the second-order term in
  \eqref{eq:Taylor2} is not dominating, and the update
  \eqref{eq:GD} decreases the objective function.
\item $h_k$ should be large enough to %
  ensure significant decrease of the objective function, which
  facilitates %
  faster convergence of the algorithm.
\end{enumerate}
A learning rate that is too high might overshoot the minimum, while a
rate that is too low results in slow convergence.
Common strategies
include, in particular, constant learning rates ($h_k=h$ for all
$k\in\N_0$), learning rate schedules such as decaying learning rates
($h_k\searrow 0$ as $k\to\infty$), and adaptive methods.
For adaptive
methods the algorithm dynamically adjusts $h_k$ based on the values of
$\objF(\Bw_j)$ or $\nabla\objF(\Bw_j)$ for $j\le k$, see Section
\ref{sec:othertraining} ahead.

\subsection{Structural conditions and existence of minimizers}
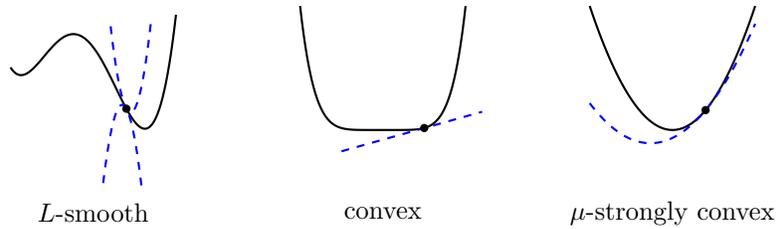
\begin{figure}
\begin{center}
\input{./plots/optimization.tex}
\end{center}
\caption{The graph of $L$-smooth functions lies between two quadratic
  functions at each point, see \eqref{eq:Lsmooth0}, the graph of convex
  function lies above the tangent at each point, see
  \eqref{eq:convex}, and the graph of $\mu$-strongly convex functions
  lies above a convex quadratic function at each point, see
  \eqref{eq:sc}.}\label{fig:optimization}
\end{figure}

We start our analysis by discussing three key notions for analyzing gradient descent algorithms, beginning with an intuitive (but loose) geometric description. A continuously differentiable objective function $\objF:\R^n\to\R$ will be called
\begin{enumerate}
\item\label{item:geosmooth} \emph{smooth} if, at each $\Bw\in\R^n$, $\objF$ is bounded above and below by a quadratic function that touches its graph at $\Bw$,
\item\label{item:geoconvex} \emph{convex} if, at each $\Bw\in\R^n$, $\objF$ lies above its tangent at $\Bw$,
\item\label{item:geosconvex} \emph{strongly convex} if, at each $\Bw\in\R^n$, $\objF$ lies above its tangent at $\Bw$ plus a convex quadratic term.
\end{enumerate}
These concepts are illustrated in Figure \ref{fig:optimization}. We next give the precise mathematical definitions.

\begin{definition}\label{def:Lsmooth}
  Let $n\in \N$ and $L>0$. A function $\objF:\R^n\to\R$ is called
  {\bf $L$-smooth} if $\objF\in C^1(\R^n)$ and
  \begin{subequations}\label{eq:Lsmooth0}
  \begin{align}\label{eq:Lsmooth}
	\objF(\Bv)&\le
	\objF(\Bw)+\dup{\nabla\objF(\Bw)}{\Bv-\Bw}
	+ \frac{L}{2}\norm{\Bw-\Bv}^2\qquad\text{for all }\Bw,\Bv\in\R^n,
                    \\ \label{eq:Lsmooth2}
                    \objF(\Bv)&\ge
	\objF(\Bw)+\dup{\nabla\objF(\Bw)}{\Bv-\Bw}
	- \frac{L}{2}\norm{\Bw-\Bv}^2\qquad\text{for all }\Bw,\Bv\in\R^n.
  \end{align}
  \end{subequations}
\end{definition}

By definition, $L$-smooth functions satisfy the geometric property
\ref{item:geosmooth}. In the literature, $L$-smoothness is
often instead defined as Lipschitz continuity of the gradient
\begin{align}\label{eq:objFLip}
	\norm{\nabla\objF(\Bw)-\nabla\objF(\Bv)}
	\le L \norm{\Bw-\Bv}\qquad\text{for all }\Bw,\Bv\in\R^n,
\end{align}
which is equivalent as the next lemma shows, e.g., \cite{bubeck,nesterov}.

\begin{lemma}\label{lemma:Lsmooth}
  Let $L>0$. Then $\objF\in C^1(\R^n)$
    is $L$-smooth if and only if \eqref{eq:objFLip} holds.
\end{lemma}

\begin{proof}
  We only show that \eqref{eq:Lsmooth0} implies \eqref{eq:objFLip}.
  The fact that \eqref{eq:objFLip} implies \eqref{eq:Lsmooth0} is left as Exercise \ref{ex:Lsmooth}.

  {\bf Step 1.} Assume first that $\objF\in C^2(\R^n)$, and that
  \eqref{eq:objFLip} does not hold.  Then we can find $\Bw\neq\Bv$
  with
  \begin{equation*}
    \norm{\Bw-\Bv} \sup_{\norm{\Be}=1}\int_0^1 \Be^\top\nabla^2\objF(\Bv+t(\Bw-\Bv))\frac{\Bw-\Bv}{\norm{\Bw-\Bv}}\dd t=\norm{\nabla\objF(\Bw)-\nabla\objF(\Bv)}
     > L \norm{\Bw-\Bv},
  \end{equation*}
  where $\nabla^2\objF\in\R^{n\times n}$ denotes the Hessian.
  Since the Hessian is symmetric,
  this implies existence of $\Bu$, $\Be\in\R^n$ with $\norm{\Be}=1$ and
  $|\Be^\top\nabla^2\objF(\Bu)\Be|>L$. Hence either
  \begin{equation}\label{eq:BenablaBe}
    \Be^\top\nabla^2\objF(\Bu)\Be>L\qquad\text{or}\qquad
    -\Be^\top\nabla^2\objF(\Bu)\Be>L.
  \end{equation}
  Assume that the first inequality holds. For $h>0$ by Taylor's formula
  \begin{equation*}
    \objF(\Bu+h\Be) = \objF(\Bu) + h\dup{\nabla\objF(\Bu)}{\Be} + \int_0^h \Be^\top\nabla^2\objF(\Bu+t\Be)\Be (h-t)\dd t.
  \end{equation*}
  Continuity of $t\mapsto \Be^\top\nabla^2\objF(\Bu+t\Be)\Be$
  and the first inequality in \eqref{eq:BenablaBe}
  imply that for $h>0$ small enough
  \begin{align*}
    \objF(\Bu+h\Be) &> \objF(\Bu) + h\dup{\nabla\objF(\Bu)}{\Be} + L\int_0^h (h-t)\dd t\\
    &= \objF(\Bu) + \dup{\nabla\objF(\Bu)}{h\Be} + \frac{L}{2}\norm{h\Be}^2.
  \end{align*}
  Hence \eqref{eq:Lsmooth} does not hold. The second case in \eqref{eq:BenablaBe} similarly leads to a violation of \eqref{eq:Lsmooth2}.

  {\bf Step 2.} Let now $\objF\in C^1$ and assume that \eqref{eq:Lsmooth0} holds. Let $\rho\in C^\infty(\R^n)$ non-negative and compactly supported with $\int_{\R^n}\rho(\Bx)\dd\Bx=1$ be a so-called mollification function (see for instance \eqref{eq:bumpfunction}). For $\eps>0$ set $\rho_\eps(\Bx)\dfn \eps^{-n}\rho(\Bx/\eps)$. It's a standard result, e.g., \cite[Appendix C.5]{evans}, that the convolution $\objF_\eps\dfn \objF*\rho_\eps\in C^\infty(\R^n)$ satisfies
  \begin{equation}\label{eq:mollification}
    \lim_{\eps\to 0} \nabla \objF_\eps(\Bv)
    =\nabla \objF(\Bv)\qquad\text{for all }\Bv\in\R^n.
  \end{equation}

  Fix $\eps>0$. By \eqref{eq:Lsmooth0}, for all $\Bv$, $\Bw\in\R^n$
  \begin{align*}
    \objF_\eps(\Bv) &= \int_{\R^n} \objF(\Bv-\Bx)\rho_\eps(\Bx)\dd\Bx\\
    &\le \int_{\R^n} \objF(\Bw-\Bx)\rho_\eps(\Bx) + \dup{\nabla\objF(\Bw-\Bx)\rho_\eps(\Bx)}{\Bv-\Bw} + \frac{L}{2}\norm{\Bw-\Bv}^2\rho_\eps(\Bx)\dd\Bx\\
    &=\objF_\eps(\Bw) +\dup{\nabla\objF_\eps(\Bw)}{\Bv-\Bw} + \frac{L}{2}\norm{\Bw-\Bv}^2,
  \end{align*}
  where we used that $\nabla\objF_\eps = \nabla\objF*\rho_\eps$. Hence $\objF_\eps$ satisfies \eqref{eq:Lsmooth}, and similarly one shows that $\objF_\eps$ satisfies \eqref{eq:Lsmooth2}. By Step 1, $\objF_\eps\in C^\infty$ satisfies \eqref{eq:objFLip}. Finally, letting $\eps\to 0$ in \eqref{eq:mollification} implies that also $\objF$ satisfies \eqref{eq:objFLip}. 
\end{proof}

\begin{definition}\label{def:convex}
Let $n \in \N$. A function $\objF:\R^n\to\R$ is called {\bf convex} if and only if
\begin{align}\label{eq:convex0}
	\objF(\lambda\Bw+(1-\lambda)\Bv)
	\le \lambda\objF(\Bw)+(1-\lambda)\objF(\Bv)
\end{align}
for all $\Bw$, $\Bv\in\R^n$, $\lambda\in (0,1)$.
\end{definition}

In case $\objF$ is continuously differentiable, this is
equivalent to the geometric property \ref{item:geoconvex} as the next
lemma shows. The proof is left as Exercise \ref{ex:convex}.
\begin{lemma}
Let $\objF\in C^1(\R^n)$. Then $\objF$ is convex if and only if
\begin{align}\label{eq:convex}
  \objF(\Bv)\ge \objF(\Bw)+\dup{\nabla\objF(\Bw)}{\Bv-\Bw} \qquad
	\text{for all }\Bw,\Bv\in\R^n.
\end{align}
\end{lemma}

  The concept of convexity is strengthened by so-called
  strong-convexity, which requires an additional positive quadratic
  term on the right-hand side of \eqref{eq:convex}, and thus
  corresponds to geometric property \ref{item:geosconvex} by
  definition.  

\begin{definition}\label{def:sc}
Let $n \in \N$ and $\mu>0$.
A differentiable function $\objF:\R^n\to\R$ is called {\bf $\mu$-strongly convex} if
\begin{align}\label{eq:sc}
	\objF(\Bv)
	\ge \objF(\Bw)+\dup{\nabla\objF(\Bw)}{\Bv-\Bw}+
	\frac{\mu}{2}\norm{\Bv-\Bw}^2
	\qquad\text{for all }\Bw,\Bv\in\R^n.
\end{align}
\end{definition}

A convex function need not be bounded from below (e.g.\ $w\mapsto w$) and thus need not have any (global) minimizers. And even if it is bounded from below, there need not exist minimizers (e.g.\ $w\mapsto \exp(w)$). However we have the following statement.
\begin{lemma}\label{lemma:unique}
Let $\objF:\R^n\to\R$. If $\objF$ is
\begin{enumerate}
\item convex, then the set of minimizers of $\objF$ is convex and has cardinality $0$, $1$, or $\infty$,
\item $\mu$-strongly convex, then $\objF$ has exactly one minimizer.
\end{enumerate}
\end{lemma}
\begin{proof}
  Let $\objF$ be convex, %
and assume that $\Bw_*$ and $\Bv_*$ are two minimizers of $\objF$. Then every convex combination $\lambda\Bw_*+(1-\lambda)\Bv_*$, $\lambda\in [0,1]$, is also a minimizer due to \eqref{eq:convex0}. This shows the first claim.

Now let $\objF$ be $\mu$-strongly convex. Then \eqref{eq:sc} implies $\objF$ to be lower bounded by a convex quadratic function. Thus $\lim_{\norm[]{\Bw}\to\infty}\objF(\Bw)=\infty$. Hence there exists a bounded sequence $(\Bw_j)_{j\in\N}$ with $\objF(\Bw_j)\to \inf_{\Bw\in\R^n}\objF(\Bw)>-\infty$. Moreover, for a subsequence we get $\Bw_{j_k}\to\Bw_*$ as $k\to\infty$. Since $\objF$ is continuous it holds $\objF(\Bw_*)=\inf_{\Bw\in\R^n}\objF(\Bw)$. Since $\Bw_*$ is a minimizer and $\objF$ is differentiable, $\nabla\objF(\Bw_*)=0$. By \eqref{eq:sc} we then have $\objF(\Bv)>\objF(\Bw_*)$ for every $\Bv\neq\Bw_*$.
\end{proof}

\subsection{Convergence of gradient descent}

As announced before, to analyze convergence, we focus on
$\mu$-strongly convex and $L$-smooth objectives only; we refer to the
bibliography section for further reading under weaker
assumptions.  %
The following theorem, which establishes linear convergence of gradient descent, is a standard result (see, e.g., \cite{nesterov,bubeck,alma99169683795301081}). The proof presented here is taken from \cite[Theorem 3.6]{2301.11235}.

  Recall that a sequence $(e_k)_{k\in\N}$ of nonnegative real numbers is said to {\bf
  converge linearly} to $0$, if and only if there exist constants
$C>0$ and $c\in [0,1)$ such that
\begin{align*}
	e_k\le Cc^k\qquad\text{for all } k\in\N_0.
\end{align*}
The constant $c$ is also referred to as the {\bf rate of convergence}.
Before giving the statement, we first note that comparing
\eqref{eq:Lsmooth} and \eqref{eq:sc} it necessarily holds $L\ge\mu$
and therefore $\kappa\dfn L/\mu\ge 1$.  This term, known as the {\bf
  condition number} of $\objF$, determines the rate of
convergence.

\begin{theorem}\label{thm:GDsc}
Let $n \in \N$ and $L\geq \mu > 0$. Let $\objF \colon \R^n \to \R$ be $L$-smooth and $\mu$-strongly convex.
Further, let $h_k=h\in (0,1/L]$ for all $k\in\N_0$, let $(\Bw_k)_{k=0}^\infty \subseteq \R^n$ be defined by \eqref{eq:GD}, and let $\Bw_*$ be the unique minimizer of $\objF$.

Then,
$\objF(\Bw_k)\to\objF(\Bw_*)$ and $\Bw_k\to\Bw_*$ converge
linearly for $k \to \infty$. For the specific choice $h=1/L$ it holds for all $k\in\N$
\begin{subequations}\label{eq:GDsc}
	\begin{align}\label{eq:GDsc1}
		\norm{\Bw_k-\Bw_*}^2&\le \Big(1-\frac{\mu}{L}\Big)^{k}\norm{\Bw_0-\Bw_*}^2\\ \label{eq:GDsc2}
		\objF(\Bw_k)-\objF(\Bw_*)&\le \frac{L}{2} \Big(1-\frac{\mu}{L}\Big)^{k}\norm{\Bw_0-\Bw_*}^2.
	\end{align}
\end{subequations}
\end{theorem}

\begin{proof}
It suffices to show \eqref{eq:GDsc1}, since \eqref{eq:GDsc2} follows
directly by %
\eqref{eq:GDsc1} and \eqref{eq:Lsmooth}
because $\nabla\objF(\Bw_*)=0$. The case $k=0$ is trivial, so let $k\in\N$.

Expanding $\Bw_{k}=\Bw_{k-1}-h\nabla\objF(\Bw_{k-1})$ and using
$\mu$-strong convexity \eqref{eq:sc} %
\begin{align}\label{eq:muscbasic}
	\norm{\Bw_{k}-\Bw_*}^2 &= \norm{\Bw_{k-1}-\Bw_*}^2-2h\dup{\nabla\objF(\Bw_{k-1})}{\Bw_{k-1}-\Bw_*} + h^2\norm{\nabla\objF(\Bw_{k-1})}^2\nonumber\\
	&\le (1-\mu h)\norm{\Bw_{k-1}-\Bw_*}^2-2h\cdot (\objF(\Bw_{k-1})-\objF(\Bw_*)) + h^2\norm{\nabla\objF(\Bw_{k-1})}^2.
\end{align}

To bound the sum of the last two terms, we first use \eqref{eq:Lsmooth} to get
\begin{align*}
	\objF(\Bw_{k})&\le
                        \objF(\Bw_{k-1}) + \dup{\nabla\objF(\Bw_{k-1})}{-h\nabla\objF(\Bw_{k-1})}+\frac{L}{2}\norm{h\nabla\objF(\Bw_{k-1})}^2\\
                      &=\objF(\Bw_{k-1})+\left(\frac{L}{2}-\frac{1}{h}\right)h^2\norm{\nabla\objF(\Bw_{k-1})}^2
\end{align*}
so that for $h< 2/L$
\begin{align*}
  h^2\norm{\nabla\objF(\Bw_{k-1})}^2&\le
  \frac{1}{1/h-L/2}(\objF(\Bw_{k-1})-\objF(\Bw_{k}))\\
  &\le
  \frac{1}{1/h-L/2}(\objF(\Bw_{k-1})-\objF(\Bw_{*})),
\end{align*}
and therefore
\begin{align*}
  &-2h\cdot (\objF(\Bw_{k-1})-\objF(\Bw_*)) + h^2\norm{\nabla\objF(\Bw_{k-1})}^2\\
  &\qquad\le  \Big(-2h+\frac{1}{1/h-L/2}\Big)(\objF(\Bw_{k-1})-\objF(\Bw_{*})).
\end{align*}
If $2h\ge 1/(1/h-L/2)$, which is equivalent to
$h\le 1/L$, then the last term is less or equal to zero.

Hence \eqref{eq:muscbasic} implies for $h\le 1/L$ 
\begin{align*}
	\norm{\Bw_{k}-\Bw_*}^2 \le (1-\mu h)\norm{\Bw_{k-1}-\Bw_*}^2\le\dots\le
  (1-\mu h)^{k} \norm{\Bw_0-\Bw_*}^2.
\end{align*}
This concludes the proof.
\end{proof}

\begin{remark}[Convex objective functions]\label{rmk:convex}
  Let $\objF \colon \R^n \to \R$ be a convex and $L$-smooth function
  with a minimizer at $\Bw_*$. As shown in Lemma~\ref{lemma:unique},
  the minimizer need not be unique, so we cannot expect
  $\Bw_k \to \Bw_*$ in general. However, the objective values still
  converge. Specifically, under these assumptions, the following holds
  \cite[Theorem~2.1.14,~Corollary~2.1.2]{nesterov}: If
  $h_k = h \in (0,2/L)$ for all $k \in \N_0$ and
  $(\Bw_k)_{k=0}^\infty \subseteq \R^n$ is generated by \eqref{eq:GD},
  then
  \begin{align*}
  \objF(\Bw_k) - \objF(\Bw_*) = O(k^{-1}) \quad \text{as } k \to \infty.
  \end{align*}
\end{remark}

\section{Stochastic gradient descent}\label{sec:SGD}
We next discuss a stochastic variant of gradient descent. The idea,
which originally goes back to Robbins and Monro \cite{RobbinsMonro},
is to replace the gradient $\nabla\objF(\Bw_{k})$ in \eqref{eq:GD} by
a random variable that we denote by $\BG_{k}$.  We interpret $\BG_k$
as an approximation to $\nabla\objF(\Bw_{k})$. %
More precisely,  throughout we assume that given $\Bw_{k}$, $\BG_k$ is an unbiased estimator of $\nabla\objF(\Bw_k)$ conditionally independent of $\Bw_0,\dots,\Bw_{k-1}$ (see Appendix \ref{sec:conddist}),
so that
\begin{align}\label{eq:Gk-1}
	\bbE[\BG_{k}|\Bw_{k}]=\bbE[\BG_{k}|\Bw_{k},\dots,\Bw_0] = \nabla\objF(\Bw_{k}).
\end{align}
After choosing some initial value $\Bw_0\in\R^n$,
the update rule becomes
\begin{align}\label{eq:sgd} 
	\Bw_{k+1}\dfn \Bw_{k}-{h}_k \BG_{k},
\end{align}
where ${h}_k>0$ denotes again the step size. Unlike in Section
\ref{sec:GD}, we focus here on $k$-dependent step sizes $h_k$.

\subsection{Motivation and decreasing learning rates}
The next example motivates the algorithm in the standard setting, e.g.\ \cite[Chapter 8]{GoodBengCour16} or \cite[Section 8.4]{pml1Book}.

\begin{example}[Empirical risk minimization]%
\label{ex:emr}
Suppose we have a data sample $S\dfn (\Bx_j,y_j)_{j=1}^{m}$, where $y_j\in\R$
is the label corresponding to the data point
$\Bx_j\in\R^d$.
Using the square loss, we wish to fit a neural network
$\Phi(\cdot,\Bw):\R^d\to\R$ depending on parameters (i.e.\ weights
and biases) $\Bw\in\R^n$, such that the empirical risk in \eqref{eq:empiricalRiskDef0}
\begin{align*}%
  \objF(\Bw)\dfn\frac{1}{m}\sum_{j=1}^{{m}}(\Phi(\Bx_j,\Bw)-y_j)^2,
\end{align*}  
is minimized. Performing one step of gradient descent requires
the computation of
\begin{align}\label{eq:fullgrad}
	\nabla\objF(\Bw)=\frac{2}{m}\sum_{j=1}^{{m}}(\Phi(\Bx_j,\Bw)-y_j)\nabla_\Bw\Phi(\Bx_j,\Bw),
\end{align}
and thus the computation of ${{m}}$ gradients of the neural network $\Phi$.
For
large ${{m}}$ (in practice ${{m}}$ can be in the millions or even larger),
this computation might be infeasible.

To reduce computational cost, we replace the full gradient \eqref{eq:fullgrad} by 
the stochastic gradient
\begin{align*}
	\BG\dfn 2(\Phi(\Bx_j,\Bw)-y_j)\nabla_\Bw\Phi(\Bx_j,\Bw)
\end{align*}
where $j\sim{\rm uniform}(1,\dots,{{m}})$ is a random variable with
uniform distribution on the discrete set $\{1,\dots,{{m}}\}$.
Then
\begin{align*}
	\bbE[\BG]=\frac{2}{m}\sum_{j=1}^{{m}}(\Phi(\Bx_j,\Bw)-y_j)\nabla_\Bw\Phi(\Bx_j,\Bw)=\nabla\objF(\Bw),
\end{align*}
meaning that $\BG$ is an unbiased estimator of $\nabla\objF(\Bw)$.
Importantly, computing (a realization of) $\BG$ merely requires a single
gradient evaluation of the neural network.

More generally, one can choose {\bf
  mini-batches}\footnote{In contrast to using the full batch, which corresponds to standard gradient descent.} of size $m_b$ (where $m_b\ll {{m}}$) and let
\begin{equation*}\BG=\frac{2}{m_b}\sum_{j\in J}(\Phi(\Bx_j,\Bw) -
  y_j)\nabla_\Bw\Phi(\Bx_j,\Bw),
\end{equation*}
where $J$ is a random subset of $\{1,\dots,{{m}}\}$ of cardinality
$m_b$. A larger mini-batch size reduces the variance of $\BG$ (thus
giving a more robust estimate of the true gradient) but increases the
computational cost.

A related common variant is the following: Let $m_bk=m$ for
$m_b$, $k$, $m\in\N$, i.e.\ the number of data points $m$ is a
$k$-fold multiple of the mini-batch size $m_b$.
In each {\bf epoch}, first a
random partition $\dot\bigcup_{i=1}^k J_i=\{1,\dots,m\}$
with $|J_i|=m_b$ for each $i$, is
determined. Then for each $i=1,\dots,k$, the weights
are updated with the gradient estimate
\begin{align*}
	\frac{2}{m_b}\sum_{j\in
	J_i}(\Phi(\Bx_j,\Bw)-y_j)\nabla_\Bw\Phi(\Bx_j,\Bw).
\end{align*}
Hence, in one epoch (corresponding to $k$ updates of the weights), the algorithm sweeps through the whole dataset, and each data point is used exactly once.
\end{example}

\begin{figure}
  \begin{center}
  \subfloat[constant learning rate for SGD]{\includegraphics[width =0.45 \textwidth]{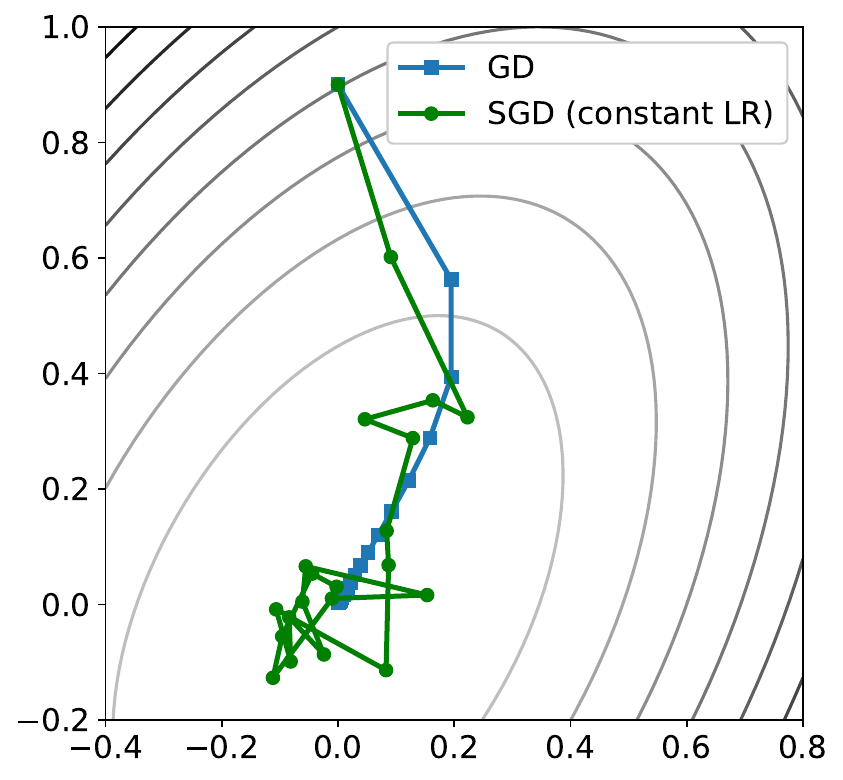}}
  \hfill
  \subfloat[decreasing learning rate for SGD]{\includegraphics[width =0.45 \textwidth]{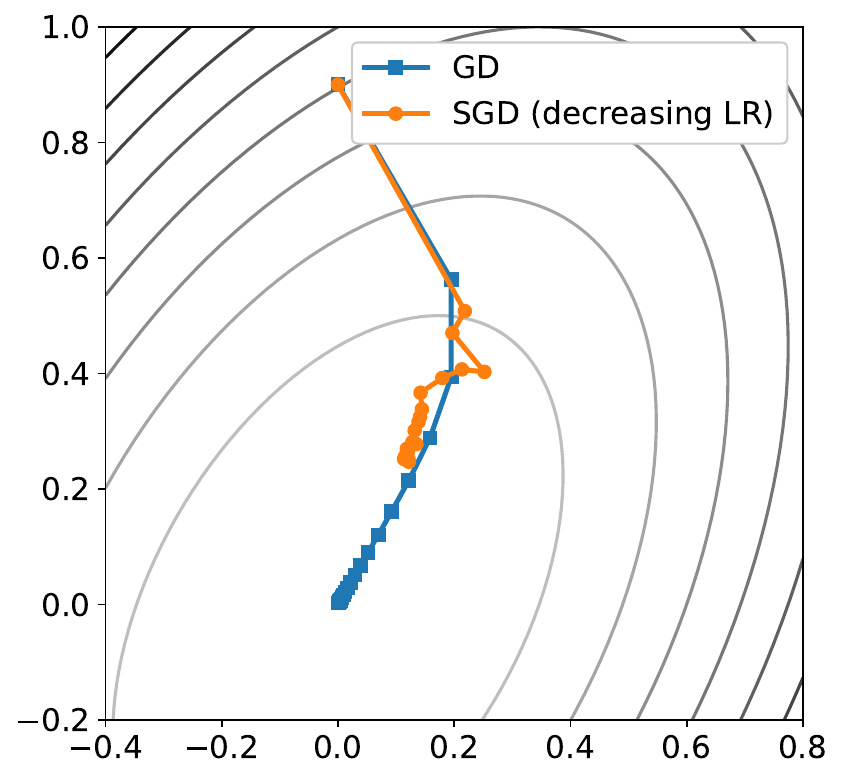}}
\end{center}
\caption{20 steps of gradient descent (GD) and stochastic gradient descent (SGD) for a strongly convex quadratic objective function. GD was computed with a constant learning rate, while SGD was computed with either a constant learning rate ($h_k=h$) or a decreasing learning rate ($h_k \sim 1/k$).}
\label{fig:gdvssgd}
\end{figure}

\begin{figure}
  \centering
  \subfloat[$\Bw_k$ far from $\Bw_*$]{\input{./plots/gd_far}}
  \hspace{0.2\textwidth}  
  \subfloat[$\Bw_k$ close to $\Bw_*$]{\input{./plots/gd_close}}
  \caption{The update vector $-h_k\BG_k$ (black) is a draw from a random variable with expectation $-h_k \nabla\objF(\Bw_k)$ (blue). In order to get convergence, the variance of the update vector should decrease as $\Bw_k$ approaches the minimizer $\Bw_*$. Otherwise, convergence will in general not hold, see Figure \ref{fig:gdvssgd} (a).}\label{fig:sgd}
 \end{figure}

 Let $\Bw_k$ be generated by \eqref{eq:sgd}. Using $L$-smoothness \eqref{eq:objFLip} we then have \cite[Lemma 4.2]{doi:10.1137/16M1080173}
 \begin{align*}
   \bbE[\objF(\Bw_{k+1})|\Bw_k]-\objF(\Bw_{k})
   &\le \bbE[\dup{\nabla\objF(\Bw_k)}{\Bw_{k+1}-\Bw_k}|\Bw_k] +\frac{L}{2}\bbE[\norm{\Bw_{k+1}-\Bw_k}^2|\Bw_k]\nonumber\\
   &= -h_k \norm{\nabla\objF(\Bw_k)}^2 + \frac{L}{2}\bbE[\norm{h_k\BG_k}^2|\Bw_k].
 \end{align*}
 For the objective function to decrease in expectation, the first term
 $h_k\norm{\nabla\objF(\Bw_k)}^2$ should dominate the second term
 $\frac{L}{2}\bbE[\norm{h_k\BG_k}^2|\Bw_k]$. As $\Bw_k$ approaches
 the minimum, we have $\norm{\nabla\objF(\Bw_k)}\to 0$, which suggests
 that $\bbE[\norm{h_k\BG_k}^2|\Bw_k]$ should also decrease over time.

 This is illustrated in Figure \ref{fig:gdvssgd} (a), which shows the
 progression of stochastic gradient descent (SGD) with a \emph{constant
 learning rate}, $h_k = h$, on a quadratic objective function and using
 artificially added gradient noise, such that
 $\bbE[\norm{\BG_k}^2|\Bw_k]$ does not tend to zero.
 The stochastic updates in \eqref{eq:sgd} cause fluctuations in the
 optimization path. Since these fluctuations do not decrease as the
 algorithm approaches the minimum, the iterates will not
 converge. Instead they stabilize within a neighborhood of the
 minimum, and oscillate around it, e.g.\ \cite[Theorem
 9.8]{2301.11235}. In practice, this might yield a good enough
 approximation to $\Bw_*$. To achieve convergence in the limit, the variance of
 the update vector, $-h_k \BG_k$, must decrease over time
 however. This can be achieved either by reducing the variance of
 $\BG_k$, for example through larger mini-batches (cf.~Example
 \ref{ex:emr}), or more commonly, by decreasing the step size $h_k$ as
 $k$ progresses. Figure \ref{fig:gdvssgd} (b) shows this for
 $h_k \sim 1/k$; also see Figure \ref{fig:sgd}.

 \subsection{Convergence of stochastic gradient descent}
 Since $\Bw_k$ in \eqref{eq:sgd} is a random variable by construction, a convergence
 statement can only be stochastic, e.g., in expectation or with high
 probability. The next theorem, which is based on
 \cite[Theorem 3.2]{pmlr-v97-qian19b} and
 \cite[Theorem 4.7]{doi:10.1137/16M1080173},
 concentrates on the former. The result
 is stated under assumption \eqref{eq:variancebound},
 which bounds the second moments of the stochastic gradients $\BG_k$ and ensures that they grow at most linearly with $\norm[]{\nabla\objF(\Bw_k)}^2$. Moreover, the analysis
 relies on the following decreasing step sizes
\begin{align}\label{eq:etak}
  {h}_k\dfn %
  \min\Big(\frac{\mu}{L^2\gamma},\frac{1}{\mu}
  \frac{(k+1)^2-k^2}{(k+1)^2}\Big)\qquad \text{for all }k\in\N_0,
\end{align}
from \cite{pmlr-v97-qian19b}. Note that $h_k=O(k^{-1})$ as $k\to\infty$, since
\begin{align}\label{eq:etakas}
  \frac{(k+1)^2-k^2}{(k+1)^2}=
  \frac{2k+1}{(k+1)^2}=\frac{2}{(k+1)}+O(k^{-2}).
\end{align}
This learning rate decay will allow us to establish a convergence rate. However, in practice, a less aggressive decay
or heuristic methods that decrease the learning rate based on the observed convergence behavior may be preferred.

\begin{theorem}%
  \label{thm:sgd}
Let $n \in \N$ and $L$, $\mu$, $\gamma > 0$. Let $\objF \colon \R^n \to \R$ be $L$-smooth and $\mu$-strongly convex. Fix $\Bw_0\in\R^n$, let $(h_k)_{k=0}^\infty$ be as in \eqref{eq:etak} and let $(\BG_{k})_{k=0}^\infty$, $(\Bw_k)_{k=1}^\infty$ be sequences of random variables as in \eqref{eq:Gk-1} and \eqref{eq:sgd}. Assume
that, for some fixed $\gamma >0$, and all $k\in\N_0$
\begin{align}\label{eq:variancebound}
	\bbE[\norm{\BG_{k}}^2|\Bw_{k}]\le\gamma (1+\norm[]{\nabla\objF(\Bw_k)}^2).
\end{align}

Then there exists a constant $C=C(\Bw_0,\gamma,\mu,L)$ such that for all $k\in\N$
\begin{align*}
	\bbE[\norm{\Bw_k-\Bw_*}^2]&\le\frac{C}{k},
                                    \nonumber\\
	\bbE[\objF(\Bw_k)]-\objF(\Bw_*)&\le\frac{C}{k}.
\end{align*}
\end{theorem}

\begin{proof}
Using \eqref{eq:Gk-1} and \eqref{eq:variancebound} it holds for $k\ge 1$
\begin{align*}
  &\bbE[\norm{\Bw_{k}-\Bw_*}^2|\Bw_{k-1}]\\
  &\quad=\norm{\Bw_{k-1}-\Bw_*}^2
	                          -2 h_{k-1}\bbE[\inp{\BG_{k-1}}{\Bw_{k-1}-\Bw_*}|\Bw_{k-1}]+h_{k-1}^2
                                          \bbE[\norm{\BG_{k-1}}^2|\Bw_{k-1}]\nonumber\\
  &\quad\le \norm{\Bw_{k-1}-\Bw_*}^2
    -2 h_{k-1}\inp{\nabla\objF(\Bw_{k-1})}{\Bw_{k-1}-\Bw_*}+h_{k-1}^2
    \gamma(1+\norm[]{\nabla\objF(\Bw_{k-1})}^2).
\end{align*}
By $\mu$-strong convexity \eqref{eq:sc}
\begin{align*}
  &-2 h_{k-1}\inp{\nabla\objF(\Bw_{k-1})}{\Bw_{k-1}-\Bw_*}\\
  &\qquad\le -\mu h_{k-1}\norm{\Bw_{k-1}-\Bw_*}^2-2h_{k-1}\cdot (\objF(\Bw_{k-1})-\objF(\Bw_*)).
\end{align*}
Moreover, $L$-smoothness, $\mu$-strong convexity and $\nabla\objF(\Bw_*)=\Bnul$ imply
\begin{equation*}
  \norm[]{\nabla\objF(\Bw_{k-1})}^2\le L^2\norm[]{\Bw_{k-1}-\Bw_*}^2
  \le \frac{2L^2}{\mu}(\objF(\Bw_{k-1})-\objF(\Bw_*)).
\end{equation*}
Combining the previous estimates we arrive at
\begin{align*}
  \bbE[\norm{\Bw_{k}-\Bw_*}^2|\Bw_{k-1}]
  \le &(1-\mu h_{k-1})\norm[]{\Bw_{k-1}-\Bw_*}^2+h_{k-1}^2\gamma \\
  &+2h_{k-1}\Big(\frac{L^2\gamma}{\mu}h_{k-1}-1\Big)(\objF(\Bw_{k-1})-\objF(\Bw_*)).
\end{align*}
The choice of $h_{k-1}\le \mu/(L^2\gamma)$ further gives
\begin{align*}
  \bbE[\norm{\Bw_{k}-\Bw_*}^2|\Bw_{k-1}]
  \le (1-\mu h_{k-1})\norm[]{\Bw_{k-1}-\Bw_*}^2+h_{k-1}^2\gamma.
\end{align*}
Note that both sides are still random variables. Taking the total expectation, and using the \emph{tower-property} (e.g., \cite[Satz 8.1.4]{klenke}), we get
\begin{align*}
  \bbE[\norm{\Bw_{k}-\Bw_*}^2]
  \le &(1-\mu h_{k-1})\bbE[\norm[]{\Bw_{k-1}-\Bw_*}^2]+h_{k-1}^2\gamma.
\end{align*}

With
$e_0\dfn \norm{\Bw_0-\Bw_*}^2$ and
$e_k\dfn \bbE[\norm{\Bw_{k}-\Bw_*}^2]$ for $k\ge 1$ we
have found
\begin{align*}
	e_{k}&\le (1-\mu h_{k-1})e_{k-1}+{h}_{k-1}^2\gamma\nonumber\\
	   &\le (1-\mu{h}_{k-1})((1-\mu{h}_{k-2})e_{k-2}+{h}_{k-2}^2\gamma)+{h}_{k-1}^2\gamma\nonumber\\
	   &\le\dots\le e_0\prod_{j=0}^{k-1}(1-\mu{h}_j)+\gamma\sum_{j=0}^{k-1}{h}_j^2\prod_{i=j+1}^{k-1}(1-\mu{h}_i).
\end{align*}
Note that there exists $k_0\in\N$ such that by \eqref{eq:etak} and \eqref{eq:etakas}
\begin{equation*}
  h_i= \frac{1}{\mu}\frac{(i+1)^2-i^2}{(i+1)^2}\quad \text{ for all } \quad i\ge k_0.
\end{equation*}
Hence there exists $\tilde C$ depending on $\gamma$, $\mu$, $L$ (but independent of $k$) such that
\begin{align*}  
	\prod_{i=j}^{k-1}(1-\mu{h}_i)
  \le \tilde C \prod_{i=j}^{k-1}\frac{i^2}{(i+1)^2}
  =\tilde C \frac{j^2}{k^2}\qquad\text{ for all }1\le j\le k
\end{align*}
and additionally
\begin{align*}  
  \prod_{i=0}^{k-1}(1-\mu{h}_i)
  \le\frac{\tilde C}{k^2}.
\end{align*}
Thus
\begin{align*}
	\bbE[\norm{\Bw_k-\Bw_*}^2]=e_k &\le \frac{\tilde C}{k^2}e_0+\tilde C
              \frac{\gamma}{\mu^2}\sum_{j=0}^{k-1} \left(\frac{(j+1)^2-j^2}{(j+1)^2}\right)^2\frac{(j+1)^2}{{k^2}}\nonumber\\
	    &\le \frac{\tilde C}{k^2}e_0+
              \frac{\tilde C\gamma}{\mu^2} \frac{1}{{k^2}}\sum_{j=0}^{k-1} \underbrace{\frac{(2j+1)^2}{(j+1)^2}}_{\le 4}\nonumber\\
	    &\le \frac{\tilde C}{k^2}e_0+
              \frac{\tilde C\gamma}{\mu^2} \frac{4k}{{k^2}}\le
              \frac{C}{k},
\end{align*}
for some $C=C(\Bw_0,\gamma,\mu,L)$.

Finally, using $L$-smoothness
\begin{align*}
	\objF(\Bw_k)-\objF(\Bw_*)\le
	\inp{\nabla\objF(\Bw_*)}{\Bw_k-\Bw_*}+\frac{L}{2}\norm{\Bw_k-\Bw_*}^2
	=\frac{L}{2}\norm{\Bw_k-\Bw_*}^2,
\end{align*}
and taking the expectation concludes the proof.
\end{proof}

The specific choice of ${h}_k$ in \eqref{eq:etak} simplifies the
calculations in the proof, but it is not necessary in order for the
asymptotic convergence to hold.
Classically, convergence of SGD can be shown under similar assumptions
as above for positive step sizes satisfying $\sum_{k\in\N} h_k=\infty$ and $\sum_{k\in\N}h_k^2<\infty$;
see for instance \cite[Section 4]{doi:10.1137/16M1080173}, \cite[Chapter 4]{books/lib/BertsekasT96},
or \cite{RobbinsMonro} for the original reference.

\section{Acceleration}\label{sec:acc}
Acceleration is an important tool for the training of neural networks
\cite{pmlr-v28-sutskever13}.
The idea was first introduced
by Polyak in 1964 under the name ``heavy ball
method'' \cite{POLYAK19641}.
It is inspired by the dynamics of a
heavy ball rolling down the valley of the loss landscape.
Since then
other types of acceleration have been proposed and analyzed, with
Nesterov acceleration being the most prominent example
\cite{nesterov83}.
In this section, we first give some intuition by
discussing the heavy ball method for a simple quadratic
loss.
Afterwards we turn to Nesterov acceleration and give a
convergence proof for $L$-smooth and $\mu$-strongly convex objective functions
that improves upon the bounds obtained for gradient descent.

\subsection{Heavy ball method}\label{sec:HB}
We follow \cite{goh2017why,Polyak,QIAN1999145} to motivate the
idea. Consider the quadratic objective function in two dimensions
\begin{align}\label{eq:2dobj}
	\objF(\Bw)\dfn \frac{1}{2}\Bw^\top \BD \Bw\qquad\text{where}\qquad
	\BD = \begin{pmatrix}
	\zeta_1 &0\\
	0 &\zeta_2
	\end{pmatrix}
\end{align}
with $\zeta_1\ge\zeta_2>0$.
Clearly, $\objF$
has a unique minimizer at $\Bw_*=\boldsymbol{0}\in\R^2$.
Starting at some $\Bw_0\in\R^2$,
gradient descent with constant step size $h>0$ computes the iterates
\begin{align*}
	\Bw_{k+1}=\Bw_k - h \BD\Bw_k = \begin{pmatrix}
	1-h\zeta_1 &0\\
	0&1-h\zeta_2
	\end{pmatrix}
	    \Bw_k
	=
	\begin{pmatrix}
	(1-h\zeta_1)^{k+1} &0\\
	0&(1-h\zeta_2)^{k+1}
	\end{pmatrix} \Bw_0.
\end{align*}
The method converges for arbitrary initialization $\Bw_0$ if and only if
\begin{align*}
	|1-h\zeta_1|<1\qquad\text{and}\qquad
	|1-h\zeta_2|<1.
\end{align*}
The optimal step size balancing %
the rate of convergence in both
coordinates is
\begin{align}\label{eq:opth}
	h_* = \argmin_{h>0} \max\{|1-h\zeta_1|,|1-h\zeta_2|\} = \frac{2}{\zeta_1+\zeta_2}.
\end{align} 
With $\kappa=\zeta_1/\zeta_2$ we then obtain the convergence rate
\begin{align}\label{eq:optconv}
	|1-h_*\zeta_1| = |1-h_*\zeta_2| = \frac{\zeta_1-\zeta_2}{\zeta_1+\zeta_2}=\frac{\kappa-1}{\kappa+1}\in [0,1).
\end{align}
If $\zeta_1\gg \zeta_2$, this term is close to $1$,
and thus the convergence will be slow.
This is consistent with our analysis for strongly convex objective functions;
by Exercise \ref{ex:quadraticobjective} the condition
number of $\objF$ equals $\kappa=\zeta_1/\zeta_2\gg 1$.
Hence, the
upper bound in Theorem \ref{thm:GDsc} %
converges only slowly.
Similar considerations hold for general quadratic objective functions
in $\R^n$ such as
\begin{align}\label{eq:quadraticobjective}
	\tilde\objF(\Bw)=\frac{1}{2}\Bw^\top \BA\Bw+\Bb^\top\Bw+c
\end{align}
with $\BA\in\R^{n\times n}$ symmetric positive definite,
$\Bb\in\R^n$ and $c\in\R$, see Exercise \ref{ex:hbm}.

\begin{remark}
Interpreting \eqref{eq:quadraticobjective} as a second-order Taylor
approximation
of some objective function %
around its minimizer,
we note that the %
described effects also occur for
general objective functions with ill-conditioned Hessians at the minimizer.
\end{remark}

\begin{figure}
\begin{center}
\includegraphics[scale=0.4]{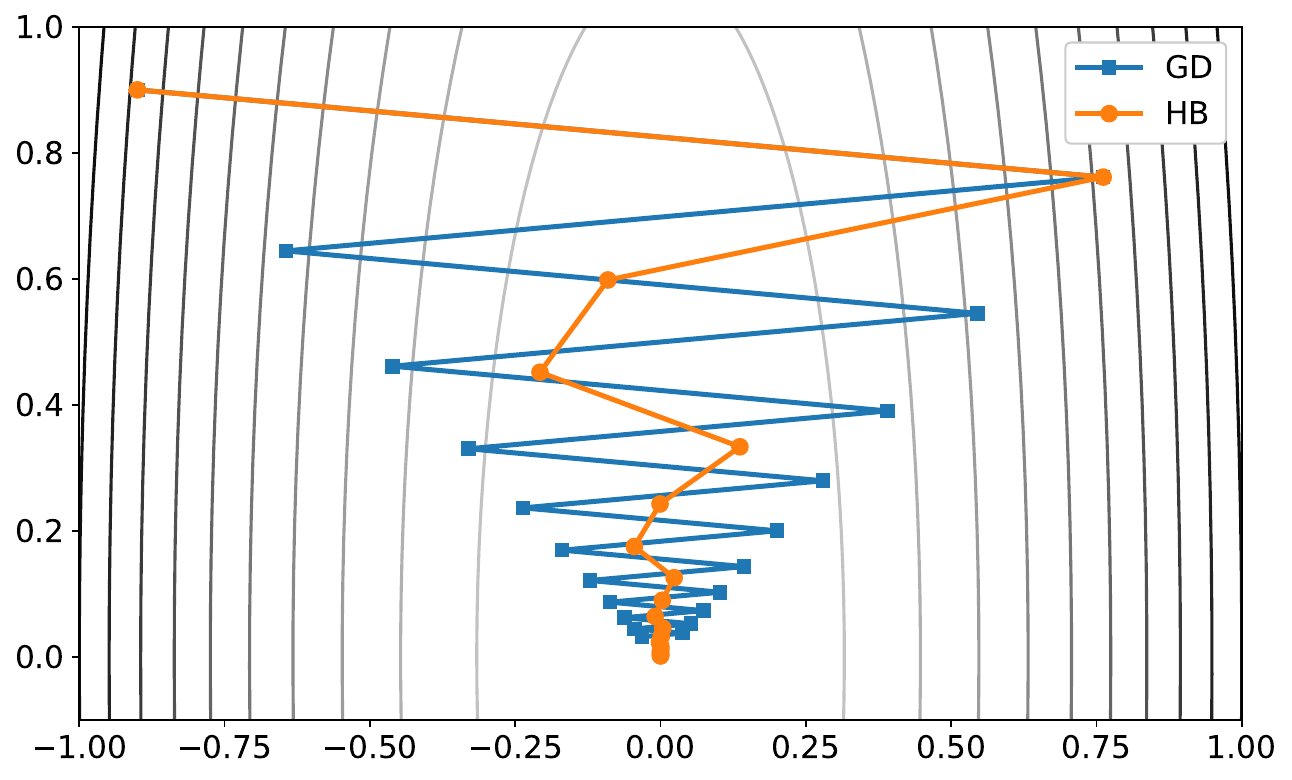}
\end{center}
\caption{20 steps of gradient descent (GD) and the heavy
ball method (HB) on the objective function \eqref{eq:2dobj} with
$\zeta_1=12\gg 1=\zeta_2$, step size $h=\alpha=h_*$ as in
\eqref{eq:opth}, and $\beta=1/3$. Figure based on \cite[Fig.~6]{Polyak}.}
\label{fig:hbm}
\end{figure}

Figure \ref{fig:hbm} gives further insight into the poor performance
of gradient descent for \eqref{eq:2dobj} with $\zeta_1\gg\zeta_2$.
The loss-landscape looks like a ravine (the derivative is much larger
in one direction than the other), and away from the floor,
$\nabla\objF$ mainly points to the opposite side.
Therefore the
iterates oscillate back and forth in the first coordinate, and make
little progress in the direction of the valley along the second
coordinate axis.
To address this problem, the heavy ball method
introduces a ``momentum'' term which can mitigate this effect to some
extent.
The idea is to choose the update not just according to the
gradient at the current location, but to add information from the
previous steps. After initializing $\Bw_0$ and, e.g.,
$\Bw_1=\Bw_0-\alpha\nabla\objF(\Bw_0)$, let for $k\in \N$

\begin{align}\label{eq:HBM}
	\Bw_{k+1} = \Bw_k - \alpha\nabla\objF(\Bw_k)+\beta(\Bw_k-\Bw_{k-1}).
\end{align}
This is known as Polyak's heavy ball method \cite{POLYAK19641,Polyak}.
Here
$\alpha>0$ and $\beta\in (0,1)$ are hyperparameters (that could also
depend on $k$) and in practice need to be carefully tuned to balance
the strength of the gradient and the momentum term.
Iteratively
expanding \eqref{eq:HBM} with the given initialization, observe that
for $k\ge 0$
\begin{align}\label{eq:HBM2}
	\Bw_{k+1}
	=\Bw_k-\alpha\Bigg(\sum_{j=0}^k \beta^j\nabla\objF(\Bw_{k-j})\Bigg).
\end{align}
Thus, $\Bw_k$ is updated using an \emph{exponentially weighted moving average}
of all past gradients.
Choosing the momentum parameter $\beta$ in
the interval $(0,1)$ ensures that the influence of previous gradients
on the update decays exponentially.
The concrete value of $\beta$
determines the balance between the impact of recent and past
gradients.

Intuitively, %
this %
linear combination of the past gradients
averages out some of the oscillation observed for gradient descent in
Figure \ref{fig:hbm}; %
the update vector is strengthened in directions where
past gradients are aligned (the second coordinate axis), while it is dampened
in directions where the gradients' signs alternate (the first coordinate
axis). Similarly, when using stochastic gradients, it
can help to reduce some of the variance.

As mentioned earlier, the heavy ball method can be interpreted as a
discretization of the dynamics of a ball rolling down the valley of
the loss landscape.
If the ball has positive mass, i.e.\ is
``heavy'', its momentum prevents the ball from bouncing back and forth
too strongly.
The following remark
elucidates this connection.

\begin{remark}
  As pointed out, e.g., in \cite{Polyak,QIAN1999145}, for suitable
  choices of $\alpha$ and $\beta$, \eqref{eq:HBM} can be interpreted
  as a discretization of the second-order ODE
\begin{align}\label{eq:AGDflow}
	m \Bw''(t) = -\nabla\objF(\Bw(t)) - r\Bw'(t).
\end{align}
This equation describes the movement of a point mass $m$ under
influence of the force field $-\nabla\objF(\Bw(t))$; the term
$-\Bw'(t)$, which points in the negative direction of the current
velocity, corresponds to friction, and $r>0$ is the friction
coefficient.
The discretization
\begin{align*}
	m\frac{\Bw_{k+1}-2\Bw_k+\Bw_{k-1}}{h^2} = -\nabla\objF(\Bw_k)-r\frac{\Bw_{k+1}-\Bw_k}{h}
\end{align*}
then leads to
\begin{align}\label{eq:AGDflowdisc}
	\Bw_{k+1} = \Bw_k-\underbrace{\frac{h^2}{m+rh}}_{=\alpha}\nabla\objF(\Bw_k)+\underbrace{\frac{m}{m+rh}}_{=\beta}(\Bw_k-\Bw_{k-1}),
\end{align}
and thus to \eqref{eq:HBM}, \cite{QIAN1999145}.

Letting $m=0$ in \eqref{eq:AGDflowdisc}, we recover the gradient
descent update \eqref{eq:GD}.  Hence, positive mass $m>0$ corresponds
to the momentum term. The gradient descent update in turn can be
interpreted as an Euler discretization of the gradient flow
\begin{equation}\label{eq:GDflow}
  \Bw'(t)=-\nabla\objF(\Bw(t)).
\end{equation}
Note that $-\nabla\objF(\Bw(t))$ represents the \emph{velocity} of
$\Bw(t)$ in \eqref{eq:GDflow}, whereas in \eqref{eq:AGDflow}, up to
the friction term, it corresponds to an \emph{acceleration}.
\end{remark}

\subsection{Nesterov acceleration}
Nesterov's accelerated gradient method (NAG)
\cite{nesterov83,nesterov}
builds on the heavy ball
method.
After initializing $\Bw_0$, $\Bv_0\in\R^n$, the update is
formulated for $k\ge 0$ as the two-step process
\begin{subequations}\label{eq:NAG}
  \begin{align}
    \Bw_{k+1} &= \Bv_k - \alpha\nabla\objF(\Bv_k)\\
    \Bv_{k+1} &= \Bw_{k+1} + \beta (\Bw_{k+1}-\Bw_{k})
\end{align}
\end{subequations}
where again $\alpha>0$ and $\beta\in (0,1)$ are hyperparameters.
Substituting
the second line into the first we get for $k\ge 1$
\begin{align*}
	\Bw_{k+1} = \Bw_k -\alpha\nabla\objF(\Bv_k) + \beta(\Bw_k-\Bw_{k-1}).
\end{align*}
Comparing with the heavy ball method \eqref{eq:HBM}, the key difference is
that the gradient is not evaluated at the current position $\Bw_k$,
but instead at the point $\Bv_k=\Bw_k+\beta(\Bw_k-\Bw_{k-1})$, which
can be interpreted as an estimate of the position at the next
iteration. This improves stability and robustness
  with respect to hyperparameter settings,
  see \cite[Sections 4 and 5]{MR3440187}.

We now discuss the convergence of NAG
for $L$-smooth
and $\mu$-strongly convex objective functions $\objF$.
To give the analysis, it
is convenient to first rewrite %
\eqref{eq:NAG}
as a three sequence update:
Let $\tau=\sqrt{\mu/L}$, $\alpha=1/L$, and
$\beta = (1-\tau)/(1+\tau)$.
After initializing $\Bw_0$, $\Bv_0\in\R^n$, \eqref{eq:NAG} can also be
written as $\Bu_0 = ((1+\tau)\Bv_0-\Bw_0)/\tau$ and
for $k\ge 0$
\begin{subequations}\label{eq:NAG3}
\begin{align}\label{eq:nag3x}
	\Bv_k &= \frac{\tau}{1+\tau}\Bu_k+\frac{1}{1+\tau}\Bw_k\\\label{eq:nag3y}
	\Bw_{k+1} &=\Bv_k - \frac{1}{L}\nabla\objF(\Bv_k)\\\label{eq:nag3z}
	\Bu_{k+1}&=\Bu_k+\tau\cdot (\Bv_k-\Bu_k)-\frac{\tau}{\mu}\nabla\objF(\Bv_k),
\end{align}
\end{subequations}
see Exercise \ref{ex:threeterm}.
The proof of the next theorem proceeds along the lines of \cite[Theorem A.3.1]{pmlr-v70-tu17a}, \cite[Proposition 10]{wilson}; also see \cite[Proposition 20]{Weissmann2022Optimization} who present a similar proof of a related result based on the same references.

\begin{theorem}\label{thm:AGDsc}
  Let $n \in \N$, $0<\mu\le L$, and let $\objF \colon \R^n \to \R$ be $L$-smooth and $\mu$-strongly convex.
  Further, let $\Bw_0$, $\Bv_0\in\R^n$ and let $\tau=\sqrt{\mu/L}$. Let $(\Bv_k,\Bw_{k+1},\Bu_{k+1})_{k=0}^\infty \subseteq \R^n$ be defined by \eqref{eq:nag3x}, and let $\Bw_*$ be the unique minimizer of $\objF$.

Then, for all $k\in\N_0$, it holds that
\begin{subequations}\label{eq:AGD}
\begin{align}\label{eq:AGD1}
	\norm{\Bu_k-\Bw_*}^2 & \le \frac{2}{\mu} \Big(1-\sqrt{\frac{\mu}{L}}\Big)^{k}\Big(\objF(\Bw_0)-\objF(\Bw_*)+\frac{\mu}{2}\norm{\Bu_0-\Bw_*}^2\Big),\\ \label{eq:AGD2}
	\objF(\Bw_k)-\objF(\Bw_*)&\le \Big(1-\sqrt{\frac{\mu}{L}}\Big)^{k}\Big(\objF(\Bw_0)-\objF(\Bw_*)+\frac{\mu}{2}\norm{\Bu_0-\Bw_*}^2\Big).
\end{align}
\end{subequations}
\end{theorem}

\begin{proof}
Define
\begin{align}\label{eq:ek}
	e_k\dfn (\objF(\Bw_k)-\objF(\Bw_*))+\frac{\mu}{2}\norm{\Bu_k-\Bw_*}^2.
\end{align}
To show \eqref{eq:AGD}, it suffices to prove with $c\dfn 1-\tau$
that $e_{k+1}\le ce_k$ for all $k\in\N_0$.

{\bf Step 1.} We bound the first term in $e_{k+1}$ defined in \eqref{eq:ek}.
Using $L$-smoothness \eqref{eq:Lsmooth} and \eqref{eq:nag3y}
\begin{align*}
	\objF(\Bw_{k+1})-\objF(\Bv_k) \le \dup{\nabla\objF(\Bv_k)}{\Bw_{k+1}-\Bv_k} + \frac{L}{2}\norm{\Bw_{k+1}-\Bv_k}^2
	= -\frac{1}{2L}\norm{\nabla\objF(\Bv_k)}^2.
\end{align*}
Thus, since $c+\tau=1$,
\begin{align}\label{eq:agdsecond}
  \objF(\Bw_{k+1})-\objF(\Bw_*)&\le (\objF(\Bv_k)-\objF(\Bw_*))-\frac{1}{2L}\norm{\nabla\objF(\Bv_k)}^2\nonumber\\
                               &= c\cdot (\objF(\Bw_k)-\objF(\Bw_*))
+c\cdot (\objF(\Bv_k)-\objF(\Bw_k))\nonumber\\
  &\quad +\tau\cdot (\objF(\Bv_k)-\objF(\Bw_*))
  -\frac{1}{2L}\norm{\nabla\objF(\Bv_k)}^2.
\end{align}

{\bf Step 2.} We bound the second term in $e_{k+1}$ defined in \eqref{eq:ek}.
By \eqref{eq:nag3z}
\begin{align}\label{eq:ek1}
	&\frac{\mu}{2}\norm{\Bu_{k+1}-\Bw_*}^2 - \frac{\mu}{2}\norm{\Bu_{k}-\Bw_*}^2\nonumber\\
	&\quad = \frac{\mu}{2} \norm{\Bu_{k+1}-\Bu_k + \Bu_k-\Bw_*}^2- \frac{\mu}{2}\norm{\Bu_{k}-\Bw_*}^2\nonumber\\
	&\quad = \frac{\mu}{2}\norm{\Bu_{k+1}-\Bu_k}^2+\mu \dup{\tau\cdot (\Bv_k-\Bu_k)-\frac{\tau}{\mu}\nabla\objF(\Bv_k)}{\Bu_k-\Bw_*}\nonumber\\
	&\quad= \frac{\mu}{2}\norm{\Bu_{k+1}-\Bu_k}^2 +\tau\dup{\nabla\objF(\Bv_k)}{\Bw_*-\Bu_k}-\tau\mu\dup{\Bv_k-\Bu_k}{\Bw_*-\Bu_k}. 
\end{align}
Using $\mu$-strong
convexity \eqref{eq:sc}, we get
\begin{align*}
	&\tau\dup{\nabla\objF(\Bv_k)}{\Bw_*-\Bu_k}=
	                                            \tau\dup{\nabla\objF(\Bv_k)}{\Bv_k-\Bu_k}+
	                                            \tau\dup{\nabla\objF(\Bv_k)}{\Bw_*-\Bv_k}\\
	&\qquad\qquad\le \tau\dup{\nabla\objF(\Bv_k)}{\Bv_k-\Bu_k} - \tau\cdot (\objF(\Bv_k)-\objF(\Bw_*))-\frac{\tau\mu}{2}\norm{\Bv_k-\Bw_*}^2.
\end{align*}
Moreover,
\begin{align*}
	&-\frac{\tau\mu}{2}\norm{\Bv_k-\Bw_*}^2-\tau \mu\dup{\Bv_k-\Bu_k}{\Bw_*-\Bu_k}\\
	&\qquad\qquad=-\frac{\tau\mu}{2}\Big(\norm{\Bv_k-\Bw_*}^2-2\dup{\Bv_k-\Bu_k}{\Bv_k-\Bw_*}+2\dup{\Bv_k-\Bu_k}{\Bv_k-\Bu_k}\Big)\\
	&\qquad\qquad=-\frac{\tau\mu}{2}(\norm{\Bu_k-\Bw_*}^2+\norm{\Bv_k-\Bu_k}^2).
\end{align*}
Thus, \eqref{eq:ek1} is bounded by
\begin{align*}
	    &\frac{\mu}{2}\norm{\Bu_{k+1}-\Bu_k}^2 + \tau\dup{\nabla\objF(\Bv_k)}{\Bv_k-\Bu_k}-\tau\cdot (\objF(\Bv_k)-\objF(\Bw_*))\nonumber\\
	&
	  -\frac{\tau\mu}{2}\norm{\Bu_k-\Bw_*}^2-\frac{\tau\mu}{2}\norm{\Bv_k-\Bu_k}^2.
\end{align*}
From \eqref{eq:nag3x} we have $\tau\cdot (\Bv_k-\Bu_k)=\Bw_k-\Bv_k$, so that with $c=1-\tau$ we arrive at
\begin{align}\label{eq:agdfirst}
	&\frac{\mu}{2}\norm{\Bu_{k+1}-\Bw_*}^2
	\le c\frac{\mu}{2}\norm{\Bu_{k}-\Bw_*}^2+\frac{\mu}{2}\norm{\Bu_{k+1}-\Bu_k}^2\nonumber\\
	&\qquad
	+ \dup{\nabla\objF(\Bv_k)}{\Bw_k-\Bv_k}-\tau\cdot (\objF(\Bv_k)-\objF(\Bw_*))
	-\frac{\mu}{2\tau}\norm{\Bw_k-\Bv_k}^2.
\end{align}

{\bf Step 3.} We show $e_{k+1}\le c e_k$. Adding \eqref{eq:agdsecond} and \eqref{eq:agdfirst} gives
\begin{align*}
  e_{k+1} &\le ce_k +c\cdot (\objF(\Bv_k)-\objF(\Bw_k))-\frac{1}{2L}\norm{\nabla\objF(\Bv_k)}^2+ \frac{\mu}{2}\norm{\Bu_{k+1}-\Bu_k}^2\nonumber\\
	&\quad+ \dup{\nabla\objF(\Bv_k)}{\Bw_k-\Bv_k}
	  -\frac{\mu}{2\tau}\norm{\Bw_k-\Bv_k}^2.
\end{align*}
Using \eqref{eq:nag3x}, \eqref{eq:nag3z} we expand
\begin{align*}
  \frac{\mu}{2}\norm{\Bu_{k+1}-\Bu_k}^2 &= \frac{\mu}{2}\normc{\Bw_k-\Bv_k-\frac{\tau}{\mu}\nabla\objF(\Bv_k)}^2\\
  &= \frac{\mu}{2}\norm{\Bw_k-\Bv_k}^2
	-\tau\dup{\nabla\objF(\Bv_k)}{\Bw_k-\Bv_k}+\frac{\tau^2}{2\mu}\norm{\nabla\objF(\Bv_k)}^2,
\end{align*}
to obtain
\begin{align*}
	e_{k+1}&\le c e_k +\Big(\frac{\tau^2}{2\mu}-\frac{1}{2L}\Big)\norm{\nabla\objF(\Bv_k)}^2 -\frac{\mu}{2\tau}\norm{\Bw_k-\Bv_k}^2\\
  &\quad+c\cdot \big(\objF(\Bv_k)-\objF(\Bw_k)+\dup{\nabla\objF(\Bv_k)}{\Bw_k-\Bv_k}\big)+\frac{\mu}{2}\norm{\Bw_k-\Bv_k}^2.
\end{align*}
The last line can be bounded using $\mu$-strong convexity \eqref{eq:sc} and $\mu\le L$
\begin{align*}
  &c\cdot(f(\Bv_k)-f(\Bw_k)+\dup{\nabla\objF(\Bv_k)}{\Bw_k-\Bv_k})+\frac{\mu}{2}\norm{\Bv_k-\Bw_k}^2\\
  &\qquad\le -(1-\tau)\frac{\mu}{2}\norm{\Bv_k-\Bw_k}^2+\frac{\mu}{2}\norm{\Bv_k-\Bw_k}^2
    \le \frac{\tau L}{2}\norm{\Bv_k-\Bw_k}^2.
\end{align*}
In all
\begin{align*}
  e_{k+1}\le ce_k+
  \Big(\frac{\tau^2}{2\mu}-\frac{1}{2L}\Big)\norm{\nabla\objF(\Bv_k)}^2 +\Big(\frac{\tau L}{2}-\frac{\mu}{2\tau}\Big)\norm{\Bw_k-\Bv_k}^2= c e_k,
\end{align*}
where the terms in brackets vanished since $\tau=\sqrt{\mu/L}$.
This concludes the proof.
\end{proof}

Comparing the result for gradient descent \eqref{eq:GDsc} with NAG
\eqref{eq:AGD}, the improvement for strongly convex objectives
lies in the convergence rate, which is
$1-\kappa^{-1}$ for gradient descent\footnote{Also see \cite[Theorem
  2.1.15]{nesterov} for the sharper rate
  $(1-\kappa^{-1})^2/(1+\kappa^{-1})^2=1-4\kappa^{-1}+O(\kappa^{-2})$.}, and $1-\kappa^{-1/2}$ for NAG, where
$\kappa=L/\mu$. %
For NAG the
convergence rate depends only on the \emph{square root} of the condition
number $\kappa$. For ill-conditioned problems where $\kappa$ is large,
we therefore expect much better performance for accelerated methods.

\section{Adaptive and coordinate-wise learning rates}\label{sec:othertraining}
In recent years, a multitude of first order (gradient descent) methods
has been proposed and studied for the training of neural networks.
Many of them incorporate some or all of the following key strategies:
stochastic gradients, acceleration, and adaptive step sizes.
The
concept of stochastic gradients and acceleration have been covered in
the Sections \ref{sec:SGD} and \ref{sec:acc}, and we will touch upon adaptive learning rates
in the present one. Specifically, following the original papers
\cite{adagrad,adadelta,rmsprop,adam} and in particular the overviews
\cite[Section 8.5]{GoodBengCour16},
\cite{1609.04747}, \cite[Chapter 11]{geron2017hands-on},
and \cite[Section 8.4]{pml1Book},
we
explain the main ideas behind AdaGrad, RMSProp, and Adam. The %
above references provide
intuitive general overviews %
including
several additional variants that are omitted
here. Moreover, in practice, various other techniques and heuristics
such as batch normalization, gradient clipping, regularization and
dropout, early stopping, specific weight initializations etc.\ are
used. We do not discuss them here, and refer for example to \cite{Bottou2012,GoodBengCour16,geron2017hands-on,prince2023understanding,pml1Book}.

\subsection{Coordinate-wise scaling}
In Section \ref{sec:HB}, we saw why plain gradient descent can be
inefficient for ill-conditioned objective functions. This issue can be
particularly pronounced in high-dimensional optimization problems,
such as when training neural networks, where certain %
parameters influence the network output much more than others. As a
result, a single learning rate may be suboptimal; directions in
parameter space with small gradients are updated too slowly, while in
directions with large gradients the algorithm might overshoot. To
address this, one approach is to precondition the gradient by
multiplying it with a matrix that accounts for the geometry of the
parameter space, e.g.\ \cite{Amari98,nocedal06}.
A simpler and computationally efficient alternative is to scale each component of the gradient individually, corresponding to a diagonal preconditioning matrix. This allows different learning rates for different coordinates and can help mitigate ill-conditioning, but only if the ill-conditioning is aligned with the coordinate axes.
The key question is how to set the
learning rates. The main idea, first proposed in \cite{adagrad}, is to
scale each component inverse proportional to the magnitude of past
gradients. In the words of the authors of \cite{adagrad}: ``Our
procedures give frequently occurring features very low learning rates
and infrequent features high learning rates.''

After initializing $\Bu_{0}=\Bnul\in\R^n$, $\Bs_{0}=\Bnul\in\R^n$, and
$\Bw_0\in\R^n$, all methods discussed below are special cases
of
\begin{subequations}\label{eq:adamgeneral}
\begin{align}
  \Bu_{k+1} &= \beta_1\Bu_{k}+\beta_2\nabla\objF(\Bw_{k})\label{eq:ag1}\\ 
  \Bs_{k+1} &= \gamma_1\Bs_{k}+\gamma_2\nabla\objF(\Bw_{k})\odot\nabla\objF(\Bw_{k})\label{eq:ag2}\\ 
	\Bw_{k+1} &= \Bw_{k}-\alpha_k\Bu_{k+1}\oslash \sqrt{\Bs_{k+1}+\eps}\label{eq:ag3}
\end{align}
\end{subequations}
for $k\in\N_0$, and certain hyperparameters $\alpha_k$, $\beta_1$,
$\beta_2$, $\gamma_1$, $\gamma_2$, and $\eps$.  Here $\odot$ and
$\oslash$ denote the componentwise (Hadamard) multiplication and
division, respectively, and $\sqrt{\Bs_{k+1}+\eps}$ is understood as
the vector $(\sqrt{v_{k+1,i}+\eps})_i$. Equation \eqref{eq:ag1}
defines an update vector and corresponds to heavy ball momentum if
$\beta_1\in (0,1)$.  If $\beta_1=0$, then $\Bu_{k+1}$ is simply a
multiple of the current gradient. Equation \eqref{eq:ag2} defines a
scaling vector $\Bs_{k+1}$ that is used to set a coordinate-wise
learning rate of the update vector in \eqref{eq:ag3}. The constant
$\eps>0$ is chosen small but positive to avoid division by zero in
\eqref{eq:ag3}.  These type of methods are often applied using
mini-batches, see Section \ref{sec:SGD}. For simplicity we present
them with the full gradients.

\begin{example}
  Consider an objective function $\objF:\R^n\to\R$,
  and its rescaled version
  \begin{equation*}
    \objF_\Bzeta(\Bw)\dfn \objF(\Bw\odot\Bzeta)\qquad\text{with gradient}\qquad
    \nabla \objF_\Bzeta(\Bw) = \Bzeta\odot \nabla \objF (\Bw\odot \Bzeta),
  \end{equation*}
  for some $\Bzeta\in (0,\infty)^n$. Gradient descent \eqref{eq:GD} applied to $\objF_\Bzeta$
  performs the update
  \begin{equation*}
    \Bw_{k+1}=\Bw_k - h_k \Bzeta\odot \nabla \objF(\Bw\odot\Bzeta).
  \end{equation*}
  Setting $\eps=0$, \eqref{eq:adamgeneral} on the other hand performs the update
  \begin{equation*}
    \Bw_{k+1} = \Bw_{k}-\alpha_k \Bigg(\beta_2\sum_{j=0}^k\beta_1^j\nabla\objF(\Bw_{k-j}\odot\Bzeta)\Bigg)\oslash \sqrt{\gamma_2\sum_{j=0}^k\gamma_1^j\nabla\objF(\Bw_{k-j}\odot\Bzeta)\odot \nabla\objF(\Bw_{k-j}\odot\Bzeta)}.
    \end{equation*}
    Note how the outer scaling factor $\Bzeta$ has vanished due to the
    division, in this sense making the update invariant to a
    componentwise rescaling of the objective.
\end{example}

\subsection{Algorithms}

\subsubsection{AdaGrad}
AdaGrad \cite{adagrad},
which stands for Adaptive Gradient Algorithm,
corresponds to \eqref{eq:adamgeneral} with
\begin{align*}
	\beta_1=0,\qquad\gamma_1=\beta_2=\gamma_2=1,\qquad
	\alpha_k=\alpha\quad\text{for all }k\in\N_0.
\end{align*}
This leaves the hyperparameters $\eps>0$ and $\alpha>0$. Here
$\alpha>0$ can be understood as a ``global'' learning rate.
The default values in tensorflow
\cite{tensorflow} are $\alpha=0.001$ and $\eps=10^{-7}$.
The AdaGrad update then reads
\begin{align*}
	\Bs_{k+1} &= \Bs_{k}+\nabla\objF(\Bw_{k})\odot\nabla\objF(\Bw_{k})\\
	\Bw_{k+1}&=\Bw_{k}-\alpha\nabla\objF(\Bw_{k})\oslash\sqrt{\Bs_{k+1}+\eps}.
\end{align*}

Due to
\begin{align}\label{eq:adagradweight}
  \Bs_{k+1}=\sum_{j=0}^{k}\nabla\objF(\Bw_j)\odot\nabla\objF(\Bw_j),
\end{align}
the algorithm therefore scales the gradient $\nabla\objF(\Bw_{k})$ in
the update componentwise by the inverse square root of the sum over
all past squared gradients plus $\eps$. Note that the scaling factor
$(s_{k+1,i}+\eps)^{-1/2}$ for component $i$ will be large, if the
previous gradients for that component were small, and vice versa.

\subsubsection{RMSProp}
RMSProp, which stands for Root Mean Squared Propagation, was introduced
by Tieleman and Hinton \cite{rmsprop}. It
corresponds to \eqref{eq:adamgeneral} with
\begin{align*}
	\beta_1=0,\qquad\beta_2=1,\qquad
	\gamma_2=1-\gamma_1\in (0,1),\qquad
	\alpha_k=\alpha\quad\text{for all }k\in\N_0,
\end{align*}
effectively leaving the hyperparameters $\eps>0$, $\gamma_1\in (0,1)$
and $\alpha>0$. The default values in tensorflow \cite{tensorflow} are
$\eps=10^{-7}$, $\alpha=0.001$ and $\gamma_1=0.9$. The algorithm is
thus given through
\begin{subequations}\label{eq:rmsprop}
\begin{align}
	\Bs_{k+1} &= \gamma_1\Bs_{k}+(1-\gamma_1)\nabla\objF(\Bw_{k})\odot\nabla\objF(\Bw_{k})\label{eq:rmspropv}\\
	\Bw_{k+1} &= \Bw_{k}- \alpha
	\nabla\objF(\Bw_{k})\oslash \sqrt{\Bs_{k+1}+\eps}.\label{eq:rmspropw}
\end{align}
\end{subequations}

The scaling vector can be expressed as
\begin{align*}
	\Bs_{k+1} = (1-\gamma_1)\sum_{j=0}^{k}\gamma_1^j\nabla\objF(\Bw_{k-j})\odot\nabla\objF(\Bw_{k-j}),
\end{align*}
and corresponds to an exponentially weighted moving average over the
past squared gradients. Unlike for AdaGrad \eqref{eq:adagradweight},
where past gradients accumulate indefinitely, RMSprop exponentially
downweights older gradients, giving more weight to recent updates.
This prevents the overly rapid decay of learning rates and slow
convergence sometimes observed in AdaGrad, e.g.\
\cite{NIPS2017_81b3833e,geron2017hands-on}. For the same reason, the
authors of Adadelta \cite{adadelta} proposed to use as a scaling
vector the average over a moving window of the past $m$ squared
gradients, for some fixed $m\in\N$. For more details on Adadelta, see
\cite{adadelta,1609.04747}. The standard RMSProp algorithm does not
incorporate momentum, however this possibility is already mentioned in
\cite{rmsprop}, also see \cite{pmlr-v28-sutskever13}.

\subsubsection{Adam}
Adam \cite{adam}, which stands for Adaptive Moment Estimation, corresponds to \eqref{eq:adamgeneral} with
\begin{align*}
	\beta_2=1-\beta_1\in (0,1),\qquad
	\gamma_2=1-\gamma_1\in (0,1),\qquad
	\alpha_k = \alpha \frac{\sqrt{1-\gamma_1^{k+1}}}{1-\beta_1^{k+1}}
\end{align*}
for all $k\in\N_0$, for some $\alpha>0$.
The default values for the remaining
parameters recommended in \cite{adam} are
$\eps=10^{-8}$, $\alpha=0.001$, $\beta_1=0.9$ and $\gamma_1=0.999$.
The update can be formulated as
\begin{subequations}\label{eq:adam}
\begin{align}
	\Bu_{k+1} &= \beta_1\Bu_{k}+(1-\beta_1)\nabla\objF(\Bw_k),
	      &&\hat \Bu_{k+1} = \frac{\Bu_{k+1}}{1-\beta_1^{k+1}}\\
	\Bs_{k+1} &= \gamma_1\Bs_{k}+(1-\gamma_1)\nabla\objF(\Bw_k)\odot\nabla\objF(\Bw_k),
	      &&\hat\Bs_{k+1} = \frac{\Bs_{k+1}}{1-\gamma_1^{k+1}}\label{eq:adamv}\\
	\Bw_{k+1}&=\Bw_k-\alpha \hat\Bu_{k+1}\oslash\sqrt{\hat\Bs_{k+1}+\eps}.\label{eq:adamw}
\end{align}
\end{subequations}

Compared to RMSProp, Adam introduces two modifications. First, due to
$\beta_1>0$,
\begin{equation*}
  \Bu_{k+1}=(1-\beta_1)\sum_{j=0}^k\beta_1^{j}\nabla\objF(\Bw_{k-j})
\end{equation*}
which
corresponds to heavy ball momentum
(cf.~\eqref{eq:HBM2}). Second, to counteract the initialization bias
from \( \Bu_0 = \Bnul \) and \( \Bs_0 = \Bnul \), Adam applies a bias
correction via
\begin{equation*}
   \hat \Bu_k = \frac{\Bu_k}{1-\beta_1^k}, \quad \hat \Bs_k = \frac{\Bs_k}{1-\gamma_1^k}.
\end{equation*}

It should be noted that there exist specific settings and convex
optimization problems for which Adam (and RMSProp and Adadelta) does
not necessarily converge to a minimizer, see \cite{adamconv1}. The
authors of \cite{adamconv1} propose a modification termed AMSGrad,
which avoids this issue.  Nonetheless, Adam remains a highly popular
algorithm for the training of neural networks. We also note that, in
the stochastic optimization setting, convergence proofs of such
algorithms in general still require $k$-dependent decrease of the
``global'' learning rate such as $\alpha=O(k^{-1/2})$ in
\eqref{eq:rmspropw} and \eqref{eq:adamw}.

\section{Backpropagation}\label{sec:backprop}
In this section we discuss how to apply gradient-based methods to the training of
feedforward neural networks.

Let $\Phi\in\CN_{d_0}^{d_{L+1}}(\sigma;L,n)$ (see Definition \ref{def:CN}) and assume that the activation function satisfies
$\sigma\in C^1(\R)$.
As earlier, we denote the neural network parameters by
\begin{align}\label{eq:params}
	\Bw =
	((\BW^{(0)},\Bb^{(0)}),\dots,(\BW^{(L)},\Bb^{(L)}))
\end{align}
with weight matrices $\BW^{(\ell)}\in\R^{d_{\ell+1}\times d_\ell}$ and
bias vectors $\Bb^{(\ell)}\in\R^{d_{\ell+1}}$.
Additionally, we fix a
differentiable loss function
$\CL:\R^{d_{L+1}}\times \R^{d_{L+1}}\to\R$, e.g.,
$\CL(\By,\tilde\By)=\norm{\By-\tilde\By}^2/2$, and assume given data
$(\Bx_j,\By_j)_{j=1}^m\subseteq \R^{d_0}\times\R^{d_{L+1}}$.
The goal is to minimize an empirical risk of the
form
\begin{align}\label{eq:bpobj}
	\objF(\Bw)
	\dfn \frac{1}{m}\sum_{j=1}^m\CL(\Phi(\Bx_j,\Bw),\By_j)
\end{align}
as a function of the neural network parameters $\Bw$.
An application of the gradient step \eqref{eq:GD} to
update the parameters requires the computation of
\begin{align*}
	\nabla\objF(\Bw) = \frac{1}{m}\sum_{j=1}^m \nabla_\Bw\CL(\Phi(\Bx_j,\Bw),\By_j).
\end{align*}
For stochastic methods, as explained in Example \ref{ex:emr}, we only
compute the average over a (random) subbatch of the dataset.
In either
case, we need an algorithm to determine
$\nabla_\Bw\CL(\Phi(\Bx,\Bw),\By)$, i.e.\ the gradients
\begin{align}\label{eq:pLWb}
	\nabla_{\Bb^{(\ell)}}\CL(\Phi(\Bx,\Bw),\By)\in\R^{d_{\ell+1}},\quad
	\nabla_{\BW^{(\ell)}}\CL(\Phi(\Bx,\Bw),\By)\in\R^{d_{\ell+1}\times d_\ell}
\end{align}
for all $\ell=0,\dots,L$.

The backpropagation algorithm \cite{rumelhart1986learning} provides an \emph{efficient} way to do so.

\subsection{Basic idea}\label{sec:bpbasic}
Due to the compositional structure of the layers in a neural network,
the objective function in \eqref{eq:bpobj} is a repeated composition
of mappings. The computation of its derivatives thus requires repeated
application of the chain rule. A direct implementation of this is
inefficient due to the occurrence of redundant calculations. The
complexity can be significantly reduced by storing and reusing certain
intermediate values. In this section we first show this in a simplified univariate
setting.

\subsubsection{Efficient use of the chain rule}
Let $f_\ell:\R\to\R$ be differentiable functions for $\ell=1,\dots,L+1$.
We wish to compute the derivative of
\begin{equation*}
  f_{L+1}\circ\cdots\circ f_1.
\end{equation*}
For $w\in\R$ denote
\begin{equation*}
  \bar f^{(1)}\dfn f_1(w)\qquad\text{and}\qquad
  \bar f^{(\ell)}\dfn f_\ell (\bar f^{({\ell-1})})\text{ for }\ell=2,\dots,L+1,
\end{equation*}
so that $\bar f^{(L+1)} = f_{L+1}\circ \cdots \circ f_1(w)$.
By the chain rule for any $w\in\R$
\begin{equation*}
  (f_{L+1}\circ\cdots\circ f_1)'(w) = f_1'(w)\cdot \prod_{\ell=2}^{L+1} f_\ell'(\underbrace{f_{\ell-1}\circ\cdots\circ f_1(w)}_{=\bar f^{(\ell-1)}}) = \frac{\partial \bar f^{(L+1)}}{\partial w}.
\end{equation*}
If we consider each evaluation of $f_\ell$ and $f_\ell'$ to be one
operation, then a naive implementation of this formula requires
$O(L^2)$ operations. If instead we first iteratively compute and store
the values $\bar f^{(\ell)}$ for $\ell=1,\dots,L$, then the
computation reduces to $O(L)$ operations.

\subsubsection{Forward and backward pass}
For neural networks the situation is slightly more complicated, since
each layer of the network corresponds to a function depending on the output
of the previous layer and on its own
parameters. We again consider this in a simplified setting, and assume
$f_1:\R\to\R$ and $f_\ell:\R\times\R\to \R$ to be differentiable for $\ell=2,\dots,L+1$.
With
\begin{equation*}
  \bar f^{(1)}\dfn f_1(w^{(0)})\qquad\text{and}\qquad
  \bar f^{(\ell)}\dfn f_\ell(w^{(\ell-1)},\bar f^{({\ell-1})})\text{ for }\ell=2,\dots,L+1,
\end{equation*}
our goal is to compute the $L+1$ partial derivatives
\begin{equation}\label{eq:allderivatives}
  \frac{\partial \bar f^{(L+1)}}{\partial w^{(0)}},\dots,\frac{\partial \bar f^{(L+1)}}{\partial w^{(L)}}.
\end{equation}
Here $w^{(\ell)}\in\R$ can be interpreted as the parameter of the $\ell$th layer.
Repeated application of the chain rule yields for $\ell=0,\dots,L$
\begin{equation*}
  \frac{\partial \bar f^{(L+1)}}{\partial w^{(\ell)}} = \frac{\partial \bar f^{(\ell+1)}}{\partial w^{(\ell)}}\underbrace{\prod_{k= \ell+1}^{L} \frac{\partial \bar f^{(k+1)}}{\partial \bar f^{(k)}}}_{\dfnn \alpha^{(\ell+1)}},
\end{equation*}
where we used that $\bar f^{(k)}$ only depends on $w^{(\ell)}$ if $k>\ell$.
Note that
\begin{equation*}
  \alpha^{(\ell)} =\frac{\partial \bar f^{(L+1)}}{\partial\bar f^{(\ell)}}
  =\alpha^{(\ell+1)}\frac{\partial \bar f^{(\ell+1)}}{\partial\bar f^{(\ell)}}
  \qquad\text{for all }\ell=1,\dots,L+1.
\end{equation*}
An efficient way, requiring only $O(L)$ operations, to compute all
derivatives in \eqref{eq:allderivatives} is thus by iteratively computing the
quantities %
\begin{equation}\label{eq:backproporder}
  \bar f^{(1)},\dots,\bar f^{(L+1)}\qquad\text{and}\qquad
  \Big(\alpha^{(L+1)},\frac{\bar f^{(L+1)}}{\partial w^{(L)}}\Big),\dots,\Big(\alpha^{(1)},\frac{\bar f^{(L+1)}}{\partial w^{(0)}}\Big)
\end{equation}
in this order. The computation of $\bar f^{(1)},\dots,\bar f^{(L+1)}$ is referred
to as the forward pass, as information is iteratively passed through
each layer of composition. The computation of the
second part in \eqref{eq:backproporder},
where the gradient information is built starting from the most outer function/layer, is called the backward pass.

\subsection{Feedforward neural networks}
We now return to the setting \eqref{eq:params}--\eqref{eq:bpobj}, and apply
the ideas in Section \ref{sec:bpbasic} for the computation of the gradients \eqref{eq:pLWb}.

Fix an input $\Bx\in\R^{d_0}$ and introduce the notation
\begin{subequations}\label{eq:defxj}
  \begin{align}
  \bar\Bx^{(1)} &\dfn \BW^{(0)}\Bx+\Bb^{(0)}\label{eq:defxj2}\\  
  \bar\Bx^{(\ell+1)}&\dfn \BW^{(\ell)}\sigma(\bar\Bx^{(\ell)})+\Bb^{(\ell)} \qquad\text{for }\ell=1,\dots,L,\label{eq:defxj3}
\end{align}
\end{subequations}
where the application of $\sigma:\R\to\R$ to a vector is,
as always, understood
componentwise.
With the notation of Definition \ref{def:nn}, 
$\Bx^{(\ell)}=\sigma(\bar\Bx^{(\ell)})\in\R^{d_\ell}$ for $\ell=1,\dots,L$ and
$\bar\Bx^{(L+1)}=\Bx^{(L+1)}=\Phi(\Bx,\Bw)\in\R^{d_{L+1}}$
is the output of the neural network.
Therefore, the $\bar\Bx^{(\ell)}$ for $\ell=1,\dots,L$ are sometimes also referred to as
the \emph{preactivations}.

In the following, we additionally fix $\By\in\R^{d_{L+1}}$ and write
\begin{align*}
	\CL\dfn \CL(\Phi(\Bx,\Bw),\By)=\CL(\bar\Bx^{(L+1)},\By).
\end{align*}                                                     
Note that $\bar\Bx^{(k)}$ depends on $(\BW^{(\ell)},\Bb^{(\ell)})$ only if
$k>\ell$. Since $\bar\Bx^{(\ell+1)}$ is a function of
$\bar\Bx^{(\ell)}$ for each $\ell$, by repeated application of the chain rule
\begin{align}\label{eq:easytocompute}
	\frac{\partial \CL}{\partial W_{ij}^{(\ell)}} =
	\underbrace{\frac{\partial \CL}{\partial \bar\Bx^{(L+1)}}}_{\in \R^{1\times d_{L+1}}}
	\underbrace{\frac{\partial \bar\Bx^{(L+1)}}{\partial \bar\Bx^{(L)}}}_{\in\R^{d_{L+1}\times d_L}}\cdots
	\underbrace{\frac{\partial \bar\Bx^{(\ell+2)}}{\partial \bar\Bx^{(\ell+1)}}}_{\in\R^{d_{\ell+2}\times d_{\ell+1}}}
	\underbrace{\frac{\partial \bar\Bx^{(\ell+1)}}{\partial W_{ij}^{(\ell)}}}_{\in\R^{d_{\ell+1}\times 1}}.
\end{align}
An analogous calculation holds for
$\partial\CL/\partial b_j^{(\ell)}$.
To avoid unnecessary computations, %
following the idea in Section \ref{sec:bpbasic},
we introduce
\begin{align}\label{eq:Balphaell}
	\Balpha^{(\ell)}\dfn \nabla_{\bar\Bx^{(\ell)}}\CL\in\R^{d_{\ell}}\qquad\text{for all }\ell=1,\dots,L+1
\end{align}
and observe that
\begin{align*}
	\frac{\partial \CL}{\partial W_{ij}^{(\ell)}} = (\Balpha^{(\ell+1)})^\top \frac{\partial \bar\Bx^{(\ell+1)}}{\partial W_{ij}^{(\ell)}}.
\end{align*}
We next formalize that the $\Balpha^{(\ell)}$ can be computed
recursively for $\ell=L+1,\dots,1$.
This explains the name ``backpropagation''.
As before, $\odot$ denotes the componentwise product.

\begin{lemma}\label{lemma:backprop}
It holds
\begin{align}\label{eq:ValphaL}
	\Balpha^{(L+1)} = \nabla_{\bar\Bx^{(L+1)}}\CL(\bar\Bx^{(L+1)},\By)
\end{align}
and
\begin{align}\label{eq:Balphaellexpl}
	\Balpha^{(\ell)} = \sigma'(\bar\Bx^{(\ell)}) \odot (\BW^{(\ell)})^\top\Balpha^{(\ell+1)}
	\qquad\text{for all }\ell=L,\dots,1.
\end{align}
\end{lemma}

\begin{proof}
Equation \eqref{eq:ValphaL} holds by definition.
For $\ell\in\{1,\dots,L\}$
by the chain rule
\begin{align*}
	\Balpha^{(\ell)} = \frac{\partial \CL}{\partial \bar\Bx^{(\ell)}}
	=
	\underbrace{\Big(\frac{\partial \bar\Bx^{(\ell+1)}}{\partial \bar\Bx^{(\ell)}}\Big)^\top}_{\in\R^{d_{\ell}\times d_{\ell+1}}}
	\underbrace{\frac{\partial \CL}{\partial \bar\Bx^{(\ell+1)}}}_{\in\R^{d_{\ell+1}\times 1}}
	= \Big(\frac{\partial \bar\Bx^{(\ell+1)}}{\partial \bar\Bx^{(\ell)}}\Big)^\top \Balpha^{(\ell+1)}.
\end{align*}
By \eqref{eq:defxj3} for $i\in\{1,\dots,d_{\ell+1}\}$, $j\in\{1,\dots,d_{\ell}\}$
\begin{align*}
	\Big(\frac{\partial \bar\Bx^{(\ell+1)}}{\partial \bar\Bx^{(\ell)}}\Big)_{ij} = 
	\frac{\partial \bar x^{(\ell+1)}_i}{\partial \bar x^{(\ell)}_j} = W_{ij}^{(\ell)}\sigma'(\bar x_j^{(\ell)}).
\end{align*}
Thus the claim follows.
\end{proof}

Putting everything together, we obtain explicit
formulas for \eqref{eq:pLWb}.

\begin{proposition}\label{prop:backprop}
It holds 
\begin{align*}
	\nabla_{\Bb^{(\ell)}}\CL
  = \Balpha^{(\ell+1)}\in\R^{d_{\ell+1}}
  \qquad\text{for }\ell=0,\dots,L
\end{align*}
and
\begin{align*}
	\nabla_{\BW^{(0)}}\CL
	= \Balpha^{(1)} %
  \Bx^\top \in\R^{d_{1}\times d_0}
\end{align*}
and
\begin{align*}
	\nabla_{\BW^{(\ell)}}\CL
	= \Balpha^{(\ell+1)} %
  \sigma(\bar\Bx^{(\ell)})^\top \in\R^{d_{\ell+1}\times d_\ell}
  \qquad   \text{for }\ell=1,\dots,L.
\end{align*}
\end{proposition}

\begin{proof}
By \eqref{eq:defxj2} for $i$, $k\in\{1,\dots,d_{1}\}$, and $j\in\{1,\dots,d_{0}\}$
\begin{align*}
	\frac{\partial \bar x_k^{(1)}}{\partial b_i^{(0)}} = \delta_{ki}\qquad
	\text{and}\qquad
	\frac{\partial \bar x_k^{(1)}}{\partial W_{ij}^{(0)}} = \delta_{ki} x_j,
\end{align*}
and by \eqref{eq:defxj3} for $\ell\in\{1,\dots,L\}$ and
  $i$, $k\in\{1,\dots,d_{\ell+1}\}$, and $j\in\{1,\dots,d_{\ell}\}$
\begin{align*}
	\frac{\partial \bar x_k^{(\ell+1)}}{\partial b_i^{(\ell)}} = \delta_{ki}\qquad
	\text{and}\qquad
	\frac{\partial \bar x_k^{(\ell+1)}}{\partial W_{ij}^{(\ell)}} = \delta_{ki}\sigma(\bar x^{(\ell)}_j).
\end{align*}
Thus, with $\Be_i=(\delta_{ki})_{k=1}^{d_{\ell+1}}$
\begin{align*}
	\frac{\partial\CL}{\partial b_i^{(\ell)}} =
	\Big(\frac{\partial\bar\Bx^{(\ell+1)}}{\partial b_i^{(\ell)}}\Big)^\top
	\frac{\partial \CL}{\partial\bar\Bx^{(\ell+1)}}
	= \Be_i^\top \Balpha^{(\ell+1)}
	= \alpha_i^{(\ell+1)}\qquad\text{for }\ell\in\{0,\dots,L\}
\end{align*}
and similarly
\begin{align*}
	\frac{\partial\CL}{\partial W_{ij}^{(0)}} =
	\Big(\frac{\partial\bar\Bx^{(1)}}{\partial W_{ij}^{(0)}}\Big)^\top \Balpha^{(1)}
	= x_j\Be_i^\top \Balpha^{(1)}
	= x_j\alpha_i^{(1)}
\end{align*}
and
\begin{align*}
	\frac{\partial\CL}{\partial W_{ij}^{(\ell)}} =
  \sigma(\bar x_j^{(\ell)})\alpha_i^{(\ell+1)}
  \qquad\text{for }\ell\in\{1,\dots,L\}.
\end{align*}
This concludes the proof.
\end{proof}

Lemma \ref{lemma:backprop} and Proposition \ref{prop:backprop}
motivate Algorithm \ref{alg:backprop}, in which a forward pass
computing $\bar\Bx^{(\ell)}$, $\ell=1,\dots,L+1$, is followed by a
backward pass to determine the $\Balpha^{(\ell)}$, $\ell=L+1,\dots,1$,
and the gradients of $\CL$ with respect to the neural network parameters.
This shows how to use gradient-based optimizers from the previous sections for
the training of neural networks.

Two important remarks are in order.
First, the objective function associated to neural networks is typically not convex as a function of the neural network weights and biases.
Thus, the analysis of the previous sections will in general not be directly applicable.
It may still give some insight about the convergence behavior locally around a (local) minimizer however.
Second, %
we assumed the activation function to be continuously differentiable, which does not hold for ReLU.
Using the concept of subgradients, gradient-based algorithms and their analysis may be generalized to some extent to also accommodate non-differentiable objective functions,
see Exercises \ref{ex:sg1}--\ref{ex:sg3}.

\begin{algorithm}[htb]
\caption{Backpropagation}\label{alg:backprop}
\begin{algorithmic}
\STATE \textbf{Input:} Network input $\Bx$, target output $\By$, neural network parameters $((\BW^{(0)},\Bb^{(0)}),\dots,(\BW^{(L)},\Bb^{(L)}))$
\STATE \textbf{Output:} Gradients of the loss function $\CL$ with respect to
neural network parameters
\STATE
\STATE{\bf Forward pass}
\STATE $\bar\Bx^{(1)} \leftarrow \BW^{(0)}\Bx+\Bb^{(0)}$
\FOR{$\ell = 1,\dots,L$}
\STATE $\bar\Bx^{(\ell+1)}\leftarrow \BW^{(\ell)}\sigma(\bar\Bx^{(\ell)})+\Bb^{(\ell)}$
\ENDFOR
\STATE
\STATE{\bf Backward pass}
\STATE $\Balpha^{(L+1)} \leftarrow \nabla_{\bar\Bx^{(L+1)}}\CL(\bar\Bx^{(L+1)},\By)$
\FOR{$\ell = L,\dots,1$}
\STATE $\nabla_{\Bb^{(\ell)}}\CL\leftarrow \Balpha^{(\ell+1)}$
\STATE $\nabla_{\BW^{(\ell)}}\CL\leftarrow \Balpha^{(\ell+1)} \sigma(\bar\Bx^{(\ell)})^\top$
\STATE $\Balpha^{(\ell)} \leftarrow \sigma'(\bar\Bx^{(\ell)}) \odot (\BW^{(\ell)})^\top\Balpha^{(\ell+1)}$
\ENDFOR
\STATE $\nabla_{\Bb^{(0)}}\CL \leftarrow \Balpha^{(1)}$
\STATE $\nabla_{\BW^{(0)}}\CL\leftarrow \Balpha^{(1)}\Bx^\top$
\end{algorithmic}
\end{algorithm}

\section*{Bibliography and further reading}
The convergence proof of gradient descent for smooth and strongly convex functions presented in Section \ref{sec:GD} follows \cite{2301.11235}, which provides a collection of simple proofs for various (stochastic) gradient descent methods together with detailed references. For standard textbooks on gradient descent and convex optimization, see for example \cite{bertsekas16,nesterov04,nesterov,boyd04,MR1857264,nocedal06,alma99169683795301081,bubeck,doi:10.1137/1.9781611974997}. These references also include convergence proofs under weaker assumptions than those considered here. For convergence results assuming for example the Polyak-\L{}ojasiewicz inequality, which does not require convexity, see, e.g., \cite{plinequality}. 

Stochastic gradient descent (SGD) discussed
in Section \ref{sec:SGD} originally dates back to Robbins and Monro
\cite{RobbinsMonro}. %
The proof presented here for strongly convex objective functions is
based on \cite{pmlr-v97-qian19b,doi:10.1137/16M1080173} and in
particular uses the step size from \cite{pmlr-v97-qian19b}; also see
\cite{BachandMoulines,10.5555/3042573.3042774,doi:10.1137/070704277,pmlr-v28-shamir13}. For
insights into the potential benefits of SGD in terms of generalization
properties, see, e.g.,
\cite{10.1016/S0893-6080(03)00138-2,pmlr-v48-hardt16,1611.03530,journals/corr/KeskarMNST16,JMLR:v19:18-188}.

The heavy ball method in Section
\ref{sec:acc} %
goes back to Polyak \cite{POLYAK19641}. To motivate the algorithm we
proceed as in \cite{goh2017why,Polyak,QIAN1999145}, and also refer to
\cite{ocw:18.065,MR3348171}. The analysis  
of Nesterov acceleration \cite{nesterov83} follows the %
arguments in \cite{pmlr-v70-tu17a,wilson}, with a similar proof
also given in \cite{Weissmann2022Optimization}.

For Section \ref{sec:othertraining} on adaptive learning rates, we follow the
overviews \cite[Section 8.5]{GoodBengCour16}, \cite{1609.04747}, and \cite[Chapter 11]{geron2017hands-on} and the original works that introduced AdaGrad \cite{adagrad}, Adadelta \cite{adadelta},
RMSProp \cite{rmsprop} and Adam \cite{adam}. %
Regarding the analysis of RMSProp and Adam, we
refer to \cite{adamconv1} which %
give an example of non-convergence,
and provide a modification of the algorithm,
termed AMSGrad, together with a convergence analysis;
also see the recent paper \cite{2411.02853}.
Convergence proofs (for variations of)
AdaGrad and Adam can furthermore be found in \cite{adamconv2}.

The backpropagation algorithm discussed in Section \ref{sec:backprop} was
popularized by Rumelhart, Hinton and Williams
\cite{rumelhart1986learning}; for further details on the historical
development we refer to \cite[Section 5.5]{SCHMIDHUBER201585}, and
for further discussion and details of the algorithm, see for instance
\cite{haykin2009neural,bishop2007,nielsenneural,berlyand2023mathematics}.

Similar discussions of gradient descent algorithms in the context
  of deep learning as given here were recently presented in
  \cite{telgarskynotes}, \cite{jentzen2023mathematical},
  and \cite{bach2025learning}: \cite[Chapter 7]{telgarskynotes} provides
  accessible convergence proofs of (stochastic) gradient descent and
  gradient flow under different smoothness and convexity assumptions,
  \cite[Part III]{jentzen2023mathematical} gives a broader overview of
  optimization techniques in deep learning, but restricts part of the
  analysis to quadratic objective functions, and \cite[Chapters 5,
  15]{bach2025learning} adds in particular the topic of variance
  reduction.
  As in \cite{doi:10.1137/16M1080173}, our
  analysis of gradient descent, stochastic gradient descent, and
  Nesterov acceleration, exclusively focused on strongly convex
  objective functions. We also refer to this paper for a more detailed
  general treatment and analysis of optimization algorithms in machine
  learning, covering various methods that are omitted here.  A
comprehensive overview skipping proofs can be found in
\cite{pml1Book,pml2Book}. Details on implementations in Python can for example
be found in \cite{geron2017hands-on}, and for recommendations and
tricks regarding the implementation we also refer to
\cite{lecun98,Bottou2012}.

\newpage
\section*{Exercises}

\begin{exercise}\label{ex:Lsmooth}
  Let $L>0$ and let $f:\R^n\to\R$ be continuously differentiable.
  Show that \eqref{eq:objFLip} implies \eqref{eq:Lsmooth0}.
\end{exercise}

\begin{exercise}\label{ex:convex}
Let $\objF\in C^1(\R^n)$.
Show that $\objF$ is convex in the
sense of Definition \ref{def:convex} if and only if
\begin{align*}
	\objF(\Bw)+\dup{\nabla\objF(\Bw)}{\Bv-\Bw}\le \objF(\Bv)\qquad
	\text{for all }\Bw,\Bv\in\R^n.
\end{align*}
\end{exercise}

\begin{definition}\label{def:subgrad}
  For convex $\objF:\R^n\to\R$, $\Bg\in\R^n$ is called a {\bf
  subgradient} (or subdifferential) of $\objF$ at $\Bv$ if and only if 
\begin{equation}\label{eq:subgrad}
  \objF(\Bw)\ge \objF(\Bv)+\inp{\Bg}{\Bw-\Bv}\qquad\text{for all }\Bw\in\R^n.
\end{equation}
The set of all subgradients of $\objF$ at $\Bv$ is denoted by
$\partial\objF(\Bv)$.
\end{definition}
For convex functions $\objF$, a subgradient always exists, i.e.\
  $\partial\objF(\Bv)$ is necessarily nonempty,
  e.g., \cite[Section 1.2]{bubeck}.
  Subgradients generalize the notion of gradients for convex
  functions, since for any convex continuously differentiable $\objF$,
  \eqref{eq:subgrad} is satisfied with $\Bg=\nabla\objF(\Bv)$. The
    following three exercises on subgradients are based on the lecture
    notes \cite{boyd2003subgrad}. Also see, e.g.,
    \cite{shor1985minimization,bubeck,alma99169683795301081} for more
    details on subgradient descent.

  \begin{exercise}\label{ex:sg1}
    Let $\objF:\R^n\to\R$ be convex and ${\rm Lip}(\objF)\le L$. Show that for any $\Bg\in\partial \objF(\Bv)$ holds $\norm{\Bg}\le L$.
  \end{exercise}

  \begin{exercise}\label{ex:sg2}
    Let $\objF:\R^n\to\R$ be convex, ${\rm Lip}(\objF)\le L$ and
    suppose that $\Bw_*$ is a minimizer of $\objF$.  Fix
    $\Bw_0\in\R^n$, and for $k\in\N_0$ define the {\bf subgradient descent} update
\begin{equation*}
  \Bw_{k+1}\dfn \Bw_{k}-h_k \Bg_{k},
\end{equation*}
where $\Bg_{k}$ is an arbitrary fixed element of
$\partial\objF(\Bw_{k})$. Show that
\begin{equation*}
\min_{i\le k}\objF(\Bw_{i}) - \objF(\Bw_*) \le \frac{ \norm{\Bw_{0} - \Bw_*}^2 + L^2 \sum \limits_{i=1}^k h_i^2 }{2 \sum \limits_{i=1}^k h_i }.
\end{equation*}

\emph{Hint}: Start by recursively expanding
  $\norm{\Bw_k-\Bw_*}^2=\cdots$, and then apply the property of the
  subgradient.
\end{exercise}

\begin{exercise}\label{ex:sg3}
  Consider the setting of Exercise \ref{ex:sg2}.
  Determine step sizes
$h_1,\dots,h_k$ (which may depend on $k$, i.e.\
$h_{k,1},\dots,h_{k,k}$) such that %
$\delta>0$
\begin{equation*}
  \min_{i\le k}\objF(\Bw_i)-\objF(\Bw_*) = O(k^{-1/2})\qquad\text{as }k\to\infty.
\end{equation*}
\end{exercise}

\begin{exercise}\label{ex:quadraticobjective}
Let $\BA\in\R^{n\times n}$ be symmetric positive semidefinite,
$\Bb\in\R^n$ and $c\in\R$.
Denote the eigenvalues of $\BA$ by
$\zeta_1\ge\dots\ge\zeta_n\ge 0$.
Show that the objective function
\begin{align}\label{eq:quadraticobjective2}
	\objF(\Bw)\dfn \frac{1}{2}\Bw^\top \BA\Bw+\Bb^\top\Bw+c
\end{align}
is convex and $\zeta_1$-smooth. Moreover, if $\zeta_n>0$, then
$\objF$ is $\zeta_n$-strongly convex.  Show that these values are
optimal in the sense that $\objF$ is neither $L$-smooth nor
$\mu$-strongly convex if $L<\zeta_1$ and $\mu>\zeta_n$.

\emph{Hint}: Note that $L$-smoothness and $\mu$-strong convexity are
invariant under shifts and the addition of constants.  That is, for
every $\alpha\in\R$ and $\Bbeta\in\R^n$, 
$\tilde\objF(\Bw)\dfn \alpha+\objF(\Bw+\Bbeta)$ is $L$-smooth or
$\mu$-strongly convex if and only if $\objF$ is. It thus suffices to
consider $\Bw^\top\BA\Bw/2$.
\end{exercise}

\begin{exercise}\label{ex:hbm}
Let $\objF$ be as in Exercise \ref{ex:quadraticobjective}.
Show that
gradient descent converges for arbitrary initialization
$\Bw_0\in\R^n$, if and only if
\begin{align*}
	\max_{j=1,\dots,n}|1-h\zeta_j|<1.
\end{align*}
Show that
$\argmin_{h>0} \max_{j=1,\dots,n}
|1-h\zeta_j|=2/(\zeta_1+\zeta_n)$ and conclude that the
convergence will be slow if $\objF$ is ill-conditioned, i.e.\ if
$\zeta_1/\zeta_n\gg 1$.

\emph{Hint}: Assume first that
$\Bb=\Bnul\in\R^n$ and $c=0\in\R$ in \eqref{eq:quadraticobjective2}, and
use the singular value decomposition
$\BA=\BU^\top{\rm diag}(\zeta_1,\dots,\zeta_n)\BU$.
\end{exercise}

\begin{exercise}\label{ex:threeterm}
Show that \eqref{eq:NAG} can equivalently be written as
\eqref{eq:NAG3} with $\tau=\sqrt{\mu/L}$, $\alpha=1/L$,
$\beta = (1-\tau)/(1+\tau)$ and the initialization
$\Bu_0 = ((1+\tau)\Bw_0-\Bs_0)/\tau$.
\end{exercise}

%% file: plots/optimization.tex
\begin{tikzpicture}[scale=1.1]
\draw [thick] (-4.5,0.7562499999999998) -- (-4.47979797979798,0.7273794575593748) -- (-4.459595959595959,0.7043855125033257) -- (-4.4393939393939394,0.6868688502438187) -- (-4.41919191919192,0.6744401499882348) -- (-4.398989898989899,0.6667200847393692) -- (-4.378787878787879,0.6633393212954319) -- (-4.358585858585858,0.6639385202500476) -- (-4.338383838383838,0.6681683359922553) -- (-4.318181818181818,0.6756894167065091) -- (-4.297979797979798,0.6861724043726768) -- (-4.277777777777778,0.6992979347660417) -- (-4.257575757575758,0.714756637457301) -- (-4.237373737373737,0.732249135812567) -- (-4.217171717171717,0.7514860469933661) -- (-4.196969696969697,0.7721879819566395) -- (-4.1767676767676765,0.7940855454547433) -- (-4.156565656565657,0.8169193360354474) -- (-4.136363636363637,0.8404399460419371) -- (-4.116161616161616,0.8644079616128117) -- (-4.095959595959596,0.8885939626820853) -- (-4.075757575757576,0.9127785229791867) -- (-4.055555555555555,0.936752210028959) -- (-4.0353535353535355,0.9603155851516603) -- (-4.015151515151516,0.9832792034629626) -- (-3.994949494949495,1.0054636138739532) -- (-3.974747474747475,1.0266993590911335) -- (-3.9545454545454546,1.0468269756164197) -- (-3.9343434343434343,1.0656969937471426) -- (-3.9141414141414144,1.083169937576047) -- (-3.893939393939394,1.0991163249912936) -- (-3.8737373737373737,1.1134166676764565) -- (-3.8535353535353534,1.1259614711105246) -- (-3.833333333333333,1.1366512345679012) -- (-3.813131313131313,1.1453964511184052) -- (-3.792929292929293,1.1521176076272688) -- (-3.7727272727272725,1.1567451847551398) -- (-3.7525252525252526,1.1592196569580797) -- (-3.7323232323232323,1.159491492487565) -- (-3.712121212121212,1.1575211533904872) -- (-3.691919191919192,1.1532790955091516) -- (-3.6717171717171717,1.1467457684812785) -- (-3.6515151515151514,1.1379116157400029) -- (-3.6313131313131315,1.1267770745138739) -- (-3.611111111111111,1.1133525758268557) -- (-3.590909090909091,1.0976585444983264) -- (-3.5707070707070705,1.0797253991430797) -- (-3.55050505050505,1.0595935521713231) -- (-3.5303030303030303,1.037313409788679) -- (-3.51010101010101,1.0129453719961838) -- (-3.4898989898989896,0.9865598325902893) -- (-3.4696969696969697,0.9582371791628613) -- (-3.4494949494949494,0.9280677931011808) -- (-3.429292929292929,0.8961520495879428) -- (-3.409090909090909,0.8626003176012565) -- (-3.388888888888889,0.827532959914647) -- (-3.3686868686868685,0.7910803330970526) -- (-3.3484848484848486,0.7533827875128274) -- (-3.3282828282828283,0.7145906673217391) -- (-3.308080808080808,0.67486431047897) -- (-3.287878787878788,0.6343740487351182) -- (-3.2676767676767673,0.5933002076361942) -- (-3.2474747474747474,0.5518331065236257) -- (-3.227272727272727,0.5101730585342528) -- (-3.2070707070707067,0.4685303706003314) -- (-3.186868686868687,0.4271253434495319) -- (-3.1666666666666665,0.38618827160493796) -- (-3.146464646464646,0.3459594433850499) -- (-3.1262626262626263,0.30668914090378085) -- (-3.106060606060606,0.26863764007045976) -- (-3.0858585858585856,0.23207521058982916) -- (-3.0656565656565657,0.19728211596204726) -- (-3.0454545454545454,0.16454861348268557) -- (-3.025252525252525,0.13417495424273085) -- (-3.005050505050505,0.10647138312858495) -- (-2.9848484848484844,0.08175813882206295) -- (-2.9646464646464645,0.06036545380039626) -- (-2.9444444444444446,0.042633554336229285) -- (-2.924242424242424,0.028912660497621534) -- (-2.904040404040404,0.019562986148047967) -- (-2.8838383838383836,0.01495473894639654) -- (-2.8636363636363633,0.015468120346970982) -- (-2.8434343434343434,0.021493325599489244) -- (-2.823232323232323,0.03343054374908372) -- (-2.8030303030303028,0.05168995763630124) -- (-2.782828282828283,0.07669174389710443) -- (-2.7626262626262625,0.10886607296286854) -- (-2.742424242424242,0.14865310906038498) -- (-2.7222222222222223,0.19650301021185856) -- (-2.7020202020202015,0.2528759282349099) -- (-2.6818181818181817,0.318242008742573) -- (-2.6616161616161618,0.3930813911432971) -- (-2.641414141414141,0.47788420864094683) -- (-2.621212121212121,0.5731505882347989) -- (-2.6010101010101008,0.6793906507195481) -- (-2.5808080808080804,0.7971245106852995) -- (-2.5606060606060606,0.9268822765175765) -- (-2.54040404040404,1.069204050397317) -- (-2.52020202020202,1.224639928300869) -- (-2.5,1.3937500000000003);
\draw [thick,dashed,blue] (-3.29,1.2542875000000002) -- (-3.285050505050505,1.2118348943985309) -- (-3.28010101010101,1.1702641988062443) -- (-3.2751515151515154,1.1295754132231406) -- (-3.2702020202020203,1.0897685376492197) -- (-3.2652525252525253,1.050843572084481) -- (-3.2603030303030303,1.0128005165289258) -- (-3.255353535353535,0.9756393709825528) -- (-3.25040404040404,0.9393601354453627) -- (-3.2454545454545456,0.9039628099173553) -- (-3.2405050505050506,0.8694473943985309) -- (-3.2355555555555555,0.8358138888888889) -- (-3.2306060606060605,0.8030622933884298) -- (-3.225656565656566,0.7711926078971536) -- (-3.220707070707071,0.7402048324150596) -- (-3.215757575757576,0.7100989669421489) -- (-3.2108080808080808,0.6808750114784206) -- (-3.2058585858585857,0.6525329660238752) -- (-3.200909090909091,0.6250728305785127) -- (-3.195959595959596,0.5984946051423325) -- (-3.191010101010101,0.5727982897153351) -- (-3.186060606060606,0.5479838842975207) -- (-3.181111111111111,0.524051388888889) -- (-3.176161616161616,0.5010008034894399) -- (-3.1712121212121214,0.4788321280991737) -- (-3.1662626262626263,0.45754536271808993) -- (-3.1613131313131313,0.43714050734618926) -- (-3.1563636363636363,0.4176175619834712) -- (-3.1514141414141412,0.39897652662993577) -- (-3.1464646464646466,0.38121740128558324) -- (-3.1415151515151516,0.36434018595041323) -- (-3.1365656565656566,0.34834488062442615) -- (-3.1316161616161615,0.3332314853076218) -- (-3.1266666666666665,0.31900000000000006) -- (-3.1217171717171714,0.30565042470156106) -- (-3.116767676767677,0.2931827594123049) -- (-3.111818181818182,0.2815970041322315) -- (-3.106868686868687,0.27089315886134085) -- (-3.1019191919191917,0.26107122359963275) -- (-3.0969696969696967,0.25213119834710745) -- (-3.092020202020202,0.244073083103765) -- (-3.087070707070707,0.23689687786960523) -- (-3.082121212121212,0.23060258264462824) -- (-3.077171717171717,0.22519019742883387) -- (-3.072222222222222,0.22065972222222227) -- (-3.0672727272727274,0.21701115702479348) -- (-3.0623232323232323,0.2142445018365474) -- (-3.0573737373737373,0.212359756657484) -- (-3.0524242424242423,0.21135692148760338) -- (-3.0474747474747472,0.21123599632690548) -- (-3.0425252525252526,0.21199698117539034) -- (-3.0375757575757576,0.21363987603305795) -- (-3.0326262626262626,0.21616468089990823) -- (-3.0276767676767675,0.2195713957759413) -- (-3.0227272727272725,0.2238600206611571) -- (-3.017777777777778,0.22903055555555563) -- (-3.012828282828283,0.2350830004591369) -- (-3.007878787878788,0.2420173553719009) -- (-3.002929292929293,0.2498336202938476) -- (-2.9979797979797977,0.25853179522497716) -- (-2.993030303030303,0.2681118801652893) -- (-2.988080808080808,0.2785738751147842) -- (-2.983131313131313,0.28991778007346203) -- (-2.978181818181818,0.3021435950413224) -- (-2.973232323232323,0.31525132001836553) -- (-2.9682828282828284,0.3292409550045914) -- (-2.9633333333333334,0.34411249999999993) -- (-2.9583838383838383,0.35986595500459156) -- (-2.9534343434343433,0.37650132001836556) -- (-2.9484848484848483,0.3940185950413223) -- (-2.9435353535353537,0.4124177800734623) -- (-2.9385858585858586,0.431698875114784) -- (-2.9336363636363636,0.45186188016528944) -- (-2.9286868686868686,0.4729067952249771) -- (-2.9237373737373735,0.4948336202938475) -- (-2.918787878787879,0.5176423553719012) -- (-2.913838383838384,0.5413330004591366) -- (-2.908888888888889,0.5659055555555558) -- (-2.903939393939394,0.5913600206611571) -- (-2.898989898989899,0.6176963957759412) -- (-2.894040404040404,0.6449146808999086) -- (-2.889090909090909,0.6730148760330574) -- (-2.884141414141414,0.7019969811753903) -- (-2.879191919191919,0.7318609963269055) -- (-2.874242424242424,0.762606921487603) -- (-2.8692929292929295,0.7942347566574843) -- (-2.8643434343434344,0.8267445018365467) -- (-2.8593939393939394,0.8601361570247936) -- (-2.8544444444444443,0.894409722222222) -- (-2.8494949494949493,0.9295651974288335) -- (-2.8445454545454547,0.9656025826446286) -- (-2.8395959595959597,1.0025218778696046) -- (-2.8346464646464646,1.0403230831037649) -- (-2.8296969696969696,1.0790061983471073) -- (-2.8247474747474746,1.1185712235996323) -- (-2.81979797979798,1.159018158861341) -- (-2.814848484848485,1.2003470041322308) -- (-2.80989898989899,1.242557759412305) -- (-2.804949494949495,1.2856504247015608) -- (-2.8,1.3296249999999996);
\draw [thick,dashed,blue] (-3.38,-0.6423500000000002) -- (-3.375252525252525,-0.6035712235996331) -- (-3.3705050505050504,-0.5656038337924703) -- (-3.3657575757575757,-0.5284478305785123) -- (-3.361010101010101,-0.4921032139577597) -- (-3.3562626262626263,-0.45656998393021175) -- (-3.3515151515151516,-0.42184814049586783) -- (-3.346767676767677,-0.38793768365472925) -- (-3.342020202020202,-0.35483861340679523) -- (-3.3372727272727274,-0.32255092975206634) -- (-3.3325252525252527,-0.2910746326905418) -- (-3.327777777777778,-0.2604097222222225) -- (-3.3230303030303032,-0.2305561983471076) -- (-3.318282828282828,-0.20151406106519754) -- (-3.3135353535353533,-0.17328331037649214) -- (-3.3087878787878786,-0.14586394628099175) -- (-3.304040404040404,-0.11925596877869615) -- (-3.299292929292929,-0.09345937786960523) -- (-3.2945454545454544,-0.0684741735537191) -- (-3.2897979797979797,-0.04430035583103764) -- (-3.285050505050505,-0.020937924701560973) -- (-3.2803030303030303,0.0016131198347107967) -- (-3.2755555555555556,0.023352777777777778) -- (-3.270808080808081,0.04428104912764008) -- (-3.266060606060606,0.06439793388429738) -- (-3.2613131313131314,0.08370343204775033) -- (-3.2565656565656567,0.10219754361799815) -- (-3.251818181818182,0.11988026859504136) -- (-3.247070707070707,0.13675160697887978) -- (-3.2423232323232325,0.15281155876951336) -- (-3.2375757575757573,0.1680601239669421) -- (-3.232828282828283,0.1824973025711662) -- (-3.228080808080808,0.19612309458218552) -- (-3.223333333333333,0.2089375000000001) -- (-3.2185858585858584,0.22094051882460974) -- (-3.2138383838383837,0.2321321510560147) -- (-3.209090909090909,0.24251239669421498) -- (-3.2043434343434343,0.2520812557392104) -- (-3.1995959595959595,0.26083872819100096) -- (-3.194848484848485,0.26878481404958676) -- (-3.19010101010101,0.2759195133149679) -- (-3.1853535353535354,0.28224282598714423) -- (-3.1806060606060607,0.2877547520661158) -- (-3.175858585858586,0.29245529155188255) -- (-3.171111111111111,0.2963444444444445) -- (-3.1663636363636365,0.2994222107438017) -- (-3.1616161616161618,0.3016885904499541) -- (-3.1568686868686866,0.30314358356290183) -- (-3.1521212121212123,0.30378719008264465) -- (-3.147373737373737,0.3036194100091828) -- (-3.1426262626262624,0.3026402433425161) -- (-3.1378787878787877,0.30084969008264467) -- (-3.133131313131313,0.29824775022956845) -- (-3.1283838383838383,0.29483442378328745) -- (-3.1236363636363635,0.2906097107438017) -- (-3.118888888888889,0.28557361111111107) -- (-3.114141414141414,0.27972612488521587) -- (-3.1093939393939394,0.2730672520661157) -- (-3.1046464646464647,0.2655969926538108) -- (-3.09989898989899,0.2573153466483012) -- (-3.095151515151515,0.24822231404958678) -- (-3.0904040404040405,0.23831789485766752) -- (-3.0856565656565658,0.2276020890725436) -- (-3.080909090909091,0.21607489669421479) -- (-3.076161616161616,0.2037363177226813) -- (-3.0714141414141416,0.1905863521579431) -- (-3.0666666666666664,0.17662499999999992) -- (-3.0619191919191917,0.16185226124885216) -- (-3.057171717171717,0.14626813590449944) -- (-3.0524242424242423,0.12987262396694213) -- (-3.0476767676767675,0.11266572543617984) -- (-3.042929292929293,0.09464744031221298) -- (-3.038181818181818,0.07581776859504115) -- (-3.0334343434343434,0.05617671028466474) -- (-3.0286868686868686,0.03572426538108334) -- (-3.023939393939394,0.01446043388429738) -- (-3.019191919191919,-0.007614784205693573) -- (-3.0144444444444445,-0.030501388888889036) -- (-3.0096969696969698,-0.05419938016528961) -- (-3.004949494949495,-0.0787087580348946) -- (-3.0002020202020203,-0.1040295224977047) -- (-2.995454545454545,-0.13016167355371927) -- (-2.990707070707071,-0.15710521120293858) -- (-2.9859595959595957,-0.1848601354453627) -- (-2.9812121212121214,-0.21342644628099153) -- (-2.9764646464646463,-0.2428041437098259) -- (-2.9717171717171715,-0.2729932277318643) -- (-2.966969696969697,-0.3039936983471082) -- (-2.962222222222222,-0.33580555555555625) -- (-2.9574747474747474,-0.3684287993572089) -- (-2.9527272727272726,-0.40186342975206646) -- (-2.947979797979798,-0.4361094467401287) -- (-2.943232323232323,-0.4711668503213957) -- (-2.9384848484848485,-0.5070356404958685) -- (-2.9337373737373738,-0.543715817263545) -- (-2.928989898989899,-0.5812073806244272) -- (-2.924242424242424,-0.6195103305785133) -- (-2.9194949494949496,-0.6586246671258043) -- (-2.9147474747474744,-0.6985503902663) -- (-2.91,-0.7392875000000003);
\fill (-3.1,0.25750000000000006) circle (0.05);
\node at (-3.5,-1) {\small smooth};
\draw [thick] (-1.0,1.5) -- (-0.9797979797979798,1.32712092569179) -- (-0.9595959595959596,1.1711748108964735) -- (-0.9393939393939394,1.0308080056610374) -- (-0.9191919191919192,0.904749695045026) -- (-0.898989898989899,0.7918084485539003) -- (-0.8787878787878788,0.6908688429887029) -- (-0.8585858585858586,0.6008881587120365) -- (-0.8383838383838383,0.5208931493303548) -- (-0.8181818181818181,0.4499768847925641) -- (-0.797979797979798,0.38729566790493686) -- (-0.7777777777777778,0.33206602426233583) -- (-0.7575757575757576,0.283561765595752) -- (-0.7373737373737373,0.24111112653615213) -- (-0.7171717171717171,0.20409397479463826) -- (-0.696969696969697,0.17193909475891933) -- (-0.6767676767676767,0.14412154450609363) -- (-0.6565656565656566,0.12016008623174468) -- (-0.6363636363636364,0.09961469009534529) -- (-0.6161616161616161,0.08208411148197692) -- (-0.5959595959595959,0.06720354168035819) -- (-0.5757575757575757,0.054642331977185835) -- (-0.5555555555555556,0.044101791167787216) -- (-0.5353535353535352,0.03531305648308409) -- (-0.5151515151515151,0.028035037932868774) -- (-0.4949494949494949,0.022052436065390653) -- (-0.4747474747474747,0.017173833143255452) -- (-0.4545454545454545,0.013229857735635396) -- (-0.43434343434343425,0.010071422726791181) -- (-0.41414141414141414,0.007568036740905535) -- (-0.3939393939393939,0.005606188983228277) -- (-0.3737373737373737,0.004087807497533107) -- (-0.3535353535353535,0.0029287908398858364) -- (-0.33333333333333326,0.002057613168724277) -- (-0.31313131313131304,0.0014140027512497054) -- (-0.2929292929292928,0.0009476938861299059) -- (-0.2727272727272727,0.0006172522425138053) -- (-0.2525252525252525,0.0003889736153576842) -- (-0.23232323232323226,0.00023585609706298905) -- (-0.21212121212121204,0.0001366456654257134) -- (-0.19191919191919182,7.49551878973741e-05) -- (-0.1717171717171716,3.845684215757019e-05) -- (-0.1515151515151515,1.8147952998128116e-05) -- (-0.13131313131313127,7.690245518831642e-06) -- (-0.11111111111111105,2.8225146347383718e-06) -- (-0.09090909090909083,8.467108950806617e-07) -- (-0.07070707070707061,1.8744261375269213e-07) -- (-0.050505050505050386,2.4894311382891464e-08) -- (-0.030303030303030276,1.1614689918801943e-09) -- (-0.010101010101010055,1.5932359285050325e-12) -- (0.010101010101010166,1.5932359285051377e-12) -- (0.030303030303030498,1.1614689918802454e-09) -- (0.05050505050505061,2.489431138289212e-08) -- (0.07070707070707072,1.8744261375269388e-07) -- (0.09090909090909105,8.467108950806741e-07) -- (0.11111111111111116,2.8225146347383883e-06) -- (0.1313131313131315,7.690245518831722e-06) -- (0.1515151515151516,1.8147952998128197e-05) -- (0.1717171717171717,3.8456842157570336e-05) -- (0.19191919191919204,7.495518789737461e-05) -- (0.21212121212121215,0.00013664566542571384) -- (0.2323232323232325,0.00023585609706299038) -- (0.2525252525252526,0.00038897361535768525) -- (0.27272727272727293,0.0006172522425138084) -- (0.29292929292929304,0.00094769388612991) -- (0.31313131313131315,0.0014140027512497084) -- (0.3333333333333335,0.002057613168724285) -- (0.3535353535353536,0.0029287908398858416) -- (0.3737373737373739,0.004087807497533122) -- (0.39393939393939403,0.005606188983228287) -- (0.41414141414141437,0.007568036740905559) -- (0.4343434343434345,0.010071422726791212) -- (0.4545454545454546,0.013229857735635417) -- (0.4747474747474749,0.0171738331432555) -- (0.49494949494949503,0.02205243606539068) -- (0.5151515151515154,0.02803503793286885) -- (0.5353535353535355,0.03531305648308418) -- (0.5555555555555556,0.044101791167787216) -- (0.5757575757575759,0.05464233197718596) -- (0.595959595959596,0.06720354168035828) -- (0.6161616161616164,0.0820841114819771) -- (0.6363636363636365,0.09961469009534539) -- (0.6565656565656568,0.1201600862317449) -- (0.6767676767676769,0.1441215445060939) -- (0.696969696969697,0.17193909475891933) -- (0.7171717171717173,0.20409397479463862) -- (0.7373737373737375,0.2411111265361524) -- (0.7575757575757578,0.2835617655957525) -- (0.7777777777777779,0.3320660242623361) -- (0.7979797979797982,0.3872956679049375) -- (0.8181818181818183,0.4499768847925648) -- (0.8383838383838385,0.5208931493303552) -- (0.8585858585858588,0.6008881587120374) -- (0.8787878787878789,0.6908688429887033) -- (0.8989898989898992,0.7918084485539015) -- (0.9191919191919193,0.9047496950450264) -- (0.9393939393939394,1.0308080056610374) -- (0.9595959595959598,1.171174810896475) -- (0.9797979797979799,1.3271209256917909) -- (1.0,1.5);
\draw [thick,dashed,blue] (-0.5,-0.2578125) -- (-0.48282828282828283,-0.25298295454545455) -- (-0.46565656565656566,-0.2481534090909091) -- (-0.4484848484848485,-0.24332386363636366) -- (-0.4313131313131313,-0.2384943181818182) -- (-0.41414141414141414,-0.2336647727272727) -- (-0.396969696969697,-0.22883522727272726) -- (-0.3797979797979798,-0.2240056818181818) -- (-0.36262626262626263,-0.21917613636363636) -- (-0.34545454545454546,-0.21434659090909092) -- (-0.3282828282828283,-0.20951704545454547) -- (-0.3111111111111111,-0.2046875) -- (-0.29393939393939394,-0.19985795454545455) -- (-0.2767676767676768,-0.1950284090909091) -- (-0.2595959595959596,-0.19019886363636362) -- (-0.24242424242424243,-0.18536931818181818) -- (-0.22525252525252526,-0.18053977272727273) -- (-0.2080808080808081,-0.17571022727272728) -- (-0.19090909090909092,-0.17088068181818183) -- (-0.17373737373737375,-0.16605113636363636) -- (-0.15656565656565657,-0.1612215909090909) -- (-0.1393939393939394,-0.15639204545454546) -- (-0.12222222222222223,-0.1515625) -- (-0.10505050505050506,-0.14673295454545454) -- (-0.08787878787878789,-0.1419034090909091) -- (-0.07070707070707072,-0.13707386363636365) -- (-0.05353535353535355,-0.1322443181818182) -- (-0.036363636363636376,-0.12741477272727272) -- (-0.019191919191919204,-0.12258522727272728) -- (-0.002020202020202033,-0.11775568181818183) -- (0.015151515151515138,-0.11292613636363635) -- (0.03232323232323231,-0.1080965909090909) -- (0.04949494949494948,-0.10326704545454546) -- (0.06666666666666665,-0.09843750000000001) -- (0.08383838383838382,-0.09360795454545455) -- (0.101010101010101,-0.08877840909090909) -- (0.11818181818181817,-0.08394886363636364) -- (0.13535353535353534,-0.0791193181818182) -- (0.1525252525252525,-0.07428977272727273) -- (0.16969696969696968,-0.06946022727272727) -- (0.18686868686868685,-0.06463068181818182) -- (0.20404040404040402,-0.059801136363636376) -- (0.2212121212121212,-0.054971590909090914) -- (0.23838383838383836,-0.05014204545454545) -- (0.25555555555555554,-0.045312500000000006) -- (0.2727272727272727,-0.04048295454545456) -- (0.2898989898989899,-0.0356534090909091) -- (0.30707070707070705,-0.030823863636363642) -- (0.3242424242424242,-0.025994318181818188) -- (0.3414141414141414,-0.021164772727272733) -- (0.35858585858585856,-0.01633522727272728) -- (0.37575757575757573,-0.011505681818181825) -- (0.3929292929292929,-0.00667613636363637) -- (0.4101010101010101,-0.0018465909090909158) -- (0.42727272727272725,0.0029829545454545386) -- (0.4444444444444444,0.007812499999999993) -- (0.4616161616161616,0.012642045454545447) -- (0.47878787878787876,0.017471590909090902) -- (0.49595959595959593,0.022301136363636356) -- (0.5131313131313131,0.02713068181818181) -- (0.5303030303030303,0.031960227272727265) -- (0.5474747474747474,0.03678977272727272) -- (0.5646464646464646,0.041619318181818174) -- (0.5818181818181818,0.04644886363636363) -- (0.598989898989899,0.05127840909090908) -- (0.6161616161616161,0.05610795454545454) -- (0.6333333333333333,0.06093749999999999) -- (0.6505050505050505,0.06576704545454545) -- (0.6676767676767676,0.0705965909090909) -- (0.6848484848484848,0.07542613636363635) -- (0.702020202020202,0.08025568181818181) -- (0.7191919191919192,0.08508522727272727) -- (0.7363636363636363,0.08991477272727272) -- (0.7535353535353535,0.09474431818181817) -- (0.7707070707070707,0.09957386363636363) -- (0.7878787878787878,0.10440340909090909) -- (0.805050505050505,0.10923295454545454) -- (0.8222222222222222,0.11406249999999998) -- (0.8393939393939394,0.11889204545454544) -- (0.8565656565656565,0.1237215909090909) -- (0.8737373737373737,0.12855113636363635) -- (0.8909090909090909,0.1333806818181818) -- (0.908080808080808,0.13821022727272725) -- (0.9252525252525252,0.14303977272727272) -- (0.9424242424242424,0.14786931818181817) -- (0.9595959595959596,0.15269886363636362) -- (0.9767676767676767,0.1575284090909091) -- (0.9939393939393939,0.16235795454545454) -- (1.011111111111111,0.1671875) -- (1.0282828282828282,0.17201704545454544) -- (1.0454545454545454,0.17684659090909088) -- (1.0626262626262626,0.18167613636363636) -- (1.0797979797979798,0.1865056818181818) -- (1.096969696969697,0.19133522727272725) -- (1.114141414141414,0.19616477272727273) -- (1.1313131313131313,0.20099431818181818) -- (1.1484848484848484,0.20582386363636362) -- (1.1656565656565656,0.21065340909090907) -- (1.1828282828282828,0.21548295454545452) -- (1.2,0.2203125);
\fill (0.5,0.0234375) circle (0.05);
\node at (0,-1) {\small convex};
\draw [thick] (2.5,1.5) -- (2.5202020202020203,1.4400061218243037) -- (2.5404040404040407,1.3812366085093357) -- (2.5606060606060606,1.3236914600550964) -- (2.580808080808081,1.2673706764615857) -- (2.601010101010101,1.2122742577288033) -- (2.621212121212121,1.1584022038567492) -- (2.6414141414141414,1.1057545148454238) -- (2.6616161616161618,1.054331190694827) -- (2.6818181818181817,1.0041322314049586) -- (2.702020202020202,0.9551576369758188) -- (2.7222222222222223,0.9074074074074074) -- (2.742424242424242,0.8608815426997245) -- (2.7626262626262625,0.8155800428527702) -- (2.782828282828283,0.7715029078665442) -- (2.8030303030303028,0.7286501377410469) -- (2.8232323232323235,0.6870217324762777) -- (2.8434343434343434,0.6466176920722375) -- (2.8636363636363638,0.6074380165289256) -- (2.883838383838384,0.5694827058463422) -- (2.904040404040404,0.5327517600244872) -- (2.9242424242424243,0.4972451790633608) -- (2.9444444444444446,0.462962962962963) -- (2.9646464646464645,0.4299051117232934) -- (2.984848484848485,0.3980716253443526) -- (3.005050505050505,0.36746250382614015) -- (3.025252525252525,0.3380777471686562) -- (3.0454545454545454,0.3099173553719007) -- (3.0656565656565657,0.28298132843587376) -- (3.0858585858585856,0.25726966636057547) -- (3.106060606060606,0.2327823691460055) -- (3.1262626262626263,0.209519436792164) -- (3.1464646464646466,0.18748086929905106) -- (3.166666666666667,0.1666666666666666) -- (3.186868686868687,0.14707682889501061) -- (3.207070707070707,0.12871135598408318) -- (3.2272727272727275,0.11157024793388429) -- (3.2474747474747474,0.09565350474441381) -- (3.2676767676767677,0.08096112641567182) -- (3.287878787878788,0.06749311294765835) -- (3.308080808080808,0.05524946434037338) -- (3.3282828282828283,0.0442301805938169) -- (3.3484848484848486,0.03443526170798897) -- (3.3686868686868685,0.025864707682889482) -- (3.388888888888889,0.018518518518518497) -- (3.409090909090909,0.012396694214876009) -- (3.4292929292929295,0.0074992347719620225) -- (3.44949494949495,0.0038261401897765356) -- (3.4696969696969697,0.0013774104683195567) -- (3.48989898989899,0.00015304560759106075) -- (3.5101010101010104,0.0001530456075910641) -- (3.5303030303030303,0.001377410468319577) -- (3.5505050505050506,0.003826140189776569) -- (3.570707070707071,0.007499234771962048) -- (3.590909090909091,0.012396694214876072) -- (3.611111111111111,0.018518518518518535) -- (3.6313131313131315,0.025864707682889572) -- (3.6515151515151514,0.03443526170798902) -- (3.6717171717171717,0.044230180593816955) -- (3.691919191919192,0.055249464340373505) -- (3.712121212121212,0.06749311294765842) -- (3.7323232323232327,0.08096112641567199) -- (3.7525252525252526,0.09565350474441389) -- (3.772727272727273,0.11157024793388447) -- (3.7929292929292933,0.12871135598408334) -- (3.813131313131313,0.14707682889501072) -- (3.8333333333333335,0.16666666666666682) -- (3.853535353535354,0.18748086929905117) -- (3.8737373737373737,0.20951943679216425) -- (3.893939393939394,0.23278236914600564) -- (3.9141414141414144,0.25726966636057574) -- (3.9343434343434343,0.2829813284358741) -- (3.9545454545454546,0.3099173553719009) -- (3.974747474747475,0.3380777471686565) -- (3.994949494949495,0.3674625038261403) -- (4.015151515151516,0.3980716253443529) -- (4.0353535353535355,0.42990511172329376) -- (4.055555555555555,0.462962962962963) -- (4.075757575757576,0.49724517906336113) -- (4.095959595959596,0.5327517600244874) -- (4.116161616161616,0.5694827058463425) -- (4.136363636363637,0.6074380165289258) -- (4.156565656565657,0.646617692072238) -- (4.1767676767676765,0.6870217324762782) -- (4.196969696969697,0.7286501377410469) -- (4.217171717171717,0.7715029078665447) -- (4.237373737373737,0.8155800428527703) -- (4.257575757575758,0.8608815426997249) -- (4.277777777777778,0.9074074074074078) -- (4.2979797979797985,0.9551576369758195) -- (4.318181818181818,1.004132231404959) -- (4.338383838383838,1.0543311906948272) -- (4.358585858585859,1.1057545148454244) -- (4.378787878787879,1.1584022038567496) -- (4.3989898989899,1.212274257728804) -- (4.41919191919192,1.267370676461586) -- (4.4393939393939394,1.3236914600550964) -- (4.45959595959596,1.3812366085093364) -- (4.47979797979798,1.4400061218243039) -- (4.5,1.5);
\draw [thick,dashed,blue] (2.5,0.3239999999999996) -- (2.5202020202020203,0.2977006427915514) -- (2.5404040404040407,0.2721359044995406) -- (2.5606060606060606,0.24730578512396684) -- (2.580808080808081,0.22321028466482984) -- (2.601010101010101,0.1998494031221303) -- (2.621212121212121,0.17722314049586751) -- (2.6414141414141414,0.15533149678604175) -- (2.6616161616161618,0.1341744719926532) -- (2.6818181818181817,0.11375206611570232) -- (2.7020202020202024,0.094064279155188) -- (2.7222222222222223,0.07511111111111068) -- (2.7424242424242427,0.056892561983471035) -- (2.762626262626263,0.039408631772268166) -- (2.782828282828283,0.022659320477502076) -- (2.803030303030303,0.006644628099173211) -- (2.8232323232323235,-0.00863544536271843) -- (2.8434343434343434,-0.023180899908172847) -- (2.8636363636363638,-0.03699173553719037) -- (2.883838383838384,-0.050067952249770786) -- (2.904040404040404,-0.062409550045913864) -- (2.9242424242424243,-0.07401652892562005) -- (2.9444444444444446,-0.08488888888888924) -- (2.9646464646464645,-0.09502662993572109) -- (2.984848484848485,-0.10442975206611582) -- (3.005050505050505,-0.11309825528007367) -- (3.0252525252525255,-0.12103213957759429) -- (3.045454545454546,-0.1282314049586779) -- (3.0656565656565657,-0.13469605142332441) -- (3.085858585858586,-0.1404260789715337) -- (3.1060606060606064,-0.14542148760330587) -- (3.1262626262626263,-0.14968227731864114) -- (3.1464646464646466,-0.1532084481175391) -- (3.166666666666667,-0.15600000000000008) -- (3.186868686868687,-0.15805693296602402) -- (3.207070707070707,-0.15937924701561068) -- (3.2272727272727275,-0.15996694214876034) -- (3.2474747474747474,-0.159820018365473) -- (3.2676767676767677,-0.15893847566574837) -- (3.287878787878788,-0.15732231404958685) -- (3.3080808080808084,-0.1549715335169881) -- (3.3282828282828287,-0.15188613406795226) -- (3.3484848484848486,-0.14806611570247935) -- (3.368686868686869,-0.14351147842056927) -- (3.3888888888888893,-0.13822222222222227) -- (3.409090909090909,-0.13219834710743802) -- (3.4292929292929295,-0.1254398530762167) -- (3.44949494949495,-0.11794674012855824) -- (3.4696969696969697,-0.10971900826446274) -- (3.48989898989899,-0.10075665748393015) -- (3.5101010101010104,-0.0910596877869605) -- (3.5303030303030307,-0.08062809917355361) -- (3.5505050505050506,-0.06946189164370971) -- (3.570707070707071,-0.057561065197428776) -- (3.5909090909090913,-0.044925619834710606) -- (3.611111111111111,-0.03155555555555549) -- (3.6313131313131315,-0.017450872359963093) -- (3.651515151515152,-0.0026115702479337227) -- (3.6717171717171717,0.012962350780532685) -- (3.691919191919192,0.02927089072543637) -- (3.7121212121212124,0.04631404958677697) -- (3.7323232323232327,0.0640918273645549) -- (3.7525252525252526,0.08260422405876969) -- (3.772727272727273,0.10185123966942179) -- (3.7929292929292933,0.1218328741965108) -- (3.813131313131313,0.14254912764003685) -- (3.8333333333333335,0.16400000000000028) -- (3.853535353535354,0.18618549127640058) -- (3.873737373737374,0.20910560146923818) -- (3.893939393939394,0.23276033057851267) -- (3.9141414141414144,0.2571496786042245) -- (3.9343434343434347,0.28227364554637313) -- (3.9545454545454546,0.3081322314049589) -- (3.974747474747475,0.334725436179982) -- (3.9949494949494953,0.362053259871442) -- (4.015151515151516,0.39011570247933935) -- (4.0353535353535355,0.41891276400367344) -- (4.055555555555555,0.44844444444444465) -- (4.075757575757576,0.47871074380165335) -- (4.095959595959596,0.5097116620752988) -- (4.116161616161617,0.5414471992653817) -- (4.136363636363637,0.5739173553719011) -- (4.156565656565657,0.6071221303948582) -- (4.176767676767677,0.6410615243342521) -- (4.196969696969697,0.6757355371900831) -- (4.217171717171717,0.7111441689623513) -- (4.237373737373738,0.7472874196510565) -- (4.257575757575758,0.7841652892561991) -- (4.277777777777778,0.8217777777777783) -- (4.2979797979797985,0.8601248852157951) -- (4.318181818181818,0.8992066115702485) -- (4.338383838383838,0.939022956841139) -- (4.358585858585859,0.9795739210284672) -- (4.378787878787879,1.020859504132232) -- (4.3989898989899,1.0628797061524342) -- (4.41919191919192,1.1056345270890733) -- (4.4393939393939394,1.1491239669421494) -- (4.45959595959596,1.1933480257116629) -- (4.47979797979798,1.2383067033976132) -- (4.5,1.2840000000000003);
\fill (3.9,0.24000000000000005) circle (0.05);
\node at (3.5,-1) {\small strongly convex};
\end{tikzpicture}

%% file: plots/gd_far.tex
\begin{tikzpicture}[scale=0.5]
\draw [thick,black!24] (1.299,0.750) -- (1.277,0.781) -- (1.251,0.810) -- (1.219,0.835) -- (1.182,0.856) -- (1.141,0.875) -- (1.094,0.889) -- (1.044,0.901) -- (0.989,0.908) -- (0.931,0.912) -- (0.868,0.912) -- (0.802,0.909) -- (0.733,0.901) -- (0.661,0.891) -- (0.586,0.876) -- (0.509,0.858) -- (0.430,0.837) -- (0.349,0.812) -- (0.267,0.784) -- (0.183,0.753) -- (0.099,0.719) -- (0.015,0.682) -- (-0.070,0.642) -- (-0.154,0.600) -- (-0.238,0.555) -- (-0.321,0.508) -- (-0.402,0.459) -- (-0.482,0.408) -- (-0.560,0.355) -- (-0.635,0.301) -- (-0.708,0.246) -- (-0.779,0.190) -- (-0.846,0.133) -- (-0.909,0.075) -- (-0.969,0.017) -- (-1.025,-0.041) -- (-1.077,-0.098) -- (-1.125,-0.156) -- (-1.168,-0.212) -- (-1.207,-0.268) -- (-1.240,-0.323) -- (-1.269,-0.377) -- (-1.292,-0.429) -- (-1.310,-0.479) -- (-1.323,-0.527) -- (-1.331,-0.573) -- (-1.333,-0.617) -- (-1.330,-0.658) -- (-1.322,-0.697) -- (-1.308,-0.733) -- (-1.289,-0.766) -- (-1.265,-0.796) -- (-1.235,-0.823) -- (-1.201,-0.846) -- (-1.162,-0.866) -- (-1.118,-0.882) -- (-1.070,-0.895) -- (-1.017,-0.905) -- (-0.960,-0.910) -- (-0.900,-0.912) -- (-0.836,-0.911) -- (-0.768,-0.905) -- (-0.697,-0.896) -- (-0.624,-0.884) -- (-0.548,-0.868) -- (-0.470,-0.848) -- (-0.390,-0.825) -- (-0.308,-0.799) -- (-0.225,-0.769) -- (-0.141,-0.736) -- (-0.057,-0.701) -- (0.028,-0.662) -- (0.112,-0.621) -- (0.196,-0.577) -- (0.279,-0.531) -- (0.361,-0.483) -- (0.442,-0.433) -- (0.521,-0.381) -- (0.598,-0.328) -- (0.672,-0.273) -- (0.744,-0.218) -- (0.813,-0.161) -- (0.878,-0.104) -- (0.940,-0.046) -- (0.998,0.012) -- (1.052,0.070) -- (1.102,0.127) -- (1.147,0.184) -- (1.188,0.241) -- (1.224,0.296) -- (1.255,0.350) -- (1.281,0.403) -- (1.302,0.454) -- (1.317,0.503) -- (1.328,0.550) -- (1.333,0.595) -- (1.332,0.638) -- (1.327,0.678) -- (1.315,0.716) -- (1.299,0.750) -- cycle;
\draw [thick,black!36] (2.382,1.375) -- (2.342,1.433) -- (2.293,1.485) -- (2.234,1.530) -- (2.167,1.570) -- (2.091,1.604) -- (2.007,1.631) -- (1.914,1.651) -- (1.814,1.665) -- (1.706,1.672) -- (1.592,1.672) -- (1.471,1.666) -- (1.344,1.653) -- (1.212,1.633) -- (1.075,1.606) -- (0.933,1.574) -- (0.788,1.534) -- (0.640,1.489) -- (0.489,1.438) -- (0.336,1.381) -- (0.182,1.318) -- (0.027,1.250) -- (-0.128,1.177) -- (-0.283,1.099) -- (-0.436,1.017) -- (-0.588,0.931) -- (-0.737,0.841) -- (-0.883,0.747) -- (-1.026,0.651) -- (-1.165,0.552) -- (-1.299,0.451) -- (-1.427,0.347) -- (-1.550,0.243) -- (-1.667,0.137) -- (-1.777,0.031) -- (-1.880,-0.075) -- (-1.975,-0.180) -- (-2.063,-0.286) -- (-2.142,-0.390) -- (-2.212,-0.492) -- (-2.274,-0.592) -- (-2.326,-0.690) -- (-2.369,-0.786) -- (-2.402,-0.878) -- (-2.426,-0.966) -- (-2.440,-1.051) -- (-2.444,-1.131) -- (-2.439,-1.207) -- (-2.423,-1.278) -- (-2.398,-1.344) -- (-2.363,-1.405) -- (-2.319,-1.459) -- (-2.265,-1.508) -- (-2.202,-1.551) -- (-2.130,-1.588) -- (-2.050,-1.618) -- (-1.961,-1.642) -- (-1.865,-1.659) -- (-1.761,-1.669) -- (-1.650,-1.673) -- (-1.532,-1.670) -- (-1.408,-1.660) -- (-1.279,-1.643) -- (-1.144,-1.620) -- (-1.005,-1.591) -- (-0.861,-1.555) -- (-0.714,-1.512) -- (-0.565,-1.464) -- (-0.413,-1.410) -- (-0.259,-1.350) -- (-0.105,-1.285) -- (0.051,-1.214) -- (0.205,-1.139) -- (0.360,-1.059) -- (0.512,-0.974) -- (0.663,-0.886) -- (0.811,-0.794) -- (0.955,-0.699) -- (1.096,-0.602) -- (1.232,-0.501) -- (1.364,-0.399) -- (1.490,-0.295) -- (1.610,-0.190) -- (1.723,-0.085) -- (1.829,0.022) -- (1.929,0.128) -- (2.020,0.233) -- (2.103,0.338) -- (2.178,0.441) -- (2.244,0.542) -- (2.301,0.642) -- (2.349,0.738) -- (2.387,0.832) -- (2.415,0.922) -- (2.434,1.009) -- (2.443,1.092) -- (2.443,1.170) -- (2.432,1.243) -- (2.412,1.312) -- (2.382,1.375) -- cycle;
\draw [thick,black!48] (3.681,2.125) -- (3.619,2.214) -- (3.543,2.294) -- (3.453,2.365) -- (3.349,2.427) -- (3.232,2.478) -- (3.101,2.520) -- (2.958,2.551) -- (2.803,2.573) -- (2.637,2.584) -- (2.460,2.584) -- (2.273,2.574) -- (2.077,2.554) -- (1.873,2.523) -- (1.661,2.483) -- (1.443,2.432) -- (1.218,2.371) -- (0.989,2.301) -- (0.756,2.222) -- (0.520,2.134) -- (0.281,2.037) -- (0.042,1.932) -- (-0.198,1.819) -- (-0.437,1.699) -- (-0.674,1.572) -- (-0.908,1.438) -- (-1.139,1.299) -- (-1.365,1.155) -- (-1.586,1.006) -- (-1.800,0.853) -- (-2.007,0.696) -- (-2.206,0.537) -- (-2.396,0.375) -- (-2.576,0.212) -- (-2.746,0.049) -- (-2.905,-0.115) -- (-3.053,-0.279) -- (-3.188,-0.441) -- (-3.310,-0.602) -- (-3.419,-0.760) -- (-3.514,-0.915) -- (-3.595,-1.067) -- (-3.661,-1.214) -- (-3.713,-1.356) -- (-3.749,-1.493) -- (-3.771,-1.624) -- (-3.777,-1.748) -- (-3.769,-1.866) -- (-3.745,-1.975) -- (-3.706,-2.077) -- (-3.652,-2.171) -- (-3.583,-2.255) -- (-3.500,-2.331) -- (-3.403,-2.397) -- (-3.292,-2.454) -- (-3.168,-2.500) -- (-3.031,-2.537) -- (-2.882,-2.563) -- (-2.721,-2.579) -- (-2.550,-2.585) -- (-2.368,-2.580) -- (-2.176,-2.565) -- (-1.976,-2.540) -- (-1.768,-2.504) -- (-1.553,-2.458) -- (-1.331,-2.403) -- (-1.104,-2.337) -- (-0.873,-2.263) -- (-0.638,-2.179) -- (-0.401,-2.086) -- (-0.162,-1.985) -- (0.078,-1.876) -- (0.318,-1.760) -- (0.556,-1.636) -- (0.791,-1.506) -- (1.024,-1.369) -- (1.253,-1.228) -- (1.476,-1.081) -- (1.694,-0.930) -- (1.905,-0.775) -- (2.108,-0.617) -- (2.302,-0.456) -- (2.487,-0.294) -- (2.663,-0.131) -- (2.827,0.033) -- (2.981,0.197) -- (3.122,0.360) -- (3.250,0.522) -- (3.366,0.682) -- (3.468,0.838) -- (3.556,0.992) -- (3.630,1.141) -- (3.689,1.286) -- (3.733,1.426) -- (3.762,1.560) -- (3.776,1.687) -- (3.775,1.808) -- (3.759,1.922) -- (3.727,2.027) -- (3.681,2.125) -- cycle;
\draw [thick,black!60] (4.763,2.750) -- (4.684,2.865) -- (4.586,2.969) -- (4.469,3.061) -- (4.334,3.140) -- (4.182,3.207) -- (4.013,3.261) -- (3.828,3.302) -- (3.627,3.329) -- (3.412,3.344) -- (3.183,3.344) -- (2.942,3.331) -- (2.688,3.305) -- (2.424,3.265) -- (2.150,3.213) -- (1.867,3.147) -- (1.577,3.069) -- (1.280,2.978) -- (0.978,2.875) -- (0.672,2.761) -- (0.364,2.636) -- (0.054,2.500) -- (-0.256,2.354) -- (-0.565,2.198) -- (-0.872,2.034) -- (-1.175,1.861) -- (-1.474,1.681) -- (-1.767,1.494) -- (-2.052,1.302) -- (-2.329,1.104) -- (-2.597,0.901) -- (-2.855,0.695) -- (-3.101,0.486) -- (-3.334,0.275) -- (-3.554,0.063) -- (-3.760,-0.149) -- (-3.951,-0.361) -- (-4.125,-0.571) -- (-4.283,-0.779) -- (-4.424,-0.984) -- (-4.547,-1.185) -- (-4.652,-1.381) -- (-4.738,-1.571) -- (-4.805,-1.755) -- (-4.852,-1.933) -- (-4.880,-2.102) -- (-4.888,-2.263) -- (-4.877,-2.414) -- (-4.846,-2.556) -- (-4.796,-2.688) -- (-4.726,-2.809) -- (-4.637,-2.919) -- (-4.530,-3.016) -- (-4.404,-3.102) -- (-4.260,-3.175) -- (-4.100,-3.236) -- (-3.923,-3.283) -- (-3.730,-3.317) -- (-3.522,-3.338) -- (-3.300,-3.346) -- (-3.064,-3.339) -- (-2.816,-3.320) -- (-2.557,-3.287) -- (-2.288,-3.241) -- (-2.009,-3.182) -- (-1.723,-3.110) -- (-1.429,-3.025) -- (-1.130,-2.928) -- (-0.826,-2.820) -- (-0.518,-2.700) -- (-0.209,-2.569) -- (0.101,-2.428) -- (0.411,-2.277) -- (0.719,-2.117) -- (1.024,-1.949) -- (1.325,-1.772) -- (1.621,-1.589) -- (1.910,-1.399) -- (2.192,-1.203) -- (2.465,-1.003) -- (2.727,-0.798) -- (2.979,-0.591) -- (3.219,-0.381) -- (3.446,-0.169) -- (3.659,0.043) -- (3.857,0.255) -- (4.040,0.466) -- (4.206,0.675) -- (4.356,0.882) -- (4.488,1.085) -- (4.602,1.283) -- (4.697,1.477) -- (4.774,1.664) -- (4.831,1.845) -- (4.869,2.018) -- (4.887,2.183) -- (4.885,2.340) -- (4.864,2.487) -- (4.823,2.624) -- (4.763,2.750) -- cycle;
\draw [black,->] (3.4,2.7) -- (2.5889310072136666,0.9896308515531562);
\draw [black,->] (3.4,2.7) -- (2.6757976362102327,1.1329940205434923);
\draw [black,->] (3.4,2.7) -- (1.5597149759272395,0.8583846385000713);
\draw [black,->] (3.4,2.7) -- (1.1568302918638587,1.2787388217160482);
\draw [black,->] (3.4,2.7) -- (2.4493405666753745,2.0043360026341);
\draw [blue,thick,->] (3.4,2.7) -- (2.2,1.4000000000000001);
\fill (3.4,2.7) circle [radius=2.5pt] node [above] {$\Bw_k$};
\fill (0,0) circle [radius=2.5pt] node [below] {$\Bw_*$};
\end{tikzpicture}

%% file: plots/gd_close.tex
\begin{tikzpicture}[scale=0.5]
\draw [thick,black!24] (1.299,0.750) -- (1.277,0.781) -- (1.251,0.810) -- (1.219,0.835) -- (1.182,0.856) -- (1.141,0.875) -- (1.094,0.889) -- (1.044,0.901) -- (0.989,0.908) -- (0.931,0.912) -- (0.868,0.912) -- (0.802,0.909) -- (0.733,0.901) -- (0.661,0.891) -- (0.586,0.876) -- (0.509,0.858) -- (0.430,0.837) -- (0.349,0.812) -- (0.267,0.784) -- (0.183,0.753) -- (0.099,0.719) -- (0.015,0.682) -- (-0.070,0.642) -- (-0.154,0.600) -- (-0.238,0.555) -- (-0.321,0.508) -- (-0.402,0.459) -- (-0.482,0.408) -- (-0.560,0.355) -- (-0.635,0.301) -- (-0.708,0.246) -- (-0.779,0.190) -- (-0.846,0.133) -- (-0.909,0.075) -- (-0.969,0.017) -- (-1.025,-0.041) -- (-1.077,-0.098) -- (-1.125,-0.156) -- (-1.168,-0.212) -- (-1.207,-0.268) -- (-1.240,-0.323) -- (-1.269,-0.377) -- (-1.292,-0.429) -- (-1.310,-0.479) -- (-1.323,-0.527) -- (-1.331,-0.573) -- (-1.333,-0.617) -- (-1.330,-0.658) -- (-1.322,-0.697) -- (-1.308,-0.733) -- (-1.289,-0.766) -- (-1.265,-0.796) -- (-1.235,-0.823) -- (-1.201,-0.846) -- (-1.162,-0.866) -- (-1.118,-0.882) -- (-1.070,-0.895) -- (-1.017,-0.905) -- (-0.960,-0.910) -- (-0.900,-0.912) -- (-0.836,-0.911) -- (-0.768,-0.905) -- (-0.697,-0.896) -- (-0.624,-0.884) -- (-0.548,-0.868) -- (-0.470,-0.848) -- (-0.390,-0.825) -- (-0.308,-0.799) -- (-0.225,-0.769) -- (-0.141,-0.736) -- (-0.057,-0.701) -- (0.028,-0.662) -- (0.112,-0.621) -- (0.196,-0.577) -- (0.279,-0.531) -- (0.361,-0.483) -- (0.442,-0.433) -- (0.521,-0.381) -- (0.598,-0.328) -- (0.672,-0.273) -- (0.744,-0.218) -- (0.813,-0.161) -- (0.878,-0.104) -- (0.940,-0.046) -- (0.998,0.012) -- (1.052,0.070) -- (1.102,0.127) -- (1.147,0.184) -- (1.188,0.241) -- (1.224,0.296) -- (1.255,0.350) -- (1.281,0.403) -- (1.302,0.454) -- (1.317,0.503) -- (1.328,0.550) -- (1.333,0.595) -- (1.332,0.638) -- (1.327,0.678) -- (1.315,0.716) -- (1.299,0.750) -- cycle;
\draw [thick,black!36] (2.382,1.375) -- (2.342,1.433) -- (2.293,1.485) -- (2.234,1.530) -- (2.167,1.570) -- (2.091,1.604) -- (2.007,1.631) -- (1.914,1.651) -- (1.814,1.665) -- (1.706,1.672) -- (1.592,1.672) -- (1.471,1.666) -- (1.344,1.653) -- (1.212,1.633) -- (1.075,1.606) -- (0.933,1.574) -- (0.788,1.534) -- (0.640,1.489) -- (0.489,1.438) -- (0.336,1.381) -- (0.182,1.318) -- (0.027,1.250) -- (-0.128,1.177) -- (-0.283,1.099) -- (-0.436,1.017) -- (-0.588,0.931) -- (-0.737,0.841) -- (-0.883,0.747) -- (-1.026,0.651) -- (-1.165,0.552) -- (-1.299,0.451) -- (-1.427,0.347) -- (-1.550,0.243) -- (-1.667,0.137) -- (-1.777,0.031) -- (-1.880,-0.075) -- (-1.975,-0.180) -- (-2.063,-0.286) -- (-2.142,-0.390) -- (-2.212,-0.492) -- (-2.274,-0.592) -- (-2.326,-0.690) -- (-2.369,-0.786) -- (-2.402,-0.878) -- (-2.426,-0.966) -- (-2.440,-1.051) -- (-2.444,-1.131) -- (-2.439,-1.207) -- (-2.423,-1.278) -- (-2.398,-1.344) -- (-2.363,-1.405) -- (-2.319,-1.459) -- (-2.265,-1.508) -- (-2.202,-1.551) -- (-2.130,-1.588) -- (-2.050,-1.618) -- (-1.961,-1.642) -- (-1.865,-1.659) -- (-1.761,-1.669) -- (-1.650,-1.673) -- (-1.532,-1.670) -- (-1.408,-1.660) -- (-1.279,-1.643) -- (-1.144,-1.620) -- (-1.005,-1.591) -- (-0.861,-1.555) -- (-0.714,-1.512) -- (-0.565,-1.464) -- (-0.413,-1.410) -- (-0.259,-1.350) -- (-0.105,-1.285) -- (0.051,-1.214) -- (0.205,-1.139) -- (0.360,-1.059) -- (0.512,-0.974) -- (0.663,-0.886) -- (0.811,-0.794) -- (0.955,-0.699) -- (1.096,-0.602) -- (1.232,-0.501) -- (1.364,-0.399) -- (1.490,-0.295) -- (1.610,-0.190) -- (1.723,-0.085) -- (1.829,0.022) -- (1.929,0.128) -- (2.020,0.233) -- (2.103,0.338) -- (2.178,0.441) -- (2.244,0.542) -- (2.301,0.642) -- (2.349,0.738) -- (2.387,0.832) -- (2.415,0.922) -- (2.434,1.009) -- (2.443,1.092) -- (2.443,1.170) -- (2.432,1.243) -- (2.412,1.312) -- (2.382,1.375) -- cycle;
\draw [thick,black!48] (3.681,2.125) -- (3.619,2.214) -- (3.543,2.294) -- (3.453,2.365) -- (3.349,2.427) -- (3.232,2.478) -- (3.101,2.520) -- (2.958,2.551) -- (2.803,2.573) -- (2.637,2.584) -- (2.460,2.584) -- (2.273,2.574) -- (2.077,2.554) -- (1.873,2.523) -- (1.661,2.483) -- (1.443,2.432) -- (1.218,2.371) -- (0.989,2.301) -- (0.756,2.222) -- (0.520,2.134) -- (0.281,2.037) -- (0.042,1.932) -- (-0.198,1.819) -- (-0.437,1.699) -- (-0.674,1.572) -- (-0.908,1.438) -- (-1.139,1.299) -- (-1.365,1.155) -- (-1.586,1.006) -- (-1.800,0.853) -- (-2.007,0.696) -- (-2.206,0.537) -- (-2.396,0.375) -- (-2.576,0.212) -- (-2.746,0.049) -- (-2.905,-0.115) -- (-3.053,-0.279) -- (-3.188,-0.441) -- (-3.310,-0.602) -- (-3.419,-0.760) -- (-3.514,-0.915) -- (-3.595,-1.067) -- (-3.661,-1.214) -- (-3.713,-1.356) -- (-3.749,-1.493) -- (-3.771,-1.624) -- (-3.777,-1.748) -- (-3.769,-1.866) -- (-3.745,-1.975) -- (-3.706,-2.077) -- (-3.652,-2.171) -- (-3.583,-2.255) -- (-3.500,-2.331) -- (-3.403,-2.397) -- (-3.292,-2.454) -- (-3.168,-2.500) -- (-3.031,-2.537) -- (-2.882,-2.563) -- (-2.721,-2.579) -- (-2.550,-2.585) -- (-2.368,-2.580) -- (-2.176,-2.565) -- (-1.976,-2.540) -- (-1.768,-2.504) -- (-1.553,-2.458) -- (-1.331,-2.403) -- (-1.104,-2.337) -- (-0.873,-2.263) -- (-0.638,-2.179) -- (-0.401,-2.086) -- (-0.162,-1.985) -- (0.078,-1.876) -- (0.318,-1.760) -- (0.556,-1.636) -- (0.791,-1.506) -- (1.024,-1.369) -- (1.253,-1.228) -- (1.476,-1.081) -- (1.694,-0.930) -- (1.905,-0.775) -- (2.108,-0.617) -- (2.302,-0.456) -- (2.487,-0.294) -- (2.663,-0.131) -- (2.827,0.033) -- (2.981,0.197) -- (3.122,0.360) -- (3.250,0.522) -- (3.366,0.682) -- (3.468,0.838) -- (3.556,0.992) -- (3.630,1.141) -- (3.689,1.286) -- (3.733,1.426) -- (3.762,1.560) -- (3.776,1.687) -- (3.775,1.808) -- (3.759,1.922) -- (3.727,2.027) -- (3.681,2.125) -- cycle;
\draw [thick,black!60] (4.763,2.750) -- (4.684,2.865) -- (4.586,2.969) -- (4.469,3.061) -- (4.334,3.140) -- (4.182,3.207) -- (4.013,3.261) -- (3.828,3.302) -- (3.627,3.329) -- (3.412,3.344) -- (3.183,3.344) -- (2.942,3.331) -- (2.688,3.305) -- (2.424,3.265) -- (2.150,3.213) -- (1.867,3.147) -- (1.577,3.069) -- (1.280,2.978) -- (0.978,2.875) -- (0.672,2.761) -- (0.364,2.636) -- (0.054,2.500) -- (-0.256,2.354) -- (-0.565,2.198) -- (-0.872,2.034) -- (-1.175,1.861) -- (-1.474,1.681) -- (-1.767,1.494) -- (-2.052,1.302) -- (-2.329,1.104) -- (-2.597,0.901) -- (-2.855,0.695) -- (-3.101,0.486) -- (-3.334,0.275) -- (-3.554,0.063) -- (-3.760,-0.149) -- (-3.951,-0.361) -- (-4.125,-0.571) -- (-4.283,-0.779) -- (-4.424,-0.984) -- (-4.547,-1.185) -- (-4.652,-1.381) -- (-4.738,-1.571) -- (-4.805,-1.755) -- (-4.852,-1.933) -- (-4.880,-2.102) -- (-4.888,-2.263) -- (-4.877,-2.414) -- (-4.846,-2.556) -- (-4.796,-2.688) -- (-4.726,-2.809) -- (-4.637,-2.919) -- (-4.530,-3.016) -- (-4.404,-3.102) -- (-4.260,-3.175) -- (-4.100,-3.236) -- (-3.923,-3.283) -- (-3.730,-3.317) -- (-3.522,-3.338) -- (-3.300,-3.346) -- (-3.064,-3.339) -- (-2.816,-3.320) -- (-2.557,-3.287) -- (-2.288,-3.241) -- (-2.009,-3.182) -- (-1.723,-3.110) -- (-1.429,-3.025) -- (-1.130,-2.928) -- (-0.826,-2.820) -- (-0.518,-2.700) -- (-0.209,-2.569) -- (0.101,-2.428) -- (0.411,-2.277) -- (0.719,-2.117) -- (1.024,-1.949) -- (1.325,-1.772) -- (1.621,-1.589) -- (1.910,-1.399) -- (2.192,-1.203) -- (2.465,-1.003) -- (2.727,-0.798) -- (2.979,-0.591) -- (3.219,-0.381) -- (3.446,-0.169) -- (3.659,0.043) -- (3.857,0.255) -- (4.040,0.466) -- (4.206,0.675) -- (4.356,0.882) -- (4.488,1.085) -- (4.602,1.283) -- (4.697,1.477) -- (4.774,1.664) -- (4.831,1.845) -- (4.869,2.018) -- (4.887,2.183) -- (4.885,2.340) -- (4.864,2.487) -- (4.823,2.624) -- (4.763,2.750) -- cycle;
\draw [black,->] (0.6,0.7) -- (0.49546726387590856,0.33212615672075085);
\draw [black,->] (0.6,0.7) -- (0.09569663569633358,0.3879190442582673);
\draw [black,->] (0.6,0.7) -- (0.2059370069636699,0.47025056689750344);
\draw [black,->] (0.6,0.7) -- (0.20064400097371426,0.2075512978950243);
\fill (0.6,0.7) circle [radius=2.5pt] node [above] {$\Bw_k$};
\draw [blue,thick,->] (0.6,0.7) -- (0.3,0.29999999999999993);
\fill (0,0) circle [radius=2.5pt] node [below] {$\Bw_*$};
\end{tikzpicture}

%% file: WideNetworks.tex
\chapter{Wide neural networks and the neural tangent
  kernel}\label{chap:wideNets}
In this chapter we explore the dynamics of training (shallow) neural
networks of large width. Throughout assume given data pairs
\begin{subequations}\label{eq:settingwide}
  \begin{equation}\label{eq:settingwidedata}
    (\Bx_i,y_i)\in\R^d\times \R\qquad i\in\{1,\dots,m\},
  \end{equation}
  for distinct $\Bx_i$. We wish to train a model (e.g.\ a neural
  network) $\Phi(\Bx,\Bw)$ depending on the input $\Bx\in\R^d$ and the
  parameters $\Bw\in\R^n$.  To this end we consider either
  minimization of the {\bf ridgeless} (unregularized) objective
  \begin{equation}\label{eq:ridgeless}
    \objF(\Bw)\dfn \sum_{i=1}^m (\Phi(\Bx_i,\Bw)-y_i)^2, 
  \end{equation}
  or, for some regularization parameter $\lambda\ge 0$, of the {\bf
    ridge} regularized objective
  \begin{equation}\label{eq:settingwiderisk}
    \objF_\lambda(\Bw)\dfn \objF(\Bw) + \lambda \norm{\Bw}^2.
  \end{equation}
\end{subequations}
The adjectives ridge and ridgeless thus indicate the presence or
absence of the regularization term $\norm{\Bw}^2$.

In the ridgeless case, the objective is a multiple of the empirical
risk $\widehat{\mathcal{R}}_S(\Phi)$ in \eqref{eq:empiricalRiskDef0}
for the sample $S = (\Bx_i,y_i)_{i=1}^m$ and the square-loss.
Regularization is a common tool in machine learning to improve model
generalization and stability, e.g.\ \cite{NIPS1991_8eefcfdf}.  The
goal of this chapter is to get some insight into the dynamics of
$\Phi(\Bx,\Bw_k)$ as the parameter vector $\Bw_k$ progresses during
training. Additionally, we want to gain some intuition about the
influence of regularization, and the behavior of the trained model
$\Bx\mapsto \Phi(\Bx,\Bw_k)$ for large $k$. We do so through the lens
of so-called kernel methods. As a training algorithm we exclusively
focus on gradient descent with constant step size.

If $\Phi(\Bx,\Bw)$ depends linearly on the parameters $\Bw$, the
objective function \eqref{eq:settingwiderisk} is convex. As
established in the previous chapter (cf.~Remark \ref{rmk:convex}),
gradient descent then finds a global minimizer. For typical neural
network architectures, $\Bw\mapsto \Phi(\Bx,\Bw)$ is not linear, and
such a statement is in general not true.  Recent results have shown
that neural network behavior tends to linearize in $\Bw$ as network
width increases \cite{jacot2018neural}. This allows to transfer some
of the techniques and statements from the linear case to the training
of neural networks.

We start this chapter in Sections \ref{sec:linreg} and
\ref{sec:kernelreg} by recalling (kernel) least-squares methods, which
describe linear (in $\Bw$) models.  Following \cite{lee2019wide}, the
subsequent sections examine why neural networks exhibit linear-like
behavior in the infinite-width limit.  In Section \ref{sec:ntk} we
introduce the so-called tangent kernel. Section \ref{sec:globmin}
presents abstract results showing, under suitable assumptions,
convergence towards a global minimizer when training the
model. Section \ref{sec:proximity} builds on this analysis and
discusses connections to kernel regression with the tangent kernel.
In Section \ref{sec:ntkdynamics} we then detail the implications for
wide neural networks.  A similar treatment of these results was
previously given by Telgarsky in \cite[Chapter 8]{telgarskynotes} for
gradient flow (rather than gradient descent), based on
\cite{NEURIPS2019_ae614c55}.

\section{Linear least-squares regression}\label{sec:linreg}
Arguably one of the simplest machine learning algorithms is linear
least-squares regression, e.g.,
\cite{demmel97,doi:10.1137/1.9781611971484,hastie_09_elements-of.statistical-learning,golub}.
Given data \eqref{eq:settingwidedata}, %
we fit a linear function $\Bx\mapsto\Phi(\Bx,\Bw)\dfn \Bx^\top\Bw$
by minimizing $\objF$ or $\objF_\lambda$ in \eqref{eq:settingwide}.
With
\begin{equation}\label{eq:Ay}
  \BA = \begin{pmatrix}
          \Bx_1^\top\\
          \vdots\\
          \Bx_m^\top
        \end{pmatrix}\in\R^{m\times d}
        \qquad\text{and}\qquad
        \By = \begin{pmatrix}
                y_1\\
                \vdots\\
                y_m
              \end{pmatrix}\in\R^m
            \end{equation}
            it holds
            \begin{equation}\label{eq:ThetaLSQ}
              \objF(\Bw)=\norm[]{\BA\Bw-\By}^2\qquad\text{and}\qquad\objF_\lambda(\Bw)=\objF(\Bw)+\lambda\norm{\Bw}^2.
            \end{equation}
            The $\Bx_1,\dots,\Bx_m$ are referred to as the {\bf
              training points} (or design points), and throughout the
            rest of Section \ref{sec:linreg}, we denote their span by
            \begin{equation}\label{eq:tildeH}
              \tilde H \dfn {\rm span}\{\Bx_1,\dots,\Bx_m\}\subseteq\R^d.
            \end{equation}
            This is the subspace spanned by the rows of $\BA$.

\begin{remark}\label{rmk:bias}
  More generally, %
  the ansatz $\Phi(\Bx,(\Bw,b))\dfn\Bw^\top \Bx+b$ corresponds to
  \begin{equation*}
    \Phi(\Bx,(\Bw,b))=(1,\Bx^\top)
    \begin{pmatrix}b\\\Bw\end{pmatrix}.
  \end{equation*}
  Therefore, additionally allowing for a bias can be treated %
  similarly.
\end{remark}

\subsection{Existence of minimizers}
We start with the ridgeless case $\lambda=0$. The model
$\Phi(\Bx,\Bw)=\Bx^\top\Bw$ is linear in both $\Bx$ and $\Bw$. In
particular, $\Bw\mapsto \objF(\Bw)$ is a convex function by Exercise
\ref{ex:quadraticobjective}.  If $\BA$ is invertible, then $\objF$ has
the unique minimizer $\Bw_*=\BA^{-1}\By$.  If ${\rm rank}(\BA)=d$,
then $\objF$ is strongly convex by Exercise
\ref{ex:quadraticobjective}, and there still exists a unique
minimizer. If however ${\rm rank}(\BA)<d$, then
$\ker(\BA)\neq\{\Bnul\}$ and there exist infinitely many minimizers of
$\objF$.  To guarantee uniqueness, we consider the {\bf minimum norm
  solution}
\begin{equation}\label{eq:min2norm}
  \Bw_*\dfn\argmin_{\Bw\in M}\norm[]{\Bw},\qquad
  M\dfn \set{\Bw\in\R^d}{\objF(\Bw)\le\objF(\Bv)~\forall\Bv\in\R^d}.
\end{equation}
It is a standard result that $\Bw_*$ is well-defined and belongs to
the span $\tilde H$ of the training points defined in
\eqref{eq:tildeH}, e.g.,
\cite{doi:10.1137/1.9781611971484,demmel97,golub}.  While one way to
prove this is through the pseudoinverse (see Theorem
\ref{thm:pseudoinverse}), we provide an alternative argument here,
which can be directly extended to the infinite-dimensional case as
discussed in Section \ref{sec:kernelreg} ahead.

\begin{theorem}\label{thm:min2norm}
  There is a unique minimum norm solution $\Bw_*\in\R^d$ in
  \eqref{eq:min2norm}.  It lies in the subspace $\tilde H$, and is the
  unique minimizer of $\objF$ in $\tilde H$, i.e.
  \begin{equation}\label{eq:repfinw}
    \Bw_*=\argmin_{\tilde\Bw\in \tilde H}%
    \objF(\tilde\Bw).
  \end{equation}
\end{theorem}

\begin{proof}
  We start with existence and uniqueness of $\Bw_*\in \R^d$ in
  \eqref{eq:min2norm}.  Let
  \begin{equation*}
    C\dfn {\rm span}\setc{\BA\Bw}{\Bw\in \R^d}\subseteq\R^m.
  \end{equation*}
  Then $C$ is a finite dimensional space, and as such it is closed and
  convex. Therefore
  $\By_*=\argmin_{\tilde\By\in C}\norm[]{\tilde\By-\By}$ exists and is
  unique (this is a fundamental property of Hilbert spaces, see
  Theorem \ref{thm:uniqueProjection}). In particular, the set
  $M=\set{\Bw\in \R^d}{\BA\Bw=\By_*}\subseteq \R^d$ of minimizers of
  $\objF$ is not empty. Clearly $M\subseteq \R^d$ is closed and
  convex.  As before, $\Bw_*=\argmin_{\Bw\in M}\norm[]{\Bw}$ exists
  and is unique.

  It remains to show \eqref{eq:repfinw}.  Decompose
  $\Bw_*=\tilde\Bw+\hat\Bw$ with $\tilde\Bw\in\tilde H$ and
  $\hat\Bw\in\tilde H^\perp$ (see Definition \ref{def:orthogonal}).
  By definition of $\BA$ it holds $\BA\Bw_*=\BA\tilde\Bw$ and
  $\objF(\Bw_*)=\objF(\tilde\Bw)$. Moreover
  $\norm[]{\Bw_*}^2=\norm[]{\tilde\Bw}^2+\norm[]{\hat\Bw}^2$.  Since
  $\Bw_*$ is the minimum norm solution, $\Bw_*=\tilde\Bw\in\tilde H$.
  To conclude the proof, we need to show that $\Bw_*$ is the only
  minimizer of $\objF$ in $\tilde H$. Assume there exists a minimizer
  $\Bv$ of $\objF$ in $\tilde H$ different from $\Bw_*$. Then
  $\Bnul\neq \Bw_*-\Bv\in\tilde H$. Thus $\BA(\Bw_*-\Bv)\neq\Bnul$ and
  $\By_*=\BA\Bw_*\neq \BA\Bv$, which contradicts that $\Bv$ minimizes
  $\objF$.
\end{proof}

Next let $\lambda>0$ in \eqref{eq:ThetaLSQ}.  Then minimizing
$\objF_\lambda$ is referred to as ridge regression or Tikhonov
regularized least squares
\cite{tikhonov1963regularization,Hoerl1,engl2000regularization,hastie_09_elements-of.statistical-learning}. The
next proposition shows that there exists a unique minimizer of
$\objF_\lambda$, which is closely connected to the minimum norm
solution, e.g.\ \cite[Theorem 5.2]{engl2000regularization}.

  \begin{theorem}\label{thm:limridge}
    Let $\lambda>0$. Then, with $\objF_\lambda$ in
    \eqref{eq:ThetaLSQ}, there exists a unique minimizer
    \begin{equation*}
      \Bw_{*,\lambda}\dfn \argmin_{\Bw\in\R^d} \objF_\lambda(\Bw).
    \end{equation*}
    It holds $\Bw_{*,\lambda}\in\tilde H$, and
    \begin{equation}\label{eq:limBwstarlambda}
      \lim_{\lambda\to 0}\Bw_{*,\lambda} = \Bw_*.
    \end{equation}
  \end{theorem}
  \begin{proof}
    According to Exercise \ref{ex:regularization},
    $\Bw\mapsto\objF_\lambda(\Bw)$ is strongly convex on $\R^d$, and
    thus also on the subspace $\tilde H\subseteq\R^d$. Therefore,
    there exists a unique minimizer of $\objF_\lambda$ in $\tilde H$,
    which we denote by $\Bw_{*,\lambda}\in\tilde H$. To show that
    there exists no other minimizer of $\objF_\lambda$ in $\R^d$, fix
    $\Bw\in \R^d\backslash\tilde H$ and decompose
    $\Bw=\tilde\Bw+\hat\Bw$ with $\tilde\Bw\in\tilde H$ and
    $\Bnul\neq\hat\Bw\in\tilde H^\perp$. Then
    \begin{equation*}
      \objF(\Bw)=\norm{\BA\Bw-\By}^2=\norm{\BA\tilde\Bw-\By}^2=\objF(\tilde\Bw)
    \end{equation*}
    and
    \begin{equation*}
      \norm{\Bw}^2=\norm{\tilde\Bw}^2+\norm{\hat\Bw}^2 > \norm{\tilde\Bw}^2.
    \end{equation*}
    Thus
    $\objF_\lambda(\Bw)>\objF_\lambda(\tilde\Bw)\ge
    \objF_\lambda(\Bw_{*,\lambda})$.

    It remains to show \eqref{eq:limBwstarlambda}. We have
    \begin{align*}
      \objF_\lambda(\Bw) &= (\BA\Bw-\By)^\top(\BA\Bw-\By)+\lambda\Bw^\top\Bw\\
                         &=\Bw^\top(\BA^\top\BA+\lambda\BI_d)\Bw-2\Bw^\top\BA^\top\By,
    \end{align*}
    where $\BI_d\in\R^{d\times d}$ is the identity matrix. The
    minimizer is reached at $\nabla \objF_\lambda(\Bw)=0$, which
    yields
    \begin{equation*}
      \Bw_{*,\lambda} = (\BA^\top\BA+\lambda\BI_d)^{-1}\BA^\top\By.
    \end{equation*}
    Let $\BA=\BU\BSigma\BV^\top$ be the singular value decomposition
    of $\BA$, where $\BSigma\in\R^{m\times d}$ contains the nonzero
    singular values $s_1\ge \dots \ge s_r>0$, and
    $\BU\in\R^{m\times m}$, $\BV\in\R^{d\times d}$ are
    orthogonal. Then
    \begin{align*}
      \Bw_{*,\lambda} &= (\BV(\BSigma^\top\BSigma+\lambda\BI_d)\BV^\top)^{-1}\BV\BSigma^\top\BU^\top\By\\
                      &=\BV\underbrace{\begin{pmatrix}
                                         \frac{s_1}{s_1^2+\lambda} &&&\\
                                                                   &\ddots&&\Bnul\\
                                                                   &&\frac{s_r}{s_r^2+\lambda}&\\
                                                                   &\Bnul&&\Bnul\\
                                       \end{pmatrix}}_{\in\R^{d\times m}}
                        \BU^\top\By,
    \end{align*}
    where $\Bnul$ stands for a zero block of suitable size. As
    $\lambda\to 0$, this converges towards $\BA^\dagger\By$, where
    $\BA^\dagger$ denotes the pseudoinverse of $\BA$, see
    \eqref{eq:pseudoinverse}. By Theorem \ref{thm:pseudoinverse},
    $\BA^\dagger\By=\Bw_*$.
  \end{proof}

\subsection{Gradient descent}\label{sec:gdlinlsq}
Consider gradient descent to minimize the objective $\objF_\lambda$ in
\eqref{eq:ThetaLSQ}. Starting with $\Bw_0\in\R^d$, the iterative
update with constant step size $h>0$ reads
\begin{equation}\label{eq:gdlsq}
  \Bw_{k+1} = \Bw_k-2h\BA^\top(\BA\Bw_k-\By)-2h\lambda\Bw_k\qquad\text{for all }k\in\N_0.
\end{equation}

Let again first $\lambda=0$, i.e.\ $\objF_\lambda=\objF$. Since
$\objF$ is convex and quadratic, by Remark \ref{rmk:convex} for
sufficiently small step size $h>0$ it holds
$\objF(\Bw_k)\to \objF(\Bw_*)$ as $k\to\infty$.  Is it also true that
$\Bw_k$ converges to the minimum norm solution $\Bw_*\in\tilde H$?
Recall that $\tilde H$ is spanned by the columns of $\BA^\top$. Thus,
if $\Bw_0\in\tilde H$, then by \eqref{eq:gdlsq}, the iterates $\Bw_k$
never leave the subspace $\tilde H$. Since there is only one minimizer
in $\tilde H$, it follows that $\Bw_k\to\Bw_*$ as $k\to\infty$.

This shows that gradient descent does not find an arbitrary optimum
when minimizing $\objF$, but converges towards the minimum norm
solution as long as $\Bw_0\in\tilde H$ (e.g.\ $\Bw_0=\Bnul$). %
It is well-known \cite[Theorem 16]{doi:10.1137/0111051}, that
iterations of type \eqref{eq:gdlsq} lead to minimum norm solutions as
made more precise by the next proposition. To state it, we write in
the following $s_{\rm max}(\BA)$ for the maximal singular value of
$\BA$, and $s_{\rm min}(\BA)$ for the minimal positive singular value,
with the convention $s_{\rm min}(\BA)\dfn \infty$ in case $\BA=0$.
The full proof is left as Exercise \ref{ex:minnorm}.

\begin{proposition}\label{prop:LSQ} 
  Let $\lambda=0$ and fix $h\in (0,s_{\rm max}(\BA)^{-2})$.  Let
  $\Bw_0=\tilde\Bw_0+\hat\Bw_0$ where $\tilde \Bw_0 \in\tilde H$ and
  $\hat\Bw_0\in\tilde H^\perp$, and let $(\Bw_k)_{k\in\N}$ be defined
  by \eqref{eq:gdlsq}.  Then
  \begin{equation*}
    \lim_{k\to\infty}\Bw_k=\Bw_*+\hat\Bw_0.
  \end{equation*}
\end{proposition}

Next we consider ridge regression, where $\lambda>0$ in
\eqref{eq:ThetaLSQ}, \eqref{eq:gdlsq}.  The condition on the step size
in the next proposition can be relaxed to
$h\in (0,(\lambda+s_{\rm max}(\BA)^2)^{-1})$, but we omit doing so for
simplicity.

\begin{proposition}\label{prop:ridgemin}
  Let $\lambda>0$, and fix
  $h\in (0,(2\lambda+2s_{\rm max}(\BA)^2)^{-1})$.  Let $\Bw_0\in\R^d$
  and let $(\Bw_k)_{k\in\N}$ be defined by \eqref{eq:gdlsq}. Then
  \begin{equation*}
    \lim_{k\to\infty}\Bw_k=\Bw_{*,\lambda}
  \end{equation*}
  and
  \begin{equation*}
    \norm{\Bw_*-\Bw_{*,\lambda}}\le \frac{\lambda}{s_{\rm min}(\BA)^3+s_{\rm min}(\BA)\lambda}\norm{\By}= O(\lambda)\qquad\text{as }\lambda\to 0.
  \end{equation*}
\end{proposition}
\begin{proof}
  By Exercise \ref{ex:quadraticobjective}, $\objF_\lambda$ is
  $(2\lambda+2s_{\rm max}(\BA)^2)$-smooth, and by Exercise
  \ref{ex:regularization}, $\objF_\lambda$ is strongly convex. Thus
  Theorem \ref{thm:GDsc} implies convergence of gradient descent
  towards the unique minimizer $\Bw_{*,\lambda}$.

  For the bound on the distance to $\Bw_*$, assume $\BA\neq 0$ (the
  case $\BA=0$ is trivial).  Expressing $\Bw_*$ via the pseudoinverse
  of $\BA$ (see Appendix \ref{app:pseudo}) we get
  \begin{equation*}
    \Bw_*=\BA^\dagger\By
    =\BV\begin{pmatrix}
          \frac{1}{s_1} &&&\\
                        &\ddots&&\Bnul\\
                        &&\frac{1}{s_r}&\\
                        &\Bnul&&\Bnul\\
        \end{pmatrix}
        \BU^\top\By,
      \end{equation*}
      where $\BA=\BU\BSigma\BV^\top$ is the singular value
      decomposition of $\BA$, and $s_1\ge\dots\ge s_r>0$ denote the
      singular values of $\BA$. The explicit formula for
      $\Bw_{*,\lambda}$ obtained in the proof of Theorem
      \ref{thm:limridge} then yields
      \begin{equation*}
        \norm{\Bw_*-\Bw_{*,\lambda}}\le \max_{i\le r} \Big|\frac{s_i}{s_i^2+\lambda}-\frac{1}{s_i}\Big|\norm{\By}.
      \end{equation*}
      This gives the claimed bound.
    \end{proof}

    By Proposition \ref{prop:ridgemin}, if we use ridge regression
    with a small regularization parameter $\lambda>0$, then gradient
    descent converges to a vector $\Bw_{*,\lambda}$ which is
    $O(\lambda)$ close to the minimum norm solution $\Bw_*$,
    regardless of the initialization $\Bw_0$.

\section{Feature methods and kernel least-squares regression}\label{sec:kernelreg} 
Linear models are often too simplistic to capture the true
relationship between $\Bx$ and $y$. Feature- and kernel-based methods
(e.g.,
\cite{cristianini2000introduction,Scholkopf2002,hastie_09_elements-of.statistical-learning})
address this by replacing $\Bx \mapsto \dup{\Bx}{\Bw}$ with
$\Bx \mapsto \dup{\phi(\Bx)}{\Bw}$ where $\phi:\R^d \to \R^n$ is a
(typically nonlinear) map. This introduces nonlinearity in $\Bx$ while
retaining linearity in the parameter $\Bw\in\R^n$. %

\begin{example}
  Let data $(x_i,y_i)_{i=1}^m\subseteq \R\times\R$ be given, and
  define for $x\in\R$
  \begin{equation*}
    \phi(x)\dfn (1,x,\dots,x^{n-1})^\top\in\R^n.
  \end{equation*}
  For $\Bw\in\R^n$, the model
  $x\mapsto \dup{\phi(x)}{\Bw}=\sum_{j=0}^{n-1}w_j x^j$ can represent
  any polynomial of degree $n-1$.
\end{example}

Let us formalize this idea. For reasons that will become apparent
later (see Remark \ref{rmk:mercer}), it is useful to allow for the
case $n=\infty$.  To this end, let $(H,\dup[H]{\cdot}{\cdot})$ be a
Hilbert space (see Appendix \ref{app:hilbert}), referred to as the
{\bf feature space}, and let $\phi:\R^d\to H$ denote the {\bf feature
  map}. The model is defined as
\begin{equation}\label{eq:KernelPredictor}
  \Phi(\Bx,w)\dfn \inp[H]{\phi(\Bx)}{w}
\end{equation}
with $w \in H$.  We may think of $H$ in the following either as $\R^n$
for some $n\in\N$, or as $\ell^2(\N)$ (see Example \ref{ex:hilbert});
in this case the components of $\phi$ are referred to as {\bf
  features}.  For some $\lambda\ge0$, the goal is to minimize the
objective
\begin{equation}\label{eq:kernelreg}
  \objF(w)\dfn \sum_{j=1}^{m}\big(\inp[H]{\phi(\Bx_j)}{w}-y_j\big)^2
  \qquad\text{or}
  \qquad
  \objF_\lambda(w)\dfn \objF(w)+\lambda\norm[H]{w}^2.
\end{equation}

In analogy to \eqref{eq:tildeH}, throughout the rest of Section
\ref{sec:kernelreg} denote by
\begin{equation*}
  \tilde H\dfn {\rm span}\{\phi(\Bx_1),\dots,\phi(\Bx_m)\}\subseteq H
\end{equation*}
the space spanned by the feature vectors at the training points.

\subsection{Existence of minimizers}
We start with the ridgeless case $\lambda=0$ in \eqref{eq:kernelreg}.
To guarantee uniqueness and regularize the problem, we again consider
the minimum norm solution
\begin{equation}\label{eq:minHnorm}
  w_*\dfn \argmin_{w\in M}\norm[H]{w},\qquad
  M\dfn \set{w\in H}{\objF(w)\le\objF(v)~\forall v\in H}.
\end{equation}

\begin{theorem}\label{thm:rep}
  There is a unique minimum norm solution $w_*\in H$ in
  \eqref{eq:minHnorm}.  It lies in the subspace $\tilde H$, and %
  is the unique minimizer of $\objF$ in $\tilde H$, i.e.
  \begin{equation}\label{eq:repw}
    w_* = \argmin_{\tilde w\in\tilde H}\objF(\tilde w).
  \end{equation}
\end{theorem}

The proof of Theorem \ref{thm:min2norm} is formulated such that it
extends verbatim to Theorem \ref{thm:rep}, upon replacing $\R^d$ with
$H$ and the matrix $\BA \in \R^{m \times d}$ with the linear map
\begin{align*}
  A:&H\to \R^m\\
    &w\mapsto (\dup[H]{\phi(\Bx_i)}{w})_{i=1}^m.
\end{align*}

For the case of ridge regression with $\lambda>0$ in
\eqref{eq:kernelreg} we let
\begin{equation}\label{eq:limBwstar}
  w_{*,\lambda}\dfn \argmin_{w\in H} \objF_\lambda(w).
\end{equation}
Similar as in the ridgeless case, Theorem \ref{thm:limridge} extends
to the current setting with small modifications. In particular
$w_{*,\lambda}\in H$ as in \eqref{eq:limBwstar} exists and is
unique. The key observation is once more that any minimizer of
$\objF_\lambda$ must belong to the \emph{finite-dimensional} subspace
$\tilde H$, since for $w=\tilde w+w^\perp$ with
$\tilde w\in \tilde H$, $w^\perp\in \tilde H^\perp$, the orthogonal
component $w^\perp$ increases the regularization term
$\lambda\norm[H]{w}^2=\lambda(\norm[H]{\tilde
  w}^2+\norm[H]{w^\perp}^2)$ but has no effect on
$\objF(w)$. Selecting a basis for $\tilde H$, the proof then proceeds
analogously. We leave it to the reader to check this, see Exercise
\ref{ex:ridge}. This leads to the following statement.
  
  \begin{theorem}\label{thm:rep2}
    Let $\lambda>0$. Then, %
    there exists a unique minimizer $w_{*,\lambda}$ in
    \eqref{eq:limBwstar}.  It holds $w_{*,\lambda}\in\tilde H$, and
    \begin{equation*}
      \lim_{\lambda\to 0}w_{*,\lambda} = w_*.
    \end{equation*}
  \end{theorem}

  Statements as in Theorems \ref{thm:rep} and \ref{thm:rep2}, which
  yield that the minimizer is attained in the finite dimensional
  subspace $\tilde H$, are known in the literature as {\bf representer
    theorems}, \cite{KimeldorfGeorgeS.1970ACBB,SchoelkopfEtAl:01}.

\subsection{The kernel trick}\label{sec:kerneltrick}
We now explain the connection to \emph{kernels}.  At first glance, %
minimizing \eqref{eq:kernelreg} in the potentially
infinite-dimensional Hilbert space $H$ seems infeasible. However, we
have already seen that the minimizer is taken in the finite
dimensional subspace $\tilde H$. Reducing the computations to this
subspace is known as the so-called kernel trick \cite{boser1992}.  To
treat the cases $\lambda=0$ and $\lambda>0$ simultaneously, we use the
notation $w_{*,0}\dfn w_*$ in the following.

\begin{definition}\label{def:kernel}
  A symmetric function $K:\R^d\times\R^d\to\R$ is called a {\bf
    kernel}, if for any $\Bx_1,\dots,\Bx_k\in\R^d$, $k\in\N$, the {\bf
    kernel matrix} $\BG=(K(\Bx_i,\Bx_j))_{i,j=1}^k\in\R^{k\times k}$
  is symmetric positive semidefinite.
\end{definition}

Given a feature map $\phi:\R^d\to H$, it is easy to check that
\begin{equation}\label{eq:featurekernel}
  K(\Bx,\Bz)\dfn \inp[H]{\phi(\Bx)}{\phi(\Bz)}\qquad\text{for all }
  \Bx,\Bz\in\R^d,
\end{equation}
defines a kernel.  The corresponding kernel matrix
$\BG\in\R^{{m}\times {m}}$ is
\begin{equation*}
  G_{ij}=\inp[H]{\phi(\Bx_i)}{\phi(\Bx_j)}=K(\Bx_i,\Bx_j).
\end{equation*}
The ansatz $\sum_{j=1}^{m}\alpha_j\phi(\Bx_j)$ for $w_{*,\lambda}$
then turns the optimization problems \eqref{eq:minHnorm} (for
$\lambda=0$) and \eqref{eq:limBwstar} into
\begin{equation}\label{eq:kernelargmin}
  \argmin_{\Balpha\in\R^m}\norm[]{\BG\Balpha-\By}^2 + \lambda \Balpha^\top\BG\Balpha.
\end{equation}
Such a minimizing $\Balpha$ need not be unique (if $\BG$ is not
regular), however, any such $\Balpha$ yields a minimizer in
$\tilde H$, and thus $w_{*,\lambda}=\sum_{j=1}^{m}\alpha_j\phi(\Bx_j)$
for any $\lambda\ge 0$ by Theorems \ref{thm:rep} and \ref{thm:rep2}.
This suggests Algorithm \ref{alg:KLSR}.

\begin{algorithm}[htb]
  \caption{Kernel least-squares regression}\label{alg:KLSR}
  \begin{algorithmic}
    \STATE \textbf{Input:} Data $(\Bx_i,y_i)_{i=1}^m\in\R^d\times \R$,
    kernel $K:\R^d\times\R^d\to \R$, regularization parameter
    $\lambda\ge 0$, evaluation point $\Bx\in\R^d$ \STATE
    \textbf{Output:} (Ridge or ridgeless) kernel least squares
    estimator at $\Bx$ \STATE \STATE compute the kernel matrix
    $\BG=(K(\Bx_i,\Bx_j))_{i,j=1}^m$ \STATE determine a minimizer
    $\Balpha\in\R^m$ of
    $\norm[]{\BG\Balpha-\By}^2+\lambda\Balpha^\top\BG\Balpha$ \STATE
    evaluate $\Phi(\Bx,w_{*,\lambda})$ via
    \begin{equation}\label{eq:klsestimator}
      \Phi(\Bx,w_{*,\lambda})=\inpc[H]{\phi(\Bx)}{\sum_{j=1}^{m}\alpha_j\phi(\Bx_j)}=\sum_{j=1}^{m}\alpha_jK(\Bx,\Bx_j)
    \end{equation}  
  \end{algorithmic}
\end{algorithm}

We refer to
\begin{equation*}
  \Bx\mapsto \Phi(\Bx,w_{*,\lambda})= \dup[H]{\phi(\Bx)}{w_{*,\lambda}}
\end{equation*}
as the {\bf (ridge or ridgeless) kernel least-squares estimator}.  By
the above considerations, its computation %
neither requires explicit knowledge of the feature map $\phi$ nor of
$w_{*,\lambda}\in H$. It is sufficient to choose a kernel
$K:\R^d\times\R^d\to\R$ and perform all computations in finite
dimensional spaces. This is known as the {\bf kernel trick}. While
Algorithm \ref{alg:KLSR} will not play a role in the rest of the
chapter, we present it here to give a more complete picture.

  \begin{remark}\label{rmk:KLSexplicit}
    Let
    \begin{subequations}\label{eq:Knotation}
      \begin{align}
        K(\Bx,\BX)&\dfn (K(\Bx,\Bx_1),\dots,K(\Bx,\Bx_m))\in\R^{1\times m}\\
        K(\BX,\BX)&\dfn (K(\Bx_i,\Bx_j))_{i,j=1}^m=\BG\in\R^{m\times m}.
      \end{align}
    \end{subequations}
    If $\lambda\ge 0$ and $K(\BX,\BX)+\lambda\VI_{m}$ is regular
    (which is always true for $\lambda>0$), then one minimizer of
    \eqref{eq:kernelargmin} is given by
    \begin{equation}\label{eq:Balphadirect}
      \Balpha = (K(\BX,\BX)+\lambda\BI_m)^{-1}\By.
    \end{equation}
    For $\lambda=0$ this follows directly by
    \eqref{eq:kernelargmin}. The case $\lambda>0$ is left as Exercise
    \ref{ex:Balphadirect}. Using the representation
    \eqref{eq:klsestimator}, the kernel least-squares estimator at
    $\Bx\in\R^d$ can thus be expressed
    \begin{equation}\label{eq:KLSexplicit}
      K(\Bx,\BX)(K(\BX,\BX)+\lambda\VI_m)^{-1}\By.
    \end{equation}
  \end{remark}
  
\begin{remark}\label{rmk:mercer}
  If $\Omega\subseteq\R^d$ is compact and $K:\Omega\times\Omega\to\R$
  is a continuous kernel, then Mercer's theorem implies existence of a
  Hilbert space $H$ and a feature map $\phi:\R^d\to H$ such that
  \begin{equation*}
    K(\Bx,\Bz)=\inp[H]{\phi(\Bx)}{\phi(\Bz)}\qquad\text{for all }\Bx,\Bz\in\Omega,
  \end{equation*}
  i.e.\ $K$ is the corresponding kernel. See for instance
  \cite[Sec.~3.2]{MR2239907} or \cite[Thm.~4.49]{steinwart}.
\end{remark}

\subsection{Gradient descent}\label{sec:kernelgd} 
In practice we may either minimize $\objF_\lambda$ in
\eqref{eq:kernelreg} (in the Hilbert space $H$) or the objective in
\eqref{eq:kernelargmin} (in $\R^m$). We now focus on the former, as
this will allow to draw connections to neural network training in the
subsequent sections. In order to use gradient descent, we assume
$H=\R^n$ equipped with the Euclidean inner product.

Initializing $\Bw_0\in\R^n$, gradient descent with constant step size
$h>0$ to minimize $\objF_\lambda$ reads
\begin{equation*}
  \Bw_{k+1} = \Bw_k-2h\BA^\top(\BA\Bw_k-\By)-2h\lambda\Bw_k\qquad\text{for all }k\in\N_0,
\end{equation*}
where now
\begin{equation*}
  \BA=\begin{pmatrix}
        \phi(\Bx_1)^\top\\
        \vdots\\
        \phi(\Bx_m)^\top
      \end{pmatrix}.
    \end{equation*}
    This corresponds to the situation discussed in Section
    \ref{sec:gdlinlsq}.

    Let $\lambda=0$. For sufficiently small step size, by Proposition
    \ref{prop:LSQ} for $\Bx\in\R^d$
    \begin{equation}\label{eq:gdkernel}
      \lim_{k\to\infty}\Phi(\Bx,\Bw_k)=\dup{\phi(\Bx)}{\Bw_*}+\dup{\phi(\Bx)}{\hat\Bw_0},
    \end{equation}
    where
    \begin{equation*}
      \Bw_0=\tilde\Bw_0+\hat\Bw_0
    \end{equation*}
    with
    $\tilde\Bw_0\in\tilde H={\rm
      span}\{\phi(\Bx_1),\dots,\phi(\Bx_m)\}\subseteq\R^n$, and
    $\hat\Bw_0\in\tilde H^\perp$.  For $\lambda=0$, gradient descent
    thus yields the ridgeless kernel least squares estimator plus an
    additional term $\dup{\phi(\Bx)}{\hat\Bw_0}$ depending on
    initialization.  Notably, on the set
    \begin{equation}\label{eq:gdkls}
      \set{\Bx\in\R^d}{\phi(\Bx)\in {\rm span}\{\phi(\Bx_1),\dots,\phi(\Bx_m)\}},
    \end{equation}
    \eqref{eq:gdkernel} always coincides with the ridgeless least
    squares estimator.
  
    Now let $\lambda>0$. For sufficiently small step size, by
    Proposition \ref{prop:ridgemin} for $\Bx\in\R^d$
    \begin{equation*}
      \lim_{k\to\infty}\Phi(\Bx,\Bw_k)=\dup{\phi(\Bx)}{\Bw_{*,\lambda}}= \dup{\phi(\Bx)}{\Bw_{*}}+O(\lambda)\qquad\text{as }\lambda\to 0.
    \end{equation*}
    Thus, for $\lambda>0$ gradient descent determines the ridge kernel
    least-squares estimator regardless of the
    initialization. Moreover, for fixed $\Bx$, the limiting model is
    $O(\lambda)$ close to the ridgeless kernel least-squares
    estimator.

  \section{Tangent kernel}\label{sec:ntk}
  Consider a general model $\Phi(\Bx,\Bw)$ with input $\Bx\in\R^d$ and
  parameters $\Bw\in\R^n$.  The goal is to minimize the square loss
  objective \eqref{eq:ridgeless} given data
  \eqref{eq:settingwidedata}. Our analysis in this and the following
  two sections focuses on the ridgeless case. We will revisit %
  ridge regression in Section \ref{sec:gaussianprocesses}, where we
  consider a simple test example of training a neural network with and
  without regularization.

  If $\Bw\mapsto \Phi(\Bx,\Bw)$ is not linear, then unlike in Sections
  \ref{sec:linreg} and \ref{sec:kernelreg}, the objective function
  \eqref{eq:ridgeless} is in general not convex, and most results on
  first order methods in Chapter \ref{chap:training} are not directly
  applicable. We thus simplify the situation by \emph{linearizing the
    model in the parameter $\Bw\in\R^n$ around initialization}: Fixing
  $\Bw_0\in\R^n$, let
   
  \begin{equation}\label{eq:Philin}
    \Phi^{\rm lin}(\Bx,\Bp)\dfn \Phi(\Bx,\Bw_0)+\dup{\nabla_\Bw\Phi(\Bx,\Bw_0)}{\Bp}
    \qquad\text{for all }\Bp\in\R^n,
  \end{equation}
  which is the first order Taylor approximation of $\Phi$ around the
  initial parameter $\Bw_0$. The parameters of the linearized model
  will always be denoted by $\Bp\in\R^n$ to distinguish them from the
  parameters $\Bw$ of the full model.  Introduce
  \begin{equation}\label{eq:deltaj}
    \delta_j\dfn y_j-\Phi(\Bx_j,\Bw_0)\qquad\text{for all }j=1,\dots,m.
  \end{equation}
  The square loss objective for the linearized model then reads
  \begin{align}\label{eq:risklin}
    \objF^{\rm lin}(\Bp)&\dfn \sum_{j=1}^m(\Phi^{\rm lin}(\Bx_j,\Bp)-y_j)^2%
                          = \sum_{j=1}^m \big(\dup{\nabla_\Bw\Phi(\Bx_j,\Bw_0)}{\Bp} - \delta_j\big)^2
  \end{align}
  where $\dup{\cdot}{\cdot}$ stands for the Euclidean inner product in
  $\R^n$.  Comparing with \eqref{eq:kernelreg}, %
  minimizing $\objF^{\rm lin}$ corresponds to kernel least squares
  regression with feature map
  \begin{equation}\label{eq:linfeatures}
    \phi(\Bx)=\nabla_\Bw\Phi(\Bx,\Bw_0)\in\R^n.
  \end{equation}
  By \eqref{eq:featurekernel} the corresponding kernel is
  \begin{equation}\label{eq:etk}
    \hat K_n(\Bx,\Bz) = \dup{\nabla_\Bw\Phi(\Bx,\Bw_0)}{\nabla_\Bw\Phi(\Bz,\Bw_0)}.
  \end{equation}
  We refer to $\hat K_n$ as the {\bf empirical tangent kernel}, as it
  arises from the first order Taylor approximation (the tangent) of
  the original model $\Phi$ around initialization $\Bw_0$. Note that
  $\hat K_n$ depends on the choice of $\Bw_0$.
  
  We point out that %
  based on the observations in Section \ref{sec:kernelgd}, minimizing
  $\objF^{\rm lin}$ with gradient descent initialized with
  $\Bp_0=\Bnul$, sufficiently small step size, and no regularization,
  yields a sequence $(\Bp_k)_{k\in\N_0}$ satisfying
  \begin{equation}\label{eq:limitlinearized}
    \lim_{k\to\infty}\Phi^{\rm lin}(\Bx,\Bp_k) =\underbrace{\Phi(\Bx,\Bw_0)}_{\text{term depending on initialization}} + \underbrace{\dup{\phi(\Bx)}{\Bp_*}}_{\substack{\text{ridgeless kernel least-squares} \\ \text{estimator with kernel $\hat K_n$ and RHS \eqref{eq:deltaj}}}}
  \end{equation}
  The first term depends on initialization $\Bw_0$. The second term
  also depends on $\Bw_0$ through the feature map in
  \eqref{eq:linfeatures} and through the right-hand side
  \eqref{eq:deltaj}.  Here we used the definition of $\Phi^{\rm lin}$
  in \eqref{eq:Philin}, the limit \eqref{eq:gdkernel}, and the fact
  that $\Bp_0=\Bnul$ so that the second term in \eqref{eq:gdkernel}
  vanishes.
  
  \section{Global minimizers}\label{sec:globmin}
  Consider a general model $\Phi:\R^d\times\R^n\to\R$, data as in
  \eqref{eq:settingwidedata}, and the ridgeless square loss
  \begin{equation*}
    \objF(\Bw) = \sum_{j=1}^m (\Phi(\Bx_j,\Bw)-y_j)^2.
  \end{equation*}
  In this section we discuss sufficient conditions under which
  gradient descent converges to a global minimizer.

  The idea is as follows: if $\Bw \mapsto \Phi(\Bx, \Bw)$ is nonlinear
  but sufficiently \emph{close to its linearization} $\Phi^{\rm lin}$
  in \eqref{eq:Philin} within some region, the objective function
  behaves almost like a convex function there. If the \emph{region is
    large enough} to contain both the initial value $\Bw_0$ and a
  global minimum, then we expect gradient descent to never leave this
  (almost convex) basin during training and find a global minimizer.

  To illustrate this, consider Figures \ref{fig:ntk1} and
  \ref{fig:ntk2} where we set the number of training samples to $m=1$
  and the number of parameters to $n=1$. For the above reasoning to
  hold, the difference between $\Phi$ and $\Phi^{\rm lin}$, as well as
  the difference in their derivatives, must remain small within a
  neighborhood of $\Bw_0$.  The %
  neighborhood should be large enough to contain the global minimizer,
  and thus depends critically on two factors: the initial error
  $\Phi(\Bx_1,w_0)-y_1$, and %
  the magnitude of the derivative
  $\frac{\partial}{\partial w}\Phi(\Bx_1,w_0)$.

    \begin{figure}
      \centering \input{plots/ntk.tex}
      \caption{Graph of $w\mapsto \Phi(\Bx_1,w)$ and the linearization
        $p\mapsto \Phi^{\rm lin}(\Bx_1,p)$ at the initial parameter
        $w_0$, s.t.\
        $\frac{\partial}{\partial w}\Phi(\Bx_1,w_0)\neq 0$. If $\Phi$
        and $\Phi^{\rm lin}$ are close, then there exists $w$ s.t.\
        $\Phi(\Bx_1,w)=y_1$ (left). If the derivatives are also close,
        the loss $(\Phi(\Bx_1,w)-y_1)^2$ is nearly convex in $w$, and
        gradient descent finds a global minimizer
        (right).}\label{fig:ntk1}
    \end{figure}
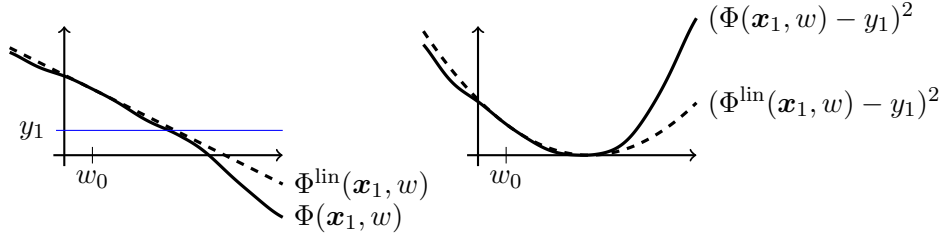
  
    \begin{figure}
      \centering \input{plots/ntk2.tex}
      \caption{Same as Figure \ref{fig:ntk1}. If $\Phi$ and
        $\Phi^{\rm lin}$ are not close, there need not exist $w$ such
        that $\Phi(\Bx_1,w)=y_1$, and gradient descent need not
        converge to a global minimizer.}\label{fig:ntk2}
    \end{figure}
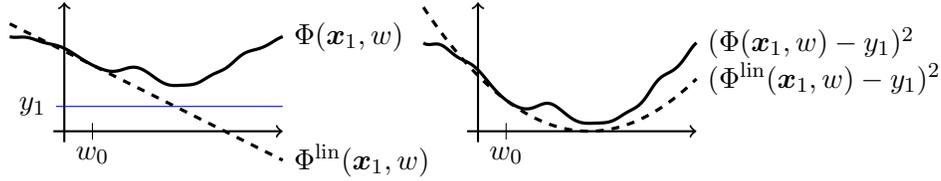

    For general $m$ and $n$, we now make the required assumptions on
    $\Phi$ precise.

  \begin{assumption}\label{ass:ntk}
    Let $\Phi\in C^1(\R^d\times\R^n)$ and $\Bw_0\in\R^n$. There exist
    constants $r,R,U,L>0$ and
    $0<\theta_{\rm min}\le \theta_{\rm max}<\infty$ such that
    $\norm{\Bx_i}\le R$ for all $i=1,\dots,m$, and it holds that
    \begin{enumerate}[label=(\alph*)]
    \item\label{item:ntk1} %
      the kernel matrix of the empirical tangent kernel
      \begin{equation}\label{eq:JBw0}
        (\hat K_n(\Bx_i,\Bx_j))_{i,j=1}^m = \big(\dup{\nabla_\Bw\Phi(\Bx_i,\Bw_0)}{{\nabla_\Bw\Phi(\Bx_j,\Bw_0)}}\big)_{i,j=1}^m\in\R^{m\times m}
      \end{equation}
      is regular and its eigenvalues belong to
      $[\theta_{\rm min},\theta_{\rm max}]$,
    \item\label{item:ntk2} for all %
      $\Bx\in\R^d$ with $\norm{\Bx}\le R$
      \begin{subequations}\label{eq:Jbound}
        \begin{align}\label{eq:JboundU}
          \norm[]{\nabla_\Bw\Phi(\Bx,\Bw)}&\le U &&\text{for all }\Bw\in B_r(\Bw_0)\\ \label{eq:JboundL}
          \norm[]{\nabla_\Bw\Phi(\Bx,\Bw)-\nabla_\Bw\Phi(\Bx,\Bv)}&\le L\norm[]{\Bw-\Bv} &&\text{for all }\Bw,~\Bv\in B_r(\Bw_0),
        \end{align}
      \end{subequations}
    \item\label{item:ntk3}
      \begin{equation}\label{eq:Lrcond}
        L\le \frac{\theta_{\rm min}^2}{8m^{3/2} U^2 \sqrt{\objF(\Bw_0)}}
        \qquad\text{and}\qquad
        r=\frac{2 \sqrt{m} U\sqrt{\objF(\Bw_0)}}{\theta_{\rm min}}.
      \end{equation}
    \end{enumerate}
  \end{assumption}

  Let us give more intuitive explanations of these technical
  assumptions: First, \ref{item:ntk1} implies that
  $(\nabla_\Bw\Phi(\Bx_i,\Bw_0)^\top)_{i=1}^m\in\R^{m\times n}$ has
  full rank $m\le n$ (thus we have at least as many parameters $n$ as
  training data $m$). In the context of Figure \ref{fig:ntk1}, this
  means that $\frac{\partial}{\partial w}\Phi(\Bx_1,w_0)\neq 0$ and
  thus $\Phi^{\rm lin}$ is a not a constant function. This guarantees
  existence of $\Bp$ such that $\Phi^{\rm lin}(\Bx_i,\Bp)=y_i$ for all
  $i=1,\dots,m$, i.e.\ the linearized model $\Phi^{\rm lin}$ is
  capable of interpolating the data.  Next, \ref{item:ntk2} formalizes
  in particular the required closeness of $\Phi$ and its linearization
  $\Phi^{\rm lin}$. %
  For example, since $\Phi^{\rm lin}$ is the first order Taylor
  approximation of $\Phi$ at $\Bw_0$ (cf.~\eqref{eq:Philin}), %
  \begin{equation*}
    |\Phi(\Bx,\Bw)-\Phi^{\rm lin}(\Bx,\Bw-\Bw_0)| = |(\nabla_\Bw\Phi(\Bx,\tilde\Bw)-\nabla_\Bw\Phi(\Bx,\Bw_0))^\top(\Bw-\Bw_0)|\le L \norm{\Bw-\Bw_0}^2,
  \end{equation*}
  for some $\tilde\Bw$ in the convex hull of $\Bw$ and $\Bw_0$.
  Finally, \ref{item:ntk3} ties together all constants, ensuring the
  full model to be sufficiently close to its linearization in a large
  enough ball of radius $r$ around $\Bw_0$. Notably, $r$ may be
  smaller for smaller initial error $\sqrt{\objF(\Bw_0)}$ and for
  larger $\theta_{\rm min}$, which aligns with our intuition from
  Figure \ref{fig:ntk1}.
  
  We are now ready to state the following theorem, which is a variant
  of \cite[Thm.~G.1]{lee2019wide}. The proof closely follows the
  arguments given there. In Section \ref{sec:ntkdynamics} we will see
  that the theorem's main requirement---Assumption \ref{ass:ntk}---is
  satisfied with high probability for certain (wide) neural networks.

  \begin{theorem}\label{thm:ntk}
    Let Assumption \ref{ass:ntk} hold. %
    Fix a positive learning rate
    \begin{equation}\label{eq:hcond}
      h \le \frac{1}{\theta_{\rm min}+\theta_{\rm max}}.
    \end{equation}
    Let $(\Bw_k)_{k\in\N}$ be generated by gradient descent, i.e.\ for
    all $k\in\N_0$
    \begin{equation}\label{eq:ntkgd}
      \Bw_{k+1}=\Bw_k- h\nabla\objF(\Bw_k).
    \end{equation}

    It then holds for all $k\in\N$
    \begin{subequations}\label{eq:ntk}
      \begin{align}
        \norm[]{\Bw_k-\Bw_0}&\le %
                              r\label{eq:ntk1}\\        
        \objF(\Bw_k)&\le (1-h\theta_{\rm min})^{2k} \objF(\Bw_0)\label{eq:ntk2}.
      \end{align}
    \end{subequations}
  \end{theorem}

  \begin{proof}
    We start with some notation.  Let
    $\tilde\Bw^1,\dots,\tilde\Bw^m\in B_r(\Bw_0)\subseteq \R^n$
    arbitrary and set $\tilde\BW=(\tilde\Bw^1,\dots,\tilde\Bw^m)$.  We
    denote the model prediction error at the $m$ data points for these
    $m$ parameter vectors by
    \begin{equation*}
      \Be(\tilde\BW)\dfn
      \begin{pmatrix}
        \Phi(\Bx_1,\tilde\Bw^1)-y_1\\
        \vdots\\
        \Phi(\Bx_m,\tilde\Bw^m)-y_m
      \end{pmatrix}\in\R^m
      \quad\text{and}\quad
      \nabla \Be(\tilde\BW) \dfn \begin{pmatrix}
                                   \nabla_\Bw\Phi(\Bx_1,\tilde\Bw^1)^\top\\
                                   \vdots\\
                                   \nabla_\Bw\Phi(\Bx_m,\tilde\Bw^m)^\top
                                 \end{pmatrix}\in\R^{m\times n}.
                               \end{equation*}
                               In case
                               $\tilde\Bw^1=\dots=\tilde\Bw^m=\tilde\Bw$,
                               we simply write $\Be(\tilde\Bw)$ and
                               $\nabla\Be(\tilde\Bw)$ instead. With
                               the empirical tangent kernel $\hat K_n$
                               in Assumption \ref{ass:ntk}
                               \ref{item:ntk1} it then holds
                               \begin{equation}\label{eq:ntkhatKn}
                                 \nabla \Be(\Bw_0) \nabla \Be(\Bw_0)^\top
                                 =(\hat K_n(\Bx_i,\Bx_j))_{i,j=1}^m\in\R^{m\times m}.
                               \end{equation}

                               By \eqref{eq:JboundU}
                               \begin{subequations}\label{eq:nablaE}
                                 \begin{equation}\label{eq:nablaEbound}
                                   \norm[]{\nabla \Be(\tilde\BW)}^2
                                   \le
                                   \norm[F]{\nabla \Be(\tilde\BW)}^2
                                   =\sum_{j=1}^m\norm{\nabla_\Bw\Phi(\Bx_j,\tilde\Bw^j)}^2
                                   \le m U^2,
                                 \end{equation}
                                 and similarly, using
                                 \eqref{eq:JboundL}
                                 \begin{align}\label{eq:nablaELip}
                                   \norm[]{\nabla \Be(\tilde\BW)-\nabla \Be(\Bw_0)}^2&\le
                                                                                       \sum_{j=1}^m\norm[]{\nabla_\Bw\Phi(\Bx_j,\tilde\Bw^j)-\nabla_\Bw\Phi(\Bx_j,\Bw_0)}^2\nonumber\\
                                                                                     &\le L^2\sum_{j=1}^m\norm[]{\tilde\Bw^j-\Bw_0}^2\le m L^2 r^2.
                                 \end{align}
                               \end{subequations}

                               {\bf Step 1.}  Denote
                               $\rho\dfn 1-h\theta_{\rm min}\in
                               (0,1)$. In the remainder of the proof
                               we use induction over $k$ to show
                               \begin{subequations}\label{eq:ntkaux}
                                 \begin{align}
                                   \sum_{j=0}^{k-1}\norm[]{\Bw_{j+1}-\Bw_j}&\le 2h\sqrt{m}U\norm[]{\Be(\Bw_0)}
                                                                             \sum_{j=0}^{k-1}\rho^{j}\label{eq:ntkaux1},        \\
                                   \norm[]{\Be(\Bw_k)}^2&\le \norm[]{\Be(\Bw_0)}^2 \rho^{2k},\label{eq:ntkaux2}
                                 \end{align}
                               \end{subequations}
                               for all $k\in\N_0$ and where an empty
                               sum is understood as zero. Since,
                               $\sum_{j=0}^\infty \rho^j =
                               (1-\rho)^{-1}$, and
                               $\norm{\Be(\Bw)}=\sqrt{\objF(\Bw)}$,
                               using \eqref{eq:Lrcond} we have
                               \begin{equation}\label{eq:sumcrbound}
                                 2h\sqrt{m}U\norm[]{\Be(\Bw_0)}\sum_{j=0}^\infty \rho^j = 2h\sqrt{m}U\sqrt{\objF(\Bw_0)}\frac{1}{h\theta_{\rm min}}= r,
                               \end{equation}
                               these inequalities directly imply
                               \eqref{eq:ntk}.
    
                               For $k=0$, \eqref{eq:ntkaux} is
                               trivial. For the induction step, assume
                               \eqref{eq:ntkaux} holds for some
                               $k\in\N_0$.
    
                               {\bf Step 2.} We show
                               \eqref{eq:ntkaux1} for $k+1$.  The
                               induction assumption \eqref{eq:ntkaux1}
                               and \eqref{eq:sumcrbound} give
                               $\Bw_k\in B_r(\Bw_0)$. Next,
                               \begin{equation}\label{eq:nablariskBwk}
                                 \nabla\objF(\Bw_k) = \nabla (\Be(\Bw_k)^\top \Be(\Bw_k)) = 2\nabla \Be(\Bw_k)^\top \Be(\Bw_k).
                               \end{equation}
                               Using the iteration rule
                               \eqref{eq:ntkgd} and the bounds
                               \eqref{eq:nablaEbound} and
                               \eqref{eq:ntkaux2}
                               \begin{align*}
                                 \norm[]{\Bw_{k+1}-\Bw_k}&=2h\norm[]{\nabla \Be(\Bw_k)^\top \Be(\Bw_k)}\\
                                                         &\le 2h \sqrt{m}U \norm[]{\Be(\Bw_k)}\\
                                                         &\le 2h \sqrt{m}U \norm[]{\Be(\Bw_0)} \rho^{k}.
                               \end{align*}
                               This shows \eqref{eq:ntkaux1} for
                               $k+1$. In particular by
                               \eqref{eq:sumcrbound}
                               \begin{equation}\label{eq:Bwkp1Bwk}
                                 \Bw_{k+1},~\Bw_k \in B_r(\Bw_0).
                               \end{equation}

                               {\bf Step 3.} We show
                               \eqref{eq:ntkaux2} for $k+1$.  Denote
                               the components of $\Be:\R^n\to\R^m$ by
                               $e_1,\dots,e_m$.  Since each $e_i$ is
                               continuously differentiable, there
                               exist $\tilde\Bw_{k}^i$ in the convex
                               hull of $\Bw_k$ and $\Bw_{k+1}$, such
                               that
                               \begin{align*}
                                 e_i(\Bw_{k+1}) &=e_i(\Bw_{k})+\int_0^1 \nabla e_i(\Bw_k+t(\Bw_{k+1}-\Bw_k)) (\Bw_{k+1}-\Bw_k)\dd t\\
                                                &=e_i(\Bw_{k})+\nabla e_i(\tilde\Bw_{k}^i)^\top (\Bw_{k+1}-\Bw_k)\\
                                                &= e_i(\Bw_k)-h\nabla e_i(\tilde\Bw_{k}^i)^\top \nabla \objF(\Bw_k).
                               \end{align*}
                               With
                               $\tilde\BW_k\dfn
                               (\tilde\Bw_{k}^1,\dots,\tilde\Bw_{k}^m)$
                               we thus have by \eqref{eq:nablariskBwk}
                               \begin{align*}
                                 \Be(\Bw_{k+1}) &= \Be(\Bw_k) -2h\nabla \Be(\tilde\BW_k) \nabla \Be(\Bw_k)^\top \Be(\Bw_k)\\
                                                &= \big(\BI_m-2h\nabla \Be(\tilde\BW_k) \nabla \Be(\Bw_k)^\top\big)\Be(\Bw_k),
                               \end{align*}
                               where $\BI_m\in\R^{m\times m}$ is the
                               identity matrix. We wish to show that
                               \begin{equation}\label{eq:ntktoshow}
                                 \norm[]{\BI_m-2h\nabla \Be(\tilde\BW_k) \nabla \Be(\Bw_k)^\top}\le \rho,
                               \end{equation}
                               which then implies \eqref{eq:ntkaux2}
                               for $k+1$ and concludes the proof.

                               Using \eqref{eq:nablaE} and the fact
                               that
                               $\Bw_k,\tilde
                               \Bw_k^1,\dots,\tilde\Bw_k^m \in
                               B_r(\Bw_0)$ by \eqref{eq:Bwkp1Bwk},
                               \begin{align*}
                                 \norm[]{\nabla \Be(\tilde\BW_k) \nabla \Be(\Bw_k)^\top-\nabla \Be(\Bw_0) \nabla \Be(\Bw_0)^\top}&\le
                                                                                                                                   \norm[]{(\nabla \Be(\tilde\BW_k)-\nabla\Be(\Bw_0))\nabla\Be(\Bw_k)^\top}\\
                                                                                                                                 &\quad+\norm[]{\nabla \Be(\Bw_0) (\nabla \Be(\Bw_k)-\nabla \Be(\Bw_0))^\top}\\
                                                                                                                                 &\le 2mULr.
                               \end{align*}
                               Since the eigenvalues of
                               $\nabla \Be(\Bw_0)\nabla
                               \Be(\Bw_0)^\top$ belong to
                               $[\theta_{\rm min},\theta_{\rm max}]$
                               by \eqref{eq:ntkhatKn} and Assumption
                               \ref{ass:ntk} \ref{item:ntk1}, as long
                               as
                               $h\le (\theta_{\rm min}+\theta_{\rm
                                 max})^{-1}$, we have
                               \begin{align*}
                                 \norm[]{\BI_m-2h\nabla \Be(\tilde\BW_k) \nabla \Be(\Bw_k)^\top}
                                 &\le \norm[]{\BI_m-2h\nabla \Be(\Bw_0) \nabla \Be(\Bw_0)^\top}+4h m ULr\\
                                 &\le 1-2h\theta_{\rm min}+4h m ULr.
                               \end{align*}
                               Due to \eqref{eq:Lrcond}
                               \begin{align*}
                                 1-2h\theta_{\rm min}+4h m ULr
                                 &\le
                                   1-2h\theta_{\rm min}+4h m U\frac{\theta_{\rm min}^2}{8 m^{3/2} U^2\sqrt{\objF(\Bw_0)}}\frac{2\sqrt{m}U\sqrt{\objF(\Bw_0)}}{\theta_{\rm min}}\\
                                 &= 1-h\theta_{\rm min}= \rho,
                               \end{align*}
                               which concludes the proof.
                             \end{proof}

                             Let us emphasize that \eqref{eq:ntk2}
                             implies that gradient descent
                             \eqref{eq:ntkgd} achieves zero loss in
                             the limit. Consequently, the limiting
                             model interpolates the training
                             data. This shows in particular
                             convergence to a global minimizer for the
                             (generally nonconvex) optimization
                             problem of minimizing $\objF(\Bw)$.

\section{Proximity to trained linearized model}\label{sec:proximity} 
The analysis in Section \ref{sec:globmin} was based on the observation
that the linearization $\Phi^{\rm lin}$ closely mimics the behavior of
the full model $\Phi$ for parameters with distance at most $r$
(cf.~Assumption \ref{ass:ntk}) to the initial parameter
$\Bw_0$. Theorem \ref{thm:ntk} states that the parameters remain
within this range throughout training. This suggests that the
predictions of the trained full model
$\lim_{k\to\infty}\Phi(\Bx,\Bw_k)$, are similar to those of the
trained linear model $\lim_{k\to\infty}\Phi^{\rm lin}(\Bx,\Bp_k)$.  In
this section we formalize this statement.

\subsection{Evolution of model predictions}
We adopt again the notation $\Phi^{\rm lin}:\R^d\times \R^n\to\R$ from
\eqref{eq:Philin} to represent the linearization of
$\Phi:\R^d\times \R^n\to\R$ around $\Bw_0$.  The parameters of the
linearized model are represented by $\Bp \in \R^n$, and the
corresponding loss function is written as $\objF^{\rm lin}(\Bp)$, as
in \eqref{eq:risklin}.  Additionally, we define
$\BX\dfn (\Bx_1,\dots,\Bx_m)$ and let for
$\tilde\BW=(\tilde\Bw^1,\dots,\tilde\Bw^m)$
\begin{align*}
  \Phi(\BX,\tilde\BW)&\dfn(\Phi(\Bx_i,\tilde\Bw^i))_{i=1}^m\in\R^m\\
  \Phi^{\rm lin}(\BX,\Bp)&\dfn(\Phi^{\rm lin}(\Bx_i,\Bp))_{i=1}^m\in\R^m    
\end{align*}
to denote the predicted values at the training points
$\Bx_1,\dots,\Bx_m$ for given parameter choices
$\tilde\Bw^1,\dots,\tilde\Bw^m$, $\Bp\in\R^n$. Moreover
\begin{equation*}
  \nabla_\Bw \Phi(\BX,\tilde\BW) = \begin{pmatrix}
                                     \nabla_\Bw\Phi(\Bx_1,\tilde\Bw^1)^\top\\
                                     \vdots\\
                                     \nabla_\Bw\Phi(\Bx_m,\tilde\Bw^m)^\top
                                   \end{pmatrix}\in\R^{m\times n}.
                                 \end{equation*}
                                 In case
                                 $\tilde\Bw^1=\dots=\tilde\Bw^m=\tilde\Bw$,
                                 we simply write $\Phi(\BX,\tilde\Bw)$
                                 and $\nabla_\Bw\Phi(\BX,\tilde\Bw)$
                                 instead.  Similarly
                                 $\nabla_\Bw\Phi^{\rm lin}(\BX,\Bp)$
                                 is defined.  With this notation, the
                                 model predictions at $\Bx\in\R^d$ and
                                 $\BX$ evolve under gradient descent
                                 as follows:

                                 \begin{itemize}
                                 \item {\bf full model}: Initialize
                                   $\Bw_0\in\R^n$, and set for step
                                   size $h>0$ and all $k\in\N_0$
                                   \begin{equation}\label{eq:GDfullmodel}
                                     \Bw_{k+1}=\Bw_k-h\nabla_\Bw\objF(\Bw_k).
                                   \end{equation}
                                   Then
                                   \begin{equation*}
                                     \nabla_\Bw\objF(\Bw) = \nabla_\Bw\norm[]{\Phi(\BX,\Bw)-\By}^2
                                     = 2\nabla_\Bw\Phi(\BX,\Bw)^\top(\Phi(\BX,\Bw)-\By).
                                   \end{equation*}
                                   Thus
                                   \begin{align*}
                                     \Phi(\Bx,\Bw_{k+1})&=
                                                          \Phi(\Bx,\Bw_{k})+(\nabla_\Bw\Phi(\Bx,\tilde \Bw_k^\Bx))^\top(\Bw_{k+1}-\Bw_k)\nonumber\\
                                                        &=\Phi(\Bx,\Bw_{k})-2h
                                                          \nabla_\Bw\Phi(\Bx,\tilde \Bw_k^\Bx)^\top\nabla_\Bw\Phi(\BX,\Bw_k)^\top(\Phi(\BX,\Bw_k)-\By),
                                   \end{align*}
                                   for some $\Bx$-dependent
                                   $\tilde\Bw_k^\Bx\in\R^n$ in the
                                   convex hull of $\Bw_k$ and
                                   $\Bw_{k+1}$. Introducing
                                   \begin{equation}\label{eq:BGk}
                                     \begin{aligned}
                                       \BG^k(\Bx,\BX)&\dfn \nabla_\Bw\Phi(\Bx,\tilde\Bw_k^\Bx)^\top\nabla_\Bw\Phi(\BX,\Bw_k)^\top\in\R^{1\times m}\\
                                       \BG^k(\BX,\BX)&\dfn \nabla_\Bw\Phi(\BX,\tilde\BW_k^\BX)\nabla_\Bw\Phi(\BX,\Bw_k)^\top\in\R^{m\times m}
                                     \end{aligned}
                                   \end{equation}
                                   where
                                   $\tilde\BW_k^\BX=(\tilde\Bw_k^{\Bx_1},\dots,\tilde\Bw_k^{\Bx_m})$,
                                   this yields
                                   \begin{subequations}\label{eq:dynamicsfull}
                                     \begin{align}\label{eq:dynamicsfullx}
                                       \Phi(\Bx,\Bw_{k+1})
                                       &=\Phi(\Bx,\Bw_{k})-2h \BG^k(\Bx,\BX)(\Phi(\BX,\Bw_k)-\By),\\ \label{eq:dynamicsfullX}
                                       \Phi(\BX,\Bw_{k+1})
                                       &=\Phi(\BX,\Bw_{k})-2h \BG^k(\BX,\BX)(\Phi(\BX,\Bw_k)-\By).
                                     \end{align}
                                   \end{subequations}

                                 \item {\bf linearized model}:
                                   Initialize $\Bp_0\dfn\Bnul\in\R^n$,
                                   and set for step size $h>0$ and all
                                   $k\in\N_0$
                                   \begin{equation}\label{eq:GDlinmodel}
                                     \Bp_{k+1}=\Bp_k-h\nabla_\Bp\objF^{\rm lin}(\Bp_k).
                                   \end{equation}
                                   Then, since
                                   $\nabla_\Bp\Phi^{\rm
                                     lin}(\Bx,\Bp)=\nabla_\Bw\Phi(\Bx,\Bw_0)$
                                   for any $\Bp\in\R^n$,
                                   \begin{equation*}
                                     \nabla_\Bp\objF^{\rm lin}(\Bp) = \nabla_\Bp\norm[]{\Phi^{\rm lin}(\BX,\Bp)-\By}^2
                                     = 2\nabla_\Bw\Phi(\BX,\Bw_0)^\top(\Phi^{\rm lin}(\BX,\Bp)-\By)
                                   \end{equation*}
                                   and
                                   \begin{align*}
                                     \Phi^{\rm lin}(\Bx,\Bp_{k+1})&=
                                                                    \Phi^{\rm lin}(\Bx,\Bp_{k})+
                                                                    \nabla_\Bw\Phi(\Bx,\Bw_0)^\top(\Bp_{k+1}-\Bp_k)\nonumber\\
                                                                  &=\Phi^{\rm lin}(\Bx,\Bp_{k})-2h
                                                                    \nabla_\Bw\Phi(\Bx,\Bw_0)^\top\nabla_\Bw\Phi(\BX,\Bw_0)^\top(\Phi^{\rm lin}(\BX,\Bp_k)-\By).
                                   \end{align*}
                                   Introducing (cf.~\eqref{eq:JBw0})
                                   \begin{equation}\label{eq:BGlin}
                                     \begin{aligned}
                                       \BG^{\rm lin}(\Bx,\BX)&\dfn\nabla_\Bw\Phi(\Bx,\Bw_0)^\top\nabla_\Bw\Phi(\BX,\Bw_0)^\top\in\R^{1\times m},\\
                                       \BG^{\rm lin}(\BX,\BX)&\dfn\nabla_\Bw\Phi(\BX,\Bw_0)\nabla_\Bw\Phi(\BX,\Bw_0)^\top=(\hat K_n(\Bx_i,\Bx_j))_{i,j=1}^m\in\R^{m\times m}
                                     \end{aligned}
                                   \end{equation}
                                   this yields
                                   \begin{subequations}\label{eq:dynamicslinear}
                                     \begin{align}\label{eq:dynamicslinearx}
                                       \Phi^{\rm lin}(\Bx,\Bp_{k+1})&=\Phi^{\rm lin}(\Bx,\Bp_{k})-2h
                                                                      \BG^{\rm lin}(\Bx,\BX)(\Phi^{\rm lin}(\BX,\Bp_k)-\By)\\ \label{eq:dynamicslinearX}
                                       \Phi^{\rm lin}(\BX,\Bp_{k+1})&=\Phi^{\rm lin}(\BX,\Bp_{k})-2h
                                                                      \BG^{\rm lin}(\BX,\BX)(\Phi^{\rm lin}(\BX,\Bp_k)-\By).
                                     \end{align}
                                   \end{subequations}
                                 \end{itemize}

                                 The full dynamics
                                 \eqref{eq:dynamicsfull} are governed
                                 by the $k$-dependent kernel matrices
                                 $\BG^k$. In contrast, the linear
                                 model's dynamics are entirely
                                 determined by the initial kernel
                                 matrix $\BG^{\rm lin}$. %
                                 The following corollary gives an
                                 upper bound on how much these
                                 matrices may deviate during training,
                                 \cite[Thm.~G.1]{lee2019wide}.

  \begin{corollary}\label{cor:LClin}
    Let $\Bp_0=\Bnul\in\R^n$, and let Assumption \ref{ass:ntk} be
    satisfied for some $r,R,U,L,\theta_{\rm min},\theta_{\rm max}>0$.
    Let $(\Bw_k)_{k\in\N}$, $(\Bp_k)_{k\in\N}$ be generated by
    gradient descent \eqref{eq:GDfullmodel}, \eqref{eq:GDlinmodel}
    with a positive step size
    \begin{equation*}
      h<\frac{1}{\theta_{\rm min}+\theta_{\rm max}}. 
    \end{equation*}
    
    Then for all $\Bx\in\R^d$ with $\norm{\Bx}\le R$
    \begin{subequations}\label{eq:LClin}
      \begin{align}\label{eq:LClinx}
        \sup_{k\in\N}\norm[]{\BG^k(\Bx,\BX)-\BG^{\rm lin}(\Bx,\BX)}&\le
                                                                     2\sqrt{m}ULr,\\ \label{eq:LClinX}
        \sup_{k\in\N}\norm[]{\BG^k(\BX,\BX)-\BG^{\rm lin}(\BX,\BX)}&\le 2 mULr.
      \end{align}
    \end{subequations}
  \end{corollary}
  \begin{proof}
    By Theorem \ref{thm:ntk} it holds $\Bw_k\in B_r(\Bw_0)$ for all
    $k\in\N$, and thus also $\tilde\Bw_k^{\Bx}\in B_r(\Bw_0)$ for
    $\tilde\Bw_k^\Bx$ in the convex hull of $\Bw_k$ and $\Bw_{k+1}$ as
    in \eqref{eq:BGk}. %
    Using Assumption \ref{ass:ntk} \ref{item:ntk2}, the definitions of
    $\BG^k$ and $\BG^{\rm lin}$ give
    \begin{align*}
      \norm[]{\BG^k(\Bx,\BX)-\BG^{\rm lin}(\Bx,\BX)}
      &\le  \norm[]{\nabla_\Bw\Phi(\Bx,\tilde\Bw_k^\Bx)}\norm{\nabla_\Bw\Phi(\BX,\Bw_k)-\nabla_\Bw\Phi(\BX,\Bw_0)}\\
      &\quad+\norm[]{\nabla_\Bw\Phi(\BX,\Bw_0)}\norm{\nabla_\Bw\Phi(\Bx,\tilde\Bw_k^\Bx)-\nabla_\Bw\Phi(\Bx,\Bw_0)}\\
      &\le %
        2\sqrt{m}ULr.
    \end{align*}
    The proof for the second inequality is similar.
  \end{proof}

  \subsection{Limiting model predictions}
  We begin by stating the main result of this section, which is based
  on and follows the arguments in \cite[Thm.~H.1]{lee2019wide}. It
  gives an upper bound on the discrepancy between the full and
  linearized models at each training step, and thus in the limit.

  \begin{theorem}\label{thm:ntkclose}
    Consider the setting of Corollary \ref{cor:LClin}, in particular
    let $r$, $R$, $\theta_{\rm min}$, $\theta_{\rm max}$ be as in
    Assumption \ref{ass:ntk}.  Then for all $\Bx\in\R^d$ with
    $\norm{\Bx}\le R$
    \begin{equation*}
      \sup_{k\in\N}|\Phi(\Bx,\Bw_k)-\Phi^{\rm lin}(\Bx,\Bp_k)|\le
      \frac{4\sqrt{m}ULr}{\theta_{\rm min}}\left(1+\frac{2mU^2}{\theta_{\rm min}}\right)\sqrt{\objF(\Bw_0)}.
    \end{equation*}
  \end{theorem}

  To prove the theorem, we first examine the difference between the
  full and linearized models on the training data.
  
  \begin{proposition}\label{prop:ntkclose}
    Consider the setting of Corollary \ref{cor:LClin} and set
    \begin{equation*}
      \alpha\dfn 4hmULr\sqrt{\objF(\Bw_0)}.
    \end{equation*}

    Then for all $k\in\N$
    \begin{equation*}
      \norm{\Phi(\BX,\Bw_k)-\Phi^{\rm lin}(\BX,\Bp_k)}\le
      \alpha k(1-h\theta_{\rm min})^{k-1}.
    \end{equation*}
  \end{proposition}
  \begin{proof}
    Throughout this proof we write for short
    \begin{equation*}
      \BG^k=\BG^k(\BX,\BX)\qquad\text{and}\qquad
      \BG^{\rm lin}=\BG^{\rm lin}(\BX,\BX),
    \end{equation*}
    and set for $k\in\N$
    \begin{equation*}
      \Be_k\dfn \Phi(\BX,\Bw_k)-\Phi^{\rm lin}(\BX,\Bp_k).
    \end{equation*}
    
    Subtracting \eqref{eq:dynamicslinearX} from
    \eqref{eq:dynamicsfullX} we get for $k\ge 0$
    \begin{align*}
      \Be_{k+1} &= \Be_k-2h\BG^k(\Phi(\BX,\Bw_k)-\By)+2h\BG^{\rm lin}(\Phi^{\rm lin}(\BX,\Bp_k)-\By)\\
                &=(\BI_m-2h\BG^{\rm lin})\Be_k-2h (\BG^k-\BG^{\rm lin})(\Phi(\BX,\Bw_k)-\By)
    \end{align*}
    where $\BI_m\in\R^{m\times m}$ is the identity.  Set
    $\rho\dfn 1-h\theta_{\rm min}$. Then by \eqref{eq:LClinX},
    \eqref{eq:ntk2}, we can bound the second term by
    \begin{equation*}
      \norm{2h (\BG^k-\BG^{\rm lin})(\Phi(\BX,\Bw_k)-\By)}
      \le \underbrace{4 hmULr
        \sqrt{\objF(\Bw_0)}}_{=\alpha} \rho^k.
    \end{equation*}
    Moreover, Assumption \ref{ass:ntk} \ref{item:ntk1} and
    $h< (\theta_{\min}+\theta_{\max})^{-1}$ yield
    \begin{equation*}
      \norm{\BI_m-2h\BG^{\rm lin}}\le 1-2h\theta_{\rm min}
      \le \rho.
    \end{equation*}
    Hence
    \begin{equation*}
      \norm{\Be_{k+1}}\le \rho\norm{\Be_k}+\alpha \rho^k\le\dots\le
      \rho^{k+1}\norm{\Be_0}+\sum_{j=0}^{k}\rho^{k-j}\alpha \rho^j 
      = \rho^{k+1} \norm{\Be_0} + %
      \alpha(k+1) \rho^k.
    \end{equation*}
    Since $\Bp_0=\Bnul$ it holds
    $\Phi^{\rm lin}(\BX,\Bp_0)=\Phi(\BX,\Bw_0)$
    (cf.~\eqref{eq:Philin}).  Thus $\norm{\Be_0}=0$ which gives the
    statement.
  \end{proof}

  We are now in position to prove the theorem.

  \begin{proof}[of Theorem \ref{thm:ntkclose}]
    Throughout this proof we write for short
    \begin{equation*}
      \BG^k=\BG^k(\Bx,\BX)\in\R^{1\times m}\qquad\text{and}\qquad
      \BG^{\rm lin}=\BG^{\rm lin}(\Bx,\BX)\in\R^{1\times m},
    \end{equation*}
    and set for $k\in\N$
    \begin{equation*}
      e_k\dfn \Phi(\Bx,\Bw_k)-\Phi^{\rm lin}(\Bx,\Bp_k).
    \end{equation*}    

    Subtracting \eqref{eq:dynamicslinearx} from
    \eqref{eq:dynamicsfullx}
    \begin{align*}
      e_{k+1} &= e_k-2h\BG^k(\Phi(\BX,\Bw_k)-\By)+2h\BG^{\rm lin}(\Phi^{\rm lin}(\BX,\Bp_k)-\By)\\
              &= e_k-2h(\BG^k-\BG^{\rm lin})(\Phi(\BX,\Bw_k)-\By)+2h \BG^{\rm lin}(\Phi^{\rm lin}(\BX,\Bp_k)-\Phi(\BX,\Bw_k)).
    \end{align*}
    Denote $\rho\dfn 1-h\theta_{\rm min}$. By \eqref{eq:LClinx} and
    \eqref{eq:ntk2}
    \begin{align*}
      2h\norm{\BG^k-\BG^{\rm lin}}&\le 4h\sqrt{m}ULr\\
      \norm{\Phi(\BX,\Bw_k)-\By}&\le \rho^k\sqrt{\objF(\Bw_0)}
    \end{align*}
    and by \eqref{eq:JboundU} (cf.~\eqref{eq:BGlin}) and Proposition
    \ref{prop:ntkclose}
    \begin{align*}
      2h\norm{\BG^{\rm lin}}&\le 2h\sqrt{m}U^2\\
      \norm{\Phi(\BX,\Bw_k)-\Phi^{\rm lin}(\BX,\Bp_k)}&\le \alpha k \rho^{k-1}.
    \end{align*}

    Hence for $k\ge 0$
    \begin{align*}
      |e_{k+1}|&\le |e_k| + \underbrace{4h\sqrt{m}ULr\sqrt{\objF(\Bw_0)}}_{\dfnn \beta_1}\rho^k + \underbrace{2h\sqrt{m}U^2\alpha}_{\dfnn\beta_2} k \rho^{k-1}.
    \end{align*}
    Repeatedly applying this bound and using
    $\sum_{j\ge 0}\rho^j=(1-\rho)^{-1}=(h\theta_{\rm min})^{-1}$ and
    $\sum_{j\ge 0}j\rho^{j-1}=(1-\rho)^{-2}=(h\theta_{\rm min})^{-2}$
    \begin{equation*}
      |e_{k+1}|\le |e_0|+\beta_1\sum_{j=0}^k \rho^j+\beta_2 \sum_{j=0}^k j \rho^{j-1}
      \le \frac{\beta_1}{h\theta_{\rm min}}+\frac{\beta_2}{(h\theta_{\rm min})^2}.
    \end{equation*}
    Here we used that due to $\Bp_0=\Bnul$ it holds
    $\Phi(\Bx,\Bw_0)=\Phi^{\rm lin}(\Bx,\Bp_0)$ so that $e_0=0$.
  \end{proof}

\section{Training dynamics for shallow neural networks}\label{sec:ntkdynamics}
In this section, following \cite{lee2019wide}, we discuss the
implications of Theorems \ref{thm:ntk} and \ref{thm:ntkclose} for wide
neural networks. As in \cite{telgarskynotes}, for ease of presentation
we focus on a shallow architecture with only one hidden layer, but
stress that similar considerations also hold for deep networks, see
the bibliography section.

\subsection{Architecture}\label{sec:ntksetting}
Let $\Phi:\R^{d}\to\R$ be a neural network of depth one and width
$n\in\N$ of type
\begin{equation}\label{eq:ntkmodel}
  \Phi(\Bx,\Bw) = \Bv^\top\sigma(\BU\Bx+\Bb)+ c.
\end{equation}
Here $\Bx\in\R^{d}$ is the input, and $\BU\in\R^{n\times d}$,
$\Bv\in\R^{n}$, $\Bb\in\R^{n}$ and $c\in\R$ are the parameters which
we collect in the vector $\Bw=(\BU,\Bb,\Bv,c)\in\R^{n(d+2)+1}$ (with
$\BU$ suitably reshaped). For future reference we note that
\begin{equation}\label{eq:nablaPhi}
  \begin{aligned}
    \nabla_{\BU}\Phi(\Bx,\Bw) &=(\Bv\odot \sigma'(\BU\Bx+\Bb))
                                \Bx^\top
                                \in\R^{n\times d}\\
    \nabla_{\Bb}\Phi(\Bx,\Bw) &= %
                                \Bv\odot \sigma'(\BU\Bx+\Bb)
                                \in\R^n\\
    \nabla_{\Bv}\Phi(\Bx,\Bw) &= \sigma(\BU\Bx+\Bb)\in\R^n\\
    \nabla_{c}\Phi(\Bx,\Bw) &= 1\in\R,
  \end{aligned}
\end{equation}
where $\odot$ denotes the Hadamard product.  We also write
$\nabla_\Bw\Phi(\Bx,\Bw)\in\R^{n(d+2)+1}$ to denote the full gradient
with respect to all parameters.

In practice, it is common to initialize the weights randomly, and in
this section we consider so-called LeCun initialization
\cite{lecun98}. The following condition on the activation function
$\sigma$, and on the distribution $\CW$ on $\R$ used for this
initialization, will be assumed throughout the rest of this
chapter. In particular it implies that $\sigma\in C^1$ grows at most
linearly, and $\CW$ has finite fourth moments; we do not aim for most
generality here.
\begin{assumption}\label{ass:gamma}
  There exist $1\le R<\infty$ such that
  \begin{enumerate}[label=(\alph*)]
  \item\label{item:gamma1} $\sigma:\R\to\R$ satisfies $|\sigma(0)|$,
    $|\sigma'(0)|$, ${\rm Lip}(\sigma)$, ${\rm Lip}(\sigma')\le R$,
  \item\label{item:gamma2} $\CW$ has expectation zero, variance one,
    and finite moments up to order four.
  \end{enumerate}
\end{assumption}
\begin{remark}
  In the rest of this chapter, constants typically depend on
  $\CW(0,1)$, which we will not state anymore.
\end{remark}

To explicitly indicate the expectation and variance in the notation,
we also write $\CW(0,1)$ instead of $\CW$, and for $\mu\in\R$ and
$\varsigma>0$ we use $\CW(\mu,\varsigma^2)$ to denote the
corresponding scaled and shifted measure with expectation $\mu$ and
variance $\varsigma^2$; thus, if $X\sim\CW(0,1)$ then
$\mu+\varsigma X\sim\CW(\mu,\varsigma^2)$. LeCun initialization sets
the variance of the weights in each layer to be reciprocal to the
input dimension of the layer: the idea is to normalize the output
variance of all network nodes. The initial parameters
\begin{equation*}
  \Bw_0=(\BU_0,\Bb_0,\Bv_0,c_0)
\end{equation*}
are thus randomly initialized with components
\begin{equation}\label{eq:ntkinit}
  U_{0;ij}\overset{\rm iid}{\sim} \CW\Big(0,\frac{1}{d}\Big),\qquad
  v_{0;i}\overset{\rm iid}{\sim} \CW\Big(0,\frac{1}{n}\Big),\qquad
  b_{0;i},~c_0=0,
\end{equation}
independently for all $i=1,\dots,n$, $j=1,\dots,d$.  For a fixed
$\varsigma>0$ one might choose variances $\varsigma^2/d$ and
$\varsigma^2/n$ in \eqref{eq:ntkinit}, which would require only minor
modifications in the rest of this section. Biases are set to zero for
simplicity, with nonzero initialization discussed in the
exercises. All expectations and probabilities in Section
\ref{sec:ntkdynamics} are understood with respect to this random
initialization.

\begin{example}
  Typical examples for $\CW(0,1)$ are the standard normal distribution
  on $\R$ or the uniform distribution on $[-\sqrt{3},\sqrt{3}]$.
\end{example}

\subsection{Neural tangent kernel}\label{sec:ntk_LeCun}
We begin our analysis by investigating the empirical tangent kernel
\begin{equation*}
  \hat K_n(\Bx,\Bz) = \dup{\nabla_\Bw\Phi(\Bx,\Bw_0)}{\nabla_\Bw \Phi(\Bz,\Bw_0)}
\end{equation*}
of the shallow network \eqref{eq:ntkmodel} with initialization
\ref{eq:ntkinit}. Scaled properly, it converges in the infinite width
limit $n\to\infty$ towards a specific kernel known as the {\bf neural
  tangent kernel} (NTK) \cite{jacot2018neural}. Importantly, this
kernel depends on both the architecture and the initialization
scheme. Since we focus only on the specific setting introduced in
Section \ref{sec:ntksetting} in the following, we simply denote it by
$K^{\rm NTK}$.
  
  \begin{theorem}\label{thm:LeCun}
    Let $\sigma$, $\CW$ satisfy Assumption \ref{ass:gamma} for some
    $R\ge 0$.  For any $\Bx$, $\Bz\in\R^d$ and
    $u_i\overset{\rm iid}{\sim}\CW(0,1/d)$, $i=1,\dots,d$, it then
    holds
    \begin{equation}\label{eq:LeCunNTK}
      \lim_{n\to\infty} \frac{1}{n}\hat K_n(\Bx,\Bz) =\bbE [\sigma(\Bu^\top\Bx)\sigma(\Bu^\top\Bz)]\dfnn K^{\rm NTK}(\Bx,\Bz)
    \end{equation}
    almost surely.

    Moreover, for every $\delta$, $\eps>0$ there exists
    $n_0(\delta,\eps,R)\in\N$ such that for all $n\ge n_0$ and
    all $\Bx$, $\Bz\in\R^d$ with $\norm[]{\Bx}$, $\norm[]{\Bz}\le R$
    \begin{equation*}
      \bbP\Bigg[\normc[]{\frac{1}{n}\hat K_n(\Bx,\Bz)-K^{\rm NTK}(\Bx,\Bz)}<\eps\Bigg]\ge 1-\delta.
    \end{equation*}
  \end{theorem}
  \begin{proof}
    Denote the preactivations by
    $\bar \Bx^{(1)}=\BU_0\Bx+\Bb_0\in\R^n$ and
    $\bar \Bz^{(1)}=\BU_0\Bz+\Bb_0\in\R^n$.  Due to the initialization
    \eqref{eq:ntkinit} and our assumptions on $\CW(0,1)$, the
    components
    \begin{equation*}
      \bar x_i^{(1)}=\sum_{j=1}^d U_{0;ij}x_j\sim\Bu^\top\Bx\qquad i=1,\dots,n
    \end{equation*}
    are i.i.d.\ with finite $p$th moment (independent of $n$) for all
    $1\le p\le 4$. The same holds for the
    $(\sigma(\bar x_i^{(1)}))_{i=1}^n$ and the
    $(\sigma'(\bar x_i^{(1)}))_{i=1}^n$, since Assumption
    \ref{ass:gamma} \ref{item:gamma1} implies the linear growth bounds
    \begin{equation*}
      |\sigma(x)|\le R\cdot (1+|x|)\quad\text{and}\quad
      |\sigma'(x)|\le R\cdot (1+|x|)\qquad\text{for all }x\in\R.
    \end{equation*}
    Similarly, the $(\sigma(\bar z^{(1)}_i))_{i=1}^n$ and
    $(\sigma'(\bar z^{(1)}_i))_{i=1}^n$ are collections of i.i.d.\
    random variables with finite $p$th moment for all $1\le p\le 4$.

    Denote $\tilde v_i=\sqrt{n}v_{0;i}$ such that
    $\tilde v_i\overset{\rm iid}{\sim}\CW(0,1)$.  By
    \eqref{eq:nablaPhi}
    \begin{align}\label{eq:hatKnlimit}
      \frac{1}{n}\hat K_n(\Bx,\Bz)%
      =(1+\Bx^\top\Bz) \frac{1}{n^2}\sum_{i=1}^n \tilde v_{i}^2\sigma'(\bar x_i^{(1)})\sigma'(\bar z_i^{(1)})
      +\frac{1}{n}\sum_{i=1}^n \sigma(\bar x_i^{(1)})\sigma(\bar z_i^{(1)}) +\frac{1}{n}.
    \end{align}
    Since
    \begin{equation}
      \frac{1}{n}\sum_{i=1}^n \tilde v_{i}^2\sigma'(\bar x_i^{(1)})\sigma'(\bar z_i^{(1)})
    \end{equation}
    is an average over i.i.d.\ random variables with finite variance
    (by the moment assumption, and since $\tilde v_i$ is independent
    of $\sigma'(\bar x_i^{(1)})\sigma'(\bar z_i^{(1)})$), the law of
    large numbers implies almost sure convergence of this expression
    towards
    \begin{align*}
      \bbE\big[\tilde v_{i}^2\sigma'(\bar x_i^{(1)})\sigma'(\bar z_i^{(1)})\big]
      &=\bbE[\tilde v_{i}^2]
        \bbE[\sigma'(\bar x_i^{(1)})\sigma'(\bar z_i^{(1)})]\\
      &= \bbE[\sigma'(\Bu^\top\Bx)\sigma'(\Bu^\top\Bz)],
    \end{align*}
    where we used that $\tilde v_{i}^2$ is independent of
    $\sigma'(\bar x_i^{(1)})\sigma'(\bar z_i^{(1)})$.  Thus the first
    term on the right-hand side of \eqref{eq:hatKnlimit} tends to $0$
    as $n\to\infty$.

    By the same argument
    \begin{equation*}
      \frac{1}{n}\sum_{i=1}^n \sigma(\bar x_i^{(1)})\sigma(\bar z_i^{(1)})\to \bbE[\sigma(\Bu^\top\Bx)\sigma(\Bu^\top\Bz)]
    \end{equation*}
    almost surely as $n\to\infty$. This shows the first statement.

    The existence of $n_0$ follows similarly by an application of
    Theorem \ref{thm:chebyscheff}.
  \end{proof}

\begin{remark}
  With the present LeCun initialization, the contribution of the term
  involving $\sigma'$ vanishes in the proof of Theorem
  \ref{thm:LeCun}. For an initialization yielding a kernel that
  includes the derivative term, see \cite{jacot2018neural}; in the
  literature this is the kernel often referred to as the NTK kernel,
  while \eqref{eq:LeCunNTK} is also known as the NNGP kernel.
\end{remark}

\begin{example}[$K^{\rm NTK}$ for ReLU]\label{ex:KLCRelu}
  Let $\sigma(x)=\max\{0,x\}$ and let $\CW(0,1)$ be the standard
  normal distribution. For $\Bx$, $\Bz\in\R^d$ denote by
  \begin{equation*}
    \vartheta=
    \arccos\left(\frac{\Bx^\top\Bz}{\norm[]{\Bx}\norm[]{\Bz}}\right)
  \end{equation*}
  the angle between these vectors. Then according to \cite[Appendix
  A]{chosaul}, it holds with $u_i\overset{\rm iid}{\sim}\CW(0,1/d)$,
  $i=1,\dots,d$,
  \begin{equation*}
    K^{\rm NTK}(\Bx,\Bz) = \bbE[\sigma(\Bu^\top\Bx)\sigma(\Bu^\top\Bz)]=\frac{\norm[]{\Bx}\norm[]{\Bz}}{2\pi d}(\sin(\vartheta)+(\pi-\vartheta)\cos(\vartheta)).
  \end{equation*}
\end{example}

  \subsection{Training dynamics and model predictions}\label{sec:ntkgrad}
  We now proceed as in \cite[Appendix G]{lee2019wide}, to show that
  the analysis in Sections \ref{sec:globmin}-\ref{sec:proximity} is
  applicable to the wide neural network \eqref{eq:ntkmodel} with high
  probability under random initialization \eqref{eq:ntkinit}.  We work
  under the following assumptions on the %
  training data \cite[Assumptions 1-4]{lee2019wide}. To avoid
  introducing further constants, we use again $R$ (as in Assumption
  \ref{ass:gamma}), and will assume in the following that $R$ is
  sufficiently large so that both Assumption \ref{ass:gamma} and
  \ref{ass:ntknn} are satisfied.

  \begin{assumption}\label{ass:ntknn}
    There exist $1\le R<\infty$ and
    $0<\theta_{\rm min}^{\rm NTK}\le \theta_{\rm max}^{\rm
      NTK}<\infty$ such that
    \begin{enumerate}[label=(\alph*)]
    \item\label{item:ntknn2} $\norm[]{\Bx_i}$, $|y_i|\le R$ for all
      training data $(\Bx_i,y_i)\in\R^d\times\R$, $i=1,\dots,m$,
    \item\label{item:ntknn3} the kernel matrix of the neural tangent
      kernel
      \begin{equation*}
        (K^{\rm NTK}(\Bx_i,\Bx_j))_{i,j=1}^m\in\R^{m\times m}
      \end{equation*}
      is regular and its eigenvalues belong to
      $[\theta_{\rm min}^{\rm NTK},\theta_{\rm max}^{\rm NTK}]$.
    \end{enumerate}
  \end{assumption}

  We start by showing Assumption \ref{ass:ntk} \ref{item:ntk1} for the
  present setting. More precisely, we give bounds for the eigenvalues
  of the empirical tangent kernel.
  
  \begin{lemma}\label{lemma:LCeigs}
    Let Assumptions \ref{ass:gamma}, \ref{ass:ntknn} be
    satisfied. Then for every $\delta>0$ there exists
    $n_0(\delta,\theta_{\rm min}^{\rm NTK},m,R)\in\R$ such that for
    all $n\ge n_0$ it holds with probability at least $1-\delta$ that
    all eigenvalues of
    \begin{equation*}
      (\hat K_n(\Bx_i,\Bx_j))_{i,j=1}^m= \big(\dup{\nabla_\Bw\Phi(\Bx_i,\Bw_0)}{\nabla_\Bw\Phi(\Bx_j,\Bw_0)}\big)_{i,j=1}^m\in\R^{m\times m}
    \end{equation*}
    belong to
    $[n\theta_{\rm min}^{\rm NTK}/2,2n\theta_{\rm max}^{\rm NTK}]$.
  \end{lemma}
  \begin{proof}
    Denote $\hat\BG_n\dfn (\hat K_n(\Bx_i,\Bx_j))_{i,j=1}^m$ and
    $\BG^{\rm NTK}\dfn (K^{\rm NTK}(\Bx_i,\Bx_j))_{i,j=1}^m$.  By
    Theorem \ref{thm:LeCun}, there exists $n_0$ such that for all
    $n\ge n_0$ holds with probability at least $1-\delta$ that
    \begin{equation*}
      \normc{\BG^{\rm NTK}-\frac{1}{n}\hat\BG_n}\le\frac{\theta_{\rm min}^{\rm NTK}}{2}.
    \end{equation*}

    Assuming this bound to hold, the smallest singular value of the
    symmetric matrix $\hat\BG_n/n$ is lower bounded by
    \begin{equation*}
      \inf_{\substack{\Ba\in\R^m\\\norm{\Ba}=1}}
      \frac{1}{n}\norm{\hat\BG_n\Ba}
      \ge \inf_{\substack{\Ba\in\R^m\\\norm{\Ba}=1}}
      \norm{\BG^{\rm NTK}\Ba}-\frac{\theta_{\rm min}^{\rm NTK}}{2}
      \ge \theta_{\rm min}^{\rm NTK}-\frac{\theta_{\rm min}^{\rm NTK}}{2}\ge \frac{\theta_{\rm min}^{\rm NTK}}{2},
    \end{equation*}
    where we have used that $\theta_{\rm min}^{\rm NTK}$ is the
    smallest eigenvalue, and thus singular value, of the symmetric
    positive definite matrix $\BG^{\rm NTK}$. Since for a positive definite
    matrix the singular- and eigenvalues coincide, this shows
    that (with probability at least $1-\delta$) the smallest
    eigenvalue of $\hat\BG_n$ is larger or equal to
    $n\theta_{\rm min}^{\rm NTK}/2$. Similarly, we conclude that the
    largest eigenvalue is bounded from above by
    $n(\theta_{\rm max}^{\rm NTK}+\theta_{\rm min}^{\rm NTK}/2)\le
    2n\theta_{\rm max}^{\rm NTK}$. This concludes the proof.
  \end{proof}

  Next we check Assumption \ref{ass:ntk} \ref{item:ntk2}. To this end
  we first provide a simple bound on the norm of a random matrix,
  which will be sufficient for the subsequent results.

  \begin{lemma}\label{lemma:whpW}
    Let $\CW$ satisfy Assumption \ref{ass:gamma} \ref{item:gamma2},
    and let $\BW\in\R^{n\times d}$ with
    $W_{ij}\overset{\rm iid}{\sim}\CW(0,1)$.  Denote the fourth moment
    of $\CW(0,1)$ by $\mu_4$. Then
    \begin{equation*}
      \bbP\Big[\norm[]{\BW}\le \sqrt{n(d+1)}\Big]\ge 1-\frac{d\mu_4}{n}.
    \end{equation*} 
  \end{lemma}
  \begin{proof}
    It holds
    \begin{equation*}
      \norm[]{\BW}%
      \le \norm[F]{\BW} =
      \Big(\sum_{i=1}^n\sum_{j=1}^dW_{ij}^2\Big)^{1/2}.
    \end{equation*}
    The $\alpha_i\dfn \sum_{j=1}^dW_{ij}^2$, $i=1,\dots,n$, are
    i.i.d.\ distributed with expectation $d$ and finite variance
    $d C$, where $C\le\mu_4$ is the variance of $W_{11}^2$. By Theorem
    \ref{thm:chebyscheff}
    \begin{equation*}
      \bbP\Big[\norm[]{\BW}>\sqrt{n(d+1)}\Big]
      \le
      \bbP\Big[\frac{1}{n}\sum_{i=1}^n\alpha_i>d+1\Big]\le
      \bbP\Big[\Big|\frac{1}{n}\sum_{i=1}^n\alpha_i-d\Big|>1\Big]\le \frac{d\mu_4}{n},
    \end{equation*}
    which concludes the proof.
  \end{proof}

  \begin{lemma}\label{lemma:LCLU}
    Let Assumption %
    \ref{ass:gamma} \ref{item:gamma1} be satisfied with some constant
    $R\ge 1$. Then there exists $M(R)$, and for all $\gamma$,
    $\delta>0$ there exists $n_0(\gamma,\delta,d,R)\in\N$ such that for
    each $n\ge n_0$ and each $\Bx\in\R^d$ with $\norm[]{\Bx}\le R$,
    it holds with probability at least $1-\delta$ that
    \begin{equation*}
      \begin{aligned}
        \norm[]{\nabla_\Bw\Phi(\Bx,\Bw)}&\le M\sqrt{n} &&\text{for all }\Bw\in B_{\gamma n^{-1/2}}(\Bw_0)\\
        \norm[]{\nabla_\Bw\Phi(\Bx,\Bw)-\nabla_\Bw\Phi(\Bx,\Bv)}&\le M\sqrt{n}\norm[]{\Bw-\Bv} &&\text{for all }\Bw,~\Bv\in B_{\gamma n^{-1/2}}(\Bw_0).
      \end{aligned}
    \end{equation*}
  \end{lemma}
  \begin{proof}
    Due to the initialization \eqref{eq:ntkinit}, by Lemma
    \ref{lemma:whpW} we can find $\tilde n_0(\delta,d)$ such that for
    all $n\ge \tilde n_0$ holds with probability at least $1-\delta$
    that
    \begin{equation}\label{eq:LCLUass}
      \norm[]{\Bv_0}\le 2\qquad\text{and}\qquad
      \norm[]{\BU_0}\le 2\sqrt{n}.
    \end{equation}
    
    For the rest of this proof we let $\Bx\in\R^d$ arbitrary with
    $\norm[]{\Bx}\le R$, we set
    \begin{equation*}
      n_0\dfn \max\{\gamma^2,\tilde n_0(\delta,d)\},
    \end{equation*}
    and we fix $n\ge n_0$ so that $n^{-1/2}\gamma\le 1$. To prove the
    lemma we need to show that the claimed inequalities hold as long
    as \eqref{eq:LCLUass} is satisfied.  We will several times use
    that for all $\Bp$, $\Bq\in\R^n$
    \begin{equation*}
      \norm[]{\Bp\odot\Bq}\le\norm[]{\Bp}\norm[]{\Bq}\qquad\text{and}\qquad
      \norm[]{\sigma(\Bp)}\le R\sqrt{n}+R\norm[]{\Bp}
    \end{equation*}
    since $|\sigma(x)|\le R\cdot (1+|x|)$ by Assumption
    \ref{ass:gamma} \ref{item:gamma1}. The same holds for $\sigma'$.

    {\bf Step 1.} We show the bound on the gradient. Fix
    \begin{equation*}
      \Bw=(\BU,\Bb,\Bv,c)\qquad\text{s.t.}\qquad
      \norm[]{\Bw-\Bw_0}\le \gamma n^{-1/2}.
    \end{equation*}
    Using formula \eqref{eq:nablaPhi} for $\nabla_\Bb\Phi$, the fact
    that $\Bb_0=\Bnul$ by \eqref{eq:ntkinit}, and the above
    inequalities %
    \begin{align}\label{eq:ntklemma2s1}
      \norm[]{\nabla_\Bb\Phi(\Bx,\Bw)}
      &\le \norm[]{\nabla_\Bb\Phi(\Bx,\Bw_0)}+
        \norm[]{\nabla_\Bb\Phi(\Bx,\Bw)-\nabla_\Bb\Phi(\Bx,\Bw_0)}\nonumber\\
      &=\norm[]{\Bv_0\odot\sigma'(\BU_0\Bx)}+
        \norm[]{\Bv\odot\sigma'(\BU\Bx+\Bb)-\Bv_0\odot\sigma'(\BU_0\Bx)}\nonumber\\
      &\le 2(R\sqrt{n}+2R^2\sqrt{n}) + \norm[]{\Bv\odot\sigma'(\BU\Bx+\Bb)-\Bv_0\odot\sigma'(\BU_0\Bx)}.
    \end{align}
    Due to
    \begin{equation}\label{eq:LCUnorm}
      \norm[]{\BU}\le\norm[]{\BU_0}+\norm[F]{\BU_0-\BU}\le 2\sqrt{n}+\gamma n^{-1/2}\le 3\sqrt{n},
    \end{equation}
    and using the fact that $\sigma'$ has Lipschitz constant $R$, the
    last norm in \eqref{eq:ntklemma2s1} is bounded by
    \begin{align*}
      &\norm[]{(\Bv-\Bv_0)\odot\sigma'(\BU\Bx+\Bb)}+\norm[]{\Bv_0\odot(\sigma'(\BU\Bx+\Bb)-\sigma'(\BU_0\Bx))}\\
      &\qquad\le \gamma n^{-1/2}(R\sqrt{n}+R\cdot (\norm[]{\BU}\norm{\Bx}+\norm{\Bb})) + 2 R\cdot (\norm[]{\BU-\BU_0}\norm{\Bx}+\norm{\Bb})\\
      &\qquad\le R\sqrt{n}+3\sqrt{n} R^2+\gamma n^{-1/2}R+2R\cdot (\gamma n^{-1/2}R+\gamma n^{-1/2})\nonumber\\
      &\qquad\le %
        \sqrt{n}(4R+5R^2)
    \end{align*}
    and therefore
    \begin{equation*}
      \norm[]{\nabla_\Bb\Phi(\Bx,\Bw)}\le \sqrt{n}(6R+9R^2).
    \end{equation*}

    For the gradient with respect to $\BU$ we use
    $\nabla_\BU\Phi(\Bx,\Bw)=\nabla_\Bb\Phi(\Bx,\Bw)\Bx^\top$, so that
    \begin{equation*}
      \norm[F]{\nabla_\BU\Phi(\Bx,\Bw)}
      =\norm[F]{\nabla_\Bb\Phi(\Bx,\Bw)\Bx^\top}
      =\norm[]{\nabla_\Bb\Phi(\Bx,\Bw)}\norm[]{\Bx}
      \le \sqrt{n} (6R^2+9R^3).
    \end{equation*}
    Next
    \begin{align*}
      \norm[]{\nabla_\Bv\Phi(\Bx,\Bw)}
      &= \norm[]{\sigma(\BU\Bx+\Bb)}\\
      &\le R\sqrt{n} + R\norm[]{\BU\Bx+\Bb}\\
      &\le R\sqrt{n} + R\cdot (3\sqrt{n}R+ \gamma n^{-1/2})\\
      &\le \sqrt{n}(2R+3R^2),
    \end{align*}
    and finally $\nabla_c\Phi(\Bx,\Bw)=1$. In all, with
    $M_1(R)\dfn (1+8R+12R^2)$
    \begin{equation*}
      \norm[]{\nabla_\Bw\Phi(\Bx,\tilde\Bw)}\le \sqrt{n} M_1(R).
    \end{equation*}

    {\bf Step 2.} We show Lipschitz continuity. Fix
    \begin{equation*}
      \Bw=(\BU,\Bb,\Bv,c)\qquad\text{and}\qquad\tilde\Bw=(\tilde\BU,\tilde\Bb,\tilde\Bv,\tilde c)
    \end{equation*}
    such that $\norm[]{\Bw-\Bw_0}$,
    $\norm[]{\tilde\Bw-\Bw_0}\le \gamma n^{-1/2}$. Then
    \begin{equation*}
      \norm[]{\nabla_\Bb\Phi(\Bx,\Bw)-\nabla_\Bb\Phi(\Bx,\tilde\Bw)}
      =\norm[]{\Bv\odot\sigma'(\BU\Bx+\Bb)-\tilde\Bv\odot\sigma'(\tilde\BU\Bx+\tilde \Bb)}.
    \end{equation*}
    Using $\norm[]{\tilde\Bv}\le \norm[]{\Bv_0}+\gamma n^{-1/2}\le 3$
    and \eqref{eq:LCUnorm}, this term is bounded by
    \begin{align*}
      &\norm[]{(\Bv-\tilde\Bv)\odot\sigma'(\BU\Bx+\Bb)}+\norm[]{\tilde\Bv\odot(\sigma'(\BU\Bx+\Bb)-\sigma'(\tilde\BU\Bx+\tilde\Bb))}\\
      &\qquad\le \norm[]{\Bv-\tilde\Bv}(R\sqrt{n}+R\cdot (\norm[]{\BU}\norm[]{\Bx}+\norm[]{\Bb}))
        +3R\cdot (\norm[]{\Bx}\norm[]{\BU-\tilde\BU}+\norm[]{\Bb-\tilde\Bb})\\
      &\qquad\le \norm[]{\Bw-\tilde\Bw}\sqrt{n}(5R+6R^2).
    \end{align*}

    For $\nabla_\BU\Phi(\Bx,\Bw)$ we obtain similar as in Step 1
    \begin{align*}
      \norm[F]{\nabla_\BU\Phi(\Bx,\Bw)-\nabla_\BU\Phi(\Bx,\tilde\Bw)}&=
                                                                       \norm{\Bx}\norm[]{\nabla_\Bb\Phi(\Bx,\Bw)-\nabla_\Bb\Phi(\Bx,\tilde\Bw)}\\
                                                                     &\le \norm[]{\Bw-\tilde\Bw}\sqrt{n}(5R^2+6R^3).
    \end{align*}
    Next
    \begin{align*}
      \norm[]{\nabla_\Bv\Phi(\Bx,\Bw)-\nabla_\Bv\Phi(\Bx,\tilde\Bw)}
      &=
        \norm[]{\sigma(\BU\Bx+\Bb)-\sigma(\tilde\BU\Bx+\tilde\Bb)}\\
      &\le R\cdot (\norm[]{\BU-\tilde\BU}\norm[]{\Bx}+\norm[]{\Bb-\tilde\Bb})\\
      &\le \norm[]{\Bw-\tilde\Bw}(R^2+R)
    \end{align*}
    and finally $\nabla_c\Phi(\Bx,\Bw)=1$ is constant.  With
    $M_2(R)\dfn R+6R^2+6R^3$ this shows
    \begin{equation*}
      \norm[]{\nabla_\Bw\Phi(\Bx,\Bw)-\nabla_\Bw\Phi(\Bx,\tilde\Bw)}\le \sqrt{n} M_2(R)\norm[]{\Bw-\tilde\Bw}.
    \end{equation*}
    In all, this concludes the proof with
    $M(R)\dfn\max\{M_1(R),M_2(R)\}$.
  \end{proof}

  Next, we show that the initial error $\objF(\Bw_0)$ remains bounded
  with high probability.

  \begin{lemma}\label{lemma:LCR0}
    Let Assumptions %
    \ref{ass:gamma} and \ref{ass:ntknn} be satisfied.  Then for every
    $\delta>0$ exists $R_0(\delta,m,R)>0$ such that for all $n\in\N$
    \begin{equation*}
      \bbP[\objF(\Bw_0)\le R_0]\ge 1-\delta.
    \end{equation*}
  \end{lemma}
  \begin{proof}
    Let $i\in\{1,\dots,m\}$, and set $\Balpha\dfn \BU_0\Bx_i$ and
    $\tilde v_j\dfn \sqrt{n} v_{0;j}$ for $j=1,\dots,n$, so that
    $\tilde v_j\overset{\rm iid}{\sim}\CW(0,1)$. Then
    \begin{equation*}
      \Phi(\Bx_i,\Bw_0) = \frac{1}{\sqrt{n}}\sum_{j=1}^n\tilde v_j\sigma(\alpha_j).
    \end{equation*}
    By Assumption \ref{ass:gamma} and \eqref{eq:ntkinit}, the
    $\tilde v_{j}\sigma(\alpha_j)$, $j=1,\dots,n$, are i.i.d.\
    centered random variables with finite variance bounded by a
    constant $C(R)$ independent of $n$. Thus the variance of
    $\Phi(\Bx_i,\Bw_0)$ is also bounded by $C(R)$. By Chebyshev's
    inequality, see Lemma \ref{lemma:chebyshev}, for every $k>0$
    \begin{equation*}
      \bbP[|\Phi(\Bx_i,\Bw_0)|\ge k\sqrt{C}]\le \frac{1}{k^2}.
    \end{equation*}
    Setting $k=\sqrt{m/\delta}$
    \begin{align*}
      \bbP\Big[\sum_{i=1}^m|\Phi(\Bx_i,\Bw_0)-y_i|^2 \ge m (k \sqrt{C}+R)^2\Big]
      &\le \sum_{i=1}^m\bbP\Big[|\Phi(\Bx_i,\Bw_0)-y_i| \ge k \sqrt{C}+R\Big]\\
      &\le \sum_{i=1}^m\bbP\Big[|\Phi(\Bx_i,\Bw_0)| \ge k \sqrt{C}\Big]\le \delta,
    \end{align*}
    which shows the claim with $R_0=m\cdot (\sqrt{Cm/\delta}+R)^2$.
  \end{proof}

  The next theorem, which corresponds to \cite[Thms.~G.1 and
  H.1]{lee2019wide}, is one of the main results of this %
  chapter.  It summarizes our findings in the present setting for a
  shallow network of width $n$: with high probability, gradient
  descent converges to a global minimizer and the limiting network
  interpolates the data. During training the network weights remain
  close to initialization. The trained network gives predictions that
  are $O(n^{-1/2})$ close to the predictions of the trained linearized
  model. In the statement of the theorem we denote again by
  $\Phi^{\rm lin}$ the linearization of $\Phi$ defined in
  \eqref{eq:Philin}, and by $\objF$, $\objF^{\rm lin}$ the
  corresponding ridgeless square loss objectives defined in
  \eqref{eq:ridgeless}, \eqref{eq:risklin}, respectively.

  \begin{theorem}\label{thm:ntkLC} 
    Let Assumptions \ref{ass:gamma} and \ref{ass:ntknn} be satisfied,
    and let the parameters $\Bw_0$ of the width-$n$ neural network
    $\Phi$ in \eqref{eq:ntkmodel} be initialized according to
    \eqref{eq:ntkinit}. %
    Fix a positive learning rate
    \begin{equation*}
      h\le \frac{2}{n(\theta_{\rm min}^{\rm NTK}+4\theta_{\rm max}^{\rm NTK})},
    \end{equation*}
    set $\Bp_0\dfn\Bnul\in\R^n$ and let for all $k\in\N_0$
    \begin{equation*}
      \Bw_{k+1}=\Bw_k-h\nabla\objF(\Bw_k)\qquad\text{and}\qquad
      \Bp_{k+1}=\Bp_k-h\nabla\objF^{\rm lin}(\Bp_k).
    \end{equation*}
    
    Then for every $\delta>0$ there exist $C>0$, $n_0\in\N$ such that
    for all $n\ge n_0$ it holds with probability at least $1-\delta$
    that for all $k\in\N$ and all $\Bx\in\R^d$ with $\norm{\Bx}\le R$
    \begin{subequations}\label{eq:ntklc}
      \begin{align}\label{eq:ntklc1}
        \norm[]{\Bw_k-\Bw_0}&\le C\sqrt{\frac{\objF(\Bw_0)}{n}}\\ \label{eq:ntklc2}
        \objF(\Bw_k)&\le %
                      \Big(1-h\frac{n\theta_{\rm min}^{\rm NTK}}{2}\Big)^{2k}
                      \objF(\Bw_0)\\ \label{eq:ntklc3}
        \norm{\Phi(\Bx,\Bw_k)-\Phi^{\rm lin}(\Bx,\Bp_k)}&\le C\sqrt{\frac{\objF(\Bw_0)}{n}}.
      \end{align}
    \end{subequations}
  \end{theorem}

  \begin{proof}
    We wish to apply Theorems \ref{thm:ntk} and \ref{thm:ntkclose}.
    This requires Assumption \ref{ass:ntk} to be satisfied.

    Fix $\delta>0$ and let $R_0(\delta/2)$ be as in Lemma
    \ref{lemma:LCR0}, so that with probability at least $1-\delta/2$
    it holds $\sqrt{\objF(\Bw_0)}\le \sqrt{R_0}$.  Next, let $M=M(R)$
    be as in Lemma \ref{lemma:LCLU}, and fix
    \begin{equation*}
      n_{0,1}=n_{0,1}(\delta,R,\theta_{\rm min}^{\rm NTK})\in\N\qquad\text{and}\qquad \gamma = \gamma(\delta,m,R,\theta_{\rm min}^{\rm NTK})>0
    \end{equation*}
    so large that for all $n\ge n_{0,1}$
    \begin{equation}\label{eq:assntk3}
      M\sqrt{n}\le \frac{n^2(\theta_{\rm min}^{\rm NTK}/2)^2}{8m^{3/2}M^2n\sqrt{R_0}}\qquad
      \text{and}\qquad
      \gamma n^{-1/2}\ge\frac{4\sqrt{m}M\sqrt{n}}{n\theta_{\rm min}^{\rm NTK}}\sqrt{R_0}.
    \end{equation}

    By Lemma \ref{lemma:LCeigs} and \ref{lemma:LCLU}, %
    we can then find
    $n_{0,2}=n_{0,2}(\delta,m,\theta_{\rm min}^{\rm NTK},R)$ such that
    for all $n\ge n_{0,2}$ with probability at least $1-\delta/2$ we
    have that Assumption \ref{ass:ntk} \ref{item:ntk1},
    \ref{item:ntk2} holds with the values
    \begin{subequations}\label{eq:ntklcvalues}
      \begin{equation}\label{eq:ntklcvalues1}
        L = M \sqrt{n},\qquad
        U = M \sqrt{n},\qquad
        r = %
        \frac{4\sqrt{m}M\sqrt{n}}{n\theta_{\rm min}^{\rm NTK}}\sqrt{\objF(\Bw_0)},
      \end{equation}
      and
      \begin{equation}
        \theta_{\rm min}=\frac{n\theta_{\rm min}^{\rm NTK}}{2},\qquad
        \theta_{\rm max}=2n\theta_{\rm max}^{\rm NTK}.
      \end{equation}
    \end{subequations}
    Together with \eqref{eq:assntk3}, this shows that Assumption
    \ref{ass:ntk} holds with probability at least $1-\delta$ as long
    as $n\ge n_0\dfn \max\{n_{0,1},n_{0,2}\}$.

    Inequalities \eqref{eq:ntklc1}, \eqref{eq:ntklc2} are then a
    direct consequence of Theorem \ref{thm:ntk} and the definition of
    $r$ in \eqref{eq:ntklcvalues1}. For \eqref{eq:ntklc3}, we plug the
    values of \eqref{eq:ntklcvalues} into the bound in Theorem
    \ref{thm:ntkclose} to obtain for $k\in\N$
    \begin{align*}
      \norm{\Phi(\Bx,\Bw_k)-\Phi^{\rm lin}(\Bx,\Bp_k)}&\le
                                                        \frac{4\sqrt{m}ULr}{\theta_{\rm min}}\left(1+\frac{2mU^2}{\theta_{\rm min}}\right)\sqrt{\objF(\Bw_0)}\\
                                                      &\le \frac{C_1}{\sqrt{n}}(1+C_2)\sqrt{\objF(\Bw_0)},
    \end{align*}
    for some $C_1$, $C_2$ depending on $\delta$, $m$, $R$,
    $\theta_{\rm min}^{\rm NTK}$, but independent of $n$.
  \end{proof}

  \subsection{Connection to Gaussian processes and kernel
    least-squares}\label{sec:gaussianprocesses}
  \subsubsection{Gaussian processes}
  Theorem \ref{thm:ntkLC} establishes that the trained neural network
  mirrors the behavior of the trained linearized model.  As pointed
  out in Section \ref{sec:ntk}, the prediction of the trained
  linearized model corresponds to a ridgeless least squares estimator
  plus a term depending on the choice of random initialization
  $\Bw_0\in\R^n$.  We should thus understand both the model at
  initialization $\Bx\mapsto\Phi(\Bx,\Bw_0)$ and the model after
  training $\Bx\mapsto\Phi(\Bx,\Bw_k)$, as random draws of a certain
  distribution over functions.  To explain this further, we introduce
  Gaussian processes.

  \begin{definition}
    Let $(\Omega,\mathfrak{A},\bbP)$ be a probability space (see
    Section \ref{sec:sigtopmeas}), and let $g:\R^d\times
    \Omega\to\R$. We call $g$ a {\bf Gaussian process} with mean
    function $\mu:\R^d\to\R$ and covariance function
    $c:\R^d\times\R^d\to\R$ if
    \begin{enumerate}[label=(\alph*)]
    \item for each $\Bx\in\R^d$ it holds that
      $\omega\mapsto g(\Bx,\omega)$ is a random variable,
    \item for all $k\in\N$ and all $\Bx_1,\dots,\Bx_k\in\R^d$ the
      random variables $g(\Bx_1,\cdot),\dots,g(\Bx_k,\cdot)$ are
      jointly Gaussian distributed with
      \begin{equation*}
        (g(\Bx_1,\omega),\dots,g(\Bx_k,\omega))\sim \gauss\Big(\mu(\Bx_i)_{i=1}^k,(c(\Bx_i,\Bx_j))_{i,j=1}^k\Big).
      \end{equation*}
    \end{enumerate}
  \end{definition}
  In words, $g$ is a Gaussian process, if
  $\omega\mapsto g(\Bx,\omega)$ defines a collection of random
  variables indexed over $\Bx\in\R^d$, and the joint distribution of
  $(g(\Bx_1,\cdot))_{j=1}^k$ is a Gaussian whose mean and variance are
  determined by $\mu$ and $c$ respectively. Fixing $\omega\in\Omega$,
  we can then interpret $\Bx\mapsto g(\Bx,\omega)$ as a random draw
  from a distribution over functions.

  As first observed in \cite{neal1995bayesian}, certain neural
  networks at initialization tend to Gaussian processes in the
  infinite width limit. For simplicity we adopt our setting from
  before, and do not aim for optimal assumption on $\sigma$ and $\CW$
  in the following proposition.
  \begin{proposition}\label{prop:NNGP}
    Let $\sigma$, $\CW$ satisfy Assumption \ref{ass:gamma}.  Consider
    width-$n$ networks $\Phi$ as in \eqref{eq:ntkmodel} with
    initialization \eqref{eq:ntkinit}.  Let
    $K^{\rm NTK}:\R^d\times\R^d$ be as in Theorem \ref{thm:LeCun}.
    
    Then for all distinct $\Bx_1,\dots,\Bx_k\in\R^d$ it holds that
    \begin{equation*}
      \lim_{n\to\infty}(\Phi(\Bx_1,\Bw_0),\dots,\Phi(\Bx_k,\Bw_0)) \sim \gauss(\Bnul,(K^{\rm NTK}(\Bx_i,\Bx_j))_{i,j=1}^k)
    \end{equation*}
    with %
    convergence in distribution.
  \end{proposition}
  \begin{proof}
    Set $\tilde v_i\dfn \sqrt{n}v_{0,i}$ and
    $\tilde\Bu_i=(U_{0,i1},\dots,U_{0,id})\in\R^d$, so that
    $\tilde v_i\overset{\rm iid}{\sim}\CW(0,1)$, and the
    $\tilde\Bu_i\in\R^d$ are also i.i.d., with each component
    distributed according to $\CW(0,1/d)$.

    Then for any $\Bx_1,\dots,\Bx_k$
    \begin{equation*}
      \BZ_i\dfn \begin{pmatrix}
                  \tilde v_i \sigma(\tilde\Bu_i^\top\Bx_1)\\
                  \vdots\\
                  \tilde v_i \sigma(\tilde\Bu_i^\top\Bx_k)
                \end{pmatrix}\in\R^k\qquad i=1,\dots,n,
              \end{equation*}
              defines $n$ centered i.i.d.\ vectors in $\R^k$ with
              finite second moments (here we use %
              that $\sigma$ only grows linearly and that $\CW(0,1)$
              has finite second moment by Assumption \ref{ass:gamma}).
              By the central limit theorem, see Theorem \ref{thm:clt},
              \begin{equation*}
                \begin{pmatrix}
                  \Phi(\Bx_1,\Bw_0)\\
                  \vdots\\
                  \Phi(\Bx_k,\Bw_0)\\
                \end{pmatrix}
                =\frac{1}{\sqrt{n}}\sum_{i=1}^n\BZ_i
              \end{equation*}
              converges %
              in distribution to $\gauss(\Bnul,\BC)$, where
              \begin{equation*}
                C_{ij}=\bbE[\tilde v_1^2\sigma(\tilde\Bu_1^\top\Bx_i)\sigma(\tilde\Bu_1^\top\Bx_j)]
                =\bbE[\sigma(\tilde\Bu_1^\top\Bx_i)\sigma(\tilde\Bu_1^\top\Bx_j)]=K^{\rm NTK}(\Bx_i,\Bx_j).
              \end{equation*}
              This concludes the proof.
            \end{proof}

  \begin{remark}
    In the present setting the covariance function is given via the
    neural tangent kernel.  For other initialization schemes this is
    not necessarily true, and the covariance in Proposition
    \ref{prop:NNGP} is also referred to as the NNGP (neural network
    Gaussian process) kernel; also see the bibliography section.
  \end{remark}

  \subsubsection{Kernel least-squares}
  In the sense of Proposition \ref{prop:NNGP}, the network
  $\Phi(\Bx,\Bw_0)$ converges to a Gaussian process as the width $n$
  tends to infinity. It can also be shown that the linearized network
  after training corresponds to a Gaussian process, with a mean and
  covariance function that depend on the data, architecture, and
  initialization.  Since the full and linearized models coincide in
  the infinite width limit (see Theorem \ref{thm:ntkLC}) we can infer
  that wide networks post-training resemble draws from a Gaussian
  process, see \cite[Section 2.3.1]{lee2019wide} and
  \cite{Matthews2017SamplethenoptimizePS}.

  To heuristically motivate the mean of this Gaussian process,
  consider again \eqref{eq:limitlinearized}, which holds for
  sufficiently small learning rates $h>0$. Using the explicit
  formulation of the kernel-least squares estimator in
  \eqref{eq:KLSexplicit}, with
  $\Phi(\BX,\Bw_0)\dfn(\Phi(\Bx_j,\Bw_0))_{j=1}^m\in\R^m$ we have for
  any $\Bx\in\R^d$
  \begin{equation*}
    \lim_{k\to\infty}\Phi^{\rm lin}(\Bx,\Bp_k) =\Phi(\Bx,\Bw_0)
    +\hat K_n(\Bx,\BX)\hat K_n(\BX,\BX)^{-1}(\By-\Phi(\BX,\Bw_0)).
  \end{equation*}
  By Theorem \ref{thm:LeCun}, $\hat K_n/n\to K^{\rm NTK}$ as
  $n\to\infty$. For large $n$ the last expression therefore resembles
  \begin{equation}\label{eq:linposttraining}
    \Phi(\Bx,\Bw_0)+K^{\rm NTK}(\Bx,\BX)K^{\rm NTK}(\BX,\BX)^{-1}(\By-\Phi(\BX,\Bw_0)).
  \end{equation}
  Equation \eqref{eq:linposttraining} corresponds to a linear
  transformation of $\Phi(\cdot,\Bw_0)$, which itself behaves like a
  Gaussian process.

  Let us consider the mean of \eqref{eq:linposttraining} (under random
  initialization of $\Bw_0$). By Proposition \ref{prop:NNGP}, for
  large $n$, we have $\bbE[\Phi(\Bx,\Bw_0)]\simeq 0$. In fact, as the
  proof of the proposition shows, even for any finite $n$ it holds
  $\bbE[\Phi(\Bx,\Bw_0)]=0$. Taking the expectation of
  \eqref{eq:linposttraining}, we thus get
  \begin{equation*}
    K^{\rm NTK}(\Bx,\BX)K^{\rm NTK}(\BX,\BX)^{-1} \By.
  \end{equation*}
  By Remark \ref{rmk:KLSexplicit}, this is precisely the ridgeless
  kernel least squares estimator for the neural tangent kernel. We
  thus expect that for large widths $n$ and large $k$
  \begin{equation}\label{eq:meankls}
    \bbE\big[\Phi(\Bx,\Bw_k)\big] \simeq
    \bbE\big[\Phi^{\rm lin}(\Bx,\Bp_k)\big] \simeq {\substack{\text{ridgeless kernel least-squares estimator} \\ \text{with kernel $K^{\rm NTK}$ evaluated at $\Bx$}}}.
  \end{equation}
  In words, after sufficient training, the \emph{mean (over random
    initializations) of the trained neural network
    $\Bx\mapsto\Phi(\Bx,\Bw_k)$ resembles the kernel least-squares
    estimator with kernel $K^{\rm NTK}$}.  Thus, under these
  assumptions, we obtain an explicit characterization of what the
  network prediction looks like after training with gradient
  descent. For more details and a precise characterization of the
  covariance function, we refer again to \cite[Section
  2.3.1]{lee2019wide}.

  \subsubsection{A simple numerical example}
\begin{figure}[htb]
  \centering \subfloat[before
  training]{\includegraphics[width=0.49\textwidth]{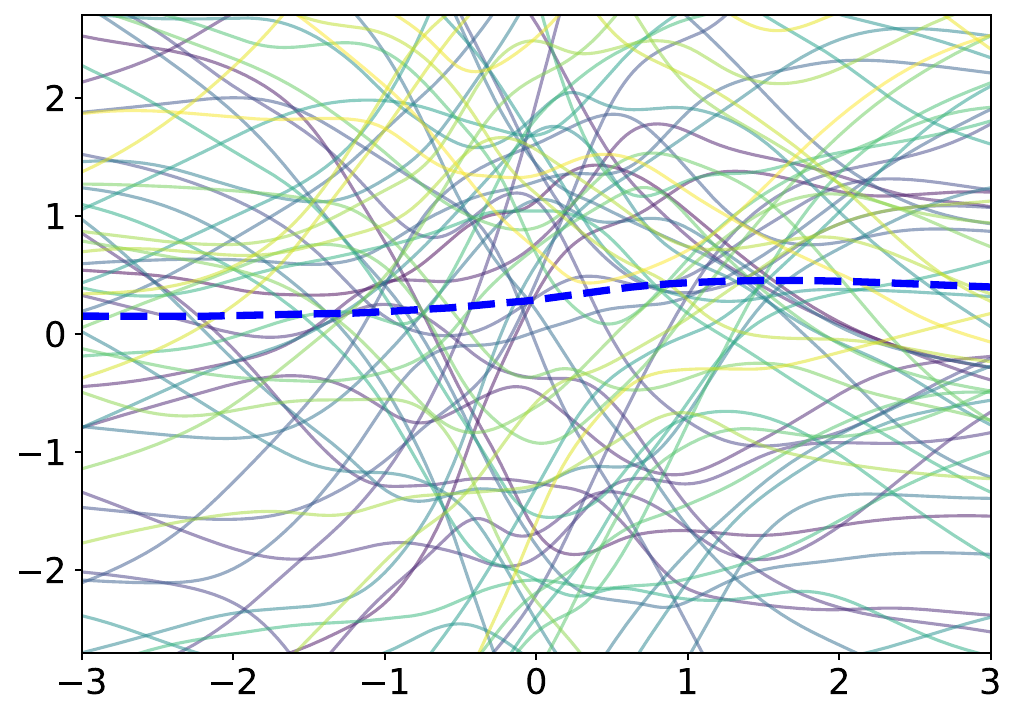}}
  \hfill \subfloat[after training without
  regularization]{\includegraphics[width=0.49\textwidth]{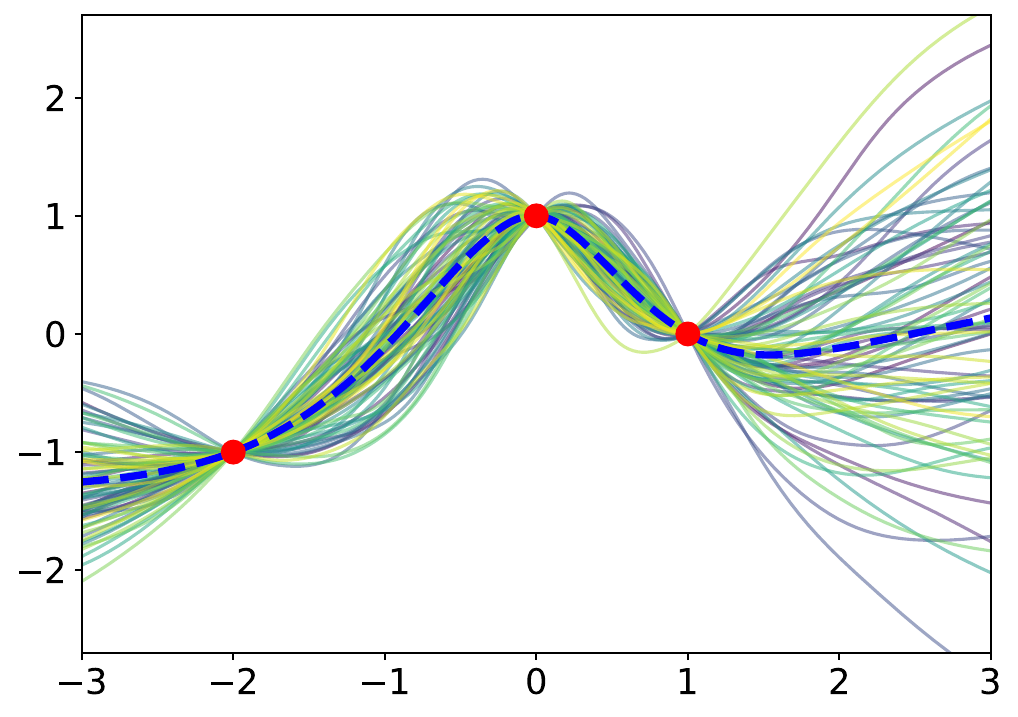}}
  \caption{80 realizations of a neural network at initialization (a)
    and after training without regularization on the red data points
    (b).  The dashed line shows the mean. Figure based on
    \cite[Fig.~2]{jacot2018neural},
    \cite[Fig.~2]{lee2019wide}.}\label{fig:ntkrealizations}
\end{figure}

  Let us now consider an experiment to visualize these
  observations. In Figure \ref{fig:ntkrealizations} we plot 80
  different realizations of a neural network before and after
  training, i.e.\ the functions
  \begin{equation}
    \Bx\mapsto\Phi(\Bx,\Bw_0)\qquad\text{and}\qquad
    \Bx\mapsto\Phi(\Bx,\Bw_k). 
  \end{equation}
  The architecture was chosen as in \eqref{eq:ntkmodel} with
  activation function $\sigma=\arctan(x)$, width $n=250$ and
  initialization
  \begin{equation}\label{eq:ex:initntk}
    U_{0;ij}\overset{\rm iid}{\sim} \gauss\Big(0,\frac{3}{d}\Big),\qquad
    v_{0;i}\overset{\rm iid}{\sim} \gauss\Big(0,\frac{3}{n}\Big),\qquad
    b_{0;i},~c_0\overset{\rm iid}{\sim}\gauss(0,2).
  \end{equation}
  This is a slight modification of the initialization
  \eqref{eq:ntkinit}, for better visualization.  The network was
  trained on the ridgeless square loss
  \begin{equation*}
    \objF(\Bw)=\sum_{j=1}^m(\Phi(\Bx_j,\Bw)-y_j)^2,
  \end{equation*}
  and a dataset of size $m=3$ with $k=5000$ steps of gradient descent
  and constant step size $h=1/n$. %
  Before training, the network's outputs resemble random draws from a
  Gaussian process with a constant zero mean function. Post-training,
  the outputs show minimal variance at the training points, since they
  essentially interpolate the data, as can be expected due to Theorem
  \ref{thm:ntkLC}, and specifically \eqref{eq:ntklc2}. Outside of the
  training points, %
  we observe increased variance.  The mean should be close to the
  ridgeless kernel least squares estimator with kernel $K^{\rm NTK}$
  by \eqref{eq:meankls}.

\begin{figure}
  \begin{center}
    \includegraphics[width=0.49\textwidth]{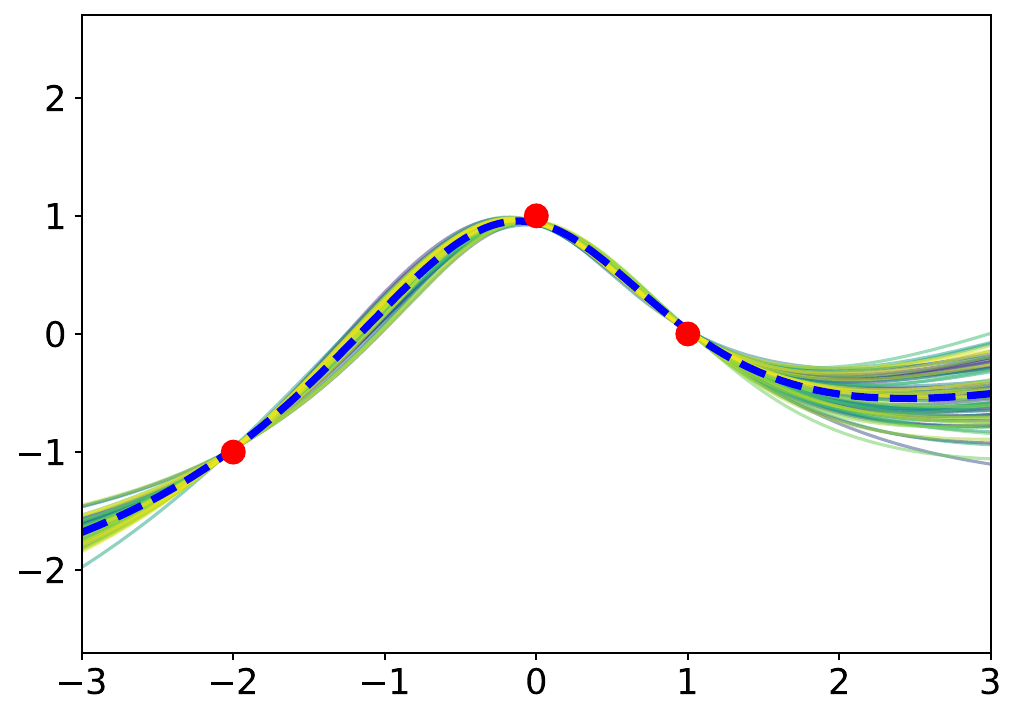}
  \end{center}
  \caption{80 realizations of the neural network in Figure
    \ref{fig:ntkrealizations} after training on the red data points
    with added ridge regularization.  The dashed line shows the
    mean.}\label{fig:ntkrealizationsridge}
\end{figure}

Figure \ref{fig:ntkrealizationsridge} shows realizations of the
network trained with ridge regularization, i.e.\ using the loss
function \eqref{eq:settingwiderisk}. Initialization and all
hyperparameters are the same as in Figure \ref{fig:ntkrealizations},
with the regularization parameter $\lambda$ set to $0.01$. For a
\emph{linear} model, the prediction after training with ridge
regularization is expected to exhibit reduced randomness, as the
trained model is $O(\lambda)$ close to the ridgeless kernel
least-squares estimator (see Section \ref{sec:kernelgd}). We note that
Theorem \ref{thm:ntkclose}, showing closeness of the trained
linearized and full model, and its analysis do not directly extend to
ridge regularization: the regularization term introduces a strong bias
towards $\Bnul$. Thus the network weights may move outside the NTK
regime during training. Nonetheless, in this example we observe a
similar effect.

\subsection{Role of initialization}\label{sec:initialization}
Consider the gradient $\nabla_\Bw\Phi(\Bx,\Bw_0)$ as in
\eqref{eq:nablaPhi} with LeCun initialization \eqref{eq:ntkinit}, so
that $v_{0;i}\overset{\rm iid}{\sim}\CW(0,1/n)$ and
$U_{0;ij}\overset{\rm iid}{\sim}\CW(0,1/d)$.  For the gradient norms
in terms of the width $n$ we obtain
\begin{equation*}
  \begin{aligned}
    \bbE[\norm{\nabla_{\BU}\Phi(\Bx,\Bw_0)}^2]&= \bbE[\norm{(\Bv_0\odot \sigma'(\BU_0\Bx))\Bx^\top}^2]&&=O(1)\\
    \bbE[\norm{\nabla_{\Bb}\Phi(\Bx,\Bw_0)}^2] &= \bbE[\norm{\Bv_0\odot \sigma'(\BU_0\Bx)}^2]&&= O(1)\\
    \bbE[\norm{\nabla_{\Bv}\Phi(\Bx,\Bw_0)}^2] &= \bbE[\norm{\sigma(\BU_0\Bx)}^2]&&=O(n)\\
    \bbE[\norm{\nabla_{c}\Phi(\Bx,\Bw_0)}^2] &= \bbE[|1|] &&=O(1).
  \end{aligned}  
\end{equation*}
Due to this different scaling, gradient descent with step size
$O(n^{-1})$ as in Theorem \ref{thm:ntkLC}, will primarily adjust the
weights $\Bv$ in the output layer, while only slightly modifying the
remaining parameters $\BU$, $\Bb$, and $c$. This is also reflected in
the expression for the obtained kernel $K^{\rm NTK}$ computed in
Theorem \ref{thm:LeCun}, which corresponds only to the contribution of
the term $\dup{\nabla_\Bv\Phi}{\nabla_\Bv\Phi}$.

LeCun initialization \cite{lecun98} sets the variance of the weight
initialization inversely proportional to the input dimension of each
layer, so that the variance of all node outputs remains stable and
does not blow up as the width increases; also see
\cite{he2015delving}. However, it does not normalize the backward
dynamics, i.e., it does not ensure that the gradients with respect to
the parameters have similar variance. To balance the normalization of
both the forward and backward dynamics, Glorot and Bengio proposed a
normalized initialization, where the variance is chosen inversely
proportional to the sum of the input and output dimensions of each
layer \cite{pmlr-v9-glorot10a}; see Exercise
\ref{ex:normalizedinit}. Importantly, the choice of initialization
strongly affects the neural tangent kernel (NTK) and, consequently,
the training dynamics as well as the predictions of the trained
network. For the so-called \emph{NTK initialization} (which does not
require learning rates decreasing like $O(n^{-1})$ as in the present
chapter), we refer in particular to the original NTK paper
\cite{jacot2018neural}. In the literature the neural tangent kernel is
often referred to as the kernel obtained with this parameterization.

\section*{Bibliography and further reading} 
The discussion on linear and kernel least-squares in Sections
\ref{sec:linreg} and \ref{sec:kernelreg} is mostly standard, and can
similarly be found in many textbooks, e.g.,
\cite{hastie_09_elements-of.statistical-learning,understanding,mohri2018foundations,bach2025learning}.
For more details on least-squares problems and algorithms see
\cite{golub,doi:10.1137/1.9781611971484,boyd04,nocedal06}, for
iterative algorithms to compute the pseudoinverse, e.g.,
\cite{doi:10.1137/0111051,PETKOVIC20111604}, and for regularization of
ill-posed problems, e.g., \cite{engl2000regularization}.  The
representer theorem was originally introduced in
\cite{KimeldorfGeorgeS.1970ACBB}, with a more general version
presented in \cite{SchoelkopfEtAl:01}.  For an easily accessible
formulation, see, e.g., \cite[Theorem 16.1]{understanding}.  The
kernel trick is commonly attributed to \cite{boser1992}, see also
\cite{Aizerman67theoretical,cortes1995support}.  For more details on
kernel methods we refer to
\cite{cristianini2000introduction,Scholkopf2002,8187598}.  For recent
works regarding in particular generalization properties of kernel
ridgeless regression see for instance
\cite{MR4124325,hastie2022surprises,MR4652869}.

The neural tangent kernel and its connection to the training dynamics
was first investigated in \cite{jacot2018neural}. %
Since then, many works have extended this idea and presented differing
perspectives on the topic, see for instance
\cite{NEURIPS2019_62dad6e2,pmlr-v97-du19c,NEURIPS2019_dbc4d84b,NEURIPS2019_ae614c55}.
Our presentation in Sections \ref{sec:ntk}--\ref{sec:ntkdynamics} is
based on and closely follows \cite{lee2019wide}, especially for the
main results in these sections, where we adhere to the arguments in
this paper. Moreover, a similar treatment of these results for
gradient flow (rather than gradient descent) was given in
\cite[Chapter 8]{telgarskynotes} based on \cite{NEURIPS2019_ae614c55}:
in particular, as in \cite{telgarskynotes}, we only consider shallow
networks and first provide an abstract analysis valid for arbitrary
function parameterizations before specifying to neural
networks. Additionally we refer to the recent textbook \cite[Chapter
9.5]{bach2025learning} by Bach. The paper \cite{lee2019wide} and some
of the other references cited above also address the case of deep
architectures.  The explicit formula for the NTK of ReLU networks as
presented in Example \ref{ex:KLCRelu} %
was given in \cite{chosaul}.

The observation that neural networks at initialization behave like
Gaussian processes discussed in Section \ref{sec:gaussianprocesses}
was first made in \cite{neal1995bayesian}. For a general reference on
Gaussian processes see the textbook
\cite{books/lib/RasmussenW06}. When only training the last layer of a
network (in which the network is affine linear), there are strong
links to random feature methods \cite{NIPS2007_013a006f}. Recent
developements on this topic can also be found in the literature under
the name ``Neural network Gaussian processes'', or NNGPs for short
\cite{lee2018deep,g.2018gaussian}.

\newpage
\section*{Exercises}

\begin{exercise}\label{ex:minnorm}
  Prove Proposition \ref{prop:LSQ}.

  \emph{Hint}: Assume first that $\Bw_0\in \ker(\BA)^\perp$ (i.e.\
  $\Bw_0\in\tilde H$). For ${\rm rank}(\BA)<d$, using
  $\Bw_{k}=\Bw_{k-1}-h\nabla\objF(\Bw_{k-1})$ and the singular value
  decomposition of $\BA$, write down an explicit formula for
  $\Bw_k$. Observe that due to $1/(1-x)=\sum_{k\in\N_0}x^k$ for all
  $x\in (0,1)$ it holds $\Bw_k\to\BA^\dagger\By$ as $k\to\infty$,
  where $\BA^\dagger$ is the Moore-Penrose pseudoinverse of $\BA$.
\end{exercise}

\begin{exercise}\label{ex:regularization}
  Let $\BA\in\R^{d\times d}$ be symmetric positive semidefinite,
  $\Bb\in\R^d$, and $c\in\R$. Let for $\lambda>0$
  \begin{equation*}
    \objF(\Bw)\dfn \Bw^\top\BA\Bw+\Bb^\top\Bw+c\qquad
    \text{and}
    \qquad
    \objF_\lambda(\Bw)\dfn \objF(\Bw)+\lambda \norm{\Bw}^2.
  \end{equation*}
  Show that $\objF$ is $2\lambda$-strongly convex.

  \emph{Hint}: Use Exercise \ref{ex:quadraticobjective}.
\end{exercise}

    \begin{exercise}\label{ex:ridge}
      Let $(H,\dup[H]{\cdot}{\cdot})$ be a Hilbert space, and
      $\phi:\R^d\to H$ a mapping.  Given
      $(\Bx_j,y_j)_{j=1}^m\in \R^d\times R$, for $\lambda>0$ denote
      \begin{equation*}
        \objF_\lambda(\Bw)\dfn \sum_{j=1}^{m}\big(\inp[H]{\phi(\Bx_j)}{\Bw}-y_j\big)^2+\lambda\norm[H]{\Bw}^2\qquad\text{for all }\Bw\in H.
      \end{equation*}
      Prove that $\objF_\lambda$ has a unique minimizer
      $\Bw_{*,\lambda}\in H$, that
      $\Bw_{*,\lambda}\in\tilde H\dfn {\rm
        span}\{\phi(\Bx_1),\dots,\phi(\Bx_m)\}$, and that
      $\lim_{\lambda\to 0}\Bw_{*,\lambda}=\Bw_*$, where $\Bw_*$ is as
      in \eqref{eq:minHnorm}.

      \emph{Hint}: Assuming existence of $\Bw_{*,\lambda}$, first show
      that $\Bw_{*,\lambda}$ belongs to the finite dimensional space
      $\tilde H$. Now express $\Bw_{*,\lambda}$ in terms of an
      orthonormal basis of $\tilde H$, and prove that
      $\Bw_{*,\lambda}\to\Bw_*$.
    \end{exercise}

\begin{exercise}
  Let $\Bx_i\in\R^d$, $i=1,\dots,m$. Show that there exists a
  ``feature map'' $\phi:\R^d\to\R^m$, such that for any configuration
  of labels $y_i\in\{-1,1\}$, there always exists a hyperplane %
  in $\R^m$ separating the two sets $\set{\phi(\Bx_i)}{y_i=1}$ and
  $\set{\phi(\Bx_i)}{y_i=-1}$.
\end{exercise}

\begin{exercise}
  Let $n\in\N$ and consider the polynomial kernel
  $K:\R^d\times\R^d\to\R$, $K(\Bx,\Bz)=(1+\Bx^\top\Bz)^n$. Find a
  Hilbert space $H$ and a feature map $\phi:\R^d\to H$, such that
  $K(\Bx,\Bz)=\inp[H]{\phi(\Bx)}{\phi(\Bz)}$.

  \emph{Hint}: Use the multinomial formula
  \begin{equation*}
    (x_1+\dots+x_k)^n = \sum_{\substack{{\Bnu\in\N_0^k}\\{|\Bnu|=n}}} \frac{n!}{\nu_1!\dots \nu_k!}x_1^{\nu_1}\dots x_k^{\nu_k}.
  \end{equation*}
\end{exercise}

\begin{exercise}\label{ex:Balphadirect}
  Let $\lambda>0$, let $K$ be a kernel (see Definition
  \ref{def:kernel}) and let $\Balpha\in\R^m$ be a minimizer in
  \eqref{eq:kernelargmin}. Show that it then holds
  \eqref{eq:Balphadirect}.
\end{exercise}

\begin{exercise}\label{ex:bias}
  Consider the network \eqref{eq:ntkmodel} with $\BU$, $\Bv$
  initialized as in \eqref{eq:ntkinit}, but with the biases instead
  initialized as
  \begin{equation}\label{eq:initbias}
    c,~b_i\overset{\rm iid}{\sim}\CW(0,1)\qquad\text{for all }i=1,\dots,n.
  \end{equation}
  Compute the corresponding NTK as in Theorem \ref{thm:ntkLC}.
\end{exercise}

\begin{exercise}\label{ex:normalizedinit}
  Consider the network \eqref{eq:ntkmodel} with initialization
  \begin{equation*}%
    U_{0;ij}\overset{\rm iid}{\sim} \CW\Big(0,\frac{1}{d+n}\Big),\qquad
    v_{0;i}\overset{\rm iid}{\sim} \CW\Big(0,\frac{1}{n+1}\Big),\qquad
    b_{0;i},~c_0=0.
  \end{equation*}
  Compute the corresponding NTK as in Theorem \ref{thm:ntkLC}.
\end{exercise}

%% file: plots/ntk.tex
\begin{tikzpicture}[scale=1.1]
\draw [->,thick] (0.528,0) -- (3.3,0);
\draw [->,thick] (0.66,-0.13199999999999998) -- (0.66,1.56);
\draw [very thick,dashed] (0.0,1.3) -- (0.01658291457286432,1.291708542713568) -- (0.03316582914572864,1.2834170854271356) -- (0.04974874371859296,1.2751256281407035) -- (0.06633165829145728,1.2668341708542714) -- (0.08291457286432159,1.2585427135678393) -- (0.09949748743718592,1.2502512562814072) -- (0.11608040201005024,1.2419597989949749) -- (0.13266331658291455,1.2336683417085428) -- (0.14924623115577887,1.2253768844221107) -- (0.16582914572864318,1.2170854271356784) -- (0.1824120603015075,1.2087939698492463) -- (0.19899497487437184,1.2005025125628142) -- (0.21557788944723616,1.192211055276382) -- (0.23216080402010048,1.18391959798995) -- (0.2487437185929648,1.1756281407035176) -- (0.2653266331658291,1.1673366834170855) -- (0.2819095477386934,1.1590452261306532) -- (0.29849246231155774,1.150753768844221) -- (0.31507537688442205,1.142462311557789) -- (0.33165829145728637,1.1341708542713569) -- (0.3482412060301507,1.1258793969849248) -- (0.364824120603015,1.1175879396984927) -- (0.3814070351758793,1.1092964824120604) -- (0.3979899497487437,1.1010050251256283) -- (0.414572864321608,1.092713567839196) -- (0.4311557788944723,1.0844221105527638) -- (0.44773869346733663,1.0761306532663317) -- (0.46432160804020095,1.0678391959798996) -- (0.48090452261306527,1.0595477386934675) -- (0.4974874371859296,1.0512562814070352) -- (0.5140703517587939,1.042964824120603) -- (0.5306532663316582,1.034673366834171) -- (0.5472361809045225,1.0263819095477387) -- (0.5638190954773868,1.0180904522613066) -- (0.5804020100502512,1.0097989949748745) -- (0.5969849246231155,1.0015075376884424) -- (0.6135678391959798,0.9932160804020101) -- (0.6301507537688441,0.9849246231155779) -- (0.6467336683417084,0.9766331658291458) -- (0.6633165829145727,0.9683417085427137) -- (0.679899497487437,0.9600502512562815) -- (0.6964824120603014,0.9517587939698493) -- (0.7130653266331657,0.9434673366834172) -- (0.72964824120603,0.9351758793969851) -- (0.7462311557788943,0.9268844221105529) -- (0.7628140703517586,0.9185929648241207) -- (0.779396984924623,0.9103015075376886) -- (0.7959798994974874,0.9020100502512564) -- (0.8125628140703517,0.8937185929648241) -- (0.829145728643216,0.885427135678392) -- (0.8457286432160803,0.8771356783919599) -- (0.8623115577889446,0.8688442211055277) -- (0.878894472361809,0.8605527638190955) -- (0.8954773869346733,0.8522613065326634) -- (0.9120603015075376,0.8439698492462313) -- (0.9286432160804019,0.8356783919597991) -- (0.9452261306532662,0.8273869346733669) -- (0.9618090452261305,0.8190954773869348) -- (0.9783919597989948,0.8108040201005027) -- (0.9949748743718592,0.8025125628140705) -- (1.0115577889447234,0.7942211055276384) -- (1.0281407035175878,0.7859296482412061) -- (1.044723618090452,0.777638190954774) -- (1.0613065326633164,0.7693467336683418) -- (1.0778894472361809,0.7610552763819096) -- (1.094472361809045,0.7527638190954775) -- (1.1110552763819095,0.7444723618090453) -- (1.1276381909547737,0.7361809045226132) -- (1.1442211055276381,0.727889447236181) -- (1.1608040201005023,0.7195979899497489) -- (1.1773869346733667,0.7113065326633167) -- (1.193969849246231,0.7030150753768846) -- (1.2105527638190954,0.6947236180904524) -- (1.2271356783919596,0.6864321608040203) -- (1.243718592964824,0.678140703517588) -- (1.2603015075376882,0.6698492462311559) -- (1.2768844221105526,0.6615577889447237) -- (1.2934673366834168,0.6532663316582916) -- (1.3100502512562813,0.6449748743718594) -- (1.3266331658291455,0.6366834170854273) -- (1.34321608040201,0.6283919597989951) -- (1.359798994974874,0.620100502512563) -- (1.3763819095477385,0.6118090452261308) -- (1.3929648241206027,0.6035175879396987) -- (1.4095477386934672,0.5952261306532665) -- (1.4261306532663314,0.5869346733668344) -- (1.4427135678391958,0.5786432160804021) -- (1.45929648241206,0.57035175879397) -- (1.4758793969849244,0.5620603015075378) -- (1.4924623115577886,0.5537688442211057) -- (1.509045226130653,0.5454773869346735) -- (1.5256281407035173,0.5371859296482414) -- (1.5422110552763817,0.5288944723618092) -- (1.558793969849246,0.5206030150753771) -- (1.5753768844221103,0.5123115577889449) -- (1.5919597989949748,0.5040201005025127) -- (1.608542713567839,0.49572864321608057) -- (1.6251256281407034,0.48743718592964835) -- (1.6417085427135676,0.47914572864321625) -- (1.658291457286432,0.47085427135678404) -- (1.6748743718592962,0.46256281407035194) -- (1.6914572864321606,0.4542713567839197) -- (1.7080402010050248,0.4459798994974876) -- (1.7246231155778893,0.4376884422110554) -- (1.7412060301507535,0.4293969849246233) -- (1.757788944723618,0.4211055276381911) -- (1.7743718592964821,0.412814070351759) -- (1.7909547738693465,0.4045226130653268) -- (1.8075376884422107,0.3962311557788947) -- (1.8241206030150752,0.38793969849246246) -- (1.8407035175879394,0.37964824120603036) -- (1.8572864321608038,0.37135678391959814) -- (1.873869346733668,0.36306532663316604) -- (1.8904522613065324,0.35477386934673383) -- (1.9070351758793966,0.3464824120603017) -- (1.923618090452261,0.3381909547738695) -- (1.9402010050251253,0.3298994974874374) -- (1.9567839195979897,0.3216080402010052) -- (1.973366834170854,0.3133165829145731) -- (1.9899497487437183,0.3050251256281409) -- (2.0065326633165825,0.2967336683417088) -- (2.0231155778894467,0.2884422110552767) -- (2.0396984924623114,0.28015075376884435) -- (2.0562814070351756,0.27185929648241225) -- (2.07286432160804,0.26356783919598015) -- (2.089447236180904,0.25527638190954804) -- (2.1060301507537686,0.24698492462311572) -- (2.122613065326633,0.23869346733668362) -- (2.139195979899497,0.23040201005025152) -- (2.1557788944723617,0.2221105527638192) -- (2.172361809045226,0.2138190954773871) -- (2.18894472361809,0.205527638190955) -- (2.2055276381909543,0.19723618090452288) -- (2.222110552763819,0.18894472361809056) -- (2.238693467336683,0.18065326633165846) -- (2.2552763819095474,0.17236180904522636) -- (2.2718592964824116,0.16407035175879425) -- (2.2884422110552762,0.15577889447236193) -- (2.3050251256281404,0.14748743718592983) -- (2.3216080402010046,0.13919597989949772) -- (2.338190954773869,0.13090452261306562) -- (2.3547738693467335,0.1226130653266333) -- (2.3713567839195977,0.1143216080402012) -- (2.387939698492462,0.10603015075376909) -- (2.404522613065326,0.09773869346733699) -- (2.4211055276381908,0.08944723618090467) -- (2.437688442211055,0.08115577889447256) -- (2.454271356783919,0.07286432160804046) -- (2.4708542713567834,0.06457286432160836) -- (2.487437185929648,0.056281407035176034) -- (2.5040201005025122,0.04798994974874393) -- (2.5206030150753764,0.03969849246231183) -- (2.5371859296482406,0.03140703517587973) -- (2.5537688442211053,0.023115577889447403) -- (2.5703517587939695,0.0148241206030153) -- (2.5869346733668337,0.006532663316583198) -- (2.603517587939698,-0.0017587939698489041) -- (2.6201005025125625,-0.010050251256281229) -- (2.6366834170854268,-0.01834170854271333) -- (2.653266331658291,-0.026633165829145433) -- (2.6698492462311556,-0.03492462311557776) -- (2.68643216080402,-0.04321608040200986) -- (2.703015075376884,-0.05150753768844196) -- (2.719597989949748,-0.059798994974874065) -- (2.736180904522613,-0.06809045226130639) -- (2.752763819095477,-0.07638190954773849) -- (2.7693467336683413,-0.0846733668341706) -- (2.7859296482412055,-0.0929648241206027) -- (2.80251256281407,-0.10125628140703502) -- (2.8190954773869343,-0.10954773869346712) -- (2.8356783919597985,-0.11783919597989923) -- (2.8522613065326627,-0.12613065326633133) -- (2.8688442211055274,-0.13442211055276365) -- (2.8854271356783916,-0.14271356783919575) -- (2.902010050251256,-0.15100502512562786) -- (2.91859296482412,-0.15929648241205996) -- (2.9351758793969847,-0.16758793969849228) -- (2.951758793969849,-0.17587939698492439) -- (2.968341708542713,-0.1841708542713565) -- (2.9849246231155773,-0.1924623115577886) -- (3.001507537688442,-0.20075376884422091) -- (3.018090452261306,-0.20904522613065302) -- (3.0346733668341703,-0.21733668341708512) -- (3.0512562814070345,-0.22562814070351722) -- (3.067839195979899,-0.23391959798994955) -- (3.0844221105527634,-0.24221105527638165) -- (3.1010050251256276,-0.25050251256281375) -- (3.117587939698492,-0.25879396984924585) -- (3.1341708542713564,-0.2670854271356782) -- (3.1507537688442206,-0.2753768844221103) -- (3.167336683417085,-0.2836683417085424) -- (3.1839195979899495,-0.2919597989949747) -- (3.2005025125628137,-0.3002512562814068) -- (3.217085427135678,-0.3085427135678389) -- (3.233668341708542,-0.316834170854271) -- (3.2502512562814068,-0.32512562814070334) -- (3.266834170854271,-0.33341708542713544) -- (3.283417085427135,-0.34170854271356754) -- (3.3,-0.34999999999999987) node [anchor=west,xshift=0] {$\Phi^{{\rm lin}}(\Bx_1,p)$};
\draw [very thick] (0.0,1.2443021834786707) -- (0.01658291457286432,1.2373808988103219) -- (0.03316582914572864,1.2300084995788854) -- (0.04974874371859296,1.2222433250407472) -- (0.06633165829145728,1.2141218919122916) -- (0.08291457286432159,1.2056963566535366) -- (0.09949748743718592,1.1970023334221473) -- (0.11608040201005024,1.1881068049800054) -- (0.13266331658291455,1.1790537738667766) -- (0.14924623115577887,1.1698774360379012) -- (0.16582914572864318,1.1606387490070034) -- (0.1824120603015075,1.1513796321551693) -- (0.19899497487437184,1.142158666068769) -- (0.21557788944723616,1.1329768073095503) -- (0.23216080402010048,1.1239226183326254) -- (0.2487437185929648,1.115019056809982) -- (0.2653266331658291,1.1062927284287465) -- (0.2819095477386934,1.0977810281143965) -- (0.29849246231155774,1.0895089819491066) -- (0.31507537688442205,1.0815049925510885) -- (0.33165829145728637,1.073779717884066) -- (0.3482412060301507,1.0663426334409587) -- (0.364824120603015,1.0591999006285686) -- (0.3814070351758793,1.0523496754166253) -- (0.3979899497487437,1.0457838675802735) -- (0.414572864321608,1.0394765029811346) -- (0.4311557788944723,1.033412231521696) -- (0.44773869346733663,1.027570320528525) -- (0.46432160804020095,1.0219150509087536) -- (0.48090452261306527,1.0164163150654928) -- (0.4974874371859296,1.0110357481012646) -- (0.5140703517587939,1.0057346920280024) -- (0.5306532663316582,1.0004779484030406) -- (0.5472361809045225,0.9952261502174169) -- (0.5638190954773868,0.9899410265778096) -- (0.5804020100502512,0.9845878466830125) -- (0.5969849246231155,0.979136475093498) -- (0.6135678391959798,0.9735551045215951) -- (0.6301507537688441,0.9678235741053837) -- (0.6467336683417084,0.9619221768505404) -- (0.6633165829145727,0.9558334817510867) -- (0.679899497487437,0.9495479412604025) -- (0.6964824120603014,0.9430616001577684) -- (0.7130653266331657,0.936370955446168) -- (0.72964824120603,0.929478255702962) -- (0.7462311557788943,0.9223913085185758) -- (0.7628140703517586,0.9151190113135025) -- (0.779396984924623,0.9076753989156781) -- (0.7959798994974874,0.9000748380138299) -- (0.8125628140703517,0.892332835733569) -- (0.829145728643216,0.8844679519377538) -- (0.8457286432160803,0.8764967227761997) -- (0.8623115577889446,0.8684383130146326) -- (0.878894472361809,0.8603088156636182) -- (0.8954773869346733,0.8521241394725666) -- (0.9120603015075376,0.8438986566597034) -- (0.9286432160804019,0.8356446483229263) -- (0.9452261306532662,0.8273725534405358) -- (0.9618090452261305,0.8190899959658313) -- (0.9783919597989948,0.8108023885167012) -- (0.9949748743718592,0.8025124621203962) -- (1.0115577889447234,0.7942204073688937) -- (1.0281407035175878,0.7859239549903777) -- (1.044723618090452,0.7776184862544608) -- (1.0613065326633164,0.7692972631547089) -- (1.0778894472361809,0.7609515724752348) -- (1.094472361809045,0.7525711937806184) -- (1.1110552763819095,0.7441443123831206) -- (1.1276381909547737,0.7356584747641746) -- (1.1442211055276381,0.7271006211499565) -- (1.1608040201005023,0.7184565284541966) -- (1.1773869346733667,0.7097139882984195) -- (1.193969849246231,0.7008598211659345) -- (1.2105527638190954,0.6918825700625907) -- (1.2271356783919596,0.6827718319345252) -- (1.243718592964824,0.6735178331347511) -- (1.2603015075376882,0.6641157043753685) -- (1.2768844221105526,0.6545584727700017) -- (1.2934673366834168,0.6448462928334537) -- (1.3100502512562813,0.6349800609487147) -- (1.3266331658291455,0.6249621531634983) -- (1.34321608040201,0.6148005760366673) -- (1.359798994974874,0.6045031565207041) -- (1.3763819095477385,0.5940865022413346) -- (1.3929648241206027,0.5835653872683247) -- (1.4095477386934672,0.5729594828063992) -- (1.4261306532663314,0.5622909421503742) -- (1.4427135678391958,0.5515828050847361) -- (1.45929648241206,0.5408595719221009) -- (1.4758793969849244,0.5301488371932629) -- (1.4924623115577886,0.5194763000079533) -- (1.509045226130653,0.5088760963619949) -- (1.5256281407035173,0.49837323131441225) -- (1.5422110552763817,0.4879969101353643) -- (1.558793969849246,0.47777343651676785) -- (1.5753768844221103,0.4677292124422225) -- (1.5919597989949748,0.4578847229824917) -- (1.608542713567839,0.44825959924746744) -- (1.6251256281407034,0.4388629412951683) -- (1.6417085427135676,0.42971153670374607) -- (1.658291457286432,0.42080974277304556) -- (1.6748743718592962,0.4121505333156955) -- (1.6914572864321606,0.4037427731833093) -- (1.7080402010050248,0.39556215519417126) -- (1.7246231155778893,0.38760913489170734) -- (1.7412060301507535,0.37986942864405965) -- (1.757788944723618,0.3723138308871528) -- (1.7743718592964821,0.3649115473124736) -- (1.7909547738693465,0.35763130929878295) -- (1.8075376884422107,0.3504712268807661) -- (1.8241206030150752,0.3433575439476726) -- (1.8407035175879394,0.33628655784413897) -- (1.8572864321608038,0.32921278965335216) -- (1.873869346733668,0.32212026041836006) -- (1.8904522613065324,0.314951509334778) -- (1.9070351758793966,0.30769406695191703) -- (1.923618090452261,0.30033002937402165) -- (1.9402010050251253,0.2928227582125062) -- (1.9567839195979897,0.28515228578934776) -- (1.973366834170854,0.2772979311836185) -- (1.9899497487437183,0.26924817128576534) -- (2.0065326633165825,0.26098847020292265) -- (2.0231155778894467,0.25251515342059955) -- (2.0396984924623114,0.24380779042855258) -- (2.0562814070351756,0.2348447186339831) -- (2.07286432160804,0.22562980494181445) -- (2.089447236180904,0.21615296039429313) -- (2.1060301507537686,0.20684647380733273) -- (2.122613065326633,0.19801460087220846) -- (2.139195979899497,0.1889469729404385) -- (2.1557788944723617,0.17961497822894995) -- (2.172361809045226,0.17000179551224923) -- (2.18894472361809,0.1601182128894839) -- (2.2055276381909543,0.1499539891957103) -- (2.222110552763819,0.1395168551965909) -- (2.238693467336683,0.128756544494643) -- (2.2552763819095474,0.11767955774447156) -- (2.2718592964824116,0.10629419726832334) -- (2.2884422110552762,0.09455970220090504) -- (2.3050251256281404,0.08247450789859898) -- (2.3216080402010046,0.0700340896242466) -- (2.338190954773869,0.05724452293242957) -- (2.3547738693467335,0.04411049382829828) -- (2.3713567839195977,0.03061404564798266) -- (2.387939698492462,0.01678028588635473) -- (2.404522613065326,0.0026164616330682094) -- (2.4211055276381908,-0.01183216284990024) -- (2.437688442211055,-0.026561195862723597) -- (2.454271356783919,-0.04156557749876924) -- (2.4708542713567834,-0.05677984671932246) -- (2.487437185929648,-0.07218536034041642) -- (2.5040201005025122,-0.08775041755811461) -- (2.5206030150753764,-0.1034195919272469) -- (2.5371859296482406,-0.11918025279146455) -- (2.5537688442211053,-0.13500476530048555) -- (2.5703517587939695,-0.15084675367137546) -- (2.5869346733668337,-0.16665878320396252) -- (2.603517587939698,-0.18243054898089092) -- (2.6201005025125625,-0.19811922166334633) -- (2.6366834170854268,-0.2137263166444641) -- (2.653266331658291,-0.22922591806892245) -- (2.6698492462311556,-0.24459108025945545) -- (2.68643216080402,-0.2598145873124085) -- (2.703015075376884,-0.2748975800805223) -- (2.719597989949748,-0.28984148357111583) -- (2.736180904522613,-0.3046349934008816) -- (2.752763819095477,-0.31928280716848156) -- (2.7693467336683413,-0.3338050323405653) -- (2.7859296482412055,-0.34820835369220315) -- (2.80251256281407,-0.3625446251973069) -- (2.8190954773869343,-0.3767568245400696) -- (2.8356783919597985,-0.39091910517632034) -- (2.8522613065326627,-0.4050137850581142) -- (2.8688442211055274,-0.41907428957577325) -- (2.8854271356783916,-0.4331048339282702) -- (2.902010050251256,-0.44713225111249444) -- (2.91859296482412,-0.46116072035499767) -- (2.9351758793969847,-0.47517734965874897) -- (2.951758793969849,-0.4892385952675013) -- (2.968341708542713,-0.5032789709148132) -- (2.9849246231155773,-0.5173196938847331) -- (3.001507537688442,-0.5313825740639952) -- (3.018090452261306,-0.5454117361998974) -- (3.0346733668341703,-0.5593872406333268) -- (3.0512562814070345,-0.5732984348158591) -- (3.067839195979899,-0.5871242890164257) -- (3.0844221105527634,-0.6008300750667944) -- (3.1010050251256276,-0.614373790816805) -- (3.117587939698492,-0.6277122645521533) -- (3.1341708542713564,-0.6408351792382153) -- (3.1507537688442206,-0.6536866107941953) -- (3.167336683417085,-0.6662079622167953) -- (3.1839195979899495,-0.6783718285640753) -- (3.2005025125628137,-0.6901543528490888) -- (3.217085427135678,-0.7015215472240789) -- (3.233668341708542,-0.7124151969595469) -- (3.2502512562814068,-0.7228202075738829) -- (3.266834170854271,-0.7327489151856763) -- (3.283417085427135,-0.7421151691770494) -- (3.3,-0.7509417869552516) node [anchor=west,xshift=0] {$\Phi(\Bx_1,w)$};
\draw (1,0.1) -- (1,-0.1) node [yshift=-6] {$w_0$};
\draw [blue] (3.3,0.3) -- (0.56,0.3) node [left,black] {$y_1$};
\draw [->,thick] (5.5280000000000005,0) -- (8.3,0);
\draw [->,thick] (5.66,-0.13200000000000003) -- (5.66,1.56);
\draw [very thick,dashed] (5.0,1.5) -- (5.016582914572864,1.4752287505366029) -- (5.033165829145728,1.4506637458650031) -- (5.049748743718593,1.4263049859852024) -- (5.066331658291458,1.4021524708971995) -- (5.082914572864322,1.378206200600995) -- (5.099497487437186,1.3544661750965887) -- (5.11608040201005,1.3309323943839801) -- (5.1326633165829145,1.30760485846317) -- (5.149246231155779,1.2844835673341586) -- (5.165829145728643,1.2615685209969443) -- (5.182412060301507,1.238859719451529) -- (5.198994974874372,1.2163571626979117) -- (5.215577889447236,1.1940608507360926) -- (5.232160804020101,1.171970783566072) -- (5.248743718592965,1.1500869611878488) -- (5.265326633165829,1.1284093836014244) -- (5.281909547738693,1.1069380508067976) -- (5.2984924623115575,1.0856729628039694) -- (5.315075376884422,1.0646141195929395) -- (5.331658291457287,1.043761521173708) -- (5.348241206030151,1.0231151675462744) -- (5.364824120603015,1.002675058710639) -- (5.381407035175879,0.9824411946668012) -- (5.397989949748744,0.9624135754147625) -- (5.414572864321608,0.9425922009545211) -- (5.431155778894472,0.9229770712860785) -- (5.447738693467336,0.9035681864094341) -- (5.464321608040201,0.8843655463245881) -- (5.4809045226130655,0.8653691510315399) -- (5.49748743718593,0.8465790005302896) -- (5.514070351758794,0.8279950948208379) -- (5.530653266331658,0.8096174339031845) -- (5.547236180904522,0.7914460177773286) -- (5.5638190954773865,0.7734808464432716) -- (5.580402010050252,0.7557219199010127) -- (5.596984924623116,0.738169238150552) -- (5.61356783919598,0.7208228011918895) -- (5.630150753768844,0.7036826090250247) -- (5.646733668341708,0.6867486616499584) -- (5.663316582914573,0.6700209590666906) -- (5.679899497487437,0.6534995012752207) -- (5.696482412060301,0.6371842882755486) -- (5.713065326633165,0.6210753200676752) -- (5.72964824120603,0.6051725966516) -- (5.7462311557788945,0.5894761180273229) -- (5.762814070351759,0.5739858841948436) -- (5.779396984924623,0.5587018951541629) -- (5.795979899497487,0.5436241509052805) -- (5.812562814070351,0.5287526514481957) -- (5.829145728643216,0.5140873967829096) -- (5.845728643216081,0.4996283869094217) -- (5.862311557788945,0.48537562182773186) -- (5.878894472361809,0.4713291015378399) -- (5.895477386934673,0.4574888260397466) -- (5.9120603015075375,0.4438547953334513) -- (5.928643216080402,0.4304270094189544) -- (5.945226130653266,0.41720546829625516) -- (5.961809045226131,0.4041901719653545) -- (5.978391959798995,0.391381120426252) -- (5.994974874371859,0.37877831367894776) -- (6.011557788944724,0.36638175172344156) -- (6.028140703517588,0.3541914345597335) -- (6.044723618090452,0.3422073621878238) -- (6.061306532663316,0.3304295346077121) -- (6.077889447236181,0.31885795181939863) -- (6.094472361809045,0.30749261382288345) -- (6.11105527638191,0.2963335206181663) -- (6.127638190954774,0.28538067220524754) -- (6.144221105527638,0.27463406858412676) -- (6.160804020100502,0.2640937097548044) -- (6.1773869346733665,0.25375959571728) -- (6.193969849246231,0.24363172647155396) -- (6.210552763819095,0.23371010201762593) -- (6.22713567839196,0.22399472235549628) -- (6.243718592964824,0.21448558748516466) -- (6.260301507537688,0.20518269740663136) -- (6.276884422110553,0.19608605211989608) -- (6.293467336683417,0.18719565162495916) -- (6.310050251256281,0.17851149592182028) -- (6.326633165829145,0.17003358501047972) -- (6.34321608040201,0.16176191889093722) -- (6.359798994974874,0.15369649756319304) -- (6.376381909547739,0.14583732102724692) -- (6.392964824120603,0.1381843892830991) -- (6.409547738693467,0.13073770233074936) -- (6.426130653266331,0.1234972601701979) -- (6.442713567839196,0.11646306280144453) -- (6.45929648241206,0.10963511022448946) -- (6.475879396984924,0.10301340243933245) -- (6.492462311557789,0.09659793944597374) -- (6.509045226130653,0.09038872124441313) -- (6.5256281407035175,0.08438574783465079) -- (6.542211055276382,0.07858901921668657) -- (6.558793969849246,0.07299853539052059) -- (6.57537688442211,0.06761429635615274) -- (6.591959798994974,0.06243630211358306) -- (6.608542713567839,0.057464552662811655) -- (6.625125628140704,0.05269904800383835) -- (6.641708542713568,0.048139788136663315) -- (6.658291457286432,0.043786773061286394) -- (6.674874371859296,0.039640002777707725) -- (6.69145728643216,0.035699477285927185) -- (6.708040201005025,0.031965196585944884) -- (6.72462311557789,0.02843716067776072) -- (6.741206030150753,0.025115369561374792) -- (6.757788944723618,0.021999823236787) -- (6.774371859296482,0.019090521703997443) -- (6.7909547738693465,0.016387464963006033) -- (6.807537688442211,0.013890653013812843) -- (6.824120603015075,0.011600085856417808) -- (6.840703517587939,0.00951576349082099) -- (6.857286432160803,0.007637685917022333) -- (6.8738693467336685,0.005965853135021885) -- (6.890452261306533,0.004500265144819603) -- (6.907035175879397,0.0032409219464155266) -- (6.923618090452261,0.002187823539809621) -- (6.940201005025125,0.001340969925001915) -- (6.9567839195979895,0.0007003611019923857) -- (6.973366834170854,0.00026599707078105053) -- (6.989949748743719,3.787783136789759e-05) -- (7.006532663316582,1.6003383752933113e-05) -- (7.023115577889447,0.00020037372793615265) -- (7.039698492462311,0.0005909888639175694) -- (7.056281407035176,0.0011878487916971625) -- (7.07286432160804,0.0019909535112749392) -- (7.089447236180904,0.0030003030226509004) -- (7.106030150753769,0.004215897325825081) -- (7.122613065326632,0.005637736420797415) -- (7.1391959798994975,0.007265820307567935) -- (7.155778894472362,0.009100148986136689) -- (7.172361809045226,0.01114072245650358) -- (7.18894472361809,0.013387540718668656) -- (7.205527638190954,0.01584060377263192) -- (7.222110552763819,0.018499911618393435) -- (7.238693467336683,0.02136546425595307) -- (7.255276381909548,0.024437261685310886) -- (7.271859296482411,0.027715303906466884) -- (7.288442211055276,0.03119959091942117) -- (7.30502512562814,0.03489012272417355) -- (7.321608040201005,0.0387868993207241) -- (7.338190954773869,0.042889920709072854) -- (7.354773869346733,0.047199186889219896) -- (7.371356783919598,0.05171469786116501) -- (7.3879396984924615,0.05643645362490831) -- (7.4045226130653266,0.0613644541804498) -- (7.421105527638191,0.06649869952778961) -- (7.437688442211055,0.07183918966692747) -- (7.454271356783919,0.0773859245978635) -- (7.470854271356783,0.08313890432059773) -- (7.4874371859296485,0.08909812883513031) -- (7.504020100502512,0.0952635981414609) -- (7.520603015075377,0.10163531223958969) -- (7.53718592964824,0.10821327112951665) -- (7.553768844221105,0.114997474811242) -- (7.5703517587939695,0.12198792328476533) -- (7.586934673366834,0.12918461655008687) -- (7.603517587939698,0.13658755460720656) -- (7.620100502512562,0.14419673745612466) -- (7.636683417085427,0.15201216509684073) -- (7.6532663316582905,0.16003383752935502) -- (7.669849246231156,0.16826175475366767) -- (7.68643216080402,0.17669591676977833) -- (7.703015075376884,0.18533632357768715) -- (7.719597989949748,0.19418297517739416) -- (7.736180904522612,0.2032358715688996) -- (7.7527638190954775,0.212495012752203) -- (7.769346733668341,0.22196039872730455) -- (7.785929648241206,0.2316320294942043) -- (7.80251256281407,0.2415099050529025) -- (7.819095477386934,0.25159402540339865) -- (7.8356783919597985,0.2618843905456929) -- (7.852261306532663,0.2723810004797854) -- (7.868844221105528,0.2830838552056764) -- (7.885427135678391,0.2939929547233653) -- (7.902010050251256,0.3051082990328523) -- (7.91859296482412,0.3164298881341375) -- (7.935175879396985,0.3279577220272213) -- (7.951758793969849,0.33969180071210286) -- (7.968341708542713,0.35163212418878265) -- (7.984924623115577,0.36377869245726063) -- (8.001507537688441,0.3761315055175372) -- (8.018090452261307,0.38869056336961155) -- (8.03467336683417,0.40145586601348415) -- (8.051256281407035,0.41442741344915485) -- (8.0678391959799,0.4276052056766241) -- (8.084422110552763,0.4409892426958911) -- (8.101005025125627,0.4545795245069564) -- (8.117587939698492,0.4683760511098199) -- (8.134170854271357,0.48237882250448194) -- (8.15075376884422,0.4965878386909417) -- (8.167336683417085,0.5110030996691998) -- (8.183919597989949,0.5256246054392564) -- (8.200502512562814,0.5404523560011107) -- (8.217085427135679,0.5554863513547633) -- (8.233668341708542,0.5707265915002141) -- (8.250251256281407,0.5861730764374635) -- (8.26683417085427,0.6018258061665105) -- (8.283417085427136,0.6176847806873559) -- (8.3,0.6337499999999998) node [anchor=west,xshift=0] {$(\Phi^{\rm lin}(\Bx_1,p)-y_1)^2$};
\draw [very thick] (5.0,1.3375599205838775) -- (5.016582914572864,1.31802442418167) -- (5.033165829145728,1.2973737139334542) -- (5.049748743718593,1.2757991258733197) -- (5.066331658291458,1.2534282499099607) -- (5.082914572864322,1.2304288356832351) -- (5.099497487437186,1.2069197792471658) -- (5.11608040201005,1.18310054557769) -- (5.1326633165829145,1.159103306024133) -- (5.149246231155779,1.1350301305918091) -- (5.165829145728643,1.1110485844385094) -- (5.182412060301507,1.0872709170730068) -- (5.198994974874372,1.0638468282520925) -- (5.215577889447236,1.0407755422734175) -- (5.232160804020101,1.0182727215001337) -- (5.248743718592965,0.996384094445149) -- (5.265326633165829,0.9751619458756084) -- (5.281909547738693,0.9546818532288952) -- (5.2984924623115575,0.9349866488674721) -- (5.315075376884422,0.9161250800734151) -- (5.331658291457287,0.8981025777131171) -- (5.348241206030151,0.8809215477438352) -- (5.364824120603015,0.8645767336716426) -- (5.381407035175879,0.849045051149252) -- (5.397989949748744,0.8342903657144863) -- (5.414572864321608,0.8202382476918118) -- (5.431155778894472,0.8068402520184508) -- (5.447738693467336,0.7940378569709707) -- (5.464321608040201,0.7817420110928823) -- (5.4809045226130655,0.7698785047380291) -- (5.49748743718593,0.7583577526168875) -- (5.514070351758794,0.7470921832977889) -- (5.530653266331658,0.7360040342983991) -- (5.547236180904522,0.7250090999191954) -- (5.5638190954773865,0.7140279302328627) -- (5.580402010050252,0.7029907797391255) -- (5.596984924623116,0.6918395277036322) -- (5.61356783919598,0.6805147182406452) -- (5.630150753768844,0.6689824891963334) -- (5.646733668341708,0.657211452309837) -- (5.663316582914573,0.6451763336786294) -- (5.679899497487437,0.6328687919934411) -- (5.696482412060301,0.6202923323962046) -- (5.713065326633165,0.6074519894032032) -- (5.72964824120603,0.5943643116042656) -- (5.7462311557788945,0.5810564113791974) -- (5.762814070351759,0.5675570971189513) -- (5.779396984924623,0.5539040856709927) -- (5.795979899497487,0.5401347168259862) -- (5.812562814070351,0.526287282432257) -- (5.829145728643216,0.5124041802634687) -- (5.845728643216081,0.49852270705754764) -- (5.862311557788945,0.48468317355438223) -- (5.878894472361809,0.4709189533655498) -- (5.895477386934673,0.45726159808248334) -- (5.9120603015075375,0.44373862307434503) -- (5.928643216080402,0.430372783915487) -- (5.945226130653266,0.4171827151835862) -- (5.961809045226131,0.4041816358677101) -- (5.978391959798995,0.39137862017155045) -- (5.994974874371859,0.378778161879454) -- (6.011557788944724,0.3663807165898128) -- (6.028140703517588,0.35418313505023596) -- (6.044723618090452,0.34217912761800395) -- (6.061306532663316,0.3303598818067502) -- (6.077889447236181,0.3187145282510875) -- (6.094472361809045,0.3072310281600211) -- (6.11105527638191,0.29589625533341257) -- (6.127638190954774,0.28469745995077045) -- (6.144221105527638,0.2736224108800181) -- (6.160804020100502,0.2626587993089068) -- (6.1773869346733665,0.2517983283110961) -- (6.193969849246231,0.24103289433777747) -- (6.210552763819095,0.23035792307829198) -- (6.22713567839196,0.21977141298376862) -- (6.243718592964824,0.2092733575045197) -- (6.260301507537688,0.19887036925915613) -- (6.276884422110553,0.18856756591949403) -- (6.293467336683417,0.1783784485214642) -- (6.310050251256281,0.16831746184980695) -- (6.326633165829145,0.15840060148298543) -- (6.34321608040201,0.1486491040095263) -- (6.359798994974874,0.13908325849660863) -- (6.376381909547739,0.1297303062008138) -- (6.392964824120603,0.12061399328495248) -- (6.409547738693467,0.1117603188809054) -- (6.426130653266331,0.10319480750119644) -- (6.442713567839196,0.0949408617214565) -- (6.45929648241206,0.08702000007974653) -- (6.475879396984924,0.07945273089211655) -- (6.492462311557789,0.0722547693977717) -- (6.509045226130653,0.0654438354471381) -- (6.5256281407035175,0.05902790835318197) -- (6.542211055276382,0.053014257330666364) -- (6.558793969849246,0.04740509209647194) -- (6.57537688442211,0.04219963305973231) -- (6.591959798994974,0.037391378626887226) -- (6.608542713567839,0.032971363153529484) -- (6.625125628140704,0.028924374697718035) -- (6.641708542713568,0.0252376241310709) -- (6.658291457286432,0.021892490923334153) -- (6.674874371859296,0.018866613184492398) -- (6.69145728643216,0.016143844481645346) -- (6.708040201005025,0.013698188258032314) -- (6.72462311557789,0.011513040774710062) -- (6.741206030150753,0.009568688447892807) -- (6.757788944723618,0.007843935206363607) -- (6.774371859296482,0.006320263461749251) -- (6.7909547738693465,0.004982051717237982) -- (6.807537688442211,0.003821017114274651) -- (6.824120603015075,0.0028198149257615424) -- (6.840703517587939,0.001975071420264066) -- (6.857286432160803,0.001280080618996499) -- (6.8738693467336685,0.000733958881464101) -- (6.890452261306533,0.00033532144708193107) -- (6.907035175879397,8.879799939087306e-05) -- (6.923618090452261,1.633790815756896e-07) -- (6.940201005025125,7.726919951422065e-05) -- (6.9567839195979895,0.00033068192592180623) -- (6.973366834170854,0.0007730758928155808) -- (6.989949748743719,0.0014185124539044403) -- (7.006532663316582,0.0022828491856623796) -- (7.023115577889447,0.0033822159820037965) -- (7.039698492462311,0.004736346624782199) -- (7.056281407035176,0.006367816034827242) -- (7.07286432160804,0.008296388869488847) -- (7.089447236180904,0.010545489075961463) -- (7.106030150753769,0.013016369163191918) -- (7.122613065326632,0.015601532452882413) -- (7.1391959798994975,0.018499162228637546) -- (7.155778894472362,0.02173883020022429) -- (7.172361809045226,0.025349299755058598) -- (7.18894472361809,0.029350371547847615) -- (7.205527638190954,0.033770708037421535) -- (7.222110552763819,0.038632259648987956) -- (7.238693467336683,0.04398648158012278) -- (7.255276381909548,0.049861115496377215) -- (7.271859296482411,0.056282907017884846) -- (7.288442211055276,0.06330857393967122) -- (7.30502512562814,0.070976009570935) -- (7.321608040201005,0.07932647990242356) -- (7.338190954773869,0.08839533246945556) -- (7.354773869346733,0.09821915905319603) -- (7.371356783919598,0.10885318860322075) -- (7.3879396984924615,0.12032010969392244) -- (7.4045226130653266,0.13265545333745457) -- (7.421105527638191,0.14585894668147004) -- (7.437688442211055,0.15996332196493818) -- (7.454271356783919,0.17500056559810162) -- (7.470854271356783,0.1909377885375948) -- (7.4874371859296485,0.2077829136775884) -- (7.504020100502512,0.22552557947473828) -- (7.520603015075377,0.24412105072611956) -- (7.53718592964824,0.2635681264954741) -- (7.553768844221105,0.28384371875119574) -- (7.5703517587939695,0.30489419294402686) -- (7.586934673366834,0.3266556299121043) -- (7.603517587939698,0.3491088518850056) -- (7.620100502512562,0.37218413848574694) -- (7.636683417085427,0.3958720926196323) -- (7.6532663316582905,0.42012010853384085) -- (7.669849246231156,0.444869167047241) -- (7.68643216080402,0.4700885582516434) -- (7.703015075376884,0.4957608413736607) -- (7.719597989949748,0.5218694636120624) -- (7.736180904522612,0.5483752128673263) -- (7.7527638190954775,0.5752667928817119) -- (7.769346733668341,0.6025632285303375) -- (7.785929648241206,0.6302611046945346) -- (7.80251256281407,0.6584480705667597) -- (7.819095477386934,0.6869996993423377) -- (7.8356783919597985,0.7160538148464709) -- (7.852261306532663,0.7455666556829532) -- (7.868844221105528,0.7756017508933545) -- (7.885427135678391,0.8061640462934949) -- (7.902010050251256,0.8373099009786353) -- (7.91859296482412,0.8690484633170086) -- (7.935175879396985,0.9013498851359436) -- (7.951758793969849,0.9343463403897281) -- (7.968341708542713,0.9678856576709419) -- (7.984924623115577,1.0020172230177509) -- (8.001507537688441,1.036795476685912) -- (8.018090452261307,1.0720815055567874) -- (8.03467336683417,1.1078196440450452) -- (8.051256281407035,1.1439752343777438) -- (8.0678391959799,1.1804842562443483) -- (8.084422110552763,1.2172422362172695) -- (8.101005025125627,1.2541191439990413) -- (8.117587939698492,1.2909750687007269) -- (8.134170854271357,1.3277562517383068) -- (8.15075376884422,1.3642772274121784) -- (8.167336683417085,1.4003367393766983) -- (8.183919597989949,1.4358171523917185) -- (8.200502512562814,1.470608463698997) -- (8.217085427135679,1.5045681143311693) -- (8.233668341708542,1.5374767965519571) -- (8.250251256281407,1.569241765532221) -- (8.26683417085427,1.5998554827257871) -- (8.283417085427136,1.6290060387433654) -- (8.3,1.656717959353046) node [anchor=west,xshift=0] {$(\Phi(\Bx_1,w)-y_1)^2$};
\draw (6,0.1) -- (6,-0.1) node [yshift=-6] {$w_0$};
\end{tikzpicture}

%% file: plots/ntk2.tex
\begin{tikzpicture}[scale=1.1]
\draw [->,thick] (0.528,0) -- (3.3,0);
\draw [->,thick] (0.66,-0.13199999999999998) -- (0.66,1.56);
\draw [very thick,dashed] (0.0,1.3) -- (0.01658291457286432,1.291708542713568) -- (0.03316582914572864,1.2834170854271356) -- (0.04974874371859296,1.2751256281407035) -- (0.06633165829145728,1.2668341708542714) -- (0.08291457286432159,1.2585427135678393) -- (0.09949748743718592,1.2502512562814072) -- (0.11608040201005024,1.2419597989949749) -- (0.13266331658291455,1.2336683417085428) -- (0.14924623115577887,1.2253768844221107) -- (0.16582914572864318,1.2170854271356784) -- (0.1824120603015075,1.2087939698492463) -- (0.19899497487437184,1.2005025125628142) -- (0.21557788944723616,1.192211055276382) -- (0.23216080402010048,1.18391959798995) -- (0.2487437185929648,1.1756281407035176) -- (0.2653266331658291,1.1673366834170855) -- (0.2819095477386934,1.1590452261306532) -- (0.29849246231155774,1.150753768844221) -- (0.31507537688442205,1.142462311557789) -- (0.33165829145728637,1.1341708542713569) -- (0.3482412060301507,1.1258793969849248) -- (0.364824120603015,1.1175879396984927) -- (0.3814070351758793,1.1092964824120604) -- (0.3979899497487437,1.1010050251256283) -- (0.414572864321608,1.092713567839196) -- (0.4311557788944723,1.0844221105527638) -- (0.44773869346733663,1.0761306532663317) -- (0.46432160804020095,1.0678391959798996) -- (0.48090452261306527,1.0595477386934675) -- (0.4974874371859296,1.0512562814070352) -- (0.5140703517587939,1.042964824120603) -- (0.5306532663316582,1.034673366834171) -- (0.5472361809045225,1.0263819095477387) -- (0.5638190954773868,1.0180904522613066) -- (0.5804020100502512,1.0097989949748745) -- (0.5969849246231155,1.0015075376884424) -- (0.6135678391959798,0.9932160804020101) -- (0.6301507537688441,0.9849246231155779) -- (0.6467336683417084,0.9766331658291458) -- (0.6633165829145727,0.9683417085427137) -- (0.679899497487437,0.9600502512562815) -- (0.6964824120603014,0.9517587939698493) -- (0.7130653266331657,0.9434673366834172) -- (0.72964824120603,0.9351758793969851) -- (0.7462311557788943,0.9268844221105529) -- (0.7628140703517586,0.9185929648241207) -- (0.779396984924623,0.9103015075376886) -- (0.7959798994974874,0.9020100502512564) -- (0.8125628140703517,0.8937185929648241) -- (0.829145728643216,0.885427135678392) -- (0.8457286432160803,0.8771356783919599) -- (0.8623115577889446,0.8688442211055277) -- (0.878894472361809,0.8605527638190955) -- (0.8954773869346733,0.8522613065326634) -- (0.9120603015075376,0.8439698492462313) -- (0.9286432160804019,0.8356783919597991) -- (0.9452261306532662,0.8273869346733669) -- (0.9618090452261305,0.8190954773869348) -- (0.9783919597989948,0.8108040201005027) -- (0.9949748743718592,0.8025125628140705) -- (1.0115577889447234,0.7942211055276384) -- (1.0281407035175878,0.7859296482412061) -- (1.044723618090452,0.777638190954774) -- (1.0613065326633164,0.7693467336683418) -- (1.0778894472361809,0.7610552763819096) -- (1.094472361809045,0.7527638190954775) -- (1.1110552763819095,0.7444723618090453) -- (1.1276381909547737,0.7361809045226132) -- (1.1442211055276381,0.727889447236181) -- (1.1608040201005023,0.7195979899497489) -- (1.1773869346733667,0.7113065326633167) -- (1.193969849246231,0.7030150753768846) -- (1.2105527638190954,0.6947236180904524) -- (1.2271356783919596,0.6864321608040203) -- (1.243718592964824,0.678140703517588) -- (1.2603015075376882,0.6698492462311559) -- (1.2768844221105526,0.6615577889447237) -- (1.2934673366834168,0.6532663316582916) -- (1.3100502512562813,0.6449748743718594) -- (1.3266331658291455,0.6366834170854273) -- (1.34321608040201,0.6283919597989951) -- (1.359798994974874,0.620100502512563) -- (1.3763819095477385,0.6118090452261308) -- (1.3929648241206027,0.6035175879396987) -- (1.4095477386934672,0.5952261306532665) -- (1.4261306532663314,0.5869346733668344) -- (1.4427135678391958,0.5786432160804021) -- (1.45929648241206,0.57035175879397) -- (1.4758793969849244,0.5620603015075378) -- (1.4924623115577886,0.5537688442211057) -- (1.509045226130653,0.5454773869346735) -- (1.5256281407035173,0.5371859296482414) -- (1.5422110552763817,0.5288944723618092) -- (1.558793969849246,0.5206030150753771) -- (1.5753768844221103,0.5123115577889449) -- (1.5919597989949748,0.5040201005025127) -- (1.608542713567839,0.49572864321608057) -- (1.6251256281407034,0.48743718592964835) -- (1.6417085427135676,0.47914572864321625) -- (1.658291457286432,0.47085427135678404) -- (1.6748743718592962,0.46256281407035194) -- (1.6914572864321606,0.4542713567839197) -- (1.7080402010050248,0.4459798994974876) -- (1.7246231155778893,0.4376884422110554) -- (1.7412060301507535,0.4293969849246233) -- (1.757788944723618,0.4211055276381911) -- (1.7743718592964821,0.412814070351759) -- (1.7909547738693465,0.4045226130653268) -- (1.8075376884422107,0.3962311557788947) -- (1.8241206030150752,0.38793969849246246) -- (1.8407035175879394,0.37964824120603036) -- (1.8572864321608038,0.37135678391959814) -- (1.873869346733668,0.36306532663316604) -- (1.8904522613065324,0.35477386934673383) -- (1.9070351758793966,0.3464824120603017) -- (1.923618090452261,0.3381909547738695) -- (1.9402010050251253,0.3298994974874374) -- (1.9567839195979897,0.3216080402010052) -- (1.973366834170854,0.3133165829145731) -- (1.9899497487437183,0.3050251256281409) -- (2.0065326633165825,0.2967336683417088) -- (2.0231155778894467,0.2884422110552767) -- (2.0396984924623114,0.28015075376884435) -- (2.0562814070351756,0.27185929648241225) -- (2.07286432160804,0.26356783919598015) -- (2.089447236180904,0.25527638190954804) -- (2.1060301507537686,0.24698492462311572) -- (2.122613065326633,0.23869346733668362) -- (2.139195979899497,0.23040201005025152) -- (2.1557788944723617,0.2221105527638192) -- (2.172361809045226,0.2138190954773871) -- (2.18894472361809,0.205527638190955) -- (2.2055276381909543,0.19723618090452288) -- (2.222110552763819,0.18894472361809056) -- (2.238693467336683,0.18065326633165846) -- (2.2552763819095474,0.17236180904522636) -- (2.2718592964824116,0.16407035175879425) -- (2.2884422110552762,0.15577889447236193) -- (2.3050251256281404,0.14748743718592983) -- (2.3216080402010046,0.13919597989949772) -- (2.338190954773869,0.13090452261306562) -- (2.3547738693467335,0.1226130653266333) -- (2.3713567839195977,0.1143216080402012) -- (2.387939698492462,0.10603015075376909) -- (2.404522613065326,0.09773869346733699) -- (2.4211055276381908,0.08944723618090467) -- (2.437688442211055,0.08115577889447256) -- (2.454271356783919,0.07286432160804046) -- (2.4708542713567834,0.06457286432160836) -- (2.487437185929648,0.056281407035176034) -- (2.5040201005025122,0.04798994974874393) -- (2.5206030150753764,0.03969849246231183) -- (2.5371859296482406,0.03140703517587973) -- (2.5537688442211053,0.023115577889447403) -- (2.5703517587939695,0.0148241206030153) -- (2.5869346733668337,0.006532663316583198) -- (2.603517587939698,-0.0017587939698489041) -- (2.6201005025125625,-0.010050251256281229) -- (2.6366834170854268,-0.01834170854271333) -- (2.653266331658291,-0.026633165829145433) -- (2.6698492462311556,-0.03492462311557776) -- (2.68643216080402,-0.04321608040200986) -- (2.703015075376884,-0.05150753768844196) -- (2.719597989949748,-0.059798994974874065) -- (2.736180904522613,-0.06809045226130639) -- (2.752763819095477,-0.07638190954773849) -- (2.7693467336683413,-0.0846733668341706) -- (2.7859296482412055,-0.0929648241206027) -- (2.80251256281407,-0.10125628140703502) -- (2.8190954773869343,-0.10954773869346712) -- (2.8356783919597985,-0.11783919597989923) -- (2.8522613065326627,-0.12613065326633133) -- (2.8688442211055274,-0.13442211055276365) -- (2.8854271356783916,-0.14271356783919575) -- (2.902010050251256,-0.15100502512562786) -- (2.91859296482412,-0.15929648241205996) -- (2.9351758793969847,-0.16758793969849228) -- (2.951758793969849,-0.17587939698492439) -- (2.968341708542713,-0.1841708542713565) -- (2.9849246231155773,-0.1924623115577886) -- (3.001507537688442,-0.20075376884422091) -- (3.018090452261306,-0.20904522613065302) -- (3.0346733668341703,-0.21733668341708512) -- (3.0512562814070345,-0.22562814070351722) -- (3.067839195979899,-0.23391959798994955) -- (3.0844221105527634,-0.24221105527638165) -- (3.1010050251256276,-0.25050251256281375) -- (3.117587939698492,-0.25879396984924585) -- (3.1341708542713564,-0.2670854271356782) -- (3.1507537688442206,-0.2753768844221103) -- (3.167336683417085,-0.2836683417085424) -- (3.1839195979899495,-0.2919597989949747) -- (3.2005025125628137,-0.3002512562814068) -- (3.217085427135678,-0.3085427135678389) -- (3.233668341708542,-0.316834170854271) -- (3.2502512562814068,-0.32512562814070334) -- (3.266834170854271,-0.33341708542713544) -- (3.283417085427135,-0.34170854271356754) -- (3.3,-0.34999999999999987) node [anchor=west,xshift=0] {$\Phi^{{\rm lin}}(\Bx_1,p)$};
\draw [very thick] (0.0,1.1438521099655388) -- (0.01658291457286432,1.1401292703876313) -- (0.03316582914572864,1.137738402790694) -- (0.04974874371859296,1.1364825106876064) -- (0.06633165829145728,1.1361362755957074) -- (0.08291457286432159,1.1364251129698115) -- (0.09949748743718592,1.13710405911444) -- (0.11608040201005024,1.1379264033346002) -- (0.13266331658291455,1.1386346820304623) -- (0.14924623115577887,1.1389902730302068) -- (0.16582914572864318,1.13881807213608) -- (0.1824120603015075,1.1379693220321778) -- (0.19899497487437184,1.1363478798681867) -- (0.21557788944723616,1.1339174648369212) -- (0.23216080402010048,1.130677067650845) -- (0.2487437185929648,1.12667609466837) -- (0.2653266331658291,1.1220109006611756) -- (0.2819095477386934,1.1168006024817527) -- (0.29849246231155774,1.1111987082416142) -- (0.31507537688442205,1.1053556070272896) -- (0.33165829145728637,1.0994293394664492) -- (0.3482412060301507,1.093572518373579) -- (0.364824120603015,1.0879021911966373) -- (0.3814070351758793,1.0825188591284922) -- (0.3979899497487437,1.0774900683804316) -- (0.414572864321608,1.0728429552098406) -- (0.4311557788944723,1.0685837313160342) -- (0.44773869346733663,1.0646716722982579) -- (0.46432160804020095,1.0610262359571647) -- (0.48090452261306527,1.0575680734459556) -- (0.4974874371859296,1.0541786364654293) -- (0.5140703517587939,1.0507473190092587) -- (0.5306532663316582,1.0471588160739196) -- (0.5472361809045225,1.0432865880377542) -- (0.5638190954773868,1.0390377343106294) -- (0.5804020100502512,1.0343159496196417) -- (0.5969849246231155,1.029051260069758) -- (0.6135678391959798,1.0231955947589937) -- (0.6301507537688441,1.0167163752456916) -- (0.6467336683417084,1.0096158465896021) -- (0.6633165829145727,1.001884774743587) -- (0.679899497487437,0.9935726505032004) -- (0.6964824120603014,0.9847129098868905) -- (0.7130653266331657,0.9753556461173325) -- (0.72964824120603,0.9655738911518361) -- (0.7462311557788943,0.9554328622515835) -- (0.7628140703517586,0.9450074784646708) -- (0.779396984924623,0.9343715628326078) -- (0.7959798994974874,0.9235977356018045) -- (0.8125628140703517,0.9127607623066244) -- (0.829145728643216,0.9019288284039849) -- (0.8457286432160803,0.8911594752366097) -- (0.8623115577889446,0.8805076515758161) -- (0.878894472361809,0.8700195008870244) -- (0.8954773869346733,0.8597316155931233) -- (0.9120603015075376,0.8496727782236106) -- (0.9286432160804019,0.8398603376107361) -- (0.9452261306532662,0.830305599489768) -- (0.9618090452261305,0.8210118653931022) -- (0.9783919597989948,0.8119833862833894) -- (0.9949748743718592,0.8032406919971955) -- (1.0115577889447234,0.7953666351616151) -- (1.0281407035175878,0.7879045197981349) -- (1.044723618090452,0.7805576706043869) -- (1.0613065326633164,0.7733548818265148) -- (1.0778894472361809,0.7663507903052938) -- (1.094472361809045,0.759614019595735) -- (1.1110552763819095,0.7532247272907641) -- (1.1276381909547737,0.7472662395978567) -- (1.1442211055276381,0.7418321453858306) -- (1.1608040201005023,0.7370067624054509) -- (1.1773869346733667,0.7328778007919973) -- (1.193969849246231,0.7295259379805941) -- (1.2105527638190954,0.7270256191763451) -- (1.2271356783919596,0.725436664470375) -- (1.243718592964824,0.7247986320678549) -- (1.2603015075376882,0.7251316806484646) -- (1.2768844221105526,0.7264316527823954) -- (1.2934673366834168,0.7286603044471376) -- (1.3100502512562813,0.731752255031126) -- (1.3266331658291455,0.735595926465993) -- (1.34321608040201,0.7400594228510422) -- (1.359798994974874,0.7449826354639052) -- (1.3763819095477385,0.7501533183507444) -- (1.3929648241206027,0.7553633640947907) -- (1.4095477386934672,0.7603870517752476) -- (1.4261306532663314,0.7649865129448353) -- (1.4427135678391958,0.7689421218833011) -- (1.45929648241206,0.7720532417315775) -- (1.4758793969849244,0.7741397020844076) -- (1.4924623115577886,0.775056337299283) -- (1.509045226130653,0.7746959608715631) -- (1.5256281407035173,0.7730075238201399) -- (1.5422110552763817,0.769966718469622) -- (1.558793969849246,0.7655981057377763) -- (1.5753768844221103,0.7599581829547268) -- (1.5919597989949748,0.7531348877615007) -- (1.608542713567839,0.7452316110762817) -- (1.6251256281407034,0.7363667701182358) -- (1.6417085427135676,0.7266758581896483) -- (1.658291457286432,0.7162758165285933) -- (1.6748743718592962,0.7053200598981904) -- (1.6914572864321606,0.6939141778099489) -- (1.7080402010050248,0.682194855999831) -- (1.7246231155778893,0.670281424558224) -- (1.7412060301507535,0.658302146651129) -- (1.757788944723618,0.6463870697386697) -- (1.7743718592964821,0.6346786862836428) -- (1.7909547738693465,0.6233358474528503) -- (1.8075376884422107,0.6124878195400352) -- (1.8241206030150752,0.6022844412766065) -- (1.8407035175879394,0.5928561014017011) -- (1.8572864321608038,0.5843243853118105) -- (1.873869346733668,0.5768174123697093) -- (1.8904522613065324,0.5703742823917738) -- (1.9070351758793966,0.565039773590563) -- (1.923618090452261,0.5608259125808495) -- (1.9402010050251253,0.5576527102908481) -- (1.9567839195979897,0.5554690041961452) -- (1.973366834170854,0.5541401849557807) -- (1.9899497487437183,0.5535306343566304) -- (2.0065326633165825,0.5534788882752492) -- (2.0231155778894467,0.5538379184652569) -- (2.0396984924623114,0.554415766035766) -- (2.0562814070351756,0.5551082166806212) -- (2.07286432160804,0.555847609679502) -- (2.089447236180904,0.5565464731930305) -- (2.1060301507537686,0.5572052071417091) -- (2.122613065326633,0.5578770777279887) -- (2.139195979899497,0.5586469401003433) -- (2.1557788944723617,0.5596181132903055) -- (2.172361809045226,0.5609663271382778) -- (2.18894472361809,0.5628403263481807) -- (2.2055276381909543,0.5653804754412552) -- (2.222110552763819,0.5687528735195522) -- (2.238693467336683,0.5730257602518571) -- (2.2552763819095474,0.5783451312153229) -- (2.2718592964824116,0.5846780455446807) -- (2.2884422110552762,0.5920208024333348) -- (2.3050251256281404,0.6002959537667062) -- (2.3216080402010046,0.6093638304574747) -- (2.338190954773869,0.6190555499712087) -- (2.3547738693467335,0.6291421199457061) -- (2.3713567839195977,0.6394520198165023) -- (2.387939698492462,0.6497580822307418) -- (2.404522613065326,0.6598426511620824) -- (2.4211055276381908,0.6695262209658703) -- (2.437688442211055,0.6787049497503436) -- (2.454271356783919,0.6873209639171787) -- (2.4708542713567834,0.6953768999788602) -- (2.487437185929648,0.7029181762157413) -- (2.5040201005025122,0.710104452761929) -- (2.5206030150753764,0.7171248501289894) -- (2.5371859296482406,0.7241984889991546) -- (2.5537688442211053,0.7315695895094567) -- (2.5703517587939695,0.7394459260739343) -- (2.5869346733668337,0.7480943967911334) -- (2.603517587939698,0.757660083382725) -- (2.6201005025125625,0.7682444695524902) -- (2.6366834170854268,0.7799568013475158) -- (2.653266331658291,0.7927088842494024) -- (2.6698492462311556,0.8064158872845338) -- (2.68643216080402,0.820901818824165) -- (2.703015075376884,0.8359160891548177) -- (2.719597989949748,0.8511953945593361) -- (2.736180904522613,0.866413342089565) -- (2.752763819095477,0.8812998683209273) -- (2.7693467336683413,0.8955195672264323) -- (2.7859296482412055,0.9088798926343367) -- (2.80251256281407,0.9211381798784863) -- (2.8190954773869343,0.9321811854663646) -- (2.8356783919597985,0.9419718352522028) -- (2.8522613065326627,0.9505321331488612) -- (2.8688442211055274,0.9579668491795128) -- (2.8854271356783916,0.9643465536867548) -- (2.902010050251256,0.9698803103934992) -- (2.91859296482412,0.9747571807574638) -- (2.9351758793969847,0.979186321533599) -- (2.951758793969849,0.9833844490412411) -- (2.968341708542713,0.9875639592506427) -- (2.9849246231155773,0.9918858564834969) -- (3.001507537688442,0.9964319866233282) -- (3.018090452261306,1.0014529471041849) -- (3.0346733668341703,1.0070114150301346) -- (3.0512562814070345,1.0132044975048877) -- (3.067839195979899,1.020008117810319) -- (3.0844221105527634,1.0274819132745743) -- (3.1010050251256276,1.0356266715016027) -- (3.117587939698492,1.044381575193924) -- (3.1341708542713564,1.053693641159227) -- (3.1507537688442206,1.0634963569325322) -- (3.167336683417085,1.0735854704067889) -- (3.1839195979899495,1.0838150621026788) -- (3.2005025125628137,1.0940767662152389) -- (3.217085427135678,1.104149247530591) -- (3.233668341708542,1.1138500479940983) -- (3.2502512562814068,1.1229434173913475) -- (3.266834170854271,1.1312564877112519) -- (3.283417085427135,1.1386087104342697) -- (3.3,1.144923371435461) node [anchor=west,xshift=0] {$\Phi(\Bx_1,w)$};
\draw (1,0.1) -- (1,-0.1) node [yshift=-6] {$w_0$};
\draw [blue] (3.3,0.3) -- (0.56,0.3) node [left,black] {$y_1$};
\draw [->,thick] (5.5280000000000005,0) -- (8.3,0);
\draw [->,thick] (5.66,-0.13200000000000003) -- (5.66,1.56);
\draw [very thick,dashed] (5.0,1.5) -- (5.016582914572864,1.4752287505366029) -- (5.033165829145728,1.4506637458650031) -- (5.049748743718593,1.4263049859852024) -- (5.066331658291458,1.4021524708971995) -- (5.082914572864322,1.378206200600995) -- (5.099497487437186,1.3544661750965887) -- (5.11608040201005,1.3309323943839801) -- (5.1326633165829145,1.30760485846317) -- (5.149246231155779,1.2844835673341586) -- (5.165829145728643,1.2615685209969443) -- (5.182412060301507,1.238859719451529) -- (5.198994974874372,1.2163571626979117) -- (5.215577889447236,1.1940608507360926) -- (5.232160804020101,1.171970783566072) -- (5.248743718592965,1.1500869611878488) -- (5.265326633165829,1.1284093836014244) -- (5.281909547738693,1.1069380508067976) -- (5.2984924623115575,1.0856729628039694) -- (5.315075376884422,1.0646141195929395) -- (5.331658291457287,1.043761521173708) -- (5.348241206030151,1.0231151675462744) -- (5.364824120603015,1.002675058710639) -- (5.381407035175879,0.9824411946668012) -- (5.397989949748744,0.9624135754147625) -- (5.414572864321608,0.9425922009545211) -- (5.431155778894472,0.9229770712860785) -- (5.447738693467336,0.9035681864094341) -- (5.464321608040201,0.8843655463245881) -- (5.4809045226130655,0.8653691510315399) -- (5.49748743718593,0.8465790005302896) -- (5.514070351758794,0.8279950948208379) -- (5.530653266331658,0.8096174339031845) -- (5.547236180904522,0.7914460177773286) -- (5.5638190954773865,0.7734808464432716) -- (5.580402010050252,0.7557219199010127) -- (5.596984924623116,0.738169238150552) -- (5.61356783919598,0.7208228011918895) -- (5.630150753768844,0.7036826090250247) -- (5.646733668341708,0.6867486616499584) -- (5.663316582914573,0.6700209590666906) -- (5.679899497487437,0.6534995012752207) -- (5.696482412060301,0.6371842882755486) -- (5.713065326633165,0.6210753200676752) -- (5.72964824120603,0.6051725966516) -- (5.7462311557788945,0.5894761180273229) -- (5.762814070351759,0.5739858841948436) -- (5.779396984924623,0.5587018951541629) -- (5.795979899497487,0.5436241509052805) -- (5.812562814070351,0.5287526514481957) -- (5.829145728643216,0.5140873967829096) -- (5.845728643216081,0.4996283869094217) -- (5.862311557788945,0.48537562182773186) -- (5.878894472361809,0.4713291015378399) -- (5.895477386934673,0.4574888260397466) -- (5.9120603015075375,0.4438547953334513) -- (5.928643216080402,0.4304270094189544) -- (5.945226130653266,0.41720546829625516) -- (5.961809045226131,0.4041901719653545) -- (5.978391959798995,0.391381120426252) -- (5.994974874371859,0.37877831367894776) -- (6.011557788944724,0.36638175172344156) -- (6.028140703517588,0.3541914345597335) -- (6.044723618090452,0.3422073621878238) -- (6.061306532663316,0.3304295346077121) -- (6.077889447236181,0.31885795181939863) -- (6.094472361809045,0.30749261382288345) -- (6.11105527638191,0.2963335206181663) -- (6.127638190954774,0.28538067220524754) -- (6.144221105527638,0.27463406858412676) -- (6.160804020100502,0.2640937097548044) -- (6.1773869346733665,0.25375959571728) -- (6.193969849246231,0.24363172647155396) -- (6.210552763819095,0.23371010201762593) -- (6.22713567839196,0.22399472235549628) -- (6.243718592964824,0.21448558748516466) -- (6.260301507537688,0.20518269740663136) -- (6.276884422110553,0.19608605211989608) -- (6.293467336683417,0.18719565162495916) -- (6.310050251256281,0.17851149592182028) -- (6.326633165829145,0.17003358501047972) -- (6.34321608040201,0.16176191889093722) -- (6.359798994974874,0.15369649756319304) -- (6.376381909547739,0.14583732102724692) -- (6.392964824120603,0.1381843892830991) -- (6.409547738693467,0.13073770233074936) -- (6.426130653266331,0.1234972601701979) -- (6.442713567839196,0.11646306280144453) -- (6.45929648241206,0.10963511022448946) -- (6.475879396984924,0.10301340243933245) -- (6.492462311557789,0.09659793944597374) -- (6.509045226130653,0.09038872124441313) -- (6.5256281407035175,0.08438574783465079) -- (6.542211055276382,0.07858901921668657) -- (6.558793969849246,0.07299853539052059) -- (6.57537688442211,0.06761429635615274) -- (6.591959798994974,0.06243630211358306) -- (6.608542713567839,0.057464552662811655) -- (6.625125628140704,0.05269904800383835) -- (6.641708542713568,0.048139788136663315) -- (6.658291457286432,0.043786773061286394) -- (6.674874371859296,0.039640002777707725) -- (6.69145728643216,0.035699477285927185) -- (6.708040201005025,0.031965196585944884) -- (6.72462311557789,0.02843716067776072) -- (6.741206030150753,0.025115369561374792) -- (6.757788944723618,0.021999823236787) -- (6.774371859296482,0.019090521703997443) -- (6.7909547738693465,0.016387464963006033) -- (6.807537688442211,0.013890653013812843) -- (6.824120603015075,0.011600085856417808) -- (6.840703517587939,0.00951576349082099) -- (6.857286432160803,0.007637685917022333) -- (6.8738693467336685,0.005965853135021885) -- (6.890452261306533,0.004500265144819603) -- (6.907035175879397,0.0032409219464155266) -- (6.923618090452261,0.002187823539809621) -- (6.940201005025125,0.001340969925001915) -- (6.9567839195979895,0.0007003611019923857) -- (6.973366834170854,0.00026599707078105053) -- (6.989949748743719,3.787783136789759e-05) -- (7.006532663316582,1.6003383752933113e-05) -- (7.023115577889447,0.00020037372793615265) -- (7.039698492462311,0.0005909888639175694) -- (7.056281407035176,0.0011878487916971625) -- (7.07286432160804,0.0019909535112749392) -- (7.089447236180904,0.0030003030226509004) -- (7.106030150753769,0.004215897325825081) -- (7.122613065326632,0.005637736420797415) -- (7.1391959798994975,0.007265820307567935) -- (7.155778894472362,0.009100148986136689) -- (7.172361809045226,0.01114072245650358) -- (7.18894472361809,0.013387540718668656) -- (7.205527638190954,0.01584060377263192) -- (7.222110552763819,0.018499911618393435) -- (7.238693467336683,0.02136546425595307) -- (7.255276381909548,0.024437261685310886) -- (7.271859296482411,0.027715303906466884) -- (7.288442211055276,0.03119959091942117) -- (7.30502512562814,0.03489012272417355) -- (7.321608040201005,0.0387868993207241) -- (7.338190954773869,0.042889920709072854) -- (7.354773869346733,0.047199186889219896) -- (7.371356783919598,0.05171469786116501) -- (7.3879396984924615,0.05643645362490831) -- (7.4045226130653266,0.0613644541804498) -- (7.421105527638191,0.06649869952778961) -- (7.437688442211055,0.07183918966692747) -- (7.454271356783919,0.0773859245978635) -- (7.470854271356783,0.08313890432059773) -- (7.4874371859296485,0.08909812883513031) -- (7.504020100502512,0.0952635981414609) -- (7.520603015075377,0.10163531223958969) -- (7.53718592964824,0.10821327112951665) -- (7.553768844221105,0.114997474811242) -- (7.5703517587939695,0.12198792328476533) -- (7.586934673366834,0.12918461655008687) -- (7.603517587939698,0.13658755460720656) -- (7.620100502512562,0.14419673745612466) -- (7.636683417085427,0.15201216509684073) -- (7.6532663316582905,0.16003383752935502) -- (7.669849246231156,0.16826175475366767) -- (7.68643216080402,0.17669591676977833) -- (7.703015075376884,0.18533632357768715) -- (7.719597989949748,0.19418297517739416) -- (7.736180904522612,0.2032358715688996) -- (7.7527638190954775,0.212495012752203) -- (7.769346733668341,0.22196039872730455) -- (7.785929648241206,0.2316320294942043) -- (7.80251256281407,0.2415099050529025) -- (7.819095477386934,0.25159402540339865) -- (7.8356783919597985,0.2618843905456929) -- (7.852261306532663,0.2723810004797854) -- (7.868844221105528,0.2830838552056764) -- (7.885427135678391,0.2939929547233653) -- (7.902010050251256,0.3051082990328523) -- (7.91859296482412,0.3164298881341375) -- (7.935175879396985,0.3279577220272213) -- (7.951758793969849,0.33969180071210286) -- (7.968341708542713,0.35163212418878265) -- (7.984924623115577,0.36377869245726063) -- (8.001507537688441,0.3761315055175372) -- (8.018090452261307,0.38869056336961155) -- (8.03467336683417,0.40145586601348415) -- (8.051256281407035,0.41442741344915485) -- (8.0678391959799,0.4276052056766241) -- (8.084422110552763,0.4409892426958911) -- (8.101005025125627,0.4545795245069564) -- (8.117587939698492,0.4683760511098199) -- (8.134170854271357,0.48237882250448194) -- (8.15075376884422,0.4965878386909417) -- (8.167336683417085,0.5110030996691998) -- (8.183919597989949,0.5256246054392564) -- (8.200502512562814,0.5404523560011107) -- (8.217085427135679,0.5554863513547633) -- (8.233668341708542,0.5707265915002141) -- (8.250251256281407,0.5861730764374635) -- (8.26683417085427,0.6018258061665105) -- (8.283417085427136,0.6176847806873559) -- (8.3,0.6337499999999998) node [anchor=west,xshift=0] {$(\Phi^{\rm lin}(\Bx_1,p)-y_1)^2$};
\draw [very thick] (5.0,1.0681295752399376) -- (5.016582914572864,1.0587257864430806) -- (5.033165829145728,1.0527084472654546) -- (5.049748743718593,1.0495544860293622) -- (5.066331658291458,1.0486858070505911) -- (5.082914572864322,1.0494104544098428) -- (5.099497487437186,1.0511148086788074) -- (5.11608040201005,1.0531809861078885) -- (5.1326633165829145,1.0549621948565018) -- (5.149246231155779,1.0558570173589514) -- (5.165829145728643,1.055423637213135) -- (5.182412060301507,1.0532888770006015) -- (5.198994974874372,1.0492166642400163) -- (5.215577889447236,1.0431275072400565) -- (5.232160804020101,1.0350365860815098) -- (5.248743718592965,1.0250900482442218) -- (5.265326633165829,1.0135528812086956) -- (5.281909547738693,1.0007448363218314) -- (5.2984924623115575,0.9870650163792951) -- (5.315075376884422,0.9728964806554412) -- (5.331658291457287,0.9586309031996447) -- (5.348241206030151,0.9446360128766763) -- (5.364824120603015,0.9311847943386936) -- (5.381407035175879,0.9185036473376356) -- (5.397989949748744,0.9067362096453122) -- (5.414572864321608,0.8959293501262195) -- (5.431155778894472,0.8860814280655167) -- (5.447738693467336,0.8770841496231213) -- (5.464321608040201,0.8687413977226951) -- (5.4809045226130655,0.8608640788569251) -- (5.49748743718593,0.8531781235512811) -- (5.514070351758794,0.8454323054993842) -- (5.530653266331658,0.8373694446554716) -- (5.547236180904522,0.8287124279352092) -- (5.5638190954773865,0.8192651591024827) -- (5.580402010050252,0.8088298707986941) -- (5.596984924623116,0.7972736097139526) -- (5.61356783919598,0.7845178024182218) -- (5.630150753768844,0.7705235438179845) -- (5.646733668341708,0.7553319745966165) -- (5.663316582914573,0.7389633555252838) -- (5.679899497487437,0.721564532289052) -- (5.696482412060301,0.7032476534486595) -- (5.713065326633165,0.6841578731138394) -- (5.72964824120603,0.6644829068744942) -- (5.7462311557788945,0.6443883553789548) -- (5.762814070351759,0.6240519709130293) -- (5.779396984924623,0.6036409195960281) -- (5.795979899497487,0.5833112037715472) -- (5.812562814070351,0.5632136277338933) -- (5.829145728643216,0.5434774716956909) -- (5.845728643216081,0.5242042877430355) -- (5.862311557788945,0.5054837003071037) -- (5.878894472361809,0.4873833470872385) -- (5.895477386934673,0.46994922224173197) -- (5.9120603015075375,0.45321024468009385) -- (5.928643216080402,0.43717377618776687) -- (5.945226130653266,0.42183604327530344) -- (5.961809045226131,0.4071800458206001) -- (5.978391959798995,0.3931904817453094) -- (5.994974874371859,0.3798767911227242) -- (6.011557788944724,0.368082154847011) -- (6.028140703517588,0.35707623065917293) -- (6.044723618090452,0.3464035121650716) -- (6.061306532663316,0.3360972662234907) -- (6.077889447236181,0.3262245894275583) -- (6.094472361809045,0.31686757051342307) -- (6.11105527638191,0.3081189801416812) -- (6.127638190954774,0.30007063362601094) -- (6.144221105527638,0.2928234670443687) -- (6.160804020100502,0.2864623655821414) -- (6.1773869346733665,0.2810747856277741) -- (6.193969849246231,0.2767387970971637) -- (6.210552763819095,0.2735263191494114) -- (6.22713567839196,0.2714945332135177) -- (6.243718592964824,0.27068081671008115) -- (6.260301507537688,0.2711054188364821) -- (6.276884422110553,0.2727659317420882) -- (6.293467336683417,0.27562448491306907) -- (6.310050251256281,0.27961501458669363) -- (6.326633165829145,0.2846157167306502) -- (6.34321608040201,0.2904784434599885) -- (6.359798994974874,0.2970143187966041) -- (6.376381909547739,0.30395701503328) -- (6.392964824120603,0.31103369003958736) -- (6.409547738693467,0.3179343561634567) -- (6.426130653266331,0.3243186858308963) -- (6.442713567839196,0.3298600705146193) -- (6.45929648241206,0.33425139454393676) -- (6.475879396984924,0.3372126856390363) -- (6.492462311557789,0.33851778541231525) -- (6.509045226130653,0.3380043829016649) -- (6.5256281407035175,0.3356041763856904) -- (6.542211055276382,0.3313030747036574) -- (6.558793969849246,0.32517239409990834) -- (6.57537688442211,0.317342295100521) -- (6.591959798994974,0.30799683975994185) -- (6.608542713567839,0.297346781252372) -- (6.625125628140704,0.2856239370951319) -- (6.641708542713568,0.2730784319428093) -- (6.658291457286432,0.2599283331398206) -- (6.674874371859296,0.24642652643380894) -- (6.69145728643216,0.23275256921953208) -- (6.708040201005025,0.21910936192909736) -- (6.72462311557789,0.20566250005930165) -- (6.741206030150753,0.19257064244221073) -- (6.757788944723618,0.17997600312321302) -- (6.774371859296482,0.16801473457881747) -- (6.7909547738693465,0.15681910537207933) -- (6.807537688442211,0.1464729560413284) -- (6.824120603015075,0.13706382515686527) -- (6.840703517587939,0.12864704419230516) -- (6.857286432160803,0.12126053412440836) -- (6.8738693467336685,0.11494181968659253) -- (6.890452261306533,0.10965337886829996) -- (6.907035175879397,0.10536912237740537) -- (6.923618090452261,0.1020452350104494) -- (6.940201005025125,0.09957737868032954) -- (6.9567839195979895,0.09789661815745508) -- (6.973366834170854,0.0968808504140376) -- (6.989949748743719,0.09641667383591314) -- (7.006532663316582,0.09637732020188437) -- (7.023115577889447,0.0966505332761616) -- (7.039698492462311,0.09709107301134846) -- (7.056281407035176,0.09762030332695014) -- (7.07286432160804,0.0981869990680722) -- (7.089447236180904,0.09872413936167349) -- (7.106030150753769,0.09923177787121422) -- (7.122613065326632,0.09975088082629069) -- (7.1391959798994975,0.1003473594349059) -- (7.155778894472362,0.10110234712262689) -- (7.172361809045226,0.10215513585006393) -- (7.18894472361809,0.10362755573222722) -- (7.205527638190954,0.10564019511814005) -- (7.222110552763819,0.10834216053752463) -- (7.238693467336683,0.11181459864165688) -- (7.255276381909548,0.11621401810691301) -- (7.271859296482411,0.12156238442270899) -- (7.288442211055276,0.1279142235807131) -- (7.30502512562814,0.1352664897729836) -- (7.321608040201005,0.1435589693929817) -- (7.338190954773869,0.1526946659511457) -- (7.354773869346733,0.16250180268353043) -- (7.371356783919598,0.17284151063625466) -- (7.3879396984924615,0.18349607412858956) -- (7.4045226130653266,0.19423010039303423) -- (7.421105527638191,0.20482444197197586) -- (7.437688442211055,0.2151261584481154) -- (7.454271356783919,0.22502629363459864) -- (7.470854271356783,0.23448433955534048) -- (7.4874371859296485,0.2435145850875287) -- (7.504020100502512,0.25227849326274193) -- (7.520603015075377,0.26098971089269785) -- (7.53718592964824,0.26991653710374885) -- (7.553768844221105,0.27937846588404147) -- (7.5703517587939695,0.28966908291446664) -- (7.586934673366834,0.3011828826534146) -- (7.603517587939698,0.31417912788282415) -- (7.620100502512562,0.32887932489973937) -- (7.636683417085427,0.3455377967396081) -- (7.6532663316582905,0.3641430669274366) -- (7.669849246231156,0.38468557634127265) -- (7.68643216080402,0.4070080572814848) -- (7.703015075376884,0.430809081922492) -- (7.719597989949748,0.45572454447513344) -- (7.736180904522612,0.4812361111456058) -- (7.7527638190954775,0.5068643053648914) -- (7.769346733668341,0.5319653324243361) -- (7.785929648241206,0.5561020854816019) -- (7.80251256281407,0.5787189577541383) -- (7.819095477386934,0.5994795768864871) -- (7.8356783919597985,0.6181917558856221) -- (7.852261306532663,0.6347880843888114) -- (7.868844221105528,0.6493805619288235) -- (7.885427135678391,0.6620345150932022) -- (7.902010050251256,0.6731094453793361) -- (7.91859296482412,0.6829458794756409) -- (7.935175879396985,0.6919410890375118) -- (7.951758793969849,0.700521457787101) -- (7.968341708542713,0.7091162970906291) -- (7.984924623115577,0.718059057602853) -- (8.001507537688441,0.7275262679881733) -- (8.018090452261307,0.7380543555017194) -- (8.03467336683417,0.7497977114743697) -- (8.051256281407035,0.7629909828917989) -- (8.0678391959799,0.7776175345691374) -- (8.084422110552763,0.7938449012124527) -- (8.101005025125627,0.8117198997367903) -- (8.117587939698492,0.8311558942322813) -- (8.134170854271357,0.8520811570857805) -- (8.15075376884422,0.8743900305738728) -- (8.167336683417085,0.8976517200367391) -- (8.183919597989949,0.9215490773685392) -- (8.200502512562814,0.9458368659642766) -- (8.217085427135679,0.9699840184560233) -- (8.233668341708542,0.9935278509299938) -- (8.250251256281407,1.0158538023416241) -- (8.26683417085427,1.0364810225430698) -- (8.283417085427136,1.054896853824343) -- (8.3,1.0708432553967988) node [anchor=west,xshift=0] {$(\Phi(\Bx_1,w)-y_1)^2$};
\draw (6,0.1) -- (6,-0.1) node [yshift=-6] {$w_0$};
\end{tikzpicture}

%% file: LossLandscapeAnalysis.tex
\chapter{Loss landscape analysis}\label{chap:LossLandscapes}
In Chapter \ref{chap:training}, we saw how the weights of neural networks get adapted during training, using, e.g., variants of gradient descent.
For certain cases, including the wide networks considered in Chapter \ref{chap:wideNets}, the corresponding iterative scheme converges to a global minimizer.
In general, this is not guaranteed, and gradient descent can for instance get stuck in non-global minima or saddle points.

To get a better understanding of these situations, in this chapter we discuss the so-called loss landscape.
This term refers to the graph of the empirical risk as a function of the weights.
We give a more rigorous definition below,
and first  
introduce notation for neural networks and their 
realizations for a fixed architecture. 

\begin{definition}\label{def:realizationetc}
	Let $\mathcal{A} = (d_0, d_1, \dots, d_{L+1}) \in \N^{L+2}$, let $\sigma\colon \R \to \R$ be an activation function, and let $B>0$.
We denote the set of neural networks $\Phi$ with $L$ layers, layer widths $d_0, d_1, \dots, d_{L+1}$, all weights bounded in modulus by $B$, and using the activation function $\sigma$ by $\mathcal{N}(\sigma; \mathcal{A}, B)$.
Additionally, we define
	\begin{align*}
		\mathcal{PN}(\mathcal{A}, B) \coloneqq \bigtimes_{\ell=0}^{L} \left([-B,B]^{d_{\ell+1} \times d_{\ell}} \times [-B,B]^{d_{\ell+1}}\right),
	\end{align*}
        and the \textbf{realization map}
	\begin{align}\label{eq:Rsigmatheta}
	\begin{split}
          R_\sigma \colon &\mathcal{PN}(\mathcal{A}, B) \to 	\mathcal{N}(\sigma; \mathcal{A}, B) \\
          &\Bw \mapsto \Phi(\cdot,\Bw),
	\end{split}
	\end{align}
	where $\Phi(\cdot,\Bw)$ is the neural network with weights and biases given by $\Bw=(\BW^{(\ell)}, \Bb^{(\ell)})_{\ell = 0}^L$
        as in Definition \ref{def:nn}.
\end{definition}

Throughout, we will identify $\mathcal{PN}(\CA,B)$ with the cube 
  $[-B,B]^{n_\CA}$, where
  $n_{\mathcal{A}} \coloneqq \sum_{\ell = 0}^{L}d_{\ell+1} (d_{\ell} + 1)$. Now we can %
introduce the loss landscape of a neural network architecture.

\begin{definition}\label{def:lossLandscape}
Let $\mathcal{A} = (d_0, d_1, \dots, d_{L+1}) \in \N^{L+2}$, let $\sigma\colon \R \to \R$.
Let $m \in \N$, and $S = (\Bx_i, \By_i)_{i=1}^m \in (\R^{d_0} \times \R^{d_{L+1}})^m$ be a sample and let $\mathcal{L}$ be a loss function.
Then, the {\bf loss landscape} is the graph of the function $\Lambda_{\mathcal{A}, \sigma, S, \mathcal{L}}$ defined as
  \begin{align*}
    \begin{split}
	 \Lambda_{\mathcal{A}, \sigma, S, \mathcal{L}}\, \colon \, \mathcal{PN}(\mathcal{A}; \infty) &\to \R \\
	 \theta &\mapsto \widehat{\mathcal{R}}_S(R_\sigma(\theta)).
	\end{split}    
    \end{align*}
 with $\widehat{\mathcal{R}}_S$ in \eqref{eq:empiricalRiskDef0} and $R_\sigma$ in \eqref{eq:Rsigmatheta}.
\end{definition}

\begin{figure}
	\centering
	\includegraphics[width = 0.85\textwidth]{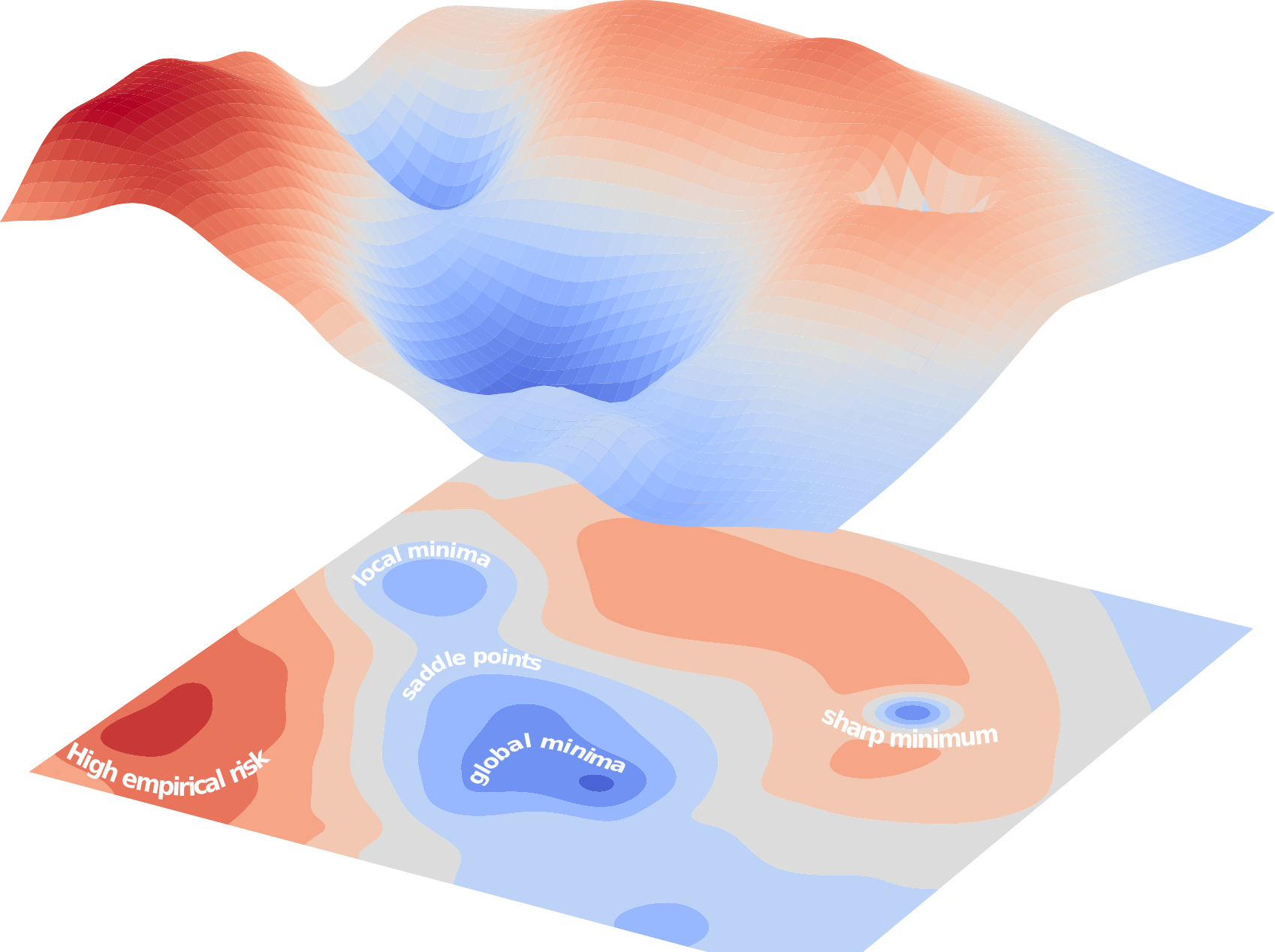}
	\caption{Two-dimensional section of a loss landscape.
The loss landscape shows a spurious valley with local minima, global minima, as well as a region where saddle points appear.
Moreover, a sharp minimum is shown.
}\label{fig:LossLandscapeExample}
\end{figure}

Identifying $\mathcal{PN}(\mathcal{A}, \infty)$ %
with $\R^{n_{\mathcal{A}}}$, we can consider $\Lambda_{\mathcal{A}, \sigma, S, \mathcal{L}}$ as a map on $\R^{n_{\mathcal{A}}}$ and the loss landscape is a subset of $\R^{n_{\mathcal{A}}} \times \R$.
The loss landscape is a high-dimensional surface, with hills and valleys.
For visualization a two-dimensional section of a loss landscape is shown in Figure \ref{fig:LossLandscapeExample}. 

Questions of interest regarding the loss landscape include for example: How likely is it that we find local instead of global minima? Are these local minima typically sharp, having small volume, or are they part of large flat valleys that are difficult to escape? How bad is it to end up in a local minimum? Are most local minima as deep as the global minimum, or can they be significantly higher? How rough is the surface generally, and how do these characteristics depend on the network architecture? While providing complete answers to these questions is hard in general, in the rest of this chapter we give some intuition and mathematical insights for specific cases.

\section{Visualization of loss landscapes}
Visualizing loss landscapes can provide valuable insights into the effects of neural network depth, width, and activation functions.
However, we can only %
visualize an at most two-dimensional surface embedded into three-dimensional space, whereas the loss landscape is a very high-dimensional object (unless
the neural networks have only very few weights and biases).

\begin{figure}[htb]
	\centering
	\includegraphics[width = 0.4\textwidth]{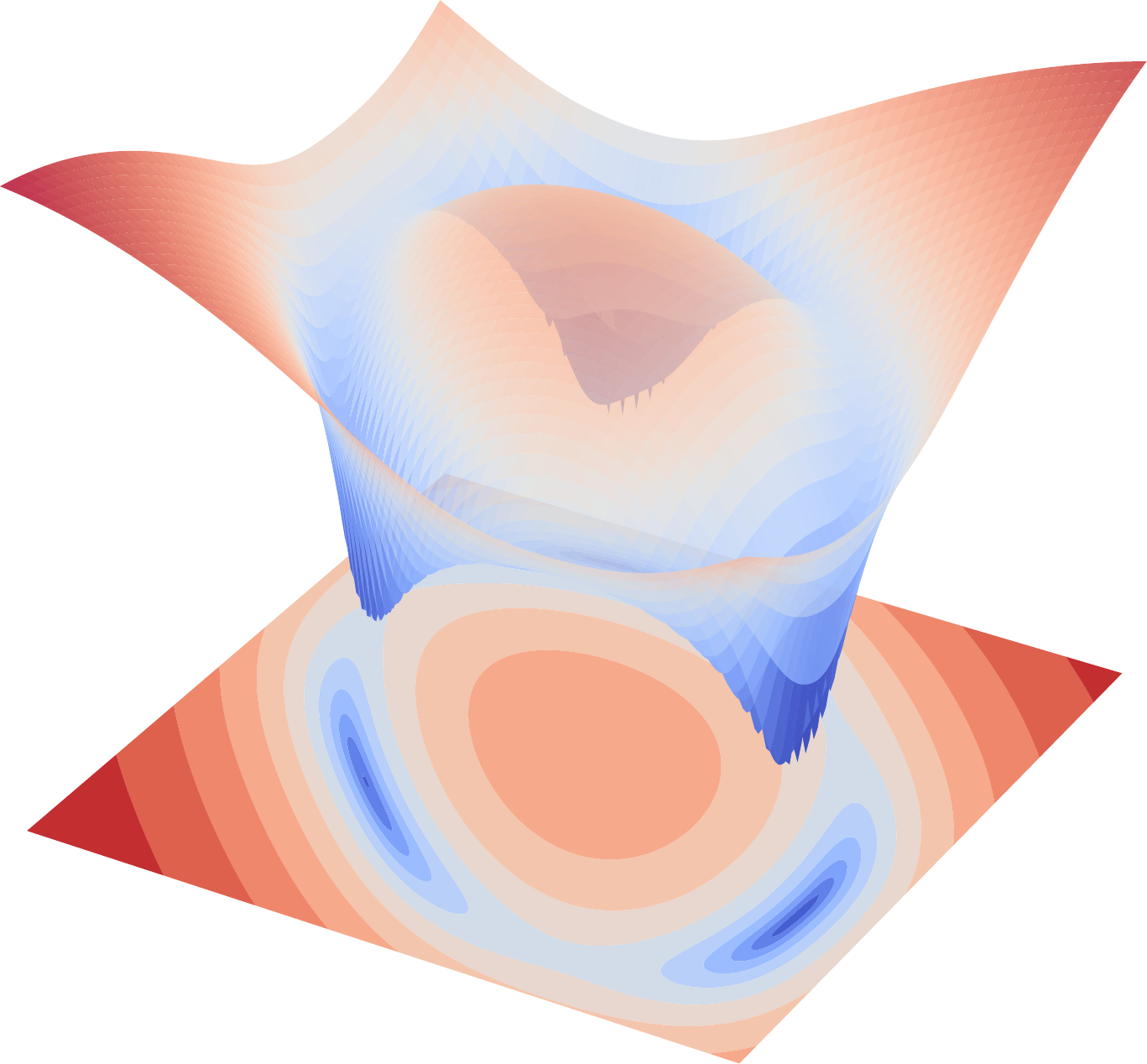} 	\includegraphics[width = 0.4\textwidth]{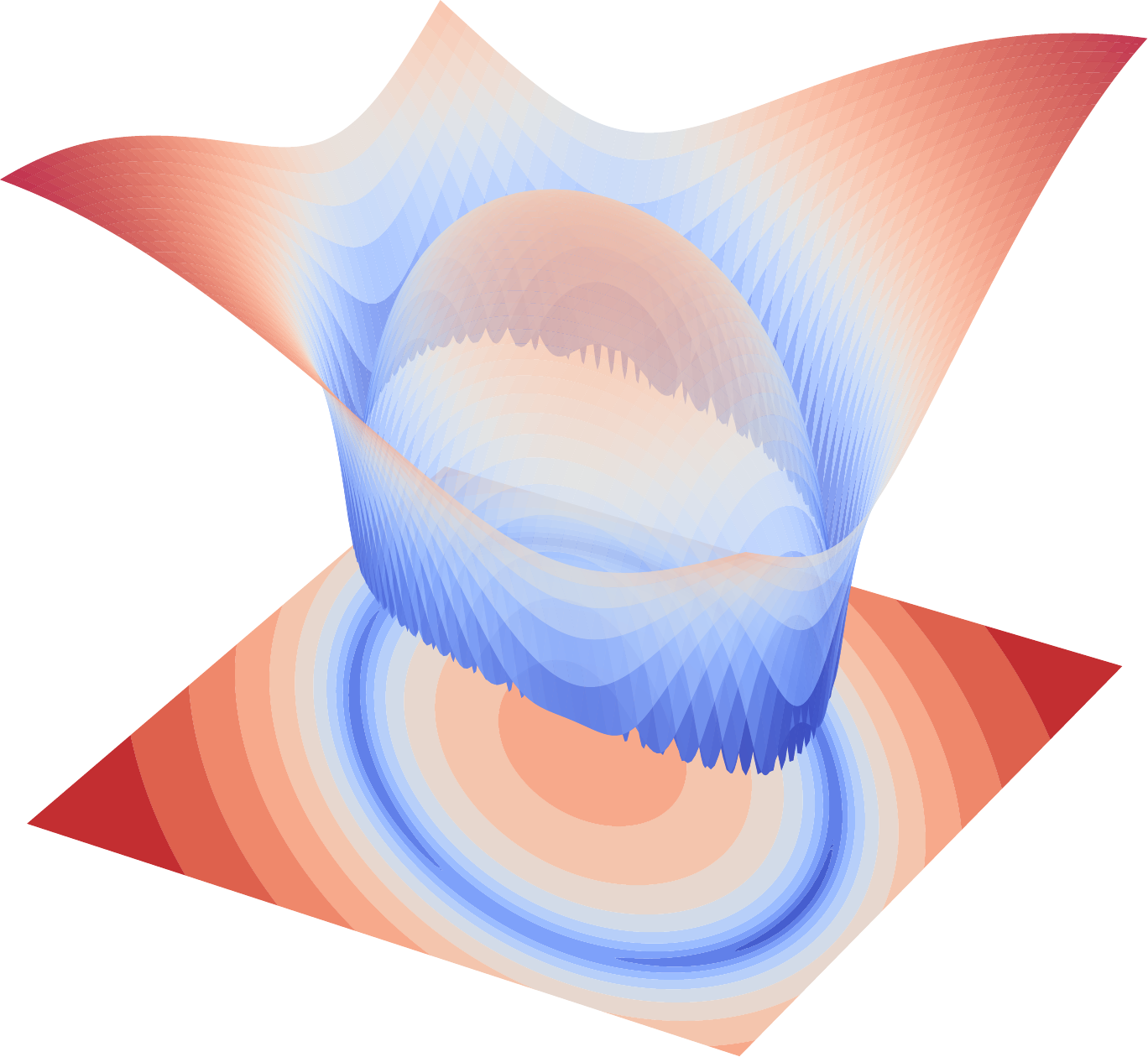}
	
	\includegraphics[width = 0.4\textwidth]{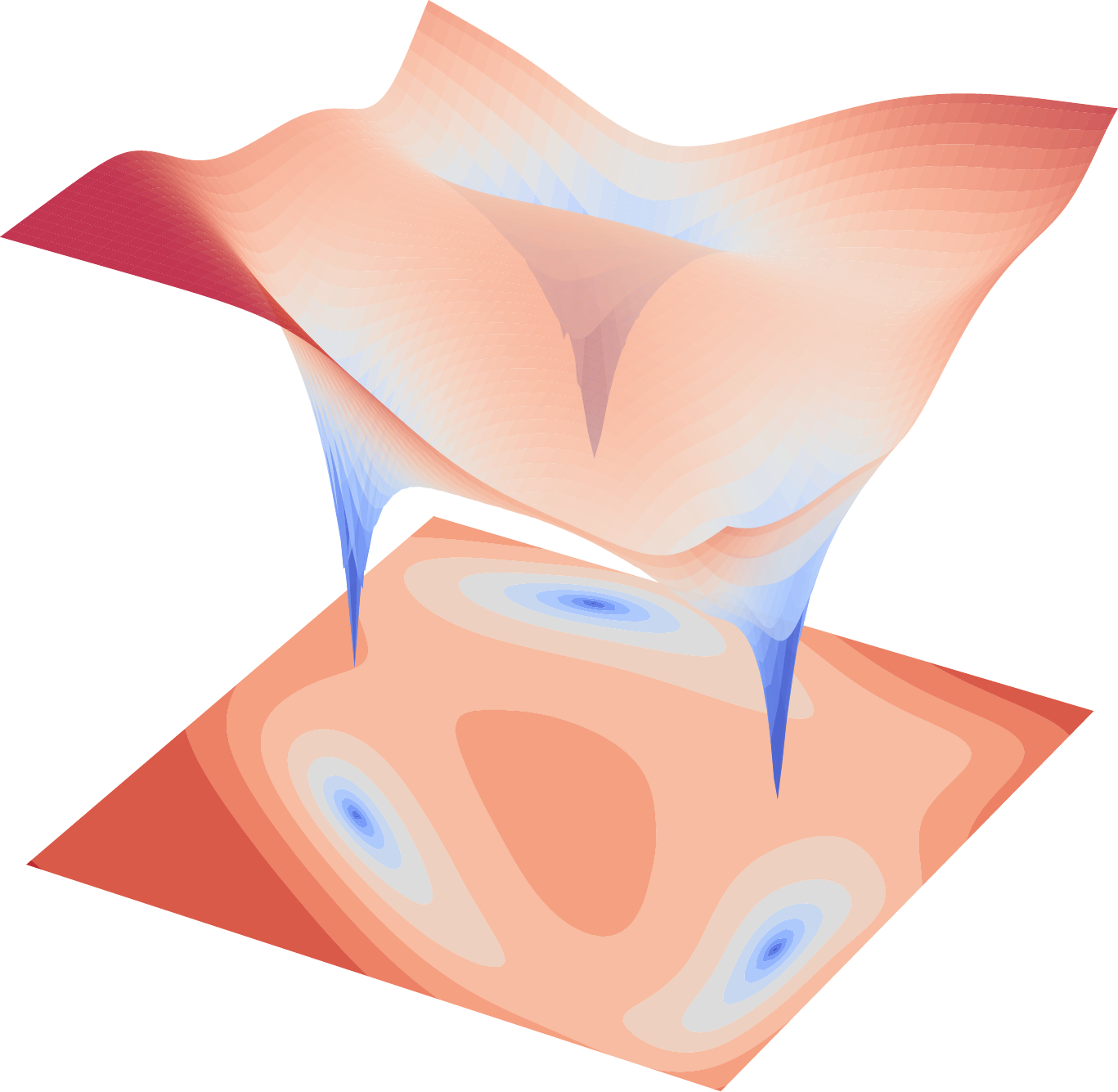} 	\includegraphics[width = 0.4\textwidth]{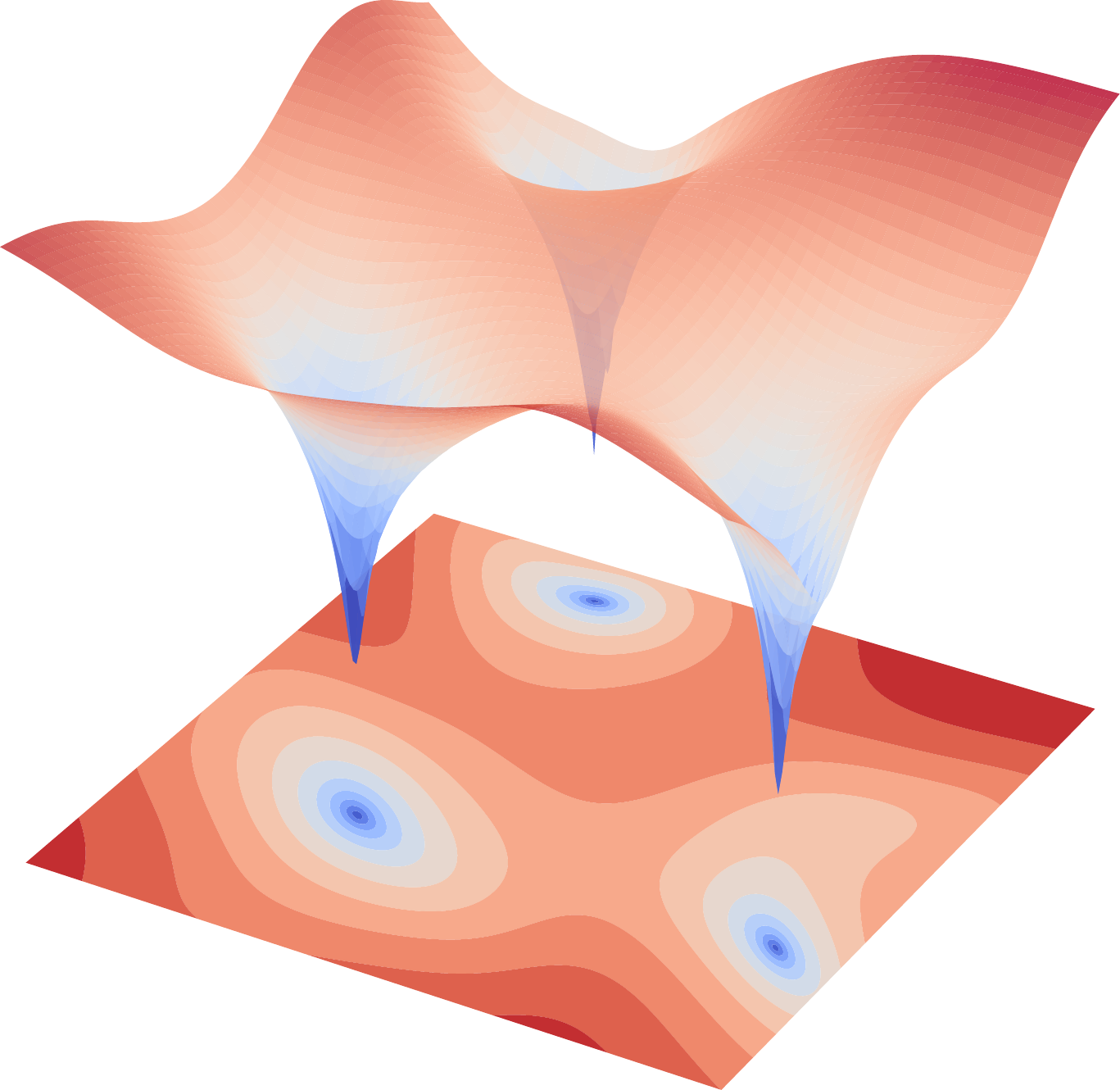}
	
	\includegraphics[width = 0.4\textwidth]{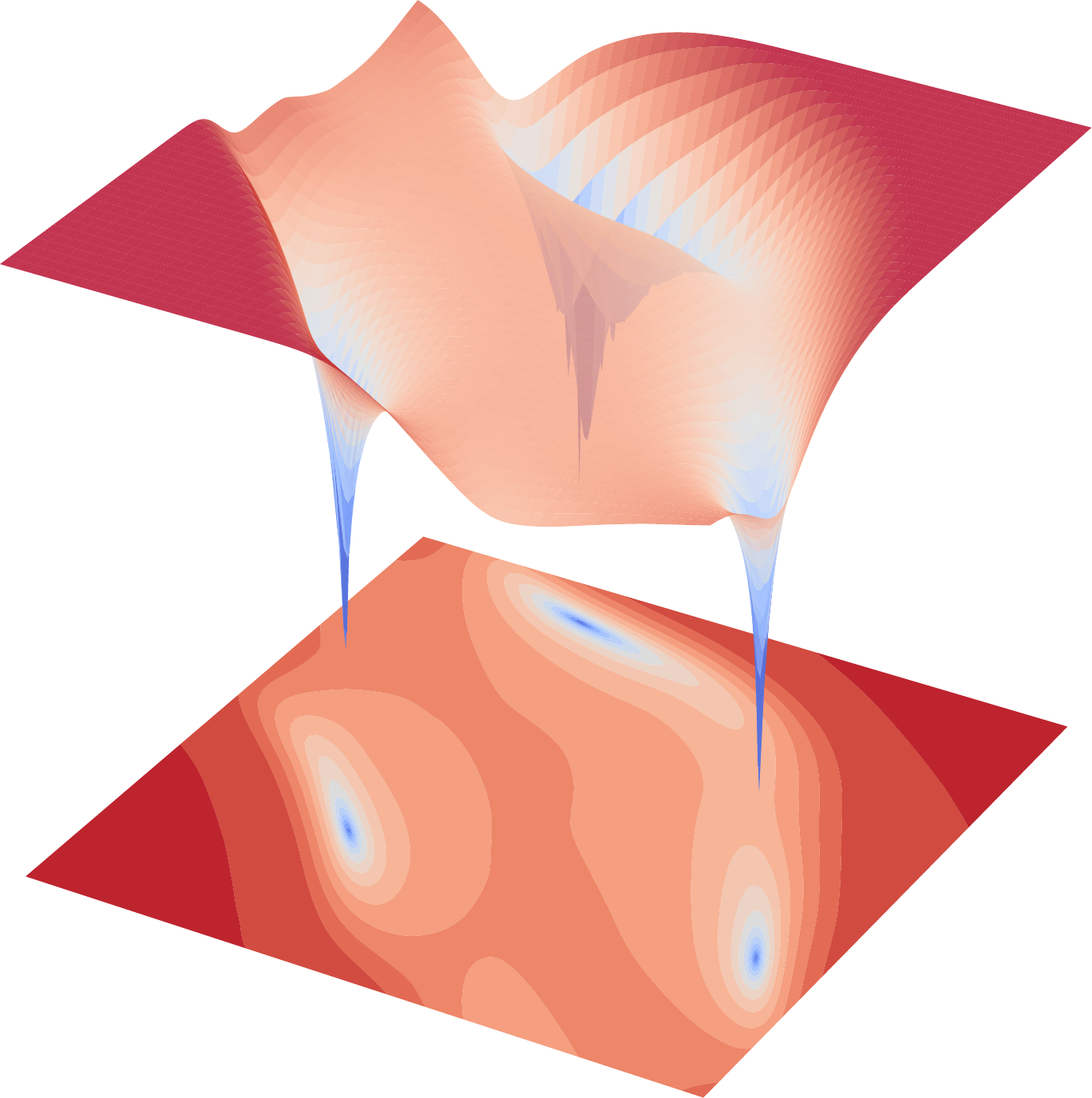} 	\includegraphics[width = 0.4\textwidth]{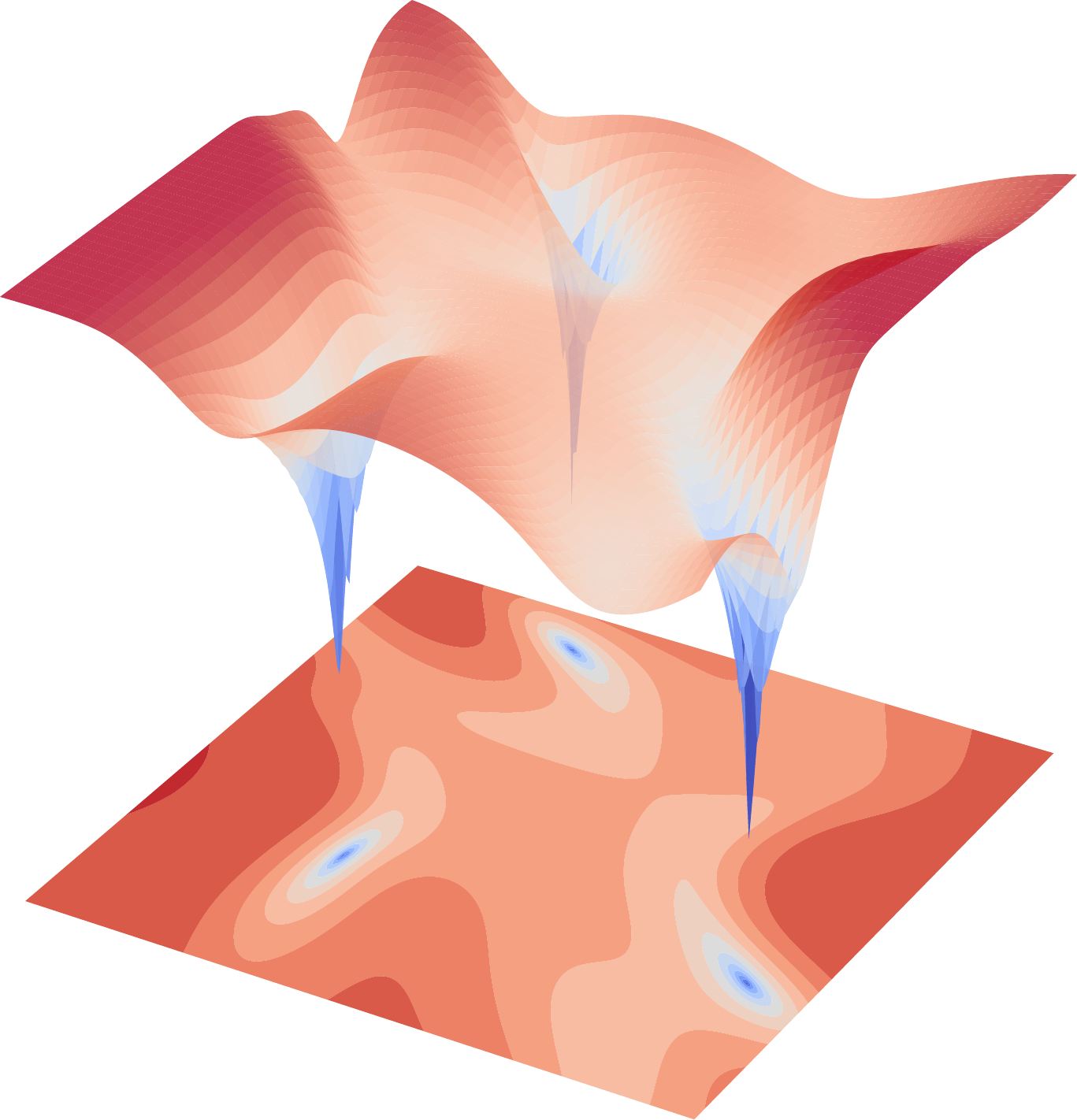}
	
	\caption{A collection of loss landscapes.
		In the left column are neural networks with ReLU activation function, the right column shows loss landscapes of neural networks with the hyperbolic tangent activation function.
		All neural networks have five dimensional input, and one dimensional output.
		Moreover, from top to bottom the hidden layers have %
                widths 1000, 20, 10, and the number of hidden layers are 1, 4, 7.}
	\label{fig:losslandscapes}
\end{figure}

To make the loss landscape accessible, we need to reduce its dimensionality.
This can be achieved by evaluating the function $\Lambda_{\mathcal{A}, \sigma, S, \mathcal{L}}$ on a two-dimensional subspace of  $\mathcal{PN}(\mathcal{A}, \infty)$. 
Specifically, we choose three-parameters $\mu$, $\theta_1$, $\theta_2$ and %
examine the function
\begin{align}\label{eq:two-dimensional-loss-landscape}
	\R^2 \ni (\alpha_1,\alpha_2) \mapsto \Lambda_{\mathcal{A}, \sigma, S, \mathcal{L}}(\mu  + \alpha_1 \theta_1 + \alpha_2 \theta_2).
\end{align}
There are various natural choices for $\mu$, $\theta_1$, $\theta_2$:
\begin{itemize}
\item \textit{Random directions:} This was, for example used in \cite{goodfellow2014qualitatively, im2016empirical}.
Here $\theta_1, \theta_2$ are %
chosen randomly, %
while $\mu$ is either a minimum of $\Lambda_{\mathcal{A}, \sigma, S, \mathcal{L}}$ or also chosen randomly.
This simple approach can offer a quick insight into how rough the surface can be.
However, as was pointed out in \cite{li2018visualizing}, random directions will very likely be orthogonal to the trajectory of the optimization procedure.
Hence, they will likely miss the most relevant features.
\item \textit{Principal components of learning trajectory:} To address the shortcomings of %
  random directions,
  another possibility is to determine $\mu$, $\theta_1$, $\theta_2$, which
  best capture some given learning trajectory; For example,
if $\theta^{(1)}, \theta^{(2)}, \dots, \theta^{(N)}$ are the parameters resulting from the training by SGD, we %
may determine $\mu$, $\theta_1$, $\theta_2$ such that the hyperplane $\set{\mu +  \alpha_1 \theta_1 + \alpha_2 \theta_2}{\alpha_1,\alpha_2 \in \R}$ minimizes the mean squared distance to the $\theta^{(j)}$ for $j \in \{1, \dots, N\}$.
This is the approach of \cite{li2018visualizing}, and can be achieved by a principal component analysis.
\item \textit{Based on critical points:} For a more global perspective, %
  $\mu$, $\theta_1$, $\theta_2$ can be chosen to ensure the observation of multiple critical points.
One way to achieve this is by running the optimization procedure three times with final parameters $\theta^{(1)}$, $\theta^{(2)}$, $\theta^{(3)}$.
If the procedures have converged, then each of these parameters is close to a critical point of $\Lambda_{\mathcal{A}, \sigma, S, \mathcal{L}}$.
We can now set $\mu = \theta^{(1)}$, $\theta_1 = \theta^{(2)}- \mu$,  $\theta_2 = \theta^{(3)}- \mu$.
This %
then guarantees
that \eqref{eq:two-dimensional-loss-landscape} passes through or at least comes very close to three critical points (at $(\alpha_1,\alpha_2) = (0,0), (0,1), (1,0)$).
We present six visualizations of this form in Figure \ref{fig:losslandscapes}.
\end{itemize}

Figure \ref{fig:losslandscapes} gives some interesting insight into the effect of depth and width on the shape of the loss landscape.
For very wide and shallow neural networks, we have the widest minima, which, in the case of the tanh activation function also seem to belong to the same valley.
With increasing depth and smaller width the minima get steeper and more disconnected. 

\section{Spurious valleys}
From the perspective of optimization,
the ideal loss landscape has one global minimum in the center of a large valley, so that gradient descent converges towards the minimum irrespective of the chosen initialization.

This situation is not realistic for deep neural networks.
Indeed, %
for a simple shallow neural network
\[
	\R^d \ni \Bx \mapsto \Phi(\Bx) = \BW^{(1)} \sigma( \BW^{(0)} \Bx + \Bb^{(0)}) + \Bb^{(1)},
\]
it is clear that for every permutation matrix $\BP$
\[
	\Phi(\Bx) =  \BW^{(1)} \BP^T \sigma( \BP \BW^{(0)} \Bx + \BP \Bb^{(0)}) + \Bb^{(1)}\qquad \text{for all } \Bx \in \R^d.
\]
Hence, %
in general there exist
multiple parameterizations %
realizing the same output function. Moreover,
if at least one global minimum with non-permutation-invariant weights exists, then there are more than one global minima of the loss landscape. 

This is not problematic; in fact, having many global minima is
beneficial. The larger issue is the existence of non-global
minima. Following \cite{venturi2019spurious}, we start by generalizing
the notion of non-global minima to \emph{spurious valleys}.

\begin{definition}\label{def:spuriousValley}
Let $\mathcal{A} = (d_0, d_1, \dots, d_{L+1}) \in \N^{L+2}$ %
and $\sigma\colon \R \to \R$.
Let $m \in \N$, and $S = (\Bx_i, \By_i)_{i=1}^m \in (\R^{d_0} \times \R^{d_{L+1}})^m$ be a sample and let $\mathcal{L}$ be a loss function.
For $c \in \R$, we define the sub-level set of $\Lambda_{\mathcal{A}, \sigma, S, \mathcal{L}}$ as 
\[
	\Omega_\Lambda(c) \coloneqq \set{\theta \in \mathcal{PN}(\mathcal{A}, \infty)}{\Lambda_{\mathcal{A}, \sigma, S, \mathcal{L}}(\theta) \leq c}.
\]
A path-connected component of $\Omega_\Lambda(c)$, which does not contain a global minimum of $\Lambda_{\mathcal{A}, \sigma, S, \mathcal{L}}$ is called a {\bf spurious valley}.
\end{definition}

The next proposition shows that spurious local minima do not exist for shallow  overparameterized neural networks, i.e., %
for neural networks that have at least as many parameters in the hidden layer as there are training samples.

\begin{proposition}
  Let $\mathcal{A} = (d_0,d_1, 1) \in \N^{3}$ and let $S = (\Bx_i, y_i)_{i=1}^m \in (\R^{d_0} \times \R)^m$ be a sample such that $m \leq d_1$.
  Furthermore, let $\sigma \in \mathcal{M}$ be not a polynomial,
    and let $\mathcal{L}$ be a convex loss function.
	Further assume that $\Lambda_{\mathcal{A}, \sigma, S, \mathcal{L}}$ has at least one global minimum. 
	Then, $\Lambda_{\mathcal{A}, \sigma, S, \mathcal{L}}$, has no spurious valleys.
\end{proposition}

\begin{proof}
	Let $\theta_a, \theta_b \in \mathcal{PN}(\mathcal{A}, \infty)$ with $\Lambda_{\mathcal{A}, \sigma, S, \mathcal{L}}(\theta_a) > \Lambda_{\mathcal{A}, \sigma, S, \mathcal{L}}(\theta_b)$. Then we will show below that there is another parameter $\theta_c$ such that 
	\begin{itemize}
		\item $\Lambda_{\mathcal{A}, \sigma, S, \mathcal{L}}(\theta_b) \geq \Lambda_{\mathcal{A}, \sigma, S, \mathcal{L}}(\theta_c)$
		\item there is a continuous path $\alpha:[0,1] \to \mathcal{PN}(\mathcal{A}, \infty)$ such that 
		$\alpha(0) = \theta_a$, $\alpha(1) = \theta_c$, and 
		$\Lambda_{\mathcal{A}, \sigma, S, \mathcal{L}}(\alpha)$ is monotonically decreasing. 
	\end{itemize}
	By Exercise \ref{ex:pathsAndValleys}, the construction above rules out the existence of spurious valleys by choosing $\theta_a$ an element of a spurious valley and $\theta_b$ a global minimum.
	
	Next, we present the construction: Let us denote 
	\[
		\theta_o = \left(\left(\BW^{(\ell)}_o, \Bb^{(\ell)}_o\right)_{\ell = 0}^1\right) \qquad \text{ for } o \in \{a,b,c\}.
	\]
	Moreover, for $j = 1, \dots, d_1$, we introduce $\Bv_o^j \in \R^{m}$ defined as
	\begin{align*}
		(\Bv_o^j)_i = \left(\sigma\left(\BW^{(0)}_o \Bx_i +  \Bb^{(0)}_o\right)\right)_j \qquad \text{ for } i = 1, \dots, m.
	\end{align*}
	Notice that, if we set $\BV_o = ((\Bv_o^j)^\top)_{j=1}^{d_1}$, then 
	\begin{align}
		\label{eq:matrix-notion-of-NNs}
		\BW^{(1)}_o \BV_o = \left(R_\sigma(\theta_o)(\Bx_i) -  \Bb^{(1)}_o\right)_{i=1}^m,
	\end{align}
	where the right-hand side is considered a row-vector.
	
	We will now distinguish between two cases. %
        For the first the result is trivial and %
        the second %
        can be transformed into the first one.
	
	\textbf{Case 1: } Assume that $\BV_a$ has rank $m$.
In this case, it is obvious from \eqref{eq:matrix-notion-of-NNs}, that there exists $\widetilde{\BW}$ such that 
	\[
	\widetilde{\BW} \BV_a = \left(R_\sigma(\theta_b)(\Bx_i)- \Bb^{(1)}_a\right)_{i=1}^m.
	\]
	We can thus set $\alpha(t) = ((\BW^{(0)}_a, \Bb^{(0)}_a),((1-t)\BW^{(1)}_a + t \widetilde{\BW}, \Bb^{(1)}_a))$. 
	
	Note that by construction $\alpha(0) = \theta_a$ and $\Lambda_{\mathcal{A}, \sigma, S, \mathcal{L}}(\alpha(1)) = \Lambda_{\mathcal{A}, \sigma, S, \mathcal{L}}(\theta_b)$.
        Moreover, %
        $t\mapsto (R_\sigma(\alpha(t))(\Bx_i))_{i=1}^m$ describes
a straight path in $\R^m$ and hence, by the convexity of $\mathcal{L}$ it is clear that $t\mapsto \Lambda_{\mathcal{A}, \sigma, S, \mathcal{L}}(\alpha(t))$ has a minimum $t^*$ on $[0,1]$ with $ \Lambda_{\mathcal{A}, \sigma, S, \mathcal{L}}(\alpha(t^*)) \leq \Lambda_{\mathcal{A}, \sigma, S, \mathcal{L}}(\theta_b)$. Moreover, $t\mapsto \Lambda_{\mathcal{A}, \sigma, S, \mathcal{L}}(\alpha(t))$ is monotonically decreasing on $[0, t^*]$. Setting $\theta_c = \alpha(t^*)$ completes this case.
	
	\textbf{Case 2: } Assume that $V_a$ has rank less than $m$.
In this case, we show that we find a continuous path from $\theta_a$ to another neural network parameter with higher rank.
The path will be such that $\Lambda_{\mathcal{A}, \sigma, S, \mathcal{L}}$ is monotonically decreasing.
	
	Under the assumptions, we have that one $\Bv_a^j$ can be written as a linear combination of the remaining $\Bv_a^i$, $i \neq j$. %
Without loss of generality, we assume $j = 1$.
Then, there exist $(\alpha_i)_{i=2}^m$ such that 
	\begin{align}
		\label{eq:replacevs}
			\Bv_a^1 = \sum_{i=2}^m \alpha_i \Bv_a^i.
	\end{align}
	Next, we observe that there exists $\Bv^* \in \R^m$ which is linearly independent from all $(\Bv_a^j)_{i=1}^m$ and can be written as $(\Bv^*)_i = \sigma((\Bw^*)^\top \Bx_i + b^*)$ for some $\Bw^* \in \R^{d_0}, b^*\in \R$.
Indeed, if we assume that such $\Bv^*$ does not exist, then for all $\Bw \in \R^{d_0}, b\in \R$ the vector $(\sigma(\Bw^\top \Bx_i + b))_{i=1}^m$ belongs to the same $m-1$ dimensional subspace. It would follow that $\mathrm{span}  \set{(\sigma(\Bw^\top \Bx_i + b))_{i=1}^m}{\Bw \in \R^{d_0}, b\in \R}$ is an $m-1$ dimensional subspace of $\R^m$ which yields a contradiction to Theorem \ref{thm:universalInterpolationThm}. 
	
	Now, we define two paths: First,
	\[
		\alpha_1(t) =  ((\BW^{(0)}_a, \Bb^{(0)}_a),(\BW^{(1)}_a(t), \Bb^{(1)}_a)), \qquad \text{ for } t \in [0,1/2]
	\] 
	where
        \begin{equation*}
          (\BW^{(1)}_a(t))_1 = (1-2t)(\BW^{(1)}_a)_1\qquad\text{and}\qquad(\BW^{(1)}_a(t))_i = (\BW^{(1)}_a)_i + 2 t \alpha_i (\BW^{(1)}_a)_1
        \end{equation*}
          for $i = 2,\dots, d_1$, for $t \in [0,1/2]$. 
	Second, 
	\[
		\alpha_2(t) =  ((\BW^{(0)}_a(t), \Bb^{(0)}_a(t)),(\BW^{(1)}_a(1/2), \Bb^{(1)}_a)), \text{ for } t \in (1/2,1],
	\] 
	where
        \begin{equation*}
          (\BW^{(0)}_a(t))_1 = 2(t-1/2) (\BW^{(0)}_a)_1  + (2t-1) \Bw^*
          \qquad\text{and}\qquad
          (\BW^{(0)}_a(t))_i = (\BW^{(0)}_a)_i
        \end{equation*}
          for $i = 2, \dots, d_1$, $(\Bb^{(0)}_a(t))_1 = 2(t-1/2)(\Bb^{(0)}_a)_1 + (2t-1) b^*$, and $(\Bb^{(0)}_a(t))_i = (\Bb^{(0)}_a)_i$ for $i = 2, \dots, d_1$. 
	
         It is clear by \eqref{eq:replacevs} that $(R_\sigma(\alpha_1)(\Bx_i))_{i=1}^m$ is constant.
	 Moreover, $R_\sigma(\alpha_2)(\Bx)$ is constant for all $\Bx \in \R^{d_0}$.
         In addition, by construction for
	\[
		\bar{\Bv}^j \coloneqq \left(\left(\sigma\left(\BW^{(0)}_a(1) \Bx_i +  \Bb^{(0)}_a(1)\right)\right)_j\right)_{i=1}^m
	\]
	it holds that $((\bar{\Bv}^j)^\top)_{j=1}^{d_1}$ has rank larger than that of $\BV_a$.
Concatenating $\alpha_1$ and $\alpha_2$ now yields a continuous path from $\theta_a$ to another neural network parameter with higher associated rank such that $\Lambda_{\mathcal{A}, \sigma, S, \mathcal{L}}$ is monotonically decreasing along the path.
	Iterating this construction, we can find a path to a neural network parameter where the associated matrix has full rank.
This reduces the problem to Case 1.	
\end{proof}

\section{Saddle points}\label{sec:SaddlePoints}

Saddle points are critical points of the loss landscape at which the loss decreases in one direction. 
In this sense, saddle points are not as problematic as local minima or spurious valleys if the updates in the learning iteration have some stochasticity. 
Eventually, a random step in the right direction could be taken and the saddle point can be escaped. 

If most of the critical points are saddle points, then, even though the loss landscape is challenging for optimization, one still has a good chance of eventually reaching the global minimum. 
Saddle points of the loss landscape were studied in \cite{dauphin2014identifying, pennington2017geometry} and we will review some of the findings in a simplified way below. 
The main observation in \cite{pennington2017geometry} is that, under some quite strong assumptions, it holds that \emph{critical points in the loss landscape associated to a large loss are typically saddle points, whereas those associated to small loss correspond to minima}.
This situation is encouraging for the prospects of optimization in deep learning, since, even if we get stuck in a local minimum, it will very likely be such that the loss is close to optimal.

The results of \cite{pennington2017geometry} use random matrix theory, which we do not %
recall here.
Moreover, it is hard to gauge if the assumptions made are satisfied for a specific problem. 
Nonetheless, we %
recall the main idea, which %
provides some intuition to support the above claim.

Let $\mathcal{A} = (d_0, d_1, 1) \in \N^3$.
Then, for a neural network parameter $\theta \in \mathcal{PN}(\mathcal{A}, \infty)$ and activation function $\sigma$, we set $\Phi_\theta \coloneqq R_\sigma(\theta)$ and define for a sample $S = (\Bx_i,y_i)_{i=1}^m$ the errors
\begin{align*}
	e_i = \Phi_\theta(\Bx_i) - y_i\qquad \text{for } i = 1, \dots, m.
\end{align*}
If we use the square loss, then %
\begin{align}\label{eq:hatRsPhitheta}
  \widehat{\mathcal{R}_S}(\Phi_\theta) = \frac{1}{m} \sum_{i=1}^m e_i^2.
\end{align}
Next, we study the Hessian of $\widehat{\mathcal{R}}_S(\Phi_\theta)$. 

\begin{proposition}\label{prop:separationOfHessian}
	Let $\mathcal{A} = (d_0, d_1, 1)$ and $\sigma : \R \to \R$.
Then, for every $\theta \in 	\mathcal{PN}(\mathcal{A}, \infty)$ where $\widehat{\mathcal{R}}_S(\Phi_\theta)$ in \eqref{eq:hatRsPhitheta} is twice continuously differentiable with respect to the weights, it holds that 
	\[
		\BH(\theta) = \BH_0(\theta) + \BH_1(\theta),
	\]
	where $\BH(\theta)$ is the Hessian of $\widehat{\mathcal{R}}_S(\Phi_\theta)$ at $\theta$, $\BH_0(\theta)$ is a positive semi-definite matrix which is independent from $(y_i)_{i=1}^m$,
	and $\BH_1(\theta)$ is a symmetric matrix that for fixed $\theta$ and $(\Bx_i)_{i=1}^m$ depends linearly on $(e_i)_{i=1}^m$.
\end{proposition}

\begin{proof}
Using the identification introduced after Definition \ref{def:lossLandscape}, we can consider $\theta$ a vector in $\R^{n_\mathcal{A}}$.
For $k = 1, \dots, n_\mathcal{A}$, we have that
\begin{align*}
	\frac{\partial \widehat{\mathcal{R}}_S(\Phi_\theta)}{\partial \theta_k} = \frac{2}{m}\sum_{i=1}^m e_i \frac{\partial \Phi_\theta(\Bx_i)}{\partial \theta_k} .
\end{align*}
Therefore, for $j  = 1, \dots, n_\mathcal{A}$, we have, by the Leibniz rule, that 
\begin{align}
	\frac{\partial^2 \widehat{\mathcal{R}}_S(\Phi_\theta)}{\partial \theta_j\partial \theta_k} &= \frac{2}{m}\sum_{i=1}^m \left(\frac{\partial \Phi_\theta(\Bx_i)}{\partial \theta_j}  \frac{\partial  \Phi_\theta(\Bx_i)}{\partial \theta_k} \right) + \frac{2}{m} \left(\sum_{i=1}^m e_i \frac{\partial^2 \Phi_\theta(\Bx_i)}{\partial \theta_j \partial \theta_k} \right) \label{eq:defOfH0andH1}\\
	&\eqqcolon \BH_0(\theta) + \BH_1(\theta).
	\nonumber
\end{align}
It remains to show that $\BH_0(\theta)$ and $\BH_1(\theta)$ have the asserted properties.
Note that, setting 
\[
	J_{i, \theta} = \begin{pmatrix}
	\frac{\partial \Phi_\theta(\Bx_i)}{\partial \theta_1}  \\
	\vdots \\
	\frac{\partial  \Phi_\theta(\Bx_i)}{\partial \theta_{n_\mathcal{A}}} 
	\end{pmatrix}\in\R^{n_{\mathcal{A}}},
\]
we have that $\BH_0(\theta) =\frac{2}{m}\sum_{i=1}^m J_{i, \theta} J_{i, \theta}^\top$ and hence $\BH_0(\theta)$ is a sum of positive semi-definite matrices, which shows that $\BH_0(\theta)$ is positive semi-definite.

The symmetry of $\BH_1(\theta)$ follows directly from the symmetry of second derivatives which holds since we assumed twice continuous differentiability at $\theta$.
The linearity of $\BH_1(\theta)$ in $(e_i)_{i=1}^m$ is clear from \eqref{eq:defOfH0andH1}.
\end{proof}

How does Proposition \ref{prop:separationOfHessian} imply the claimed relationship between the size of the loss and the prevalence of saddle points? 

Let $\theta$ correspond to a critical point.
If $\BH(\theta)$ has at least one negative eigenvalue, then $\theta$ cannot be a minimum, but instead must be either a saddle point or a maximum.
While we do not know anything about $\BH_1(\theta)$ other than that it is symmetric, it is not unreasonable to assume that it has a negative eigenvalue especially if $n_{\mathcal{A}}$ is very large.
With this consideration, let us consider the following model:

Fix a parameter $\theta$.
Let $S^0 = (\Bx_i, y_i^0)_{i=1}^m$ be a sample and $(e_i^0)_{i=1}^m$ be the associated errors.
Further let $\BH^0(\theta), \BH_0^0(\theta), \BH_1^0(\theta)$ be the matrices according to Proposition \ref{prop:separationOfHessian}.

Further let for $\lambda >0$, $S^\lambda = (\Bx_i, y_i^\lambda)_{i=1}^m$ be such that the associated errors are $(e_i)_{i=1}^m = \lambda (e_i^0)_{i=1}^m$.
The Hessian of $\widehat{\mathcal{R}}_{S^\lambda}(\Phi_\theta)$ at $\theta$ is then $\BH^\lambda(\theta)$ satisfying
\[
  \BH^\lambda(\theta) = \BH^0_0(\theta) + \lambda %
  \BH_1^0(\theta).
\]
Hence, if $\lambda$ is large, then $\BH^\lambda(\theta)$ is  perturbation of an amplified version of %
 $\BH_1^0(\theta)$.
Clearly, if $\Bv$ is an eigenvector of $\BH_1(\theta)$ with negative eigenvalue $-\mu$, then
\[
	\Bv^\top \BH^\lambda(\theta)\Bv \leq (\|\BH^0_0(\theta)\| - \lambda \mu) \|\Bv\|^2,
\]
which we can expect to be negative for large $\lambda$.
Thus, $\BH^\lambda(\theta)$ has a negative eigenvalue for large $\lambda$.

On the other hand, if $\lambda$ is small, then 	$\BH^\lambda(\theta)$ is merely a perturbation of $\BH^0_0(\theta)$ and we can expect its spectrum to resemble that of $\BH^0_0$ more and more.

What we see is that, the same parameter, is more likely to be a saddle point for a sample that produces a high empirical risk than for a sample with small risk. 
Note that, since $\BH^0_0(\theta)$ was only shown to be \emph{semi}-definite the argument above does not rule out saddle points even for very small $\lambda$.
But it does show that for small $\lambda$, every negative eigenvalue would be very small. 

A more refined analysis where we compare different parameters but for the same sample and quantify the likelihood of local minima versus saddle points requires the introduction of a probability distribution on the weights.
We refer to \cite{pennington2017geometry} for the details.

\section*{Bibliography and further reading}

The results on visualization of the loss landscape are inspired by \cite{li2018visualizing,goodfellow2014qualitatively,im2016empirical}. 
Results on the non-existence of spurious valleys can be found in \cite{venturi2019spurious} with similar results in \cite{Nguyen_2018}.
In \cite{choromanska2015loss} the loss landscape was studied by linking it to so-called spin-glass models. There it was found that under strong assumptions 
critical points associated to lower losses are more likely to be minima than saddle points. In \cite{pennington2017geometry}, random matrix theory is used to provide similar results, that go beyond those established in Section \ref{sec:SaddlePoints}. On the topic of saddle points, \cite{dauphin2014identifying} identifies the existence of saddle points as more problematic than that of local minima, and an alternative saddle-point aware optimization algorithm is introduced.

Two essential topics associated to the loss landscape that have not been discussed in this chapter are mode connectivity and the sharpness of minima. Mode connectivity, roughly speaking describes the phenomenon, that local minima found by SGD over deep neural networks are often connected by simple curves of equally low loss \cite{garipov2018loss, draxler2018essentially}. Moreover, the sharpness of minima has been analyzed and linked to generalization capabilities of neural networks, with the idea being that wide neural networks are easier to find and also yield robust neural networks \cite{hochreiter1997flat, chaudhari2019entropy, jiang2019fantastic}. However, this does not appear to prevent sharp minima from generalizing well \cite{dinh2017sharp}. 

\newpage
\section*{Exercises}

\begin{exercise}
	In view of Definition \ref{def:spuriousValley}, show that a local minimum of a differentiable function is contained in a spurious valley. 
\end{exercise}

\begin{exercise}\label{ex:pathsAndValleys}
	Show that if there exists a continuous path $\alpha$ between a parameter $\theta_1$ and a global minimum $\theta_2$ such that $\Lambda_{\mathcal{A}, \sigma, S, \mathcal{L}}(\alpha)$ is monotonically decreasing, then $\theta_1$ cannot be an element of a spurious valley.	
\end{exercise}

\begin{exercise}
  Find an example of a spurious valley for a simple architecture.
  
  \emph{Hint}: Use a single neuron ReLU neural network and observe that, for two networks one with positive and one with negative slope, every continuous path in parameter space that connects the two has to pass through a parameter corresponding to a constant function.
\end{exercise}

%% file: ShapeOfNNSpaces.tex
\chapter{Shape of neural network spaces}\label{chap:shape}
As we have seen in the previous chapter, the loss landscape of neural networks %
can be quite intricate and is typically not convex.
In some sense, the reason for this is that we take the point of view of a map from the parameterization of a neural network.
Let us consider a convex loss function $\mathcal{L}:\R \times \R  \to \R$ and a sample $S = (\Bx_i, y_i)_{i=1}^m \in (\R^d \times \R)^m$.

Then, for two neural networks $\Phi_1, \Phi_2$ and for $\alpha \in (0,1)$ it holds that 
\begin{align*}
	\widehat{\mathcal{R}}_S(\alpha \Phi_1 + (1-\alpha) \Phi_2) &= \frac{1}{m}\sum_{i=1}^m \mathcal{L}(\alpha \Phi_1(\Bx_i) + (1-\alpha) \Phi_2 (\Bx_i), y_i)\\
	& \leq   \frac{1}{m}\sum_{i=1}^m \alpha \mathcal{L}(\Phi_1(\Bx_i), y_i) + (1- \alpha) \mathcal{L}(\Phi_2(\Bx_i), y_i)\\
	& = \alpha 	 \widehat{\mathcal{R}}_S(\Phi_1) + (1-\alpha) \widehat{\mathcal{R}}_S(\Phi_2).
\end{align*}
Hence, the empirical risk %
is convex %
when considered as a map depending on the neural network functions rather then the neural network parameters.
A convex function does not have spurious minima or saddle points.
As a result, the issues from the previous section are avoided if we take the perspective of neural network sets. 

So why do we not optimize over the sets of neural networks instead of the parameters? To understand this, %
we will now study the set of neural networks associated with a fixed architecture as a subset of other function spaces. 

We start by investigating the realization map $R_\sigma$ introduced in Definition \ref{def:realizationetc}.
Concretely, we %
show in Section \ref{sec:LipschitzParam}, that if $\sigma$ is Lipschitz, then the set of neural networks is the image of $\mathcal{PN}(\CA, \infty)$ under a locally Lipschitz map.
We will use this fact to show in Section \ref{sec:convexityofNNSpaces} that sets of neural networks are typically non-convex, and even have arbitrarily large holes. 
Finally, in Section \ref{sec:ClosednessBestApprox}, we study the extent to which there exist best approximations to arbitrary functions, in the set of neural networks.
We will demonstrate that the lack of best approximations causes the weights of neural networks to grow infinitely during training.

\section{Lipschitz parameterizations}\label{sec:LipschitzParam}

In this section, we study the Lipschitz continuity of the realization map $R_\sigma$.
The main result is the following simplified version of \cite[Proposition 4]{petersen2021topological}.
For its formulation we introduce a sparsity vector $\Bs$ whose entries consist of zeros and ones. Its purpose is to switch off certain parameters of the network, by an elementwise multiplication of the network weights $\Bw$ with the sparsity vector $\Bs$. This will allow us to connect to the (sparse) networks constructed in Chapters \ref{chap:ReLUNNs}-\ref{chap:DReLUNN} to show convergence rates.

Recall that for an architecture $\CA=(d_0,\dots,d_{L+1})$, the number of parameters in a fully connected network is
  \begin{equation*}
    n_\CA= \sum_{\ell=0}^Ld_{\ell+1}(d_\ell+1).
  \end{equation*}
  In the following we use again the notation for the parameter range $\CP\CN(\CA,B)$, and
  \begin{equation*}
    R_\sigma:\Bw\mapsto \Phi(\cdot,\Bw),
  \end{equation*}
  for the realization map, where $\Bw\in\R^{n_\CA}$ (cf.~Definitions
  \ref{def:nn} and \ref{def:realizationetc}).

	\begin{proposition}\label{prop:LipschitzOfRealizationMap}
          Let $\CA = (d_0, d_1, \dots, d_{L+1}) \in \N^{L+2}$, let $\sigma\colon \R \to \R$ be $C_\sigma$-Lipschitz continuous with $C_\sigma \geq 1$, let $|\sigma(x)| \leq C_\sigma\cdot (1+|x|)$ for all $x \in \R$, and let $B\geq 1$.

          Let $\Bs\in\{0,1\}^{n_\CA}$. %
          Then, for all $\Bw$,
          $\Bv \in \mathcal{PN}(\CA, B)$,
		\begin{align*}
			\norm[{L^{\infty}([-1,1]^{d_0})}]{R_\sigma(\Bw\odot\Bs) - R_\sigma(\Bv\odot\Bs)} \leq  (3C_\sigma B d_{\rm max})^{L} |\Bs| \norm[\infty]{\Bw - \Bv},  
		\end{align*} 
                where $d_{\rm max} = \max_{\ell = 0, \dots, L+1} d_\ell$,
                and $|\Bs|=\sum_{j=1}^{n_\CA}s_j\le n_\CA$.
	\end{proposition}

          Before we come to the proof of Proposition \ref{prop:LipschitzOfRealizationMap}, we show two lemmata, investigating Lipschitz properties of feedforward neural networks, and upper bounds on the output of each layer.

		\begin{lemma}\label{lem:LipschitzEstimateNN2324}
                  Let $\CA=(d_0,\dots,d_{L+1})\in\N^{L+2}$, let $\sigma:\R\to\R$ be $C_\sigma$-Lipschitz continuous with $C_\sigma\ge 0$, and let $B\ge 0$.
                  Then it
                  holds 
                  for all $\Phi \in \mathcal{N}(\sigma; \CA,  B)$ %
		\begin{align} \label{eq:LipschitzEstimateNN2324}
			\|\Phi(\Bx) - \Phi(\By)\|_\infty \leq C_\sigma^L \cdot (B d_{\rm max})^{L+1} \|\Bx-\By \|_\infty
		\end{align}
		for all $\Bx$, $\By \in \R^{d_0}$.
		\end{lemma}
		\begin{proof}
			We start with the case, where $L = 1$.
Then, for $\CA=(d_0, d_1,d_2)$, we have
			\[
				\Phi(\Bx) = \BW^{(1)} \sigma(\BW^{(0)}\Bx + \Bb^{(0)}) + \Bb^{(1)},
			\]
			for %
                        certain
                        $\BW^{(0)}, \BW^{(1)}, \Bb^{(0)}, \Bb^{(1)}$ with the absolute value of all entries bounded by $B$.
                        Thus
		\begin{align*}
			\|\Phi(\Bx) - \Phi(\By)\|_\infty &= \left\|\BW^{(1)} \left( \sigma(\BW^{(0)}\Bx + \Bb^{(0)}) - \sigma(\BW^{(0)}\By + \Bb^{(0)})\right) \right\|_\infty\\
			&\leq d_{1}B \left\| \sigma(\BW^{(0)}\Bx + \Bb^{(0)}) - \sigma(\BW^{(0)}\By + \Bb^{(0)})\right\|_\infty\\
			&\leq d_{1}B C_\sigma\left\| \BW^{(0)}(\Bx - \By)\right\|_\infty\\
                                                          &\leq d_{1}d_0 B^2 C_\sigma \left\|\Bx - \By\right\|_\infty \\
                  &\leq C_\sigma \cdot (d_{\rm max} B)^2 \left\|\Bx - \By\right\|_\infty,
		\end{align*}
		where we used the Lipschitz continuity of $\sigma$ and the fact that $\|\BA\Bz\|_\infty \leq n \max_{i,j}|A_{ij}| \|\Bz\|_\infty$ for every matrix $\BA=(A_{ij})_{i = 1, j = 1}^{m,n} \in \R^{m\times n}$.
		
		The induction step from $L$ to $L+1$ follows similarly.
This concludes the proof of the lemma.		
		\end{proof}

                In the next Lemma, $\Bx^{(\ell)}$ as in \eqref{eq:xell} denotes the output of the $\ell$th layer.
		
		\begin{lemma} \label{lem:xellestimate}
                  Consider the setting of Proposition \ref{prop:LipschitzOfRealizationMap}. Then
		 \begin{align}
		 	\label{eq:xellestimate}
			\|\Bx^{(\ell)}\|_\infty \leq (3 C_\sigma B d_{\rm max})^\ell \quad \text{ for all } \Bx \in [-1,1]^{d_{0}}.
		\end{align}
		\end{lemma}	
		\begin{proof}
                  Clearly it suffices to consider the case $\Bs=(1)_{j=1}^{n_\CA}$, which allows for arbitrary weights bounded in modulus by $B$ (in particular zero weights).
                  
                  Per Definitions \eqref{eq:defxell} and \eqref{eq:defxLplus1}, we have that for $\ell = 1, \dots, L+1$ 
		\begin{align*}
			\|\Bx^{(\ell)}\|_\infty &\leq C_\sigma\left\|\BW^{(\ell-1)}\Bx^{(\ell-1)} + \Bb^{(\ell-1)} \right\|_\infty+C_\sigma\\
			&\leq  C_\sigma B d_{\rm max} \|\Bx^{(\ell-1)}\|_\infty + B C_\sigma+C_\sigma,
		\end{align*}
                where we used $|\sigma(y)|\le C_\sigma\cdot (1+|y|)$, the triangle inequality,
                and again the estimate $\|\BA\Bx\|_\infty \leq n \max_{i,j}|A_{ij}| \|\Bx\|_\infty$.
                Since $B$, $d_{\max}\ge 1$, we thus have
		\begin{align*}
                  \|\Bx^{(\ell)}\|_\infty
                                  \leq 3 C_\sigma B d_{\rm max}\cdot \max\{1,\|\Bx^{(\ell-1)}\|_\infty\}.
		\end{align*}
		Resolving the recursive estimate of $\|\Bx^{(\ell)}\|_\infty$ by $2 C_\sigma B d_{\rm max}(\max\{1,\|\Bx^{(\ell-1)}\|_\infty\})$, we conclude that 
		\begin{align*}
			\|\Bx^{(\ell)}\|_\infty \leq (3 C_\sigma B d_{\rm max})^\ell \cdot \max\{1,\|\Bx^{(0)}\|_\infty\} = (3 C_\sigma B d_{\rm max})^\ell.
		\end{align*}
		This concludes the proof of the lemma.
		\end{proof}	

                We are now in position to prove the proposition.

\begin{proof}[of Proposition \ref{prop:LipschitzOfRealizationMap}]
          Let $\Bw$, $\Bv \in \mathcal{PN}(\CA, B)$. %
          There is no loss of generality in assuming that $\Bs=(1,\dots,1,0,\dots,0)$, i.e.\ the first $s\dfn|\Bs|$ entries of $\Bs$ contain ones.
          
          For $j=0,\dots,s-1$ let
            \begin{equation*}
              \Bu_j\dfn (w_1,\dots,w_{n_\CA-j},v_{n_{\CA}-j+1},\dots,v_{n_\CA})^\top,
            \end{equation*}
            so that $\Bu_j$, $\Bu_{j+1}$ differ only in one entry, and in particular $\Bu_0\odot\Bs=\Bw\odot\Bs$ and $\Bu_{s}\odot \Bs=\Bv\odot \Bs$.
            For any $\Bx\in [-1,1]^{d_0}$, the triangle inequality then gives
		\begin{align}\label{eq:estimateThis}
                  \|R_\sigma(\Bw)(\Bx) - R_\sigma(\Bv)(\Bx)\|_\infty \leq \sum_{j=0}^{s-1} \|R_\sigma(\Bu_j)(\Bx) - R_\sigma(\Bu_{j+1})(\Bx)\|_\infty.
		\end{align}

                Fix $j\in\{0,\dots,s-1\}$, 
                  and let $\ell$ be such that
                  the differing entry between $\Bu_j$, $\Bu_{j+1}$ belongs
                  to a parameter in the
                $\ell$th layer. We first assume $\ell<L$.
                Then for $\Bx\in [-1,1]^{d_0}$
		\begin{align*}
			|R_\sigma(\Bu_j)(\Bx) - R_\sigma(\Bu_{j+1})(\Bx)| = |\Phi^\ell( \sigma(\BW^{{(\ell)}}\Bx^{(\ell)} + \Bb^{(\ell)})) - \Phi^\ell( \sigma(\overline{\BW^{{(\ell)}}}\Bx^{(\ell)} + \overline{\Bb^{(\ell)}}))|,
		\end{align*}
		where $\Phi_\ell \in \mathcal{N}(\sigma; \CA_\ell, B)$ for 
		$\CA_\ell = (d_{\ell+1}, \dots, d_{L+1})$ and $(\BW^{{(\ell)}}, \Bb^{{(\ell)}})$, $(\overline{\BW}^{{(\ell)}}, \overline{\Bb}^{{(\ell)}})$ differ in one entry only.

		Using the Lipschitz continuity of $\Phi_\ell$ of Lemma \ref{lem:LipschitzEstimateNN2324}, we have %
		\begin{align*}
			&|R_\sigma(\Bu_j)(\Bx) - R_\sigma(\Bu_{j+1})(\Bx)| \\
			&\qquad \leq  C_\sigma^{L-\ell - 1} (B d_{\rm max})^{L-\ell}|\sigma(\BW^{{(\ell)}}\Bx^{(\ell)} + \Bb^{(\ell)}) - \sigma(\overline{\BW^{{(\ell)}}}\Bx^{(\ell)} + \overline{\Bb^{(\ell)}})|\\
			&\qquad \leq C_\sigma^{L-\ell} (B d_{\rm max})^{L-\ell} \| \BW^{{(\ell)}}\Bx^{(\ell)} + \Bb^{(\ell)} -  \overline{\BW^{{(\ell)}}}\Bx^{(\ell)} - \overline{\Bb^{(\ell)}}\|_\infty\\
			&\qquad \leq 2 C_\sigma^{L-\ell} (B d_{\rm max})^{L-\ell} %
                          \|\Bw\odot\Bs - \Bv\odot\Bs\|_{\infty}
                          \max\{1,\|\Bx^{(\ell)}\|_\infty\},
		\end{align*}
		Invoking Lemma \ref{lem:xellestimate}, we conclude that 
		\begin{align*}
			|R_\sigma(\Bu_j)(\Bx) - R_\sigma(\Bu_{j+1})(\Bx)| &\leq  (3 C_\sigma B d_{\rm max})^\ell C_\sigma^{L-\ell}\cdot (B d_{\rm max})^{L-\ell} \delta\\
			&\leq  (3C_\sigma B d_{\rm max})^{L} \|\Bw\odot\Bs - \Bv\odot\Bs\|_{\infty}.
		\end{align*}
                For the case $\ell=L$, a similar estimate can be shown.
		Combining this with \eqref{eq:estimateThis} yields the result.
	\end{proof}

Using Proposition \ref{prop:LipschitzOfRealizationMap}, we can now consider the set of neural networks with a fixed architecture $\mathcal{N}(\sigma; \CA, \infty)$ as a subset of $L^\infty([-1,1]^{d_0})$.
Moreover, for the (non-sparse) case $\Bs=(1)_{j=1}^{n_\CA}$,
  we have shown that $\mathcal{N}(\sigma; \CA, \infty)$ is the image of $\mathcal{PN}(\CA, \infty)$ under a \textbf{locally Lipschitz map}. 

\section{Convexity of neural network spaces}\label{sec:convexityofNNSpaces}

As a first step towards understanding $\mathcal{N}(\sigma; \CA, \infty)$ as a subset of $L^\infty([-1,1]^{d_0})$, we notice that it is star-shaped with few centers.
Let us first introduce the necessary terminology.

\begin{definition}
Let $Z$ be a subset of a linear space.
A point $x \in Z$ is called a \textbf{center of $Z$} if, for every $y \in Z$ it holds that 
\[
  \set{tx+ (1-t)y}{t \in [0,1]} \subseteq Z.
\] 
A set is called \textbf{star-shaped} if it has at least one center.
\end{definition}

The following proposition follows directly from the definition of a neural network and is the content of Exercise \ref{ex:proofOfPropositionStarShaped}.

\begin{proposition}\label{prop:NNsAreStarShaped}
Let $L\in \N$ and $\CA = (d_0, d_1, \dots, d_{L+1})\in \N^{L+2}$ and $\sigma \colon \R \to \R$.
Then $\mathcal{N}(\sigma; \CA, \infty)$ is scaling invariant, i.e.
for every $\lambda \in \R$ it holds that $\lambda f \in \mathcal{N}(\sigma; \CA, \infty)$ if $f \in \mathcal{N}(\sigma; \CA, \infty)$, and hence $0 \in \mathcal{N}(\sigma; \CA, \infty)$ is a center of $\mathcal{N}(\sigma; \CA, \infty)$.
\end{proposition}

Knowing that $\mathcal{N}(\sigma; \CA , B)$ is star-shaped with center $0$, we can also ask ourselves if $\mathcal{N}(\sigma; \CA, B)$ has more than this one center.  %
It is not hard to see that also every constant function is a center. %
The following theorem,
which corresponds to \cite[Proposition C.4]{petersen2021topological},
yields an upper bound on the number of \emph{linearly independent} centers. 

\begin{theorem}%
  \label{thm:numberofcentersBound}
Let $L\in \N$ and $\CA = (d_0, d_1, \dots, d_{L+1})\in \N^{L+2}$, and let $\sigma: \R \to \R$ be Lipschitz continuous.
Then, $\mathcal{N}(\sigma; \CA, \infty)$ contains at most $n_{\CA} = \sum_{\ell = 0}^{L} (d_{\ell} + 1) d_{\ell+1}$
linearly independent centers.
\end{theorem}

\begin{proof}
Assume %
by contradiction, that there are functions $(g_i)_{i=1}^{n_{\CA}+1} \subseteq \mathcal{N}( \sigma ; \CA, \infty) \subseteq L^\infty([-1,1]^{d_0})$ that are linearly independent and centers of $\mathcal{N}(\sigma; \CA, \infty)$. 

By the Theorem of Hahn-Banach, there exist $(g_i')_{i=1}^{n_{\CA}+1} \subseteq (L^\infty([-1,1]^{d_0}))'$ such that $g_i'(g_j) = \delta_{ij}$, for all $i,j \in \{1, \dots, L+1\}$.
We define 
\begin{align*}
	T\colon L^\infty([-1,1]^{d_0}) \to \R^{n_{\CA}+1}, \qquad
	g \mapsto \left(\begin{array}{c}
	 g_1'(g) \\
	 g_2'(g)\\
	 \vdots\\
	 g_{n_{\CA}+1}'(g)
	\end{array} \right).
\end{align*}
Since $T$ is continuous and linear, we have that $T \circ R_\sigma$ is locally Lipschitz continuous by Proposition \ref{prop:LipschitzOfRealizationMap}. 
Moreover, since the $(g_i)_{i=1}^{n_{\CA}+1}$ are linearly independent, we have that $T(\mathrm{span}((g_i)_{i=1}^{n_{\CA}+1}))= \R^{n_{\CA}+1}$. %
We denote $V \coloneqq \mathrm{span}((g_i)_{i=1}^{n_{\CA}+1})$. 

Next, we would like to establish that $\mathcal{N}(\sigma; \CA, \infty) \supset V$.
Let $g \in V$ then 
\[
	g = \sum_{\ell = 1}^{n_{\CA}+1} a_\ell g_\ell, 
\]
for some $a_1, \dots, a_{n_{\CA}+1} \in \R$. %
We show by induction that $\tilde{g}^{(m)} \coloneqq \sum_{\ell = 1}^{m} a_\ell g_\ell \in \mathcal{N}(\sigma; \CA, \infty)$ for every $m \leq n_{\CA}+1$. 
This is obviously true for $m = 1$. 
Moreover, we have that  $\tilde{g}^{(m+1)} = a_{m+1}g_{m+1} + \tilde{g}^{(m)}$. 
Hence, the induction step holds true if $a_{m+1}= 0$. 
If $a_{m+1} \neq 0$, then we have that 
\begin{align}
	\widetilde{g}^{(m+1)} &= 2 a_{m+1} \cdot \left(\frac{1}{2} g_{m+1} + \frac{1}{2a_{m+1}}\widetilde{g}^{(m)}\right). \label{eq:oihfwoihdo}
\end{align}
By the induction assumption $\widetilde{g}^{(m)} \in \mathcal{N}(\sigma; \CA,  \infty)$ and hence by Proposition \ref{prop:NNsAreStarShaped}  $\widetilde{g}^{(m)}/(a_{m+1}) \in \mathcal{N}(\sigma; \CA, \infty)$.
Additionally, since $g_{m+1}$ is a center of $\mathcal{N}(\sigma; \CA, \infty)$, we have that $\frac{1}{2} g_{m+1} + \frac{1}{2a_{m+1}}\widetilde{g}^{(m)} \in \mathcal{N}(\sigma; \CA,  \infty)$.
By Proposition \ref{prop:NNsAreStarShaped}, we conclude that $\widetilde{g}^{(m+1)}\in \mathcal{N}( \sigma; \CA,\infty)$. 

The induction shows that $g \in \mathcal{N}( \sigma; \CA,\infty)$ and thus $V\subseteq \mathcal{N}( \sigma; \CA,\infty)$.
As a consequence, $T \circ R_\sigma(\mathcal{PN}(\CA, \infty)) \supseteq T(V) = \R^{n_\CA+1}$.

It is a well known fact of basic analysis that for every $n \in \N$ there does not exist a surjective and locally Lipschitz continuous map from $\R^n$ to $\R^{n+1}$. 
We recall that $n_{\CA} = \mathrm{dim}(\mathcal{PN}(\CA, \infty))$.
This yields the contradiction.
\end{proof}

For a convex set $X$, the line between all two points of $X$ is a subset of $X$.
Hence, every point of a convex set is a center.
This yields the following corollary.

\begin{corollary}\label{cor:NonConvexity}
Let $\CA = (d_0, d_1, \dots, d_{L+1})$, let, and let $\sigma: \R \to \R$ be Lipschitz continuous.
If $\mathcal{N}( \sigma; \CA,\infty)$ contains more than $n_\CA= \sum_{\ell = 0}^{L} (d_{\ell} + 1) d_{\ell+1}$ linearly independent functions, then $\mathcal{N}( \sigma; \CA,\infty)$ is not convex.
\end{corollary}

Corollary \ref{cor:NonConvexity} tells us that we cannot expect convex sets of neural networks, if the set of neural networks has many linearly independent elements. 
Sets of neural networks contain for each $f \in \mathcal{N}( \sigma; \CA,\infty)$ also all shifts of this function, i.e., $f(\cdot + \Bb)$ for a $\Bb \in \R^d$ are elements of $\mathcal{N}( \sigma; \CA,\infty)$. 
For a set of functions, being shift invariant and having only finitely many linearly independent functions at the same time, is a very restrictive condition. 
Indeed, it was shown in \cite[Proposition C.6]{petersen2021topological} that if $\mathcal{N}( \sigma; \CA,\infty)$ has only finitely many linearly independent functions and $\sigma$ is differentiable in at least one point and has non-zero derivative there, then $\sigma$ is necessarily a polynomial.

\begin{figure}
  \begin{center}
    \includegraphics[width = 0.5\textwidth]{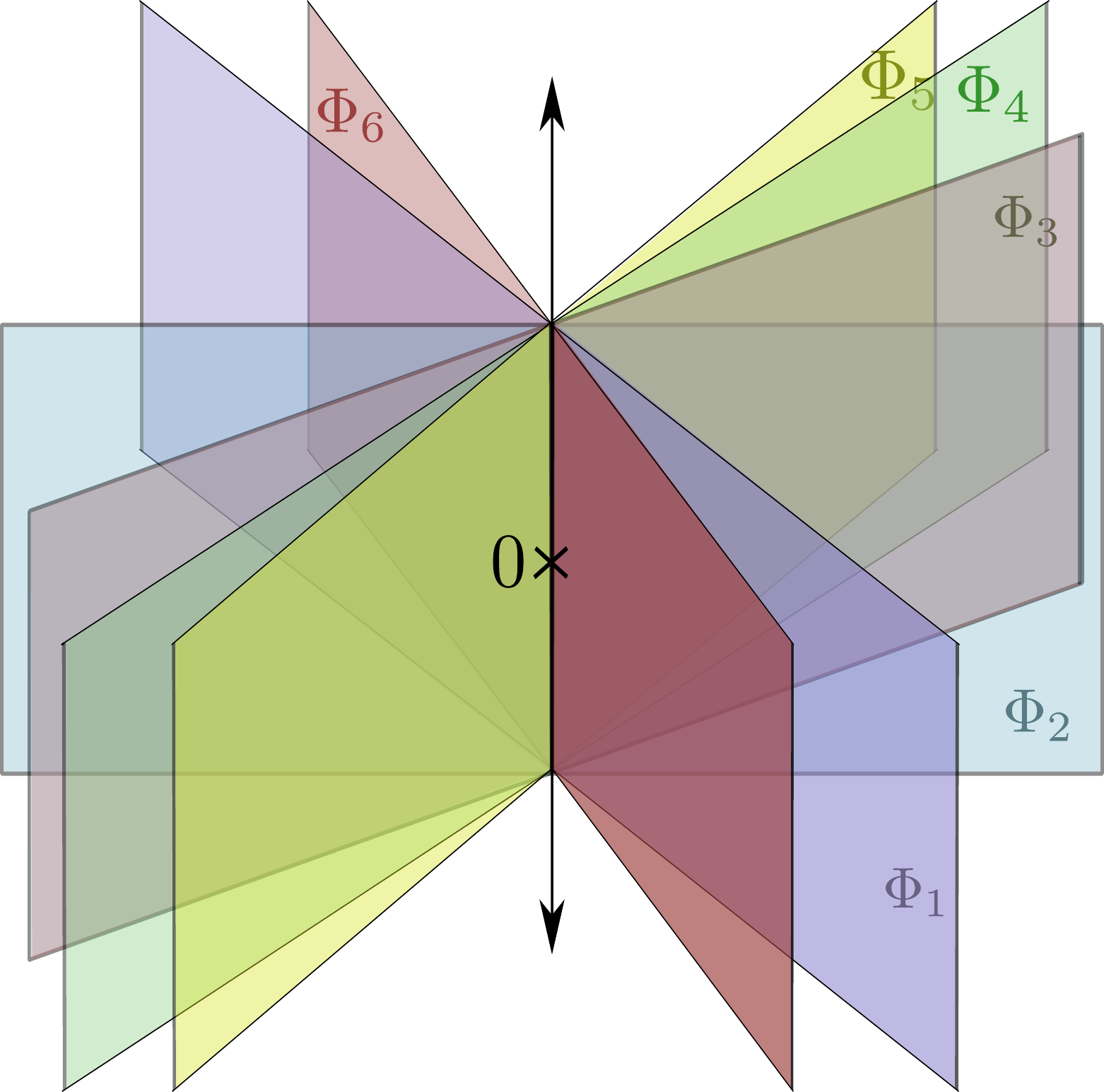}
  \end{center}
	\caption{Sketch of the space of neural networks in 3D.
          The vertical axis corresponds to the constant neural network functions, each of which is a center. 
The set of neural networks %
consists of many low-dimensional linear subspaces spanned by certain neural networks ($\Phi_1, \dots, \Phi_6$ in this %
sketch) and linear functions.
Between these low-dimensional subspaces, there is not always a straight-line connection by Corollary \ref{cor:NonConvexity} and Theorem \ref{thm:noEpsConvexity}.
}
	\label{fig:ShapeOfNNSpaces}
      \end{figure}

We conclude that the set of neural networks is in general non-convex and star-shaped with 0 and constant functions being centers.
One could visualize this set in 3D as in Figure \ref{fig:ShapeOfNNSpaces}.

The fact, that the neural network space is not convex, could also mean that it %
merely fails to be convex at one point.
For example $\R^2 \setminus \{0\}$ is not convex, but for an optimization algorithm this would likely not pose a problem.

We will next observe that $\mathcal{N}( \sigma; \CA,\infty)$ does not have such a benign non-convexity and in fact, has \emph{arbitrarily large holes}.

To make this claim mathematically precise, we first introduce the notion of $\eps$-convexity. 

\begin{definition}
For $\eps > 0$, we say that a subset $A$ of a normed vector space $X$ is $\eps$-convex if
\[
	\co(A) \subseteq A + B_\eps(0),
\]
where $\co(A)$ denotes the convex hull of $A$ and $B_\eps(0)$ is an $\eps$ ball around $0$ with respect to the norm of $X$.
\end{definition}

Intuitively speaking, a set that is convex when one fills up all holes smaller than $\eps$ is $\eps$-convex.
Now we show that there is no $\eps>0$ such that $\mathcal{N}( \sigma; \CA,\infty)$ is $\eps$-convex.

\begin{theorem}\label{thm:noEpsConvexity}
Let $L \in \N$ and $\CA = (d_0, d_1, \dots, d_{L}, 1) \in \N^{L+2}$.
Let $K \subseteq \R^{d_0}$ be compact and let $\sigma \in \CM$, with $\CM$ as in \eqref{eq:M} and assume that $\sigma$ is not a polynomial.
Moreover, assume that there exists an open set, where $\sigma$ is differentiable and not constant.

If there exists an $\eps>0$ such that $\mathcal{N}( \sigma; \CA,\infty)$ is $\eps$-convex, then $\mathcal{N}( \sigma; \CA,\infty)$ is dense in $C(K)$.
\end{theorem}

\begin{proof}

\textbf{Step 1.} We show that $\eps$-convexity %
implies $\overline{\mathcal{N}( \sigma; \CA,\infty)}$ to be convex.
By Proposition \ref{prop:NNsAreStarShaped}, we have that $\mathcal{N}( \sigma; \CA,\infty)$ is scaling invariant. 
This implies that $\co(\mathcal{N}( \sigma; \CA,\infty))$ is scaling invariant as well. 
Hence, if there exists $\eps >0$ such that $\mathcal{N}( \sigma; \CA,\infty)$ is $\eps$-convex, then for every $\eps' >0$
\begin{align*}
	\co(\mathcal{N}( \sigma; \CA,\infty)) &= \frac{\eps'}{\eps}\co(\mathcal{N}( \sigma; \CA,\infty)) \subseteq \frac{\eps'}{\eps}\left(\mathcal{N}( \sigma; \CA,\infty) + B_\eps(0)\right)\\
	&= \mathcal{N}( \sigma; \CA,\infty) + B_{\eps'}(0).
\end{align*}
This yields that $\mathcal{N}( \sigma; \CA,\infty)$ is $\eps'$-convex.
Since $\eps'$ was arbitrary, we have that $\mathcal{N}( \sigma; \CA,\infty)$ is $\eps$-convex for all $\eps>0$. 

As a consequence, we have that 
\begin{align*}
	\co(\mathcal{N}( \sigma; \CA,\infty)) \subseteq & \bigcap_{\eps>0} (\mathcal{N}( \sigma; \CA,\infty) + B_\eps(0))\\
	 \subseteq &\bigcap_{\eps>0} \overline{(\mathcal{N}( \sigma; \CA,\infty) + B_\eps(0))} = \overline{\mathcal{N}( \sigma; \CA,\infty)}.
\end{align*}
Hence, $\overline{\co}(\mathcal{N}( \sigma; \CA,\infty)) \subseteq \overline{\mathcal{N}( \sigma; \CA,\infty)}$ and, by the well-known fact that in every metric vector space $\co(\overline{A}) \subseteq \overline{\co}(A)$, we conclude that $\overline{\mathcal{N}( \sigma; \CA,\infty)}$ is convex.

\textbf{Step 2.} %
We show that $\CN_d^1(\sigma; 1)\subseteq\overline{\CN(\sigma;\CA,\infty)}$.
If $\mathcal{N}( \sigma; \CA,\infty)$ is %
$\eps$-convex, then %
by Step 1
$\overline{\mathcal{N}( \sigma; \CA,\infty)}$ is convex. %
The scaling invariance of $\mathcal{N}( \sigma; \CA,\infty)$ %
then shows that $\overline{\mathcal{N}( \sigma; \CA,\infty)}$ is a closed linear subspace of $C(K)$.

Note that, by Proposition \ref{prop:Identity1} for every $\Bw \in \R^{d_0}$ and $b \in \R$ there exists a function $f \in \overline{\mathcal{N}( \sigma; \CA,\infty)}$ such that
\begin{align}\label{eq:singleNeuronsAreIn}
	f(\Bx) = \sigma(\Bw^\top \Bx + b) \qquad\text{for all }\Bx\in K.
\end{align}
By definition, every constant function is an element of $\overline{\mathcal{N}( \sigma; \CA,\infty)}$.

Since $\overline{\mathcal{N}( \sigma; \CA,\infty)}$ is a closed vector space, this implies that for all $n \in \N$ and all $\Bw_1^{(1)}, \dots, \Bw_n^{(1)} \in \R^{d_0}$, $w_1^{(2)}, \dots, w_n^{(2)} \in \R$, $b_1^{(1)}, \dots, b_n^{(1)} \in \R$, $b^{(2)} \in \R$
\begin{align} \label{eq:almostNNRepresentation}
	\Bx \mapsto \sum_{i=1}^n w_i^{(2)}\sigma((\Bw_i^{(1)})^\top \Bx + b_i^{(1)})  + b^{(2)} \in \overline{\mathcal{N}( \sigma; \CA,\infty)}.
\end{align}

\textbf{Step 3.} %
From \eqref{eq:almostNNRepresentation}, we conclude that $\CN_d^1(\sigma; 1) \subseteq \overline{\mathcal{N}( \sigma; \CA,\infty)}$.
In words, the whole set of shallow neural networks of arbitrary width is contained in the closure of the set of neural networks with a fixed architecture. 
By Theorem \ref{thm:universal}, we have that $\CN_d^1(\sigma;1)$ is dense in $C(K)$, which yields the result.
\end{proof}

For any activation function of practical relevance,
a set of neural networks with fixed architecture %
is not dense in $C(K)$.
This is only %
the case for very strange activation functions such as the one discussed in Subsection \ref{sec:kolmogorov}. 
Hence, Theorem \ref{thm:noEpsConvexity} shows that in %
general, sets of neural networks of fixed architectures have arbitrarily large holes.

\section{Closedness and best-approximation property}\label{sec:ClosednessBestApprox}

The non-convexity of the set of neural networks can have some serious consequences for the way we think of the approximation or learning problem by neural networks.

Consider $\CA = (d_0, \dots, d_{L+1}) \in \N^{L+2}$ %
and an activation function $\sigma$.
Let $H$ be a normed function space on $[-1,1]^{d_{0}}$ such that $\mathcal{N}( \sigma; \CA,\infty) \subseteq H$.
For $h \in H$ we would like to find a neural network that best %
approximates $h$, i.e.\ to find %
${\Phi_*} \in \mathcal{N}( \sigma; \CA,\infty)$ such that
\begin{align}\label{eq:BestApproximationProblem}
	\| {\Phi_*} - h \|_{H} = \inf_{\Phi \in \mathcal{N}( \sigma; \CA,\infty)} \|\Phi - h \|_{H}.
\end{align} 
We say that $\mathcal{N}( \sigma; \CA,\infty)\subseteq H$ has
\begin{itemize}
	\item the {\bf best approximation property}, if for all $h\in H$ there exists at least one ${\Phi_*}\in \mathcal{N}( \sigma; \CA,\infty)$ such that \eqref{eq:BestApproximationProblem} holds,
	\item the {\bf unique best approximation property}, if for all $h\in H$ there exists exactly one ${\Phi_*}\in \mathcal{N}( \sigma; \CA,\infty)$ such that \eqref{eq:BestApproximationProblem} holds,
	\item the {\bf continuous selection property}, if there exists a continuous function $\phi \colon H \to \mathcal{N}( \sigma; \CA,\infty)$ such that ${\Phi_*} = \phi(h)$ satisfies \eqref{eq:BestApproximationProblem} for all $h \in H$.
\end{itemize}
We will see in the sequel, that, in the absence of the best approximation property, we will be able to prove that the learning problem necessarily requires the weights of the neural networks to tend to infinity, which may or may not be desirable in applications. 

Moreover, having a continuous selection procedure is desirable as it implies the existence of a
stable selection algorithm; that is, an algorithm which, for similar problems yields similar neural networks satisfying \eqref{eq:BestApproximationProblem}.

Below, we will study the properties above for $L^p$ spaces, $p \in [1,\infty)$. As we will see, neural network classes typically neither satisfy the continuous selection nor the best approximation property.

\subsection{Continuous selection}

As shown in \cite{kainen1999approximation}, %
neural network spaces essentially never admit the continuous selection property.
To give the argument, we first recall the following result from \cite[Theorem 3.4]{kainen1999approximation} without proof.
\begin{theorem}\label{thm:UBAImpliesConvex}
  Let $p \in (1, \infty)$. Every subset of $L^p([-1,1]^{d_0})$ with the unique best approximation property is convex.
\end{theorem} %
This allows to show the next proposition.

\begin{proposition}\label{prop:noContinuousSelection}
Let $L\in \N$, $\CA = (d_0, d_1, \dots, d_{L+1})\in \N^{L+2}$, let $\sigma: \R \to \R$ be Lipschitz continuous and not a polynomial, and let $p \in (1,\infty)$.

Then, $\mathcal{N}( \sigma; \CA,\infty)\subseteq L^{p}([-1,1]^{d_0})$ does not have the continuous selection property.
\end{proposition}

\begin{proof}
	We observe from Theorem \ref{thm:numberofcentersBound} and the discussion below, that under the present assumptions, %
        $\mathcal{N}( \sigma; \CA,\infty)$ is not convex. %
	
	We conclude from Theorem \ref{thm:UBAImpliesConvex} that $\mathcal{N}( \sigma; \CA,\infty)$ does not have the unique best approximation property. %
	Moreover, if the set $\mathcal{N}( \sigma; \CA,\infty)$ does not have the best approximation property, then it is obvious that it cannot have continuous selection. 
	Thus, we can assume without loss of generality, that $\mathcal{N}( \sigma; \CA,\infty)$ has the best approximation property and there exists a point $h \in L^p([-1,1]^{d_0})$ and two different $\Phi_1$,$\Phi_2$ such that 
	\begin{align}\label{eq:hnotinN}
		\| \Phi_1 - h \|_{L^p} = 	\| \Phi_2 - h \|_{L^p} = \inf_{\Phi \in \mathcal{N}( \sigma; \CA,\infty)} \| \Phi - h \|_{L^p}.
	\end{align}
	Note that \eqref{eq:hnotinN} implies that $h \not \in \mathcal{N}( \sigma; \CA,\infty)$. 
	
	Let us consider the following function:
	\begin{align*}
		[-1,1] \ni \lambda \mapsto P(\lambda) = \left\{ \begin{array}{ll}
			(1+ \lambda) h - \lambda \Phi_1 & \text{ for }\lambda \leq 0,\\
			(1- \lambda)h + \lambda \Phi_2& \text{ for }\lambda \geq 0.\\
		\end{array}\right.
	\end{align*}
	It is clear that $P(\lambda)$ is a continuous path in $L^{p}$.
Moreover, for $\lambda  \in (-1,0)$
	\[
		\|\Phi_1 -P(\lambda)\|_{L^p} = (1+\lambda) \|\Phi_1 - h\|_{L^p}.
	\]
	Assume towards a contradiction, that there exists $\Phi_* \in \mathcal{N}( \sigma; \CA,\infty)$ with $\Phi_* \neq \Phi_1$ such that for $\lambda \in (-1,0)$ %
	\begin{align*}
		\|\Phi_* -P(\lambda)\|_{L^p} \leq \|\Phi_1 - P(\lambda)\|_{L^p}.
	\end{align*}
	Then
	\begin{align}
		\|\Phi_* - h\|_{L^p} &\leq \|\Phi_* - P(\lambda)\|_{L^p} +  \|P(\lambda) - h\|_{L^p}  \nonumber\\
					&\leq  \|\Phi_1 - P(\lambda)\|_{L^p} +   \|P(\lambda) - h\|_{L^p} \nonumber\\
					& = (1+\lambda) \| \Phi_1 - h\|_{L^p} +  |\lambda| \| \Phi_1 - h\|_{L^p} =  \| \Phi_1 - h\|_{L^p} .\label{eq:equalityAfterBootstrapping}
	\end{align} %
	Since $\Phi_1$ is a best approximation to $h$ this implies that every inequality in the estimate above is an equality.
Hence, we have that 
	\begin{align*}
	\|\Phi_* - h\|_{L^p} = \|\Phi_* - P(\lambda)\|_{L^p} +  \|P(\lambda) - h\|_{L^p}.
	\end{align*}
	However, in a strictly convex space like $L^p([-1,1]^{d_0})$ for $p > 1$ this implies that 
	\[
		\Phi_* - P(\lambda) = c\cdot (P(\lambda) - h)
	\]
	for a constant $c \neq 0$.
This yields that 
	\[
		\Phi_* =   h +  (c+1) \lambda \cdot (h - \Phi_1) 
	\]
	and plugging into \eqref{eq:equalityAfterBootstrapping} yields $| (c+1) \lambda| = 1$.
If $(c+1) \lambda = -1$, then we have $\Phi_* = \Phi_1$ which produces a contradiction.
If $(c+1) \lambda = 1$, then 
	\begin{align*}
	\|\Phi_* -  P(\lambda)\|_{L^p} &= \|2h - \Phi_1 - (1+\lambda) h + \lambda \Phi_1\|_{L^p} \\
	&= \|(1-\lambda) h - (1-\lambda) \Phi_1\|_{L^p} > \|P(\lambda) - \Phi_1\|_{L^p},
	\end{align*}
	which is another contradiction. 

	Hence, for every $\lambda <0$ we have that $\Phi_1$ is the unique minimizer to $P(\lambda)$ in $\mathcal{N}( \sigma; \CA,\infty)$.
The same argument holds for $\lambda >0$ and $\Phi_2$.
We conclude that for every selection function $\phi \colon L^p([-1,1]^{d_0})
\to \mathcal{N}( \sigma; \CA,\infty)$ such that $\Phi = \phi(h)$ satisfies \eqref{eq:BestApproximationProblem} for all $h \in L^p([-1,1]^{d_0})$ it holds that 
	\[
		\lim_{\lambda \downarrow 0} \phi(P(\lambda)) = \Phi_2 \neq \Phi_1 = \lim_{\lambda \uparrow 0} \phi(P(\lambda)).
	\]
	As a consequence, $\phi$ is not continuous, which shows the result.
\end{proof}

\subsection{Existence of best approximations}

We have seen in Proposition \ref{prop:noContinuousSelection} that under very mild assumptions, the continuous selection property %
cannot hold.
Moreover, the next result shows that in many cases, 
also the best approximation property fails to be satisfied. %
We provide below a simplified version of  \cite[Theorem 3.1]{petersen2021topological}.
We also refer to \cite{girosi1990networks} for earlier work on this problem.

\begin{proposition}\label{prop:non-closednessSpecificArchitecture}
  Let $\CA = (1, 2, 1)$ and let $\sigma: \R \to \R$ be Lipschitz continuous.
  Additionally assume that there exist $r >0$ and $\alpha' \neq \alpha$ such that $\sigma$ is differentiable for all $|x|>r$ and $\sigma'(x) \to \alpha$ for $x \to \infty$, $\sigma'(x) \to \alpha'$ for $x \to -\infty$.
  
Then, there exists a sequence in $\mathcal{N}( \sigma; \CA,\infty)$ which converges in  $L^{p}([-1,1])$, for every $p \in (1,\infty)$, and the limit of this sequence is discontinuous.
In particular, the limit of the sequence does not lie in $\mathcal{N}(\sigma; \CA', \infty)$ for any $\CA'$.
\end{proposition}

\begin{proof} 
  For all $n\in\N$ let
\begin{align*}
	f_n(x) = \sigma(n x+ 1) -  \sigma(n x)\qquad \text{for all } x \in \R.
\end{align*}
Then $f_n$ can be written as a neural network with architecture $(\sigma; 1,2,1)$, i.e., $\CA = (1,2,1)$.
Moreover, for $x > 0$ we observe with the fundamental theorem of calculus and using integration by substitution that
\begin{align}\label{eq:RHSOfThisConvergestoAlphaPrime}
	f_n(x) = \int_{x}^{x+1/n} n \sigma'(n z) dz =  \int_{nx}^{nx+1} \sigma'(z) dz.
\end{align}
It is not hard to see that the right hand side of \eqref{eq:RHSOfThisConvergestoAlphaPrime} converges to $\alpha$ for $n \to \infty$.

Similarly, for $x < 0$, we observe that $f_n(x)$ converges to $\alpha'$ for $n \to \infty$. 
We conclude that 
\[
	f_n \to \alpha \ind_{\R_+} +  \alpha' \ind_{\R_-} 
\]
almost everywhere as $n\to\infty$.
Since $\sigma$ is Lipschitz continuous, we have that $f_n$ is bounded. 
Therefore, we conclude that $f_n \to  \alpha \ind_{\R_+} +  \alpha' \ind_{\R_-}$ in $L^p$ for all $p \in [1,\infty)$ by the dominated convergence theorem.
\end{proof}

There is a straight-forward extension of Proposition \ref{prop:non-closednessSpecificArchitecture} to arbitrary architectures, that will be the content of Exercises \ref{ex:extendToArbitraryWidth} and \ref{ex:extendToArbitraryDepth}.

\begin{remark}
The proof of Theorem \ref{prop:non-closednessSpecificArchitecture} does not extend to the $L^\infty$ norm.
This, of course, does not mean that generally $\mathcal{N}( \sigma; \CA,\infty)$ is a closed set in $L^\infty([-1,1]^{d_0})$.
In fact, almost all activation functions used in practice still give rise to non-closed neural network sets, see \cite[Theorem 3.3]{petersen2021topological}.
However, there is one notable exception.
For the ReLU activation function, it %
can
be shown that $\mathcal{N}(\sigma_{\rm ReLU}; \CA, \infty)$ is a closed set in $L^\infty([-1,1]^{d_0})$ if $\CA$ has only one hidden layer.
The closedness of deep ReLU spaces in $L^\infty$ is an open problem.  
\end{remark}

\subsection{Exploding weights phenomenon}

Finally, we discuss one of the consequences of the non-existence of best approximations of Proposition \ref{prop:non-closednessSpecificArchitecture}.

Consider a regression problem, where we aim to learn a function $f$ using neural networks with a fixed architecture $\CN(\CA;\sigma,\infty)$. 
As discussed in the Chapters \ref{chap:training} and \ref{chap:wideNets}, we wish to produce a 
sequence of neural networks $(\Phi_n)_{n=1}^\infty$ such that the risk defined in \eqref{eq:riskDef0} converges to $0$.
If the loss $\mathcal{L}$ is the squared loss, $\mu$ is a probability measure on $[-1,1]^{d_0}$, and the data is given by $(\Bx, f(\Bx))$ for $\Bx \sim \mu$, then
 \begin{align}\label{eq:L2convergenceNonClosedness}
 	\begin{split}
 	\mathcal{R}(\Phi_n) &= \| \Phi_n - f \|_{L^2([-1,1]^{d_0}, \mu)}^2\\
 	&= \int_{[-1,1]^{d_0}}|\Phi_n(\Bx) - f(\Bx)|^2 d \mu(\Bx) \to 0 \qquad \text{ for } n \to \infty.
	\end{split}
\end{align}

According to Proposition \ref{prop:non-closednessSpecificArchitecture}, for a given $\CA$, and an activation function $\sigma$, it is possible that \eqref{eq:L2convergenceNonClosedness} holds, but $f \not \in \mathcal{N}( \sigma; \CA,\infty)$.
The following result shows that in this situation, the weights of $\Phi_n$ diverge.

\begin{proposition}\label{prop:ExplodingWeights}
Let $\CA = (d_0, d_1, \dots, d_{L+1}) \in \N^{L+2}$, let $\sigma\colon \R \to \R$ be Lipschitz continuous with $C_\sigma \geq 1$, and $|\sigma(x)| \leq C_\sigma|x|$ for all $x \in \R$, and let $\mu$ be a measure on $[-1,1]^{d_0}$.

Assume that there exists a sequence $\Phi_n  \in \mathcal{N}( \sigma; \CA,\infty)$ and $f \in L^2([-1,1]^{d_0}, \mu) \setminus \mathcal{N}( \sigma; \CA,\infty)$
such that
\begin{align} \label{eq:ThisEquationWillBeReferencedInOneOfTheExercises}
	\| \Phi_n - f \|_{L^2([-1,1]^{d_0}, \mu)}^2 \to 0.
\end{align}
Then 
\begin{align}\label{eq:limsupequalsInfinity}
	\limsup_{n \to \infty}\max \setc{\|\BW_n^{(\ell)}\|_\infty, \|\Bb_n^{(\ell)}\|_\infty}{\ell = 0, \dots L} = \infty.
\end{align}
\end{proposition}

\begin{proof}
We assume towards a contradiction that the left-hand side of \eqref{eq:limsupequalsInfinity} is finite.
As a result, there exists $C >0$ such that $\Phi_n \in \mathcal{N}(\sigma; \CA, C)$ for all $n \in \N$. 

By Proposition \ref{prop:LipschitzOfRealizationMap}, we conclude that $\mathcal{N}(\sigma; \CA,  C)$ is the image of a compact set under a continuous map and hence is itself a compact set in $L^2([-1,1]^{d_0}, \mu)$. 
In particular, we have that $\mathcal{N}(\sigma; \CA, C)$ is closed.
Hence, \eqref{eq:ThisEquationWillBeReferencedInOneOfTheExercises} implies $f \in \mathcal{N}(\sigma; \CA, C)$.
This %
gives a contradiction.
\end{proof}

Proposition \ref{prop:ExplodingWeights} can be extended to all $f$ for which there is no best approximation in $\mathcal{N}( \sigma; \CA,\infty)$, see Exercise \ref{eq:noBestApproxYieldsExplodingWeights}.
The results imply that for functions we wish to learn that lack a best approximation within a neural network set, we must expect the weights of the approximating neural networks to grow to infinity.
This can be undesirable because, as we will see in the following sections on generalization, a bounded parameter space facilitates many generalization bounds. 

\section*{Bibliography and further reading}

The properties of neural network sets were first studied with a focus on the continuous approximation property in \cite{kainen1999approximation, kainen2003best, kainen2001continuity} and \cite{girosi1990networks}. 
The results in \cite{kainen1999approximation, kainen2001continuity, kainen2003best} already use the non-convexity of sets of shallow neural networks. 
The results on convexity and closedness presented in this chapter follow mostly the arguments of \cite{petersen2021topological}. %
Similar results were also derived for other norms in \cite{mahan2021nonclosedness}.

\newpage
\section*{Exercises}
\begin{exercise}\label{ex:proofOfPropositionStarShaped}
Prove Proposition \ref{prop:NNsAreStarShaped}.
\end{exercise}

\begin{exercise}\label{ex:extendToArbitraryWidth}
  Extend Proposition \ref{prop:non-closednessSpecificArchitecture} to 
  $\CA=(d_0,d_1,1)$
  for arbitrary $d_0$, $d_1 \in \N$, $d_1\geq 2$. 
\end{exercise}

\begin{exercise}\label{ex:extendToArbitraryDepth}
Use Proposition \ref{prop:Identity1}, to extend Proposition \ref{prop:non-closednessSpecificArchitecture} to arbitrary depth. 
\end{exercise}

\begin{exercise}\label{eq:noBestApproxYieldsExplodingWeights}
Extend Proposition \ref{prop:ExplodingWeights} to functions $f$ for which there is no best-approximation in $\mathcal{N}( \sigma; \CA,\infty)$.
To do this, replace \eqref{eq:ThisEquationWillBeReferencedInOneOfTheExercises} by 
\[
	\| \Phi_n - f \|_{L^2}^2 \to \inf_{\Phi \in \mathcal{N}( \sigma; \CA,\infty)} \| \Phi - f \|_{L^2}^2.
\]

\end{exercise}

%% file: GeneralizationOfNNs.tex
\chapter{Generalization properties of deep neural
  networks}\label{chap:VC}
As discussed in the introduction in Section \ref{sec:highlevOverview},
we generally learn based on a finite data set. For example, given data
$(x_i,y_i)_{i=1}^m$, we try to find a network $\Phi$ that
(approximately) satisfies $\Phi(x_i)=y_i$ for $i=1,\dots,m$. The field
of generalization is concerned with how well such $\Phi$ performs on
\emph{unseen} data, which refers to any $x$ outside of training data
$\{x_1,\dots,x_m\}$. In this chapter we discuss generalization through
the use of covering numbers.

In Sections \ref{sec:LearningSetup} and \ref{sec:ERM} we revisit and
formalize the general setup of learning and empirical risk
minimization in a general context. Although some notions introduced in
these sections have already appeared in the previous chapters, we
reintroduce them here for a more coherent presentation. In Sections
\ref{sec:GenBounds}-\ref{sec:estimateOfCoveringNumbers}, we first
discuss the concepts of generalization bounds and covering numbers,
and then apply these arguments specifically to neural networks. In
Section \ref{sec:ApproxComplTradeOff} we explore the so-called
\emph{approximation-complexity trade-off}, and finally in Sections
\ref{sec:PACVC}-\ref{sec:VCLowerBoundsOnApprox} we introduce the
\emph{VC dimension} and give some implications for classes of neural
networks.

\section{Learning setup}\label{sec:LearningSetup}

A general learning problem \cite{mohri2018foundations, understanding,
  cucker2002mathematical} requires a \textbf{feature space} ${X}$ and
a \textbf{label space} ${Y}$, which we assume throughout to be
measurable spaces. We observe joint data pairs
$(x_i,y_i)_{i=1}^m\subseteq {X}\times{Y}$, and aim to identify a
connection between the $x$ and $y$ variables.  Specifically, we assume
a relationship between features $x$ and labels $y$ modeled by a
probability distribution $\CD$ over ${X} \times {Y}$, that generated
the observed data $(x_i,y_i)_{i=1}^m$.  While this distribution is
unknown, our goal is to extract %
information %
from it, so that we can make possibly good predictions of $y$ for a
given $x$.  Importantly, the relationship between $x$ and $y$ need not
be deterministic.

To make these concepts more concrete, we next present an example that
will serve %
to explain ideas throughout this chapter. This example is of high
relevance for many mathematicians, as ensuring a steady supply of
high-quality coffee is essential for maximizing the output of our
mathematical work.

		\begin{figure}[h]
                  \centering \includegraphics[width =
                  0.9\textwidth]{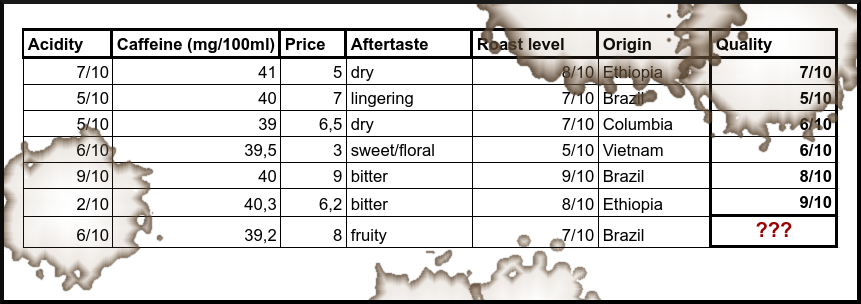}
                  \caption{Collection of coffee data.  The last row
                    lacks a ``Quality'' label.  Our aim is to predict
                    the label without the need for an (expensive)
                    taste test.  }
                  \label{fig:CoffeeDB}
		\end{figure}

	\begin{example}[Coffee Quality]\label{ex:coffee}%
          Our goal is to determine the quality of different
          coffees. To this end we model the quality as a number in
          \begin{equation*}
            Y = \Big\{\frac{0}{10},\dots,\frac{10}{10}\Big\},
          \end{equation*}
          with higher numbers indicating better quality.  Let us
          assume that our subjective assessment of quality of coffee
          is related to six features: ``Acidity'', ``Caffeine
          content'', ``Price'', ``Aftertaste'', ``Roast level'', and
          ``Origin''. The feature space $X$ thus corresponds to the
          set of six-tuples describing these attributes, which can be
          either numeric or categorical (see Figure
          \ref{fig:CoffeeDB}).

          We aim to understand the relationship between elements of
          $X$ and elements of $Y$, but we can neither afford, nor do
          we have the time to taste all the coffees in the
          world. Instead, we can sample some coffees, taste them, and
          grow our database accordingly as depicted in Figure
          \ref{fig:CoffeeDB}. This way we obtain samples of pairs in
          $X\times Y$. The distribution $\CD$ from which they are
          drawn depends on various external factors. For instance, we
          might have avoided particularly cheap coffees, believing
          them to be inferior. As a result they do not occur in our
          database. Moreover, if a colleague contributes to our
          database, he might have tried the same brand and arrived at
          a different rating. In this case, the quality label is not
          deterministic anymore.

          Based on our database, we wish to predict the quality of an
          untasted coffee. Before proceeding, we first formalize what
          it means to be a ``good'' prediction.
        \end{example}

        Characterizing how good a predictor is requires a notion of
        discrepancy in the label space. %
        This is the purpose of the so-called \textbf{loss function},
        which is a measurable mapping
        $\mathcal{L} \colon {Y} \times {Y} \to \R_+$.
	\begin{definition}\label{def:genError}
          Let $\mathcal{L} \colon {Y} \times {Y} \to \R_+$ be a loss
          function and let $\CD$ be a distribution on
          ${X} \times {Y}$.  For a measurable function
          $h \colon {X} \to {Y}$ we call
          \begin{align*}
            \risk(h) = \bbE_{(x,y) \sim \CD}\left[\mathcal{L}(h(x), y) \right]
          \end{align*}
          the \textbf{(population) risk of $h$}.
	\end{definition}
	Based on the risk, we can now formalize what we consider a
        good predictor.  The best predictor is one such that its risk
        is as close as possible to the smallest that any function can
        achieve.  More precisely, we would like a risk that is close
        to the so-called \textbf{Bayes risk}
	\begin{align}\label{eq:BayesRiskGenSection}
          R^* \coloneqq \inf_{h\colon {X} \to {Y}} \risk(h),
	\end{align}
        where the infimum is taken over all %
        measurable $h:X\to Y$. %

	\begin{example}[Loss functions]%
          The choice of a loss function $\mathcal{L}$ %
          usually depends on the application.  For a regression
          problem, i.e., a learning problem where ${Y}$ is a
          non-discrete subset of a Euclidean space, a common choice is
          the square loss $\mathcal{L}_2(\By,\By') = \|\By-\By'\|^2$.
	
          For binary classification problems, i.e.\ when ${Y}$ is a
          discrete set of cardinality two, the ``$0-1$ loss''
          \begin{equation*}
            \mathcal{L}_{0-1}(y,y') = \begin{cases}
                                        1 & y\neq y'\\
                                        0 & y= y'
                                      \end{cases}
                                    \end{equation*}
                                    seems more natural.

                                    Another frequently used loss for
                                    binary classification
                                    ($Y=\{0,1\}$), especially when the
                                    hypothesis returns probabilities
                                    (i.e., $h:X\to [0,1]$) is the
                                    binary cross-entropy loss
                                    \begin{equation*}
                                      \mathcal{L}_{\rm ce}(y, y') = -(y \log(y') + (1 - y) \log(1 - y')).
                                    \end{equation*}
                                    In contrast to the $0-1$ loss, the
                                    cross-entropy loss is
                                    differentiable, which is desirable
                                    in deep learning as we saw in
                                    Chapter \ref{chap:training}.

                                    In the coffee quality prediction
                                    problem, the quality is given as a
                                    fraction of the form $k/10$ for
                                    $k = 0, \dots, 10$. While this is
                                    a discrete set, it makes %
                                    sense to more heavily penalize
                                    predictions that are wrong by a
                                    larger amount. For example,
                                    predicting $4/10$ instead of
                                    $8/10$ should produce a higher
                                    loss than predicting
                                    $7/10$. Hence, we would not use
                                    the $0-1$ loss %
                                    but, for example, the square
                                    loss. %
                                  \end{example}
        
                                  How do we find a function
                                  $h\colon {X} \to {Y}$ with a risk
                                  that is as close as possible to the
                                  Bayes risk? We will introduce a
                                  procedure to tackle this task in the
                                  next section.
	
                                  \section{Empirical risk
                                    minimization}\label{sec:ERM}
	
                                  Finding a minimizer of the risk
                                  constitutes a considerable
                                  challenge.  First, we cannot search
                                  through all measurable functions.
                                  Therefore, we need to restrict
                                  ourselves to a %
                                  specific set
                                  \begin{equation*}
                                    \mathcal{H} \subseteq \set{h:{X}\to{Y}}{h\text{ is measurable}}
                                  \end{equation*}
                                  called the \textbf{hypothesis set}.
                                  In the following, this set will %
                                  be some set of neural networks.
                                  Second, we are faced with the
                                  problem that we cannot evaluate
                                  $\risk(h)$ for non-trivial loss
                                  functions, because %
                                  the distribution $\CD$ is %
                                  typically unknown so that
                                  expectations with respect to $\CD$
                                  cannot be computed.  To approximate
                                  the risk, we will assume access to
                                  an i.i.d.\ sample of $m$
                                  observations drawn from $\CD$.  This
                                  is precisely the situation described
                                  in the coffee quality example of
                                  Figure \ref{fig:CoffeeDB}, where %
                                  $m=6$ coffees were
                                  sampled.\footnote{In practice, the
                                    assumption of independence of the
                                    samples is often unclear and
                                    typically not satisfied. For
                                    instance, the selection of the six
                                    previously tested coffees might be
                                    influenced by external factors
                                    such as personal preferences or
                                    availability at the local store,
                                    which introduce bias into the
                                    dataset.  } For a given hypothesis
                                  $h$ we can then check how well it
                                  performs on our sampled data.
                                  \begin{definition}\label{def:erisk}
                                    Let $m\in \N$, let
                                    $\mathcal{L} \colon {Y} \times {Y}
                                    \to \R$ be a loss function and let
                                    $S = (x_i, y_i)_{i=1}^m \in ({X}
                                    \times {Y})^m$ be a sample.  For
                                    $h \colon {X} \to {Y}$, we call
                                    \begin{align*}
                                      \widehat{\risk}_S(h) = \frac{1}{m}\sum_{i=1}^m \mathcal{L}(h(x_i), y_i)
                                    \end{align*}
                                    the \textbf{empirical risk} of
                                    $h$.
                                  \end{definition}
                                  If the sample $S$ is drawn i.i.d.\
                                  according to $\CD$, then we
                                  immediately see from the linearity
                                  of the expected value that
                                  $\widehat{\risk}_S(h)$ is an
                                  unbiased estimator of $\risk(h)$,
                                  i.e.,
                                  $\bbE_{S \sim
                                    \CD^m}[\widehat{\risk}_S(h)] =
                                  \risk(h)$.  Moreover, the weak law
                                  of large numbers states that the
                                  sample mean of an i.i.d.\ sequence
                                  of integrable random variables
                                  converges to the expected value in
                                  probability.  Hence, there is some
                                  hope that, at least for large
                                  $m \in \N$, minimizing the empirical
                                  risk instead of the %
                                  population risk might lead to a good
                                  hypothesis.  We formalize this
                                  approach in the next definition.
                                  \begin{definition}\label{def:empiricalRiskMinimizer}
                                    Let
                                    $\mathcal{H} \subseteq \{h \colon
                                    {X} \to {Y}\}$ be a hypothesis
                                    set.  Let $m\in \N$, let
                                    $\mathcal{L} \colon {Y} \times {Y}
                                    \to \R_+$ be a loss function and
                                    let
                                    $S = (x_i, y_i)_{i=1}^m \in ({X}
                                    \times {Y})^m$ be a sample.  We
                                    call a function $h_S$ such that
                                    \begin{align*}
                                      \widehat{\risk}_S(h_S) = \inf_{h \in \mathcal{H}}\widehat{\risk}_S(h)
                                    \end{align*}
                                    an \textbf{empirical risk
                                      minimizer}.
                                  \end{definition}
                                  From a generalization perspective,
                                  supervised deep learning is
                                  empirical risk minimization over
                                  sets of neural networks.  The
                                  question we want to address next is
                                  how effective this approach is at
                                  producing hypotheses that achieve a
                                  risk close to the Bayes risk.

                                  Let $\mathcal{H}$ be some hypothesis
                                  set, such that an empirical risk
                                  minimizer $h_S$ exists for all %
                                  $S\in ({X}\times{Y})^m$; see
                                  Exercise
                                  \ref{ex:existenceOfERminimiser} for
                                  an explanation of why this is a
                                  reasonable assumption. Moreover, let
                                  $g \in \mathcal{H}$ be arbitrary. %
                                  Then
                                  \begin{align}
                                    \risk(h_S) - R^* &= \risk(h_S)-\widehat{\risk}_S(h_S) + \widehat{\risk}_S(h_S) - R^* \nonumber\\
                                                     &\leq |\risk(h_S)-\widehat{\risk}_S(h_S)| + \widehat{\risk}_S(g) - R^*\label{eq:errorDec123}\\
                                                     &\leq 2 \sup_{h\in \mathcal{H}}|\risk(h)-\widehat{\risk}_S(h)| + \risk(g) - R^*,\nonumber
                                  \end{align}
                                  where in the first inequality we
                                  used that $h_S$ is the empirical
                                  risk minimizer.  By taking the
                                  infimum over all $g$, we conclude
                                  that
                                  \begin{align}
                                    \risk(h_S) - R^* &\leq 2 \sup_{h\in \mathcal{H}}|\risk(h)-\widehat{\risk}_S(h)| + \inf_{g \in \mathcal{H}}\risk(g) - R^* \nonumber \\
                                                     &\eqqcolon 2\eps_{\mathrm{gen}} + \eps_{\mathrm{approx}}.\label{eq:biasVarianceTradeOff}
                                  \end{align}
                                  Similarly, considering only
                                  \eqref{eq:errorDec123}, yields that
                                  \begin{align}
                                    \risk(h_S) &\leq \sup_{h\in \mathcal{H}}|\risk(h)-\widehat{\risk}_S(h)| + \inf_{g \in \mathcal{H}}\widehat{\risk}_S(g) \nonumber \\
                                               &\eqqcolon \eps_{\mathrm{gen}} + \eps_{\mathrm{int}}.\label{eq:interpolationVarianceTradeOff}
                                  \end{align}
	
                                  How to choose $\mathcal{H}$ to
                                  reduce the \textbf{approximation
                                    error} $\eps_{\mathrm{approx}}$ or
                                  the \textbf{interpolation error}
                                  $ \eps_{\mathrm{int}}$ was discussed
                                  at length in the previous chapters.
                                  The final piece is to figure out how
                                  to bound the \textbf{generalization
                                    error}
                                  $\sup_{h\in
                                    \mathcal{H}}|\risk(h)-\widehat{\risk}_S(h)|$.
                                  This will be discussed in the
                                  sections below.

                                  \section{Generalization
                                    bounds}\label{sec:GenBounds}

                                  We have seen that one aspect of
                                  successful learning is to bound the
                                  generalization error
                                  $\eps_{\rm gen}$ in
                                  \eqref{eq:biasVarianceTradeOff}.
                                  Let us first formally describe this
                                  problem.
                                  \begin{definition}[Generalization
                                    bound]\label{def:GenBound}
                                    Let
                                    $\mathcal{H} \subseteq \{h \colon
                                    {X} \to {Y} \}$ be a hypothesis
                                    set, and let
                                    $\mathcal{L} \colon {Y} \times {Y}
                                    \to \R_+$ be a loss function.  Let
                                    $\kappa \colon (0,1) \times \N \to
                                    \R_+$ be such that for every
                                    $\delta \in (0,1)$ holds
                                    $\kappa(\delta, m) \to 0$ for
                                    $m \to \infty$. %
                                    We call $\kappa$ a
                                    \textbf{generalization bound for
                                      $\mathcal{H}$} if for every
                                    distribution $\CD$ on
                                    ${X} \times {Y}$, every $m \in \N$
                                    and every $\delta \in (0,1)$, it
                                    holds with probability at least
                                    $1-\delta$ over the random sample
                                    $S \sim \CD^m$ that
                                    \begin{align*}
                                      \sup_{h \in \mathcal{H}}|\risk(h) - \widehat{\risk}_S(h)| \leq \kappa(\delta, m).
                                    \end{align*}
                                  \end{definition}
                                  \begin{remark}
                                    For a generalization bound
                                    $\kappa$ it holds that
                                    \begin{align*}
                                      \bbP\left[ \left|\risk(h_S) - \widehat{\risk}_S(h_S)\right| \leq \eps \right] \geq 1-\delta
                                    \end{align*}
                                    as soon as $m$ is %
                                    so large that
                                    $\kappa(\delta, m) \leq \eps$.  If
                                    there exists an empirical risk
                                    minimizer $h_S$ such that
                                    $\widehat{\risk}_S(h_S) = 0$, %
                                    then with high probability the
                                    empirical risk minimizer will also
                                    have a small risk $\CR(h_S)$.
                                    Empirical risk minimization is
                                    often referred to as a ``PAC''
                                    algorithm, %
                                    which stands for \emph{probably
                                      ($\delta$) approximately correct
                                      ($\eps$)}.
                                  \end{remark}

                                  Definition \ref{def:GenBound}
                                  requires %
                                  the upper bound $\kappa$ on the
                                  discrepancy between the empirical
                                  risk and the risk to be independent
                                  from the distribution $\CD$.  Why
                                  should this be possible?  After all,
                                  we could have an underlying
                                  distribution that is not uniform and
                                  hence, certain data points could
                                  appear very rarely in the sample.
                                  As a result, it should be very hard
                                  to produce a correct prediction for
                                  such points.  At first sight, this
                                  suggests that non-uniform
                                  distributions should be much more
                                  challenging than uniform
                                  distributions.  This intuition is
                                  incorrect, as the following argument
                                  based on Example \ref{ex:coffee}
                                  demonstrates.

	\begin{example}[Generalization in the coffee quality problem]
          In Example \ref{ex:coffee}, the underlying distribution %
          describes both our process of choosing coffees and the
          relation between the attributes and the quality.  Suppose we
          do not enjoy drinking coffee that costs less than
          $1$\texteuro{}.  Consequently, we do not have a single
          sample of such coffee in the dataset, and therefore we have
          no chance %
          of learning the quality of cheap coffees.

          However, the absence of coffee samples costing less than
          1\texteuro{} in our dataset is due to our \emph{general
            avoidance} of such coffee. As a result, we run a low risk
          of incorrectly classifying the quality of a coffee that is
          cheaper than $1$\texteuro{}, since it is unlikely that we
          will choose such a coffee in the future.
        \end{example}

	To establish generalization bounds, we %
        use stochastic tools that guarantee that the empirical risk %
        converges to the true risk as the sample size increases.  This
        is typically %
        achieved through concentration inequalities.  One of the
        simplest and most well-known is Hoeffding's inequality, see
        Theorem \ref{thm:hoeffdings}. %
        We will now apply Hoeffding's inequality to obtain a first
        generalization bound.  This generalization bound is well-known
        and can be found in many textbooks on machine learning, e.g.,
        \cite{mohri2018foundations, understanding}.  Although the
        result %
        does not yet encompass neural networks, %
        it %
        forms the basis for a similar result applicable to neural
        networks, as we discuss subsequently.
	
\begin{proposition}[Finite hypothesis set]\label{prop:finitehypothesis}
  Let $\mathcal{H} \subseteq \{h \colon {X} \mapsto {Y} \}$ be a
  finite hypothesis set.  Let
  $\mathcal{L} \colon {Y} \times {Y} \to \R$ be such that
  $\mathcal{L}( {Y} \times {Y}) \subseteq [c_1,c_2]$ with
  $c_2 - c_1 = C>0$.
		
  Then, for every $m \in \N$ and every distribution $\CD$ on
  ${X} \times {Y}$ it holds with probability at least $1-\delta$ over
  the %
  sample $S \sim \CD^m$ that
  \begin{align*}
    \sup_{h \in \mathcal{H}}|\risk(h) - \widehat{\risk}_S(h)| \leq C\sqrt{\frac{\log(|\mathcal{H}|) + \log(2/\delta)}{2m}}.
  \end{align*}  	
\end{proposition}
\begin{proof}
  Let $\mathcal{H} = \{h_1, \dots, h_n\}$.  Then it holds by a union
  bound that
  \begin{align*}
    \bbP\left[\exists h_i \in \mathcal{H} \colon |\risk(h_i) - \widehat{\risk}_S(h_i)| > \eps\right] \leq \sum_{i=1}^n \bbP\left[|\risk(h_i) - \widehat{\risk}_S(h_i)| > \eps\right].
  \end{align*}
  Note that %
  $\widehat{\risk}_S(h_i)$ is the mean of independent random variables
  which take their values almost surely in $[c_1,c_2]$.  Additionally,
  $\risk(h_i)$ is the expectation of $\widehat{\risk}_S(h_i)$.  The
  proof can therefore be finished by applying Theorem
  \ref{thm:hoeffdings}.  This will be %
  addressed in Exercise \ref{ex:finishProofFiniteHyp}.
\end{proof}

Consider now a \emph{non-finite} set of neural networks $\CH$, and
assume that it can be covered by a \emph{finite} set of (small)
balls. Applying Proposition \ref{prop:finitehypothesis} to the centers
of these balls, then allows to derive a similar bound as in the
proposition for $\CH$. This intuitive argument will be made rigorous
in the following section.
	
\section{Generalization bounds from covering
  numbers}\label{sec:GenBoundsCov}

To derive a generalization bound for classes of neural networks, we
start by introducing the notion of covering numbers.
\begin{definition}\label{def:coveringNumber}
  Let $A$ be a relatively compact subset of a metric space $(X, d)$.
  For $\eps >0$, we call
  \begin{align*}
    \mathcal{G}(A, \eps, (X,d)) \coloneqq \min\setc{n \in \N}{\exists\, (x_i)_{i=1}^n \subseteq X \text{ s.t.\ } \bigcup_{i=1}^n B_{\eps}(x_i) \supset A },
  \end{align*}
  where $B_\eps(x) = \set{ z \in X}{ d(z,x) \leq \eps}$, the
  $\eps$-covering number of $A$ in $X$. In case $X$ or $d$ are clear
  from context, we also write $\mathcal{G}(A, \eps, d)$ or
  $\mathcal{G}(A, \eps, X)$ instead of $\mathcal{G}(A, \eps, (X,d))$.
\end{definition}

\begin{figure}[htb]
  \centering \includegraphics[width =
  0.8\textwidth]{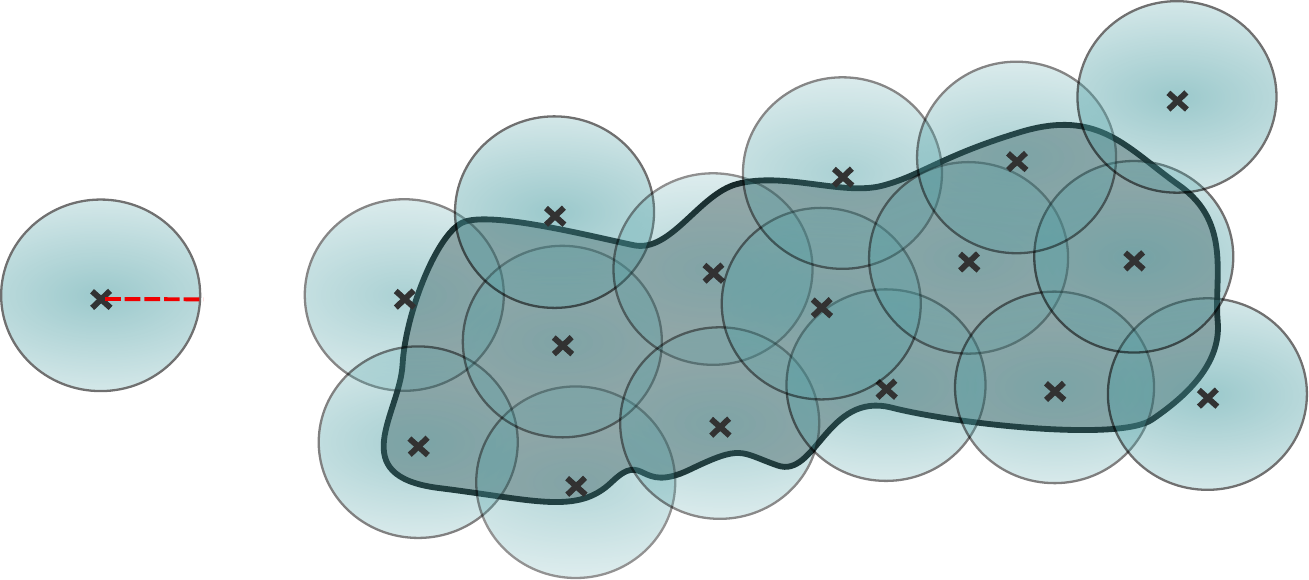} \put(-0.71\textwidth,
  0.15\textwidth){{\color{red}$\eps$}}
  \caption{Illustration of the concept of covering numbers of
    Definition \ref{def:coveringNumber}.  The %
    shaded set $A\subseteq\R^2$ is covered by sixteen Euclidean balls
    of radius $\eps$.  Therefore,
    $\mathcal{G}(A, \eps, \R^2) \leq 16$.}\label{fig:vizOfCovNum}
\end{figure}

A visualization of Definition \ref{def:coveringNumber} is given in
Figure \ref{fig:vizOfCovNum}.
As we will see, for Lipschitz continuous activation functions, it is
possible to upper bound the $\eps$-covering numbers of neural
networks %
as a subset of %
$C^0([0,1]^d)$ assuming the weights are confined to a fixed bounded
set.  The precise estimates are postponed to Section
\ref{sec:estimateOfCoveringNumbers}.  Before that, let us show how a
finite covering number facilitates a generalization bound.  We only
consider Euclidean feature spaces $X$ in the following result. %
A more general version could be easily derived.  For the statement,
recall that for $h:X\to\R$, $\norm[\infty]{h}\dfn \sup_{x\in X}|h(x)|$
denotes the supremum norm.

	\begin{theorem}\label{thm:coveringNumberGeneralizationBound} 
          Let $C_{Y}$, $C_{\CL} >0$ and $\alpha >0$.  Let
          ${Y} \subseteq [-C_{Y},C_{Y}]$, $X \subseteq \R^d$ for some
          $d \in \N$, and
          $\mathcal{H} \subseteq \{h \colon {X} \to {Y}\}$.  Further,
          let $\mathcal{L} \colon {Y} \times {Y} \to \R$ be
          $C_{\CL}$-Lipschitz in the first coordinate, i.e.\
          \begin{equation*}
            |\mathcal{L}(y,\tilde y)-\mathcal{L}(z,\tilde y)|\leq C_{\CL}|y-z| \qquad \text{for all } \tilde y,y,z \in Y.
          \end{equation*}

          Then, for every distribution $\CD$ on ${X} \times {Y}$ and
          every $m \in \N$ it holds with probability at least
          $1-\delta$ over %
          the sample $S \sim \CD^m$ that for all $h \in \mathcal{H}$
          \begin{align*}
            |\risk(h) - \widehat{\risk}_S(h)| \leq 4 C_{Y} C_{\CL} \sqrt{\frac{\log(\mathcal{G}(\mathcal{H}, m^{-\alpha}, %
            (X,\norm[\infty]{\cdot}))) + \log(2/\delta)}{m}} + \frac{2C_{\CL}}{m^{\alpha}}.
          \end{align*}  	
	\end{theorem}
	\begin{proof}
          Let
          \begin{equation}\label{eq:defMN}
            M = \mathcal{G}(\mathcal{H},m^{-\alpha}, (X,\norm[\infty]{\cdot}))
          \end{equation}
          and let $\CH_M = (h_i)_{i=1}^M\subseteq %
          \{h:X\to Y\}$ be such that for every $h \in \CH$ there
          exists $h_i \in \CH_M$ %
          with $\norm[\infty]{h - h_i} \leq m^{-\alpha}$. The
          existence of $\CH_M$ follows by Definition
          \ref{def:coveringNumber}.

          Fix for the moment such $h\in\CH$ and $h_i\in\CH_M$.  By the
          reverse and normal triangle inequalities, we have %
          \begin{align*}
            |\risk(h) - \widehat{\risk}_S(h) |- | \risk(h_i) - \widehat{\risk}_S(h_i)| &\leq |\risk(h) -   \risk(h_i) | + |\widehat{\risk}_S(h) - \widehat{\risk}_S(h_i)|.
          \end{align*}
          Moreover, from the monotonicity of the expected value and
          the Lipschitz property of $\mathcal{L}$ it follows that %
          \begin{align*}
            |\risk(h) -   \risk(h_i) | &\leq \bbE| \mathcal{L}(h(x), y) - \mathcal{L}(h_i(x), y) |\\
                                       & \leq C_{\CL} \bbE |h(x) - h_i(x)| \leq \frac{C_{\CL}}{m^{\alpha}}.
          \end{align*}
          A similar estimate yields
          $|\widehat{\risk}_S(h) - \widehat{\risk}_S(h_i)| \leq
          C_{\CL}/m^{\alpha}$.
        
          We thus conclude that for every $\eps>0$
          \begin{align}\label{eq:PSDm0}
            &\bbP_{S \sim \CD^m}\left[\exists h \in %
              \CH
              \colon |\risk(h) - \widehat{\risk}_S(h) |\geq \eps \right]\nonumber \\
            &\qquad \leq \bbP_{S \sim \CD^m}\left[\exists h_i \in \CH_M \colon |\risk(h_i) - \widehat{\risk}_S(h_i) |\geq \eps - \frac{2C_{\CL}}{m^{\alpha}} \right].
          \end{align}
          From Proposition \ref{prop:finitehypothesis}, we know that
          for $\eps>0$ and $\delta \in (0,1)$
          \begin{align}\label{eq:PSDm}
            \bbP_{S \sim \CD^m}\left[\exists h_i \in \CH_M\colon |\risk(h_i) - \widehat{\risk}_S(h_i) |\geq \eps - \frac{2C_{\CL}}{m^{\alpha}} \right] \leq \delta
          \end{align}
          as long as
          \[
            \eps - \frac{2C_{\CL}}{m^{\alpha}} > C \sqrt{\frac{\log(M)
                + \log(2/\delta)}{2m}},
          \]
          where $C$ is such that
          $\mathcal{L}({Y} \times {Y}) \subseteq [c_1,c_2]$ with
          $c_2 - c_1 \leq C$.  By the Lipschitz property of
          $\mathcal{L}$ we can choose $C = 2\sqrt{2} C_{\CL} C_{Y}$.
	
          Therefore, the definition of $M$ in \eqref{eq:defMN}
          together with \eqref{eq:PSDm0} and \eqref{eq:PSDm} give that
          with probability at least $1-\delta$ it holds for all
          $h\in\CH$
          \begin{align*}
            |\risk(h) - \widehat{\risk}_S(h)| \leq 2 \sqrt{2} C_{\CL}  C_{{Y}} \sqrt{\frac{\log(\mathcal{G}(\mathcal{H}, m^{-\alpha}, %
            (X,\norm[\infty]{\cdot})
            )) + \log(2/\delta)}{2m}} + \frac{2C_{\CL}}{m^{\alpha}}.
          \end{align*}
          This concludes the proof.
	\end{proof}

	\section{Covering numbers of deep neural
          networks}\label{sec:estimateOfCoveringNumbers}
	
	We have seen in Theorem
        \ref{thm:coveringNumberGeneralizationBound}, %
        that estimating %
        $\norm[\infty]{\cdot}$-covering numbers is crucial for
        understanding the generalization error. How can we determine
        these covering numbers?  The set of neural networks of a fixed
        architecture can be a quite complex set (see Chapter
        \ref{chap:shape}). Therefore it is not immediately clear how
        to cover it with balls, let alone identify the number of
        required balls.  The following lemma %
        suggest a simpler approach.
	\begin{lemma}\label{lemma:CoveringNumbersLipschitzProperty}
          Let ${X}_1$, ${X}_2$ be two metric spaces and let
          $f\colon {X}_1 \to {X}_2$ be $C_{\rm Lip}$-Lipschitz
          continuous.  For every relatively compact
          $A \subseteq {X}_1$ it then holds that for all $\eps>0$
          \begin{align*}
            \mathcal{G}(f(A), C_{\rm Lip} \eps, X_2) \leq \mathcal{G}(A, \eps, X_1).
          \end{align*} 
	\end{lemma}
	The proof of Lemma
        \ref{lemma:CoveringNumbersLipschitzProperty} is left as
        Exercise \ref{ex:CoveringNumbersLipschitzProperty}.  If we can
        represent the set of neural networks as the image under a
        Lipschitz map of another set %
        with known covering numbers, then Lemma
        \ref{lemma:CoveringNumbersLipschitzProperty} gives a direct
        way to bound the covering number of the neural network class.

\begin{figure}[htb]
          \centering \includegraphics[width =
          0.8\textwidth]{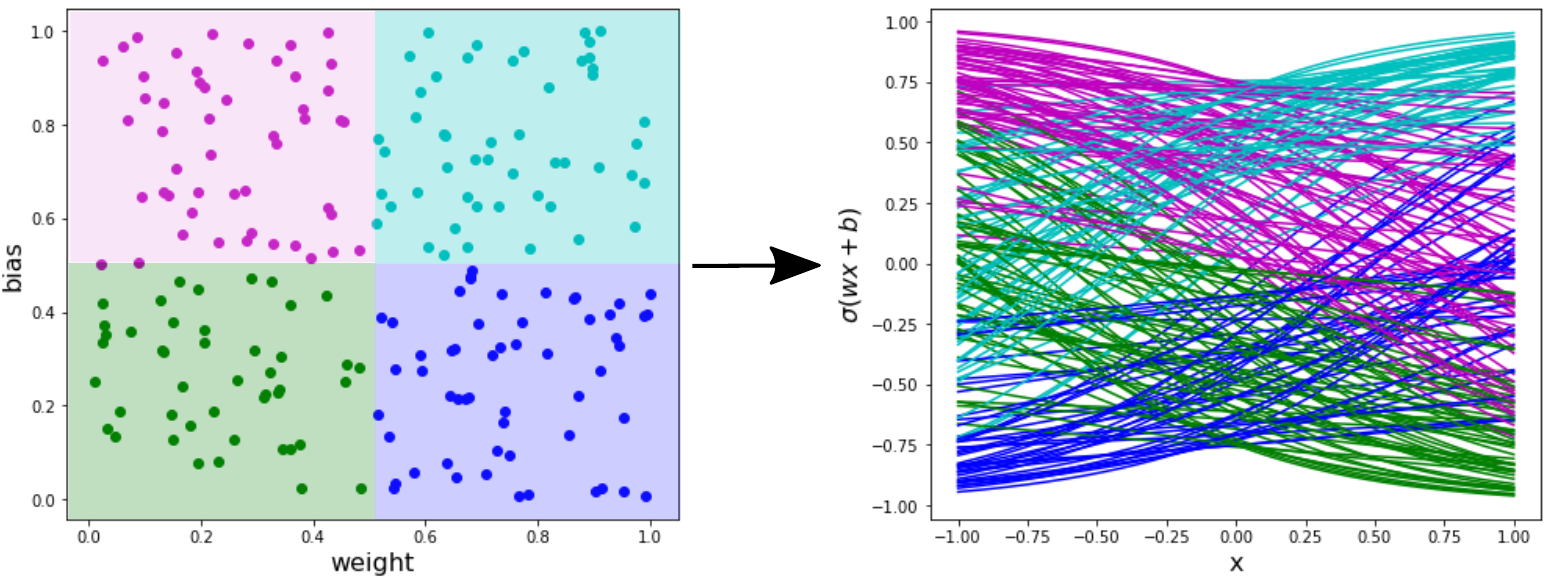} \put(-0.43\textwidth,
          0.18\textwidth){$R_\sigma$}
          \caption{Illustration of the main idea to deduce covering
            numbers of neural network spaces.  Points $\theta\in\R^2$
            in parameter space in the left figure correspond to
            functions $R_\sigma(\theta)$ %
            in the right %
            figure (with matching colors).  By Lemma
            \ref{lemma:CoveringNumbersLipschitzProperty}, a covering
            of the parameter space on the left translates to a
            covering of the function space on the right.  }
          \label{fig:CoveringNumberLipschitzIdea}
	\end{figure}

        Conveniently, we have already observed in Proposition
        \ref{prop:LipschitzOfRealizationMap}, that the set of neural
        networks is the image of the parameter range
        $\mathcal{PN}(\mathcal{A}, B)$ as in Definition
        \ref{def:realizationetc} under the Lipschitz continuous
        realization map $R_\sigma$.  It thus suffices to establish the
        $\eps$-covering number of $\mathcal{PN}(\mathcal{A}, B)$ or
        equivalently of $[-B,B]^{n_\CA}$.  Then, %
        we can apply Lemma
        \ref{lemma:CoveringNumbersLipschitzProperty} to find the
        covering numbers of $\CN(\sigma; \mathcal{A}, B)$.  This idea
        is depicted in Figure \ref{fig:CoveringNumberLipschitzIdea}.
	
	\begin{proposition}\label{prop:CoveringNumbers}
          Let $B$, $\eps >0$ and $s \in \N$. Then
          \[
            \mathcal{G}([-B,B]^{s}, \eps, (\R^{s},
            \norm[\infty]{\cdot})) \leq \lceil B/\eps \rceil^s.
          \]
	\end{proposition}
	\begin{proof}
          Let at first $s=1$. The interval $[-B,B]$ has length
          $2B$. Thus it can be covered by
          $\lceil 2B/(2\eps)\rceil=\lceil B/\eps\rceil$ closed
          intervals of length $2\eps$. Hence
          $\mathcal{G}([-B,B], \eps,\R) \leq \lceil B/\eps \rceil$.

          For $s>1$, the Cartesian products of these intervals cover
          $[-B,B]^s$. Since there are exactly $\lceil B/\eps\rceil^s$
          such Cartesian products, each of which corresponds to a ball
          of radius $\eps$ w.r.t.\ to the norm $\norm[\infty]{\cdot}$,
          we have
          $\mathcal{G}([-B,B]^s, \eps,\norm[\infty]{\cdot}) \leq
          \lceil B/\eps \rceil^s$.
	\end{proof}

	Having established a covering number for $[-B,B]^{s}$, we can
        now estimate the covering numbers of deep neural networks by
        combining Lemma \ref{lemma:CoveringNumbersLipschitzProperty}
        and Propositions \ref{prop:LipschitzOfRealizationMap} and
        \ref{prop:CoveringNumbers}. First we formalize the set of
        networks with $s$ trainable parameters (also see Definition
        \ref{def:realizationetc} for $\CN(\sigma;\CA,B)$). Therefore
        $s$ stands in the following for the size of these networks.

          \begin{definition}\label{def:Nsp}
            Let $B\ge 0$, $L$, $s$, $d_0$, $d_{L+1}\in\N$, and for
            $\ell=1,\dots,L$
            \begin{align*}
              \CA_\ell&\dfn (d_0,\underbrace{s,\dots,s}_{\text{$\ell$ times}},d_{L+1})\in\N^{\ell+2},\\
              n_{\CA_\ell}&\dfn(d_0+1)s + (\ell-1)s(s+1) + (s+1)d_{L+1}\in\N.
            \end{align*}
            Then
            \begin{align*}
              \CN^{\rm sp}(\sigma;L,B,s)\dfn
              \Big\{\Phi(\cdot,\Bw\odot\Bs)\;\Big|\;&\Phi\in\CN(\sigma;\CA_\ell,B)\text{ with }1\le\ell\le L,\\
                                                    &\Bw\in [-B,B]^{n_{\CA_\ell}},~\Bs\in\{0,1\}^{n_{\CA_\ell}},~\sum_{j=1}^{n_{\CA_\ell}}s_j=s\Big\}
            \end{align*}
            is the set of {\bf sparsely connected networks} of depth
            at most $L$.
          \end{definition}

          In words, $\CN^{\rm sp}(\sigma;L,B,s)$ denotes the set of
          feedforward neural networks with activation function
          $\sigma$, at least one hidden layer and at most $L$ hidden
          layers, at most $N$ nonzero weights, and all weights bounded
          in absolute value by $B$. The number $n_{\CA_\ell}$ is the
          maximum number of possible weights for the architecture
          $\CA_\ell$ in the definition.

      \begin{theorem}\label{thm:NNcover}
        Let $\sigma\colon \R \to \R$ be $C_\sigma$-Lipschitz
        continuous with $C_\sigma \geq 1$ and such that
        $|\sigma(x)|\le C_\sigma(1+|x|)$. Moreover let
        $\CN^{\rm sp}(\sigma;L,B,s)$ %
        be as in Definition \ref{def:Nsp} and assume that
        $N\ge\max\{d_0,d_{L+1}\}$.
        
        Then %
        for all $\eps>0$
        \begin{align*}
          \mathcal{G}\big(\CN^{\rm sp}(\sigma;L,B,s), \eps, %
          C^0([0,1]^{d_0})\big) \leq (s+1)^{7sL}\lceil 3C_\sigma B\rceil^{2sL}\lceil 1/\eps\rceil^s.
        \end{align*}
      \end{theorem}

\begin{proof}
  We use the notation from Definition \ref{def:Nsp}, and additionally
  let $S_\ell\dfn \set{\Bs\in\{0,1\}^{n_{\CA_\ell}}}{|\Bs|=s}$.  For
  $\Bs\in S_\ell$, denote by
  $\CN^{\rm sp}(\sigma; \mathcal{A}_\ell, B, \Bs)$ the neural networks
  in $\CN^{\rm sp}(\sigma; L, B, s)$ which have architecture
  $\CA_\ell$ and nonzero weights only at the positions where
  $s_j=1$. Then
$$
\CN^{\rm sp}(\sigma; L, B,s) = \bigcup_{\ell=1}^L \bigcup_{\Bs \in
  S_\ell} \CN^{\rm sp}(\sigma; \mathcal{A}_\ell, B, \Bs).
$$
Thus, for $\eps >0$, the covering number of
$\CN^{\rm sp}(\sigma;L,B,s)$ is bounded by
\begin{align}\label{eq:IwillKeepUpperBoundingThis12243}
  \sum_{\ell=1}^L \sum_{\Bs \in S_\ell} \mathcal{G}(\CN(\sigma; \mathcal{A}_\ell,  B, \Bs), \eps, C^0([0,1]^{d_0})).
\end{align}

Fix $1\le \ell\le L$. The set $\CN(\sigma; \mathcal{A}_\ell, B, \Bs)$
is the image of $[-B,B]^s$ under the realization map $R_\sigma$.  By
Proposition \ref{prop:LipschitzOfRealizationMap}, $R_\sigma$ is
$C_{\rm Lip,\ell}$-Lipschitz continuous with
\begin{equation*}
  C_{\rm Lip,\ell}\dfn (3C_\sigma B d_{\max})^\ell s
  \le (3C_\sigma B s)^\ell s.
\end{equation*}
For the inequality we used $d_0$, $d_{L+1}\le s$ so that
$d_{\max}\le s$.  Using Proposition \ref{prop:CoveringNumbers} and
Lemma \ref{lemma:CoveringNumbersLipschitzProperty} we find
\begin{equation*}
  \CG\big(\CN(\sigma; \mathcal{A}_\ell,  B, \Bs),\eps,C^0([0,1]^{d_0})\big)\le
  \Big\lceil\frac{B C_{\rm Lip,\ell}}{\eps} \Big\rceil^s
  \le \Big\lceil \frac{(3C_\sigma Bs)^{L+1}s}{\eps}\Big\rceil^s.
\end{equation*}
Next, the cardinality of $S_\ell$ is equal to
\begin{equation*}
  \binom{n_{\CA_\ell}}{s}
  \le n_{\CA_\ell}^s\le %
  ((\ell+1)s(s+1))^s
  \le (L+1)^s(s+1)^{2s}.
\end{equation*}
Therefore \eqref{eq:IwillKeepUpperBoundingThis12243} is bounded by
\begin{equation*}
  (L+1)^{s+1}(s+1)^{2s}
  \Big\lceil \frac{(3C_\sigma Bs)^{L+1}s}{\eps}\Big\rceil^s
  \le %
  (L+1)^{s+1}(s+1)^{2s+s(L+2)}
  \Big\lceil\frac{(3C_\sigma B)^{L+1}}{\eps} \Big\rceil^s.
\end{equation*}
Since $s$, $L\ge 1$ we have $L+1\le (s+1)^L$ so that
\begin{equation*}
  (L+1)^{s+1}(s+1)^{2s+s(L+2)}
  \le (s+1)^{L(s+1)+2s+s(L+2)}
  \le (s+1)^{7Ls}.
\end{equation*}
This concludes the proof.
\end{proof}

The bound in Theorem \ref{thm:NNcover} is rather crude, but it will
suffice in the following.

We end this section, by applying the previous theorem to the
generalization bound of Theorem
\ref{thm:coveringNumberGeneralizationBound} with $\alpha = 1/2$.  To %
simplify the analysis, we restrict the discussion to neural networks
with range $[-1,1]$.  To this end, denote
\begin{align}\label{eq:CNstar}
  \CN^{\rm sp,*}(\sigma; L, B,s) \dfn \setc{\Phi \in \CN( \sigma;L, B,s)}{\Phi(\Bx) \in [-1,1] \text{ for all } \Bx \in [0,1]^{d_0}}.
\end{align} %
Since
$\CN^{\rm sp,*}( \sigma; L, B,s) \subseteq \CN^{\rm
  sp}(\sigma;L,B,s)$, %
its covering number is bounded by that of
$\CN^{\rm sp}(\sigma; L, B,s)$. This yields the following result.
\begin{theorem}\label{thm:coveringNumberGeneralizationBoundForNNs}
  Let $C_{\CL} >0$ and let
  $\mathcal{L} \colon [-1,1] \times [-1,1] \to \R$ be
  $C_{\CL}$-Lipschitz continuous.  Let $L$, $d_0$, $d_{L+1}\in\N$,
  $B\ge 1$, let $\sigma\colon \R \to \R$ be $C_\sigma$-Lipschitz
  continuous with $C_\sigma \geq 1$, and
  $|\sigma(x)| \leq C_\sigma\cdot(1+|x|)$ for all $x \in \R$.
		
  Then, for every $m \in \N$, every distribution $\CD$ on
  $[0,1]^{d_0} \times [-1,1]$, and every $s\ge\max\{d_0,d_{L+1}\}$ it
  holds with probability at least $1-\delta$ over %
  $S \sim \CD^m$ that for all
  $\Phi \in \CN^{\rm sp,*}( \sigma;L, B,s)$
  \begin{align*}
    |\risk(\Phi) -\widehat{\risk}_S(\Phi)|\lesssim &C_{\CL} \sqrt{\frac{%
                                                     sL\log(C_\sigma Bs)+s\log(m^{1/2})+\log(2/\delta)
                                                     }{m}} + \frac{2 C_{\CL}}{\sqrt{m}},
  \end{align*}
  where the hidden constant is an absolute constant independent of all
  other quantities.
\end{theorem}

\section{The approximation-complexity
  trade-off}\label{sec:ApproxComplTradeOff}

We recall %
the decomposition of the error %
in \eqref{eq:biasVarianceTradeOff}
\begin{align*}
  \risk(h_S) - R^* &\leq 2\eps_{\mathrm{gen}} + \eps_{\mathrm{approx}},
\end{align*}
where $R^*$ is the Bayes risk defined in
\eqref{eq:BayesRiskGenSection}.  We make the following observations
about the approximation error $\eps_{\mathrm{approx}}$ and
generalization error $\eps_{\mathrm{gen}}$ in the context of neural
network based learning:
\begin{itemize}
\item \textit{Scaling of the generalization error:} %
  For the hypothesis class
  $\mathcal{H} = \CN^{{\rm sp},*}(\sigma; L, B,s)$, and for %
  sample of size $m\in \N$, if follows from Theorem \ref{thm:coveringNumberGeneralizationBoundForNNs} that the generalization error
  $\eps_{\mathrm{gen}}$ essentially scales like
  \[
    \eps_{\mathrm{gen}} \lesssim %
    \sqrt{\frac{sL\log(mBs)}{m}} + \sqrt{\frac{\log(2/\delta)}{m}}, %
  \]
  in terms of the network size $s$, network depth $L$, and upper bound
  $B$ on the weights; for each $m$ this bound holds with probability
  at least $1-\delta$.

\item \textit{Scaling of the approximation error:} Assume there exists %
  $h^*$ such that $\risk(h^*) = R^*$, and let the loss function
  $\mathcal{L}$ be $C_\CL$-Lipschitz continuous in the first
  coordinate. Then
  \begin{align*}
    \eps_{\mathrm{approx}} = \inf_{h \in \mathcal{H}} \risk(h) -\risk(h^*) &=  \inf_{h \in \mathcal{H}} \bbE_{(x,y) \sim \CD}[ \mathcal{L}(h(x), y) -  \mathcal{L}(h^*(x), y)  ]\\
                                                                           & \leq C_{\CL} \inf_{h \in \mathcal{H}}\|h - h^*\|_{%
                                                                             \infty}.
  \end{align*}
  We have seen in Chapters \ref{chap:ReLUNNs} and \ref{chap:DReLUNN}
  several results stating for networks of size $s$ an approximation
  error $s^{-\alpha}$ where the algebraic convergence rate $\alpha$
  depends on the smoothness of the functions and the underlying
  dimension; more precisely, this assumes that the depth $L$ and the
  upper bound $B$ increase suitably with $s$.
\end{itemize}

In this scenario, for an empirical risk minimizer $\Phi_S$ from
$\CN^{{\rm sp},*}(\sigma; L, B,s)$, it then holds the following: with
probability at least $1- \delta$
\begin{align}\label{eq:upperBoundForReLUNNs}
  \risk(\Phi_S) - R^* \lesssim %
  \underbrace{%
  \sqrt{\frac{sL\log(mBs)}{m}} + \sqrt{\frac{\log(2/\delta)}{m}}    
  }_{\text{generalization error}}
  + \underbrace{s^{-\alpha}}_{\text{approximation error}}.
\end{align}

\begin{figure}[htb]
  \centering \includegraphics[width =
  0.8\textwidth]{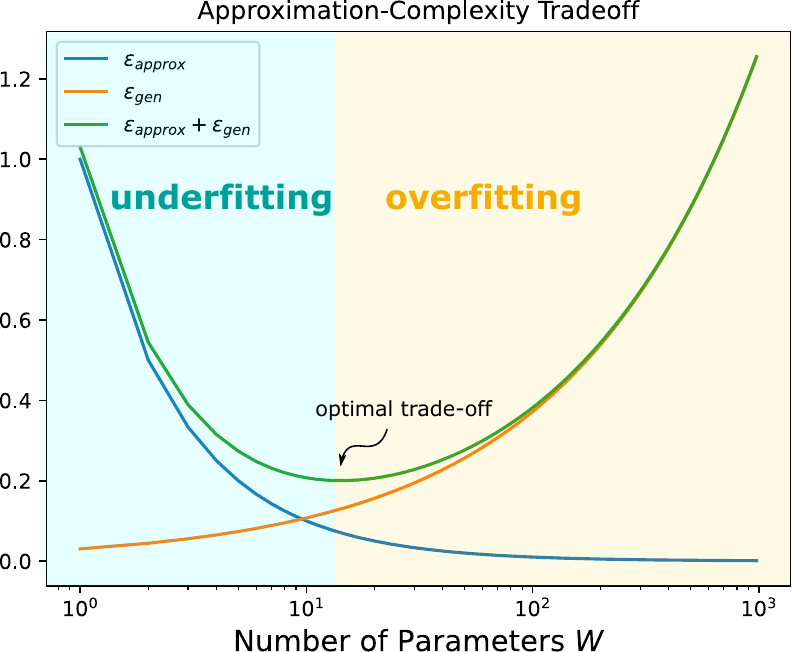}
  \caption{Illustration of the approximation-complexity-trade-off of
    Equation \eqref{eq:upperBoundForReLUNNs}.  Here we chose $r = 1$
    and $m = 10.000$, also all implicit constants are assumed to be
    equal to 1.  }
  \label{fig:classApprox-Comp-TradeOff}
\end{figure}

Increasing the network size $s$, depth $L$, and weight bound $B$ has
opposite effects on these terms: While the %
approximation error decreases, the term associated to generalization
increases.  This trade-off is known as
\textbf{approximation-complexity trade-off}.  The situation is
depicted in Figure \ref{fig:classApprox-Comp-TradeOff}.  The figure
and \eqref{eq:upperBoundForReLUNNs} suggest that, the perfect model
achieves the optimal trade-off between approximation and
generalization error.  Using this notion, we can %
categorize all models into three classes:
\begin{itemize}
\item \textit{Underfitting}: The model is not expressive enough for
  the data. The total error is high because the approximation error
  $\eps_{\rm approx}$ dominates the generalization error
  $\eps_{\rm gen}$. Increasing model complexity reduces the total
  error.
\item \textit{Optimal}: The model's complexity achieves balance
  between the approximation and the generalization error, and the sum
  of both error terms $\eps_{\rm approx}+\eps_{\rm gen}$ reaches a
  minimum. Either increasing or decreasing the model complexity
  increases the total error.
\item \textit{Overfitting:} The model is too expressive for the
  data. The total error is high because the generalization error
  $\eps_{\rm gen}$ dominates the approximation error
  $\eps_{\rm approx}$. While it may perfectly fit the data, the model
  fits noise or introduces unwanted features that do not generalize
  the true distribution. Reducing model complexity decreases the
  error.
\end{itemize}

In Chapter \ref{chap:GenOverparameterized}, we will see that deep
learning often operates in the regime where the number $s$ of
trainable parameters exceeds the optimal trade-off point. For certain
architectures used in practice, the number of parameters can be so
large that the theory of the approximation-complexity trade-off
suggests that learning should be impossible.  However, we emphasize,
that the present analysis only provides upper bounds. It does not
prove that learning is impossible or even impractical in the
overparameterized regime. Moreover, in Chapter \ref{chap:wideNets} we
have already seen indications in Section \ref{sec:gaussianprocesses}
that learning in the overparameterized regime need not necessarily
lead to large generalization errors.

\section{Risk bounds at the optimal
  trade-off}\label{risk-bound-trade-off}

\newcommand{\spaceparam}{N} %

Let us now identify the generalization bounds when the set of neural
networks is chosen according to the optimal trade-off of the previous
section.  We first make an abstract assumption on the approximability
of our underlying function class by neural networks.

Specifically we assume that the depth increases (at most)
logarithmically compared to the total number of network parameters,
and the error decays algebraically. Additionally, the network weights
are also allowed to increase algebraically. This is the typical
situation we encountered for our approximation results in the previous
chapters. %

\begin{assumption}\label{ass:assumptionOnApproximationSpaces}
  Let $d \in \N$, $c$, $\alpha$, $\beta > 0$, and let
  $\sigma\colon \R \to \R$. Let $\CC \subset \{h:[0,1]^d\to\R\}$ be
  such that for all $\spaceparam \in \N$ there exists
  $L_{\spaceparam}$, $B_{\spaceparam} >0$, %
  $s_{\spaceparam}\in\N$ such that
  \begin{enumerate}
  \item\label{ass:Approxbeta} $%
    s_{\spaceparam}\leq c \spaceparam \log(\spaceparam) , \qquad
    L_{\spaceparam} \le c\log(\spaceparam+1), \qquad B_{\spaceparam}
    \leq c \cdot \spaceparam^\beta$,
  \item\label{ass:Approxalpha} for all $h \in \CC$ there exists
    $\Phi_{h,\spaceparam} \in \CN^{{\rm sp},*}( \sigma;
    L_{\spaceparam}, B_{\spaceparam}, s_{\spaceparam})$ such that
	$$
	\norm[{L^{\infty}([0,1]^d)}]{h - \Phi_{h,\spaceparam}} \le c
        \spaceparam^{-\alpha}.
	$$
      \end{enumerate}
    \end{assumption}
    In the following we denote by $D_\CC$ the set of distributions for
    which the Bayes risk is attained for at least one $h\in\CC$.

    By \eqref{eq:upperBoundForReLUNNs} and under Assumption
    \ref{ass:assumptionOnApproximationSpaces}, with probability at
    least $1-\delta$ over a sample $S \sim \mathcal{D}^m$ for
    $\mathcal{D} \in D_\CC$, an empirical risk minimizer $\Phi_S$ from
    $\CN^*( \sigma; \mathcal{A}_{\spaceparam}, B_{\spaceparam})$
    satisfies
    \begin{align}\label{eq:findBestk}
      \risk(\Phi_S) - R^*\lesssim  \sqrt{\frac{{\spaceparam\log(\spaceparam)^3}\log(m)}{m}} +  \frac{1}{\spaceparam^\alpha} + \sqrt{\frac{\log(2/\delta)}{m}},
    \end{align}
    where we used $\log(Nm)\lesssim \log(N)\log(m)$ for $N$, $m\gg 1$
    and where the implicit constant depends on $\beta$.  For fixed
    sample size $m$, we wish to choose $\spaceparam$ optimal to
    minimize the right-hand side. %
    The last term is not affected by $\spaceparam$. For the first and
    second term, the ansatz $N=m^\gamma$ leads to
    \begin{equation}\label{eq:theTermsThatNeedToBeEquilibrated}
      \sqrt{\frac{{\spaceparam\log^{3}(\spaceparam)}\log(m)}{m}} = m^{(\gamma-1)/2} \log^{2}(m) \qquad\text{and}\qquad \frac{1}{\spaceparam^\alpha}= m^{-\alpha\gamma},
    \end{equation}
    for the two $N$-dependent terms. The asymptotic rate of the second
    term increases in $\gamma$, and the asymptotic rate of the first
    decreases in $\gamma$. We thus have to equilibrate them.
	
    For $\gamma = 1/(2\alpha + 1)$ and
    $\spaceparam = \spaceparam^* \coloneqq m^{1/(2\alpha + 1)}$ the
    terms \eqref{eq:theTermsThatNeedToBeEquilibrated} are both bounded
    by $c' m^{-\alpha/(2\alpha + 1)} \log^{2}(m)$ for a constant
    $c'>0$.  Clearly, for
    $\spaceparam > \spaceparam^* \cdot (c' \log^{2}(m))$ the first
    term in \eqref{eq:theTermsThatNeedToBeEquilibrated} exceeds
    $c m^{-\alpha/(2\alpha + 1)} \log^{2}(m)$.  Similarly, for
    $N < \spaceparam^*/ (c' \log^{2}(m))^{1/\alpha}$ the second term
    exceeds $c m^{-\alpha/(2\alpha + 1)} \log^{2}(m)$.  Therefore, we
    can conclude that, up to an at most logarithmic factor in $m$,
    $\spaceparam^*$ yields the best upper bound in
    \eqref{eq:findBestk}.  We summarize these observations in the
    following theorem.

\begin{theorem}\label{thm:optimalTradeOffGeneralization}
  Consider the setting of Theorem
  \ref{thm:coveringNumberGeneralizationBoundForNNs} and let Assumption
  \ref{ass:assumptionOnApproximationSpaces} be satisfied.
    Then for every $m \in \N$, every distribution $\CD \in D_{\CC}$, and
  every $\delta\in (0,1)$ it holds with probability at least
  $1-\delta$ over $S \sim \CD^m$ that an empirical risk
  minimizer\footnote{Here we implicitly assume that this empirical
    risk minimizer is measurable as a function of the sample $S$. Such
    statements can be made rigorous under certain assumptions (cf.,
    e.g., \cite[Prop.~5]{Nickl.2007}, \cite{MR1468737}), but we
    refrain from going into further detail.} %
  $\Phi_S \in \CN^{{\rm sp},*}( \sigma;
  L_{\spaceparam^*},B_{\spaceparam^*},s_{\spaceparam^*})$ where
  $\spaceparam^* = m^{1/(2\alpha +1)}$ satisfies
  \begin{align*}
    \risk(\Phi_S) -R^*\lesssim m^{- \frac{\alpha}{1+2\alpha}} \log^{2}(m) + \sqrt{\frac{\log(2/\delta)}{m}}.
  \end{align*}
\end{theorem}

\begin{remark}
  Let us now %
  discuss two results from this book to which Theorem
  \ref{thm:optimalTradeOffGeneralization} can be applied.
  \begin{itemize}
  \item \textit{$C^{k,s}$ functions (Theorem \ref{thm:Cks}):} For
    $\CC$ the unit ball in $C^{k,s}([0,1]^d)$, by Theorem
    \ref{thm:Cks}, Assumption
    \ref{ass:assumptionOnApproximationSpaces} \ref{ass:Approxbeta} is
    satisfied for ReLU neural networks %
    with $\alpha=(k+s)/d$; we did not explicitly derive upper bounds
    on the weights in Theorem \ref{thm:Cks}, but carefully checking
    the proof shows that they can be chosen such that Assumption
    \ref{ass:assumptionOnApproximationSpaces} holds for some $\beta$.
    Hence, it holds with probability at least $1-\delta$ that
	$$
	\risk(\Phi_S) \lesssim m^{- \frac{k+s}{2k +2s + d}}
        \log^{2}(m) + \sqrt{\frac{\log(2/\delta)}{m}}.
	$$       
      \item \textit{Barron functions (Theorem \ref{thm:BarronLight}):}
        The theorem only yields approximation rates in $L^2$ which is
        not sufficient for Assumption
        \ref{ass:assumptionOnApproximationSpaces}. However, the
        approximation result can be extended to $L^\infty$,
        \cite{barron1992neural}, which then yields that Assumption
        \ref{ass:assumptionOnApproximationSpaces} is satisfied with %
        $\alpha=1/2$ (and again some $\beta>0$). This implies the risk
        bound
	$$
	\risk(\Phi) \lesssim m^{- 1/4}\log^{2}(m) +
        \sqrt{\frac{\log(2/\delta)}{m}}
	$$
	which holds with probability $1-\delta$.
      \end{itemize}
    \end{remark}

\begin{remark}
  The rates established use only the Lipschitz property of the loss
  functions and are %
  not necessarily optimal for %
  specific losses. For example, for the square loss, %
  the rate can usually be doubled: \cite{schmidt2020nonparametric}
  establishes for $C^{k,s}$ regular functions an upper bound on the
  risk of the order of
  $m^{-{2}/((k+s)/d + 2)} = m^{-(2k+2s)/(d + 2k + 2s)}$ and
  \cite{barron1994approximation} demonstrates a risk decaying like
  $m^{-1/2}$ for Barron regular functions and appropriately chosen
  neural network spaces.
\end{remark}

\section{PAC learning from VC dimension}\label{sec:PACVC}

In addition to covering numbers, there are %
several other tools to analyze the generalization capacity of
hypothesis sets.  In the context of classification problems, one of
the most %
important is the so-called Vapnik--Chervonenkis (VC) dimension.
	
	\subsection{Definition and examples}
        Let $\CH$ be a hypothesis set of functions mapping from $\R^d$
        to $\{0,1\}$.  A set $S=\{\Bx_1,\dots,\Bx_n\} \subseteq \R^d$
        is said to be {\bf shattered} by $\CH$ if for every
        $(y_1,\dots,y_n) \in \{0,1\}^n$ there exists $h\in\CH$ such
        that $h(\Bx_j)=y_j$ for all %
        $j=1,\dots,n$.
	
	The VC dimension quantifies the complexity of a function class
        via the number of points that can %
        be shattered.
	
	\begin{definition}\label{def:VCDim}%
          The {\bf VC dimension} of $\CH$ is the cardinality of the
          largest set $S\subseteq\R^d$ that is shattered by $\CH$.  We
          denote the VC dimension by $\mathrm{VCdim}(\mathcal{H})$.
	\end{definition}
        
\begin{figure}
          \centering \includegraphics[width =
          \textwidth]{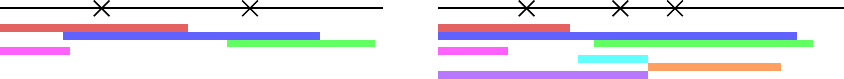}
          \caption{Different ways to classify two or three points.
            The colored-blocks correspond to intervals that produce
            different classifications of the points.}
          \label{fig:VCIntervals}
	\end{figure}

	\begin{example}[Intervals]
          Let $\mathcal{H} = \set{\ind_{[a,b]}}{a, b \in \R}$.  It is
          clear that $\mathrm{VCdim}(\mathcal{H}) \geq 2$ since for
          $x_1 < x_2$ the functions
          \[
            \ind_{[x_1-2,x_1-1]}, \quad \ind_{[x_1-{1},x_1]}, \quad
            \ind_{[x_1,x_2]}, \quad \ind_{[x_2,x_2+1]},
          \]
          are all different, when restricted to %
          $S=\{x_1,x_2\}$.
	
          On the other hand, if $x_1 < x_2 < x_3$ then, since
          $h^{-1}(\{1\})$ is an interval for all $h \in \mathcal{H}$,
          we have that $h(x_1) = 1 = h(x_3)$ implies $h(x_2) = 1$.
          Hence, no set of three elements can be shattered.
          Therefore, $\mathrm{VCdim}(\mathcal{H}) = 2$.  The situation
          is depicted in Figure \ref{fig:VCIntervals}.
	\end{example}

\begin{figure}[htb]
          \vspace{1cm}

          \centering \includegraphics[width =
          0.7\textwidth]{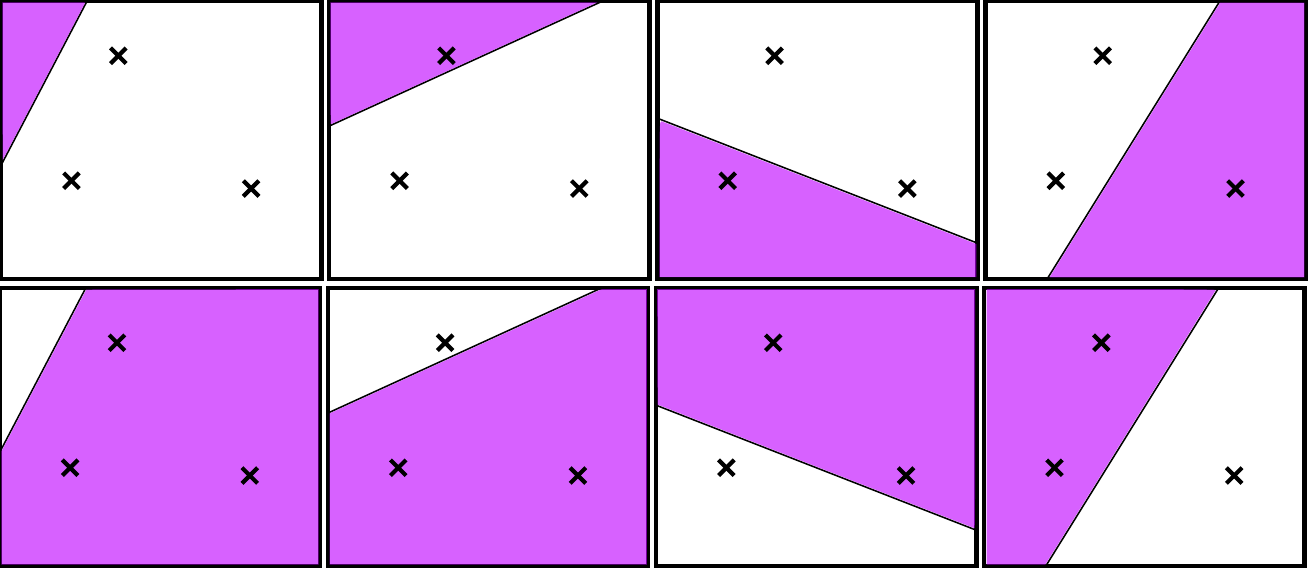}
          \caption{Different ways to classify three points by a
            half-space, \cite[Figure 1.4]{Scholkopf2002}.}
          \label{fig:VCHyperplanes}
	\end{figure}

\begin{example}[Half-spaces]\label{ex:half-spaces}        
  Let $\mathcal{H}_2 = \set{ \ind_{[0,\infty)}(\langle \Bw%
    , \cdot\rangle + b )}{ %
    \Bw\in \R^2, b \in \R}$ be a hypothesis set of rotated and shifted
  two-dimensional half-spaces.  In Figure \ref{fig:VCHyperplanes} we
  see that $\mathcal{H}_2$ shatters a set of three points.  More
  general, for $d\ge 2$ with
  \begin{align*}
    \mathcal{H}_d\dfn \set{\Bx\mapsto
    \ind_{[0,\infty)}(\Bw^\top\Bx+b)}{\Bw\in\R^d,~b\in\R}
  \end{align*}
  the VC dimension of $\CH_d$ equals $d+1$.
\end{example}

	In the example above, the VC dimension coincides with the
        number of parameters.  However, this is not true in general as
        the following example shows.
	
	\begin{example}[Infinite VC dimension]
          Let for $x\in\R$
          \begin{align*}
            \CH\dfn \set{x\mapsto
            \ind_{[0,\infty)}(\sin(wx))}{w\in\R}.
          \end{align*}
          Then the VC dimension of $\CH$ is infinite (Exercise
          \ref{ex:InfiniteVCDimension}).
	\end{example}

	\subsection{Generalization based on VC dimension}
	In the following, we consider a classification problem.
        Denote by $\CD$ the data-generating distribution on
        $\R^{d}\times \{0,1\}$.  Moreover, we let $\CH$ be a %
        set of functions from $\R^d\to \{0,1\}$.
	
        In the binary classification set-up, the natural choice of a
        loss function is the $0-1$ loss
        $\mathcal{L}_{0-1}(y,y') = \ind_{y \neq y'}$.  Thus, given a
        sample $S$, the empirical %
        risk of a function $h \in\CH$ is
	\begin{align*}
          \widehat{\CR}_S(h) = \frac{1}{m}\sum_{i=1}^{m}\ind_{h(\Bx_i)\neq y_i}.
	\end{align*}
	Moreover, the risk can be written as
	\begin{align*}
          \risk(h) =  \bbP_{(\Bx,y)\sim \CD}[h(\Bx)\neq y],
	\end{align*}
	i.e., the probability under $(\Bx,y)\sim \CD$ of $h$
        misclassifying the label $y$ of $\Bx$.
	
	We can now give a generalization bound in terms of the VC
        dimension of $\CH$, see, e.g., \cite[Corollary
        3.19]{mohri2018foundations}:

	\begin{theorem}\label{thm:gen}
          Let $d$, $k \in \N$ and
          $\CH \subseteq \{h\colon\R^d\to \{0,1\}\}$ have VC dimension
          $k$.  Let $\CD$ be a distribution on $\R^d\times \{0,1\}$.
          Then, for every %
          $\delta \in(0,1)$ and $m \in \N$ with $m \geq k$, it holds
          with probability at least $1-\delta$ over a sample
          $S \sim \CD^m$ that for every $h \in \CH$
          \begin{align}\label{eq:VCgenBound}
            |\risk(h)- \widehat{\risk}_S(h)| \leq \sqrt{\frac{2k\log(e m /k)}{m}} + \sqrt{\frac{\log(1/\delta)}{2m}},
          \end{align}
          where $e$ is the base of the natural logarithm.
	\end{theorem}
	In words, Theorem \ref{thm:gen} tells us that if a hypothesis
        class has finite VC dimension, then a hypothesis with a small
        empirical risk will have a small risk if the number of samples
        is large.  This shows that empirical risk minimization is a
        viable strategy in this scenario.  Will this approach also
        work if the VC dimension is not bounded?  No, in fact, in that
        case, no learning algorithm will succeed in reliably producing
        a hypothesis for which the risk is close to the best possible.
        We omit the technical proof of the following theorem from
        \cite[Theorem 3.23]{mohri2018foundations}.

        \begin{theorem}
          \label{thm:fundamentalTheoremOfLearningLowerBound}
          Let $k \in \N$ and let
          $\mathcal{H} \subseteq \{h \colon {X} \to \{0,1\}\}$ be a
          hypothesis set with VC dimension $k$.  Then, for every
          $m \in \N$ and every learning algorithm\footnote{The term
            learning algorithm refers here to any mapping
            $\mathrm{A} \colon ({X} \times \{0,1\})^m \to \mathcal{H}$
            such that $\risk\circ A$ is measurable.}
          $\mathrm{A} \colon ({X} \times \{0,1\})^m \to \mathcal{H}$
          there exists a distribution $\CD$ on ${X} \times \{0,1\}$
          such that
          \begin{align*}
            \bbP_{S \sim \CD^m} \left[ \risk(\mathrm{A}(S)) - \inf_{h \in \mathcal{H}} \risk(h) > \sqrt{\frac{k}{320 m}}  \right] \geq \frac{1}{64}.
          \end{align*}
	\end{theorem}
	Theorem \ref{thm:fundamentalTheoremOfLearningLowerBound}
        immediately implies the following statement for the
        generalization bound.
	\begin{corollary}\label{cor:NoLuckInOverParamRegime}
          Let $k \in \N$ and let
          $\mathcal{H} \subseteq \{h \colon {X} \to \{0,1\}\}$ be a
          hypothesis set with VC dimension $k$.  Then, for every
          $m \in \N$ there exists a distribution $\CD$ on
          ${X} \times \{0,1\}$ such that
          \begin{align*}
            \bbP_{S \sim \CD^m} \left[ \sup_{h\in \mathcal{H}}|\risk(h) - \widehat{\risk}_S(h)| > \sqrt{\frac{k}{1280 m}}  \right] \geq \frac{1}{64}.
          \end{align*}
	\end{corollary}
	\begin{proof}
          For a sample $S$, let $h_S\in\CH$ be an empirical risk
          minimizer, i.e.,
          $\widehat{\risk}_S(h_S) = \min_{h \in \mathcal{H}}
          \widehat{\risk}_S(h)$.  Let $\CD$ be the distribution of
          Theorem \ref{thm:fundamentalTheoremOfLearningLowerBound}.
          Moreover, for $\delta >0$, let $h_{\delta} \in \mathcal{H}$
          be such that
          \begin{align*}
            \risk(h_{\delta}) - \inf_{h \in \mathcal{H}} \risk(h) < \delta.
          \end{align*} 
          Then,
          \begin{align*}
            2\sup_{h \in \mathcal{H}}|\risk(h) - \widehat{\risk}_S(h)| &\geq
                                                                         |\risk(h_S) - \widehat{\risk}_S(h_S)| + |\risk(h_\delta) - \widehat{\risk}_S(h_\delta)|\\
                                                                       &\geq \risk(h_S) - \widehat{\risk}_S(h_S) + \widehat{\risk}_S(h_\delta) - \risk(h_\delta)\\
                                                                       &\geq \risk(h_S) - \risk(h_\delta)\\
                                                                       &>\risk(h_S) - \inf_{h \in \mathcal{H}} \risk(h) - \delta,
          \end{align*}
          where we used %
          $\widehat{\risk}_S(h_\delta)\ge \widehat{\risk}_S(h_S)$ for
          the third inequality.  The proof is completed by applying
          Theorem \ref{thm:fundamentalTheoremOfLearningLowerBound}
          with $\mathrm{A}(S) = h_S$ and using that $\delta>0$ was
          arbitrary.
	\end{proof}
	We have seen now, that we have a generalization bound scaling like %
        $O(1/\sqrt{m})$ for $m\to \infty$ if and only if the VC
        dimension of a hypothesis class is finite.  In more
        quantitative terms, we require the VC dimension of a neural
        network to be smaller than $m$.
		
        What does this imply for neural network functions? For ReLU
        neural networks there holds the following \cite[Theorem
        8.8]{MR1741038}.

        \begin{theorem}%
          \label{thm:ReLUVC}
          Let $\CA \in \N^{L+2}$, $L\in\N$ and set
          \begin{align*}
            \CH\dfn \set{\ind_{[0,\infty)}\circ \Phi}{\Phi\in\CN(\sigma_{\rm ReLU}; \mathcal{A}, \infty)}.
          \end{align*}
          Then, there exists a constant $C>0$ independent of $L$ and
          $\CA$ such that
          \begin{align*} {\rm VCdim}(\CH)\le C\cdot
            (n_{\mathcal{A}}L\log(n_{\mathcal{A}})+n_{\mathcal{A}}L^2).
          \end{align*}
        \end{theorem}
        The bound \eqref{eq:VCgenBound} is meaningful if %
        the sample size $m$ is significantly larger than the VC
        dimension $k$.  For ReLU neural networks as in Theorem
        \ref{thm:ReLUVC}, this means
        $m\gg n_{\mathcal{A}} L\log(n_{\mathcal{A}})+n_{\mathcal{A}}
        L^2$.  Fixing $L=1$ this amounts to
        $m\gg n_{\mathcal{A}} \log(n_{\mathcal{A}})$ for a shallow
        neural network with $n_{\mathcal{A}}$ parameters.  This is in
        contrast to what we assumed in Chapter \ref{chap:wideNets},
        where it was crucial that $n_{\mathcal{A}}$ is sufficiently
        large.  If the VC dimension of the neural network sets %
        scale like %
        $O(n_{\mathcal{A}}\log (n_{\mathcal{A}}))$, then Theorem
        \ref{thm:fundamentalTheoremOfLearningLowerBound} and Corollary
        \ref{cor:NoLuckInOverParamRegime} %
        indicate that, at least for %
        certain distributions, generalization should not be possible
        in the overparameterization regime $n_\CA\gg m$.  We will
        discuss the resolution of this potential paradox in Chapter
        \ref{chap:GenOverparameterized}.

        \section{Lower bounds on achievable approximation
          rates} \label{sec:VCLowerBoundsOnApprox} We %
        conclude this chapter on the complexities and generalization
        bounds of neural networks by using the established VC
        dimension bound of Theorem \ref{thm:ReLUVC} to deduce
        limitations to the approximation capacity of ReLU neural
        networks.  The result described below was %
        first given in \cite{yarotsky}. The proof is conceptually
        similar to the analysis in Section \ref{sec:cod}; the
        difference is that there is no assumption on continuous weight
        assignment, and we specify the argument to ReLU neural
        networks.

\begin{theorem}\label{thm:lowerBoundViaVC}
  Let $k$, $d\in\N$.  Assume that for every $\eps>0$ there exists
  $L_\eps\in\N$ and $\mathcal{A}_\eps$ with $L_\eps$ layers and input
  dimension $d$ such that
  \begin{align*}
    \sup_{\norm[{C^k([0,1]^d)}]{f}\le 1}
    \inf_{\Phi\in\CN(\sigma_{\rm ReLU}; \mathcal{A}_\epsilon, \infty)}\|f-\Phi\|_{C^0({[0,1]^d})}<\frac{\eps}{2}.
  \end{align*} %
  Then there exists $C>0$ solely depending on $k$ and $d$, such that
  for all $\eps\in (0,1)$
  \begin{align*}
    n_{\mathcal{A_\eps}} L_\eps\log(n_{\mathcal{A_\eps}}) +n_{\mathcal{A_\eps}} L_\eps^2
    \ge C \eps^{-\frac{d}{k}}.
  \end{align*}
\end{theorem}

\begin{proof}
  For $\Bx\in\R^d$ recall the ``bump function'' introduced in
  \eqref{eq:bumpfunction}
  \begin{align*}
    \psi(\Bx)\dfn
    \begin{cases}
      \exp\left(1-\frac{1}{1-\norm[2]{\Bx}^2}\right) &\text{if }\|\Bx\|_2<1\\
      0 &\text{otherwise.}
    \end{cases}
  \end{align*}
  For $\eps\in (0,1)$ we consider the scaled version
  \begin{align*}
    \tilde \psi_\eps(\Bx) \coloneqq \eps \tilde \psi\left(2\eps^{-1/k} \Bx\right).
  \end{align*}
  As in Lemma \ref{lemma:bumpfunction}, it holds for some constant
  $\tau_k$ and for all $\eps\in(0,1)$
  \[
    \supp(\tilde \psi_\eps)\subseteq
    \Big[-\frac{\eps^{-1/k}}{2},\frac{\eps^{-1/k}}{2}\Big]^d
    \qquad\text{and}\qquad \norm[C^k]{\tilde \psi_\eps}\le \tau_k.
  \]

  Consider the equispaced point set
  $\{\Bx_1,\dots,\Bx_{N(\eps)}\}= \eps^{1/k}\Z^d \cap [0,1]^d$.  The
  cardinality of this set is $N(\eps)\simeq \eps^{-d/k}$.  Given
  $\By\in\{0,1\}^{N(\eps)}$, let for $\Bx \in \R^d$
  \begin{align}\label{eq:fy}
    f_{\By}(\Bx) \coloneqq \tau_k^{-1} \sum_{j=1}^{N(\eps)} y_j \tilde \psi_{\eps}(\Bx-\Bx_j).
  \end{align}
  Then $f_{\By}(\Bx_j)= \tau_k^{-1} \eps y_j$ for all
  $j = 1, \dots, N(\eps)$ and $\norm[C^k]{f_{\By}}\le 1$.

  For every $\By\in\{0,1\}^{N(\eps)}$ let
  $\Phi_{\By} \in \CN(\sigma_{\rm ReLU};\mathcal{A}_{\tau_k^{-1}
    \eps},\infty)$ be such that
  \begin{equation*}
    \sup_{\Bx\in [0,1]^d}|f_{\By}(\Bx)-\Phi_{\By}	(\Bx)|<\frac{\eps}{2 \tau_k}.
  \end{equation*}
  Then
  \begin{equation*}
    \ind_{[0,\infty)}\Big(\Phi_y(\Bx_j)-\frac{\eps}{2 \tau_k}\Big)=y_j\qquad\text{for all }j = 1, \dots, N(\eps).
  \end{equation*}
  Hence, the VC dimension of
  $ \CN(\sigma_{\rm ReLU};\mathcal{A}_{\tau_k^{-1} \eps},\infty)$ is
  larger or equal to $N(\eps)$.  Theorem \ref{thm:ReLUVC} thus implies
  \begin{align*}
    N(\eps)\simeq\eps^{-\frac{d}{k}}\le C \cdot \Big(n_{\mathcal{A}_{\tau_k^{-1} \eps}} L_{\tau_k^{-1} \eps}\log(n_{\mathcal{A}_{\tau_k^{-1} \eps}})+n_{\mathcal{A}_{\tau_k^{-1} \eps}} L_{\tau_k^{-1} \eps}^2\Big)
  \end{align*}
  or equivalently
  \begin{align*}
    \tau_k^{-\frac{d}{k}}\eps^{-\frac{d}{k}}\le C \cdot \Big(n_{\mathcal{A}_{\eps}} L_{ \eps}\log(n_{\mathcal{A}_{\eps}})+n_{\mathcal{A}_{\eps}} L_{\eps}^2\Big).
  \end{align*} %
  This completes the proof.
\end{proof}

	\begin{figure}[htb]
          \centering \includegraphics[width =
          0.6\textwidth]{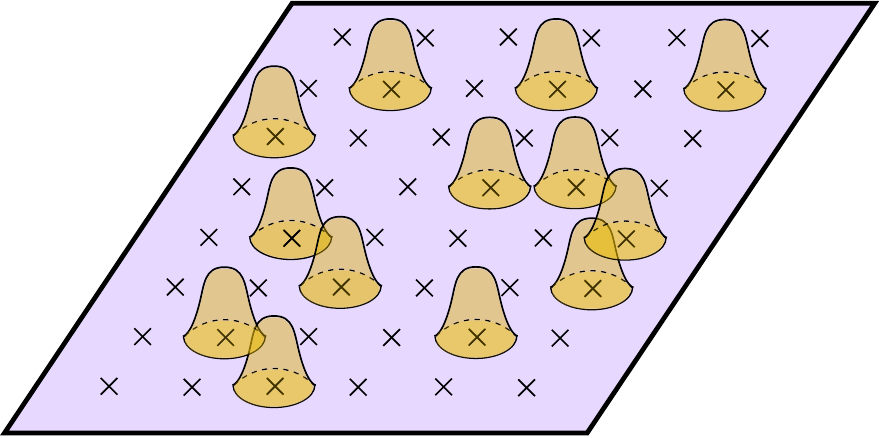}
          \caption{Illustration of $f_\By$ from Equation \eqref{eq:fy}
            on $[0,1]^2$.}
        \end{figure}

        To interpret Theorem \ref{thm:lowerBoundViaVC}, we consider
        two situations:
        \begin{enumerate}
        \item %
          {\bf depth increases at most logarithmically in $\eps$:} In
          this case, reaching uniform error $\eps$ for all
          $f\in C^k([0,1]^d)$ with $\|f\|_{C^k({[0,1]^d})}\le 1$
          requires
          \begin{align*}
	    n_{\mathcal{A}_\eps}\log(n_{\mathcal{A}_\eps})\log(\eps)+n_{\mathcal{A}_\eps}\log(\eps)^2
	    \ge C \eps^{-\frac{d}{k}}.
	  \end{align*}
          In terms of the neural network size, this (necessary)
          condition becomes
          $n_{\mathcal{A}_\eps}\ge C \eps^{-d/k}/\log(\eps)^2$.  As we
          have shown in Chapter \ref{chap:DReLUNN}, in particular
          Theorem \ref{thm:Cks}, up to log terms, %
          the network size $O(\eps^{-d/k})$ is also sufficient to
          achieve error $\eps$.  Hence, while the constructive proof
          of Theorem \ref{thm:Cks} might have seemed rather specific,
          under the assumption of the depth increasing at most
          logarithmically (which the construction in Chapter
          \ref{chap:DReLUNN} satisfies), it was essentially optimal
          (cf.~Remark \ref{rmk:ReLUessopt1}).
        \item {\bf depth is allowed to increase faster than
            logarithmically in $\eps$:} In this case the lower bound
          on the required neural network size improves.  Fixing for
          example $\CA_\eps$ with $L_\eps$ layers %
          such that $n_{\mathcal{A}_\eps} \le WL_\eps$ for some fixed
          $\eps$ independent $W\in\N$, the (necessary) condition on
          the depth becomes
          \begin{align*}
	    W\log(WL_\eps)L_\eps^2+W L_\eps^3
	    \ge C \eps^{-\frac{d}{k}}
	  \end{align*}
          and hence $L_\eps\gtrsim \eps^{-d/(3k)}$.

          We add that, for arbitrary depth the upper bound on the VC
          dimension of Theorem \ref{thm:ReLUVC} can be improved to
          $n_\mathcal{A}^2$, \cite[Theorem 8.6]{MR1741038}, and using
          this, would improve the just established lower bound to
          $L_\eps\gtrsim \eps^{-d/(2k)}$.

          For fixed width, this corresponds to neural networks of size
          $O(\eps^{-d/(2k)})$, which would mean twice the convergence
          rate proven in Theorem \ref{thm:Cks}.  Indeed, it turns out
          that neural networks can achieve this rate in terms of the
          neural network size \cite{NEURIPS2020_979a3f14}. However, by
          Theorem \ref{thm:codparam}, this is only attainable for
          discontinuous weight assignment. Therefore a stable (in the
          sense of continuously depending on the data) training
          algorithm achieving this rate does not exist.
        \end{enumerate}

        To sum up, in order to get error $\eps$ uniformly for all
        $f\in C^k([0,1]^d)$ with $\norm[{C^k([0,1]^d)}]{f}\le 1$, the
        size of a ReLU neural network is required to increase at least
        like $O(\eps^{-d/(2k)})$ as $\eps\to 0$; the best possible
        attainable convergence rate is therefore $2k/d$.  It has been
        proven, that this rate is also achievable, and thus the bound
        is sharp. Achieving this rate requires neural network
        architectures that grow faster in depth than in width.

\section*{Bibliography and further reading}
Classical statistical learning theory is based on the foundational
work of Vapnik and Chervonenkis \cite{vapnik2015uniform}. This led to
the formulation of the probably approximately correct (PAC) learning
model in \cite{valiant1984theory}, which is primarily utilized in this
chapter. A streamlined mathematical introduction to statistical
learning theory can be found in \cite{cucker2002mathematical}, and we
also refer to the recent textbook \cite{bach2025learning}.

Since statistical learning theory is well-established, there exists a
substantial amount of excellent expository work describing this
theory. Some highly recommended books on the topic are
\cite{mohri2018foundations, understanding, MR1741038}. The specific
approach of characterizing learning via covering numbers has been
discussed extensively in \cite[Chapter 14]{MR1741038}. Specific
results for ReLU activation used in this chapter were derived in
\cite{schmidt2020nonparametric, berner2020analysis}. The results of
Section \ref{sec:VCLowerBoundsOnApprox} describe some of the findings
in \cite{yarotsky, NEURIPS2020_979a3f14}. Other scenarios in which the
tightness of the upper bounds were shown are, for example, if
quantization of weights is assumed, \cite{bolcskei2019optimal,
  9363169, petersen2018optimal}, or when some form of continuity of
the approximation scheme is assumed as discussed in Section
\ref{sec:cod}.

\newpage
\section*{Exercises}

\begin{exercise}\label{ex:existenceOfERminimiser}
  Let $\mathcal{H}$ %
  be a set of neural networks with fixed architecture, where the
  weights are taken from a compact set.  Moreover, assume that the
  activation function is continuous.  Show that for every sample $S$
  there always exists an empirical risk minimizer $h_S$.
\end{exercise}
\begin{exercise}\label{ex:finishProofFiniteHyp}
  Complete the proof of Proposition \ref{prop:finitehypothesis}.
\end{exercise}
\begin{exercise} Prove Lemma
  \ref{lemma:CoveringNumbersLipschitzProperty}.
\end{exercise}
\begin{exercise}\label{ex:CoveringNumbersLipschitzProperty}
  Show that, the VC dimension of $\mathcal{H}$ of Example
  \ref{ex:half-spaces} is indeed 3, by demonstrating that no set of
  four points can be shattered by $\mathcal{H}$.
\end{exercise}
\begin{exercise}\label{ex:InfiniteVCDimension}
  Show that the VC dimension of
  \begin{align*}
    \CH\dfn \set{x\mapsto
    \ind_{[0,\infty)}(\sin(wx))}{w\in\R}
  \end{align*}
  is infinite.
	   	
\end{exercise}

%% file: GenInOverparameterization.tex
\chapter{Generalization in the overparameterized regime}\label{chap:GenOverparameterized}
In the previous chapter, we discussed the theory of generalization for deep neural networks trained by minimizing the empirical risk.
A key conclusion was that good generalization is possible as long as we choose an architecture that has a moderate number of neural network parameters relative to the number of training samples.
Moreover, we saw in Section \ref{sec:ApproxComplTradeOff} that the best performance can be expected when the neural network size is chosen to balance the generalization and approximation errors, by minimizing their sum.

\begin{figure}[htb]
	\centering
	\includegraphics[width=0.7\textwidth]{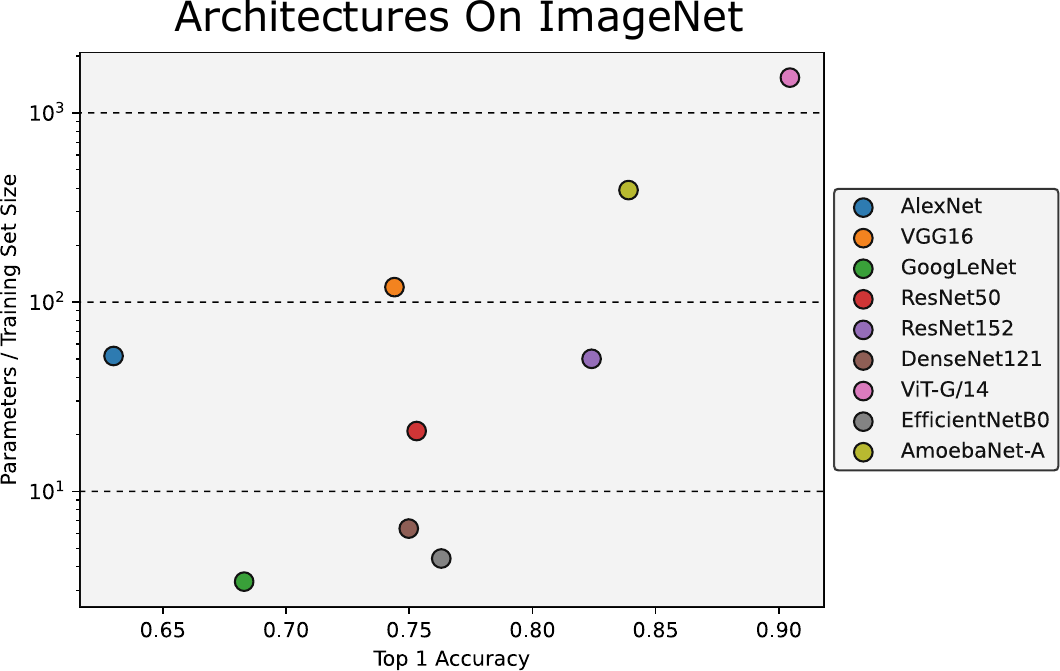}
	\caption{%
          ImageNet Classification Competition:
          Final score on the test set in the Top 1 category vs.\
Parameters-to-Training-Samples Ratio. 
		Note that all architectures have more parameters than training samples. Architectures include AlexNet \cite{krizhevsky2012imagenet}, VGG16 \cite{simonyan2014very}, GoogLeNet \cite{szegedy2015going}, ResNet50/ResNet152 \cite{he2016deep}, DenseNet121 \cite{huang2017densely}, ViT-G/14 \cite{zhai2022scaling}, EfficientNetB0 \cite{tan2019efficientnet}, and AmoebaNet \cite{real2019regularized}.}
	\label{fig:Overparameterizations}
	\label{fig:ratios}
\end{figure}

Surprisingly, successful neural network architectures do not necessarily follow these theoretical observations.
Consider the neural network architectures in Figure \ref{fig:Overparameterizations}.
They represent some of the most renowned image classification models, and all of them participated in the ImageNet Classification Competition \cite{5206848}. The training set consisted of 1.2 million images.
The $x$-axis shows the model performance, and the $y$-axis displays the ratio of the number of parameters to the size of the training set; notably, all architectures have a ratio larger than one, i.e.\ have more parameters than training samples. For the largest model, there are by a factor $1000$ more neural network parameters than training samples.

Given that the practical application of deep learning appears to operate in a regime significantly different from the one analyzed in Chapter \ref{chap:VC}, we must ask: Why do these methods still work effectively?

\section{The double descent phenomenon}

\begin{figure}[htb]
\centering
\includegraphics[width = 0.95\textwidth]{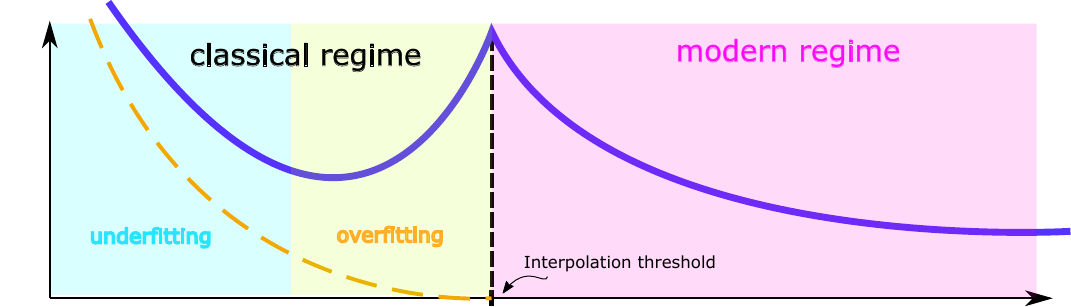}
\put(-85, -10){Expressivity of $\mathcal{H}$}
\put(-350, 75){{\color{blue} $\mathcal{R}(h)$}}
\put(-340, 35){\color{orange} $\widehat{\mathcal{R}}_S(h)$}
\caption{Illustration of the double descent phenomenon.}\label{fig:doubleDescentSketch}
\end{figure}

The success of deep learning in a regime not covered by traditional statistical learning theory puzzled researchers for some time.
In \cite{belkin2019reconciling}, an intriguing set of experiments was performed. 
These experiments indicate that while the risk %
follows the upper bound from Section \ref{sec:ApproxComplTradeOff} for neural network architectures that do not interpolate the data, 
the curve does not expand to infinity in the way that Figure \ref{fig:classApprox-Comp-TradeOff} suggests. 
Instead, after surpassing the so-called ``interpolation threshold'',
the risk starts to decrease again.
This behavior, known as double descent, is illustrated in Figure \ref{fig:doubleDescentSketch}.

\subsection{Least-squares regression revisited}
To gain further insight, we consider ridgeless kernel least-squares
regression as introduced in Section \ref{sec:kernelreg}.
   Consider a data sample $(\Bx_j,y_j)_{j=1}^m\subseteq \R^d\times\R$
  generated by some ground-truth function $f$, i.e.\
  \begin{equation}\label{eq:datarunge}
    y_j=f(\Bx_j)\qquad\text{for }j=1,\dots,m.
  \end{equation}
  Let $\phi_j: \R^d\to\R$, $j\in\N$, be a sequence of \emph{ansatz
    functions}. %
    For $n\in\N$, we wish to fit a
  function %
  $\Bx \mapsto \sum_{i=1}^n w_i \phi_i(\Bx)$ to the data using linear
  least-squares. To this end, we introduce the feature map
\begin{equation*}
  \R^d \ni \Bx \mapsto \phi(\Bx)\dfn (\phi_1(\Bx),\dots,\phi_n(\Bx))^\top\in\R^n.
\end{equation*}
The goal is to determine coefficients $\Bw\in\R^n$ minimizing
the empirical risk
\begin{equation*}
  \widehat\risk_S(\Bw)=  \frac{1}{m}\sum_{j=1}^m\Big(\sum_{i=1}^n w_i\phi_i(\Bx_j) -y_j\Big)^2 = \frac{1}{m}\sum_{j=1}^m(\inp{\phi(\Bx_j)}{\Bw}-y_j)^2.
\end{equation*}
With
\begin{equation}\label{eq:Anrunge}
  \BA_n
  \dfn
    \begin{pmatrix}
    \phi_1(\Bx_1) &\dots &\phi_n(\Bx_1)\\
    \vdots &\ddots &\vdots\\
    \phi_1(\Bx_m) &\dots &\phi_n(\Bx_m)
  \end{pmatrix}
  =
  \begin{pmatrix}
    \phi(\Bx_1)^\top\\
    \vdots\\
    \phi(\Bx_m)^\top
  \end{pmatrix}
  \in\R^{m\times n}
\end{equation}
and $\By=(y_1,\dots,y_m)^\top$ it holds
\begin{equation}\label{eq:hatriskSBwOv}
\widehat\risk_{S}(\Bw) = \frac{1}{m}\norm[]{\BA_n\Bw-\By}^2.
\end{equation}

As discussed in Sections \ref{sec:linreg}-\ref{sec:kernelreg}, a
unique minimizer of \eqref{eq:hatriskSBwOv} only exists if $\BA_n$ has
rank $n$. %
For a minimizer $\Bw_n$, the fitted function reads
\begin{equation}\label{eq:fnpredfunc}
  f_n(x)\dfn \sum_{j=1}^n w_{n,j}\phi_j(x).
\end{equation}
We are interested in the behavior of the $f_n$ as a function of $n$
(the number of ansatz functions/parameters of our model), and
distinguish between two cases:
\begin{itemize}
\item \emph{Underparameterized}: If $n<m$ we have fewer parameters $n$ than training points $m$. For the least squares problem of minimizing $\widehat\risk_S$, this means that there are more conditions $m$ than free parameters $n$. 
Thus, in general, we cannot interpolate the data, and we have $\min_{\Bw\in\R^n}\widehat\risk_S(\Bw)>0$.
\item \emph{Overparameterized}: If $n\ge m$, then we have at least as many
  parameters $n$ as training points $m$. If the $\Bx_j$ and the $\phi_j$
  are such that $\BA_n\in\R^{m\times n}$ has full rank $m$, then there
  exists $\Bw$ such that $\widehat\risk_S(\Bw)=0$. If $n>m$, then $\BA_n$
  necessarily has a nontrivial kernel, and there exist infinitely many
  parameters choices $\Bw$ that yield zero empirical risk
  $\widehat\risk_S$. Some of them lead to better, and some lead to worse
  prediction functions $f_n$ in \eqref{eq:fnpredfunc}.
\end{itemize}

\begin{figure}
  \centering
  \subfloat[ansatz functions $\phi_j$]{\includegraphics[width=0.49\textwidth]{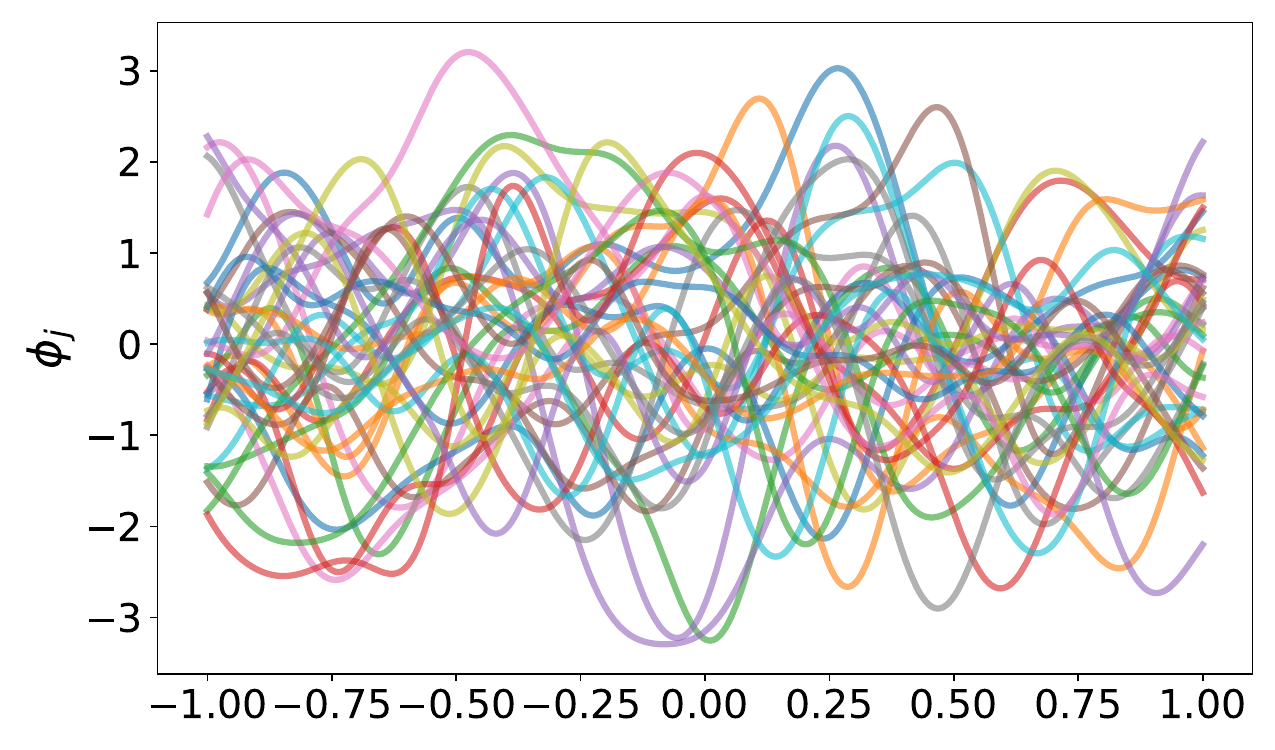}}
  \hfill
  \subfloat[Runge function $f$ and data points]{\includegraphics[width=0.49\textwidth]{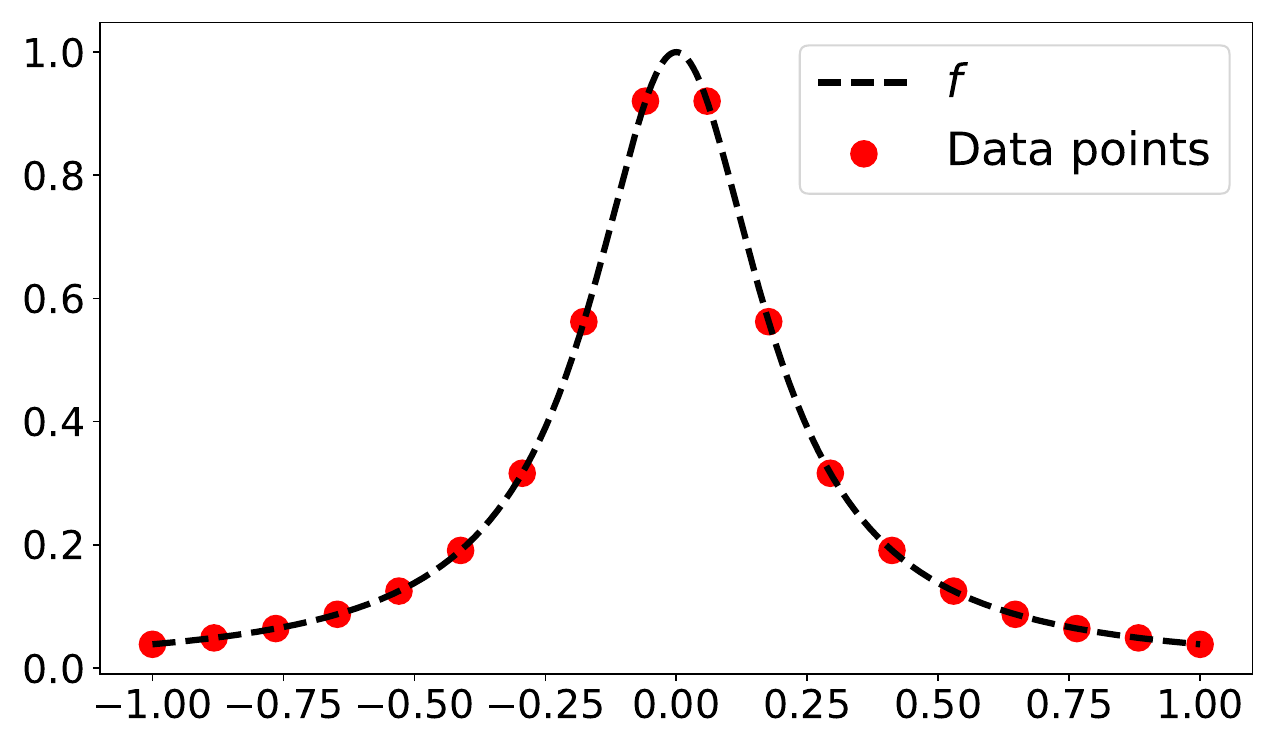}}
\caption{Ansatz functions $\phi_1, \dots, \phi_{40}$ drawn from a Gaussian process, along with the Runge function and $18$ equispaced data points.}\label{fig:runge}
\end{figure}

In the overparameterized case, there exist many minimizers of
$\widehat\risk_S$. The training algorithm we use to compute a minimizer
determines the type of prediction function $f_n$ we obtain. We argued
in Chapter \ref{chap:wideNets}, that for suitable initialization, gradient
descent converges towards the minimal norm minimizer
\footnote{Here, the index $n$ emphasizes the dimension of $\Bw_{n,*}\in\R^n$. This notation should not be confused with the ridge-regularized minimizer $\Bw_{\lambda,*}$ introduced in Chapter \ref{chap:wideNets}.}
\begin{equation}\label{eq:Bwnstar}
  \Bw_{n,*}=\argmin_{\Bw\in M}\norm{\Bw}\in\R^n,\qquad
    M=\set{\Bw\in\R^n}{\widehat\risk_S(\Bw)\le\widehat\risk_S(\Bv)~\forall\Bv\in\R^n}.
\end{equation}

\subsection{An example}\label{sec:anExample}
\begin{figure}
  \centering
    \subfloat[$n=2$ (underparameterization)]{\includegraphics[width=0.49\textwidth]{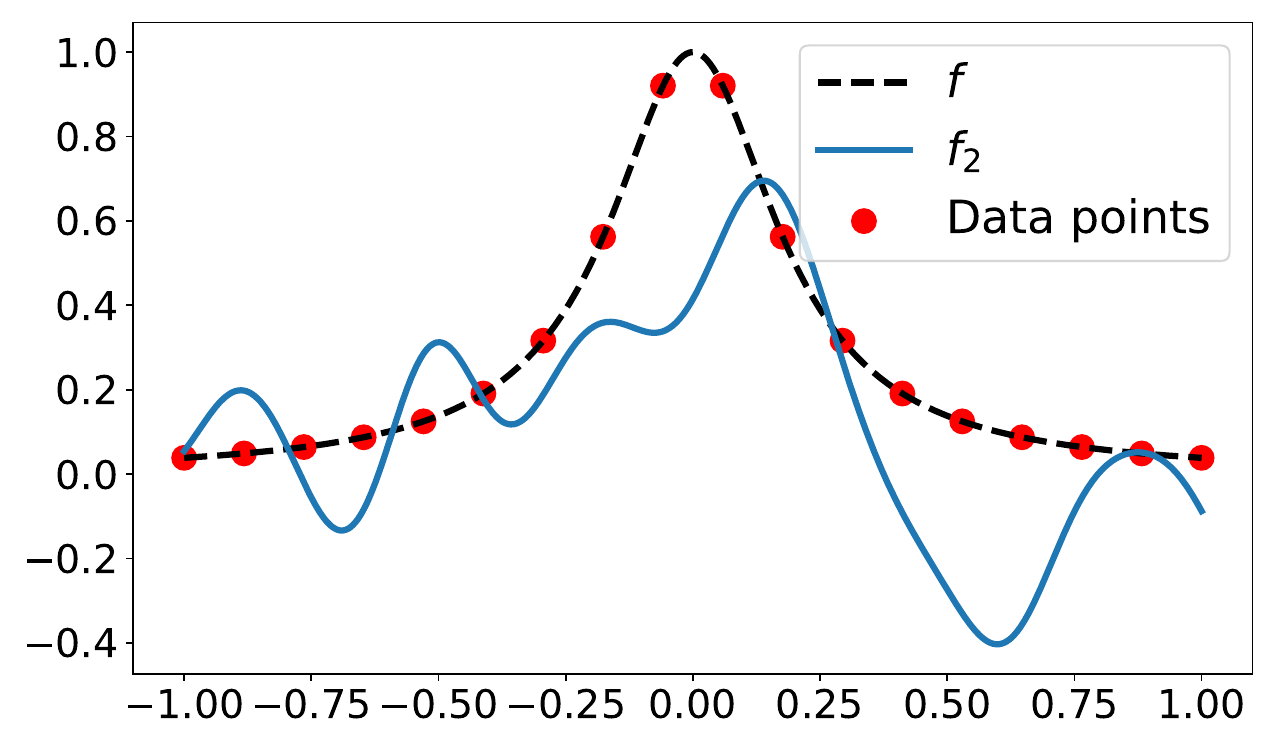}}
    \hfill%
    \subfloat[$n=15$ (balance of appr.\ and gen.\ error)]{\includegraphics[width=0.49\textwidth]{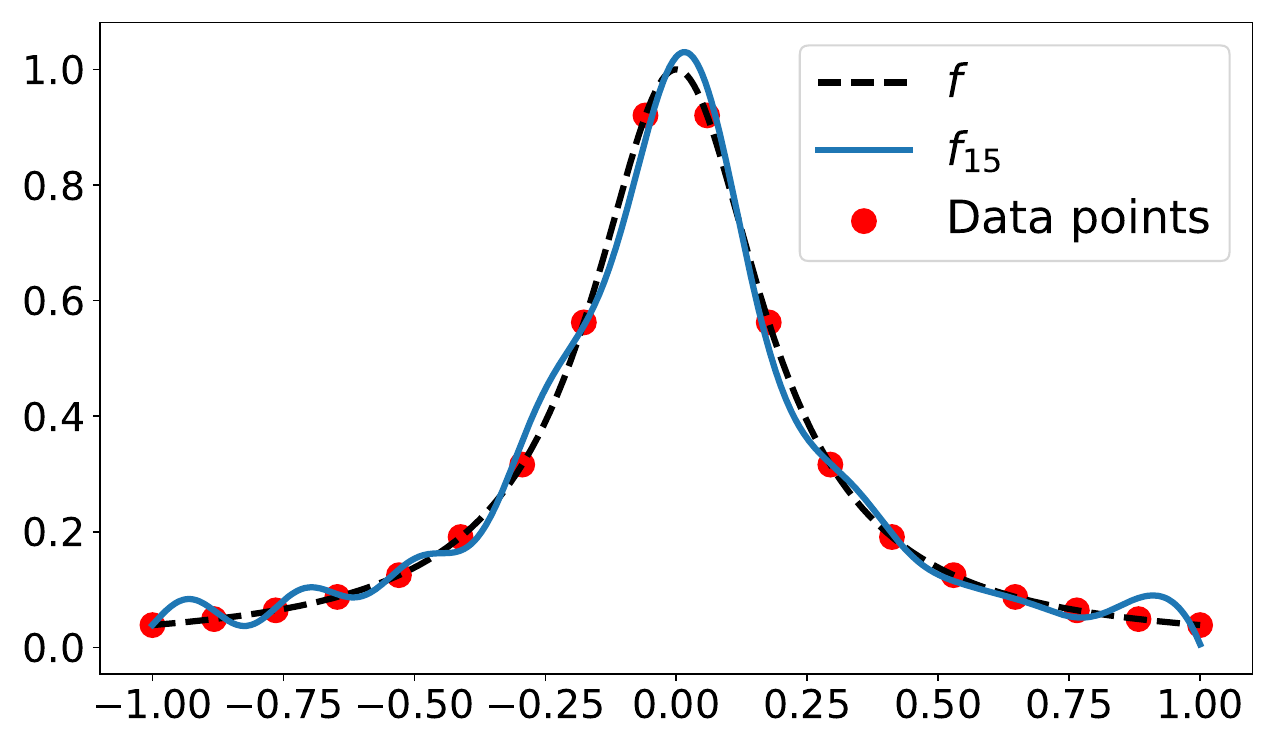}}\\
    \subfloat[$n=18$ (interpolation threshold)]{\includegraphics[width=0.49\textwidth]{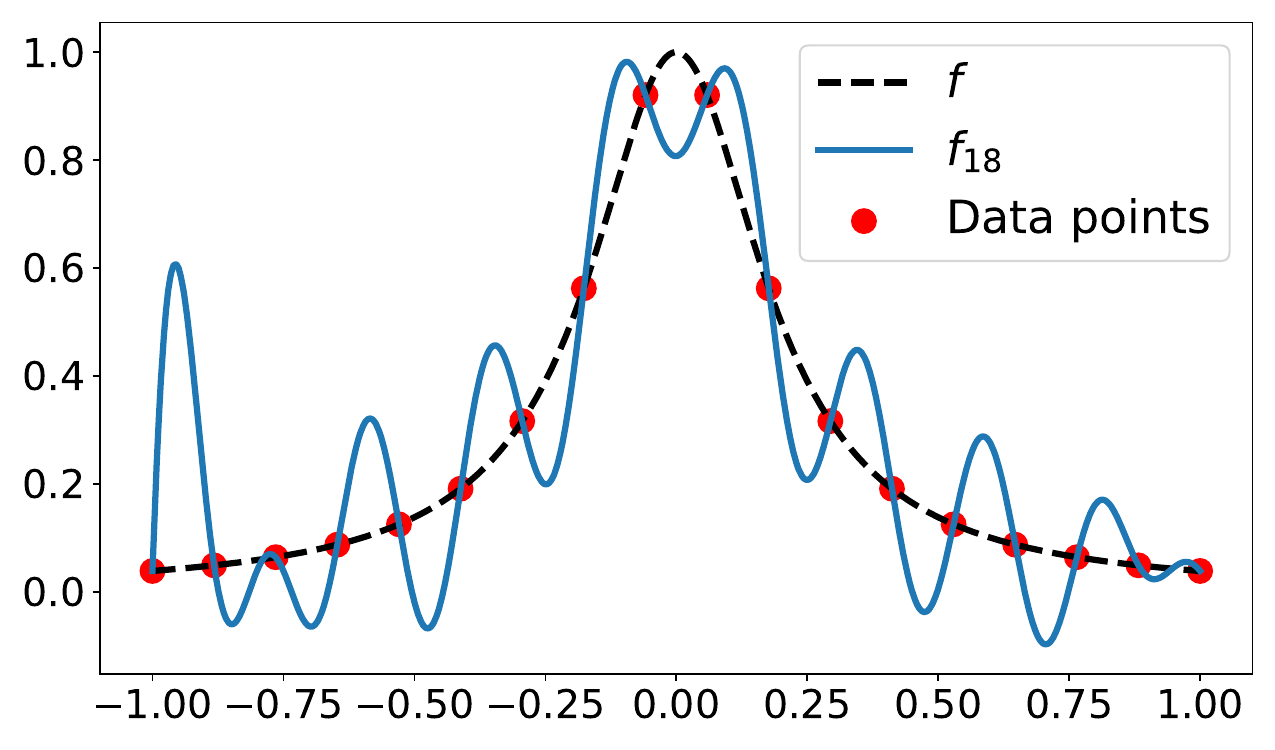}}
    \hfill%
    \subfloat[$n=40$ (overparameterization)]{\includegraphics[width=0.49\textwidth]{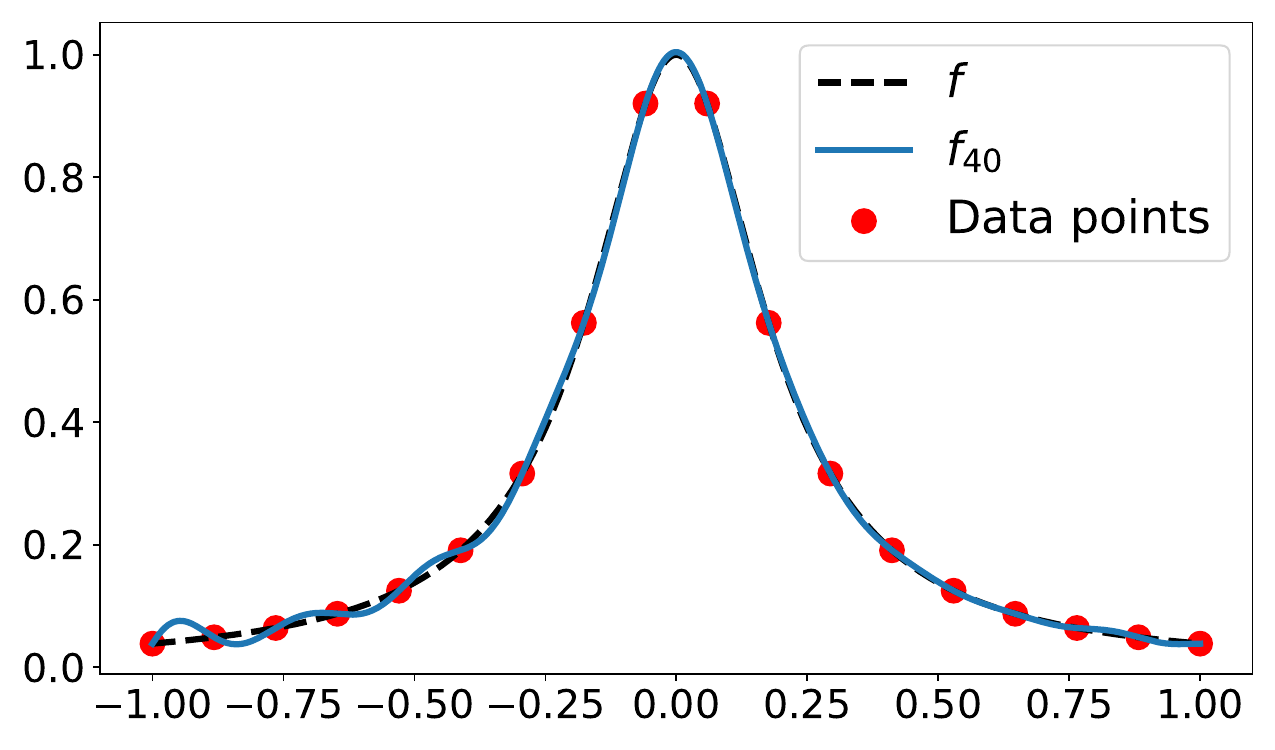}}
    \caption{Fit of the $m=18$ red data points using the ansatz functions $\phi_1, \dots, \phi_n$ from Figure \ref{fig:runge}, employing equations \eqref{eq:Bwnstar} and \eqref{eq:fnpredfunc} for different numbers of ansatz functions $n$.}\label{fig:runge_fit}
\end{figure}

\begin{figure}
    \subfloat[$\|f-f_n\|_{L^2([-1,1])}$]
    {\includegraphics[width=0.495\textwidth]{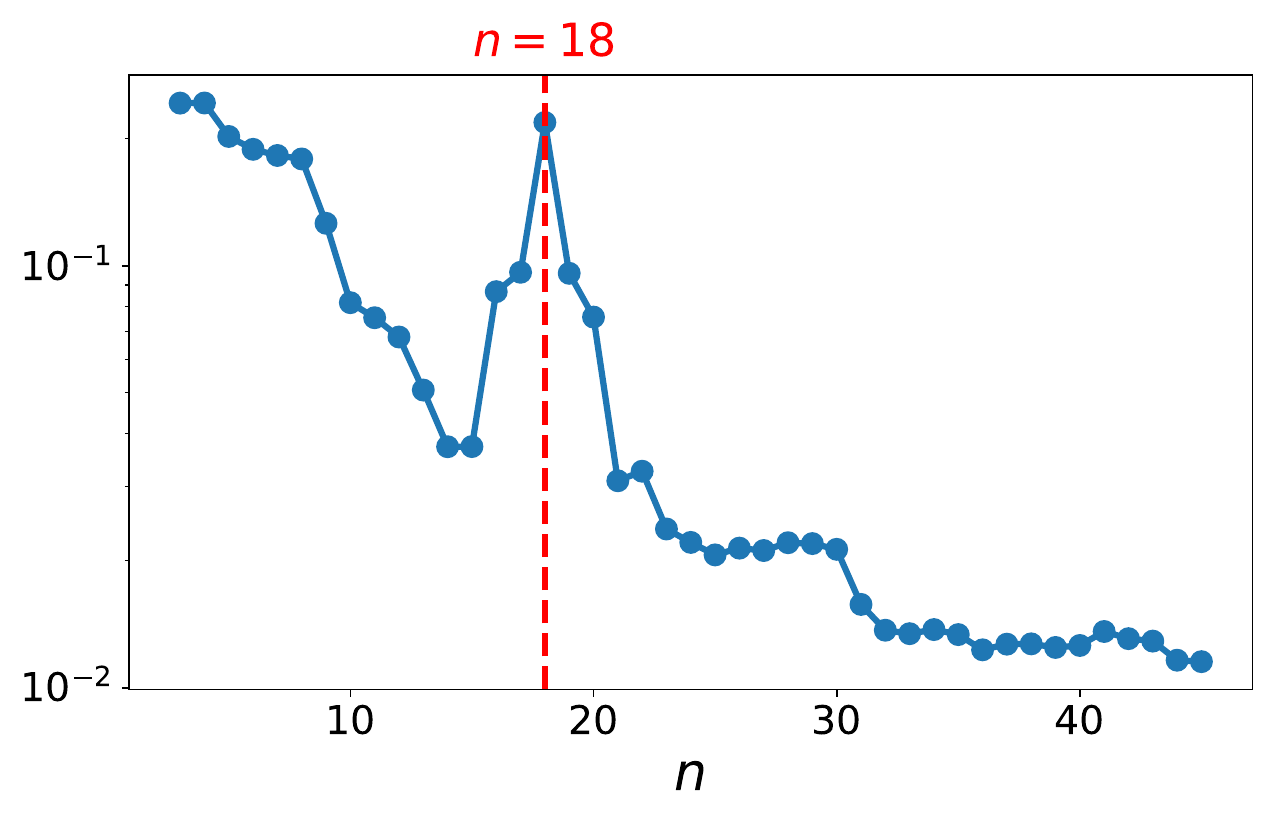}}
    \hfill%
    \subfloat[$\|\Bw_{n,*}\|$]{\includegraphics[width=0.495\textwidth]{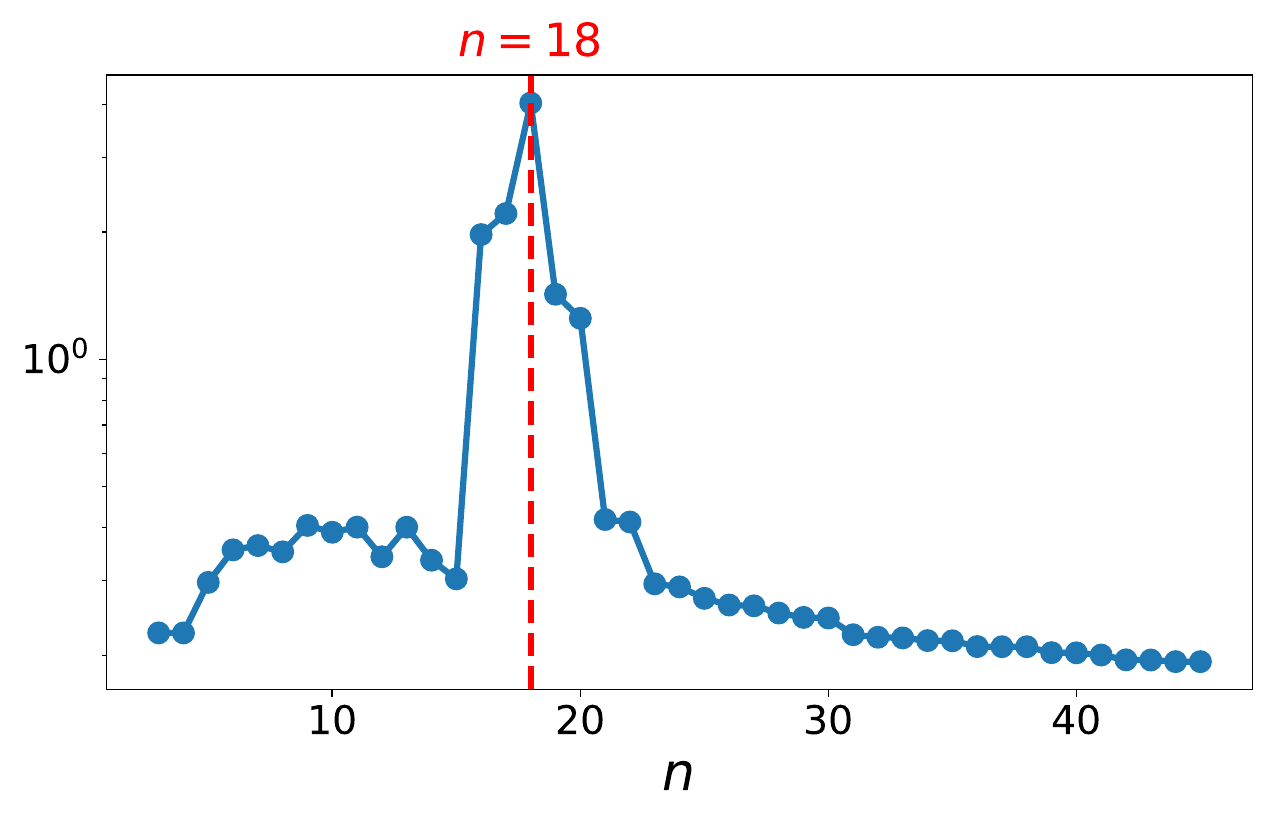}}\\
    \caption{The $L^2$-error for the fitted functions in Figure \ref{fig:runge_fit},
      and the Euclidean norm of the corresponding coefficient vector
      $\Bw_{n,*}$ defined in \eqref{eq:Bwnstar}.
    }\label{fig:runge_error}
\end{figure}

We consider a concrete example. In Figure \ref{fig:runge} we plot a set
of $40$ ansatz functions $\phi_1,\dots,\phi_{40}$, which are %
drawn from a Gaussian process. %
Additionally, the figure shows a plot of the Runge function $f$,
and $m=18$ equispaced points %
which are used as the training data points. We then fit a function in %
${\rm span}\{\phi_1,\dots,\phi_n\}$ via \eqref{eq:Bwnstar} and
  \eqref{eq:fnpredfunc}. The result is displayed in Figure \ref{fig:runge_fit}:

  \begin{itemize}
  \item $n=2$: The model can only represent functions in
    ${\rm span}\{\phi_1,\phi_2\}$. It is not yet expressive enough to
    give a meaningful approximation of $f$.
  \item $n=15$: The model has sufficient expressivity to capture the
    main characteristics of $f$. Since $n=15<18=m$, it is not yet able
    to interpolate the data. Thus it allows to strike a good balanced
    between the approximation and generalization error, which
    corresponds to the scenario discussed in Chapter \ref{chap:VC}.
  \item $n=18$: We are at the interpolation threshold. The model is
    capable of interpolating the data, and there is a unique $\Bw$
    such that $\widehat\risk_S(\Bw)=0$. Yet, in between data points the
    behavior of the predictor $f_{18}$ seems erratic, and displays
    strong oscillations.
  This is referred to as {\bf
      overfitting}, and is to be expected due to our analysis in
    Chapter \ref{chap:VC}; while
    the approximation error at the data points has improved compared
    to the case $n=15$, the generalization error has gotten worse.
  \item $n=40$: This is the overparameterized regime, where we have
    significantly more parameters than data points. Our prediction
    $f_{40}$ interpolates the data and appears to be the best overall
    approximation to $f$ so far, due to a ``good'' choice of minimizer of
    $\widehat\risk_S$, namely \eqref{eq:Bwnstar}. We also note that, while
    quite good, the fit is not perfect.  We cannot expect significant
    improvement in performance by further increasing $n$,
    since %
    at this point the main limiting factor is the amount of available data.
    Also see Figure \ref{fig:runge_error} (a).
  \end{itemize}

Figure \ref{fig:runge_error} (a) displays the error
$\norm[{L^2([-1,1])}]{f-f_n}$ over $n$. We observe the
characteristic double descent curve, where the error initially
decreases and then peaks at the interpolation threshold, which is
marked by the dashed red line. Afterwards, in the overparameterized
regime, it starts to decrease again. Figure \ref{fig:runge_error} (b)
displays $\norm{\Bw_{n,*}}$. Note how the Euclidean norm of the
coefficient vector also peaks at the interpolation threshold.

We emphasize that the precise nature of the convergence curves depends
strongly on various factors, such as the distribution and number of
training points $m$, the ground truth $f$, and the choice of ansatz
functions $\phi_j$ (e.g., the specific kernel used to generate the
$\phi_j$ in Figure \ref{fig:runge} (a)). In the present setting we
  achieve a good approximation of $f$ for $n=15<18=m$ corresponding to
  the regime where the approximation and interpolation errors are
  balanced. However, as Figure \ref{fig:runge_error} (a) shows, it can
  be difficult to determine a suitable value of $n<m$ a priori, and
  the acceptable range of $n$ values can be quite narrow.  For
  overparameterization ($n\gg m$), the precise choice of $n$ is less
  critical, potentially making the algorithm more stable in this
  regime. We encourage the reader to conduct similar experiments and
explore different settings to get a better feeling for the double descent
phenomenon.

\section{Size of weights}\label{sec:sizeOfWeights}

In Figure \ref{fig:runge_error}, we observed that the norm of the coefficients $\norm[]{\Bw_{n,*}}$ exhibits similar behavior to the $L^2$-error, peaking at the interpolation threshold $n=18$. In machine learning, large weights are %
usually undesirable, as they are associated with large derivatives or oscillatory behavior. This is evident in the example shown in Figure \ref{fig:runge_fit} for $n=18$. Assuming that the data in \eqref{eq:datarunge} was generated by a ``smooth'' function $f$, e.g.\ a function with moderate Lipschitz constant, these large derivatives of the prediction function could lead to poor generalization. %
Such a smoothness assumption about $f$ may or may not be satisfied. However, if $f$ is not smooth, there is little hope of accurately recovering $f$ from limited data (see the discussion in Section \ref{sec:OptIntLip}).

The next result gives an explanation for the observed behavior of
$\norm{\Bw_{n,*}}$.  

\begin{proposition}
  Assume that $\Bx_1,\dots,\Bx_m$ and the $(\phi_j)_{j\in\N}$ are such that
  $\BA_n$ in \eqref{eq:Anrunge} has full rank $n$ for all $n\le m$.
  Given $\By\in \R^m$, denote by $\Bw_{n,*}(\By)$ the vector in \eqref{eq:Bwnstar}.
  Then
  \begin{equation*}
    n\mapsto\sup_{\norm{\By}=1}\norm[]{\Bw_{n,*}(\By)}\quad\text{is monotonically}\quad
    \begin{cases}
      \text{increasing} &\text{for }n< m,\\
      \text{decreasing} &\text{for }n\ge m.
      \end{cases}
  \end{equation*}
\end{proposition}
\begin{proof}
  We start with the case $n\ge m$. By assumption $\BA_m$ has full rank
  $m$, and thus $\BA_n$ has rank $m$ for all $n\ge m$,
  see \eqref{eq:Anrunge}. 
  In particular, there exists $\Bw_n\in\R^n$
  such that $\BA_n\Bw_{n}=\By$. Now fix $\By\in\R^m$ and let $\Bw_n$
  be any such vector. Then $\Bw_{n+1}\dfn (\Bw_n,0)\in\R^{n+1}$
  satisfies $\BA_{n+1}\Bw_{n+1}=\By$ and $\norm[]{\Bw_{n+1}}=\norm{\Bw_n}$. Thus
  necessarily $\norm{\Bw_{n+1,*}}\le\norm{\Bw_{n,*}}$ for
  the minimal norm solutions defined in \eqref{eq:Bwnstar}. Since this holds
  for every $\By$, we obtain the statement for $n\ge m$.

  Now let $n<m$. Recall that the minimal norm solution can be written
  through the pseudo inverse
  \begin{equation*}
    \Bw_{n,*}(\By)=\BA_n^\dagger \By,
  \end{equation*}
  see %
  Appendix \ref{app:pseudo}.
  That is,
  \begin{equation*}
    \BA_n^\dagger = \BV_n\begin{pmatrix}
      s_{n,1}^{-1}& & &\\
      &\ddots &&\Bnul\\
      && s_{n,n}^{-1}&
      \end{pmatrix}
      \BU_n^\top\in\R^{n\times m}
  \end{equation*}  
  where $\BA_n=\BU_n\BSigma_n\BV_n^\top$ is the singular value
  decomposition of $\BA_n$, and %
  \begin{equation*}
    \BSigma_n = \begin{pmatrix}
      s_{n,1} & &\\
      & \ddots &\\
      &&s_{n,n}\\
      &\Bnul&
      \end{pmatrix}\in\R^{m\times n}
  \end{equation*}  
  contains the singular values $s_{n,1}\ge\dots\ge s_{n,n}>0$ of $\BA_n\in\R^{m\times n}$ ordered by decreasing size. Since $\BV_n\in\R^{n\times n}$ and
  $\BU_n\in\R^{m\times m}$ are orthogonal matrices, we have
  \begin{equation*}
    \sup_{\norm{\By}=1}\norm{\Bw_{n,*}(\By)}=\sup_{\norm{\By}=1}\norm{\BA_n^\dagger\By}
    =s_{n,n}^{-1}.
  \end{equation*}
  Finally, since the minimal singular value $s_{n,n}$ of $\BA_n$ can be written as
  \begin{equation*}
    s_{n,n}=\inf_{\substack{\Bx\in\R^n\\ \norm{\Bx}=1}}\norm{\BA_n\Bx}\ge
    \inf_{\substack{\Bx\in\R^{n+1}\\ \norm{\Bx}=1}}\norm{\BA_{n+1}\Bx} = s_{n+1,n+1},
  \end{equation*}
  we observe that $n\mapsto s_{n,n}$ is monotonically decreasing for $n\le m$.
  This concludes the proof.
\end{proof}

\section{Theoretical justification}\label{sec:theoreticalJust}
Let us now %
examine one possible explanation of the double descent phenomenon for neural networks. 
While there are many alternative arguments available %
in the literature (see the bibliography section),
the explanation %
presented here is based on a simplification of the ideas in \cite{bartlett1996valid}.

The key assumption %
underlying our analysis is
that large overparameterized neural networks %
tend to be
Lipschitz continuous with a Lipschitz constant independent of the size. 
This is a consequence of %
neural networks typically having relatively small weights. %
To motivate this, let us consider the class of neural networks $\CN(\sigma;\mathcal{A},B)$
for an architecture $\CA$ of depth $d\in\N$ and width $L\in\N$. If $\sigma$
  is $C_\sigma$-Lipschitz %
continuous with $C_\sigma \geq 1$, such that $B \leq c_B \cdot (d C_\sigma)^{-1}$ for some
$c_B >0$, then by Lemma \ref{lem:LipschitzEstimateNN2324} %
\begin{align}\label{eq:NNCoveringBoundedB}
	\mathcal{N}(\sigma; \mathcal{A},  B) \subseteq \mathrm{Lip}_{c_B^L}(\R^{d_0}),
\end{align}
An assumption of the type $B \leq c_B \cdot (d C_\sigma)^{-1}$, i.e.\ a scaling
of the weights by the reciprocal $1/d$ of the width, is %
not unreasonable in practice: Standard initialization schemes such as LeCun \cite{lecun98}
or He \cite{he2015delving} initialization, use random weights with variance scaled inverse
proportional to the input dimension of each layer.
Moreover, as we saw in Chapter \ref{chap:wideNets}, for very wide neural networks, the weights do not move significantly from their initialization during training.
Additionally, many training routines use regularization terms on the weights, thereby encouraging the optimization routine to find small weights. 

We %
study the generalization capacity of Lipschitz functions through %
the covering-number-based learning results of Chapter \ref{chap:VC}.
The set $\mathrm{Lip}_C(\Omega)$ of $C$-Lipschitz functions on a compact $d$-dimensional Euclidean domain $\Omega$ has covering numbers %
bounded according to
\begin{align}\label{eq:VoceringNumOfLipschitz}
	\log( \mathcal{G}(\mathrm{Lip}_C(\Omega), \eps, L^\infty)) \le C_{\rm cov} %
  \cdot \left(\frac{C}{\eps}\right)^d%
  \quad \text{ for all } \eps >0
\end{align}
for some constant $C_{\rm cov}$ independent of $\eps>0$.
A proof can be found in \cite[Lemma 7]{gottlieb2017efficient}, see also \cite{tikhomirov1993varepsilon}. 

As a result of these considerations, we can identify two regimes:
\begin{itemize}
\item \textit{Standard regime:} %
  For small neural network size $n_{\CA}$,
  we %
  consider neural networks as a set parameterized by $n_\mathcal{A}$ parameters. As we have seen before, this yields a bound on the generalization error that scales linearly with $n_\mathcal{A}$. As long as $n_\mathcal{A}$ is small in comparison %
  to
  the number of samples, we can expect good generalization by Theorem \ref{thm:coveringNumberGeneralizationBoundForNNs}.
\item \textit{Overparameterized regime:} %
  For large neural network size $n_{\CA}$ and small weights,
  we %
          consider neural networks as a subset of $\mathrm{Lip}_C(\Omega)$ for a constant $C>0$. This set has a covering number bound that is independent %
          of the number of parameters %
          $n_\mathcal{A}$.
\end{itemize}

Choosing the better of the two generalization bounds for each regime yields the following result.
Recall that $\CN^*(\sigma;\CA,B)$ denotes all neural networks in
  $\CN(\sigma;\CA,B)$ with a range contained in $[-1,1]$ (see \eqref{eq:CNstar}).

\begin{theorem}\label{thm:BoundForNNsOverparam}
Let $C$, $C_{\mathcal{L}} >0$ and let $\mathcal{L} \colon [-1,1] \times [-1,1] \to \R$ be $C_{\mathcal{L}}$-Lipschitz.
Further, let $\mathcal{A} = (d_0, d_1, \dots, d_{L+1}) \in \N^{L+2}$, let $\sigma\colon \R \to \R$ be $C_\sigma$-Lipschitz continuous with $C_\sigma \geq 1$, and $|\sigma(x)| \leq C_\sigma|x|$ for all $x \in \R$, and let $B>0$. 
	
Then, there exist $c_1$, $c_2>0$, such that for every $m \in \N$ and every distribution $\mathcal{D}$ on $[-1,1]^{d_0} \times [-1,1]$ it holds with probability at least $1-\delta$ over %
$S \sim \mathcal{D}^m$ that
for all $\Phi \in \mathcal{N}^*(\sigma; \mathcal{A}, B) \cap \mathrm{Lip}_C([-1,1]^{d_0})$
	\begin{align}\label{eq:OverParamEstimate}
		|\mathcal{R}(\Phi) -\widehat{\mathcal{R}}_S(\Phi) |&\leq  g(\mathcal{A}, C_\sigma , B,m) + 4 C_{\mathcal{L}}  \sqrt{\frac{\log(4/\delta)}{m}},
	\end{align}  	
	where
	\begin{align*}\nonumber
          g(\mathcal{A}, C_\sigma , B, m) = \min\left\{c_1 \sqrt{\frac{n_{\mathcal{A}}  \log(n_{\mathcal{A}} \lceil \sqrt{m}\rceil) + L n_{\mathcal{A}}\log(d_{\rm max})}{m} }, c_2 m ^{-\frac{1}{2+d_0}}\right\}.
	\end{align*}
\end{theorem}

\begin{proof}
  Applying Theorem \ref{thm:coveringNumberGeneralizationBound} with $\alpha = 1/(2+d_0)$
  and \eqref{eq:VoceringNumOfLipschitz},
we obtain that with probability at least $1-\delta/2$ it holds
for all $\Phi \in \mathrm{Lip}_C([-1,1]^{d_0})$
\begin{align*}
		|\mathcal{R}(\Phi) - \widehat{\mathcal{R}}_S(\Phi)| &\leq 4 C_{\mathcal{L}} \sqrt{\frac{{C_{\rm cov}} (m^{\alpha}C )^{d_0} + \log(4/\delta)}{m}} + \frac{2C_{\mathcal{L}}}{m^{\alpha}}\\
& \leq 4 C_{\mathcal{L}} \sqrt{{C_{\rm cov}} C^{d_0} (m^{d_0/(d_0+2) - 1})} + \frac{2C_{\mathcal{L}}}{m^{\alpha}} + 4 C_{\mathcal{L}} \sqrt{\frac{\log(4/\delta)}{m}}\\
& = 4 C_{\mathcal{L}} \sqrt{{C_{\rm cov}} C^{d_0} (m^{-2/(d_0+2)})} + \frac{2C_{\mathcal{L}}}{m^{\alpha}} +  4 C_{\mathcal{L}}\sqrt{\frac{\log(4/\delta)}{m}}\\
& = \frac{(4 C_{\mathcal{L}} \sqrt{{C_{\rm cov}} C^{d_0}}  + 2C_{\mathcal{L}})}{m^{\alpha}} +  4 C_{\mathcal{L}}\sqrt{\frac{\log(4/\delta)}{m}},
\end{align*}
where we used in the second inequality that $\sqrt{x+y}\le \sqrt{x}+\sqrt{y}$ for all $x$, $y\ge 0$.

In addition, Theorem \ref{thm:coveringNumberGeneralizationBoundForNNs} yields that with probability at least $1-\delta/2$
it holds for all $\Phi \in \mathcal{N}^*(\sigma; \mathcal{A}, B)$
\begin{align*}
		|\mathcal{R}(\Phi) - \widehat{\mathcal{R}}_S(\Phi)| &\leq 4 C_{\mathcal{L}} \sqrt{\frac{n_{\mathcal{A}} \log( \lceil n_{\mathcal{A}} \sqrt{m}\rceil )  + L n_{\mathcal{A}} \log( \lceil 2 C_\sigma B d_{\rm max}  \rceil ) + \log(4/\delta)}{m}}\\
		&\qquad  + \frac{2C_{\mathcal{L}}}{\sqrt{m}}\\
& \leq 6 C_{\mathcal{L}} \sqrt{\frac{n_{\mathcal{A}} \log( \lceil n_{\mathcal{A}} \sqrt{m}\rceil )  + L n_{\mathcal{A}} \log( \lceil 2 C_\sigma B d_{\rm max}  \rceil )}{m}}\\
		&\qquad  +4 C_{\mathcal{L}} \sqrt{\frac{ \log(4/\delta))}{m}}.
\end{align*}

Then, for $\Phi \in \mathcal{N}^*(\sigma; \mathcal{A}, B) \cap \mathrm{Lip}_C([-1,1]^{d_0})$ the minimum of both upper bounds holds with probability at least $1-\delta$.
\end{proof}

The two regimes in  Theorem \ref{thm:BoundForNNsOverparam}
correspond to the two terms %
comprising the minimum %
in the definition of $g(\mathcal{A}, C_\sigma , B, m)$. The first term increases with  $n_\mathcal{A}$ while the second is constant.
In the first regime, where the first term is smaller, the generalization gap 
$|\mathcal{R}(\Phi) - \widehat{\mathcal{R}}_S(\Phi)|$ increases with $n_\mathcal{A}$.

In the second regime, %
where the second term is smaller, the generalization gap is constant with $n_\mathcal{A}$.
Moreover, %
it is reasonable to assume that the empirical risk $\widehat{\mathcal{R}}_S$ will decrease with increasing number of parameters $n_\mathcal{A}$.

By \eqref{eq:OverParamEstimate} we can bound the risk by
$$
\mathcal{R}(\Phi) \leq \widehat{\mathcal{R}}_S +  g(\mathcal{A}, C_\sigma , B,m) + 4 C_{\mathcal{L}}  \sqrt{\frac{\log(4/\delta)}{m}}.
$$
In the second regime, this upper bound is
monotonically decreasing. %
In the first regime it may both decrease and increase.
In some cases, this behavior can lead to an upper bound on the risk %
resembling the curve of Figure \ref{fig:doubleDescentSketch}.
The following section describes a specific scenario where this is the case.

\begin{remark}\label{rmk:toooptimistic}
  Theorem \ref{thm:BoundForNNsOverparam} %
  assumes $C$-Lipschitz continuity of the neural networks.
As we saw in Sections \ref{sec:anExample} and \ref{sec:sizeOfWeights}, this assumption %
may not hold near the interpolation threshold. 
  Hence, Theorem \ref{thm:BoundForNNsOverparam} likely gives a too optimistic upper bound %
  near the interpolation threshold.
\end{remark}

\section*{Bibliography and further reading}

The discussion on kernel regression and the effect of the number of parameters on the norm of the weights was already given in \cite{belkin2019reconciling}. Similar analyses, with more complex ansatz systems and more precise asymptotic estimates, are found in \cite{mei2022generalization, hastie2022surprises}. Our results in Section \ref{sec:theoreticalJust} are inspired by \cite{bartlett1996valid}; see also \cite{neyshabur2015norm}.

For a detailed account of further arguments justifying the surprisingly good generalization capabilities of overparameterized neural networks, we refer to \cite[Section 2]{berner2021modern}. Here, we only briefly mention two additional directions of inquiry. First, if the learning algorithm introduces a form of robustness, this can be leveraged to yield generalization bounds \cite{arora2018stronger, xu2012robustness, bousquet2002stability, poggio2004general}. Second, for very overparameterized neural networks, it was stipulated in \cite{jacot2018neural} that neural networks become linear kernel interpolators as discussed in Chapter \ref{chap:wideNets}.
Thus, for large neural networks, generalization can be studied through kernel regression \cite{jacot2018neural, lee2019wide, belkin2018understand, li2021generalization}.

\newpage
\section*{Exercises}

\begin{exercise}\label{ex:GOI1}
  Let $f:[-1,1]\to\R$ be a continuous function, and let
  $-1\le x_1<\dots<x_m\le 1$ for some fixed $m\in\N$.
  As in Section \ref{sec:anExample}, we wish to approximate $f$ by
  a least squares approximation. To this end we use the Fourier ansatz
  functions
  \begin{equation}\label{eq:ansatzfourier}
    b_0(x)\dfn \frac{1}{2}\qquad\text{and}\qquad
    b_j(x)\dfn \begin{cases}
      \sin(\lceil\frac j2\rceil\pi x) &j\ge 1\text{ is odd}\\
      \cos(\lceil\frac j2\rceil\pi x) &j\ge 1\text{ is even}.
      \end{cases}
  \end{equation}
  With the empirical risk
\begin{equation*}
  \widehat\risk_S(\Bw)=  \frac{1}{m}\sum_{j=1}^m\Big(\sum_{i=0}^n w_ib_i(x_j) -y_j\Big)^2,
\end{equation*}
denote by $\Bw_{*}^n\in\R^{n+1}$ the minimal norm minimizer
of $\widehat\risk_S$, and set $f_n(x)\dfn \sum_{i=0}^n w_{*,i}^nb_i(x)$.

Show that in this case generalization fails in the overparameterized
regime: for sufficiently large $n\gg m$, $f_n$ is 
\emph{not} necessarily a good approximation to $f$. What does $f_n$ converge to as
$n\to\infty$?
\end{exercise}

\begin{exercise}
  Consider the setting of Exercise \ref{ex:GOI1}. We
  adapt the ansatz functions in \eqref{eq:ansatzfourier} by rescaling them via
  \begin{equation*}
    \tilde b_j\dfn c_j b_j.
  \end{equation*}
  Choose real numbers $c_j\in\R$, such that the corresponding minimal
  norm least squares solution avoids the phenomenon encountered in
  Exercise \ref{ex:GOI1}.

  \emph{Hint:} Should ansatz functions corresponding to large
  frequencies be scaled by large or small numbers to avoid
  overfitting?
\end{exercise}

\begin{exercise}
  Prove \eqref{eq:VoceringNumOfLipschitz} for $d = 1$.
\end{exercise}

%% file: RobustnessAndAdversarialEx.tex
\chapter{Robustness and adversarial examples}\label{chap:adversarial}
\begin{figure}[b]\centering
\includegraphics[width = \textwidth]{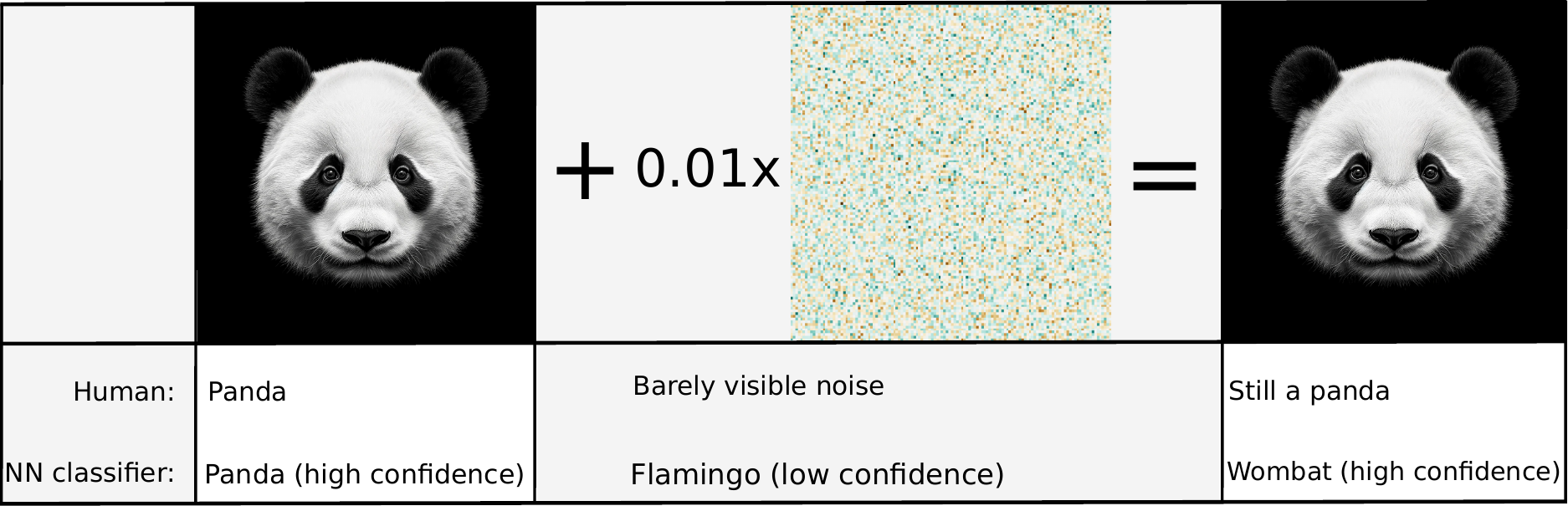}
\caption{Sketch of %
  an adversarial example.}
\label{fig:adversarialExamples}
\end{figure}

How sensitive is the output of a neural network to small changes in its input? %
Real-world observations of trained neural networks often reveal that %
even barely noticeable modifications of the input can lead to drastic variations in the network's predictions.
This intriguing behavior was first documented in the context of image classification in \cite{szegedy2013intriguing}.

Figure \ref{fig:adversarialExamples} illustrates %
this concept.
The left panel shows a picture of a panda that the neural network %
correctly classifies as %
a panda.
By adding an almost imperceptible amount of noise to the image, we obtain the modified image in the right panel.
To a human, there is no visible difference, %
but the neural network %
classifies the perturbed image as a wombat.
This phenomenon, where a correctly classified image is misclassified after a slight perturbation, is termed an \emph{adversarial example}.

In practice, such behavior is highly undesirable. It indicates that our learning algorithm might not be very reliable and poses a potential security risk, as malicious actors could exploit it to trick the algorithm. In this chapter, we describe the basic mathematical principles behind adversarial examples and investigate simple conditions under which they might or might not occur. For simplicity, we restrict ourselves to a binary classification problem but note that the main ideas remain valid in more general situations.

\section{Adversarial examples}
Let us start by formalizing the notion of an adversarial example. We
consider the problem of assigning a label $y\in\{-1,1\}$ to a vector
$\Bx\in\R^d$. It is assumed that the relation between $\Bx$ and $y$ is
described by a distribution $\CD$ on $\R^d\times\{-1,1\}$. In
particular, for a given $\Bx$, both values $-1$ and $1$ could have
positive probability, i.e.\ the label is not necessarily deterministic.
Additionally, we let
\begin{equation}\label{eq:featureSupport}
  D_\Bx\dfn \set{\Bx\in\R^d}{\exists y\text{ s.t. }(\Bx,y)\in\supp(\CD)},
\end{equation}
and refer to $D_\Bx$ as the {\bf feature support}. %

Throughout this chapter we denote by
\begin{equation*}
  g \colon \R^{d} \to \{-1,0,1\}
\end{equation*}
a fixed so-called \emph{ground-truth classifier}, satisfying\footnote{To be more precise, the conditional distribution of $y|\Bx$ is only well-defined almost everywhere w.r.t.\ the marginal distribution of $\Bx$. Thus \eqref{eq:gBxBx} can only be assumed to hold for \emph{almost every $\Bx\in D_\Bx$ w.r.t.\ to the marginal distribution of $\Bx$}.}
\begin{equation}\label{eq:gBxBx}
  \bbP[y=g(\Bx)|\Bx]\ge \bbP[y=-g(\Bx)|\Bx]\qquad\text{for all }\Bx\in D_\Bx.
\end{equation}
Note that we allow $g$ to take the value $0$, which is to be
understood as an additional label corresponding to nonrelevant or
nonsensical input data $\Bx$. We will refer to $g^{-1}(0)$ as the {\bf
  nonrelevant class}. The ground truth $g$ is interpreted as how a
human would classify the data, as the following example illustrates.
\begin{example}
  We wish to classify whether an image shows a panda ($y=1$) or a
  wombat ($y=-1$). Consider again Figure
  \ref{fig:adversarialExamples}, and denote the three images by
  $\Bx_1$, $\Bx_2$, $\Bx_3$. The first image $\Bx_1$ is a photograph
  of a panda. Together with a label $y$, it can be interpreted as a
  draw $(\Bx_1,y)$ from a distribution of images $\CD$, i.e.\ $\Bx_1\in D_\Bx$ and
  $g(\Bx_1)=1$. The second image $\Bx_2$ displays noise and
  corresponds to nonrelevant data as it shows neither a panda nor a
  wombat. In particular, $\Bx_2\in D_\Bx^c$ and $g(\Bx_2)=0$. The
  third (perturbed) image $\Bx_3$ also belongs to $D_\Bx^c$, as it is %
  not a photograph but a noise corrupted version of
  $\Bx_1$. Nonetheless, it is \emph{not} nonrelevant, as a human would
  classify it as a panda. Thus $g(\Bx_3)=1$.
\end{example}
 Additional to the ground truth $g$, we denote
by
\begin{equation*}
  h \colon \R^{d} \to \{-1,1\}
\end{equation*}
some trained classifier.

\begin{definition}\label{def:adversarialExamples}
  Let $g \colon \R^{d} \to \{-1,0,1\}$ be the ground-truth classifier,
  let $h \colon \R^{d} \to \{-1,1\}$ be a classifier, and let
  $\norm[*]{\cdot}$ be a norm on $\R^d$. For $\Bx\in\R^d$ and $\delta >0$, we call
  $\Bx' \in \R^{d}$ an {\bf adversarial example} to $\Bx \in \R^{d}$
  with perturbation $\delta$, if and only if
  \begin{enumerate}
	\item %
          $\norm[*]{\Bx' - \Bx} \leq \delta$,
	\item %
          $g(\Bx)g(\Bx')>0$,
	\item %
          $h(\Bx) = g(\Bx)$ and $h(\Bx')\neq g(\Bx')$.
 \end{enumerate}
\end{definition}

In words, $\Bx'$ is an adversarial example to $\Bx$ with perturbation
$\delta$, if (i) the distance of $\Bx$ and $\Bx'$ is at most $\delta$,
(ii) $\Bx$ and $\Bx'$ belong to the same (not nonrelevant) class
according to the ground truth classifier, and (iii) the classifier $h$
correctly classifies $\Bx$ but misclassifies $\Bx'$.

\begin{remark}
  We emphasize that the concept of a ground-truth classifier $g$ differs from a minimizer of the Bayes risk \eqref{eq:BayesRiskGenSection} for two reasons. First, we allow for an additional label $0$ corresponding to the nonrelevant class, which does not exist for the data generating distribution $\CD$. Second, $g$ should correctly classify points \emph{outside of $D_\Bx$}; small perturbations of images as we find them in adversarial examples, are not regular images in $D_\Bx$. Nonetheless, a human classifier can still classify these images, and $g$ models this property of human classification.
\end{remark}

\section{Bayes classifier}\label{sec:BayesClassifier}
At first sight, an adversarial example seems to be no more than %
a misclassified sample.
Naturally, these exist if the model does not generalize well.
In this section we present the more nuanced view of \cite{stutz2019disentangling}.

To avoid edge cases, we assume in the following that for all $\Bx\in D_\Bx$
\begin{equation}\label{eq:bayesunique}
  \text{either} \quad\bbP[y=1|\Bx]>\bbP[y=-1|\Bx]\quad\text{or}\quad
  \bbP[y=1|\Bx]<\bbP[y=-1|\Bx]
\end{equation}
so that
\eqref{eq:gBxBx} uniquely defines $g(\Bx)$ for $\Bx\in D_\Bx$.
We say that the distribution
{\bf exhausts the domain} if
$D_\Bx \cup g^{-1}(0) = \R^{d}$. This means that every point is either
in the feature support $D_\Bx$ %
or %
it belongs to the nonrelevant class. Moreover, we say that $h$ is a
{\bf Bayes classifier} if %
	\[
		\mathbb{P}[h(\Bx) | \Bx] \geq \mathbb{P}[-h(\Bx) | \Bx]\qquad\text{for all }\Bx\in D_\Bx.
	\]
        By \eqref{eq:gBxBx}, the ground truth $g$ is a Bayes
        classifier, and \eqref{eq:bayesunique} ensures that $h$
        coincides with $g$ on $D_\Bx$ if $h$ is a Bayes classifier. It
        is easy to see that a Bayes classifier minimizes the Bayes
        risk.
        
        With these two %
        notions, we now distinguish between four cases.
\begin{enumerate}
\item \textit{Bayes classifier/exhaustive distribution:} If $h$ is a Bayes
  classifier and the data exhausts the domain, then there are \emph{no
    adversarial examples}. This is because every $\Bx\in \R^d$
  either belongs to the nonrelevant class or is classified the same
  by $h$ and $g$.
\item\label{item:BayesNonexhaustive} \textit{Bayes classifier/non-exhaustive distribution:} If $h$ is a Bayes
  classifier and the distribution does not exhaust the domain, then
  \emph{adversarial examples can exist}.
  Even though the learned classifier $h$ %
  coincides with the ground truth $g$ on the feature support,
  adversarial examples can
  be constructed for data points on the complement
  of %
  $D_\Bx \cup g^{-1}(0)$, which is not empty.
	\item \textit{Not a Bayes classifier/exhaustive distribution:}
          The set $D_\Bx$ can be covered by the four subdomains
          \begin{equation}\label{eq:thisNeedsToBeReferencedInTheExercise}
            \begin{aligned}
		C_1 &= h^{-1}(1) \cap g^{-1}(1), \quad F_1 = h^{-1}(-1) \cap g^{-1}(1),\\
		C_{-1} &= h^{-1}(-1) \cap g^{-1}(-1), \quad F_{-1} = h^{-1}(1) \cap g^{-1}(-1).
              \end{aligned}
	\end{equation}
	If $\mathrm{dist}(C_1 \cap D_\Bx, F_1 \cap D_\Bx)$ or $\mathrm{dist}(C_{-1} \cap D_\Bx, F_{-1} \cap D_\Bx)$ is smaller than $\delta$, then there exist points $\Bx$, $\Bx' \in D_\Bx$ such that $\Bx'$ is an adversarial example to $x$ with perturbation $\delta$. 
	Hence, \emph{adversarial examples in the feature support can exist}.
	This is, however, not guaranteed to happen.
For example, $D_\Bx$ does not need to be connected if $g^{-1}(0) \neq \emptyset$, see
Exercise \ref{ex:AdversarialExampleExample}.
Hence, even for classifiers that have incorrect predictions on the data, adversarial examples \emph{do not need to exist}.
	
	\item\label{item:NonBayesExhaustive} \textit{Not a Bayes classifier/non-exhaustive distribution:}	
	In this case \emph{everything is possible}.
Data points and their associated adversarial examples can appear in the feature support of the distribution and adversarial examples to elements in the feature support of the distribution can be created by leaving the feature support of the distribution.
We will see examples %
in the following section.	
\end{enumerate}

\section{Affine classifiers}\label{sec:affineClass}

For linear classifiers, %
  a simple argument outlined in %
  \cite{szegedy2013intriguing} and \cite{goodfellow2014explaining}
  showcases that the high-dimensionality of the input, %
common in image classification problems, is a potential %
cause for the existence of adversarial examples. 

A linear classifier is a map of the form
\begin{equation*}
  \Bx\mapsto \mathrm{sign}(\Bw^\top \Bx)\qquad \text{where }\Bw, \Bx \in \R^d.
\end{equation*}
Let
\begin{equation*}
  \Bx' \dfn \Bx - 2 |\Bw^\top \Bx| \frac{\mathrm{sign}(\Bw^\top \Bx) \mathrm{sign}(\Bw)}{\|\Bw\|_1}
\end{equation*}
where $\mathrm{sign}(\Bw)$ is understood coordinate-wise.
Then %
$\|\Bx - \Bx'\|_\infty \leq 2 |\Bw^\top \Bx| /\|\Bw\|_1$
and it is not hard to see that $\mathrm{sign}(\Bw^\top \Bx') \neq \mathrm{sign}(\Bw^\top \Bx)$.

For high-dimensional vectors $\Bw$, $\Bx$ %
  chosen at random but possibly dependent %
  such that $\Bw$ is uniformly distributed on a $d-1$ dimensional sphere, it holds with high probability that
\begin{align*}%
	\frac{|\Bw^\top \Bx|}{\|\Bw\|_1} \leq \frac{\|\Bx\| \|\Bw\|}{\|\Bw\|_1} \ll \|\Bx\|.
\end{align*}
This can be seen by noting that for every $c>0$ %
\begin{align}\label{eq:HighDimBalls}
\mu(\set{\Bw \in \R^d }{\|\Bw\|_1 > c, \|\Bw\| \leq 1} ) \to 1 \text{ for } d \to \infty,
\end{align}
where $\mu$ is the uniform probability measure on the $d$-dimensional Euclidean unit ball, see Exercise \ref{ex:highDimBalls}. 
Thus, if $\Bx$ has a moderate Euclidean norm, %
the perturbation of $\Bx'$ is likely small for large dimensions.

Below we give a sufficient condition for the existence of adversarial examples, in case both $h$ and the ground truth $g$ are linear classifiers.

\begin{theorem}\label{thm:LinearClassifierAdversarial}
  Let $\Bw$, $\overline{\Bw} \in \R^{d}$ be nonzero. For $\Bx \in \R^d$, let
  $h(\Bx) = \mathrm{sign}(\Bw^\top \Bx)$ be a classifier and
  let $g(\Bx) = \mathrm{sign}(\overline{\Bw}^\top x)$ be the ground-truth classifier.

  For every $\Bx\in \R^{d}$ with $h(\Bx)g(\Bx)>0$ and all $\eps \in (0, |\Bw^\top \Bx|)$ such that 
\begin{align}\label{eq:adversarialExampleEquation}
	\frac{|\overline{\Bw}^\top \Bx|}{\|\overline{\Bw}\|}  > \frac{\eps  + |\Bw^\top \Bx|}{\|\Bw\|}  \frac{|\Bw^\top \overline{\Bw}|}{\|\Bw\| \| \overline{\Bw}\|}
\end{align}
it holds that 
\begin{align}\label{eq:definitionOfTheAdversarialExample}
	\Bx' = \Bx - h(\Bx) \frac{\eps + |\Bw^\top \Bx|}{\|\Bw\|^2} \Bw
\end{align}
is an adversarial example to $\Bx$ with perturbation $\delta =  (\eps  + |\Bw^\top \Bx|)/{\|\Bw\|}$.
\end{theorem}

Before we present the proof, we give some interpretation of this result. First, note that $\set{\Bx\in\R^d}{\Bw^\top\Bx = 0}$ is the decision boundary of $h$, meaning that points lying on opposite sides of this hyperplane, are classified differently by $h$. Due to $|\Bw^\top\overline{\Bw}|\le \norm{\Bw}\norm{\overline{\Bw}}$, \eqref{eq:adversarialExampleEquation} implies that an adversarial example always exists whenever
  \begin{equation}\label{eq:decisionmargin}
    \frac{|\overline{\Bw}^\top\Bx|}{\norm{\overline{\Bw}}}> \frac{|\Bw^\top\Bx|}{\norm{\Bw}}.
  \end{equation}
  The left term is the decision margin of $\Bx$ for $g$, i.e.\ the distance of $\Bx$ to the decision boundary of $g$. Similarly, the term on the right is the decision margin of $\Bx$ for $h$. Thus we conclude that adversarial examples exist if the decision margin of $\Bx$ for the ground truth $g$ is larger than that for the classifier $h$.

Second, the term $(\Bw^\top \overline{\Bw})/{(\|\Bw\| \| \overline{\Bw}\|)}$ describes the alignment of the two classifiers.
If the classifiers are not aligned, i.e., $\Bw$ and $\overline{\Bw}$ have a large angle between them, then adversarial examples exist even if the margin of the classifier is larger than that of the ground-truth classifier.

Finally, adversarial examples with small perturbation are possible if $|\Bw^\top\Bx|\ll\norm{\Bw}$. The extreme case $\Bw^\top\Bx=0$ means that $\Bx$ lies on the decision boundary of $h$, and if $|\Bw^\top\Bx|\ll \norm{\Bw}$ then $\Bx$ is close to the decision boundary of $h$.

\begin{proof}[of Theorem \ref{thm:LinearClassifierAdversarial}]
  We verify that $\Bx'$ in \eqref{eq:definitionOfTheAdversarialExample} satisfies the conditions of an adversarial example in Definition \ref{def:adversarialExamples}.
    In the following we will use that due to $h(\Bx)g(\Bx)>0$
    \begin{equation}\label{eq:BwBxg0}
      g(\Bx) = {\rm sign}(\overline{\Bw}^\top \Bx)
      ={\rm sign}(\Bw^\top \Bx) = h(\Bx)\neq 0.
    \end{equation}

  First, it holds
\begin{align*}
	\|\Bx - \Bx'\| = \left\| \frac{\eps + |\Bw^\top \Bx|}{\|\Bw\|^2} \Bw\right\| %
  =\frac{\eps + |\Bw^\top \Bx|}{\|\Bw\|} = \delta.
\end{align*}

Next we show $g(\Bx)g(\Bx') > 0$, i.e.\ that
$(\overline{\Bw}^\top \Bx) (\overline{\Bw}^\top \Bx')$ is positive.
Plugging in the definition of $\Bx'$, this term reads
\begin{align}\label{eq:divideTheRHSofThis}
	 \overline{\Bw}^\top \Bx \left(\overline{\Bw}^\top\Bx - h(\Bx) \frac{\eps + |\Bw^\top \Bx|}{\|\Bw\|^2}\overline{\Bw}^\top \Bw\right) &= |\overline{\Bw}^\top \Bx|^2 - |\overline{\Bw}^\top \Bx| \frac{\eps + |\Bw^\top \Bx|}{\|\Bw\|^2}\overline{\Bw}^\top \Bw\nonumber\\
	 &\geq |\overline{\Bw}^\top \Bx|^2 - |\overline{\Bw}^\top \Bx| \frac{\eps + |\Bw^\top \Bx|}{\|\Bw\|^2}|\overline{\Bw}^\top \Bw|,
\end{align}
where %
the equality holds because
$h(\Bx) = g(\Bx) = \mathrm{sign}(\overline{\Bw}^\top \Bx)$ by \eqref{eq:BwBxg0}. 
Dividing the right-hand side of \eqref{eq:divideTheRHSofThis} by $|\overline{\Bw}^\top \Bx| \|\overline{\Bw}\|$, which is positive by \eqref{eq:BwBxg0}, we obtain 
\begin{align}\label{eq:ThisIsPositive143}
	\frac{|\overline{\Bw}^\top \Bx|}{\|\overline{\Bw}\|} - \frac{\eps + |\Bw^\top \Bx|}{\|\Bw\|} \frac{{|\overline{\Bw}^\top \Bw|}}{\|\Bw\| \|\overline{\Bw}\|}.
\end{align}
The term \eqref{eq:ThisIsPositive143} is positive thanks to \eqref{eq:adversarialExampleEquation}. 

Finally, we check that ${0\neq }h(\Bx') \neq h(\Bx)$,
i.e.\ $(\Bw^\top \Bx) (\Bw^\top \Bx')<0$. 
We have that 
\begin{align*}
	(\Bw^\top \Bx) (\Bw^\top \Bx') & = |\Bw^\top \Bx|^2 - \Bw^\top \Bx h(\Bx) \frac{\eps + |\Bw^\top \Bx|}{\|\Bw\|^2}\Bw^\top \Bw\\
	& = |\Bw^\top \Bx|^2 - |\Bw^\top \Bx| (\eps + |\Bw^\top \Bx|)<0,
\end{align*}
where we used that $h(\Bx) = \mathrm{sign}(\Bw^\top \Bx)$.
This completes the proof.
\end{proof}

Theorem \ref{thm:LinearClassifierAdversarial} readily implies the following proposition for \emph{affine} classifiers.

\begin{proposition}\label{prop:AdversarialAffine}
  Let $\Bw$, $\overline{\Bw} \in \R^{d}$ and $b$, $\overline{b}\in\R$. For $\Bx\in\R^d$ let
  $h(\Bx) = \mathrm{sign}(\Bw^\top \Bx + b)$ be a classifier and let $g(\Bx) = \mathrm{sign}(\overline{\Bw}^\top \Bx + \overline{b})$ %
  be the ground-truth classifier. 
	
	For every $\Bx\in \R^{d}$ with $\overline{\Bw}^\top\Bx \neq 0$, $h(\Bx)g(\Bx)>0$, and all $\eps \in (0, |\Bw^\top \Bx + b|)$ such that 
	\begin{align*}
		\frac{|\overline{\Bw}^\top \Bx + \overline{b}|^2}{\|\overline{\Bw}\|^2 + b^2}  > \frac{(\eps  + |\Bw^\top \Bx + b|)^2}{\|\Bw\|^2 + b^2}  \frac{(\Bw^\top \overline{\Bw} + b\overline{b})^2}{(\|\Bw\|^2 + b^2) (\| \overline{\Bw}\|^2 + \overline{b}^2)}
	\end{align*}
	it holds that 
	\[
	\Bx' = \Bx - h(\Bx) \frac{\eps + |\Bw^\top \Bx + b|}{\|\Bw\|^2} \Bw
	\] is an adversarial example with perturbation $\delta =  (\eps  + |\Bw^\top \Bx + b|)/{\|\Bw\|}$ to $\Bx$.
\end{proposition}

The proof is left to the reader, see Exercise \ref{ex:proofOfCorollaryAffine}.

\begin{figure}[htb]
	\centering
	\includegraphics[width = 0.7\textwidth]{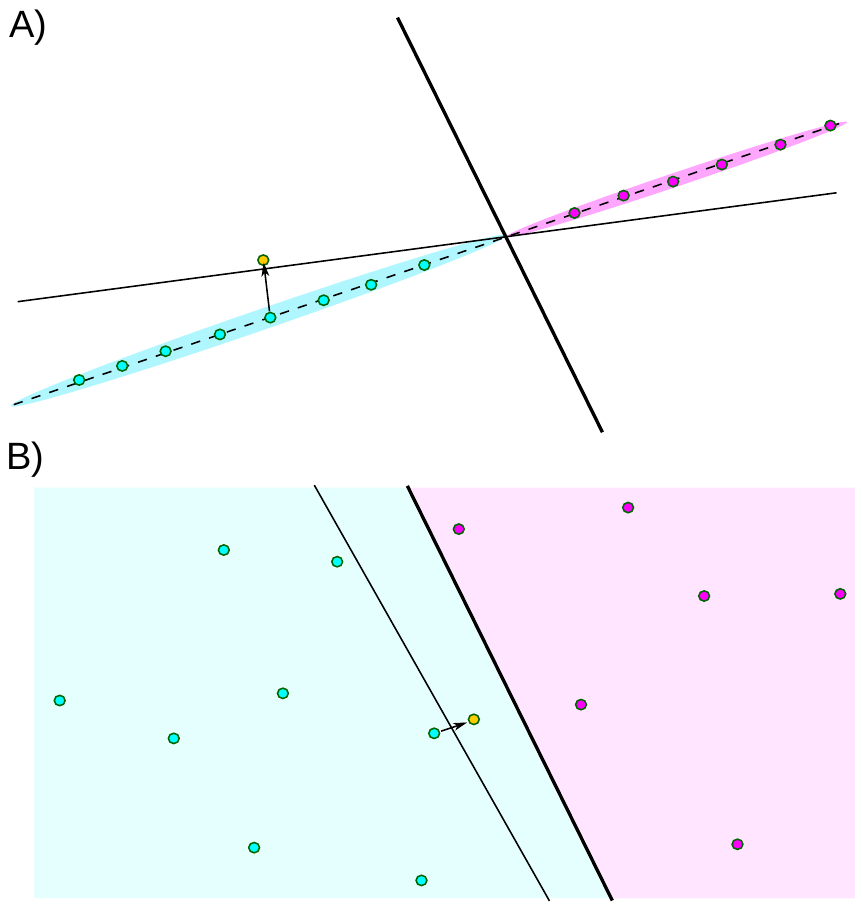}
	\put(-150, 290){${\rm DB}_g$}
	\put(-280, 250){${\rm DB}_h$}
	\put(-230, 215){$x$}
	\put(-230, 258){$x'$}
	\put(-142, 110){${\rm DB}_g$}
	\put(-195, 90){${\rm DB}_h$}
	\put(-173, 60){$x$}
	\put(-143, 70){$x'$}
	\caption{Illustration of the two types of adversarial examples %
          in Examples \ref{example:AdversarialsLinearI} and \ref{example:AdversarialsLinearII}.
In %
panel A) %
the feature support $D_\Bx$ corresponds to the dashed line.
We depict the two decision boundaries
${\rm DB}_h = \set{\Bx}{\Bw^\top\Bx = 0}$ of
$h(\Bx) = \mathrm{sign}(\Bw^\top\Bx)$ and
${\rm DB}_g = \set{\Bx}{\overline{\Bw}^\top\Bx = 0}$
$g(\Bx) = \mathrm{sign}(\overline{\Bw}^\top\Bx)$. %
Both $h$ and $g$ perfectly classify every data point in $D_\Bx$.
One data point $\Bx$ is shifted outside of the support of the distribution in a way to change its label according to $h$.
This creates an adversarial example $\Bx'$.
In panel B) the data distribution is globally supported.
However, $h$ and $g$ do not coincide.
Thus the decision boundaries ${\rm DB}_h$ and ${\rm DB}_g$ do not coincide.
Moving data points %
across ${\rm DB}_h$
can create adversarial examples, as depicted by %
$\Bx$ and $\Bx'$.}
	\label{fig:IllustrationAdversarialExamplesLinear}
\end{figure}

Let us now study two cases of linear classifiers, %
which allow for different types of adversarial examples.
In the following two examples, the ground-truth classifier $g:\R^{d} \to \{-1,1\}$ is given by $g(\Bx) = \mathrm{sign}(\overline{\Bw}^\top \Bx)$ for $\overline{\Bw} \in \R^d$ with $\|\overline{\Bw}\| = 1$.

For the first example, we construct a Bayes classifier $h$ admitting adversarial examples in the complement of the feature support. This corresponds to case \ref{item:BayesNonexhaustive} in Section \ref{sec:BayesClassifier}.

\begin{example}\label{example:AdversarialsLinearI}
	 Let $\CD$ be the uniform distribution on 
	 \[
	 	\set{ (\lambda \overline{\Bw}, g(\lambda \overline{\Bw}))}{\lambda \in [-1,1] \setminus \{0\}}\subseteq \R^d\times \{-1,1\}.
	 \]
         The feature support equals
	 \[
	 D_\Bx = \set{\lambda \overline{\Bw}}{\lambda \in [-1,1] \setminus \{0\}} \subseteq {\rm span}\{\overline{\Bw}\}.
	 \]	 
	 Next %
         fix $\alpha \in (0,1)$ and set $\Bw \dfn \alpha\overline{\Bw} + (1-\alpha) \Bv$ for
           some $\Bv \in \overline{\Bw}^\perp$ with $\|\Bv\| = 1$, so that $\norm{\Bw}=1$. We let $h(\Bx) \dfn \mathrm{sign}(\Bw^\top \Bx)$. We now show that every $\Bx\in D_\Bx$ satisfies the assumptions of Theorem \ref{thm:LinearClassifierAdversarial}, and therefore admits an adversarial example.

	 Note that $h(\Bx) = g(\Bx)$ for every $\Bx \in D_\Bx$.
         Hence $h$ is a Bayes classifier. Now fix $\Bx\in D_\Bx$.
         Then
         $|\Bw^\top \Bx| \leq \alpha|\overline{\Bw}^\top \Bx|$, so that
         \eqref{eq:adversarialExampleEquation} is satisfied. %
	 Furthermore, for every $\eps >0$ %
         it holds that
	 \[
           \delta\dfn \frac{\eps  + |\Bw^\top \Bx|}{\|\Bw\|} \leq \eps  + \alpha.
              \]
	 Hence, for $\eps < |\Bw^\top \Bx|$ it holds by Theorem \ref{thm:LinearClassifierAdversarial} %
         that there exists an adversarial example with perturbation less than $\eps  + \alpha$.
For %
small $\alpha$, the situation is depicted in the upper panel of Figure \ref{fig:IllustrationAdversarialExamplesLinear}.
\end{example}

For the second example, we construct a distribution with global feature support and a classifier which is not a Bayes classifier. This corresponds to case \ref{item:NonBayesExhaustive} in Section \ref{sec:BayesClassifier}.

\begin{example}\label{example:AdversarialsLinearII}
  Let $\mathcal{D}_\Bx$ be a distribution on $\R^d$ with positive Lebesgue density everywhere %
  outside the decision boundary ${\rm DB}_g = \set{\Bx}{\overline{\Bw}^\top\Bx = 0}$ of $g$.
  We define $\mathcal{D}$ to be the distribution of $(X, g(X))$ for $X \sim \CD_\Bx$.
In addition, let $\Bw \notin \{\pm \overline{\Bw}\}$, $\|\Bw\| = 1$ and $h(\Bx) = \mathrm{sign}(\Bw^\top \Bx)$.
We exclude $\Bw=-\overline{\Bw}$ because, in this case, every prediction of $h$ is wrong. Thus no adversarial examples are possible.

By construction the feature support is given by $D_\Bx=\R^d$.
Moreover, $h^{-1}(\{-1\})$, $h^{-1}(\{1\})$ and $g^{-1}(\{-1\})$, $g^{-1}(\{1\})$ are half spaces, which implies in the notation of \eqref{eq:thisNeedsToBeReferencedInTheExercise} that %
	 \[
	 	\mathrm{dist}(C_{\pm 1} \cap D_\Bx, F_{\pm 1} \cap D_\Bx) = \mathrm{dist}(C_{\pm 1}, F_{\pm 1}) = 0.
	 \]
	 Hence, for every $\delta>0$ there is a positive probability of observing $\Bx$ to which an adversarial example with perturbation $\delta$ exists. 

The situation is depicted in the lower panel of Figure \ref{fig:IllustrationAdversarialExamplesLinear}.
\end{example}

\section{ReLU neural networks}

So far we discussed classification by affine classifiers.
A binary classifier based on a ReLU neural network is a function $\R^d \ni \Bx\mapsto \mathrm{sign}(\Phi(\Bx))$, where $\Phi$ is a ReLU neural network. 
As noted in \cite{szegedy2013intriguing}, %
the arguments for affine classifiers, see Proposition \ref{prop:AdversarialAffine}, %
can be %
applied to the affine pieces of $\Phi$, to show existence of adversarial examples.

Consider a ground-truth classifier $g\colon \R^d \to \{-1,0,1\}$.
For each %
$\Bx \in \R^d$ we define the geometric margin of $g$ at $\Bx$ as 
\begin{align}\label{eq:geometricMarginOfgatBx}
	\mu_g(\Bx) \dfn \mathrm{dist}(\Bx, g^{-1}(\{g(\Bx)\})^c ),	
\end{align}
i.e., as the distance of $\Bx$ to the closest element that is classified differently from $\Bx$ or the infimum over all distances to elements from other classes if no closest element exists.
Additionally, we denote the distance of $\Bx$ to the closest adjacent affine piece by
\begin{align}\label{eq:distanceToNearestPiece}
	\nu_{\Phi}(\Bx) \dfn \mathrm{dist}(\Bx, A_{\Phi, \Bx}^c),
\end{align}
where $A_{\Phi, \Bx}$ is the largest connected region on which $\Phi$ is affine and which contains $\Bx$.
We have the following theorem.

\begin{theorem}\label{thm:AdversarialExForReLU}
  Let $\Phi\colon \R^d \to \R$ and for $\Bx\in\R^d$ let $h(\Bx) = \mathrm{sign}(\Phi(\Bx))$.
  Denote by $g\colon \R^d \to \{-1,0,1\}$ the ground-truth classifier.
  Let $\Bx\in\R^d$ and $\eps>0$ be such that $\nu_{\Phi}(\Bx)>0$, $g(\Bx)\neq 0$, $\nabla\Phi(\Bx) \neq 0$ and
\begin{align*}
	\mu_g(\Bx), \nu_{\Phi}(\Bx) > \frac{\eps + |\Phi(\Bx)|}{\| \nabla\Phi(\Bx)\|}.
\end{align*}

Then
\begin{align*}
	\Bx' \dfn \Bx - h(\Bx) \frac{\eps + |\Phi(\Bx)|}{\| \nabla\Phi(\Bx)\|^2} \nabla\Phi(\Bx)
\end{align*}
is an adversarial example to $\Bx$ with perturbation $\delta = ({\eps + |\Phi(\Bx)|})/{\| \nabla\Phi(\Bx)\|}$.
\end{theorem}

\begin{proof}
We %
show that $\Bx'$ %
satisfies the properties in Definition \ref{def:adversarialExamples}.

By construction $\|\Bx - \Bx'\| \leq \delta$.
Since $\mu_g(\Bx)>\delta$ it follows that $g(\Bx) = g(\Bx')$.
Moreover, by assumption $g(\Bx) \neq 0$, and thus $g(\Bx)g(\Bx')>0$.

It only remains to show that $h(\Bx') \neq h(\Bx)$.
Since $\delta < \nu_{\Phi}(\Bx)$, we have that $\Phi(\Bx) = \nabla \Phi(\Bx)^\top \Bx + b$ and $\Phi(\Bx') = \nabla \Phi(\Bx)^\top \Bx' + b$ for some $b \in \R$.
Therefore, 
\begin{align*}
	\Phi(\Bx) - \Phi(\Bx') &= \nabla \Phi(\Bx)^\top(\Bx - \Bx') = \nabla \Phi(\Bx)^\top\left(h(\Bx) \frac{\eps + |\Phi(\Bx)|}{\| \nabla\Phi(\Bx)\|^2} \nabla\Phi(\Bx)\right)\\
	&=h(\Bx) (\eps + |\Phi(\Bx)|).
\end{align*}
	Since $h(\Bx)|\Phi(\Bx)| = \Phi(\Bx)$ it follows that $\Phi(\Bx') = - h(\Bx) \eps$.
Hence, $h(\Bx') = - h(\Bx)$, which completes the proof.
\end{proof}

\begin{remark}
We look at the key %
parameters in Theorem \ref{thm:AdversarialExForReLU} to understand which factors facilitate adversarial examples.
\begin{itemize}
\item \emph{The geometric margin of the ground-truth classifier $\mu_{g}(\Bx)$:} To make %
the construction possible, we need to be sufficiently far away from points that belong to a different class than $\Bx$ or to the nonrelevant class.
\item \emph{The %
    distance to the next affine piece
    $\nu_{\Phi}(\Bx)$:} Since we are looking for an adversarial example within the same affine piece as $\Bx$, we need this piece to %
  be sufficiently large.
\item \emph{The perturbation $\delta$:} The perturbation is given by $({\eps + |\Phi(\Bx)|})/{\| \nabla\Phi(\Bx)\|}$, which %
  depends on
  the classification margin $|\Phi(\Bx)|$ of the ReLU classifier and its sensitivity to inputs $\| \nabla\Phi(\Bx)\|$.
  For adversarial examples to be possible,
  we either want a small classification margin of $\Phi$ or a %
  high sensitivity of $\Phi$ to its inputs. 
\end{itemize}
\end{remark}

\section{Robustness}
Having established that adversarial examples can arise in various ways
under mild assumptions, we now turn our attention to conditions that prevent their existence.

\subsection{Global Lipschitz regularity}

We have repeatedly observed in the previous sections that a large value of $\|\Bw\|$ for linear classifiers ${\rm sign}(\Bw^\top\Bx)$, or $\|\nabla\Phi(\Bx)\|$ for ReLU classifiers ${\rm sign}(\Phi(\Bx))$, facilitates the occurrence of adversarial examples.
Naturally, both these values are upper bounded by the Lipschitz constant %
of the classifier's inner functions $\Bx\mapsto\Bw^\top\Bx$ and $\Bx\mapsto\Phi(\Bx)$.
Consequently, it was stipulated %
early on that bounding the Lipschitz constant %
of the inner functions could be an effective measure against
adversarial examples \cite{szegedy2013intriguing}.

We have the following result for general classifiers of the form $\Bx \mapsto \mathrm{sign}(\Phi(\Bx))$.

\begin{proposition}\label{prop:LipschitzPreventsAdversarials}
Let $\Phi \colon \R^d \to \R$ be $C_L$-Lipschitz with $C_L >0$, and let $s>0$.
Let $h(\Bx) =  \mathrm{sign}(\Phi(\Bx))$ be a classifier, and let $g\colon  \R^d \to \{-1,0,1\}$ be a ground-truth classifier.
Moreover, %
let $\Bx\in\R^d$ be such that %
\begin{align}\label{eq:propOfxPreventingAdversarials}
	\Phi(\Bx) g(\Bx) \geq s. %
\end{align}
Then there does not exist an adversarial example to $\Bx$ of perturbation $\delta < s/C_L$.
\end{proposition}

\begin{proof}
Let $\Bx\in \R^d$ satisfy \eqref{eq:propOfxPreventingAdversarials} and assume that $\|\Bx' - \Bx \| \leq \delta$. %
The Lipschitz continuity of $\Phi$ implies
\begin{align*}
	|\Phi(\Bx') - \Phi(\Bx) | < s.
\end{align*}
Since $|\Phi(\Bx)| \geq s$ we conclude that $\Phi(\Bx')$ has the same sign as $\Phi(\Bx)$ which shows that $\Bx'$ cannot be an adversarial example to $\Bx$. 
\end{proof}

\begin{remark}
  As we have seen in Lemma \ref{lem:LipschitzEstimateNN2324}, 
we can %
bound the Lipschitz constant of ReLU neural networks by restricting the magnitude and number of their weights and the number of layers.
\end{remark}

There has been some criticism to results of this form, see, e.g., \cite{huster2019limitations}, since an assumption on the Lipschitz constant may potentially restrict the capabilities of the neural network too much.
We %
next present a result that shows under which assumptions on the training set, %
there exists a neural network that %
classifies the training set correctly, but does not allow for %
adversarial examples within the training set.

\begin{theorem}\label{thm:LipschitzInterpolationAdversarials}
Let $m \in \N$, let $g \colon{\R^d} \to \{-1,0,1\}$ be a ground-truth classifier, and let 
$(\Bx_i, g(\Bx_i))_{i=1}^m \in (\R^{d} \times \{-1,1\})^m$.
Assume that
\[
	\sup_{i \neq j} \frac{|g(\Bx_i) - g(\Bx_j)|}{\norm{\Bx_i - \Bx_j}} \dfnn \widetilde{M} >0.
\]
Then there exists a ReLU neural network $\Phi$ with  $\depth(\Phi)=O(\log(m))$ and $\wdth(\Phi)=O(dm)$ such that for all $i = 1,\dots, m$
\[
	\mathrm{sign}(\Phi(\Bx_i)) = g(\Bx_i) 
\]
and there is no adversarial example of perturbation $\delta = 1/\widetilde{M}$ to $\Bx_i$. 
\end{theorem}

\begin{proof}
The result follows directly from Theorem \ref{thm:optimalLipschitz} and Proposition \ref{prop:LipschitzPreventsAdversarials}.
The reader is invited to complete the argument in Exercise \ref{ex:proofOfLipschitzInterpolationAdversarials}.
\end{proof}

\subsection{Local regularity}

One issue with %
upper bounds involving global Lipschitz %
constants such as those in Proposition \ref{prop:LipschitzPreventsAdversarials}, is that these bounds may be %
quite large for deep neural networks. 
For example, the upper bound given in Lemma \ref{lem:LipschitzEstimateNN2324} is
\[
	\|\Phi(\Bx) - \Phi(\Bx')\|_\infty \leq C_\sigma^L \cdot (B d_{\rm max})^{L+1} \|\Bx-\Bx' \|_\infty
\]
which grows exponentially with the depth of the neural network.
However, %
in practice this bound may be pessimistic, and locally the neural network might have significantly smaller gradients than the global Lipschitz constant.

Because of this, it is reasonable to study results preventing adversarial examples under \emph{local} Lipschitz bounds.
Such a result together with an algorithm providing bounds on the local Lipschitz constant was proposed in \cite{hein2017formal}.
We %
state the theorem adapted to our set-up.

\begin{theorem}\label{thm:Hein-local-Lipschitz}
Let $h \colon \R^d \to \{-1,1\}$ be a classifier of the form $h(\Bx) = \mathrm{sign}(\Phi(\Bx))$ and let $g \colon \R^d \to \{-1,0,1\}$ be the ground-truth classifier.
Let $\Bx \in \R^d$ %
satisfy $g(\Bx) \neq 0$, and set
\begin{align}\label{eq:heinRequirementForLowerBound}
  \alpha \dfn \max_{R >0} \min \left\{
  \Phi(\Bx) g(\Bx) \Big/ \sup_{\substack{\norm[\infty]{\By-\Bx}\le R\\ \By\neq\Bx}} \frac{|\Phi(\By) - \Phi(\Bx)|}{\|\Bx-\By\|_\infty}
  , R \right\},
\end{align}
where the minimum is understood to be $R$ in case the supremum is zero.
Then there are no adversarial examples to $\Bx$ with perturbation $\delta< \alpha$.
\end{theorem}

\begin{proof}
	Let $\Bx \in \R^d$ be as in the statement of the theorem. 
	Assume, towards a contradiction, that for $0<\delta < \alpha$ satisfying \eqref{eq:heinRequirementForLowerBound}, there exists an adversarial example $\Bx'$ to $\Bx$ with perturbation $\delta$.

        If the supremum in \eqref{eq:heinRequirementForLowerBound} is zero,
        then $\Phi$ is constant on a ball of radius $R$ around
        $\Bx$. In particular for $\norm{\Bx'-\Bx}\le\delta<R$ it holds that $h(\Bx')=h(\Bx)$ and $\Bx'$ cannot be an adversarial example.

      Now assume the supremum in \eqref{eq:heinRequirementForLowerBound} is not zero.
	It holds by \eqref{eq:heinRequirementForLowerBound} for $\delta<R$, that 
	\begin{align}\label{eq:SimplifiedHeinEq}
		\delta < \Phi(\Bx) g(\Bx)\Big/ \sup_{\substack{\norm[\infty]{\By-\Bx}\le R\\ \By\neq\Bx}} %
          \frac{|\Phi(\By) - \Phi(\Bx)|}{\|\Bx-\By\|_\infty}.
	\end{align}
	Moreover, 
	\begin{align*}
          |\Phi(\Bx') - \Phi(\Bx)| &\leq %
                                     \sup_{\substack{\norm[\infty]{\By-\Bx}\le R\\ \By\neq\Bx}}
                                           \frac{|\Phi(\By) - \Phi(\Bx)|}{\|\Bx-\By\|_\infty} \| \Bx - \Bx'\|_\infty\\
                                   &\leq %
                                     \sup_{\substack{\norm[\infty]{\By-\Bx}\le R\\ \By\neq\Bx}}
           \frac{|\Phi(\By) - \Phi(\Bx)|}{\|\Bx-\By\|_\infty} \delta < \Phi(\Bx)g(\Bx),
	\end{align*}
	where we applied \eqref{eq:SimplifiedHeinEq} in the last line.
	It follows that 
	\begin{align*}
		g(\Bx) \Phi(\Bx') &= g(\Bx) \Phi(\Bx) + g(\Bx) (\Phi(\Bx') - \Phi(\Bx))\\
		&\geq g(\Bx) \Phi(\Bx) - |\Phi(\Bx') - \Phi(\Bx)|
		> 0.
	\end{align*}
	This rules out $\Bx'$ as an adversarial example.
\end{proof}

The supremum in \eqref{eq:heinRequirementForLowerBound} is bounded by the Lipschitz constant of $\Phi$ on $B_R(\Bx)$. Thus Theorem \ref{thm:Hein-local-Lipschitz} depends only on the local Lipschitz constant of $\Phi$.
One obvious criticism of this %
result is that the computation of \eqref{eq:heinRequirementForLowerBound} is potentially prohibitive. 
We %
next show a different result, for which the assumptions can immediately be checked by applying a simple algorithm that we present subsequently.
  
To state the following proposition, for a continuous function $\Phi:\R^d\to\R$ and $\delta>0$ we define for $\Bx \in \R^d$ and $\delta >0$
\begin{align}
	z^{\delta, \mathrm{max}} &\dfn \max\set{\Phi(\By)}{\|\By - \Bx\|_\infty \leq \delta}\\
	z^{\delta, \mathrm{min}} &\dfn \min\set{\Phi(\By)}{\|\By - \Bx\|_\infty \leq \delta}.
\end{align}

\begin{proposition}\label{prop:maxMinPreventAdversarialEx}
	Let $h\colon \R^d \to \{-1,1\}$ be a classifier of the form $h(\Bx) = \mathrm{sign}(\Phi(\Bx))$ and $g \colon \R^d \to \{-1,0,1\}$, let $\Bx$ be such that $h(\Bx) = g(\Bx)$. 
	Then $\Bx$ does not have an adversarial example of perturbation $\delta$ if $z^{\delta, \mathrm{max}}  z^{\delta, \mathrm{min}}>0$. 
\end{proposition}

\begin{proof}
The proof is immediate, since $z^{\delta, \mathrm{max}}  z^{\delta, \mathrm{min}}>0$ implies that all points in a $\delta$ neighborhood of $\Bx$ are classified the same.
\end{proof}

To apply \eqref{prop:maxMinPreventAdversarialEx}, %
we only have to compute $z^{\delta, \mathrm{max}}$ and $z^{\delta, \mathrm{min}}$.
It turns out that if $\Phi$ is a neural network, then $z^{\delta, \mathrm{max}}$, $z^{\delta, \mathrm{min}}$ can be %
approximated by a computation similar to a forward pass of $\Phi$.
Denote by $|\BA|$ the matrix obtained by taking the absolute value of each entry of the matrix $\BA$. Additionally, we define
\[
	\BA^+ = (|\BA| + \BA)/2 \text{ and } \BA^- = (|\BA| - \BA)/2.
\]
The idea behind the Algorithm \ref{alg:ForwardPassPlusCertificate} is common in the area of neural network verification, see, e.g., \cite{gehr2018ai2, fischer2019dl2, baader2019universal, wang2022interval}.

\begin{algorithm}
	\caption{Compute $\Phi(\Bx)$, $z^{\delta, \mathrm{max}}$ and $z^{\delta, \mathrm{min}}$ for a given neural network
}
	\label{alg:ForwardPassPlusCertificate}
	\begin{algorithmic}%
          \STATE \textbf{Input:} weight matrices $\BW^{(\ell)}\in\R^{d_{\ell+1}\times d_\ell}$ and bias vectors $\Bb^{(\ell)}\in\R^{d_{\ell+1}}$ for $\ell = 0, \dots, L$ with $d_{L+1}= 1$, monotonous activation function $\sigma$, input vector $\Bx \in \R^{d_0}$, neighborhood size $\delta>0$
          \STATE \textbf{Output:} Bounds for $z^{\delta,\max}$ and $z^{\delta,\min}$
          \STATE
		\STATE $\Bx^{(0)}= \Bx$
		\STATE $\delta^{(0), \mathrm{up}} = \delta \ind \in \R^{d_0}$
		\STATE $\delta^{(0), \mathrm{low}} = \delta \ind \in \R^{d_0}$

		\FOR{$\ell = 0,\dots,L-1$}
		\STATE $\Bx^{(\ell+1)} = \sigma(\BW^{(\ell)} \Bx^{(\ell)} + \Bb^{(\ell)})$
		\STATE $\delta^{(\ell+1), \mathrm{up}} = \sigma(\BW^{(\ell)} \Bx^{(\ell)} + (\BW^{(\ell)})^+ \delta^{(\ell), \mathrm{up}}  + (\BW^{(\ell)})^- \delta^{(\ell), \mathrm{low}}+ \Bb^{(\ell)}) - \Bx^{(\ell+1)}$
		\STATE $\delta^{(\ell+1), \mathrm{low}} =  \Bx^{(\ell+1)} - \sigma(\BW^{(\ell)} \Bx^{(\ell)} -  (\BW^{(\ell)})^+ \delta^{(\ell), \mathrm{low}}  - (\BW^{(\ell)})^- \delta^{(\ell), \mathrm{up}} + \Bb^{(\ell)})$
		\ENDFOR
		\STATE $\Bx^{(L+1)} = \BW^{(L)} \Bx^{(L)} + \Bb^{(L)}$
		\STATE $\delta^{(L+1), \mathrm{up}} = (\BW^{(L)})^+ \delta^{(L), \mathrm{up}}  + (\BW^{(L)})^- \delta^{(L), \mathrm{low}}$
		\STATE $\delta^{(L+1), \mathrm{low}} = (\BW^{(L)})^+ \delta^{(L), \mathrm{low}}  + (\BW^{(L)})^- \delta^{(L), \mathrm{up}}$

		\RETURN $\Bx^{(L+1)}$, $\Bx^{(L+1)} + \delta^{(L+1), \mathrm{up}}$, $\Bx^{(L+1)} - \delta^{(L+1), \mathrm{low}}$
	\end{algorithmic}
      \end{algorithm}

\begin{remark}
Up to constants, Algorithm \ref{alg:ForwardPassPlusCertificate} has the same computational complexity as a %
forward pass, also see Algorithm \ref{alg:backprop}. %
In addition, in contrast to upper bounds based on estimating the global Lipschitz constant of $\Phi$ via its weights, the upper bounds found via Algorithm \ref{alg:ForwardPassPlusCertificate} include the effect of the activation function $\sigma$.
For example, if $\sigma$ is the ReLU, then we may often end up in a situation, where $\delta^{(\ell), \mathrm{up}}$ or $\delta^{(\ell), \mathrm{low}}$ can have many entries that are $0$.
If an entry of 
$\BW^{(\ell)} \Bx^{(\ell)} + \Bb^{(\ell)}$
is nonpositive, then it is guaranteed that the associated entry in $\delta^{(\ell), \mathrm{low}}$ will be zero.
Similarly, if $\BW^{(\ell)}$ has only few positive entries, then most of the entries of $\delta^{(\ell), \mathrm{up}}$ are not propagated to $\delta^{(\ell+1), \mathrm{up}}$.
\end{remark}

Next, we prove that Algorithm \ref{alg:ForwardPassPlusCertificate} indeed produces sensible output.

\begin{proposition}
Let $\Phi$ be a neural network with weight matrices $\BW^{(\ell)}\in\R^{d_{\ell+1}\times d_\ell}$ and bias vectors $\Bb^{(\ell)}\in\R^{d_{\ell+1}}$ for $\ell = 0, \dots, L$, and a monotonically increasing activation function $\sigma$. 

Let $\Bx \in \R^d$.
Then the output of Algorithm \ref{alg:ForwardPassPlusCertificate} satisfies
\[
	\Bx^{L+1} + \delta^{(L+1), \mathrm{up}} > z^{\delta, \mathrm{max}} \text{ and } \Bx^{L+1} - \delta^{(L+1), \mathrm{low}} < z^{\delta, \mathrm{min}}.
\]
\end{proposition}

\begin{proof}
Fix $\By$, $\Bx \in \R^d$ with $\|\By - \Bx\|_\infty \leq \delta$ and let $\By^{(\ell)}$, $\Bx^{(\ell)}$ for $\ell = 0, \dots, L+1$ be %
as in Algorithm \ref{alg:ForwardPassPlusCertificate} applied to $\By$, $\Bx$, respectively.
Moreover, let $\delta^{\ell, \mathrm{up}}$, $\delta^{\ell, \mathrm{low}}$  for $\ell = 0, \dots, L+1$ be as in Algorithm \ref{alg:ForwardPassPlusCertificate} applied to $\Bx$.
We will prove by induction over $\ell = 0, \dots, L+1$ that
\begin{align}\label{eq:thiswillbeshownviaInductionoverell}
	\By^{(\ell)} - \Bx^{(\ell)} \leq \delta^{\ell, \mathrm{up}}\qquad\text{and}\qquad
	\Bx^{(\ell)} - \By^{(\ell)} \leq \delta^{\ell, \mathrm{low}},
\end{align}
where the inequalities are understood entry-wise for vectors.
Since $\By$ was arbitrary %
this then proves the result.

The case $\ell = 0$ follows immediately from $\|\By - \Bx\|_\infty \leq \delta$.
Assume now, that the statement was shown for $\ell < L$. 
We have that 
\begin{align*}
	\By^{(\ell+1)} - \Bx^{(\ell+1)} - \delta^{\ell+1, \mathrm{up}}= 	&\sigma(\BW^{(\ell)} \By^{(\ell)} + \Bb^{(\ell)})\\
	&- \sigma\big(\BW^{(\ell)} \Bx^{(\ell)} + (\BW^{(\ell)})^+ \delta^{(\ell), \mathrm{up}}  + (\BW^{(\ell)})^- \delta^{(\ell), \mathrm{low}} + \Bb^{(\ell)}\big).
\end{align*}
The monotonicity of $\sigma$ implies that 
\[
	\By^{(\ell+1)} - \Bx^{(\ell+1)} \leq \delta^{\ell+1, \mathrm{up}}
\]
if 
\begin{align}
	\label{eq:firstPartOfCertificationEstimate}
	\BW^{(\ell)} \By^{(\ell)} \leq \BW^{(\ell)} \Bx^{(\ell)} + (\BW^{(\ell)})^+ \delta^{(\ell), \mathrm{up}}  + (\BW^{(\ell)})^- \delta^{(\ell), \mathrm{low}}.
\end{align}
To prove \eqref{eq:firstPartOfCertificationEstimate}, we observe that 
\begin{align*}
	\BW^{(\ell)} (\By^{(\ell)} - \Bx^{(\ell)}) &= (\BW^{(\ell)})^+ (\By^{(\ell)} - \Bx^{(\ell)}) -  (\BW^{(\ell)})^- (\By^{(\ell)} - \Bx^{(\ell)})\\
	&=(\BW^{(\ell)})^+ (\By^{(\ell)} - \Bx^{(\ell)}) +  (\BW^{(\ell)})^- (\Bx^{(\ell)} - \By^{(\ell)})\\
	&\leq (\BW^{(\ell)})^+ \delta^{(\ell), \mathrm{up}}  + (\BW^{(\ell)})^- \delta^{(\ell), \mathrm{low}},
\end{align*}
where we used the induction assumption in the last line.
This shows the first estimate %
in \eqref{eq:thiswillbeshownviaInductionoverell}.
Similarly, 
\begin{align*}
	&\Bx^{(\ell+1)}  - \By^{(\ell+1)} - \delta^{\ell+1, \mathrm{low}} \\
	&= \sigma(\BW^{(\ell)} \Bx^{(\ell)} - (\BW^{(\ell)})^+ \delta^{(\ell), \mathrm{low}}  - (\BW^{(\ell)})^- \delta^{(\ell), \mathrm{up}} + \Bb^{(\ell)}) - \sigma(\BW^{(\ell)} \By^{(\ell)} + \Bb^{(\ell)}).
\end{align*}
Hence, $\Bx^{(\ell+1)}  - \By^{(\ell+1)} \leq \delta^{\ell+1, \mathrm{low}}$ if 
\begin{align}
	\label{eq:firstPartOfCertificationEstimate2}
	\BW^{(\ell)} \By^{(\ell)} \geq \BW^{(\ell)} \Bx^{(\ell)} - (\BW^{(\ell)})^+ \delta^{(\ell), \mathrm{low}}  - (\BW^{(\ell)})^- \delta^{(\ell), \mathrm{up}}.
\end{align}
To prove \eqref{eq:firstPartOfCertificationEstimate2}, we observe that 
\begin{align*}
	\BW^{(\ell)} (\Bx^{(\ell)} -  \By^{(\ell)}) &= (\BW^{(\ell)})^+ (\Bx^{(\ell)} -  \By^{(\ell)}) - (\BW^{(\ell)})^- (\Bx^{(\ell)} -  \By^{(\ell)})\\
	&= (\BW^{(\ell)})^+ (\Bx^{(\ell)} -  \By^{(\ell)}) + (\BW^{(\ell)})^- (\By^{(\ell)} -  \Bx^{(\ell)})\\
	&\leq (\BW^{(\ell)})^+ \delta^{(\ell), \mathrm{low}}  + (\BW^{(\ell)})^- \delta^{(\ell), \mathrm{up}},
\end{align*}
where we used the induction assumption in the last line. 
This completes the proof of \eqref{eq:thiswillbeshownviaInductionoverell} for all $\ell \leq L$. 

The case $\ell = L+1$ follows %
by the same argument, but replacing $\sigma$ by the identity.
\end{proof}

\section*{Bibliography and further reading}
This chapter %
begins with the foundational paper \cite{szegedy2013intriguing}, but it should be remarked that adversarial examples for non-deep-learning models in machine learning
were studied earlier in
\cite{huang2011adversarial}.

The results in this chapter are inspired by various results in the literature, %
though they may not be found in precisely the same form. %
The overall setup is inspired by \cite{szegedy2013intriguing}. %
The explanation %
based on the high-dimensionality of the data given in Section \ref{sec:affineClass} was first formulated in \cite{szegedy2013intriguing} and \cite{goodfellow2014explaining}. The formalism reviewed in Section \ref{sec:BayesClassifier} is inspired by \cite{stutz2019disentangling}. The results on robustness via local Lipschitz properties are due to \cite{hein2017formal}. Algorithm \ref{alg:ForwardPassPlusCertificate} is covered by results in the area of network verifiability \cite{gehr2018ai2, fischer2019dl2, baader2019universal, wang2022interval}.
For a more comprehensive overview of modern approaches, we refer to the survey article \cite{ruan2021adversarial}.

Important directions not discussed %
in this chapter are the transferability of adversarial examples, defense mechanisms, and alternative adversarial operations. Transferability refers to the phenomenon that adversarial examples for one model often %
also fool other models, \cite{papernot2017practical,moosavi2017universal}. Defense mechanisms, i.e.,
techniques for specifically training a neural network to prevent adversarial examples,
include for example the Fast Gradient Sign Method of \cite{goodfellow2014explaining},
and more sophisticated recent approaches %
such as \cite{carlini2017towards}.
Finally, adversarial examples can be generated not only through additive perturbations, but also through smooth transformations of images, as demonstrated in \cite{alaifari2018adef, xiao2018spatially}.

\newpage
\section*{Exercises}

\begin{exercise}\label{ex:highDimBalls}
Prove \eqref{eq:HighDimBalls} by comparing the volume of the $d$-dimensional Euclidean unit ball with the volume of the $d$-dimensional 1-ball of radius $c$ for a given $c>0$.  
\end{exercise}

\begin{exercise}\label{ex:AdversarialExampleExample}
  Fix $\delta>0$.
  For a pair of classifiers $h$ and $g$ such that $C_1 \cup C_{-1} = \emptyset$ in \eqref{eq:thisNeedsToBeReferencedInTheExercise}, there trivially cannot %
exist any adversarial examples. 
Construct an example, of $h$, $g$, $\mathcal{D}$ such that $C_1$, $C_{-1} \neq \emptyset$, $h$ is not a Bayes classifier, and $g$ is such that no adversarial examples with a perturbation $\delta$ exist. 

Is this also possible if $g^{-1}(0) = \emptyset$? 	
\end{exercise}

\begin{exercise}\label{ex:proofOfCorollaryAffine}
  Prove Proposition \ref{prop:AdversarialAffine}.

  \emph{Hint}: Repeat the proof of Theorem \ref{thm:LinearClassifierAdversarial}. In the first part set $\Bx^{(\rm ext)} = (\Bx, 1)$, $\Bw^{(\rm  ext)} = (\Bw, b)$ and $\overline{\Bw}^{(\rm  ext)} = (\overline{\Bw}, \overline{b})$. Then show that $h(\Bx') \neq h(\Bx)$ by plugging in the definition of $\Bx'$.
\end{exercise}

\begin{exercise}\label{ex:proofOfLipschitzInterpolationAdversarials}
	Complete the proof of Theorem \ref{thm:LipschitzInterpolationAdversarials}.
      \end{exercise}

%% file: Architectures.tex
\chapter{Modern architectures}\label{chap:ModArchitectures}
Up to this point, this book has discussed the most classical type of
neural networks---feed-forward neural networks.  However, in practice,
a wide range of modifications and variants are employed.  In this
chapter, we discuss the most common ones, identify cases where it is
sensible to use them, and prove intuitive theoretical statements where
possible.

\section{Residual neural networks}
One of the key concepts in deep learning is that deep architectures
often outperform shallow ones.  Throughout this book, we have seen
many theoretical %
benefits of the power of depth, especially in terms of expressivity.
In practice, however, it is often hard to harness these advantages
because deep architectures prove to be more %
difficult to train.  One reason, which we have seen in Section
\ref{sec:activationFunctions}, is that deep architectures are prone to
the vanishing or exploding gradient phenomenon.

\begin{figure}[htb]
  \centering \includegraphics[width = 0.8\textwidth]{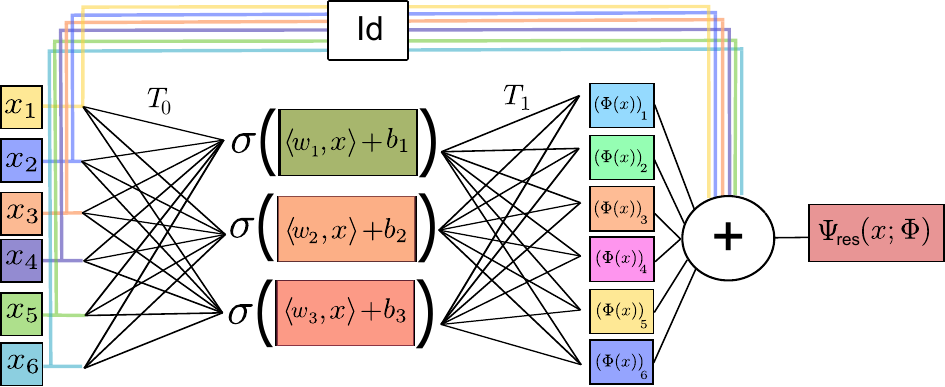}
  \caption{Sketch of a residual block with activation function
    $\sigma$, $d = 6$, $N = 3$, and a neural network $\Phi$.}
  \label{fig:ResidualNetwork}
\end{figure}

This issue was studied in \cite{he2016deep}, %
whose authors proposed the residual neural network %
as a solution.  This idea enabled the training of very deep neural
networks, %
making \cite{he2016deep} %
one of the most cited works in %
the field. The idea is to introduce so-called residual blocks, which
complement a regular neural network with an identity.

\begin{definition}\label{def:ResBlock}
  Let $d_0$, $d_1 \in \N$, and $\sigma \colon \R \to \R$. Then a
  \textbf{residual block} is a function of the form
  \begin{align*}
    \Bx\mapsto \Psi_{\rm res}(\Bx; \Phi) \dfn \Bx + \Phi(\Bx), 
  \end{align*}
  where $\Phi$ is a neural network with architecture
  $(\sigma; d_0, d_1, d_0)$.
\end{definition}
A sketch of a residual block is shown in Figure
\ref{fig:ResidualNetwork}. If %
a function $f$ is learned with a residual block, then $\Phi$ needs to
approximate the \emph{residual} $\Bx \mapsto f(\Bx) - \Bx$, i.e., the
additive change that needs to be applied to the input.  A residual
neural network is a composition of multiple residual blocks.
\begin{definition}\label{def:resnn}
  Let $d_0, \dots, d_{L+1} \in \N$ and $\sigma \colon \R \to \R$. A
  \textbf{residual neural network (ResNet)} is a function
  $\Psi:\R^{d_0}\to\R^{d_{L+1}}$ of the form
  \begin{align*}
    \Psi
    =\BW^{(L+1)} \; (\Psi_{\rm res}(\cdot ; \Phi^{(L)})) \circ \dots \circ (\Psi_{\rm res}(\cdot ; \Phi^{(1)})), 
  \end{align*}
  where $\Phi^{(\ell)}:\R^{d_0}\to\R^{d_0}$ is a neural network with
  architecture $(\sigma; d_0, d_\ell, d_0)$ for $\ell = 1, \dots, L$,
  and $\BW^{(L+1)} \in \R^{d_{L+1} \times d_{0}}$. The final layer
  acts as this linear map. We call $L$ the depth of the residual
  neural network.
\end{definition}

Analogous to Definition \ref{def:nn}, we can define the output of each
residual layer by
\begin{subequations}\label{eq:reslayers}
  \begin{align}
    \Bx_{\rm res}^{(0)} &\dfn \Bx\label{eq:reslayers1}\\
    \Bx_{\rm res}^{(\ell)}  &\dfn  \Bx_{\rm res}^{(\ell - 1)} + \Phi^{(\ell)}(\Bx_{\rm res}^{(\ell - 1)})\qquad \text{for } \ell = 1 \dots, L\label{eq:reslayers2}\\
    \Bx_{\rm res}^{(L+1)} &\dfn \BW^{(L+1)}\Bx_{\rm res}^{(L)}\label{eq:reslayers3}
  \end{align}
\end{subequations}
so that $\Psi(\Bx)=\Bx_{\rm res}^{(L+1)}$.

\begin{remark}
  In our definition, each residual block has the same input and output
  dimension $d_0$.  While this assumption is necessary if one uses
  identity skip connections, there are alternative constructions.  In
  \cite{he2016deep}, also skip connections featuring projections or
  embeddings instead of identities are studied; this allows for
  dimensional change.  We will not further discuss these but refer to
  \cite{he2016deep}.
\end{remark}

\subsection{Backpropagation and vanishing gradients}
Let us now %
discuss backpropagation (cf.~Section \ref{sec:backprop}) for residual
neural networks. %
Denote by $\BW^{(\ell,1)}\in\R^{d_\ell\times d_0}$,
$\BW^{(\ell,2)}\in\R^{d_0\times d_\ell}$,
$\Bb^{(\ell,1)}\in\R^{d_\ell}$, $\Bb^{(\ell,2)}\in\R^{d_0}$ the
weights of $\Phi^{(\ell)}$ such that
  $$
  \Phi^{(\ell)}(\Bz) \;=\; \BW^{(\ell,2)}\,\sigma
  \bigl(\BW^{(\ell,1)}\,\Bz \;+\; \Bb^{(\ell,1)}\bigr) \;+\;
  \Bb^{(\ell,2)}\qquad \text{for all } \Bz \in \R^{d_0}.
$$
As earlier, %
we collect all weights of the residual neural network, i.e.
\begin{equation}\label{eq:ResWeights}
  (\BW^{(\ell,1)},\BW^{(\ell,2)},\Bb^{(\ell,1)},\Bb^{(\ell,2)})_{\ell = 1, \dots, L} \text{ and } \BW^{(L+1)},
\end{equation}
into a vector $\Bw$ of suitable size. We then also write
$\Psi(\Bx,\Bw)$ for the residual neural network in Definition
\ref{def:resnn} to emphasize the dependence on the weights in
\eqref{eq:ResWeights}.

Let $\CL:\R^{d_0}\times\R^{d_0}\to\R_+$ be a (differentiable) loss
function and fix $\Bx\in\R^{d_0}$, $\By\in\R^{d_0}$. For brevity we
write
\begin{equation*}
  \CL\dfn \CL(\Psi(\Bx,\Bw),\By)=\CL(\Bx^{(L+1)}_{\rm res},\By).
\end{equation*}
The goal is to compute the derivative of the loss with respect to each
of these parameters, i.e.\ the gradients
\begin{subequations}\label{eq:pLWb-resnet}
  \begin{align}\label{eq:pLWb-resnet1}
    \nabla_{\Bb^{(\ell, 1)}}\CL\in\R^{d_{\ell}},\quad
    \nabla_{\Bb^{(\ell, 2)}}\CL\in\R^{d_{0}},\quad\nabla_{\BW^{(\ell, 1)}}\CL\in\R^{d_{\ell}\times d_0},\quad
    \nabla_{\BW^{(\ell, 2)}}\CL\in\R^{d_{0}\times d_\ell}
  \end{align}
  for $\ell = 1, \dots, L$, and finally
  \begin{align}\label{eq:pLWb-resnet2}
    \nabla_{\BW^{(L+1)}}\CL\in\R^{d_{L+1}\times d_0}.
  \end{align}
\end{subequations}
As in Section \ref{sec:backprop}, this requires an efficient
application of the chain rule. Following the ideas in Section
\ref{sec:bpbasic} and similar to \eqref{eq:Balphaell}, we introduce
the helper quantities
\begin{equation*}
  \Bbeta^{(\ell)}\dfn \nabla_{\Bx_{\rm res}^{(\ell)}}\CL\qquad\text{for all }\ell=1,\dots,L+1.
\end{equation*}
Similar to Lemma \ref{lemma:backprop}, these vectors can be computed
recursively starting from the outer layer.
\begin{lemma}\label{lemma:backpropbeta}
  It holds
  \begin{subequations}\label{eq:Bbetares}
    \begin{align}\label{eq:Bbetares1}
      \Bbeta^{(L+1)} = \nabla_{\Bx_{\rm res}^{(L+1)}}\CL(\Bx_{\rm res}^{(L+1)},\By)\in\R^{d_{L+1}}
    \end{align}
    and
    \begin{align}\label{eq:Bbetares2}
      \Bbeta^{(L)} = (\BW^{(L+1)})^\top \Bbeta^{(L+1)}\in\R^{d_0}
    \end{align}
    and with the identity matrix $\BI_{d_0}\in\R^{d_0\times d_0}$
    \begin{align}\label{eq:Bbetares3}
      \Bbeta^{(\ell)} = \Big(\BI_{d_0}+ \frac{\partial \Phi^{(\ell+1)}(\Bx_{\rm res}^{(\ell)})}{\partial \Bx_{\rm res}^{(\ell)}}
      \Big)^\top\;\Bbeta^{(\ell+1)}\in\R^{d_0}
      \qquad\text{for all }\ell=L-1,\dots,1.
    \end{align}
  \end{subequations}
\end{lemma}
\begin{proof}
  Equation \eqref{eq:Bbetares1} holds by definition. By
  \eqref{eq:reslayers3}
  \begin{equation*}
    \frac{\partial \Bx_{\rm res}^{(L+1)}}{\partial \Bx_{\rm res}^{(L)}}=\BW^{(L+1)}\in\R^{d_{L+1}\times d_0}
  \end{equation*}
  so that by the chain rule
  \begin{equation*}
    \frac{\partial\CL}{\partial\Bx_{\rm res}^{(L)}} =
    \frac{\partial\CL}{\partial\Bx_{\rm res}^{(L+1)}}     \frac{\partial\Bx_{\rm res}^{(L+1)}}{\partial\Bx_{\rm res}^{(L)}}= (\Bbeta^{(L+1)})^\top\BW^{(L+1)},
  \end{equation*}
  which gives \eqref{eq:Bbetares2}.

  For $\ell\in\{1,\dots,L-1\}$, by \eqref{eq:reslayers2} we have
  $\Bx_{\rm res}^{(\ell+1)}=\Bx_{\rm
    res}^{(\ell)}+\Phi^{(\ell+1)}(\Bx_{\rm res}^{(\ell)})$. Thus
  \begin{equation*}
    \frac{\partial \Bx_{\rm res}^{(\ell+1)}}{\partial \Bx_{\rm res}^{(\ell)}}
    =\BI_{d_0}+\Big(\frac{\partial \Phi^{(\ell+1)}(\Bx_{\rm res}^{(\ell)})}{\partial \Bx_{\rm res}^{(\ell)}}\Big)\in\R^{d_0\times d_0}.
  \end{equation*}
  Similar as before, the chain rule then gives \eqref{eq:Bbetares3}.
\end{proof}

Now we can compute the gradients in \eqref{eq:pLWb-resnet} as follows:
First, with $\Bx_{\rm res}^{(L+1)} = \BW^{(L+1)} \Bx_{\rm res}^{(L)}$,
applying the chain rule we get for
$\CL=\CL(\Bx_{\rm res}^{(L+1)},\By)$
\begin{equation*}
  \nabla_{\BW^{(L+1)}}\CL =
  \Bbeta^{(L+1)} (\Bx_{\rm res}^{(L)})^\top \in\R^{d_{L+1}\times d_0}.
\end{equation*}
For the remaining gradients in \eqref{eq:pLWb-resnet1} we only show
the computation $\Bb^{(\ell, 1)}$, as the others terms can be treated
similarly. Using
$\Bx_{\rm res}^{(\ell)}=\Bx^{(\ell-1)}_{\rm
  res}+\Phi^{(\ell)}(\Bx_{\rm res}^{(\ell-1)})$, and the chain rule
for all $\ell=1,\dots,L$
\begin{align*}
  \nabla_{\Bb^{(\ell,1)}}\CL = 
  \Big(\frac{\partial\Bx_{\rm res}^{(\ell)}}{\partial\Bb^{(\ell,1)}}\Big)^\top
  \Big(\frac{\partial\CL}{\partial\Bx_{\rm res}^{(\ell)}}\Big)^\top
  =\frac{\partial\Phi^{(\ell)}(\Bx_{\rm res}^{(\ell)})}{\partial\Bb^{(\ell,1)}} \Bbeta^{(\ell)}.
\end{align*}

Overall, all computations are reduced to computing the
$\Bbeta^{(\ell)}$ with a backward pass as explained in Lemma
\ref{lemma:backpropbeta}, and additionally computing derivatives of
the feedforward neural networks $\Phi^{(\ell)}$; the latter can be
done using regular backpropagation as explained in Section
\ref{sec:backprop}.

The main %
insight is that \eqref{eq:Bbetares3} counteracts the vanishing
gradient problem. At least two points contribute to this:
\begin{itemize}
\item For standard feedforward neural networks (without residual
  blocks), a small activation derivative causes $\Balpha^{(\ell)}$ to
  decrease by that factor relative to $\Balpha^{(\ell+1)}$ (see the
  recursive formula in Lemma \ref{lemma:backprop}). For large depths
  $L$, backpropagating this effect to the first layers can results in
  negligible size of the $\Balpha^{(\ell)}$ for small $\ell$. Since
  these factors scale the gradients with respect to the network
  weights (cf.~Algorithm \ref{alg:backprop}), the vanishing gradient
  phenomenon occurs.

  In residual neural networks, small activation derivatives dampen the
  derivatives $\partial\Phi^{(\ell+1)}/\partial\Bx_{\rm res}^{(\ell)}$
  of the intermediate networks. However, due to the presence of the
  identity matrix in \eqref{eq:Bbetares3}, this does in general not
  cause a decay of $\Bbeta^{(\ell)}$ relative to $\Bbeta^{(\ell+1)}$.

  In the extreme case where the intermediate residual block
  $\Phi^{(\ell)}$ is constant (i.e. has zero derivative), we obtain
  $\Bbeta^{(\ell-1)} = \Bbeta^{(\ell)}$ in \eqref{eq:Bbetares3}. In
  contrast, if the derivative of layer $\ell$ in a standard
  feedforward network is zero (i.e.\ $\sigma'(\bar\Bx^{(\ell)})=\Bnul$
  in \eqref{eq:Balphaellexpl}) then the derivatives with respect to
  all weights in the layers $1, \dots, \ell$ vanish.

\item While \eqref{eq:Bbetares3} can still lead to a reduction of
  $\|\Bbeta^{(\ell-1)}\|$ compared to $\|\Bbeta^{(\ell)}\|$, this is
  less likely than in the standard case. Without the identity, the
  Jacobian $\partial \Phi^{(\ell)}/\partial \Bx_{\rm res}^{(\ell-1)}$
  merely needs to act as a contraction %
  for the gradient to be small.

  In contrast, due to the presence of the identity in
  \eqref{eq:Bbetares3}, the Jacobian
  $\partial \Phi^{(\ell)}/\partial \Bx_{\rm res}^{(\ell-1)}$ needs to
  map $\Bbeta^{(\ell)}$ close to $-\Bbeta^{(\ell)}$ to decrease the
  gradient. The first operation relies solely on reducing the
  amplitude of $\Bbeta^{(\ell)}$, while in the second case, also the
  directions need to align.
\end{itemize}

\subsection{Universality}

As for standard neural networks, we consider the universal
approximation property of residual neural networks.  Recall that by
Theorem \ref{thm:universal}, shallow networks of arbitrary width have
the universal approximation property, as long as the activation
function is not a polynomial. Therefore, a single residual block
$\Phi^{(1)}$ suffices to show universality of ResNets with one
residual block, if we do not limit the width of $\Phi^{(1)}$. The main
role of residual blocks is, however, to enable the training of
\emph{deep} and not necessarily very wide architectures. %
We thus investigate whether universality can be achieved for ResNets
of fixed width but arbitrary depth.

If the activation function is the ReLU, then a form of universality
(in Lebesgue spaces) in fact holds, and it only requires width one for
the $\Phi^{(\ell)}$, i.e.\ $d_1=\dots=d_L=1$ in Definition
\ref{def:resnn}, \cite{lin2018resnet}. We will prove a similar result,
but for simplicity, we allow $d_1=\dots=d_L=2$, and leave the proof of
the stronger result as Exercise \ref{ex:resnet}. We start with the
case of univariate functions.

\begin{lemma}\label{lemma:resnet}
  Let $a_0<a_1<\dots<a_n$ and $b_0,\dots,b_{n-1}\in\R$. Then for any
  $0<\delta<\min_i(a_{i+1}-a_i)$ there exists a ReLU ResNet
  $\Psi:\R\to\R$ with $d_1=\dots=d_L=2$, such that
  \begin{equation}\label{eq:psiclip}
    \Psi(x)\in [\min_i b_i,\max_j b_j]\qquad\text{for all }x\in\R
  \end{equation}
  and for all $i=0,\dots,n-1$
  \begin{equation}\label{eq:psibi}
    \Psi(x)=b_i\qquad\text{if }x\in [a_{i}+\delta,a_{i+1}].
  \end{equation}
\end{lemma}
\begin{proof}
  Let $c\dfn \max_i|a_i|+\max_j|b_j|$.  We show by induction over
  $\ell=1,\dots,n+1$, that there exists a one layer ReLU network
  $\Phi^{(\ell)}$ of width two such that for all
  $x_{\rm res}^{(0)}\in [a_0,a_n]$ and with
  \begin{equation*}
    x_{\rm res}^{(\ell)}\dfn x_{\rm res}^{(\ell-1)}+\Phi^{(\ell)}(x_{\rm res}^{(\ell-1)}),
  \end{equation*}
  it holds
  \begin{enumerate}
  \item\label{item:res1} $x_{\rm res}^{(\ell)}=x_{\rm res}^{(0)}+c$ if
    $x_{\rm res}^{(0)}\le a_{n-\ell+1}$,
  \item\label{item:res2} $x_{\rm res}^{(\ell)}=b_j$ if
    $x_{\rm res}^{(0)}\in [a_j+\delta,a_{j+1}]$ for all
    $j\in\{n-\ell+1,\dots,n-1\}$.
  \end{enumerate}
  For $\ell=n+1$, this gives a ResNet satisfying \eqref{eq:psibi}.
  Finally, we will clip the network to satisfy \eqref{eq:psiclip}.
  
  {\bf Step 1.}  For the base case $\ell=1$ of the induction let
  $\Phi^{(1)}(x)\dfn c$, which can be realized by a one layer ReLU
  network of width two. Then $x_{\rm res}^{(1)}=x_{\rm res}^{(0)}+c$
  satisfies item \ref{item:res1}, and item \ref{item:res2} %
  holds since $\{n-1+1,\dots,n-1\}$ is empty.

  For the induction step, assume \ref{item:res1}-\ref{item:res2} hold
  for some $\ell\le n$. Let $\Phi^{(\ell+1)}$ be a continuous
  piecewise linear function on three pieces such that
  \begin{equation*}
    \Phi^{(\ell+1)}(x)=\begin{cases}
                         0 &\text{if }x\le a_{n-\ell}+c\\
                         b_{n-\ell}-x &\text{if }x\ge a_{n-\ell}+c+\delta,
                       \end{cases}
                     \end{equation*}
                     and uniquely extended on
                     $[a_{n-\ell}+c,a_{n-\ell}+c+\delta]$.  Then
                     $\Phi^{(\ell+1)}$ is the linear combination of
                     two ReLUs, and thus of width two. We verify the
                     two properties:
                     \begin{enumerate}
                     \item Let $x_{\rm res}^{(0)}\le a_{n-\ell}$. By
                       induction assumption \ref{item:res1},
                       $x_{\rm res}^{(\ell)}=x_{\rm res}^{(0)}+c\le
                       a_{n-\ell}+c$, and
                       \begin{equation*}
                         x_{\rm res}^{(\ell+1)}=x_{\rm res}^{(\ell)}+\Phi^{(\ell+1)}(x_{\rm res}^{(\ell)}) = x_{\rm res}^{(\ell)}=x_{\rm res}^{(0)}+c.
                       \end{equation*}
                     \item Let $j\in\{n-\ell+1,\dots,n-1\}$ and
                       $x_{\rm res}^{(0)}\in
                       [a_j+\delta,a_{j+1}]$. Then by induction
                       assumption \ref{item:res2},
                       $x_{\rm res}^{(\ell)}=b_j\le a_{n-\ell}+c$ so
                       that $\Phi^{(\ell+1)}(x_{\rm
                         res}^{(\ell)})=0$. Thus
                       $x_{\rm res}^{(\ell+1)}=b_j$.

                       Next let
                       $x_{\rm res}^{(0)}\in
                       [a_{n-\ell}+\delta,a_{n-\ell+1}]$. Then
                       $x_{\rm res}^{(\ell)}=x_{\rm res}^{(0)}+c$ by
                       induction assumption \ref{item:res1}. Thus
                       $x_{\rm res}^{(\ell)}\ge a_{n-\ell}+c+\delta$
                       and
                       \begin{equation*}
                         x_{\rm res}^{(\ell+1)}=x_{\rm res}^{(\ell)}+\Phi^{(\ell+1)}(x_{\rm res}^{(\ell)})=x_{\rm res}^{(\ell)}+b_{n-\ell}-x_{\rm res}^{(\ell)}=b_{n-\ell}.
                       \end{equation*}
                     \end{enumerate}
                     The construction is visualized in Figure
                     \ref{fig:resnet}.

                     {\bf Step 2.} To satisfy the bounds
                     \eqref{eq:psiclip} let
                     $b_{\rm min}\dfn \min_ib_i$,
                     $b_{\rm max}\dfn \max_j b_j$ and
                     \begin{equation*}
                       \Phi^{(n+2)}(x)\dfn \begin{cases}
                                             b_{\rm min}-x &\text{if }x\le b_{\rm min}\\
                                             b_{\rm max}-x &\text{if }x\ge b_{\rm max}\\
                                             0 &\text{otherwise},
                                           \end{cases}
                                         \end{equation*}
                                         which is again a linear
                                         combination of two ReLUs.
                                         Clearly
                                         $x_{\rm res}^{(n+2)}=x_{\rm
                                           res}^{(n+1)}+\Phi^{(n+2)}(x_{\rm
                                           res}^{(n+1)})$ satisfies
                                         \eqref{eq:psiclip} regardless
                                         of the value of
                                         $x_{\rm
                                           res}^{(n+1)}$. Moreover,
                                         $x_{\rm res}^{(n+1)}=x_{\rm
                                           res}^{(n+2)}$ in case
                                         $x_{\rm res}^{(n+1)}\in
                                         [b_{\rm min},b_{\rm max}]$.
                                       \end{proof}

                                       \begin{figure}
                                         \begin{center}
                                           \begin{tikzpicture}
                                             \fill [blue!20]
                                             (-0.13,2.03) rectangle
                                             (3.2,3.1);
      
                                             \draw [->,thick] (-0.2,0)
                                             -- (3.1,0); \draw
                                             [->,thick] (0,-0.1) --
                                             (0,2.8);

                                             \draw [] (0.05,0.3) --
                                             (-0.05,0.3) node [left]
                                             {\footnotesize $b_{0}$};
                                             \draw [] (0.05,0.7) --
                                             (-0.05,0.7) node [left]
                                             {\footnotesize $b_{1}$};
                                             \draw [] (0.05,2.03) --
                                             (-0.05,2.03) node [left]
                                             {\footnotesize
                                               $a_{1}+c$};

                                             \draw [] (2.8,0.05) --
                                             (2.8,-0.05) node [below]
                                             {\footnotesize $a_{2}$};
                                             \draw [] (1.5,0.05) --
                                             (1.5,-0.05) node [below]
                                             {\footnotesize $a_{1}$};
                                             \draw [] (0.2,0.05) --
                                             (0.2,-0.05) node [below]
                                             {\footnotesize $a_{0}$};
    
                                             \draw [thick] (-0.2,1) --
                                             (3.1,3); \node at
                                             (1.5,-1)
                                             {$x_{\rm
                                                 res}^{(1)}=x+c$};

                                             \fill [blue!20]
                                             (3.87,1.24) rectangle
                                             (7.2,3.1); \draw
                                             [->,thick] (3.8,0) --
                                             (7.1,0); \draw [->,thick]
                                             (4,-0.1) -- (4,2.8);

                                             \draw [] (4.05,0.3) --
                                             (3.95,0.3) node [left]
                                             {\footnotesize $b_{0}$};
                                             \draw [] (4.05,0.7) --
                                             (3.95,0.7) node [left]
                                             {\footnotesize $b_{1}$};
                                             \draw [] (4.05,1.24) --
                                             (3.95,1.24) node [left]
                                             {\footnotesize
                                               $a_{0}+c$};

                                             \draw [] (6.8,0.05) --
                                             (6.8,-0.05) node [below]
                                             {\footnotesize $a_{2}$};
                                             \draw [] (5.5,0.05) --
                                             (5.5,-0.05) node [below]
                                             {\footnotesize $a_{1}$};
                                             \draw [] (4.2,0.05) --
                                             (4.2,-0.05) node [below]
                                             {\footnotesize $a_{0}$};
                                             \draw [] (5.7,0.05) --
                                             (5.7,-0.05);

                                             \draw [thick] (3.8,1) --
                                             (5.5,2.03) -- (5.7,0.7)
                                             -- (7.1,0.7); \node at
                                             (5.5,-1)
                                             {$x_{\rm res}^{(2)}$};

                                             \draw [->,thick] (7.8,0)
                                             -- (11.1,0); \draw
                                             [->,thick] (8,-0.1) --
                                             (8,2.8);

                                             \draw [] (8.05,0.3) --
                                             (7.95,0.3) node [left]
                                             {\footnotesize $b_{0}$};
                                             \draw [] (8.05,0.7) --
                                             (7.95,0.7) node [left]
                                             {\footnotesize $b_{1}$};
    
                                             \draw [] (10.8,0.05) --
                                             (10.8,-0.05) node [below]
                                             {\footnotesize $a_{2}$};
                                             \draw [] (9.5,0.05) --
                                             (9.5,-0.05) node [below]
                                             {\footnotesize $a_{1}$};
                                             \draw [] (8.2,0.05) --
                                             (8.2,-0.05) node [below]
                                             {\footnotesize $a_{0}$};

                                             \draw [thick] (7.8,1) --
                                             (8.2,1.24) -- (8.4,0.3)
                                             -- (9.5,0.3); \draw
                                             [thick, dashed] (9.5,0.3)
                                             -- (9.7,0.7); \draw
                                             [thick] (9.7,0.7) --
                                             (11.1,0.7);

                                             \draw [] (9.7,0.05) --
                                             (9.7,-0.05); \draw []
                                             (8.4,0.05) --
                                             (8.4,-0.05);

                                             \node at (9.5,-1)
                                             {$x_{\rm res}^{(3)}$};
    
                                           \end{tikzpicture}
                                         \end{center}
                                         \caption{The output of the
                                           ResNet constructed in Step
                                           1 of Lemma
                                           \ref{lemma:resnet} in
                                           different layers. The
                                           dashed part corresponds to
                                           a continuous piecewise
                                           linear (but not necessarily
                                           linear) connection. The
                                           subsequent ResNet block
                                           only affects values in the
                                           blue shaded
                                           region.}\label{fig:resnet}
                                       \end{figure}

                                       Recall that the step functions
                                       are dense in $L^p(K)$ for any
                                       compact set $K\subseteq\R$ and
                                       any $p\in [1,\infty)$.  Since
                                       we can approximate step
                                       functions arbitrarily well with
                                       ResNets, it is clear that
                                       residual neural networks are
                                       dense in $L^p([a_0,a_n])$ on
                                       any compact interval. We now
                                       come to the multivariate
                                       case. The proof follows the
                                       ideas in \cite[Theorem
                                       3.1]{lin2018resnet}.

\begin{theorem}\label{thm:resnet}
  Let $p\in [1,\infty)$, $d \in \N$ and let $f \colon [0,1]^d \to \R$
  be continuous. Then, for every $\eps >0$, there exist $L \in \N$ and
  a ReLU residual neural network $\Psi$ with $d_0 = d$,
  $d_1=\dots=d_L= 2$, and $d_{L+1} = 1$, such that
  \begin{align*}
    \int_{[0,1]^d} |\Psi(\Bx) - f(\Bx)|^p \dd\Bx < \eps.
  \end{align*}
\end{theorem}

\begin{proof}
  The idea of the proof is to use a quantization technique as in
  Proposition \ref{prop:kolmogorov}. For convenience of the reader we
  repeat the construction (with slight adjustments) in Step 1. Step 2
  then reapproximates this construction with a ResNet.

  {\bf Step 1.}  Let $n\in\N$ and $a_i=i/n$ for $i=0,\dots,n$. For
  $\Bnu\in\Lambda_n\dfn \{0,\dots,n-1\}^d$ set
  $Q_\Bnu\dfn \times_{j=1}^d [a_{\nu_j},a_{\nu_j+1})$ and denote by
  $f_\Bnu$ the value of $f$ at the midpoint of the cube $Q_\Bnu$. Then
  with
  \begin{equation*}
    F_n(\Bx)\dfn \sum_{\Bnu\in\Lambda_n}f_\Bnu \cdot \ind_{Q_\Bnu}(\Bx),
  \end{equation*}
  we have
  \begin{equation}\label{eq:limitfnresnet}
    \lim_{n\to\infty}\int_{[0,1]^d}|f(\Bx)-F_n(\Bx)|^p\dd\Bx =0 .
  \end{equation}

  We now rewrite $F_n$ as a composition of functions.  To each
  $\Bnu\in\{0,\dots,n-1\}^d$ we assign the unique number
  \begin{equation*}
    b_\Bnu \dfn \sum_{j=1}^d \nu_j\cdot n^{j-1}.
  \end{equation*}
  We remark that
  $\set{b_\Bnu}{\Bnu\in\Lambda_n}=\{0,\dots,n^{d}-1\}$. Next, define
  \begin{equation*}
    g(x)\dfn \sum_{\Bnu\in\Lambda_n} f_\Bnu\cdot \ind_{[b_\Bnu-1/2,b_\Bnu+1/2]}(x)\qquad\text{for all }x\in\R,
  \end{equation*}
  and for $j=1,\dots,d$
  \begin{equation*}
    h_j(x_j)\dfn n^{j-1}\cdot \sum_{i=0}^{n-1} i\cdot\ind_{[a_i,a_{i+1})}(x_j)\qquad\text{for all }x_j\in\R.
  \end{equation*}
  Then for any $\Bx\in Q_\Bmu=\times_{j=1}^d [a_{\mu_j},a_{\mu_j+1})$
  \begin{equation*}
    h(\Bx)\dfn \sum_{j=1}^d h_j(x_j) = \sum_{j=1}^d \mu_j\cdot n^{j-1}=b_\Bmu,
  \end{equation*}
  so that
  \begin{equation*}
    g\circ h(\Bx) = \sum_{\Bnu\in\Lambda_n}f_\Bnu\cdot \ind_{[b_\Bnu-1/2,b_\Bnu+1/2]}(b_\Bmu) = f_\Bmu = F_n(\Bx).
  \end{equation*}

  {\bf Step 2.}  Due to \eqref{eq:limitfnresnet}, it suffices to show
  that $F_n$ can be approximated arbitrarily well in $L^p([0,1]^d)$ by
  a residual neural network with the properties stated in the theorem.
  We now construct a ResNet mimicking the function $g\circ h$.

  First, applying Lemma \ref{lemma:resnet} to the first coordinate,
  and leaving the other coordinates unchanged, we can find a ResNet
  $\Psi^{(1)}:\R^d\to\R^d$, whose residual blocks have width two, such
  that
  \begin{equation*}
    \Psi^{(1)}(\Bx) = \begin{pmatrix}
                        \tilde h_1(x_1)\\
                        x_2\\
                        \vdots\\
                        x_d
                      \end{pmatrix}
                    \end{equation*}
                    where
                    \begin{equation*}
                      \tilde h_1(x_1) = h_1(x_1)\qquad \text{for all $x_1\in [0,1]$ s.t.\ $|x_1-a_i|>\delta$ for all $i$}.
                    \end{equation*}
                    Analogously we construct $\Psi^{(j)}$ for
                    $j=2,\dots,d$. Moreover, using
                    \begin{equation*}
                      x_1+\sigma_{\rm ReLU}\Big(\sum_{j=2}^dx_j\Big)-\sigma_{\rm ReLU}\Big(-\sum_{j=2}^dx_j\Big)=\sum_{j=1}^dx_j
                    \end{equation*}
                    and Lemma \ref{lemma:resnet}, we find that there
                    exists a ResNet $\Psi^{(d+1)}:\R^d\to\R$, whose
                    residual blocks have width two, such that
                    \begin{equation*}
                      \Psi^{(d+1)}(\Bx) = \begin{pmatrix}
                                            \tilde g(\sum_{j=1}^d x_j)\\
                                            x_2\\
                                            \vdots\\
                                            x_d
                                          \end{pmatrix}
                                        \end{equation*}
                                        where
                                        \begin{equation*}
                                          \tilde g(x) = g(x)\qquad \text{for all }x\in\set{x\in [0,n^{d}-1]}{|x-i|<1/4~\text{for some }i\in\N_0},
                                        \end{equation*}
                                        and additionally
                                        $\min_\Bnu f_\Bnu\le \tilde
                                        g(x)\le\max_\Bnu f_\Bnu$ for
                                        all $\Bx\in [0,1]^d$.

                                        Then
                                        \begin{equation*}
                                          \Psi^{(d)}\circ\dots\circ\Psi^{(1)}(\Bx) = \begin{pmatrix}
                                                                                       \tilde h_1(x_1)\\
                                                                                       \vdots\\
                                                                                       \tilde h_d(x_d)
                                                                                     \end{pmatrix}
                                                                                   \end{equation*}
                                                                                   and
                                                                                   \begin{equation*}
                                                                                     \Psi(\Bx)\dfn (1,0,\dots,0)~\Psi^{(d+1)}\circ\dots\circ\Psi^{(1)}(\Bx) = \tilde g\Big(\sum_{j=1}^d\tilde h_j(x_j)\Big).
                                                                                   \end{equation*}
                                                                                   By
                                                                                   construction
                                                                                   this
                                                                                   function
                                                                                   coincides
                                                                                   with
                                                                                   $F_n$
                                                                                   for
                                                                                   all
                                                                                   $\Bx$
                                                                                   in
                                                                                   \begin{equation*}
                                                                                     Z \dfn \set{\Bx\in [0,1]^d}{|x_j-a_i|>\delta~ \text{ for all } i,j}.
                                                                                   \end{equation*}
                                                                                   The
                                                                                   volume
                                                                                   of
                                                                                   $Z$
                                                                                   tends
                                                                                   to
                                                                                   zero
                                                                                   as
                                                                                   $\delta\to
                                                                                   0$
                                                                                   and
                                                                                   $|\Psi|$
                                                                                   is
                                                                                   uniformly
                                                                                   upper
                                                                                   bounded
                                                                                   by
                                                                                   $\sup_{\Bx\in
                                                                                     [0,1]^d}|f(\Bx)|$
                                                                                   (independent
                                                                                   of
                                                                                   $\delta$).
                                                                                   Thus
                                                                                   $\int_{[0,1]^d}|F_n(\Bx)-\Psi(\Bx)|^p\dd\Bx\to
                                                                                   0$
                                                                                   as
                                                                                   $\delta\to
                                                                                   0$.
                                                                                   Since
                                                                                   $\Psi$
                                                                                   is
                                                                                   a
                                                                                   residual
                                                                                   neural
                                                                                   network,
                                                                                   whose
                                                                                   residual
                                                                                   blocks
                                                                                   all
                                                                                   have
                                                                                   width
                                                                                   two
                                                                                   in
                                                                                   the
                                                                                   hidden
                                                                                   layer,
                                                                                   this
                                                                                   concludes
                                                                                   the
                                                                                   proof.
                                                                                 \end{proof}

\begin{remark}
  The residual neural network $\Psi$ constructed in the proof of
  Theorem \ref{thm:resnet} has $N=O(n^d)$ parameters. For
  $f\in C^{0,s}([0,1]^d)$ and $p\in [1,\infty)$, it is easy to see
  that the error scales like
  $\norm[{L^p([0,1]^d)}]{f-\Psi}=O(n^{-s})$. Thus the proof does not
  just give universality, but for $C^{0,s}$ functions, similar as in
  Section \ref{sec:HoelderRates}, we obtain the convergence rate
  $N^{-s/d}$ in terms of the network size $N$.

  Since every ReLU ResNet corresponds to a standard feedforward ReLU
  neural network (by using that
  $x=\sigma_{\rm ReLU}(x)-\sigma_{\rm ReLU}(-x))$, we find that in
  $L^p$, $p\in [1,\infty)$, the convergence rate in Theorem
  \ref{thm:hoelder} can also be achieved by fixing a large enough
  width, and increasing the depth.
\end{remark}

\subsection{When should a residual network be used?}
As we have described in this section, residual neural networks enable
very deep architectures.  Training these can be extremely challenging
without the residual structure.  Therefore, in an application where
very complex dependencies need to be resolved and a very powerful deep
architecture is used, it can be very beneficial to add residual
blocks.  On the other hand, if a problem is simple and can be solved
with a shallow neural network, then residual connections are often not
necessary.

\section{Convolutional neural networks}

In this section, we will review one of the most common architectures
in deep learning, the convolutional neural network, introduced in
\cite{lecun2002gradient}, \cite{lecun1989backpropagation}.

To motivate this architecture, %
let us start by discussing the concept of \emph{features}.  With
$\Bx$, $\Bw\in\R^d$, consider a single neuron
$\Bx \mapsto \nu(\Bx) = \sigma(\langle \Bw, \Bx \rangle)$, and assume
$\sigma:\R\to\R$ is monotonically increasing; here and in the
following $\dup{\cdot}{\cdot}$ is again the Euclidean inner product.
For an input $\Bx$ with $\|\Bx\| = 1$, the response $\nu$ is %
larger if $\Bx$ is aligned with $\Bw$, and smaller otherwise.  Thus
$\nu$ measures how much of $\Bw$ is in $\Bx$.  For example, let
$\Bw=(w_i)_{i=1}^d$ with $w_i = \sin(2 \pi k i/d )$, and let
$\Bx\in\R^d$ be an acoustic signal.  Then $\nu$ %
quantifies the contribution of frequency $k$ to the signal. %
In this sense, each neuron %
can be interpreted as checking for a specific \emph{feature} of the
data.

\begin{figure}[htb]
  \centering
  \includegraphics[width=0.9\linewidth]{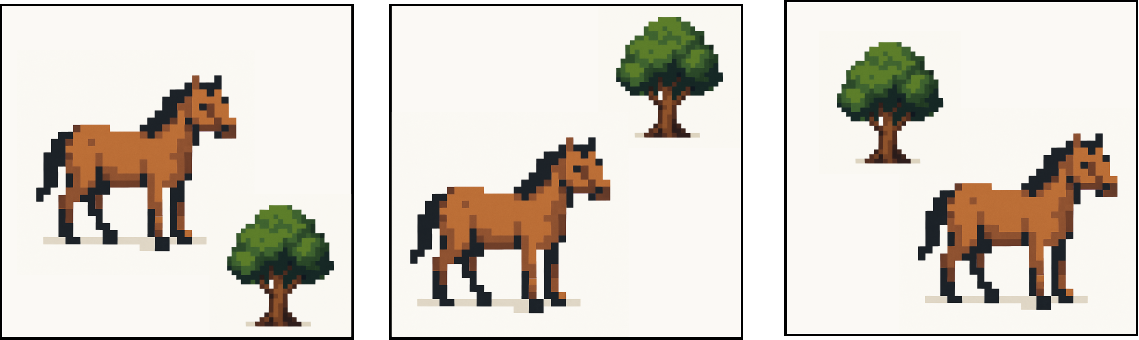}
  \caption{Three images of a horse standing next to a tree. This
    picture serves as a motivation for translation-agnostic and local
    features in this section.}
  \label{fig:translationForCNNs}
\end{figure}

Consider now %
a special type of signal, namely a picture
$\BX=(X_{i,j})_{i,j=1}^d\in\R^{d\times d}$ (a $d\times d$ pixel
matrix).  Figure \ref{fig:translationForCNNs} shows an example of
three such images.  All of %
them are different, but they %
share the exact same content---a horse next to a tree.  With the goal
of classifying the objects in the picture, a useful feature should
possess the following properties:
\begin{itemize}
\item \emph{Translation-agnostic features:} For most of the content of
  a picture, it does not matter where it appears.  For example, in
  Figure \ref{fig:translationForCNNs}, each image contains two
  objects---a horse and a tree.  For the classification, it is not
  relevant where either of these two objects is.  To formalize this,
  we define a (periodic) translation operator on $(d \times d)$-pixel
  images: First, letting $\bar i\dfn ((i-1)\;{\rm mod}\;d)+1$ for
  $i\in\Z$, we periodically extend the pixel matrix via
  \begin{equation}\label{eq:periodicextensionx}
    X_{i,j}\dfn X_{\bar i,\bar j}\quad
    \text{for all }i,j\in\Z.
  \end{equation}
  Then set for %
  $k_1$, $k_2\in\Z$
  \begin{align} \label{eq:translationOp} (T_{k_1,k_2}(\BX))_{i,j} =
    X_{i-k_1, j-k_2}\qquad\text{for all }i,j=1,\dots,d.
  \end{align}

  This defines a translation operator
  $T_{k_1,k_2}:\R^{d\times d}\to\R^{d\times d}$. We can now
  reformulate our observation on the translation property of features:
  If $\BW\in\R^{d\times d}$ is a relevant feature, we expect for any
  $k_1$, $k_2\in\Z$ that $T_{k_1,k_2}(\BW)$ is a relevant feature too,
  and vice versa.
	
\item \emph{Locality:} Consider again Figure
  \ref{fig:translationForCNNs}. %
  Comparing the pictures, only small parts are shifted, but not the
  whole images.  Indeed, %
  meaningful features often correspond to local patches.  In other
  words, a meaningful feature $\BW$ should satisfy that $\supp(\BW)$
  is contained in a small $p \times p$ patch with $p \ll d$.  It is an
  empirical fact that in images, nearby pixels are most correlated and
  the correlation decreases the further away the pixels are
  \cite{lecun2002gradient}.  Moreover, from a practical viewpoint,
  having features of small size drastically reduces the number of
  parameters of the underlying model, making it faster to compute and
  easier to store.
\end{itemize}

To comply with the translation-agnostic features principle, %
given $\BW \in \R^{d \times d}$, we %
should compute
\begin{align*}
  \BX \mapsto \sigma(\langle T_{k_1,k_2}\BW, \BX\rangle)\qquad\text{for all }k_1, k_2 = 0,\dots,d-1,
\end{align*}
or equivalently for $k_1$, $k_2\in\{1,\dots,d\}$.  Denote by $*$ the
discrete two-dimensional convolution operator, i.e., for $\BX$,
$\BW\in\R^{d\times d}$
\begin{equation*}
  \BW*\BX=\Big(\sum_{i,j=1}^d W_{k_1-i,k_2-j}X_{i,j}\Big)_{k_1,k_2=1}^d.
\end{equation*}
With the symmetric extension of matrices \eqref{eq:periodicextensionx}
it then holds for $k_1$, $k_2\in\Z$
\begin{align}\label{eq:howtousecheck}
  \langle T_{k_1,k_2}\BW, \BX\rangle &= \sum_{i,j= 1}^d (T_{k_1,k_2}\BW)_{i,j} X_{i,j}\nonumber \\
  =& \sum_{i,j= 1}^d W_{i-k_1,j-k_2} X_{i,j}\nonumber\\
  =& \sum_{i,j= 1}^d \widecheck{W}_{k_1-i,k_2-j} X_{i,j}
     = (\widecheck{\BW}*\BX)_{k_1,k_2},
\end{align}
where we use the reflection notation
(cf.~\eqref{eq:periodicextensionx})
\begin{equation}\label{eq:checknotation}
  \widecheck{W}_{i,j} \dfn {W}_{-i,-j}\qquad\text{for all }i,j=1,\dots,d.
\end{equation}

Therefore,
$(\sigma(\langle T_{k_1,k_2}\BW, \BX\rangle))_{k_1,k_2 =1}^d$ can be
computed by a %
convolution followed by a coordinate-wise application of $\sigma$.
Note that,
$(\sigma(\langle T_{k_1,k_2}\BW, \cdot\rangle))_{k_1,k_2 =1}^d$ is a
neural network with $d^2$ output neurons and $d^2$ input neurons. Yet
it has only as many free parameters as we allow $\BW$ to have (at most
$d^2$, but not $O(d^4)$, which is what a fully connected feedforward
neural network would give).  This concept is referred to as
\emph{parameter sharing} since the parameters used to compute the
outputs of each of the neurons are the same.  These observations
motivate the following definition of a convolutional neural network.

\begin{definition}%
  \label{def:convblock}
  Let $d$, $C_{\rm in}$, $C_{\rm out} \in \N$, let
  $\BW \in \R^{d \times d \times C_{\rm in} \times C_{\rm out}}$, let
  $\Bb \in \R^{C_{\rm out}}$, and finally, let
  $\sigma \colon \R \to \R$.
	
  The function
  $\mathcal{C} \colon \R^{d \times d \times C_{\rm in}} \to \R^{d
    \times d \times C_{\rm out}}$ defined as
  \begin{align}\label{eq:convblockdef}
    \BX \mapsto \left(\sigma\left(\sum_{k=1}^{C_{\rm in}} \BW_{\cdot, \cdot, k, m} * \BX_{\cdot, \cdot, k}  + b_m \right)\right)_{m = 1}^{C_{\rm out}},
  \end{align}
  where $\sigma$ and the addition of $b_m\in\R$ are applied
  coordinate-wise, is called a \textbf{convolutional block} with
  \textbf{input channel size $C_{\rm in}\in\N$}, \textbf{output
    channel size $C_{\rm out}\in\N$}, and activation function
  $\sigma$. We define the \textbf{size} of a convolutional block as
  $\size(\mathcal{C}) = \|\Bb\|_0 + \sum_{k = 1}^{C_{\rm in}} \sum_{m
    = 1}^{C_{\rm out}}\|\BW_{\cdot ,\cdot,k,m}\|_0$.
\end{definition}
In the definition of a convolutional block (and also for convolutional
neural networks as defined below) we restrict ourselves to
two-dimensional inputs. Similar constructions can be made for
one-dimensional inputs, high-dimensional inputs, or even more general
structures. %

Typically, a convolutional block is combined with further blocks of
either pooling or flattening layers.
\begin{itemize}
\item \textbf{Pooling:} %
  A pooling layer reduces the size of the input
  $\BX \in \R^{d\times d \times C_{\rm in}}$ in the first two
  coordinates. For example, for a divisor $s \in \N$%
  of $d$, %
  the $d \times d$ components can be split into $(d/s)^2$ patches of
  size $s \times s$.  On each patch, a procedure is carried out to
  reduce the patch to a single number; typical examples include %
  the mean, the maximum, or some other aggregate. %
  These aggregates then form a vector of shape $(d/s) \times (d/s)$.
	
\item \textbf{Flattening:} It is often useful to combine a
  convolutional neural network with a standard feedforward neural
  network. To achieve this, %
  a flattening layer is applied, which takes an input
  $\BX \in \R^{d\times d \times C_{\rm in}}$ and returns a vector in
  $\R^{d^2 C_{\rm in}}$ with the same entries, but arranged
  appropriately.
\end{itemize}

\begin{figure}[htb]
  \includegraphics[width=0.95\textwidth]{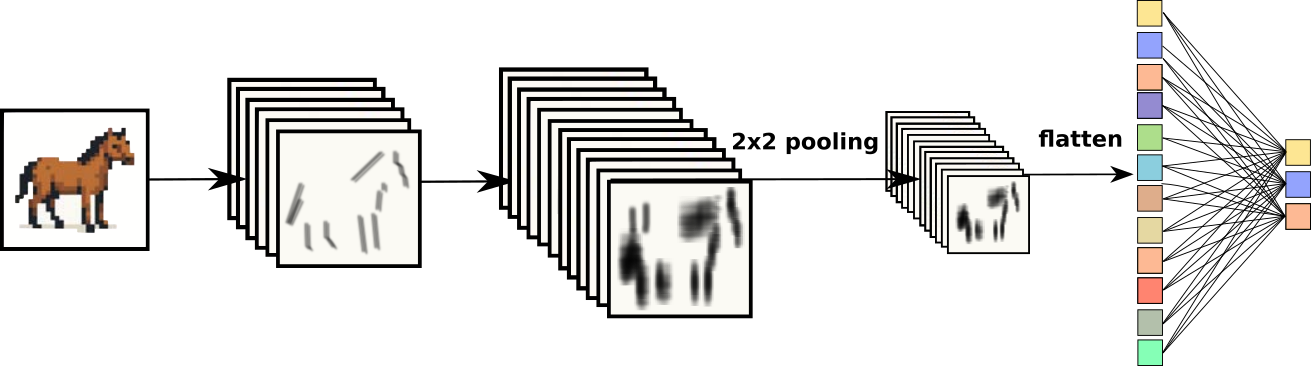}
  \put(-0.83\textwidth, 0.15\textwidth){$\mathcal{C}_1$}
  \put(-0.64\textwidth, 0.15\textwidth){$\mathcal{C}_2$}
  \caption{Sketch of a convolutional neural network with two
    convolutional blocks $\mathcal{C}_1, \mathcal{C}_2$. These are
    followed by a 2x2 pooling block, which reduces the spatial size in
    each coordinate by a factor of 2. Then, a flattening layer is
    applied, and a one-layer neural network follows.}
  \label{fig:CNN}
\end{figure}

A convolutional neural network is a composition of convolutional
blocks, often with different activation functions (e.g., the identity
and the ReLU), pooling layers, and a flattening layer followed by a
regular neural network. A sketch is shown in Figure \ref{fig:CNN}.
This flexibility in terms of the usable components makes it hard to
give precise statements that are specific to the convolutional part of
a convolutional neural network.  Therefore, we study in the sequel a
more pure form of convolutional neural networks, without pooling or
flattening layers.

\begin{definition}%
  \label{def:convnet}
  Let $L$, $d \in \N$, $C_0, \dots, C_{L} \in \N$. For each
  $\ell=1,\dots,L$ let $\sigma_\ell \colon \R \to \R$ and let
  \begin{equation*}
    \mathcal{C}^{(\ell)}:\R^{d\times d\times C_{\ell-1}}\to\R^{d\times d\times C_\ell}
  \end{equation*}
  be a convolutional block %
  depending on weights
  $\BW^{(\ell-1)}\in\R^{d\times d\times C_{\ell-1}\times C_\ell}$ and
  biases $\Bb^{(\ell-1)}\in\R^{C_\ell}$ and using activation function
  $\sigma_{\ell-1}$ (cf.~Definition \ref{def:convblock}).

  A {\bf convolutional neural network} is a function
  $\Psi:\R^{d\times d\times C_0}\to\R^{d\times d\times C_{L}}$ of the
  form
  \begin{align*}
    \Psi = \CC^{(L)}\circ\dots\circ \CC^{(1)}(\BX).
  \end{align*}
  We define the \textbf{size} of $\Psi$ as
  $\size(\Psi) = \sum_{\ell = 1}^{L} \size(\CC^{(\ell)})$.
\end{definition}

A natural question to ask is how convolutional neural networks relate
to regular neural networks. %
The latter are typically %
take a vector as input, which is why we first need to introduce an
appropriate isomorphism between $\R^{d \times d \times q}$ and
$\R^{d^2 q}$, which we call $\mathcal{I}_{\rm flat}$. To keep the
notation simple we omit the parameters $d$ and $q$ since they will
always be clear from the context.  To compare these two types of
architectures, we %
next investigate whether for every neural network $\Phi$, there exists
a convolutional neural network $\Psi$ of comparable size, such that
\begin{align}\label{eq:transferencePrinciple}
  \mathcal{I}_{\rm flat} \circ\Psi \circ\mathcal{I}_{\rm flat}^{-1} = \Phi.
\end{align}

\begin{definition}\label{def:treq}
  Let $d$, $q_1$, $q_2 \in \N$. A function
  $f \colon \R^{d \times d \times q_1} \to \R^{d \times d \times q_2}$
  is called {\bf translation equivariant} in the spatial coordinates,
  if for all $\BX \in \R^{d \times d \times q_1}$ and all $k_1$,
  $k_2 = 0, \dots, d-1$
	$$
	f \left( (T_{k_1,k_2} \BX_{\cdot, \cdot,
            i})_{i=1}^{q_1}\right) = \left(T_{k_1,k_2} (f(\BX)_{\cdot,
            \cdot,j})\right)_{j=1}^{q_2}.
	$$
      \end{definition}

      If $f$, $g$ are translation equivariant in the spatial
      coordinates and $f \circ g$ is well-defined, then $f \circ g$ is
      translation equivariant in the spatial coordinates.  The
      convolution operator is translation equivariant (see Exercise
      \ref{ex:convolution}), adding a bias is constant in the spatial
      coordinates, and the activation function is applied
      coordinate-wise. All of these operations are translation
      equivariant, and therefore, %
      every convolutional block is translation equivariant in the
      spatial coordinates. Thus the same holds for every convolutional
      neural network.

\begin{proposition}\label{prop:CNNsAreTranslationEqui}
  Every convolutional neural network is translation equivariant in the
  spatial coordinates.
\end{proposition}

This observation allows us to answer part of the above posed question,
as the next example shows: there exist $\Phi$ for which
\eqref{eq:transferencePrinciple} cannot even hold approximately for
any convolutional neural network $\Psi$.

\begin{example}
  Consider $q_1=q_2=1$ in Definition \ref{def:treq}, and let
  \begin{align*}
    f:&\R^{2\times 2}\to\R^{2\times 2}\\
      &\begin{pmatrix}
         X_{11}&X_{12}\\
         X_{21}&X_{22}
       \end{pmatrix}
        \mapsto
        \begin{pmatrix}
          X_{11}&X_{11}\\
          X_{11}&X_{11}
        \end{pmatrix}.        
  \end{align*}
  Evidently, this function is not translation equivariant. Let
  \begin{equation*}
    \BX\dfn \begin{pmatrix}
              1 &0\\
              0 &0
            \end{pmatrix}\qquad\text{and}\qquad
            T_{1,1}\BX =
            \begin{pmatrix}
              0 &0\\
              0 &1
            \end{pmatrix}.
          \end{equation*}
          For any convolutional neural network
          $\Psi:\R^{2\times 2}\to\R^{2\times 2}$, due to the
          translation equivariance, it must hold
          \begin{equation*}
            T_{1,1}\Psi(\BX)=\Psi(T_{1,1}\BX).
          \end{equation*}
          Due to $T_{1,1}\circ T_{1,1}$ being the identity on
          $\R^{2\times 2}$ we thus have
          \begin{equation*}
            \Psi(\BX)=T_{1,1}\Psi(T_{1,1}\BX).
          \end{equation*}
          Using that %
          \begin{equation*}
            f(\BX)=\Bone\in\R^{2\times 2}\qquad\text{and}\qquad
            f(T_{1,1}\BX)=\Bnul\in\R^{2\times 2}
          \end{equation*}
          are the constant $1$ and $0$ matrix, it must hold that
          either
          \begin{equation*}
            \norm{\Psi(\BX)-f(\BX)}=\norm{\Psi(\BX)-\Bone}\ge \frac{1}{2}%
          \end{equation*}
          or
          \begin{align*}
            \norm{\Psi(T_{1,1}\BX)-f(T_{1,1}\BX)} &=
                                                    \norm{T_{1,1}\Psi(T_{1,1}\BX)-T_{1,1}f(T_{1,1}\BX)}\\
                                                  &= \norm{\Psi(\BX)-\Bnul}\ge \frac{1}{2}.
          \end{align*}
          In either case, no convolutional neural network can
          approximate $f$ on $[0,1]^{2\times 2}$ to accuracy less than
          $1/2$.  For feedforward neural networks on the other hand,
          universal approximation ensures under mild conditions on the
          activation function that there exists a neural network
          approximating $f$ to arbitrary accuracy, see Chapter
          \ref{chap:UA}.
        \end{example}

        Next, we observe that every spatially translation equivariant
        function is completely determined by its values in one spatial
        output variable. Indeed
$$
(f(\BX))_{i,j,q} = ( T_{1-i, 1-j} f(\BX))_{1,1,q} = (f( T_{-i -1, -j
  -1} \BX))_{1,1,q}.
$$
Therefore, we can %
identify every function that is translation equivariant in the spatial
coordinates as a function with just one spatial output dimension; that
is, every such
$f:\R^{d\times d\times q_1}\to \R^{d\times d\times q_2}$ can be
identified with some $g:\R^{d\times d \times q_1}\to\R^{q_2}$.

Motivated by this, we now study the following question: For a neural
network $\Phi$ with one-dimensional output, under which condition does
there exist a convolutional neural network $\Psi$ such that
\begin{align}\label{eq:transferencePrincipleB}
  (\mathcal{I}_{\rm flat}\circ \Psi\circ \mathcal{I}_{\rm flat}^{-1})_{1,1,1} = \Phi
\end{align}
holds? Conversely, given a convolutional neural network $\Psi$, under
which condition does $\Phi$ as in \eqref{eq:transferencePrincipleB}
exist?  These questions were answered under more general conditions in
\cite{petersen2020equivalence}, and we give some simplified statements
below.

\begin{theorem}\label{thm:EveryNNCanBeACNN}
  Let $L$, $d \in \N$, $d_0, \dots, d_{L+1} \in \N$ with $d^2 = d_0$
  and $d_{L+1}=1$, and let $\sigma \colon \R \to \R$. Let $\Phi$ be a
  neural network with architecture $(\sigma; d_0, \dots, d_{L+1})$.
	
  Then, there exists a convolutional neural network $\Psi$ with $L+1$
  layers, channel sizes $1, d_1, \dots, d_L, 1$, activation functions
  $\sigma_{\ell} = \sigma$ for $\ell =0, \dots, L-1$ and
  $\sigma_{L} = {\rm id}$, such that \eqref{eq:transferencePrincipleB}
  holds and $\size(\Phi) = \size(\Psi)$.
\end{theorem}
\begin{proof}
  We construct the convolutional neural network $\Psi$ block by block
  and keep track of its size. To distinguish between $\Phi$ and
  $\Psi$, it will be convenient to denote the weights and biases of
  $\Phi$ by
  \begin{equation*}
    \BV^{(\ell)}\in\R^{d_{\ell+1}\times d_{\ell}},~\Bc^{(\ell)}\in\R^{d_{\ell+1}}\qquad\text{for all }\ell=0,\dots,L.
  \end{equation*}
	
  \textbf{Step 1 (First block).}  The first block
  \begin{equation*}
    \CC^{(1)}:\R^{d\times d\times 1}\to\R^{d\times d \times d_1}
  \end{equation*}
  of $\Psi$ will serve to emulate the first layer of $\Phi$, which
  maps from $\R^{d_0}=\R^{d^2}\to\R^{d_1}$. Concretely, we will define
  weights
  \begin{equation*}
    \BW^{(0)}\in \R^{d\times d \times 1 \times d_1}\qquad\text{and}\qquad
    \Bb^{(0)}\in\R^{d_1}
  \end{equation*}
  with at most $\|\BV^{(0)}\|_0 + \|\Bc^{(0)}\|_0$ nonzero entries, so
  that the corresponding convolutional block $\CC^{(1)}$ reproduces
  $\Bx \mapsto \sigma(\BV^{(0)} \Bx + \Bc^{(0)}):\R^{d_0}\to\R^{d_1}$.
	
  We start by reshaping each row of
  $\BV^{(0)}\in\R^{d_1\times d_0}=\R^{d_1\times d^2}$ into a
  $d \times d$ matrix by applying $\mathcal{I}_{\rm flat}^{-1}$.
  Denote the resulting matrices by $(\BW_{k}^{(0)})_{k=1}^{d_1}$, and
  set with the notation from \eqref{eq:checknotation}
	$$
	\BW^{(0)}_{\cdot, \cdot, 1,k} = \widecheck{\left(T_{-1,-1}
            \BW_{k}^{(0)}\right)} %
        \qquad\text{for all }k=1,\dots,d_1.
	$$
        Then with
        $\BX = \mathcal{I}_{\rm flat}^{-1}(\Bx)\in\R^{d\times d}$
        (cf.~\eqref{eq:howtousecheck})
	\begin{align*}
          \left(\BW^{(0)}_{\cdot, \cdot, 1, k} * \BX\right)_{1,1} &= \left( \widecheck{\left(T_{-1,-1}\BW_{k}^{(0)}\right)}%
                                                                    *\BX \right)_{1,1}\\
                                                                  & = \sum_{i,j=1}^d \left(T_{-1,-1}\BW^{(0)}_k\right)_{i-1,j-1} X_{i,j} \\ 
                                                                  &= \sum_{i,j=1}^d(\BW_k^{(0)})_{i,j} X_{i,j} = (\BV^{(0)}\Bx)_k.
	\end{align*}
        Next, set the bias of the first convolutional block
        $\mathcal{C}^{(1)}$ to the bias of the first layer of $\Phi$,
        i.e.\ $\Bb^{(0)}\dfn\Bc^{(0)}\in\R^{d_1}$.  Then by
        \eqref{eq:convblockdef}
	$$
	\mathcal{C}^{(1)}(\BX)_{1,1,k} = (\sigma(\BV^{(0)} \Bx +
        \Bc^{(0)}))_k
	$$
	for $k = 1, \dots, d_1$. By construction
        $\size(\mathcal{C}_1) = \|\BV^{(0)}\|_0 + \|\Bc^{(0)}\|_0$.
	
	\textbf{Step 2 (Blocks $2$ to $L$).}  For $\ell = 2, \dots, L$
        we construct the convolutional block $\mathcal{C}^{(\ell)}$ by
        setting $\Bb^{(\ell-1)} = \Bc^{(\ell-1)}$ and defining
        $\BW^{(\ell-1)}\in\R^{d\times d\times d_{\ell-1}\times
          d_{\ell}}$ via
        \begin{equation*}
          W^{(\ell-1)}_{i,j,k, m}\dfn          
          \begin{cases}
            V^{(\ell-1)}_{m, k}&\text{if }i=j= d,\\ 
            0 &\text{otherwise}.
          \end{cases}
        \end{equation*}
        Let $\BX \in \R^{d \times d \times d_{\ell-1}}$.  Then for
        $m = 1, \dots, d_{\ell}$ and $k = 1, \dots, d_{\ell-1}$
	\begin{align*}
          \left(\BW^{(\ell-1)}_{\cdot, \cdot, k, m} * \BX_{\cdot, \cdot, k}\right)_{1,1} =  V^{(\ell-1)}_{m, k} X_{1, 1, k}.
	\end{align*}  
        Therefore
        $(\mathcal{C}^{(\ell)}(\BX))_{1,1,m} = \sigma(\BV^{(\ell-1)}
        \BX_{1, 1, \cdot} + \Bc^{(\ell-1)})_m$ for
        $m = 1, \dots d_{\ell}$. We have that
        $\size(\mathcal{C}^{(\ell)}) = \|\BV^{(\ell-1)}\|_0 +
        \|\Bc^{(\ell-1)}\|_0$.
	
	\textbf{Step 3 (Block $L+1$).}  We construct
        $\mathcal{C}^{(L+1)}$ precisely like the blocks in Step 2, but
        with the identity as activation function. Naturally,
        $\size(\mathcal{C}^{(L+1)}) = \|\BV^{(L)}\|_0 +
        \|\Bc^{(L)}\|_0$.
	
	\textbf{Step 4.}  Per construction the whole convolutional
        neural network $\Psi$ has
	$$
	\size(\Psi) = \sum_{\ell=0}^{L} \|\BV^{(\ell)}\|_0 +
        \|\Bc^{(\ell)}\|_0 = \size(\Phi).
	$$ 
	Moreover, \eqref{eq:transferencePrincipleB} holds by
        construction. This concludes the proof
      \end{proof}

      By definition, a convolutional neural network alternatingly
      applies linear transformations and an elementwise nonlinear
      activation function to the input
      $\BX\in\R^{d \times d\times C_0}$.  It can thus be viewed as a
      standard feedfoward neural network with the same number of
      layers. However, due to the weight sharing in the convolution
      (the same weights are applied at $d^2$ spatial positions) the
      number of non-zero entries in the weight matrices increases by a
      factor of $d^2$ when written in the standard form.  The next
      theorem makes this statement precise.

\begin{theorem}\label{thm:ACNNCanBeANN}
  Let $L$, $d \in \N$, and let $\sigma \colon \R \to \R$. Let
  $\Psi:\R^{d\times d\times C_0}\to\R^{d\times d\times C_{L}}$ be a
  convolutional neural network as in Definition \ref{def:convnet} with
  $L$ layers using $\sigma$ as activation function in each layer, and
  with channel sizes $C_0=1$ and $C_1,\dots,C_L\in\N$.
  
  Then, there exists a neural network $\Phi:\R^{d^2}\to\R$ with $L$
  layers and architecture $(\sigma; d_0, \dots, d_{L}, 1)$, where
  $d_j=C_j d^2$ for $j=0,\dots,L$ and such that
  \eqref{eq:transferencePrincipleB} holds.  Moreover, it holds that
  $\size(\Phi) \leq d^2 \size(\Psi)$.
\end{theorem}
      
\begin{proof}
  Let $\BW^{(0)}\in\R^{d\times d\times C_0\times C_1}$ and
  $\BX\in\R^{d\times d\times C_0}$. For each $m\in\{1,\dots,C_1\}$,
  there exists a matrix $\BV_m^{(0)}\in\R^{d^2\times C_0 d^2}$ such
  that
  \begin{equation*}
    \mathcal{I}_{\rm flat}\Big(\sum_{k=1}^{C_0}\BW^{(0)}_{\cdot, \cdot, k, m} * \BX_{\cdot, \cdot, k}\Big) = \BV^{(0)}_m \mathcal{I}_{\rm flat}(\BX)
  \end{equation*}
  since the left-hand side defines a linear map in $\BX$.  It is not
  hard to see that
  $\|\BV_m^{(0)}\|_0 \leq d^2 \|\BW^{(0)}_{\cdot, \cdot, 1, m}\|_0$.
	
  Define the matrix
	$$
	\BV^{(0)} = \left( \begin{array}{c} \BV^{(0)}_1\\
                             \vdots\\
                             \BV^{(0)}_{C_1} \end{array}\right)
  \in\R^{C_1 d^2\times C_0^2 d^2},
	$$
	and %
        a vector
        $\Bc^{(0)} = (b^{(0)}_1, b^{(0)}_1, \dots, b^{(0)}_2, \dots,
        b^{(0)}_{C_{1}}) \in \R^{d^2 C_{1}}$ such that each
        $b^{(0)}_m$ appears $d^2$ times.

        For a vector $\Bz \in \R^{d^2C_1}$ denote by
        $\Bz_{d^2 [m-1, m]}$ the subvector associated to the indices
        in the interval $[d^2(m-1), d^2m]$. Then
        \begin{equation*}
          (\BV^{(0)} \CI_{\rm flat}(\BX) + \Bc^{(0)})_{d^2 [m-1, m]} = \mathcal{I}_{\rm flat}\left(\sum_{k=1}^{C_0}\BW^{(0)}_{\cdot, \cdot, k, m} * \BX_{\cdot, \cdot, k}  + b^{(0)}_m\right).
        \end{equation*}
	Naturally,
        $\|\BV^{(0)}\|_0 \leq d^2 \sum_{k=1}^{C_0}
        \sum_{m=1}^{C_1}\|\BW^{(0)}_{\cdot, \cdot, k, m}\|_0$ and
        $ \norm[0]{\Bc^{(0)}} \leq d^2\|\Bb^{(0)}\|_0$. After applying
        $\sigma$ componentwise to the left and right-hand side, we
        have constructed the flattened output of the first
        convolutional layer $\CC^{(1)}$ of $\Psi$ as a one layer
        standard feedforward neural network.

	Using the same construction, each subsequent layer of the
        convolutional neural network can also be transformed into a
        regular neural network layer (simply by replacing $C_0$, $C_1$
        with $C_{\ell-1}$, $C_\ell$ in the above calculation). The two
        networks $\Phi$ and $\Psi$ are then related by flattening, and
        it holds $\size(\Phi)\le d^2\size(\Psi)$.
      \end{proof}

      From Theorems \ref{thm:EveryNNCanBeACNN} and
      \ref{thm:ACNNCanBeANN}, we see that convolutional neural
      networks and standard feedforward neural networks are closely
      related. A neural network can be rebuilt by considering only the
      first spatial output coordinate of a convolutional neural
      network with the same size.  On the other hand, to rebuild a
      convolutional neural network, we require a neural network that
      is larger than the convolutional neural network. In fact, this
      gap is not just an artifact of our proof.  It was shown in
      \cite{sepliarskaia2024vc} that when considering function classes
      of convolutional neural networks and neural networks with the
      same number of parameters, then the class of convolutional
      neural networks is considerably larger. More specifically, it
      has a significantly larger VC dimension.

\begin{remark}
  The translation equivariance property that was central in this
  section will cease to be present when pooling is used. In fact, it
  was demonstrated in \cite{yarotsky2022universal} that depending on
  the amount of pooling that is performed throughout a convolutional
  neural network architecture, the resulting function will move
  between translationally equivariant and translationally invariant.
\end{remark}

\subsection{When should a convolutional neural network be used?}
Convolutional neural networks excel if the underlying data exhibits
features that are translation-agnostic and where local relationships
between coordinates are meaningful.  If this is not satisfied, then
the translation agnosticity to features of the resulting functions
will not be meaningful and potentially introduce unnecessary overhead.
Also, if long-range dependencies, as in natural language processing,
are important for the model, then convolutional neural networks with
small filter sizes may have problems replicating these functions.  In
addition, we have seen that convolutional neural networks are more
expressive than their fully-connected counterparts. As a result, they
will generalize worse in scenarios where little training data is
available.

\section{Transformers}

We saw in the previous section that convolutional neural networks are
highly effective for modeling \emph{local dependencies}.  This is
because the convolutional filters act on small neighborhoods of the
input.  However, convolutional layers are ill-suited to capture
long-range correlations.

Transformers, introduced in \cite{vaswani2017attention}, provide a
mechanism for harnessing \emph{global dependencies}.  These models
have been extremely successful at tasks where data exhibits such
structures.  For example, transformers such as BERT
\cite{devlin2019bert}, GPT-3 \cite{brown2020language}, LLaMA
\cite{touvron2023llama}, and PaLM \cite{chowdhery2023palm} are the
basis of chatbots that assist millions of people.  Moreover, they %
underlie \emph{AlphaFold 2} \cite{jumper2021highly} for the prediction
of protein folding, %
whose creators were awarded the Nobel prize in Chemistry in 2024.

Below, we will review the associated architecture.  Transformers act
on a sequence of vectors. To understand them, it is helpful to
consider the application of natural language processing, which will be
our guiding example. To this end, we first explain in Subsection
\ref{sec:Embeddings} how to map text to a sequence of vectors.  The
main building block of a transformer, self-attention, is introduced in
Subsection \ref{sec:attentionMech}.  We then define a specific type of
transformers in Subsection \ref{sec:trafoBlock}.  Finally, in
Subsection \ref{sec:TextGen}, we explain how to generate %
text with transformers.

Throughout this section the following notation will be convenient: for
a function $f:\R^d\to\R^d$, we denote by
$f^{\rm col}:\R^{d\times n}\to\R^{d\times n}$ the columnwise
application, i.e.\ for $\BX=(\Bx_1,\dots,\Bx_n)\in\R^{d\times n}$
\begin{equation}\label{eq:columnnotation}
  f^{\rm col}(\BX)=[f(\Bx_1),\dots,f(\Bx_n)].
\end{equation}
In particular $f^{\rm col}$ acts independently on each column and does
not mix information between them.

\subsection{%
  Embeddings and positional encodings}\label{sec:Embeddings}

The first step in natural language processing is typically so-called
\emph{tokenization}: Let $\mathcal{V}$ be a set of tokens, which could
be syllables, parts of words, or whole words.  Each of these tokens is
then mapped to an embedding vector $\Be_v \in \mathbb{R}^d$ for all
$v \in \mathcal{V}$.  This embedding should maintain some structure of
the language, for example in the sense that tokens with similar
meaning correspond to some form of similarity in the embedding
vectors.

Additional to the tokens themselves, we would like to convey
information about the position of a token within a sentence to the
algorithm. Consider for example
\begin{quote}
  The dog chased the cat.
\end{quote}
or
\begin{quote}
  The cat chased the dog.
\end{quote}
While these sentences consist of the same words (or tokens), clearly
the position of the words ``cat'' and ``dog'' plays a key role in
their meaning. So-called positional encodings allow to store this
information. One possibility is to concatenate the embedding vector
with a positional encoding vector. The %
paper \cite{vaswani2017attention} proposes instead to add a positional
encoding to the embedding. Specifically, for a token $v\in\CV$ at
position $j$, the positional encoding vector $\Bs_j\in \R^d$, defined
via
\begin{equation}\label{eq:posenc}
  s_{j,2i} = \sin\!\left(\frac{j}{10000^{2i/d}}\right),
  \quad
  s_{j,2i+1} = \cos\!\left(\frac{j}{10000^{2i/d}}\right),
\end{equation}
is added to the embedding vector $\Be_v$ of this token.

The motivation for this positional encoding is the following: For each
embedding dimension $2i$ or $2i+1\in\{1,\dots,d\}$ it forms a
sinusoidal wave with wavelengths between $2\pi$ and $2\pi\cdot
10000$. The number $10000$ is a hyperparameter, that is chosen
sufficiently large depending on the maximum {\bf context length} (the
maximum number of tokens), to avoid repetition, i.e.\ $\Bs_j\neq\Bs_k$
for all $j\neq k$ less than a finite maximal context length. This
requires sufficiently large wavelengths. Moreover, the inner product
of $\Bs_j$ and $\Bs_k$ only depends on the distance $j-k$, see
Exercise \ref{ex:posenc}. This type of property is desirable since in
natural language, dependencies often exist in terms of relative
distance of the words rather than absolute position.

\subsection{The causal self-attention
  mechanism}\label{sec:attentionMech}

Let $n \in \N$ %
and let again $d \in \N$ be the embedding dimension.  A transformer
operates on a sequence of length $n$, which we represent via the data
matrix $\BX \in \R^{d \times n}$. Throughout the rest of this section
the sequence length (or context length) $n$ should not be understood
as fixed but as variable; the architecture can accommodate arbitrary
$n\in\N$ as we will see. This is in contrast to the feedforward neural
networks discussed in the first part of this book, which took a vector
of fixed length as input. In the context of natural language, each
column of $\BX$ corresponds to the embedding of a word %
as explained in Subsection \ref{sec:Embeddings}.  Mathematically, the
(masked) self-attention mechanism takes the matrix $\BX$ and returns
\begin{equation}\label{eq:attention}
  \BV{\rm softmax}^{\rm col}\Big(\frac{\BK^\top\BQ}{\sqrt{d}}+\BM\Big)\in\R^{d\times n}
\end{equation}
where
\begin{subequations}\label{eq:KQV}
  \begin{align}
    \BK &\dfn \BW_K \BX\in\R^{d\times n}&&\text{for some }\BW_K\in\R^{d\times d}\\
    \BQ &\dfn \BW_Q \BX\in\R^{d\times n}&&\text{for some }\BW_Q\in\R^{d\times d}\\
    \BV &\dfn \BW_V \BX\in\R^{d\times n}&&\text{for some }\BW_V\in\R^{d\times d},
  \end{align}
\end{subequations}
and $\BM\in\{0,-\infty\}^{n\times n}$. The matrix $\BM$ is called the
masking matrix and is fixed and not learned. If $\BM$ is constant
zero, this corresponds to unmasked self-attention. We will focus on
\begin{equation}\label{eq:masking}
  M_{ij}=\begin{cases}
           0 &\text{if }i\le j\\
           -\infty &\text{if }i>j,
         \end{cases}
       \end{equation}
       i.e.\ $\BM$ has $-\infty$ entries below the diagonal, and zero
       entries on and above the diagonal. In this case
       \eqref{eq:attention} is also referred to as {\bf causal
         self-attention}, and we will explain the terminology later.
       The softmax function, which will be recalled below, acts here
       as a mapping from $\R^n$ to $\R^n$ that is applied columnwise
       (cf.~\eqref{eq:columnnotation}).  The matrices $\BW_K$, $\BW_Q$
       and $\BW_V$ contain learnable parameters (independent of $n$).

       To unpack the meaning of \eqref{eq:attention}, we start by
       giving the usual interpretation of the remaining matrices.
       This interpretation is entirely heuristic, and merely serves as
       a motivation in the following.

       \begin{itemize}
       \item $\BQ$: The mapping represented by the matrix $\BW_Q$
         extracts the information a \emph{token seeks}. This is why
         the columns of $\BQ$ are referred to as the {\bf queries}.
       \item $\BK$: The mapping represented by the matrix $\BW_K$
         extracts the information a \emph{token has to offer}. This is
         why columns of $\BK$ are referred to as the {\bf keys}.
       \item $\VV$: The mapping represented by the matrix $\BW_V$
         extracts the \emph{meaning of a token}. This is why the
         columns of $\BV$ are referred to as the {\bf values}.
       \end{itemize}

       Equation \eqref{eq:attention} can now be interpreted as
       computing linear combinations of the values of the tokens
       (stored in the $n$ columns of $\BV$). The coefficients of these
       linear combinations are determined by the softmax term. Let us
       go into more detail on these coefficients.

\begin{definition}\label{def:importanceScores}
  Let $d$, $n\in \N$, denote the columns of $\BX\in\R^{d\times n}$ by
  $(\Bx_1,\dots,\Bx_n)$, and for $t\le n$ set
  $\BX_{[t]}\dfn (\Bx_1,\dots,\Bx_t)\in\R^{d\times t}$.  For
  $t = 1, \dots n$, the \textbf{importance score}
  $\tilde{\mathcal{I}}_{[t]}\in\R^t$ is given by
	$$
	\tilde{\mathcal{I}}_{[t]}(\BX) \coloneqq \frac{1}{\sqrt{d}}
        (\BW_K \BX_{[t]})^ \top (\BW_Q \Bx_t) = \frac{1}{\sqrt{d}}
        \left( \Bx_i^\top\BW_K^ \top \BW_Q
          \Bx_t\right)_{i=1}^t\in\R^t.
	$$
      \end{definition}

      The importance score $\tilde{\mathcal{I}}_{[t]}(\BX)$ compares
      the $t$-th column of $\BX$ with itself and all previous columns,
      by computing the scalar products of $\BW_K \Bx_i$ and
      $\BW_Q \Bx_t$ for $i=1,\dots,t$. %
      If the key embedding $\BW_K\Bx_i$ of $\Bx_i$ points into a
      similar direction as the query embedding of $\BW_q\Bx_t$ of
      $\Bx_t$, %
      then the word (or token) from the key embedding fits the query
      embedding and we believe that the words are important for each
      other. The purpose of the learned matrices $\BW_K$, $\BW_Q$ is
      thus to determine whether a token $\Bx_i$ is relevant to $\Bx_t$
      via $\Bx_i^\top\BW_K^\top\BW_Q\Bx_t$, or how well the \emph{key
        at position $i$ fits the query at position $t$}. The larger
      this value, the larger the important score of $\Bx_i$ for
      $\Bx_t$, and the more \emph{attention} will be given to it in
      the following.

      To make the importance scores better comparable and
      interpretable, one usually normalizes them. Therefore, we
      introduce the \textbf{normalized importance scores} next. %
      This is the purpose of the \textbf{softmax} function
      \begin{equation} \label{eq:softmaxDef}
        \begin{aligned}
          \mathrm{softmax} \colon &\R^t \to \R^t\\
                                  &\By\mapsto \Big(\frac{e^{y_i}}{\sum_{j=1}^t e^{y_j}}\Big)_{i=1}^t.
        \end{aligned}
      \end{equation}

\begin{definition}\label{def:importancescore}
  Let $d$, $n\in \N$ and $\BW_Q$, $\BW_K \in \R^{d \times d}$.  For
  $t=1,\dots,n$ and $\BX \in \R^{d \times n}$ the \textbf{normalized
    importance score} is given by
  $\mathcal{I}_{[t]}(\BX) =
  \mathrm{softmax}(\tilde{\mathcal{I}}_{[t]}(\BX))\in\R^t$.
\end{definition}

\begin{figure}[htb]
  \centering
  \includegraphics[width=0.85\textwidth]{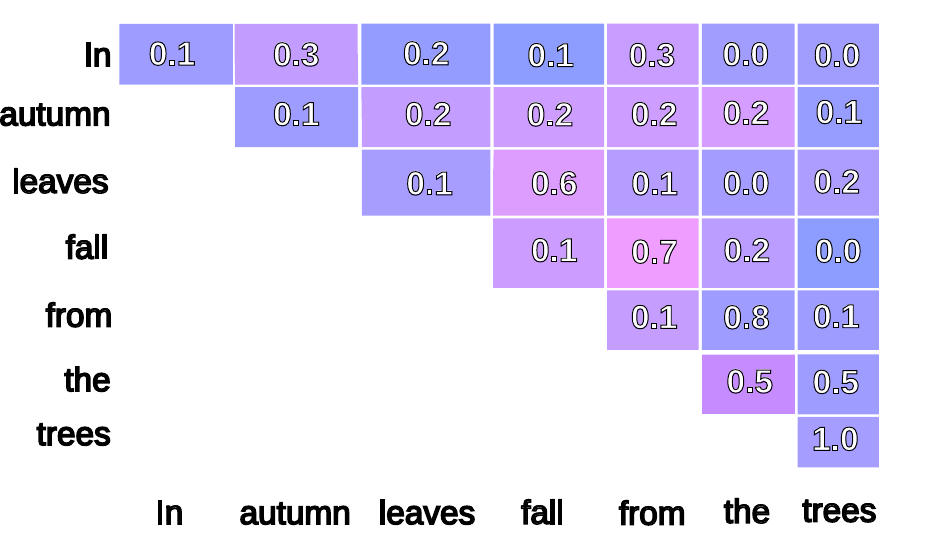}
  \caption{Normalized importance scores for an input $\BX$
    representing words. The columns correspond to the query and the
    rows to the keys.  For the query ``fall'', the largest importance
    is assigned to ``leaves'' instead of the word ``autumn'' even
    though the words ``fall'' and ``autumn'' can mean the same thing
    if fall is considered a noun. This is because the query and the
    key embedding make the importance scores measure how well words
    combine rather than merely measuring their similarity.}
  \label{fig:similariy scores}
\end{figure}

Based on the normalized importance scores, we now define the main
component of a (causal) transformer, which is the causal
self-attention mechanism. For the query at index $t$, it computes a
linear combination of the first $t$ values (the first $t$ columns of
$\BV$) weighted by the importance weights $\CI_{[t]}$ (corresponding
to the keys with indices $1,\dots,t$). Its role is so central, that
the authors of the original paper \cite{vaswani2017attention} named
their article ``Attention is all you need''.

\begin{definition}\label{def:attention}
  Let $d,n\in \N$ and $\BW_Q, \BW_K, \BW_V \in \R^{d \times d}$. We
  define for $\BX \in \R^{d \times n}$ the \textbf{self-attention}
  operator as
  \begin{align*}
    \Phi_{\rm at}(\BX; \BW_Q, \BW_K, \BW_V) \coloneqq [(\BW_V \BX_{[1]}) \mathcal{I}_{[1]}(\BX), \dots, (\BW_V \BX_{[n]}) \mathcal{I}_{[n]}(\BX)] \in \R^{d \times n}.
  \end{align*}
  We sometimes also write $\Phi_{\rm at}(\BX)$ for short.
\end{definition}

Since $\exp(-\infty)=0$, the $t$th column of the softmax term in
\eqref{eq:attention} corresponds exactly to $\mathcal{I}_{[t]}(\BX)$
padded with zeros. Therefore, \eqref{eq:attention} is compact notation
for the self-attention operator $\Phi_{\rm at}(\BX)$ in Definition
\ref{def:attention}.

The masking matrix $\BM$ in \eqref{eq:masking} ensures that the
normalized important scores below the diagonal vanish, and are thus
consistent with Definition \ref{def:importancescore}.  Concretely, the
$t$th column of $\Phi_{\rm at}(\BX)$ contains a convex combination of
all value embeddings of words at or before position $t$.  The
coefficients correspond to the normalized importance scores.  This
means that the value of the word at position $t$ is updated with
respect to the context of the previous tokens.  For text generation,
this masking is a crucial ingredient of the architecture: the model
should learn to predict the next word in a sentence. If it were
allowed to access the whole sentence, all it would learn is to shift
the words by one, but it would fail to learn the structure and
semantics required to generate new text. This is where the terminology
causality stems from. The term self-attention stems from the process
of computing importance scores of the tokens among each other.

\begin{remark}
  In practical implementations, attention is typically computed in
  parallel across several independent ``heads''.  Each head has its
  own learned projections, producing its own attention distribution.
  The outputs of all heads are then concatenated and linearly
  transformed to form the final result.  This construction is called
  {\bf multi-head attention} and allows the model to capture different
  types of dependencies simultaneously.  For example, one head may
  focus on syntactic relations while another captures semantic
  similarity.
\end{remark}

\subsection{Transformer blocks}\label{sec:trafoBlock}

As we have seen, %
self-attention updates the values of words with respect to the context
of the text. For more complicated texts, we would like to iterate this
process. Composing multiple attention blocks with one another is
precisely what leads to a transformer.

For added flexibility and stability, besides (multi-headed) attention,
the standard construction of a transformer includes three more key
ingredients: residual connections, feedforward layers, and layer
normalization. We are already familiar with the first two. Let us now
introduce normalization.

\begin{definition}\label{def:normalization}
  For $\Bx \in \mathbb{R}^d$, we define the \textbf{layer
    normalization} by
  \begin{align*}
    \Phi_{\rm la}:&\R^d\to\R^d\\
                  &\Bx\mapsto \Big(\frac{x_i - \mu(\Bx)}{\sqrt{v(\Bx)}}\Big)_{i=1}^d,
  \end{align*}
  where $\mu(\Bx) = \sum_{i=1}^d x_i/d$ and
  $v(\Bx) = \sum_{i=1}^d (x_i - \mu(\Bx))^2/d$.
\end{definition}
\begin{remark}
  Even though statisticians and applied mathematicians would expect
  the definition of $v$ to be the unbiased estimator of the variance
  given by $v(\Bx) = \sum_{i=1}^d (x_i - \mu(\Bx))^2/(d-1)$, the $v$
  defined in Definition \ref{def:normalization} is indeed used in
  layer normalization, \cite{vaswani2017attention}.
\end{remark}

We are now in position to introduce a transformer block, and remind
the reader again of the columnwise notation \eqref{eq:columnnotation}.

\begin{definition}\label{def:transformerBlock}
  Let $d$, $d_1$, $n \in \N$. Let
  $\Phi_{\rm at}:\R^{d\times n}\to\R^{d\times n}$ be a self-attention
  operator as in Definition \ref{def:attention} with parameters
  \begin{equation*}
    \BW_Q, \BW_K, \BW_V \in \R^{d \times d}
  \end{equation*}
  and let $\Phi:\R^d\to\R^d$ be a shallow (one hidden layer)
  feedforward neural network of width $\max\{d,d_1\}$ (cf.~Definition
  \ref{def:nn}) with weights and biases
  \begin{equation}\label{eq:trffwdparams}
    \BW^{(0)} \in \R^{d_1\times d},\quad \BW^{(1)} \in \R^{d \times d_1},\quad\Bb^{(0)} \in \R^{d_1},\quad\Bb^{(1)} \in \R^d,
  \end{equation}
  and activation function $\sigma \colon \R \to \R$.

  For $\BX \in \R^{d \times n}$, a {\bf transformer block}
  $\Phi_{\rm tr}:\R^{d\times n}\to\R^{d\times n}$ computes
  \begin{align*}
    \BY &\coloneqq \Phi_{\rm la}^{\rm col}\big(\BX + \Phi_{\rm at}(\BX)\big),\\
    \Phi_{\rm tr}(\BX) &\coloneqq \Phi_{\rm la}^{\rm col}\big(\BY + \Phi^{\rm col}(\BY)\big).
  \end{align*}
\end{definition}

\begin{figure}[htb]
  \centering
  \includegraphics[width=0.8\linewidth]{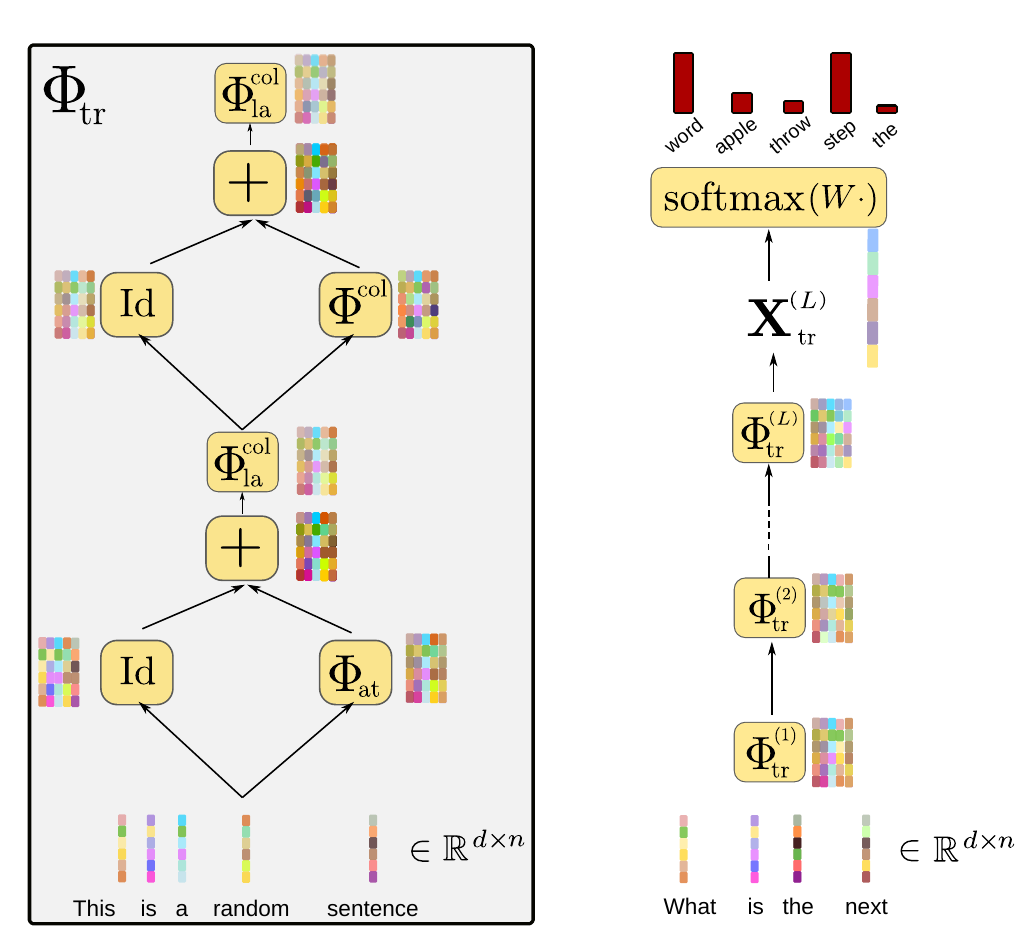}
  \caption{Sketch of a transformer. On the left, the transformer block
    of Definition \ref{def:transformerBlock} is shown. On the right,
    we see how %
    a transformer of depth $L$ %
    can be used to generate a probability distribution of potential
    next words in a sentence.}
  \label{fig:transformerBlock}
\end{figure}

A sketch of a transformer block is shown in Figure
\ref{fig:transformerBlock}.

\begin{remark}
  The columnwise application of the feedforward neural network $\Phi$
  to $\BY$ guarantees that the causality is not destroyed, i.e.\ no
  information from later tokens is propagated to earlier tokens by
  this operation.
\end{remark}

\begin{remark}
  In \cite{vaswani2017attention}, the %
  width $d_1$ of the hidden layer of the neural network is chosen as
  $4d$ and the activation function $\sigma$ is the ReLU.
\end{remark}

\begin{remark}
  Definition \ref{def:transformerBlock} shows the architecture as
  introduced in the original work
  \cite{vaswani2017attention}. However, it was found later that
  putting the normalization block before the attention block and
  neural network leads to more stable implementations
  \cite{xiong2020layer, budzinskiy2025numerical}.
\end{remark}

A transformer is simply a composition of multiple transformer blocks.
\begin{definition}
  Let $d$, $L$, $n \in \N$. Let
  $\Phi_{\rm tr}^{(1)}, \dots, \Phi_{\rm tr}^{(L)}$ be transformer
  blocks with input dimensions $d \times n$.  A \textbf{transformer}
  is a function
  $\Phi_{\rm tr} \colon \R^{d \times n} \to \R^{d \times n}$ such that
  for $\BX \in \R^{d \times n}$ it holds that
  $\Phi_{\rm tr}(\BX) = \BX_{\rm tr}^{(L)}$, where
  \begin{align*}
    \BX_{\rm tr}^{(0)} &= \BX\\
    \BX_{\rm tr}^{(j)} &= \Phi_{\rm tr}^{(j)}(\BX_{\rm tr}^{(j-1)}) \qquad \text{ for } j = 1, \dots, L.
  \end{align*}
  We call $L$ the depth of the transformer.
\end{definition}

As we have discussed before, an application of self-attention and,
therefore, also an application of a transformer block, will update the
value at each entry in a way that incorporates the context of the
text, specifically the earlier tokens.  Ideally, if this happens
multiple times, as in a transformer, the updated values at each
position will include all the relevant information from its preceding
text to make the word understandable.  The following example explains
how multiple layers of a transformer can improve the contextualization
of values compared to a single application.

\begin{example}
  Consider the sentence: \emph{``The leaf that fell from the tree was
    green.''}  After a single attention block, the query for ``was''
  may have an equally large attention value at both ``leaf'' and
  ``tree'', since both are nouns in its vicinity.  However, as the
  representations are refined through successive layers, the key for
  ``leaf'' increasingly encodes that it is the subject of the
  sentence, while the key for ``tree'' encodes that its role is to
  specify the nature of the subject leaf.  Thus in deeper layers, the
  attention of ``was'' concentrates on ``leaf''.
	
  This illustrates how multiple layers allow transformers to
  progressively learn grammatical structure, not in one step but
  through the accumulation of contextual information.
\end{example}

Once the input sentence is sufficiently transformed, it can be used to
generate text, as we will see in the following subsection.

\subsection{Text generation with transformers} \label{sec:TextGen}

A transformer produces a sentence one token at a time. Here we assume
there are $N_T \in \N$ available tokens in our dictionary, and we want
to identify one of these to extend a text fragment that we already
have.  Concretely, we assume we have a sentence of length $n$, which
corresponds to a %
matrix $\BX \in \R^{d \times n}$. Let $\Phi_{\rm tr}$ be a transformer
of depth $L\in \N$, with input and output dimension $d \times n$.

Set $\BX^{\rm context} = \Phi_{\rm tr}(\BX)$. Then, the output
$\BX^{\rm context}$ is a representation of the sentence corresponding
to $\BX$ that has undergone $L$ applications of transformers. Hence
the $j$th column %
contains a contextualized version under the previous tokens. In
particular, the last column has had access to all of $\BX$ during its
transformation.

To predict the next token, we therefore consider only the last column
$\Bx^{\rm context}_n$ %
of
$\BX^{\rm context} = [\Bx^{\rm context}_1, \dots, \Bx^{\rm
  context}_n]$ and compute for a matrix
\begin{equation}\label{eq:transformerW}
  \BW \in \R^{N_T \times d}
\end{equation}
the vector
$$
\Bp(\BX) \dfn \mathrm{softmax}\left( \BW \Bx^{\rm
    context}_n\right)\in\R^{N_T}.
$$
The vector $\Bp$ represents a probability distribution over all %
tokens. The process is depicted in Figure \ref{fig:transformerBlock}.
We now choose a token that has a high probability assigned by $\Bp$ to
create a new text of total length $n+1$.  This process is looped until
a sufficiently long text is generated.

\subsection{When should a transformer be used?}
Transformers are particularly effective in situations where long-range
dependencies appear, i.e., where inputs contain interactions between
elements far apart in a sequence or space (e.g., subject-verb
agreement across long sentences of words).  The architecture is
flexible enough to allow the context window to change and adapt to the
input.  This is in contrast to the fixed and local windows of a
convolutional neural network. However, transformers may not be ideal
when we encounter small data sets.

Moreover, one of the key disadvantages of transformers is their
quadratically increasing cost in the number $n$ of input tokens. While
the parameter matrices
\begin{equation*}
  \BW_K,~\BW_Q,~\BW_V\in\R^{d\times d}
\end{equation*}
as well as the parameters of the feedforward layers in
\eqref{eq:trffwdparams} and of the final embedding
\eqref{eq:transformerW}, are independent of $n$, the evaluation of
\eqref{eq:attention} has cost $O(n^2)$.  This is due to the term
\begin{equation*}
  \mathrm{softmax}^{\rm col}(\BX^\top\BW_K^\top\BW_Q\BX+\BM)\in\R^{n\times n}.
\end{equation*}
We also point out that while $\BX^\top\BW_K^\top\BW_Q\BX$ has rank at
most $d$, after the columnwise application of the softmax, the matrix
need not be low rank. Different strategies have been proposed in the
literature to address this quadratic scaling, and we will discuss one
in Exercise \ref{ex:linearatt}.

  \section*{Bibliography and further reading}
  The literature on deep learning architectures is incredibly
  vast. This is especially so because each architecture has numerous
  variations that have been studied and employed. Therefore, the
  references below are by no means comprehensive, but only offer a few
  pointers to where mathematically interesting results could be found.
  
  ResNets were introduced by \cite{he2016deep} to enable the training
  of very deep networks via skip connections.  Their universality
  already with only one neuron per layer was demonstrated in
  \cite{lin2018resnet}.  ResNets admit interpretations as
  discretizations of continuous-time dynamical systems, an observation
  that has motivated connections to control theory and differential
  equations. In particular, \cite{haber2017stable} interprets deep
  residual architectures as numerical schemes for optimal control
  problems. Moreover, \cite{chen2018neural} formalized the
  continuous-depth limit in the framework of neural ordinary
  differential equations.

  CNNs were developed in early work on handwritten character and
  zip-code recognition, where convolutional architectures were
  successfully trained using backpropagation
  \cite{lecun2002gradient,lecun1989backpropagation}.  Mathematical
  theory has focused on approximation and invariance properties:
  universality results were obtained by \cite{zhou2020universality}
  and \cite{yarotsky2022universal}. Moreover, approximation
  equivalence between CNNs and fully connected networks was shown in
  \cite{petersen2020equivalence}. Invariance and equivariance have
  been studied through harmonic analysis and learning-theoretic
  lenses, notably in the group-invariant scattering framework
  \cite{mallat2012group} or in the framework of group convolutional
  neural networks \cite{cohen2016group}. VC theory for deep group
  convolutional networks was studied in \cite{sepliarskaia2024vc}.

Transformers were introduced in \cite{vaswani2017attention} and form the basis of influential large language models \cite{devlin2019bert,touvron2023llama,chowdhery2023palm}. 
Transformer architectures have also achieved breakthrough results in scientific applications, most notably protein structure prediction with AlphaFold \cite{jumper2021highly}. 
Early results on the expressive power of transformers and their ability to approximate sequence-to-sequence mappings were established by \cite{1912.10077, yun2020n, alberti2023sumformer, 2507.10581}. Different variations of universality were studied, such as for in-context learning \cite{furuya2025transformers} or next-token prediction \cite{Sander2024TowardsUT}. For the training (in)stability of transformers we refer to \cite{2004.08249}, and a mean field analysis perspective analyzing the dynamics of transformers in simplified settings is provided in \cite{pmlr-v151-sander22a,MR4926874}.
Finally, for layer normalization we refer to the original paper \cite{1607.06450} and for further insights on its effect (in general and on transformers) to \cite{NEURIPS2019_2f4fe03d,xiong2020layer}.
    
  \newpage
  \section*{Exercises}
  \begin{exercise}\label{ex:resnet}
    Show that the construction in the proof of Step 1 of Lemma
    \ref{lemma:resnet} can be replaced with two residual blocks
    (instead of one) using one hidden ReLU (instead of two). Conclude
    that Lemma \ref{lemma:resnet} also holds with
    $d_1=\dots=d_L=1$. Convince yourself that also the proof of
    Theorem \ref{thm:resnet} can be adapted to $d_1=\dots=d_L=1$.

    \emph{Hint:} You need to construct a ReLU ResNet with $L=2$ and
    $d_1=d_2=1$, realizing the following type of function:

    \begin{center}
      \begin{tikzpicture}
        \draw [->,thick] (3.8,0) -- (7.1,0); \draw [->,thick] (4,-0.1)
        -- (4,2.5);

        \draw [thick] (3.8,1) -- (5.5,2.03) -- (5.7,0.7) -- (7.1,0.7);
      \end{tikzpicture}
    \end{center}

  \end{exercise}

  \begin{exercise}\label{ex:convolution}
    Let $\BX\in\R^{d\times d}$, $\BW\in\R^{d\times d}$ and $k_1$,
    $k_2\in\{0,\dots,d_1\}$. Show that
    \begin{equation*}
      T_{k_1,k_2}(\BW*\BX) = (T_{k_1,k_2}\BW)*\BX = \BW*(T_{k_1,k_2}\BX).
    \end{equation*}    
  \end{exercise}

  \begin{exercise}\label{ex:posenc}
    Let $\Bs_j$ be as in \eqref{eq:posenc}. Show that
    $\Bs_j^\top\Bs_k$ is a function of $j-k$ for all $j$, $k\in\Z$.

    \emph{Hint:}
    $\cos(\alpha-\beta) = \cos\alpha\cos\beta + \sin\alpha\sin\beta$
  \end{exercise}
  
  The following exercise is based on
  \cite{pmlr-v119-katharopoulos20a}.
  
  \begin{exercise}\label{ex:linearatt} 
    Suppose that $\phi:\R^d\to\R^m$ is a (feature) map such that for
    $\Bx$, $\By\in\R^d$
    \begin{equation*}
      \dup{\phi(\Bx)}{\phi(\By)}
    \end{equation*}
    is an approximation to $\exp(\Bx^\top\By/\sqrt{d})$.

    Denote for $\BX\in\R^{d\times n}$, as in
    \eqref{eq:attention}-\eqref{eq:masking},
    \begin{equation*}
      \Phi_{\rm att}(\BX)=\BV{\rm softmax}^{\rm col}\Big(\frac{\BK^\top\BQ}{\sqrt{d}}+\BM\Big)\in\R^{d\times n}.
    \end{equation*}
    Show that the $i$th column of $\Phi_{\rm att}(\BX)$ can then be
    approximated by
    \begin{equation}\label{eq:linatt}
      \frac{\sum_{j=1}^i\dup{\phi(\Bk_j)}{\phi(\Bq_i)}\Bv_j}{\sum_{j=1}^i\dup{\phi(\Bk_j)}{\phi(\Bq_i)}}.
    \end{equation}
    Here $\Bk_i$, $\Bq_i$, $\Bv_i$ denote the $i$th column of $\BK$,
    $\BQ$, $\BV$ in \eqref{eq:KQV}, respectively. Based on
    \eqref{eq:linatt}, propose an algorithm of complexity $O(n)$
    w.r.t.\ context length $n$, to approximately evaluate
    $\Phi_{\rm att}:\R^{d\times n}\to\R^{d\times n}$. What is a
    suitable choice of $\phi$?
  \end{exercise}

%% file: Appendix.tex
\appendix

\chapter{Probability theory}
This appendix provides some basic notions and results in probability theory required in the main text. It is intended as a revision for a reader already familiar with these concepts. For more details and proofs, we refer for example to the standard textbook \cite{klenke}.

\section{Sigma-algebras, topologies, and measures}\label{sec:sigtopmeas}
Let $\Omega$ be a set, and denote by $2^\Omega$ the powerset of $\Omega$. 

\begin{definition}\label{def:sigmaalgebra}
A subset $\mathfrak{A}\subseteq 2^\Omega$ is called a {\bf sigma-algebra}\footnote{We use this notation instead of the more common ``$\sigma$-algebra'' to avoid confusion with the activation function $\sigma$.} on $\Omega$ if it satisfies
\begin{enumerate}
\item $\Omega\in\mathfrak{A}$,
\item $A^c\in\mathfrak{A}$ whenever $A\in\mathfrak{A}$,
\item $\bigcup_{i\in\N}A_i\in\mathfrak{A}$ whenever $A_i\in\mathfrak{A}$ for all $i\in\N$.
\end{enumerate}
\end{definition}

For a sigma-algebra $\mathfrak{A}$ on $\Omega$, the tuple $(\Omega,\mathfrak{A})$ is also referred to as a {\bf measurable space}. For a measurable space, a subset $A\subseteq\Omega$ is called {\bf measurable}, if $A\in\mathfrak{A}$. Measurable sets are also called {\bf events}. 

Another key system of subsets of $\Omega$ is that of a topology.
\begin{definition}\label{def:topology}
  A subset $\mathfrak{T}\subseteq 2^\Omega$ is called a {\bf topology} on $\Omega$ if it satisfies
\begin{enumerate}
\item $\emptyset$, $\Omega\in\mathfrak{T}$,
\item $\bigcap_{j=1}^n O_j\in\mathfrak{T}$ whenever $n\in\N$ and $O_1,\dots,O_n\in\mathfrak{T}$,
\item $\bigcup_{i\in I}O_i\in\mathfrak{T}$ whenever for an index set $I$ holds
  $O_i\in\mathfrak{T}$ for all $i\in I$.
\end{enumerate}
If $\mathfrak{T}$ is a topology on $\Omega$, we call
$(\Omega,\mathfrak{T})$ a {\bf topological space}, and a set
$O\subseteq \Omega$ is called {\bf open} if and only if
$O\in\mathfrak{T}$.
\end{definition}

\begin{remark}
  The two notions differ in that a topology allows for unions of \emph{arbitrary} (possibly uncountably many) sets, but only for \emph{finite} intersection, whereas a sigma-algebra allows for countable unions and intersections.
\end{remark}

\begin{example}
  Let $d\in\N$ and denote by
  $B_\eps(\Bx)=\set{\By\in\R^d}{\norm{\By-\Bx}<\eps}$ the set of
  points whose Euclidean distance to $\Bx$ is less than $\eps$. Then
  for every $A\subseteq\R^d$, 
  the smallest topology on $A$ containing $A\cap B_\eps(\Bx)$ for all
  $\eps>0$, $\Bx\in\R^d$, is called the {\bf Euclidean
    topology} on $A$.
\end{example}

If $(\Omega,\mathfrak{T})$ is a topological space, then the {\bf Borel sigma-algebra} refers to the smallest sigma-algebra on $\Omega$ containing all open sets, i.e.\ all elements of $\mathfrak{T}$. Throughout this book, subsets of $\R^d$ are always understood to be equipped with the Euclidean topology and the Borel sigma-algebra. The Borel sigma-algebra on $\R^d$ is denoted by $\mathfrak{B}_d$.

We can now introduce measures.

\begin{definition}\label{def:probMeasure}
  Let $(\Omega,\mathfrak{A})$ be a measurable space. A mapping $\mu:\mathfrak{A}\to [0,\infty]$ is called a {\bf measure} if it satisfies
  \begin{enumerate}
  \item $\mu(\emptyset)=0$,
  \item for every sequence $(A_i)_{i\in\N}\subseteq \mathfrak{A}$ such that $A_i\cap A_j=\emptyset$ whenever $i\neq j$, it holds
    \begin{equation*}
      \mu\Big(\bigcup_{i\in\N}A_i\Big) = \sum_{i\in\N}\mu(A_i).
    \end{equation*}
  \end{enumerate}
    We say that the measure is {\bf finite} if $\mu(\Omega)<\infty$, and it is {\bf sigma-finite} if there exists a sequence $(A_i)_{i\in\N}\subseteq\mathfrak{A}$ such that $\Omega=\bigcup_{i\in\N}A_i$ and $\mu(A_i)<1$ for all $i\in\N$. In case $\mu(\Omega)=1$, the measure is called a {\bf probability measure}.  
  \end{definition}

\begin{example}
  One can show that there exists a unique measure $\lambda$ on $(\R^d,\mathfrak{B}_d)$, such that
  for all sets of the type $\times_{j=1}^d [a_i,b_i)$ with $-\infty<a_i\le b_i<\infty$ holds
  \begin{equation*}
    \lambda(\times_{i=1}^d [a_i,b_i)) = \prod_{i=1}^d (b_i-a_i).
  \end{equation*}
  This measure is called the {\bf Lebesgue measure}.
\end{example}

If $\mu$ is a measure on the measurable space $(\Omega,\mathfrak{A})$, then the triplet $(\Omega,\mathfrak{A},\mu)$ is called a {\bf measure space}. In case $\mu$ is a probability measure, it is called a {\bf probability space}.

Let $(\Omega,\mathfrak{A},\mu)$ be a measure space. A subset $N\subseteq\Omega$ is called a {\bf null-set}, if $N$ is measurable and $\mu(N)=0$. Moreover, an equality or inequality is said to hold $\mu$-{\bf almost everywhere} or $\mu$-{\bf almost surely}, if it is satisfied on the complement of a null-set. In case $\mu$ is clear from context, we simply write ``almost everywhere'' or ``almost surely'' instead. Usually this refers to the Lebesgue measure.

\section{Random variables}
\subsection{Measurability of functions}
To define random variables, we first need to recall the measurability of functions.
\begin{definition}
  Let $(\Omega_1,\mathfrak{A}_1)$ and $(\Omega_2,\mathfrak{A}_2)$ be two measurable spaces.
A function $f:\Omega_1\to \Omega_2$ is called {\bf measurable} if
  \begin{equation*}
    f^{-1}(A_2)\dfn \set{\omega\in\Omega_1}{f(\omega)\in A_2}\in\mathfrak{A}_1\qquad \text{for all }A_2\in\mathfrak{A}_2.
  \end{equation*}
  A mapping $X:\Omega_1\to\Omega_2$ is called a {\bf $\Omega_2$-valued random variable} if it is measurable.
\end{definition}

  \begin{remark}
    We again point out the parallels to topological spaces: A function $f:\Omega_1\to\Omega_2$
    between two topological spaces $(\Omega_1,\mathfrak{T}_1)$ and $(\Omega_2,\mathfrak{T}_2)$
    is called {\bf continuous} if $f^{-1}(O_2)\in\mathfrak{T}_1$ for all $O_2\in\mathfrak{T}_2$.
  \end{remark}

Let $\Omega_1$ be a set and let $(\Omega_2,\mathfrak{A}_2)$ be a measurable space. For $X:\Omega_1\to\Omega_2$, we can ask for the smallest sigma-algebra $\mathfrak{A}_X$ on $\Omega_1$, such that $X$ is measurable as a mapping from $(\Omega_1,\mathfrak{A}_X)$ to $(\Omega_2,\mathfrak{A}_2)$. Clearly, for every sigma-algebra $\mathfrak{A}_1$ on $\Omega_1$, $X$ is measurable as a mapping from $(\Omega_1,\mathfrak{A}_1)$ to $(\Omega_2,\mathfrak{A}_2)$ if and only if every $A\in \mathfrak{A}_X$ belongs to $\mathfrak{A}_1$; or in other words, $\mathfrak{A}_X$ is a sub sigma-algebra of $\mathfrak{A}_1$. It is easy to check that $\mathfrak{A}_X$ is given through the following definition.

\begin{definition}
Let $X:\Omega_1\to\Omega_2$ be a random variable. Then
\begin{equation*}
  \mathfrak{A}_X\dfn \set{X^{-1}(A_2)}{A_2\in\mathfrak{A}_2}\subseteq 2^{\Omega_1}
\end{equation*}
is the {\bf sigma-algebra induced by $X$} on $\Omega_1$.
\end{definition}

\subsection{Distribution and expectation}\label{sec:distAndExp}
Now let $(\Omega_1,\mathfrak{A}_1,\bbP)$ be a probability space, and let $(\Omega_2,\mathfrak{A}_2)$ be a measurable space. Then $X$ naturally induces a measure on $(\Omega_2,\mathfrak{A}_2)$ via
\begin{equation*}
  \bbP_X[A_2] \dfn \bbP[X^{-1}(A_2)]\qquad\text{for all }A_2\in\mathfrak{A}_2.
\end{equation*}
Note that due to the measurability of $X$ it holds
$X^{-1}(A_2)\in\mathfrak{A}_1$, so that $\bbP_X$ is well-defined. 
\begin{definition}\label{def:distribution}
  The measure $\bbP_X$ is called the {\bf distribution} of $X$.
  If $(\Omega_2,\mathfrak{A}_2)=(\R^d,\mathfrak{B}_d)$, and there exists
  a function $f_X:\R^d\to\R$ such that
  \begin{equation*}
    \bbP[A] = \int_A f_X(\Bx)\dd \Bx\qquad \text{ for all } A\in\mathfrak{B}_d,
  \end{equation*}
  then $f_X$ is called the {\bf (Lebesgue) density} of $X$.
\end{definition}

\begin{remark}
  The term distribution is often used without specifying an underlying
  probability space and random variable. In this case, ``distribution'' stands
  interchangeably for ``probability measure''. For example,
    \textit{$\mu$ is a distribution on $\Omega_2$}
  states that $\mu$ is a probability measure on the measurable
  space $(\Omega_2,\mathfrak{A}_2)$. In this case, there always exists
  a probability space $(\Omega_1,\mathfrak{A}_1,\bbP)$ and a random variable
  $X:\Omega_1\to\Omega_2$ such that $\bbP_X=\mu$; namely
  $(\Omega_1,\mathfrak{A}_1,\bbP)=(\Omega_2,\mathfrak{A}_2,\mu)$ and $X(\omega)=\omega$.
\end{remark}

\begin{example}
  Some important distributions include the following.
  \begin{itemize}
  \item {\bf Bernoulli distribution}: A random variable $X:\Omega\to\{0,1\}$ is
    Bernoulli distributed if there exists $p\in [0,1]$ such that $\bbP[X=1]=p$
    and $\bbP[X=0]=1-p$.
  \item {\bf Uniform distribution}: A random variable $X:\Omega\to\R^d$ is
    uniformly distributed on a measurable set $A\in\mathfrak{B}_d$,
    if its density equals
    \begin{equation*}
      f_X(\Bx)=\frac{1}{|A|}\ind_A(\Bx)
    \end{equation*}
      where $|A|<\infty$ is
    the Lebesgue measure of $A$.
  \item {\bf Gaussian distribution}: A random variable $X:\Omega\to\R^d$ is
    Gaussian distributed with mean $\Bm\in\R^d$ and the regular covariance
    matrix $\BC\in\R^{d\times d}$, if its density equals
    \begin{equation*}
      f_X(\Bx) = \frac{1}{(2\pi\det(\BC))^{d/2}}\exp\left(-\frac{1}{2}(\Bx-\Bm)^\top\BC^{-1}(\Bx-\Bm)\right).
    \end{equation*}
    We denote this distribution by $\gauss(\Bm,\BC)$.
  \end{itemize}
\end{example}

Let $(\Omega,\mathfrak{A},\bbP)$ be a probability space, let
$X:\Omega\to\R^d$ be an $\R^d$-valued random variable. We then
call the Lebesgue integral
\begin{equation}\label{eq:expectationDefinition}
  \bbE[X]\dfn \int_{\Omega}X(\omega)\dd\bbP(\omega) = \int_{\R^d}\Bx\dd\bbP_X(\Bx)
\end{equation}
the {\bf expectation} of $X$. Moreover, for $k\in\N$ %
we say that $X$ has 
{\bf finite $k$-th moment} if $\bbE[\norm[]{X}^k]<\infty$.
Similarly, for a probability measure $\mu$ on $\R^d$
and $k\in\N$, we say that $\mu$ has finite $k$-th moment if
\begin{equation*}
  \int_{\R^d}\norm{\Bx}^k\dd\mu(\Bx)<\infty.
\end{equation*}
Furthermore, the matrix
\begin{equation*}
  \int_{\Omega}(X(\omega)-\bbE[X])(X(\omega)-\bbE[X])^\top\dd\bbP(\omega)\in\R^{d\times d}
\end{equation*}
is the {\bf covariance} of $X:\Omega\to\R^d$. For $d=1$, it is called the {\bf variance} of $X$
and denoted by $\bbV[X]$.

Finally, we recall different variants of convergence for random variables.

\begin{definition}\label{def:Xnconvergence}
  Let $(\Omega,\mathfrak{A},\bbP)$ be a probability space,
  let $X_j:\Omega\to\R^d$, $j\in\N$, be a sequence of random variables,
  and let $X:\Omega\to\R^d$ also be a random variable.
  The sequence is said to
  \begin{enumerate}
  \item {\bf converge almost surely} to $X$, if
    \begin{equation*}
      \bbP\left[\setc{\omega\in\Omega}{\lim_{j\to\infty}X_j(\omega)=X(\omega)}\right]=1,
    \end{equation*}
  \item {\bf converge in probability} to $X$, if
    \begin{equation*}
      \text{for all }\eps>0:\quad\lim_{j\to\infty}\bbP\left[\setc{\omega\in\Omega}{|X_j(\omega)-X(\omega)|>\eps}\right]=0,
    \end{equation*}
  \item {\bf converge in distribution} to $X$, if for all bounded continuous
    functions $f:\R^d\to\R$
    \begin{equation*}
      \lim_{j\to\infty}\bbE[f\circ X_j]=\bbE[f\circ X].
    \end{equation*}
  \end{enumerate}
\end{definition}

The notions in Definition \ref{def:Xnconvergence} are ordered by
decreasing strength, i.e.\ almost sure convergence implies convergence
in probability, and convergence in probability implies
convergence in distribution, see for example \cite[Chapter 13]{klenke}. Since
$\bbE[f\circ X] = \int_{\R^d}f(x)\dd\bbP_X(x)$, the
notion of %
convergence in distribution only depends on the distribution $\bbP_X$
of $X$. We thus also say that a sequence of random variables converges
in distribution
towards a measure $\mu$.  

\section{Conditionals, marginals, and independence}
In this section, we concentrate on $\R^d$-valued random variables, although
the following concepts can be extended to more general spaces.

\subsection{Joint and marginal distribution}
Let again $(\Omega,\mathfrak{A},\bbP)$ be a probability space, and let
$X:\Omega\to\R^{d_X}$, $Y:\Omega\to \R^{d_Y}$ be two 
random variables. Then
\begin{equation*}
  Z\dfn (X,Y):\Omega\to\R^{d_X+d_Y}
\end{equation*}
is also a random variable. Its distribution $\bbP_Z$ is a measure on the measurable space $(\R^{d_X+d_Y},\mathfrak{B}_{d_X+d_Y})$, and $\bbP_Z$ is referred to as the {\bf joint distribution} of $X$ and $Y$. On the other hand, $\bbP_{X}$, $\bbP_Y$ are called the {\bf marginal distributions} of $X$, $Y$. %
Note that
\begin{equation*}
  \bbP_{X}[A] = \bbP_Z[A\times\R^{d_Y}]\qquad\text{for all }A\in \mathfrak{B}_{d_X},
\end{equation*}
and similarly for $\bbP_{Y}$. Thus the marginals $\bbP_X$, $\bbP_Y$, can be constructed from the joint distribution $\bbP_Z$. In turn, knowledge of the marginals is not sufficient to construct the joint distribution.

\subsection{Independence}
The concept of independence serves to formalize the situation, where knowledge of one random variable provides no information about another random variable. We first give the formal definition, and afterwards discuss the roll of a die as a simple example. 

\begin{definition}
  Let $(\Omega,\mathfrak{A},\bbP)$ be a probability space. Then two events $A$, $B\in \mathfrak{A}$ are
  called {\bf independent} if
  \begin{equation*}
    \bbP[A\cap B] =\bbP[A]\bbP[B].
  \end{equation*}
  Two random variables $X:\Omega\to\R^{d_X}$ and $Y:\Omega\to\R^{d_Y}$ are called {\bf independent},
  if
  \begin{equation*}
    A,~B\text{ are independent for all }A\in \mathfrak{A}_{X},~B\in\mathfrak{A}_{Y}.
  \end{equation*}
\end{definition}

Two random variables are thus independent, if and only if all events in their induced sigma-algebras are independent. This turns out to be equivalent to the joint distribution $\bbP_{(X,Y)}$ being equal to the product measure $\bbP_X\otimes\bbP_Y$; the latter is characterized as the unique measure $\mu$ on $\R^{d_X+d_Y}$ satisfying $\mu(A\times B)=\bbP_X[A]\bbP_Y[B]$ for all $A\in\mathfrak{B}_{d_x}$, $B\in\mathfrak{B}_{d_Y}$.

\begin{example}\label{ex:dice}
  Let $\Omega=\{1,\dots,6\}$ represent the outcomes of rolling a fair die, let $\mathfrak{A}=2^\Omega$ be the sigma-algebra, and let $\bbP[\omega]=1/6$ for all $\omega\in\Omega$.
  Consider the three random variables
  \begin{equation*}
    X_1(\omega) = \begin{cases}
      0 &\text{if $\omega$ is odd}\\
      1 &\text{if $\omega$ is even}
    \end{cases}\quad
    X_2(\omega) = \begin{cases}
      0 &\text{if }\omega\le 3\\
      1 &\text{if }\omega\ge 4
    \end{cases}\quad
    X_3(\omega) = \begin{cases}
      0 &\text{if }\omega\in\{1,2\}\\
      1 &\text{if }\omega\in\{3,4\}\\
      2 &\text{if }\omega\in\{5,6\}.      
    \end{cases}
  \end{equation*}
These random variables can be interpreted as follows:
\begin{itemize}
\item $X_1$ indicates whether the roll yields an odd or even number.
\item $X_2$ indicates whether the roll yields a number at most $3$ or at least $4$.
\item $X_3$ categorizes the roll into one of the groups $\{1,2\}$, $\{3,4\}$ or $\{5,6\}$.
\end{itemize}
  The induced sigma-algebras are
  \begin{align*}
    \mathfrak{A}_{X_1}&=\{\emptyset,\Omega,\{1,3,5\},\{2,4,6\}\}\\
    \mathfrak{A}_{X_2}&=\{\emptyset,\Omega,\{1,2,3\},\{4,5,6\}\}\\
    \mathfrak{A}_{X_3}&=\{\emptyset,\Omega,\{1,2\},\{3,4\},\{5,6\},\{1,2,3,4\},\{1,2,5,6\},\{3,4,5,6\}\}.
  \end{align*}
  We leave it to the reader to formally check that $X_1$ and $X_2$ are
  not independent, but $X_1$ and $X_3$ are independent. This reflects
  the fact that, for example, knowing the outcome to be odd, makes it
  more likely that the number belongs to $\{1,2,3\}$ rather than
  $\{4,5,6\}$. However, this knowledge provides no information on the
  three categories $\{1,2\}$, $\{3,4\}$, and $\{5,6\}$.
\end{example}

If $X:\Omega\to\R$, $Y:\Omega\to\R$ are two independent random variables, then,
  due to $\bbP_{(X,Y)}=\bbP_X\otimes\bbP_Y$
  \begin{align*}
    \bbE[XY]&=\int_{\Omega} X(\omega)Y(\omega)\dd\bbP(\omega)\\
    &=\int_{\R^{2}} x y \dd\bbP_{(X,Y)}(x,y)\\
    &=\int_{\R}x \dd\bbP_X(x)\int_{\R}y \dd\bbP_X(y)\\
    &=\bbE[X]\bbE[Y].
  \end{align*}
Using this observation, it is easy to see that for a sequence of independent $\R$-valued random variables $(X_i)_{i=1}^n$ with bounded second moments, there holds \textbf{Bienaym\'e's identity}
\begin{align}\label{eq:BienaymeIdentity}
	 \bbV\left[\sum_{i=1}^n X_i\right] = \sum_{i=1}^n \bbV\left[X_i\right].
\end{align}

\subsection{Conditional distributions}\label{sec:conddist}

Let $(\Omega,\mathfrak{A},\bbP)$ be a probability space, and let $A$, $B\in\mathfrak{A}$ be two events.
In case $\bbP[B]>0$, we define
\begin{equation}\label{eq:PAB}
  \bbP[A|B]\dfn \frac{\bbP[A\cap B]}{\bbP[B]},
\end{equation}
and call $\bbP[A|B]$ the {\bf conditional probability of $A$ given $B$}.

\begin{example}
  Consider the setting of Example \ref{ex:dice}. Let $A=\set{\omega\in\Omega}{X_1(\omega)=0}$
  be the event that the outcome of the die roll was an odd number and let
  $B=\set{\omega\in\Omega}{X_2(\omega)=0}$ be the event that the outcome yielded
  a number at most $3$. Then $\bbP[B]=1/2$, and $\bbP[A\cap B]=1/3$. Thus
  \begin{equation*}
    \bbP[A|B]=%
    \frac{\bbP[A\cap B]}{\bbP[B]} = \frac{1/3}{1/2} = \frac{2}{3}.
  \end{equation*}
  This reflects that, given we know the outcome to be at most $3$, the probability
  of the number being odd, i.e.\ in $\{1,3\}$, is larger than the probability of the
  number being even, i.e.\ equal to $2$.
\end{example}

The conditional probability in \eqref{eq:PAB} is only well-defined if
$\bbP[B]>0$. In practice, we often encounter the case where we would
like to condition on an event of probability zero.

\begin{example}\label{ex:Xp}
  Consider the following procedure: We first draw a random number
  $p\in [0,1]$ according to a uniform distribution on $[0,1]$.
  Afterwards we draw a random number $X\in\{0,1\}$ according to a
  $p$-Bernoulli distribution, i.e.\ $\bbP[X=1]=p$ and
  $\bbP[X=0]=1-p$. Then $(p,X)$ is a joint random variable
  taking values in $[0,1]\times\{0,1\}$.
  What is $\bbP[X=1|p=0.5]$ in
  this case? Intuitively, it should be $1/2$, but note that
  $\bbP[p=0.5]=0$, so that \eqref{eq:PAB} is not meaningful here.
\end{example}

\begin{definition}[regular conditional distribution]\label{def:rvcd1}
  Let $(\Omega,\mathfrak{A},\bbP)$ be a probability space, and let
  $X:\Omega\to \R^{d_X}$ and $Y:\Omega\to \R^{d_Y}$ be two random variables.
  Let $\tau_{X|Y}:\mathfrak{B}_{d_X}\times \R^{d_Y} \to [0,1]$ satisfy
  \begin{enumerate}
  \item $y\mapsto \tau_{X|Y}(A,y):\R^{d_Y}\to [0,1]$ is measurable for every
    fixed $A\in\mathfrak{B}_{d_X}$,
  \item $A\mapsto \tau_{X|Y}(A,y)$ is a probability measure on
    $(\R^{d_X},\mathfrak{B}_{d_X})$ for every %
    $y\in Y(\Omega)$,
  \item\label{item:gen} for all $A\in\mathfrak{B}_{d_X}$
    and all $B\in\mathfrak{B}_{d_Y}$ holds
    \begin{equation*}
      \bbP[X\in A,Y\in B]=\int_{B}\tau_{X|Y}(A,y)\bbP_{Y}(y).
    \end{equation*}
  \end{enumerate}
  Then $\tau$ is called a {\bf regular (version of the) conditional distribution
    of $X$ given $Y$}. In this case, we denote
  \begin{equation*}
    \bbP[X\in A|Y=y]:=\tau_{X|Y}(A,y),
  \end{equation*}
  and refer to this measure as the conditional distribution of $X|Y=y$.
\end{definition}

 Definition \ref{def:rvcd1} provides a mathematically rigorous way
  of assigning a distribution to a random variable conditioned on an
  event that may have probability zero, as in Example
  \ref{ex:Xp}. %
  Existence and uniqueness of these conditional distributions hold in
  the following sense, see for example \cite[Chapter 8]{klenke} or
  \cite[Chapter 3]{ScheichlZech2021} for the specific statement given
  here.

\begin{theorem}
  Let $(\Omega,\mathfrak{A},\bbP)$ be a probability space,
  and let $X:\Omega\to\R^{d_X}$, $Y:\Omega\to\R^{d_Y}$
  be two random variables. Then there exists a regular
  version of the conditional distribution $\tau_1$.
  
  Let $\tau_2$ be another regular version of the conditional distribution.
  Then there exists a $\bbP_Y$-null set $N\subseteq\R^{d_Y}$,
  such that for all $y\in N^c\cap Y(\Omega)$,
  the two probability measures $\tau_1(\cdot,y)$ and $\tau_2(\cdot,y)$ coincide.
\end{theorem}

In particular, conditional distributions are only well-defined in a
$\bbP_Y$-almost everywhere sense.

\begin{definition}
  Let $(\Omega,\mathfrak{A},\bbP)$ be a probability space, and let
  $X:\Omega\to\R^{d_X}$, $Y:\Omega\to\R^{d_Y}$, $Z:\Omega\to\R^{d_Z}$
  be three random variables. We say that $X$ and $Z$ are {\bf
    conditionally independent given $Y$}, if the two distributions
  $X|Y=y$ and $Z|Y=y$ are independent for $\bbP_Y$-almost every
  $y\in Y(\Omega)$.
\end{definition}

\section{Concentration inequalities}
Let $X_i:\Omega\to\R$, $i\in\N$, be a sequence of %
random variables with finite first moments.
The centered average over the first $n$ terms
\begin{equation}\label{eq:Sn}
  S_n\dfn \frac{1}{n}\sum_{i=1}^n (X_i-\bbE[X_i])
\end{equation}
is another random variable, and by linearity of the expectation it holds
$\bbE[S_n]=0$. %
The sequence is said to satisfy the {\bf strong law of large numbers}
if
\begin{equation*}
  \bbP\Big[\limsup_{n\to\infty} |S_n|=0\Big]=1.
\end{equation*}
This is for example the case if there exists $C<\infty$ such that
$\bbV[X_i]\le C$ for all $i\in\N$. Concentration inequalities provide
bounds on the rate of this convergence.

We start with Markov's inequality.
\begin{lemma}[Markov's inequality]
  Let $X:\Omega\to\R$ be a random variable,
  and let $\varphi:[0,\infty)\to [0,\infty)$ be monotonically increasing. Then for all $\eps>0$
  \begin{equation*}
    \bbP[|X|\ge\eps]\le \frac{\bbE[\varphi(|X|)]}{\varphi(\eps)}.
  \end{equation*}
\end{lemma}
\begin{proof}
  We have
  \begin{equation*}
    \bbP[|X|\ge\eps]
    =\int_{X^{-1}([\eps,\infty))} 1\dd\bbP(\omega)
    \le
    \int_{\Omega} \frac{\varphi(|X(\omega)|)}{\varphi(\eps)}\dd\bbP(\omega)
    =\frac{\bbE[\varphi(|X|)]}{\varphi(\eps)},
  \end{equation*}
  which gives the claim.
\end{proof}

Applying Markov's inequality with $\varphi(x)\dfn x^2$ to the random
variable $X-\bbE[X]$ directly gives Chebyshev's inequality.

\begin{lemma}[Chebyshev's inequality]
  Let $X:\Omega\to\R$ be a random variable with finite variance. Then for all $\eps>0$\label{lemma:chebyshev}%
  \begin{equation*}
    \bbP[|X-\bbE[X]|\ge\eps]\le \frac{\bbV[X]}{\eps^2}.
  \end{equation*}
\end{lemma}

From Chebyshev's inequality we obtain the next result,
which is a quite general concentration inequality
for random variables with finite variances.
  
  \begin{theorem}\label{thm:chebyscheff}
    Let $X_1,\dots,X_n$ be $n\in\N$ independent real-valued
    random variables such that for some $\varsigma>0$ holds
    $\bbE[|X_i-\mu|^2]\le \varsigma^2$ for all $i=1,\dots,n$. Denote
    \begin{equation}\label{eq:bbEmu}
      \mu\dfn \bbE\Big[\frac{1}{n}\sum_{j=1}^n X_j\Big].
    \end{equation}
    Then for all $\eps>0$
    \begin{equation*}
      \bbP\Bigg[\Big|\frac{1}{n}\sum_{j=1}^n X_j-\mu\Big|\ge \eps\Bigg]\le \frac{\varsigma^2}{\eps^2n}.
    \end{equation*}
  \end{theorem}

  \begin{proof}
    Let $S_n=\sum_{j=1}^n (X_i-\bbE[X_i])/n=(\sum_{j=1}^nX_i)/n-\mu$. By Bienaym\'e's identity \eqref{eq:BienaymeIdentity}, it holds that
    \begin{align*}
      \bbV[S_n]&=\frac{1}{n^2}\sum_{j=1}^n\bbE[(X_i-\bbE[X_i])^2] \le \frac{\varsigma^2}{n}.
    \end{align*}
    Since $\bbE[S_n]=0$, Chebyshev's inequality applied to $S_n$ gives the statement.
  \end{proof}

  If we have additional information about the random variables, then
  we can derive sharper bounds. In case of uniformly bounded random
  variables (rather than just bounded variance), Hoeffding's
  inequality, which we recall next, shows an exponential rate of
  concentration around the mean.
  
  \begin{theorem}[Hoeffding's inequality]\label{thm:hoeffdings}
    Let %
    $a$, $b \in \R$. Let $X_1,\dots,X_n$ be $n\in\N$
    independent real-valued random variables such that
    $a \leq X_i \leq b$ almost surely for all $i = 1, \dots, n$, and
    let $\mu$ be as in \eqref{eq:bbEmu}.
		Then, for every $\eps >0$
		\begin{align*}
                  \mathbb{P}\left[ \left| \frac{1}{n}\sum_{j=1}^n X_j - \mu \right| > \eps \right] \leq 2 e^{-\frac{2n \eps^2}{(b-a)^2}}.
		\end{align*}
	\end{theorem}
A proof can, for example, be found in \cite[Section B.4]{understanding}, where this version is also taken from.

Finally, we recall the central limit theorem, in its multivariate
formulation.
We say that $(X_j)_{j\in\N}$ is an {\bf i.i.d.\ sequence
  of random variables}, if the random variables are (pairwise)
independent and identically distributed. For a proof see \cite[Theorem
15.58]{klenke}.

  \begin{theorem}[Multivariate central limit theorem]\label{thm:clt}
    Let $(\BX_n)_{n\in\N}$ be an i.i.d.\ sequence of $\R^d$-valued
    random variables, such that $\bbE[\BX_n]=\Bnul\in\R^d$ and
    $\bbE[X_{n,i}X_{n,j}]=C_{ij}$ for all $i$, $j=1,\dots,d$.  Let
    \begin{equation*}
      \BY_n\dfn \frac{\BX_1+\dots+\BX_n}{\sqrt{n}}\in\R^d.
    \end{equation*}
      Then $\BY_n$ converges in distribution to
      $\gauss(\Bnul,\BC)$ as $n\to\infty$.
  \end{theorem}

  \chapter{Linear algebra and functional analysis}
This %
appendix provides some basic notions and results in linear algebra and functional analysis required in the main text. It is intended as a revision for a reader already familiar with these concepts. For more details and proofs, we refer for example to the standard textbooks \cite{doi:10.1137/1.9781611971484,rudin-rca, rudin-fa, conway2019course, grochenig2013foundations}.

\section{Singular value decomposition and pseudoinverse}\label{app:pseudo}
Let $\BA\in\R^{m\times n}$, $m$, $n\in\N$. Then the square root of the
positive eigenvalues of $\BA^\top\BA$ (or equivalently of
$\BA\BA^\top$) are referred to as the {\bf singular values} of $\BA$.
We denote them in the following by $s_1\ge s_2\dots\ge s_r>0$, where
$r\dfn {\rm rank}(\BA)$, so that $r\le \min\{m,n\}$. Every matrix
allows for a {\bf singular value decomposition (SVD)} as stated in the
next theorem, e.g.\ \cite[Theorem
1.2.1]{doi:10.1137/1.9781611971484}. Recall that a matrix
$\BV\in\R^{n\times n}$ is called {\bf orthogonal}, if $\BV^\top\BV$ is
the identity.

\begin{theorem}[Singular value decomposition]\label{thm:svd}
  Let $\BA\in\R^{m\times n}$. Then there exist orthogonal matrices
  $\BU\in\R^{m\times m}$, $\BV\in\R^{n\times n}$ such that with
  \begin{equation*}
    \BSigma \dfn \begin{pmatrix}
                s_1 &&&\\
                &\ddots&&&\Bnul\\
                    &&& s_r&\\
                    &\Bnul&&&\Bnul\\
              \end{pmatrix}\in\R^{m\times n}
  \end{equation*}
  it holds that $\BA=\BU\BSigma\BV^\top$, where $\Bnul$ stands for a zero
  block of suitable size. 
\end{theorem}
Given $\By\in\R^m$, consider the linear system
\begin{equation}\label{eq:lsappendix}
  \BA\Bw=\By.
\end{equation}
If $\BA$ is not a regular square matrix, then in general
there need not be a unique solution $\Bw\in\R^n$ to \eqref{eq:lsappendix}.  However, there
exists a unique {\bf minimal norm solution}
\begin{equation}\label{eq:minnormappendix}
  \Bw_*=\argmin_{\Bw\in M}\norm{\Bw},\qquad M = \set{\Bw\in\R^m}{\norm{\BA\Bw-\By}\le \norm{\BA\Bv-\By}~\forall\Bv\in\R^n}.
\end{equation}
The minimal norm solution can be expressed via the {\bf Moore-Penrose
  pseudoinverse} $\BA^\dagger\in\R^{n\times m}$ of $\BA$;
given an (arbitrary) SVD $\BA=\BU\Sigma\BV^\top$, it is defined as
\begin{equation}\label{eq:pseudoinverse}
 \BA^\dagger\dfn \BV\BSigma^\dagger \BU^\top\qquad\text{where}\qquad \BSigma^\dagger\dfn \begin{pmatrix}
                s_1^{-1} &&&\\
                &\ddots&&&\Bnul\\
                    &&& s_r^{-1}&\\
                    &\Bnul&&&\Bnul\\
              \end{pmatrix}\in\R^{n\times m}.
\end{equation}
The following theorem makes this precise, e.g., \cite[Theorem
1.2.10]{doi:10.1137/1.9781611971484}.
\begin{theorem}\label{thm:pseudoinverse}
  Let $\BA\in\R^{m\times n}$. Then there exists a unique minimum norm
  solution $\Bw_*\in\R^n$ in \eqref{eq:minnormappendix} and it holds $\Bw_*=\BA^\dagger\By$.
\end{theorem}
\begin{proof}
  Denote by $\BSigma_r\in\R^{r\times r}$ the upper left quadrant of $\BSigma$.
  Since $\BU\in\R^{m\times m}$ is orthogonal,
  \begin{equation*}
    \norm{\BA\Bw-\By} = \normc{\begin{pmatrix}
                                \BSigma_r&\Bnul\\
                                \Bnul&\Bnul
                              \end{pmatrix}
                              \BV^\top\Bw-\BU^\top\By
                                }.
\end{equation*}
We can thus write $M$ in \eqref{eq:minnormappendix} as
  \begin{align*}
    M &= \setc{\Bw\in\R^n}{\big(\begin{pmatrix}
                            \BSigma_r \,\Bnul
                          \end{pmatrix}\BV^\top\Bw\big)_{i=1}^r = (\BU^\top\By)_{i=1}^r
    }\\
      &= \setc{\Bw\in\R^n}{(\BV^\top\Bw)_{i=1}^r = \BSigma_r^{-1}(\BU^\top\By)_{i=1}^r}\\
    &= \setc{\BV\Bz}{\Bz\in\R^n,~(\Bz)_{i=1}^r = \BSigma_r^{-1}(\BU^\top\By)_{i=1}^r}
  \end{align*}
  where $(\Ba)_{i=1}^r$ denotes the first $r$ entries of a vector $\Ba$,
  and for the last equality we used
  orthogonality of $\BV\in\R^{n\times n}$. Since $\norm{\BV\Bz}=\norm{\Bz}$, the
  unique minimal norm solution is obtained by setting components $r+1,\dots,m$ of
  $\Bz$ to zero, which yields
  \begin{equation*}
    \Bw_* = \BV \begin{pmatrix}
          \BSigma_r^{-1}(\BU^\top\By)_{i=1}^r\\
          \Bnul
        \end{pmatrix}=\BV\BSigma^\dagger\BU^\top\By = \BA^\dagger\By
  \end{equation*}
  as claimed.
\end{proof}

\section{Vector spaces}

\begin{definition}
  Let $\bbK\in\{\R,\C\}$.
	A \textbf{vector space (over $\bbK$)} is a set $X$ such that the following %
        holds:
	\begin{enumerate}
        \item \textit{Properties of addition:} For every $x$, $y \in X$ there exists %
          $x+y \in X$ such that for all $z \in X$
		\begin{align*}
			x+y = y+x\quad\text{ and }\quad x + (y+z) = (x+y) + z. 
		\end{align*}
		Moreover, there exists a unique element $0 \in X$ such that $x + 0 = x$	for all $x \in X$ and for each $x\in X$ there exists a unique $-x \in X$ such that $x + (-x) = 0$.
		\item \textit{Properties of scalar multiplication:} 
		There exists a map $(\alpha, x) \mapsto \alpha x$ from $\bbK\times X$ to $X$ called scalar multiplication. %
                It satisfies $1x = x$ and $(\alpha \beta) x = \alpha(\beta x)$ for all $x\in X$.
	\end{enumerate}
	We call the elements of a vector space \textbf{vectors}.
    \end{definition}

    If the field is clear from context, we simply refer to $X$ as a vector space. We will primarily consider the case $\bbK=\R$, and in this case we also say that $X$ is a real vector space. 

To introduce a notion of convergence on a vector space $X$, it needs to be equipped with a topology, see Definition \ref{def:topology}. A \textbf{topological vector space} is a vector space which is also a topological space, and in which addition and scalar multiplication are continuous maps. We next discuss the most important instances of topological vector spaces.

\subsection{Metric spaces}

An important class of topological vector spaces consists of vector spaces that are also metric spaces.

\begin{definition}
For a set $X$, we call a map $d_X \colon X\times X \to [0,\infty)$ a \textbf{metric}, if 
\begin{enumerate}
\item $d_X(x, y) = 0$ if and only if $x = y$,
\item $d_X(x, y) = d(y, x)$ for all $x$, $y \in X$,
\item $d_X(x, z) \leq d_X(x, y) + d_X(y, z)$ for all  $x$, $y$, $z \in X$.
\end{enumerate}
We call $(X,d_X)$ a {\bf metric space}.
\end{definition}

In a metric space $(X,d_X)$, we denote the \textbf{open ball with center $x$ and radius $r>0$} by
\begin{equation}\label{eq:ball}
	B_r(x) \dfn \set{y \in X}{d_X(x, y) < r}.
\end{equation}
        Every metric space is naturally equipped with a topology: A set $A\subseteq X$ is open if and only if for every $x\in A$ exists $\eps>0$ such that $B_\eps(x)\subseteq A$. Therefore every metric vector space is a topological vector space.

\begin{definition} 
A metric space  $(X, d_X)$ is called \textbf{complete}, if every Cauchy sequence with respect to $d$ converges to an element in $X$.
\end{definition}

For complete metric spaces, an immensely powerful tool is Baire's category theorem. To state it, we require the notion of density of sets. Let $A$, $B \subseteq X$ for a topological space $X$. Then $A$ is \textbf{dense} in $B$ if the closure of $A$, denoted by $\overline{A}$, satisfies $\overline{A} \supseteq B$. 

\begin{theorem}[Baire's category theorem]\label{thm:BaireCat}
Let $X$ be a complete metric space. Then the intersection of every countable collection of dense open subsets of $X$ is dense in $X$. 
\end{theorem}
Theorem \ref{thm:BaireCat} implies that if $X = \bigcup_{i=1}^\infty V_i$ for a sequence of sets $V_i$, then at least one of the $V_i$ has to contain an open set. Indeed, assuming all $V_i$'s have empty interior implies that $V_i^c = X \setminus V_i$ is dense for all $i \in \N$. 
By De Morgan's laws, it then holds that $\emptyset = \bigcap_{i=1}^\infty V_i^c$ which contradicts Theorem \ref{thm:BaireCat}.

\subsection{Normed spaces}

A norm is a way of assigning a length to a vector. A normed space is a vector space with a norm. 

\begin{definition}\label{def:normAppendix}
  Let $X$ be a vector space over a field $\bbK\in\{\R,\C\}$.
  A map $\norm[X]{\cdot}:X\to [0,\infty)$
  is called a \textbf{norm} if the following %
  hold for all $x$, $y\in X$ and all $\alpha\in \bbK$:
  \begin{enumerate}
		\item \textbf{triangle inequality}: $\norm[X]{x+y} \leq \norm[X]{x} + \norm[X]{y}$, %
		\item \textbf{absolute homogeneity}:
                  $\norm[X]{\alpha x} = |\alpha| \norm[X]{x}$,
  \item \textbf{positive definiteness}:  %
    $\norm[X]{x}=0$ if and only if $x=0$.                  
\end{enumerate}  

We call $(X, \norm[X]{\cdot})$ a \textbf{normed space} and omit $\norm[X]{\cdot}$ from the notation if it is clear from the context. 
\end{definition}	
Every norm \emph{induces a metric $d_X$ and hence a topology} via $d_X(x,y) \dfn \|x-y\|_X$.
In particular, 
every normed vector space is a topological vector space with respect to %
this topology. 

\subsection{Banach spaces}\label{app:BanachSpaces}
\begin{definition}
  A normed vector space is called a {\bf Banach space} if and only if it is complete.
\end{definition}
Before presenting the main results on Banach spaces, we collect a couple of important examples.
\begin{itemize}
\item \textit{Euclidean spaces:} Let $d\in\N$. Then $(\R^d,\norm[]{\cdot})$ is a Banach space.
\item \textit{Continuous functions:} Let $d \in \N$ and let
  $K \subseteq \R^d$ be compact. The set of continuous functions
  from $K$ to $\R$ is denoted by $C(K)$. For $\alpha$, $\beta \in \R$
  and $f$, $g \in C(K)$, we define addition and scalar multiplication
  by $(\alpha f + \beta g)(\Bx) = \alpha f(\Bx) + \beta g(\Bx)$ for all
  $\Bx \in K$. The vector space $C(K)$ equipped with the {\bf supremum norm}
\begin{align*}
\norm[\infty]{f} \dfn \sup_{\Bx \in K}|f(\Bx)|,
\end{align*}
is a Banach space. 
\item \textit{Lebesgue spaces:} Let $(\Omega, \mathfrak{A}, \mu)$ be a measure space and let $1 \leq p < \infty$. Then the \textbf{Lebesgue space $L^p(\Omega, \mu)$} is defined as %
  the vector space of all equivalence classes of measurable functions $f:\Omega\to\R$ that coincide $\mu$-almost everywhere and satisfy 
\begin{equation}\label{eq:Lpnorm}
	\|f\|_{L^p(\Omega, \mu)} \dfn \left(\int_{\Omega} |f(x)|^p d\mu(x)\right)^{1/p} < \infty.
\end{equation}
The integral is independent of the choice of
representative of the equivalence class of $f$. Addition and scalar multiplication are defined pointwise as for $C(K)$. It then holds that $L^p(\Omega, \mu)$ is a Banach space. 
If $\Omega$ is a measurable subset of $\R^d$ for $d \in \N$, and $\mu$ is the Lebesgue measure,
we typically omit %
$\mu$ from the notation and %
simply write $L^p(\Omega)$. If $\Omega = \N$ and the measure is the counting measure, %
we denote these spaces by $\ell^p(\N)$ or simply $\ell^p$.

The definition can be extended to complex or $\R^d$-valued functions. In the latter case the integrand in \eqref{eq:Lpnorm} is replaced by $\norm{f(x)}^p$. We denote these spaces again by $L^p(\Omega,\mu)$ with the precise meaning being clear from context.
\item \textit{Essentially bounded functions:}
  Let $(\Omega, \mathfrak{A}, \mu)$ be a measure space.
  The $L^p$ spaces can be extended to $p=\infty$ by defining the $L^\infty$-norm
\begin{align*}
\norm[{L^\infty(\Omega, \mu)}]{f} \dfn \inf\set{C \geq 0}{ \mu(\{|f| > C\}) = 0)}.
\end{align*}
This is indeed a norm on the space of equivalence classes of measurable functions from $\Omega\to\R$ that coincide $\mu$-almost everywhere. Moreover, with this norm, $L^\infty(\Omega, \mu)$ is a Banach space. %
If $\Omega = \N$ and $\mu$ is the counting measure, we denote the resulting space by $\ell^\infty(\N)$ or simply $\ell^\infty$. As in the case $p<\infty$, it is straightforward to extend the definition to complex or $\R^d$-valued functions, for which the same notation will be used.
\end{itemize}

We continue by introducing the concept of dual spaces.
\begin{definition}\label{def:DualSpace}
  Let $(X, \norm[X]{\cdot})$ be a normed vector space over $\bbK\in\{\R,\C\}$.
  Linear maps from $X\to\bbK$ are called
  \textbf{linear functionals}. 
	The vector space of all continuous linear functionals on $X$ is called the \textbf{(topological) dual space of $X$} and is denoted by $X'$.
      \end{definition}

        Together with the natural
        addition and scalar multiplication %
        (for all $h$, $g\in X'$, $\alpha\in\bbK$ and $x\in X$)
	$$
	(h+g)(x) \coloneqq h(x) + g(x)\quad\text{and}\quad (\alpha h) (x) \coloneqq \alpha (h(x)),%
	$$
        $X'$ is a vector space. %
        We equip $X'$ with the norm
        \begin{equation*}
          \norm[X']{f}\dfn \sup_{\substack{x\in X\\ \norm[X]{x}=1}}|f(x)|.
        \end{equation*}
        The space $(X',\norm[X']{\cdot})$ is always a Banach space, even if $(X,\norm[X]{\cdot})$ is not complete \cite[Theorem 4.1]{rudin-fa}.

The dual space can often be used to characterize the original Banach space. One way in which the dual space $X'$ captures certain algebraic and geometric properties of the Banach space $X$ is through the Hahn-Banach theorem. In this book, we use one specific variant of this theorem and its implication for the existence of dual bases, see for instance \cite[Theorem 3.5]{rudin-fa}.

\begin{theorem}[Geometric Hahn-Banach, subspace version] \label{thm:banachSeparation} %
Let $M$ %
be a subspace of a Banach space $X$ and let
$x_0 \in X$. If $x_0$ is not in the closure of $M$, then there exists $f \in X'$ such that
$f(x_0) = 1$ %
and $f(x) = 0$ for every $x \in M$.
\end{theorem}
An immediate consequence of Theorem \ref{thm:banachSeparation} that will be used throughout this book is the existence of a \textbf{dual basis}.
Let $X$ be a Banach space and let $(x_i)_{i\in\N} \subseteq X$ be such that for all $i \in \N$
\begin{align*}
	x_i \not \in \overline{\mathrm{span}\set{x_j}{j\in\N,~j \neq i}}.
\end{align*}
Then, for every $i \in \N$, there exists $f_i \in X'$ such that $f_i(x_j) = 0$ if $i \neq j$ and $f_i(x_i) = 1$.

\subsection{Hilbert spaces}\label{app:hilbert}

Often, we require more structure than that provided by normed spaces. An inner product offers additional tools to compare vectors by introducing notions of angle and orthogonality. For simplicity we restrict ourselves to real vector spaces in the following.

\begin{definition}\label{def:innerProduct}
  Let $X$ be a real vector space. %
  A map
  $\langle \cdot , \cdot \rangle_X:X \times X\to \R$
  is called an \textbf{inner product} on $X$ if the following hold
  for all $x$, $y$, $z \in X$ and all $\alpha$, $\beta\in \R$: %
\begin{enumerate}
\item {\bf linearity}: $\langle \alpha x + \beta y, z\rangle_X = \alpha \langle x, z\rangle_X + \beta \langle y, z \rangle_X$,
\item {\bf symmetry}: $\langle x, y\rangle_X  = \langle y, x\rangle_X$,
\item {\bf positive definiteness}: $\langle x, x\rangle_X  > 0$ for all $x \neq 0$.  
        \end{enumerate}
        
\end{definition}

\begin{example}\label{ex:hilbert}
  For $p=2$, the Lebesgue spaces $L^2(\Omega)$
  and $\ell^2(\N)$ are Hilbert spaces with inner products
  \begin{align*}
    \dup[L^2(\Omega)]{f}{g}=\int_\Omega f(x)g(x)\dd x\qquad\text{for all }f,~g\in L^2(\Omega),
  \end{align*}
  and
  \begin{align*}
    \dup[\ell^2(\N)]{\Bx}{\By}=\sum_{j\in\N}x_jy_j\qquad\text{for all }\Bx=(x_j)_{j\in\N},~\By=(y_j)_{j\in\N}\in\ell^2(\N).
  \end{align*}
\end{example}

On inner product spaces the so-called Cauchy-Schwarz inequality holds.
\begin{theorem}[Cauchy-Schwarz inequality]
  Let $X$ be a vector space with inner product $\langle \cdot , \cdot \rangle_X$.
  Then it holds for all $x$, $y \in X$
\begin{align*}
  |\langle x, y \rangle_X| \leq \sqrt{\dup[X]{x}{x}\dup[X]{y}{y}
  }.
\end{align*}
Moreover, equality %
holds if and only if $x$ and $y$ are linearly dependent.
\end{theorem}
\begin{proof}
  Let $x$, $y\in X$. If $y=0$ then $\dup[X]{x}{y}=0$ and thus the statement is trivial.
  Assume in the following $y\neq 0$, so that $\dup[X]{y}{y}>0$.
  Using the linearity and symmetry properties it holds for all $\alpha\in\R$
  \begin{equation*}
    0\le \dup[X]{x-\alpha y}{x-\alpha y} = \dup[X]{x}{x}-2\alpha\dup[X]{x}{y} +\alpha^2\dup[X]{y}{y}.
  \end{equation*}
  Letting $\alpha\dfn \dup[X]{x}{y}/\dup[X]{y}{y}$ we get
  \begin{equation*}
    0\le \dup[X]{x}{x}-2\frac{\dup[X]{x}{y}^2}{\dup[X]{y}{y}}+\frac{\dup[X]{x}{y}^2}{\dup[X]{y}{y}}=\dup[X]{x}{x}-\frac{\dup[X]{x}{y}^2}{\dup[X]{y}{y}}.
  \end{equation*}
  Rearranging terms gives the claim.
\end{proof}

Every inner product $\langle \cdot , \cdot \rangle_X$ induces a norm via
\begin{equation}\label{eq:Hilbertnorm}
  \norm[X]{x} \dfn \sqrt{\langle x, x \rangle}\qquad\text{for all }x \in X.
\end{equation}
The properties of the inner product immediately yield the \textbf{polar identity}
\begin{align}\label{eq:polarIdentity}
\|x + y\|_X^2 = \|x\|_X^2 + 2 \langle x, y\rangle_X + \|y\|_X^2.
\end{align}
The fact that \eqref{eq:Hilbertnorm} indeed defines a norm follows by
an application of the Cauchy-Schwarz inequality %
to \eqref{eq:polarIdentity}, which yields that $\norm[X]{\cdot}$ satisfies the triangle inequality.
This gives rise to the definition of a Hilbert space.

\begin{definition}
  Let $H$ be a real vector space with inner product $\dup[H]{\cdot}{\cdot}$.
  Then $(H,\dup[H]{\cdot}{\cdot})$ is called a {\bf Hilbert space} if and only if $H$ is complete with respect to the norm $\norm[H]{\cdot}$ induced by the inner product.
\end{definition}

A standard example of a Hilbert space is $L^2$: Let $(\Omega, \mathfrak{A}, \mu)$ be a measure space. Then %
\begin{align*}
	\langle f, g \rangle_{L^2(\Omega, \mu)} = \int_{\Omega} f(x)g(x) \dd\mu(x)\qquad \text{ for all }f,g \in L^2(\Omega, \mu),
\end{align*}
defines an inner product on $L^2(\Omega,\mu)$ compatible with the $L^2(\Omega, \mu)$-norm.

In a Hilbert space, we can compare vectors not only via their distance, measured by the norm, but also by using the inner product, which %
corresponds to their relative orientation. This %
leads to the concept of orthogonality. 
\begin{definition}\label{def:orthogonal}
Let $(H,\dup[H]{\cdot}{\cdot})$ be a Hilbert space and let $f$, $g \in H$. We say that $f$ and $g$ are \textbf{orthogonal} if $\dup[H]{f}{g} = 0$, denoted by %
$f \perp g$. For $F$, $G \subseteq H$ we write $F \perp G$ if $f\perp g$ for all $f \in F$, $g \in G$. Finally, for $F\subseteq H$, the set $F^\perp=\set{g\in H}{g\perp f~\forall f\in F}$ is called the {\bf orthogonal complement} of $F$ in $H$.
\end{definition}

For orthogonal vectors, the polar identity immediately implies the Pythagorean theorem. 

\begin{theorem}[Pythagorean theorem]
Let $(H,\dup[H]{\cdot}{\cdot})$ be a Hilbert space, $n\in \N$, and let $f_1, \dots, f_n \in H$ be pairwise orthogonal vectors. 
Then, 
\begin{align*}
	\left \|\sum_{i=1}^n f_i \right \|_H^2 = \sum_{i=1}^n \left \| f_i \right \|_H^2.
\end{align*}
\end{theorem}

A final property of Hilbert spaces that we encounter in this book is the existence of unique \textbf{projections} onto convex sets. For a proof, see for instance \cite[Thm.~4.10]{rudin-rca}.

\begin{theorem}\label{thm:uniqueProjection}
	Let $(H,\dup[H]{\cdot}{\cdot})$ be a Hilbert space and let $K\neq\emptyset$ be a closed convex subset
	of $H$. Then for all $h\in H$ exists a unique %
        $k_0 \in K$ such that
	\begin{align*}
          \norm[H]{h - k_0} = \inf\set{\norm[H]{h - k}}{k \in K}.	
	\end{align*}	
\end{theorem}